\documentclass{article}
\usepackage{iclr2027_conference,times}

\usepackage[T1]{fontenc}
\usepackage{booktabs}
\usepackage{amsfonts}
\usepackage{nicefrac}
\usepackage{microtype}
\usepackage{amsthm}
\usepackage{longtable}
\usepackage{needspace}
\usepackage{graphicx}
\usepackage{xcolor}
\graphicspath{{./}}
\usepackage{amsmath}
\usepackage{amssymb}
\usepackage{siunitx}
\usepackage{afterpage}
\usepackage{float}
\usepackage{multirow}

\usepackage{titletoc}
\titlecontents{lsection}[1.8em]{\addvspace{3pt}\bfseries}{\contentslabel{1.8em}}{\hspace*{-1.8em}}{\hfill\contentspage}
\titlecontents{lsubsection}[4.3em]{}{\contentslabel{2.5em}}{\hspace*{-2.5em}}{\titlerule*[0.5pc]{.}\contentspage}
\usepackage{hyperref}
\usepackage{url}

\makeatletter
\def\input@path{{./}}
\makeatother

\newtheorem{proposition}{Proposition}
\newtheorem{theorem}{Theorem}
\newcommand{\R}{\mathbb{R}}
\newcommand{\WA}{\mathrm{WA}}

\newcommand{\ConservativeMode}{\mathrm{C}}
\newcommand{\DirectHeadMode}{\mathrm{V}}
\newcommand{\introaddition}[1]{#1}
\newcommand{\explanationaddition}[1]{#1}
\newcommand{\explanationcolor}{}
\newcommand{\secondreviewercolor}{}

\title{WallpaperNet: A General Blueprint for Wallpaper-Group-Equivariant Graph Neural Networks}

\author{%
\makebox[\dimexpr\textwidth-2\tabcolsep\relax][c]{%
\begin{tabular}{@{}c@{\hspace{4em}}c@{}}
Rostislav Fedorov\thanks{Corresponding author:
\texttt{rostislav.fedorov@helmholtz-berlin.de}.}
& Lilliana Fl\'orido Martins \\
\normalfont Helmholtz-Zentrum Berlin f\"ur
& \normalfont Ruprecht-Karls-Universit\"at \\
\normalfont Materialien und Energie
& \normalfont Heidelberg \\[1ex]
Kevin Maik Jablonka
& Ganna Gryn'ova \\
\normalfont Friedrich-Schiller-Universit\"at Jena
& \normalfont University of Birmingham
\end{tabular}%
}%
}

\iclrfinalcopy
\begin{document}

\maketitle
\lhead{}

\begin{abstract}
Equivariant graph neural networks are typically designed for continuous Euclidean symmetry groups such as $E(3)$ or $SE(3)$, where arbitrary global translations are symmetries and can be removed using relative coordinates. In many settings, however, a graph is defined relative to an underlying periodic structure, so its symmetry is no longer the full Euclidean group. In two dimensions, these symmetries are the 17 wallpaper groups. This structure creates an opportunity: rather than explicitly representing the entire periodic system as a large graph, one can represent only the smaller graph of interest while incorporating the symmetry of the surrounding periodic structure into the model. We present a general blueprint for wallpaper-group-equivariant message passing by separating the symmetry problem into two components: point-group equivariance and lattice-relative translational symmetry. We demonstrate how it can be applied to a molecular adsorption problem on a periodic graphene substrate, providing high accuracy at a fraction of the cost of full-system message passing.
\end{abstract}

\section{Introduction}

\providecommand{\introaddition}[1]{#1}
\providecommand{\explanationaddition}[1]{#1}
\providecommand{\explanationcolor}{}

\begin{figure}[h]
    \centering
    \includegraphics[width=\textwidth]{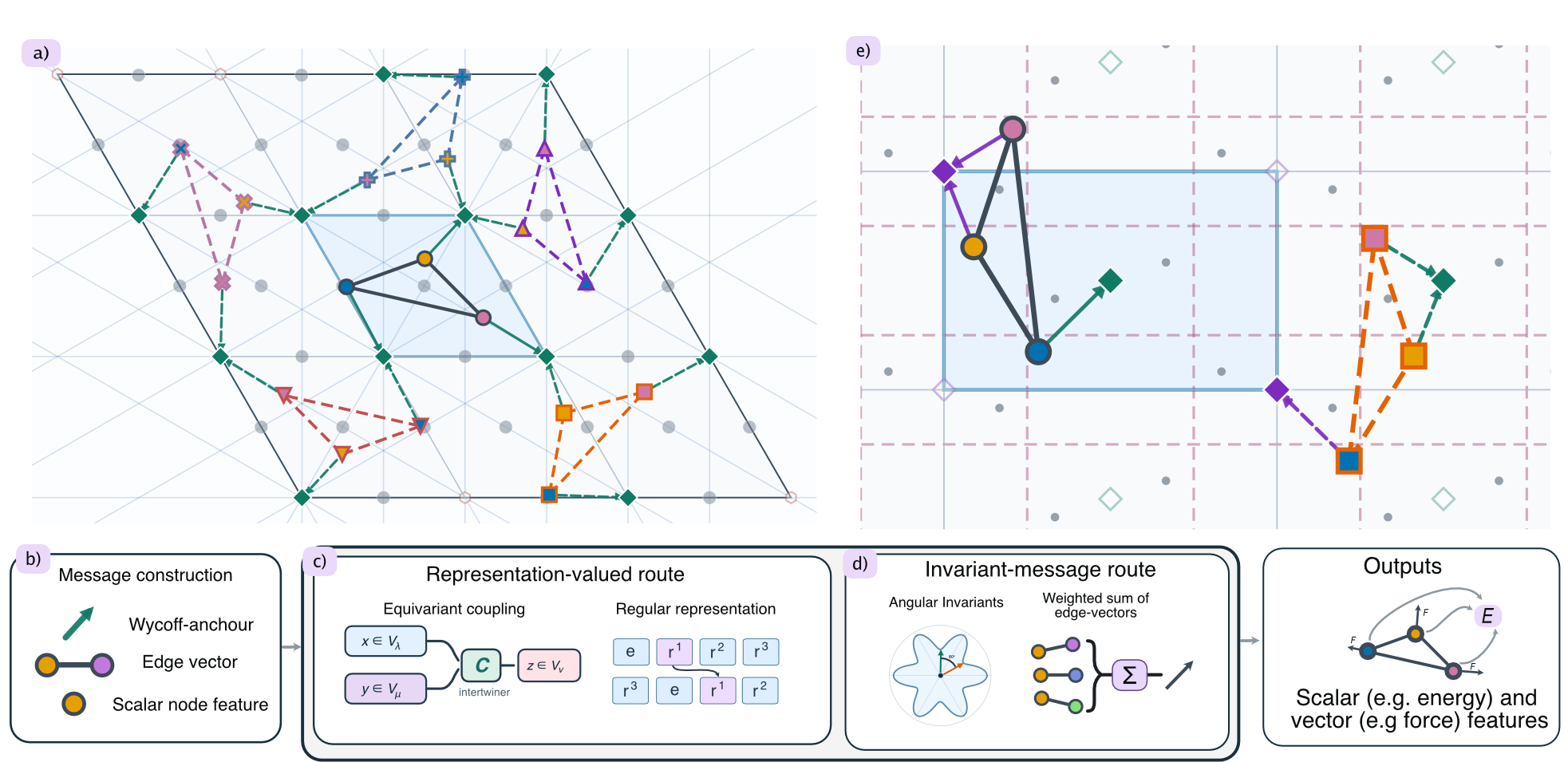}
    {\explanationcolor
    \caption{\textbf{Wallpaper-equivariant message passing on periodic substrates.} (a) Molecular graph on a graphene substrate (gray dots) with wallpaper-group symmetry \(p6m\) (graphene). Colored copies illustrate symmetry-related configurations under lattice translations, rotations, and reflections. Green vectors illustrate Wyckoff-anchors (WA) for each graph node. (b) Message construction combines scalar node features, relative edge vectors, and WA vectors. (c) The representation-valued route propagates features through equivariant couplings (intertwiners). Regular-representation features transform by permutations of their group-indexed channels, illustrated schematically by a cyclic shift. For simplicity, the \(C_4\) regular representation is shown. Rotating a vector by \(90^\circ\) cyclically permutes the four regular channels. (d) The invariant-message route instead constructs point-group-invariant angular features, illustrated by their sixfold rotational periodicity, and reconstructs covariant vectors as weighted sums of edge vectors with invariant scalar coefficients. Both approaches produce invariant scalar predictions (e.g., energies) and equivariant vector predictions (e.g., forces). (e) Extension to the nonsymmorphic wallpaper group \(pgg\), where glide reflections relate graph configurations and exchange distinct translation-suborbit members of the WA orbit (green and purple). In this case, WA vectors are supplemented by permutation-valued member information to preserve the full lattice registry.}
    }
    \label{fig:wallpaper-graph-overview}
\end{figure}

\introaddition{Molecular adsorption on periodic crystalline surfaces plays an important role
in heterogeneous catalysis, chemical sensing, and molecular
electronics~\citep{medfordSabatierPrinciplePredictive2015,kongMolecularAdsorptionGraphene2014}.
Understanding these interfaces requires predicting how the interaction energy
and atomic forces depend on the adsorbate's molecular conformation,
orientation, and position relative to the underlying surface. First-principles
studies of molecular adsorption on a periodic surface have demonstrated that
different adsorption sites and molecular orientations can lead to different
binding energies and electronic interactions. Traditionally, these properties
have been investigated using atomistic electronic-structure calculations. The
computational scale of surface-adsorbate modeling is illustrated by large-scale
initiatives such as the Open Catalyst
Project~\citep{chanussotOpenCatalyst20202021}. In a common simulation scenario,
the periodic surface is fixed and has one of the 17 wallpaper-group symmetries,
while the adsorbate is
relaxed~\citep{hofmannFirstprinciplesCalculationsHybrid2021,dipasqualeSolidLiquidInterfacial2025,zhouApplicabilityThermodynamicMechanical2025}. This motivates a general framework for learning on graphs interacting with a
periodic environment whose periodicity and symmetry are treated implicitly.
Although application of this framework can go beyond material science, atomistic simulation is a major direct application of this design pattern. Therefore,
we emphasize molecular adsorption on periodic substrates as a concrete
application of our framework. In this setting, the molecular
graph typically contains substantially fewer nodes than an explicit graph of
the periodic substrate, motivating a more computationally efficient
representation.}

Message-passing neural networks provide a standard language for learning on
molecular and atomistic graphs~\citep{gilmer2017mpnn}. Modern geometric graph
networks exploit Euclidean symmetry to improve data efficiency and guarantee
physically meaningful transformation laws. A major family of methods uses
invariant scalar messages and reconstructs covariant coordinates or vectors
from invariant scalar coefficients, as in $E(n)$-equivariant graph neural
networks (EGNNs)~\citep{satorras2021egnn}. A second family propagates
representation-valued features, including vectors and higher-order tensors, as
in Tensor Field Networks~\citep{thomas2018tfn}, SE(3)-Transformers~\citep{fuchs2020se3}, SEGNNs~\citep{brandstetter2022segnn}, e3nn~\citep{geiger2022e3nn}, and equivariant interatomic potentials such as
NequIP~\citep{batznerE3equivariantGraphNeural2022}.

These methods commonly remove global translations through relative coordinates,
which is appropriate when jointly translating the modeled system leaves the
target unchanged. For example, relative edge vectors $r_{ij}=x_i-x_j$ are
exactly invariant under a common translation. For a graph interacting with a periodic
substrate, however, an arbitrary in-plane translation is generally \emph{not} a
symmetry: translating the graph relative to the lattice changes its adsorption
registry or local environment. The relevant transformations are the affine
isometries of a crystallographic plane group, i.e., one of the 17 wallpaper
groups tabulated in the International Tables for Crystallography~\citep{internationaltables2016}.

This distinction creates two separate symmetry problems. First, rotations and
reflections must act correctly on graph features. Second, the model must retain
the graph's position relative to the periodic lattice modulo the discrete translation lattice. Relative edges provide covariant geometric inputs for the first
problem, but do not encode the relative position inside the cell required for the second. In surface science, the placement of a molecule relative to the substrate lattice is often described as its \emph{registry}, and we adopt this terminology for an arbitrary graph position with respect to the periodic lattice.

Wallpaper symmetry naturally separates into a finite point-group action and a periodic lattice registry. We handle the former with invariant or representation-valued message passing and the latter with Wyckoff-anchor features.
Figure~\ref{fig:wallpaper-graph-overview} summarizes the physical setting,
geometric inputs, architectural routes, and nonsymmorphic extension used
throughout this work.

Our main contributions are:
\begin{enumerate}
    \item We introduce a general framework for wallpaper-group-equivariant
    message passing on compact graphs coupled to periodic two-dimensional
    environments. The framework separates finite point-group equivariance
    from lattice registry, encoded using Wyckoff-anchor geometry and
    translation-suborbit member information for nonsymmorphic groups.

    \item We develop \texttt{we3nn}, a finite-group equivariant neural-network
    library for cyclic and dihedral point groups, supporting equivariant
    linear and bilinear maps, regular representations, and
    representation-valued message passing.

    \item We benchmark invariant and representation-valued realizations of
    the framework on synthetic scalar and vector prediction tasks spanning
    all 17 wallpaper groups, including a comparison with full-cell baselines.

    \item We apply the framework to a wallpaper-equivariant interatomic
    potential for molecular adsorption on a fixed \(p6m\) graphene substrate,
    demonstrating its use in a realistic periodic atomistic setting.
\end{enumerate}

\paragraph{Related work.}
\label{sec:related-work}

Our architectures build on equivariant geometric
GNNs~\citep{satorras2021egnn,thomas2018tfn,geiger2022e3nn,PaiNN,batznerE3equivariantGraphNeural2022}
and finite-group equivariant
constructions~\citep{cohen2016group,finziPracticalMethodConstructing2021,cesa2021ENsteerable}.
Unlike crystal graph models~\citep{xie2018cgcnn,kaba2022crystal,ruff2024connectivity},
we treat the substrate as a fixed periodic background and supply its registry
to an adsorbate-only graph. An extended discussion is given in
Appendix~\ref{sec:extended-related-work}.

\section{Problem formulation and crystallographic symmetry}
\label{sec:problem}

\subsection{A graph coupled to a periodic background}

Let $\mathcal G=(\mathcal V,\mathcal E)$ be a geometric graph with node
features $h_i$ and positions $x_i\in\R^3$ with in-plane parts
$x_{i,xy}\in\R^2$. The graph is coupled to a periodic background in the \(xy\) plane, which is not represented as an explicit large graph. Instead, the model acts on the small graph
$\mathcal G$ together with symmetry-aware geometric features encoding its
registry relative to the periodic environment.

Let $G$ denote the wallpaper group of the periodic background. An operation
$g\in G$ acts on the in-plane coordinates as
\begin{equation}
    gx = Q_gx+\tau_g,
    \qquad
    Q_g\in O(2),
    \qquad
    \tau_g\in\R^2.
    \label{eq:affine-wallpaper-action}
\end{equation}
The linear part $Q_g$ is a rotation or reflection. The translation $\tau_g$ may
also contain the fractional shift of a glide reflection.

We write $X=\{(h_i,x_i)\}_{i=1}^{N}$ for the graph with $N$ nodes embedded in
three-dimensional space. A wallpaper operation $g$ acts on the in-plane
coordinates while leaving the out-of-plane coordinate unchanged,
\begin{equation}
x_i
\longmapsto
\widetilde Q_g x_i+\widetilde \tau_g,
\qquad
\widetilde Q_g=
    \begin{pmatrix}
        Q_g & 0\\
        0 & 1
    \end{pmatrix},
\qquad
\widetilde \tau_g=
    \begin{pmatrix}
        \tau_g\\
        0
    \end{pmatrix}.
\end{equation}
We denote the transformed graph by $gX$.

For a symmetry-compatible graph construction, $g$ induces a permutation
$\pi_g$ of the graph nodes and edges. For invariant scalar predictions such
as an energy $E$, and covariant vector predictions such as forces
$F_i\in\mathbb{R}^3$, the target transformation laws are
\begin{equation}
    E(gX)=E(X),
    \label{eq:target-invariance}
\end{equation}
and
\begin{equation}
F_{\pi_g(i)}(gX)
=
\widetilde Q_g F_i(X).
\label{eq:target-equivariance}
\end{equation}

Panel (a) of Figure~\ref{fig:wallpaper-graph-overview} illustrates the problem setting. The graph $\mathcal G$ represents a small molecule on a periodic substrate, and symmetry-related configurations $gX$ are shown with dashed edges.

\subsection{Translation subgroup and point group}
\label{sec:problem-formulation}
For a planar translation vector \(t\in\mathbb R^2\), let
\(T_t(x)=x+t\). The primitive Bravais lattice
(Equation~\eqref{bravais_lattice}) consists of the translations that
preserve the periodic background:
\begin{equation}
    \Lambda
    =
    \{t\in\mathbb R^2:T_t\in G\}.
\end{equation}
Thus, \(\Lambda\) describes the exact translational repetitions of the periodic background. Primitive cells of \(\Lambda\) are outlined by blue parallelograms in panels (a) and (e) of Figure~\ref{fig:wallpaper-graph-overview}.

The linear-part map
\begin{equation}
    \rho:G\rightarrow O(2),
    \qquad
    \rho(g)=Q_g,
\end{equation}
preserves group composition and maps precisely the pure translations
in \(G\) to the identity matrix. Its kernel is therefore the translation
subgroup, which is consequently normal in \(G\). The finite
crystallographic point group is
\begin{equation}
    P:=\rho(G)\cong G/\Lambda.
    \label{eq:quotient-point-group}
\end{equation}
Equivalently, wallpaper symmetry has the short exact sequence
\begin{equation}
    1
    \longrightarrow
    \Lambda
    \longrightarrow
    G
    \longrightarrow
    P
    \longrightarrow
    1,
    \label{eq:wallpaper-group-extension}
\end{equation}
where \(\Lambda\) is the normal subgroup of lattice translations and the
quotient map sends each wallpaper operation to its point-group class in
\(P\simeq G/\Lambda\). Two wallpaper operations belong to the same quotient class exactly
when they differ by a lattice translation. For a symmorphic wallpaper group this extension splits,
\begin{equation}
    G\simeq\Lambda\rtimes P,
\end{equation}
whereas for a nonsymmorphic wallpaper group it does not. This distinction
will determine whether the selected Wyckoff-anchor orbit requires an
additional translation-suborbit member encoding.

We obtain generators, Wyckoff
positions, and standard settings from the International Tables and the Bilbao
Crystallographic Server~\citep{aroyo2006bcs1}. The table of wallpaper groups is reproduced in Appendix Table~\ref{tab:selected-wa-orbits}. The distinction between symmorphic and nonsymmorphic crystallographic groups is
standard~\citep{internationaltables2016}. Panel (a) shows the symmorphic wallpaper group \(p6m\), with green diamonds marking sites in the selected Wyckoff-anchor orbit. Panel (e) shows the nonsymmorphic group \(pgg\), with green and purple diamonds distinguishing the two translation suborbits of the selected anchor orbit.

When the periodic background is represented explicitly, a joint Euclidean transformation of the graph and the cell preserves their relative geometry. For a common translation $x_i\mapsto x_i+t$, the relative edge vector satisfies $(x_i+t)-(x_j+t)=r_{ij}$. This is the usual mechanism by which Euclidean geometric GNNs remove arbitrary
translations~\citep{duvalHitchhikersGuideGeometric2023}. In the periodic problem, however, $\{r_{ij}\}_{(i,j)\in\mathcal{E}}$ is unchanged if every graph node is shifted by a non-lattice vector, even though
the graph has moved to a different position inside the periodic cell. Relative graph
geometry is therefore insufficient to uniquely represent the graph position in the periodic environment.

\section{Wallpaper-group-equivariant message passing}
\label{sec:wallpaper-mp}

\explanationaddition{Section~\ref{sec:problem} showed that relative edge vectors capture
orientation-dependent geometry but discard the graph's position within the
periodic cell. WallpaperNet therefore solves two distinct problems:
transforming features correctly under rotations and reflections, and retaining
lattice-relative registry. Section~\ref{sec:point-group-mp} develops two
approaches to the first problem, Sections~\ref{sec:wa}--\ref{sec:nonsymmorphic}
construct the registry encoding, and Section~\ref{sec:full-proof} shows that
combining these ingredients yields equivariance under the full wallpaper group.}

\subsection{Point-group equivariance}
\label{sec:point-group-mp}

\explanationaddition{This component handles rotations and reflections but does not retain position
within the periodic cell.}

We consider two realizations of point-group equivariance. The first maps
geometric inputs to invariant scalars and uses an ordinary multilayer
perceptron (MLP). The second retains representation-valued features and uses
equivariant linear and bilinear maps.

\subsubsection{Invariant-message scalarization}
\label{sec:invariant-mp}

The planar crystallographic point groups are cyclic or dihedral groups
$C_n$ and $D_n$. The construction below applies for arbitrary $n\in\mathbb N$.
For wallpaper groups, crystallographic restriction limits the occurring
rotation orders to $n\in\{1,2,3,4,6\}$.

For an in-plane vector $v=(v_x,v_y)$, define
\begin{equation}
    z(v)=v_x+\mathrm{i}v_y,
    \qquad
    \hat z(v)=\frac{z(v)}{\|v\|}
\end{equation}
for $v\neq0$. For $C_n$, a complete angular invariant for a single nonzero
vector can be represented by
\begin{equation}
    \psi_{C_n}(v)
    =
    \left(
        \operatorname{Re}[\hat z(v)^n],
        \operatorname{Im}[\hat z(v)^n],
        \|v\|
    \right).
    \label{eq:cn-invariant}
\end{equation}
For $D_n$, let $\alpha$ specify the orientation of one reflection axis. A
reflection-invariant encoding is
\begin{equation}
    \psi_{D_n}(v)
    =
    \left(
        \operatorname{Re}
        \left[e^{-\mathrm{i}n\alpha}\hat z(v)^n\right],
        \|v\|
    \right).
    \label{eq:dn-invariant}
\end{equation}
These features are invariant under the corresponding cyclic or dihedral
point-group action. We write $\psi_P$ for the encoding of the point group $P$
at hand. The derivation and the treatment of $v=0$ are given in
Appendix~\ref{sec:scalarization-details}.
\explanationaddition{Panel (d) of Figure~\ref{fig:wallpaper-graph-overview} visualizes a scalar angular function invariant under \(60^\circ\) rotations.}

For an oriented edge $j\to i$, let \(s_{ij}\in\mathbb R^{d_s}\) denote the
concatenation of the invariant message inputs: invariant node features,
distance and radial features, $\psi_P(r_{ij,xy})$, $\psi_P(v_{\WA,i})$ for the WA
vector $v_{\WA,i}$ of node $i$ (Section~\ref{sec:wa}), and optional joint invariants (Appendix~\ref{sec:scalarization-details}). With the
message MLP \(M:\mathbb R^{d_s}\to\mathbb R^{d_m}\) and the coefficient head
\(A:\mathbb R^{d_m}\to\mathbb R\), we obtain an invariant message
$m_{ij}=M(s_{ij})$ and an invariant scalar coefficient $w_{ij}=A(m_{ij})$.
As in EGNN-style models~\citep{satorras2021egnn}, the in-plane vector output is
reconstructed from the neighbors $\mathcal N(i)$ of node $i$ as
\begin{equation}
    F_{i,xy}
    =
    \sum_{j\in\mathcal N(i)}w_{ij}r_{ij,xy}.
\end{equation}
Since $w_{ij}$ is invariant and $r_{ij,xy}\mapsto Q_gr_{ij,xy}$,
\begin{equation}
    F_{\pi_g(i),xy}(gX)
    =Q_gF_{i,xy}(X).
\end{equation}
Thus an invariant-message network can produce a covariant vector output without
propagating nontrivial hidden representations.

\subsubsection{Representation-valued message passing}
\label{sec:regular-mp}

Alternatively, features may retain nontrivial point-group transformation laws.
Each feature space carries a real orthogonal representation of the point group
\(P\). Scalars, planar vectors, regular fields, and anchor-member features
carry their respective trivial, standard, regular, and permutation
representations.

\explanationaddition{For real representation spaces \((U,\rho_U)\) and \((V,\rho_V)\) of the finite
point group \(P\), a map \(f:U\to V\) is \(P\)-equivariant if
}
\begin{equation}
    \explanationaddition{f\!\left(\rho_U(p)x\right)
    =
    \rho_V(p)f(x)
    \qquad
    \text{for all }p\in P\text{ and }x\in U.}
    \label{eq:equivariant-map-definition}
\end{equation}
\explanationaddition{For linear maps, this condition defines the intertwiner space}
\begin{equation}
    \operatorname{Hom}_P(U,V)
    =\left\{W\in\operatorname{Hom}_{\mathbb R}(U,V):
      W\rho_U(p)=\rho_V(p)W\quad\forall p\in P\right\}.
    \label{eq:mp-hom}
\end{equation}
Here we use \(\rho_U,\rho_V\) to denote feature representations of
\(P\), distinct from the linear-part map \(\rho:G\to O(2)\) introduced in
Section~\ref{sec:problem-formulation}. Bilinear interactions of inputs
\(x\in U_1\) and \(y\in U_2\) take the form \(T(x,y)=C(x\otimes y)\) with \(C\in\operatorname{Hom}_P(U_1\otimes U_2,V)\).
Thus linear and bilinear interactions are parameterized by bases of the
appropriate intertwiner spaces. In \texttt{we3nn}, these bases are fixed and
their coefficients are learned (Appendices~\ref{sec:mp-details}
and~\ref{sec:we3nn}).

For nonlinear processing, representation-valued features are mapped to regular
fields, on which the group acts by permuting group-indexed coordinates. One
regular field is a function \(u:P\to\mathbb R\), with one scalar coordinate per
group element, and the regular representation
\(\rho_{\mathrm{reg}}(p)=\lambda(p)\) acts as
\begin{equation}
    [\lambda(p)u](k)=u(p^{-1}k),
    \label{eq:mp-regular-action}
\end{equation}
where the element \(p\) acts, while \(k\) indexes a group coordinate. A scalar
activation \(\sigma:\mathbb R\to\mathbb R\), applied componentwise, commutes
with this permutation action and therefore preserves equivariance. The result
is then mapped to the desired output type (conventions in
Appendix~\ref{sec:mp-details}). \explanationaddition{Panel (c) of Figure~\ref{fig:wallpaper-graph-overview} illustrates the regular representation of \(C_4\): applying a \(90^\circ\) rotation cyclically permutes its four group-indexed coordinates.}

\subsection{Wyckoff-anchor encoding of periodic registry}
\label{sec:wa}

\explanationaddition{The WA construction supplies the registry information that relative graph
edges discard.}

\paragraph{Selection of one anchor orbit.}

We select one crystallographic anchor orbit for each wallpaper
group. Let $A^0=\{q_1,\ldots,q_m\}$ contain the representatives of this orbit modulo primitive translations. Its
periodic lift is
\begin{equation}
    \mathcal A
    =A^0+\Lambda
    =\{q_\alpha+t:\alpha=1,\ldots,m,\ t\in\Lambda\}.
    \label{eq:periodic-anchor}
\end{equation}
The orbit is $G$-closed, $g\mathcal A=\mathcal A$ for all $g\in G$.

For the selected symmorphic cases, the orbit has one translation-suborbit
member per primitive cell, $m=1$, whereas the selected nonsymmorphic cases $pg$,
$pmg$, $pgg$, and $p4g$ require $m=2$ and are treated in
Section~\ref{sec:nonsymmorphic}. The complete choices are listed in
Appendix~\ref{sec:selected-anchor-orbits}, Table~\ref{tab:selected-wa-orbits}.

\paragraph{Nearest-anchor vector.}

For a point $x\in\R^2$, define
\begin{equation}
    a(x)
    \in
    \operatorname*{argmin}_{a\in\mathcal A}\|a-x\|.
\end{equation}
Whenever the minimizer is unique (see Appendix~\ref{voronoi_boundaries}), define
\begin{equation}
    v_{\WA}(x)=a(x)-x.
    \label{eq:wa-vector}
\end{equation}

\begin{proposition}[WA covariance]
If $\mathcal A$ is closed under $G$ and the nearest anchor is unique, then
\begin{equation}
    a(gx)=ga(x)
\end{equation}
and
\begin{equation}
    v_{\WA}(gx)=Q_gv_{\WA}(x).
    \label{eq:wa-covariance}
\end{equation}
\end{proposition}
The proof is in Appendix~\ref{sec:wa-covariance-proof}. For $m=1$, $\mathcal A=q_1+\Lambda$ and
\begin{equation}
    x\equiv q_1-v_{\WA}(x)\pmod{\Lambda}.
\end{equation}
Thus the WA vector retains complete position information modulo primitive translations, except at nearest-anchor ties. WA vectors of the \(p6m\) wallpaper group are visualized as green arrows in panel (a) of Figure~\ref{fig:wallpaper-graph-overview}.

\subsection{Nonsymmorphic translation-suborbit encoding}
\label{sec:nonsymmorphic}

\explanationaddition{The WA vector is sufficient for the selected symmorphic orbits, while
nonsymmorphic groups additionally require a discrete member label.} The vector covariance in Equation~\eqref{eq:wa-covariance} does not by itself
retain which representative of a multi-member anchor orbit was selected. This
matters in nonsymmorphic groups because different translation-suborbit members
are permuted by wallpaper operations. \explanationaddition{Panel (e) of Figure~\ref{fig:wallpaper-graph-overview} illustrates a selected \(pgg\) anchor orbit with two translation suborbits, shown by green and purple diamonds. Correspondingly colored arrows show the nearest-anchor vectors of symmetry-related nodes. A glide reflection exchanges the two suborbit labels and maps one vector into the other.}

\paragraph{Member permutation representation.}

Write $\mathcal A=\coprod_{\alpha=1}^{m}(q_\alpha+\Lambda)$.
Let \(\alpha(x)\) denote the translation-suborbit member containing the
nearest anchor, and for node \(i\) define
\begin{equation}
    v_i=v_{\WA}(x_{i,xy}),
    \qquad
    \ell_i=\operatorname{onehot}(\alpha(x_{i,xy}))\in\mathbb R^m.
\end{equation}

For \(g\in G\), let \(p_g:=g\Lambda\in G/\Lambda\simeq P\).
The induced member permutation representation
\(\rho_{\mathrm{mem}}:P\to O(m)\) is defined by
\begin{equation}
    \rho_{\mathrm{mem}}(p_g)e_\alpha=e_\beta
    \quad\Longleftrightarrow\quad
    gq_\alpha\equiv q_\beta\pmod{\Lambda},
    \label{eq:member-representation}
\end{equation}
where \(e_\alpha\) is the \(\alpha\)-th standard basis vector of
\(\mathbb R^m\). Its image consists of permutation matrices. Hence
(Appendix~\ref{sec:translation-suborbit-action})
\begin{equation}
    v_i\mapsto Q_gv_i,
    \qquad
    \ell_i\mapsto\rho_{\mathrm{mem}}(p_g)\ell_i.
\end{equation}

\explanationaddition{The three geometric ingredients now have distinct roles. Relative edge vectors
encode molecular geometry, \(v_i\) records continuous position within the
primitive cell, and \(\ell_i\) distinguishes translation-suborbit members that
the vector alone cannot identify. The latter is needed only for the selected
nonsymmorphic groups. Together, these quantities provide the
symmetry-compatible geometric input assumed by
Theorem~\ref{thm:wallpaper-equivariance}.}

\paragraph{Deterministic joint orbit lift.}
\explanationaddition{Let \(Q:P\to O(2)\), \(p\mapsto Q_p\), be the standard planar
representation, so that \(Q_{p_g}=Q_g\). For each \(p\in P\), define the
\(p\)-indexed block of the joint orbit lift by}
\begin{equation}
    \bar X_{i,p}
    =
    \left[
        Q_pv_i,\,
        \rho_{\mathrm{mem}}(p)\ell_i,\,
        \bigl(\rho_{\mathrm{mem}}(p)\ell_i\bigr)
        \otimes(Q_pv_i)
    \right].
    \label{eq:joint-orbit-lift}
\end{equation}

\explanationaddition{Stacking these blocks over all \(p\in P\) gives the group-indexed field
\(\bar X_i=(\bar X_{i,p})_{p\in P}\).}

\explanationaddition{The tensor-product term places the transformed WA vector in the slot selected by the transformed one-hot member label, while setting all other member slots to zero. It therefore keeps each WA vector associated with the correct translation-suborbit member. After the fixed group-index conversion detailed in
Appendix~\ref{translation-suborbit-regular-representation}, the resulting
regular field transforms as}
\begin{equation}
    X^{\mathrm{reg}}_{\pi_g(i),k}(gX)
    =
    X^{\mathrm{reg}}_{i,p_g^{-1}k}(X),
    \qquad k\in P.
    \label{eq:regular-registry-lift}
\end{equation}

\paragraph{Learned regular encoder.}

\explanationaddition{Because \(X_i^{\mathrm{reg}}\) transforms as a regular field, it can be
processed by a \(P\)-equivariant regular-to-regular encoder
\(f_{\mathrm{reg}}\), where \(d_{\mathrm{reg}}\) denotes the number of
regular-representation copies in its output. We define the learned registry
feature by}
\begin{equation}
    Z_i
    =
    f_{\mathrm{reg}}\!\left(X_i^{\mathrm{reg}}\right).
    \label{eq:learned-regular-registry}
\end{equation}
\explanationaddition{By equivariance, \(Z_i\) obeys the same transformation law as
\(X_i^{\mathrm{reg}}\) in Equation~\eqref{eq:regular-registry-lift}.
The encoder dimensions, index convention, and layerwise equivariance argument
are given in Appendix~\ref{translation-suborbit-regular-representation}.} In
Appendix~\ref{sec:wallpaper-action-visualizations}, we visualize all point-group
actions of all 17 wallpaper groups on a randomly sampled graph with the
corresponding WAs.

\paragraph{Invariant and equivariant uses.}

For the invariant-message architecture, we project the learned regular
feature onto the trivial representation using the Reynolds average,
\begin{equation}
    L_i
    =
    \frac{1}{|P|}
    \sum_{k\in P} Z_{i,k}.
    \label{eq:registry-reynolds}
\end{equation}
Since the group action only permutes the regular group index,
\(L_{\pi_g(i)}(gX)=L_i(X)\) for all \(g\in G\).
The components of \(L_i\) are therefore invariant scalar features and may be
passed directly to the invariant-message MLP. For the fully equivariant
architecture, the group axis is retained, and \(Z_i\) is passed directly to the
equivariant message-passing network as \(d_{\mathrm{reg}}\) copies of the
regular representation. Retaining the group axis preserves access to both
trivial and nontrivial regular components. A small identifiability experiment
in Appendix~\ref{sec:wa_encoding_exp} isolates the information supplied by the
WA encoding. For the nonsymmorphic groups $pg$, $pmg$, $pgg$, and $p4g$, it
shows that the translation-suborbit member representation can distinguish
configurations.

\subsection{Full wallpaper equivariance}
\label{sec:full-proof}

Let \(G\) be the wallpaper group with primitive translation lattice
\(\Lambda\), and let \(p_g:=g\Lambda\in G/\Lambda\simeq P\).
All feature representations below are representations of \(P\) and are
evaluated at \(p_g\), whereas \(g\) acts affinely on positions. Let
\(\pi_g\) denote the corresponding node reindexing, with
\(\pi_g=\mathrm{id}\) when node indices are retained. Let \(h_i^{(r)}\in V_r\) denote the feature of node \(i\) after
message-passing layer \(r\), where \(V_r\) carries the point-group
representation
\(\rho_r:P\to O(V_r)\). \explanationaddition{The complete layerwise notation for edge
inputs, messages, aggregation weights, and readouts is introduced with the
proof in Appendix~\ref{sec:full-proof-appendix}.}

\begin{theorem}[Wallpaper equivariance of representation-valued graph fields]
\label{thm:wallpaper-equivariance}
Assume that the graph construction and geometric encodings are compatible
with the wallpaper action, including the prescribed anchor-member actions
and unique nearest anchors wherever WA encoding is used. Assume that the
initial node fields transform in their assigned point-group representations,
and that all message functions, node updates, and readouts are equivariant
for these representations. Aggregation is a sum, possibly with invariant
scalar weights. Then every node field \(h_i^{(r)}\) produced by the
network satisfies
{\explanationcolor
\begin{equation}
    h_{\pi_g(i)}^{(r)}(gX)
    =
    \rho_r(p_g)h_i^{(r)}(X),
    \qquad g\in G.
    \label{eq:all-field-wallpaper-equivariance}
\end{equation}
}
\end{theorem}

\explanationaddition{Theorem~\ref{thm:wallpaper-equivariance} is a compositional guarantee. Once
the geometric inputs obey the transformation laws established above, any
stack of compatible equivariant message maps, updates, aggregations, and
readouts remains equivariant under the full wallpaper group. The theorem does
not prescribe one unique architecture, it specifies the conditions that every
WallpaperNet realization must satisfy.}

The proof follows by induction over message-passing layers. The encoded
geometry transforms in its assigned representation, equivariant message maps
preserve this law, and symmetry-compatible sum aggregation commutes with node
reindexing. The complete proof is given in
Appendix~\ref{sec:full-proof-appendix}. The theorem covers both message-passing
constructions: the invariant-message model is recovered by assigning trivial
representations to its hidden features and combining invariant scalar
coefficients with covariant geometric vectors. A permutation-invariant graph
readout into the trivial representation therefore yields an invariant scalar
prediction, as required by Equation~\eqref{eq:target-invariance}.

\explanationaddition{Architecturally, these conditions give the two routes shown in panels (c) and
(d) of Figure~\ref{fig:wallpaper-graph-overview}: retain representation-valued
features through equivariant layers, or scalarize messages to invariants before
ordinary MLP processing. Section~\ref{sec:experiments} compares their concrete
realizations.}

\section{Experiments}
\label{sec:experiments}
\explanationaddition{The experiments address two questions. First, across all 17 wallpaper groups, can a
realization permitted by Theorem~\ref{thm:wallpaper-equivariance} achieve
accuracy comparable to a full-cell model (where every atom is explicitly
modeled), and which permitted design yields the best accuracy? Second, can this
framework handle a realistic catalytic simulation scenario involving a molecule
on a graphene sheet?}


\subsection{Synthetic benchmark and learning targets}

To test whether WallpaperNet can reach the accuracy of full-cell atomistic
models, we introduce a synthetic benchmark covering all 17 wallpaper groups
with 1,000 configurations each: a three- to five-atom carbon fragment above a
fixed, symmetry-constrained carbon support, with varying conformation, height,
and lateral registry. Reference energies and forces are computed with GFN2-xTB
on an explicit $(3\times3)$ support cluster
(Appendix~\ref{sec:si_synthetic_dataset}). We predict the total energy and the
atomic forces, either with a per-node equivariant force head or as the negative
energy gradient. This benchmark provides a controlled setting for testing whether WallpaperNet can learn atomistic energy and force targets across all 17 wallpaper groups. It is designed to isolate geometric and symmetry effects from much of the chemical complexity while retaining a nontrivial target function provided by a realistic simulator.

\subsubsection{Model choices}

We test seven WallpaperNet architectures of increasing complexity, from the
distance-based invariant WP-Inv and its extension WP-InvExt with covariant edge
and Wyckoff-anchor directions, through the regular-representation models
WP-Equi, WP-VecEqui, WP-Reg, and WP-RegTP, to WP-O3, which couples
three-dimensional $O(3)$ tensor features with wallpaper-group-equivariant
processing. The baselines are three full-cell models without wallpaper
symmetry: an invariant EGNN-style GNN and e3nn models with maximum angular degree $\ell_{\max}=1$ and
$2$. All models share the optimizer and learning-rate schedule and have
approximately matched parameter counts
(Appendix~\ref{sec:toy-optimization} and
Table~\ref{tab:toy_model_parameter_counts}). Table~\ref{tab:toy-model-all-groups} summarizes
the results averaged over all 17 groups, and Appendix~\ref{sec:all_toys}
reports the individual tables for every group.

\begin{table*}[t]
\centering
\caption{Toy-model performance averaged over all 17 wallpaper groups. For each seed, metrics are first averaged over groups. Entries report the mean across three seeds with the sample standard deviation as a subscript. C/V denotes combined/vector-head training. Energy MAE is taken from combined training. Energy MAE is in meV atom$^{-1}$ and force-component MAE in meV\,\AA$^{-1}$. The mean group-wise dataset energy SD is 17.52 meV atom$^{-1}$. Bold marks the best mean for each metric and training mode.}
\label{tab:toy-model-all-groups}
\fontsize{8}{9.5}\selectfont
\setlength{\tabcolsep}{3pt}
\begin{tabular}{@{}lccccccc@{}}
\toprule
 & Energy & \multicolumn{2}{c}{Rel.\ force $L_2$} & \multicolumn{2}{c}{Force comp.} & \multicolumn{2}{c}{Force comp.} \\
 & MAE $\downarrow$ & \multicolumn{2}{c}{(\%) $\downarrow$} & \multicolumn{2}{c}{MAE $\downarrow$} & \multicolumn{2}{c}{$R^2$ $\uparrow$} \\
\cmidrule(lr){3-4}\cmidrule(lr){5-6}\cmidrule(lr){7-8}
Model & & C & V & C & V & C & V \\
\midrule
\multicolumn{8}{@{}l}{\textit{WallpaperNet models}} \\
WP-Inv & \(9.42_{\pm 0.09}\) & \(31.5_{\pm 0.6}\) & \(14.1_{\pm 0.4}\) & \(714.4_{\pm 7.7}\) & \(298.6_{\pm 8.4}\) & \(0.900_{\pm 0.004}\) & \(0.980_{\pm 0.001}\) \\
WP-InvExt & \(6.31_{\pm 0.26}\) & \(24.7_{\pm 0.9}\) & \(14.2_{\pm 0.3}\) & \(541.7_{\pm 14.4}\) & \(301.8_{\pm 5.5}\) & \(0.937_{\pm 0.004}\) & \(0.979_{\pm 0.001}\) \\
WP-Equi & \(11.16_{\pm 0.78}\) & \(34.8_{\pm 2.7}\) & \(17.7_{\pm 0.8}\) & \(781.2_{\pm 57.3}\) & \(386.1_{\pm 14.5}\) & \(0.874_{\pm 0.024}\) & \(0.967_{\pm 0.003}\) \\
WP-VecEqui & \(10.93_{\pm 1.05}\) & \(32.9_{\pm 0.4}\) & \(24.3_{\pm 0.9}\) & \(739.7_{\pm 11.7}\) & \(518.3_{\pm 16.9}\) & \(0.890_{\pm 0.002}\) & \(0.938_{\pm 0.004}\) \\
WP-Reg & \(10.53_{\pm 0.28}\) & \(32.9_{\pm 0.7}\) & \(19.1_{\pm 0.8}\) & \(737.6_{\pm 17.6}\) & \(394.7_{\pm 16.6}\) & \(0.891_{\pm 0.005}\) & \(0.963_{\pm 0.003}\) \\
WP-RegTP & \(10.03_{\pm 0.32}\) & \(30.8_{\pm 0.5}\) & \(25.6_{\pm 1.1}\) & \(706.2_{\pm 5.5}\) & \(580.0_{\pm 20.6}\) & \(0.904_{\pm 0.003}\) & \(0.931_{\pm 0.007}\) \\
WP-O3 & \(\mathbf{4.67}_{\pm 0.22}\) & \(\mathbf{18.6}_{\pm 0.1}\) & \(12.3_{\pm 0.5}\) & \(\mathbf{418.9}_{\pm 0.9}\) & \(264.5_{\pm 8.0}\) & \(\mathbf{0.965}_{\pm 0.000}\) & \(0.984_{\pm 0.002}\) \\
\midrule
\multicolumn{8}{@{}l}{\textit{Full-cell baselines}} \\
Invariant GNN & \(7.17_{\pm 0.62}\) & \(26.9_{\pm 1.4}\) & \(17.5_{\pm 2.3}\) & \(594.3_{\pm 31.2}\) & \(373.3_{\pm 51.1}\) & \(0.926_{\pm 0.008}\) & \(0.956_{\pm 0.019}\) \\
e3nn ($\ell_{\max}=1$) & \(5.09_{\pm 0.25}\) & \(20.6_{\pm 0.7}\) & \(11.9_{\pm 0.7}\) & \(456.7_{\pm 12.5}\) & \(\mathbf{251.6}_{\pm 19.0}\) & \(0.957_{\pm 0.003}\) & \(\mathbf{0.986}_{\pm 0.002}\) \\
e3nn ($\ell_{\max}=2$) & \(5.42_{\pm 0.12}\) & \(20.7_{\pm 0.5}\) & \(\mathbf{11.9}_{\pm 0.5}\) & \(466.2_{\pm 12.4}\) & \(254.2_{\pm 15.3}\) & \(0.957_{\pm 0.002}\) & \(0.985_{\pm 0.002}\) \\
\bottomrule
\end{tabular}
\end{table*}

\subsubsection{Discussion}
A per-atom energy error below approximately 43 meV/atom is commonly associated
with chemical accuracy~\citep{rockenEnhancingMachineLearning2025,wuMachineLearningInteratomic2025,ihalageFormulaGraphSelfAttention2022}.
Here, the mean group-wise energy standard deviation of 17.52 meV/atom is a more
informative reference, and because the systems are far from equilibrium, we
compare forces through the force-component \(R^2\) and the relative force
\(L_2\) error \(\lVert f^{\mathrm{pred}}-f\rVert_2/\lVert f\rVert_2\) over all flattened
molecular force components. WP-O3 is the strongest WallpaperNet architecture and slightly
outperforms both full-cell e3nn baselines for energy (\(4.67\) versus \(5.09\)
and \(5.42\,\mathrm{meV\,atom^{-1}}\)) and for all combined-training force
metrics (\(18.6\%\) versus approximately \(20.6\)--\(20.7\%\) relative force
error). For direct vector-force training, the e3nn baselines are slightly
better (\(11.9\%\) versus \(12.3\%\)). For every architecture, the direct force
head is more accurate than the conservative energy gradient. Overall,
WallpaperNet reaches accuracy comparable to full-cell models with less message
passing per system. Further discussion is given in
Appendix~\ref{sec:all_toys}.

\subsection{WallpaperNet for molecular adsorption on graphene}
\label{sec:real-system}

We next test whether the registry-aware construction can learn a realistic
interatomic potential for molecular adsorption on graphene, which is relevant to
sensing, catalysis, electronics, and energy
storage~\citep{kongMolecularAdsorptionGraphene2014,hongRecentProgressOrganic2013,ersanAdsorptionOrganicContaminants2017}
and is commonly modeled with the graphene atoms held
fixed~\citep{chenCrystalShapeControlled2017,londeroDesorptionNalkanesGraphene2012,cunhaWaterDispersibleFewLayer2018,songInfluenceGrapheneElectronic2024,carlsonModelingWaterInteractions2024}.
Graphene has wallpaper group $p6m$, with in-plane point group
$C_{6v}\cong D_6$.

\subsubsection{Dataset and experimental setup}
The dataset contains $15{,}089$ adsorption systems, each with one organic
molecule on frozen, three-layer AAA-stacked graphene, relaxed with
GFN2-xTB~\citep{Bannwarth2019}. The WallpaperNet graph contains only the
adsorbate atoms, on average 24.93 per system versus 386.8 in the full system.
The data are split 80/10/10, and we train for three objectives: (1) the
full-system energy with energy-derived adsorbate forces as an auxiliary term,
(2) adsorbate forces as the negative energy gradient with a small energy term,
and (3) adsorbate forces from a direct equivariant head. Besides WP-Inv,
WP-Equi, WP-Reg, WP-RegTP, and WP-O3, we test WP-InvEq and WP-VecInv, which
transfer the design of WP-InvExt, against the three full-cell baselines scaled
to approximately one million parameters. Details are given in
Appendices~\ref{sec:dataset-construction},
\ref{sec:wallpapernet-architectures_graphene},
and~\ref{sec:real-system-optimization}.

\begin{table*}[t]
\centering
\caption{
Test-set performance averaged over three seeds (mean \(\pm\) sample standard
deviation). Energy metrics use combined training, force columns use
force-weighted conservative (C) or direct vector-head (V) training. Units are
meV atom$^{-1}$ for energy and meV\,$\mathring{\mathrm A}^{-1}$ for forces;
arrows indicate the direction of improvement. Full results, including the
zero-force score, are given in Table~\ref{tab:real_system_test_full}.
}
\label{tab:real_system_test}
\fontsize{8}{9.5}\selectfont
\setlength{\tabcolsep}{2.5pt}
\begin{tabular}{@{}lccccccc@{}}
\toprule
 & Energy & \multicolumn{2}{c}{Rel.\ force $L_2$} & \multicolumn{2}{c}{Force comp.} & \multicolumn{2}{c}{Force comp.} \\
 & MAE $\downarrow$ & \multicolumn{2}{c}{(\%) $\downarrow$} & \multicolumn{2}{c}{MAE $\downarrow$} & \multicolumn{2}{c}{$R^2$ $\uparrow$} \\
\cmidrule(lr){3-4}\cmidrule(lr){5-6}\cmidrule(lr){7-8}
Model & & C & V & C & V & C & V \\
\midrule
\multicolumn{8}{@{}l}{\textit{Wallpaper-symmetry models}} \\
WP-Inv & $\mathbf{1.73}_{\pm 0.02}$ & $80.79_{\pm 1.35}$ & $41.35_{\pm 7.79}$ & $139.9_{\pm 0.6}$ & $75.0_{\pm 16.3}$ & $0.35_{\pm 0.02}$ & $0.82_{\pm 0.07}$ \\
WP-InvEq & $4.75_{\pm 5.09}$ & $78.68_{\pm 1.15}$ & $25.67_{\pm 0.63}$ & $137.1_{\pm 2.0}$ & $43.5_{\pm 1.0}$ & $0.38_{\pm 0.02}$ & $0.93_{\pm 0.00}$ \\
WP-Equi & $1.90_{\pm 0.17}$ & $80.59_{\pm 0.94}$ & $27.95_{\pm 0.80}$ & $140.7_{\pm 1.4}$ & $48.8_{\pm 0.9}$ & $0.35_{\pm 0.02}$ & $0.92_{\pm 0.00}$ \\
WP-VecInv & $2.24_{\pm 0.06}$ & $62.49_{\pm 15.95}$ & $26.62_{\pm 0.93}$ & $109.5_{\pm 25.4}$ & $44.8_{\pm 0.8}$ & $0.59_{\pm 0.21}$ & $0.93_{\pm 0.01}$ \\
WP-Reg & $2.09_{\pm 0.12}$ & $30.70_{\pm 1.90}$ & $27.15_{\pm 1.07}$ & $54.0_{\pm 2.7}$ & $45.7_{\pm 1.8}$ & $0.91_{\pm 0.01}$ & $0.93_{\pm 0.01}$ \\
WP-RegTP & $1.86_{\pm 0.24}$ & $30.49_{\pm 1.33}$ & $26.21_{\pm 1.14}$ & $52.7_{\pm 2.3}$ & $44.6_{\pm 1.1}$ & $0.91_{\pm 0.01}$ & $0.93_{\pm 0.01}$ \\
WP-O3 & $1.80_{\pm 0.08}$ & $\mathbf{19.78}_{\pm 0.70}$ & $\mathbf{19.22}_{\pm 0.75}$ & $\mathbf{32.9}_{\pm 0.9}$ & $\mathbf{31.3}_{\pm 0.2}$ & $\mathbf{0.96}_{\pm 0.00}$ & $\mathbf{0.96}_{\pm 0.00}$ \\
\midrule
\multicolumn{8}{@{}l}{\textit{Full-cell baselines}} \\
Invariant GNN & $2.58_{\pm 0.23}$ & $67.51_{\pm 1.03}$ & $47.22_{\pm 6.46}$ & $119.9_{\pm 2.5}$ & $83.6_{\pm 12.2}$ & $0.54_{\pm 0.01}$ & $0.77_{\pm 0.06}$ \\
e3nn ($\ell_{\max}=1$) & $1.92_{\pm 0.07}$ & $25.89_{\pm 0.89}$ & $25.31_{\pm 3.50}$ & $44.7_{\pm 2.1}$ & $43.9_{\pm 7.2}$ & $0.93_{\pm 0.00}$ & $0.94_{\pm 0.02}$ \\
e3nn ($\ell_{\max}=2$) & $2.04_{\pm 0.22}$ & $21.43_{\pm 0.75}$ & $25.55_{\pm 5.82}$ & $36.2_{\pm 1.2}$ & $45.0_{\pm 11.1}$ & $0.95_{\pm 0.00}$ & $0.93_{\pm 0.03}$ \\
\bottomrule
\end{tabular}
\end{table*}

\subsubsection{Discussion}
Table~\ref{tab:real_system_test} shows that WallpaperNet learns an accurate
interatomic potential for molecular adsorption on graphene without representing
the substrate explicitly. In energy-dominated training, all WallpaperNet models
remain far below the chemical-accuracy threshold of
\(43\,\mathrm{meV\,atom^{-1}}\), and the simplest invariant model, WP-Inv, gives
the lowest energy MAE of \(1.73\,\mathrm{meV\,atom^{-1}}\), below both full-cell
e3nn baselines. The ordering changes for forces: with a direct vector head,
WP-O3 gives the lowest error of all ten models, \(19.22\%\) relative force error
and \(31.3\,\mathrm{meV\,\mathring{A}^{-1}}\) component MAE, compared with
approximately \(25\%\) for the full-cell e3nn baselines. After rebalancing the
energy and force terms, WP-O3 recovers almost the same accuracy conservatively
(\(19.78\%\)), whereas several other architectures do not, showing that an
equivariant force output alone does not guarantee an energy surface with
accurate derivatives. These results are obtained on graphs with approximately
\(15\times\) fewer nodes and \(45\times\) fewer edges than the full-cell
representation and with lower training memory. The full discussion, including
the loss weighting and the memory and timing benchmarks, is given in
Appendix~\ref{sec:real-system-discussion}.

\section{Conclusion}
\label{sec:conclusion}
We introduced WallpaperNet, a general blueprint for wallpaper-group-equivariant
graph neural networks. It separates the symmetry of a graph embedded in a
periodic structure into point-group equivariance and a lattice-relative
registry, which is encoded by a Wyckoff-anchor vector and, for nonsymmorphic
groups, a translation-suborbit member label. We showed how to use the blueprint
by building seven architectures and evaluating them on a synthetic benchmark
covering all 17 wallpaper groups, where the strongest model, WP-O3, is on par
with full-cell e3nn baselines. We then applied it to atoms by learning an
interatomic potential for molecular adsorption on graphene. Because the
substrate enters through its symmetry rather than as explicit nodes, the
message-passing graph has roughly \(15\times\) fewer nodes and \(45\times\) fewer
edges. Every WallpaperNet model therefore needs less training memory than the
full-cell baselines, all of them predict energies well below chemical accuracy,
and in direct-force training WP-O3 both reaches the lowest force error of all
models and trains faster than every full-cell baseline.

\paragraph{Limitations and future work.}
WallpaperNet assumes a fixed periodic support with a known symmetry group and
is trained for that support. Standard MLIPs, in contrast, apply to arbitrary
atomic systems, and pretrained universal potentials require no system-specific
training. At the same time, crystals offer many more fixed geometric settings
than the 17 wallpaper groups, e.g., the 230 space groups. Extending the blueprint to these
settings is a promising direction for future work.

\subsection*{AI use statement}

In this work, we used generative AI tools to assist with research execution, software development, literature discovery, and the writing and presentation of the paper.

Generative AI was used during the development of \texttt{we3nn}, including assistance with implementing methods and writing supporting software. All AI-assisted implementation was reviewed by a human author. Where possible, the correctness of the implementation was additionally assessed by comparing its behavior and results with the established e3nn and escnn libraries. Generative AI was also used to create and modify auxiliary scripts and tools, including scripts for extracting data from experiment logs and for submitting computational jobs to a cluster.

We used AI-based chat tools for literature discovery by asking them to identify potentially relevant related work. The resulting references were subsequently checked, and the relevant papers were read and assessed by a human author rather than relying on AI-generated summaries or descriptions.

Generative AI was also used to provide feedback on experimental design and execution. Suggestions produced by the AI were reviewed by the main author before being incorporated into the research process. In addition, we used AI tools to inspect equations and notation for potential inconsistencies or collisions in notation. We also used AI to convert equations from LaTeX to Python code, which was then reviewed and tested by a human author. We also used AI to describe the code in a human-readable format, which was then reviewed and edited by a human author.

For preparation of the manuscript, generative AI was used for text editing, formatting, and stylistic feedback. Generative AI was used to draft passages from author-provided outlines, to correct grammatical errors and typographical mistakes, and to assess whether passages were understandable to readers from different research backgrounds. We also used AI tools to obtain suggestions about which material could potentially be moved to the appendix.

AI was used to generate synthetic data sets, in the sense that it was used to write code, that after execution, generated synthetic data sets. But we never used AI output as synthetic data directly. 

All AI-assisted research, code, literature suggestions, and manuscript edits were reviewed by the main author. The authors take responsibility for the final content of this work, including all text, claims, code, results, and other artifacts produced with the assistance of generative AI.

AI was not used to develop theoretical models or conceptual frameworks, formulate mathematical claims, or provide critical ingredients for proving mathematical claims. AI was not used for assistance with translation, cleaning or reformat datasets and it was not used for independent data analysis. We did not use generative AI to generate any of the research ideas.

\subsection*{Reproducibility statement}

All finite-group equivariant layers are built with our \texttt{we3nn} library,
which is publicly available at
\url{https://github.com/lamalab-org/we3nn/tree/main} and described in
Appendix~\ref{sec:we3nn}. The WallpaperNet architectures and the full-cell
baselines are described in Appendix~\ref{architecture-descriptions}, and their
training objectives and optimization settings for the synthetic and the
molecule--graphene experiments in
Appendices~\ref{sec:toy-objectives}--\ref{sec:real-system-optimization}. The
construction of the synthetic benchmark and of the molecule--graphene dataset is
described in Appendices~\ref{sec:si_synthetic_dataset}
and~\ref{sec:dataset-construction}, the Wyckoff-anchor identifiability
experiment in Appendix~\ref{sec:wa_encoding_exp}, and the per-group and per-seed
results in Appendices~\ref{sec:all_toys} and~\ref{sec:mol-grap-results}.

\clearpage
\appendix
\startcontents[appendix]
\section*{Appendix contents}
\printcontents[appendix]{l}{1}{\setcounter{tocdepth}{2}\setlength{\parskip}{1.5pt}}
\clearpage

\section{Extended related work}
\label{sec:extended-related-work}

\paragraph{Equivariant geometric learning.}
\citet{satorras2021egnn} construct EGNN vector updates from invariant scalar
messages and relative displacement vectors. \citet{thomas2018tfn} instead
introduce representation-valued features and spherical-harmonic filters
coupled through equivariant tensor products. \citet{geiger2022e3nn} provide
the \texttt{e3nn} framework for constructing such $E(3)$-equivariant
operations. \citet{PaiNN} propagate scalar and vector features through
equivariant message passing, while \citet{batznerE3equivariantGraphNeural2022}
apply higher-order $E(3)$-equivariant interactions to interatomic potentials
for molecular and materials systems.
These approaches provide building blocks for our scalar- and
representation-valued architectures. Their Euclidean equivariance should,
however, be distinguished from the registry problem considered here:
transforming an adsorbate and substrate jointly preserves their relative
configuration, whereas translating the adsorbate relative to a fixed periodic
substrate changes its registry.

\paragraph{Finite-group equivariant constructions.}
\citet{cohen2016group} introduce group-equivariant convolutional networks that
represent transformations through group-indexed features and shared filters.
\citet{finziPracticalMethodConstructing2021} construct equivariant layers by
solving representation constraints, while \citet{cesa2021ENsteerable} provide
a general Wigner--Eckart construction of steerable kernel spaces. Our
\texttt{we3nn} implementation
builds on these representation-theoretic principles to parameterize finite
planar-group linear and bilinear maps. Regular fields support componentwise
nonlinearities, while intertwiner spaces specify the allowed interactions.
The general mathematical equivariance criterion is therefore established. Our focus is on its extension to graphs and its use with cell-registry encodings.

\paragraph{Crystalline and surface models.}
\citet{xie2018cgcnn} represent crystalline environments as atomic graphs for
property prediction in CGCNN. \citet{kaba2022crystal} explicitly incorporate
crystalline symmetry into generalized message-passing operations, and
\citet{ruff2024connectivity} exploit asymmetric-unit graph representations to
reduce the graph size for crystalline materials. These are close precedents for
crystallographic inductive biases and symmetry-based graph reduction.
In contrast to explicit substrate-atom
representations, our formulation treats the substrate as a fixed periodic
background and supplies its registry to an adsorbate-only graph through
Wyckoff-anchor geometry and equivariant member labels. The emphasis is on
retaining this registry, including nonsymmorphic member permutations,
without constructing substrate nodes in the message-passing graph.

\paragraph{Wallpaper-group symmetry.}
The 17 wallpaper, or crystallographic plane, groups were established in
foundational classifications by \citet{fedorov1891symmetry} and
\citet{polya1924analogie}.
These groups combine lattice translations with compatible rotations,
reflections, and glide reflections and form the two-dimensional analogue of
the crystallographic space groups. Standard crystallographic tabulations
specify their symmetry operations, general and special positions, and Wyckoff
positions~\citep{aroyo2006bcs1,internationaltables2016}.
Representation-theoretic data for the plane groups have likewise been
tabulated systematically~\citep{litvin1991character}. Our construction uses
this structure as an inductive bias: Wyckoff-related geometry identifies the
registry of a graph relative to the periodic background, while finite-group representations determine how the corresponding features transform. Additionally, our blueprint is applicable to GNNs on graphs with wallpaper symmetry, which has applications beyond chemistry and materials science.

\section{Theory and implementation details}
\subsection{Point-group-equivariant message passing: derivations and conventions}
\label{sec:mp-details}

\paragraph{Cyclic and dihedral scalarization.}
\label{sec:scalarization-details}
For \(\psi_{C_n}\) in Equation~\eqref{eq:cn-invariant}: under a generator
rotation by $2\pi/n$, the phase acquires $e^{2\pi\mathrm{i}/n}$ and its $n$th
power is unchanged. In \(\psi_{D_n}\), Equation~\eqref{eq:dn-invariant}, the
imaginary part is omitted because it changes sign under reflection. At $v=0$,
continuous polynomial versions using $z(v)^n$ may be used instead of
normalized angular features. Besides $\psi_P(r_{ij,xy})$ and
$\psi_P(v_{\WA,i})$, the invariant message input may contain joint invariants
such as dot and (if reflection is absent) out-of-plane cross products. The
radial bases, the joint invariants used by each architecture, and the input
dimensions are given in Appendix~\ref{sec:wallpapernet-architectures_graphene}.

\paragraph{Scalar-weighted vector reconstruction.}
Although each edge term \(w_{ij}r_{ij,xy}\) is parallel to its own edge, the
aggregated vector \(F_{i,xy}\) is generally not parallel to any individual
edge. Two non-collinear in-plane neighbors already span $\R^2$.

\paragraph{Equivariant linear and bilinear maps.}
\label{sec:equivariant-maps-details}
Irreducible representations are representations with no nonzero proper
invariant subspace. Direct sums of irreducible representations describe
general finite-group feature types. Bilinear interactions are described by the
same construction as Equation~\eqref{eq:mp-hom} with a tensor-product input.
Writing \(\rho_{12}(p)=\rho_1(p)\otimes\rho_2(p)\), their linearizations belong to
\begin{equation}
\begin{split}
    \operatorname{Hom}_P(U_1\otimes U_2,V)
    =\bigl\{C\in\operatorname{Hom}_{\mathbb R}(U_1\otimes U_2,V):
    \qquad\qquad\\
    C\rho_{12}(p)=\rho_V(p)C\quad\forall p\in P\bigr\}.
\end{split}
    \label{eq:mp-bilinear-hom}
\end{equation}
Specifically, \(T(x,y)=C(x\otimes y)\) is bilinear and equivariant precisely
when \(C\) belongs to this space.

We parameterize affine linear layers and bilinear
interactions as
\begin{equation}
\begin{aligned}
    L_\theta(x)&=W_\theta x+b,
    &W_\theta&=\sum_a\theta_a B_a,\\
    T_\eta(x,y)&=C_\eta(x\otimes y),
    &C_\eta&=\sum_c\eta_c C_c,
\end{aligned}
    \label{eq:mp-hom-parameterization}
\end{equation}
where \(\{B_a\}\) and \(\{C_c\}\) are fixed bases of the corresponding
intertwiner spaces for the selected representation blocks. Biases satisfy
\(b\in V^P=\{v\in V:\rho_V(p)v=v\ \forall p\in P\}\).
Multiplicity-channel connectivity is specified separately. Tensor-product
coefficients may also be generated from invariant contextual features, for example, a radial MLP may generate tensor-product coefficients from
\(\lVert r_{ij}\rVert\), while preserving equivariance of the complete operation.

\paragraph{Construction of the symmetry bases.}
\label{sec:basis-construction}
Both spaces are instances of \(\operatorname{Hom}_P(D,V)\), with
\(D=U\) for linear maps and \(D=U_1\otimes U_2\) for bilinear interactions.
Their bases can be obtained by Reynolds averaging followed by numerical
basis extraction~\citep{thomasComparingCrystalStructures2021}, by solving the linear intertwiner constraints as a
null-space problem~\citep{finziPracticalMethodConstructing2021}, or through an
irreducible-decomposition construction combining Clebsch--Gordan
coefficients and the required real endomorphism
bases~\citep{cesa2021ENsteerable}. These are alternative constructions of the
same target intertwiner space, not different equivariance requirements.
For blocks involving regular representations, \texttt{we3nn} exploits
analytic formulas for these spaces, as detailed in
Appendix~\ref{sec:we3nn}.

\paragraph{Regular fields.}
\label{sec:regular-field-details}
For \(H\in\mathbb N\) copies of the regular field of
Equation~\eqref{eq:mp-regular-action}, let
\(\mathcal H=\mathbb R[P]^{\oplus H}\), with
\begin{equation}
    \rho_{\mathcal H}(p)=I_H\otimes\lambda(p),
\end{equation}
where \(I_H\) is the identity on the copy index. Flattening uses copy-major
order, with the group coordinate varying fastest. Invariant gates have one
value per regular copy and are shared over its group coordinates.

\subsection{\texttt{we3nn}: finite-group equivariant linear and bilinear maps}
\label{sec:we3nn}

\paragraph{Notation.} Appendices~\ref{sec:we3nn}--\ref{sec:we3nn-tp-weights}
describe \texttt{we3nn} as a standalone
library. Here \(G\) denotes a finite cyclic or dihedral group, the point group
\(P\) of the main text, with elements \(g,h,q\), and \(\rho\) denotes a generic
feature representation, written \(\rho_U,\rho_V\) in the main text. These
symbols do not refer to the wallpaper group or the linear-part map. The indices
\(\lambda,\mu,\nu\) label irreducible representations \(V_\lambda\) with
matrices \(\rho_\lambda\). They are unrelated to the left-regular action
\(\lambda\) of the main text, and the regular representation is written
\(\rho_{\mathrm{reg}}\). Dimensions of intertwiner spaces are denoted \(K\),
and \(p\) indexes their basis elements.

To support representation-valued GNNs for finite cyclic and
dihedral groups, we developed \texttt{we3nn}. The library constructs exact
equivariant linear and bilinear maps between finite-dimensional
representations of

$$
    G\in\{C_n,D_n\}.
$$

Although the library also exposes
a generic finite-group interface, the discussion below is restricted to its
built-in \(C_n\) and \(D_n\) representations. The central design principle is to parameterize only the vector space of maps
that commute with the prescribed group actions.

\paragraph{Motivation.}
\texttt{we3nn} is motivated by the gap between \texttt{escnn} and \texttt{e3nn}: it brings the flexible point-group equivariance concepts developed in \texttt{escnn} to graph and message-passing neural networks, compatible with \texttt{e3nn}. Many useful abstractions and features already exist in \texttt{escnn} and can be adapted, and \texttt{we3nn} deliberately borrows much of its terminology, notation, and API style. Graph networks, however, also require additional primitives, most notably bilinear tensor products. While the interface is inspired by \texttt{escnn}, the core implementation is different: rather than relying on explicit Clebsch--Gordan decompositions and Wigner--Eckart constructions, \texttt{we3nn} builds equivariant maps directly from finite-group representation constraints and intertwiners.

\paragraph{Representations and feature types.}

Let \(V\) be a finite-dimensional real vector space carrying a representation

$$
    \rho:G\rightarrow \operatorname{GL}(V),
$$
Where \(\operatorname{GL}(V)\) is the group of invertible linear maps on \(V\). A feature \(x\in V\) transforms according to

$$
    x\longmapsto \rho(g)x.
$$

Feature types may contain several representations,

$$
    V
    =
    \bigoplus_{\lambda \in \operatorname{Irr}(G)}
    m_{\lambda}V_{\lambda},
$$

where \(\operatorname{Irr}(G)\) is the set of irreducible representations, and \(m_{\lambda}\) denotes the number of copies, or multiplicity, of representation \(V_{\lambda}\). The learnable operators are constructed separately for each compatible pair
or triple of representation types. Equal representation occurrences share
the same symmetry basis, while independent trainable coefficients mix their multiplicity channels.

\subsubsection{Equivariant linear maps}

Consider a linear map

$$
    y=Wx,
    \qquad
    W:V_{\mathrm{in}}\rightarrow V_{\mathrm{out}},
$$

where the input and output transform according to
\(\rho_{\mathrm{in}}\) and \(\rho_{\mathrm{out}}\). Equivariance requires
\begin{equation}
\rho_{\mathrm{out}}(g)W
=
W\rho_{\mathrm{in}}(g)
\qquad
\forall g\in G.
\label{eq:we-linear-condition}
\end{equation}
The admissible matrices therefore form the intertwiner space
\begin{equation}
\operatorname{Hom}_{G}
(V_{\mathrm{in}},V_{\mathrm{out}})
=
\left\{
W:
\rho_{\mathrm{out}}(g)W
=
W\rho_{\mathrm{in}}(g)
\quad
\forall g\in G
\right\}.
\label{eq:linear-hom-space}
\end{equation}

Let

$$
    \{B_p\}_{p=1}^{K}
$$

be a basis of this space, where

$$
    K
    =
    \dim
    \operatorname{Hom}_{G}
    (V_{\mathrm{in}},V_{\mathrm{out}}),
$$

then \texttt{WE\_Linear} parameterizes the learned matrix as
\begin{equation}
W
=
\sum_{p=1}^{K}
\theta_p B_p.
\label{eq:we-linear-expansion}
\end{equation}
The basis matrices \(B_p\) are fixed by symmetry and only the reduced
coefficients \(\theta_p\) are trainable. Consequently,
Equation~\eqref{eq:we-linear-condition} holds for every value of the learned
parameters.

When multiple copies of the input \(V_{\mathrm{\lambda}}\) and output representations \(V_{\mathrm{\mu}}\) are present,
the coefficients become

$$
    \theta_{rcp}^{\mathrm{\mu} \mathrm{\lambda}},
$$

where \(c\) indexes input multiplicity and \(r\) output multiplicity:
\begin{equation}
W^{\mathrm{\mu} \mathrm{\lambda}}_{rc}
=
\sum_{p=1}^{K_{\mathrm{\mu} \mathrm{\lambda}}}
\theta_{rcp}^{\mathrm{\mu} \mathrm{\lambda}}B_{p}^{\mathrm{\mu} \mathrm{\lambda}}.
\label{eq:linear-multiplicity}
\end{equation}

\paragraph{Generic representation-to-representation linear map.}

When neither representation is regular, the default structural backend constructs the
intertwiner basis through finite-group averaging. Starting from the elementary
matrix basis \(E_{oi} \in \mathbb R^{d_{\mathrm{out}}\times d_{\mathrm{in}}}\), whose \((o,i)\) entry is 1 and all other entries are 0, each matrix is
projected with the Reynolds operator
\begin{equation}
\Pi_{\mathrm{lin}}(A)
=
\frac{1}{|G|}
\sum_{g\in G}
\rho_{\mathrm{out}}(g)
A
\rho_{\mathrm{in}}(g)^{\mathsf T}.
\label{eq:linear-reynolds}
\end{equation}
The representations used here are orthogonal, so
\(\rho(g)^{-1}=\rho(g)^{\mathsf T}\). Hence
\(\Pi_{\mathrm{lin}}(A)\) is the group average of the natural action
\(A\mapsto\rho_{\mathrm{out}}(g)A\rho_{\mathrm{in}}(g)^{-1}\) on linear maps.
The projected elementary matrices span the complete intertwiner space.  The
implementation flattens each matrix to a vector, applies a singular-value
decomposition (SVD) to remove linear dependencies, and reshapes the retained
right-singular vectors into a Frobenius-orthonormal basis
\(\{B_p\}_{p=1}^{K}\).  The optional generic backend can equivalently obtain
the same space by solving the linear null-space constraints in
Equation~\eqref{eq:we-linear-condition}.

For irreducible representations, the statement can be sharpened.  If
\(V_\lambda\) and \(V_\mu\) are inequivalent irreps, then according to Schur's lemma \citep{serreLinearRepresentationsFinite2014},
\[
    \operatorname{Hom}_{G}(V_\lambda,V_\mu)=\{0\}.
\]
If they are equivalent, the intertwiner space is isomorphic to the real
endomorphism algebra
\[
    \operatorname{End}_{G}(V_\lambda)
    =
    \operatorname{Hom}_{G}(V_\lambda,V_\lambda).
\]
Over \(\mathbb R\), this algebra need not be one-dimensional.  In particular,
a non-one-dimensional real irrep of \(C_n\) is of complex type and has a
two-dimensional real endomorphism algebra.  In a standard rotation basis,
these two directions may be represented by the identity matrix and the
\(90^\circ\) rotation matrix.  The implementation therefore constructs the
complete real intertwiner space instead of assuming one scalar degree of
freedom for every equal-irrep pair.

\subsubsection{Irrep-to-regular linear map}

Let \(V_{\lambda}\) be a representation of dimension \(d_{\lambda}\), and let

$$
    \mathbb R[G]
    =
    \operatorname{span}
    \{\delta_h:h\in G\}
$$

denote the regular representation. For each \(h \in G\), let \(\delta_h\) denotes the basis vector with the  group element \(h\) in the \(h\)-th coordinate and zeros elsewhere.  The left-regular action of \(G\) on \(\mathbb R[G]\) is defined by
\begin{equation}
\rho_{\mathrm{reg}}(g)\delta_h
=
\delta_{gh}.
\label{eq:left-regular-definition}
\end{equation}

Here \(g,\in G\) denotes the group element acting on the feature, \(h,q\in G\) indexes the original regular coordinate, and the transformed regular coordinate respectively. Consequently, if
\(f=\sum_{h\in G}f_h\delta_h\in\mathbb R[G]\), then its transformed
coordinates obey
\[
    [\rho_{\mathrm{reg}}(g)f]_q
    =
    f_{g^{-1}q},
    \qquad q\in G.
\]

For an equivariant map
\[
    W:
    V_\lambda
    \longrightarrow
    \mathbb R[G],
\]
one has
\begin{equation}
\dim_{\mathbb R}
\operatorname{Hom}_{G}
(V_\lambda,\mathbb R[G])
=
d_\lambda.
\end{equation}
Thus a single \(V_\lambda\)-to-regular block has \(d_\lambda\) independent
symmetry-allowed directions.

Let \(\{\mathbf e_i\}_{i=1}^{d_\lambda}\) be the coordinate basis of
\(V_\lambda\), where \(i\) indexes coordinates within the representation.
The analytic Frobenius-orthonormal basis used by \texttt{we3nn} for
\(\operatorname{Hom}_G(V_\lambda,\mathbb R[G])\) is
\begin{equation}
(B_\alpha)_{h,i}
=
\frac{1}{\sqrt{|G|}}
[\rho_\lambda(h)]_{i\alpha},
\qquad
\alpha,i\in\{1,\ldots,d_\lambda\},
\quad h\in G.
\label{eq:irrep-to-reg-basis}
\end{equation}
Here \(B_\alpha\) is the \(\alpha\)-th basis intertwiner,
\(\alpha\) labels the \(d_\lambda\) independent equivariant maps,
\(h\) labels a coordinate of the regular output, and \(i\) labels a
coordinate of the input representation \(V_\lambda\).

The learned equivariant map is a linear combination of these fixed basis
intertwiners,
\begin{equation}
W
=
\sum_{\alpha=1}^{d_\lambda}
\theta_\alpha B_\alpha,
\end{equation}
where
\[
\theta
=
(\theta_1,\ldots,\theta_{d_\lambda})^{\mathsf T}
\in\mathbb R^{d_\lambda}
\]
contains the learned scalar coefficients.

For an input
\[
x=(x_1,\ldots,x_{d_\lambda})^{\mathsf T}\in V_\lambda,
\]
the output \(f=Wx\in\mathbb R[G]\) contains one scalar coordinate
\(f_h\) for each group element \(h\in G\). Its \(h\)-th coordinate is
\begin{align}
f_h
&=
\sum_{i=1}^{d_\lambda}W_{hi}x_i
\\
&=
\frac{1}{\sqrt{|G|}}
\sum_{\alpha=1}^{d_\lambda}
\sum_{i=1}^{d_\lambda}
\theta_\alpha
[\rho_\lambda(h)]_{i\alpha}
x_i
\\
&=
\frac{1}{\sqrt{|G|}}
\theta^{\mathsf T}
\rho_\lambda(h)^{\mathsf T}x.
\label{eq:irrep-to-reg}
\end{align}
Equivalently,
\begin{equation}
f_h
=
\frac{1}{\sqrt{|G|}}
[\rho_\lambda(h)\theta]^{\mathsf T}x.
\end{equation}
Thus, the same learned probe \(\theta\) is transformed by every group element
\(h\), and the corresponding transformed probe
\(\rho_\lambda(h)\theta\) is compared with the input \(x\) through an inner
product. The resulting scalar is stored in the regular coordinate indexed by
\(h\).

For the standard two-dimensional \(C_6\) vector representation, for example,
the output contains six coordinates,
\[
(f_0,\ldots,f_5),
\]
and each \(f_h\) measures the inner product of the input vector with a
different \(C_6\)-rotated copy of the learned probe \(\theta\).

\subsubsection{Regular-to-irrep linear map}

Consider the reverse map
\[
    W:
    \mathbb R[G]
    \longrightarrow
    V_\lambda.
\]
Again,
\begin{equation}
\dim_{\mathbb R}
\operatorname{Hom}_{G}
(\mathbb R[G],V_\lambda)
=
d_\lambda.
\end{equation}
An equivariant map is completely determined by the image of the identity
basis vector \(\delta_e\).  Indeed, if \(W\delta_e=v\in V_\lambda\), then for
any \(h\in G\),
\begin{align}
W\delta_h
&=
W\rho_{\mathrm{reg}}(h)\delta_e
\\
&=
\rho_\lambda(h)W\delta_e
\\
&=
\rho_\lambda(h)v.
\end{align}

The analytic orthonormal basis is
\begin{equation}
(B_\alpha)_{i,h}
=
\frac{1}{\sqrt{|G|}}
[\rho_\lambda(h)]_{i\alpha},
\qquad
\alpha,i\in\{1,\ldots,d_\lambda\},
\quad h\in G.
\label{eq:reg-to-irrep-basis}
\end{equation}
For a regular input \(f=\sum_{h\in G}f_h\delta_h\) and learned coefficient
vector
\(\theta=(\theta_1,\ldots,\theta_{d_\lambda})^{\mathsf T}\), the output is
\begin{equation}
z
=
\frac{1}{\sqrt{|G|}}
\sum_{h\in G}
f_h
\rho_\lambda(h)\theta.
\label{eq:reg-to-irrep}
\end{equation}
In components,
\[
    z_i
    =
    \frac{1}{\sqrt{|G|}}
    \sum_{h\in G}
    \sum_{\alpha=1}^{d_\lambda}
    f_h[\rho_\lambda(h)]_{i\alpha}\theta_\alpha.
\]
With this orthonormal convention,
\(\theta=\sqrt{|G|}\,W\delta_e\).  Thus the layer synthesizes an ordinary
representation-valued feature by combining group-indexed regular coordinates
with symmetry-related copies of a learned template.

\subsubsection{Regular-to-regular linear map}

Let
\[
    W:
    \mathbb R[G]
    \longrightarrow
    \mathbb R[G],
\]
and write the input and output as
\(f=\sum_{h\in G}f_h\delta_h\) and
\(z=\sum_{q\in G}z_q\delta_q\), respectively.  Equivariance under the
left-regular action implies
\begin{equation}
W_{q,h}
=
\frac{1}{\sqrt{|G|}}
\theta_{h^{-1}q},
\qquad q,h\in G,
\label{eq:reg-reg-linear}
\end{equation}
where \(W_{q,h}\) is the matrix entry from input coordinate \(h\) to output
coordinate \(q\), and \(\theta_\eta\) is one learned scalar for each relative
group element \(\eta=h^{-1}q\in G\).  Equivalently,
\[
    z_q
    =
    \frac{1}{\sqrt{|G|}}
    \sum_{h\in G}
    \theta_{h^{-1}q}f_h.
\]
This is a finite-group convolution in the indexing convention induced by the
library's left-regular action.

An unconstrained regular-to-regular matrix contains \(|G|^2\) coefficients,
whereas the equivariant map contains only \(|G|\) independent coefficients.
With \(U\) regular input copies and \(M\) regular output copies,
\[
    \mathbb R[G]^{\oplus U}
    \longrightarrow
    \mathbb R[G]^{\oplus M},
\]
the operation is
\[
    z_{m,q}
    =
    \frac{1}{\sqrt{|G|}}
    \sum_{u=1}^{U}
    \sum_{h\in G}
    \theta_{m u,\,h^{-1}q}f_{u,h},
\]
where \(u\) indexes an input copy, \(m\) indexes an output copy, and the
learnable tensor has shape
\begin{equation}
M\times U\times |G|.
\end{equation}

The factor \(1/\sqrt{|G|}\) in the analytic regular-representation formulas
makes the symmetry bases orthonormal in the Euclidean/Frobenius inner product.
It is a normalization convention rather than an additional symmetry
constraint, and it can be absorbed into the learned coefficients.

\subsubsection{Equivariant bias and nonlinearities}

For an affine map
\[
    y=Wx+b,
\]
equivariance additionally requires
\begin{equation}
\rho_{\mathrm{out}}(g)b=b
\qquad
\text{for every }g\in G.
\end{equation}
The allowed bias therefore belongs to the invariant subspace
\[
    V_{\mathrm{out}}^G
    =
    \{v\in V_{\mathrm{out}}:
      \rho_{\mathrm{out}}(g)v=v
      \text{ for every }g\in G\}
    \cong
    \operatorname{Hom}_{G}
    (\mathbb R_{\mathrm{triv}},V_{\mathrm{out}}),
\]
where \(\mathbb R_{\mathrm{triv}}\) is the one-dimensional trivial
representation, for which \(\rho_{\mathrm{triv}}(g)=1\).  A nontrivial vector
irrep has no invariant vector and therefore admits no arbitrary vector bias.
A regular field has a one-dimensional invariant subspace, corresponding to a
bias that is constant over all group coordinates.

The regular representation is a permutation representation.  More generally,
if a representation acts by permuting its coordinates, then any scalar
nonlinearity \(\sigma:\mathbb R\to\mathbb R\) applied identically to every
coordinate is equivariant.  For a regular feature \(f\),
\begin{equation}
\sigma(\rho_{\mathrm{reg}}(g)f)
=
\rho_{\mathrm{reg}}(g)\sigma(f),
\label{eq:reg-nonlinearity}
\end{equation}
where \(\sigma(f)\) denotes coordinatewise application.  This permits
ordinary nonlinearities such as ReLU between regular-to-regular equivariant
linear layers. 

\subsection{Equivariant tensor products}

Whereas equivariant linear layer parameterizes equivariant linear maps,
the Tensor Product and Fully Connected TensorProduct parameterize
equivariant bilinear maps.

Let
\[
    x\in V_\lambda,
    \qquad
    y\in V_\mu,
\]
where \(V_\lambda\) and \(V_\mu\) have dimensions \(d_\lambda\) and
\(d_\mu\), and transform as
\[
    x\longmapsto\rho_\lambda(g)x,
    \qquad
    y\longmapsto\rho_\mu(g)y.
\]
Their raw tensor product belongs to \(V_\lambda\otimes V_\mu\) and transforms
according to
\begin{equation}
\rho_{\lambda\otimes\mu}(g)
=
\rho_\lambda(g)\otimes\rho_\mu(g).
\label{eq:tensor-product-representation}
\end{equation}

Suppose the desired output belongs to \(V_\nu\), with dimension \(d_\nu\).
An equivariant bilinear map
\[
    B:V_\lambda\times V_\mu\longrightarrow V_\nu
\]
is equivalent to a linear intertwiner
\[
    C:
    V_\lambda\otimes V_\mu
    \longrightarrow
    V_\nu.
\]
Thus all equivariant bilinear maps form the real vector space
\begin{equation}
\operatorname{Hom}_{G}
(V_\lambda\otimes V_\mu,V_\nu).
\label{eq:tp-hom-space}
\end{equation}

Let
\[
    K_{\nu;\lambda\mu}
    =
    \dim_{\mathbb R}
    \operatorname{Hom}_{G}
    (V_\lambda\otimes V_\mu,V_\nu),
\]
and let
\(\{C_p\}_{p=1}^{K_{\nu;\lambda\mu}}\) be a basis of this space.  A basis
tensor has entries
\[
    (C_p)_{oij},
    \qquad
    o=1,\ldots,d_\nu,\quad
    i=1,\ldots,d_\lambda,\quad
    j=1,\ldots,d_\mu.
\]
For trainable scalar coefficients \(w_p\), the bilinear map is
\begin{equation}
B(x,y)_o
=
\sum_{p=1}^{K_{\nu;\lambda\mu}}
w_p
\sum_{i=1}^{d_\lambda}
\sum_{j=1}^{d_\mu}
(C_p)_{oij}
x_i y_j.
\label{eq:learned-tp}
\end{equation}
The tensors \(C_p\) are fixed by group symmetry, while the coefficients
\(w_p\) are learned or supplied externally.

\subsubsection{Generic irrep tensor product}

When the left, right, and output representations are non-regular irreps,
\  constructs the complete real coupling space
\[
    \operatorname{Hom}_{G}
    (V_\lambda\otimes V_\mu,V_\nu).
\]
View a rank-three tensor
\(T\in\mathbb R^{d_\nu\times d_\lambda\times d_\mu}\) as a matrix from
\(V_\lambda\otimes V_\mu\) to \(V_\nu\).  The tensor-product Reynolds
projection is
\begin{equation}
\Pi_{\mathrm{TP}}(T)
=
\frac{1}{|G|}
\sum_{g\in G}
\rho_\nu(g)\,
T\,
\bigl(\rho_\lambda(g)\otimes\rho_\mu(g)\bigr)^{\mathsf T}.
\label{eq:tp-reynolds}
\end{equation}
In coordinate form,
\[
\begin{split}
[\Pi_{\mathrm{TP}}(T)]_{oij}
={}&
\frac{1}{|G|}
\sum_{g\in G}
\sum_{o'=1}^{d_\nu}
\sum_{i'=1}^{d_\lambda}
\sum_{j'=1}^{d_\mu}
[\rho_\nu(g)]_{oo'}
[\rho_\lambda(g)]_{ii'}
[\rho_\mu(g)]_{jj'}
T_{o'i'j'}.
\end{split}
\]
The implementation applies this projection to the elementary rank-three
tensor basis, flattens the projected tensors, and uses an SVD to retain all
linearly independent directions.  The resulting tensors
\[
    C_p\in
    \mathbb R^{d_\nu\times d_\lambda\times d_\mu}
\]
are orthonormal in the entrywise Frobenius inner product and span the complete
real equivariant coupling space.

For real irreps, the number
\(K_{\nu;\lambda\mu}\) is the real Hom-space dimension, not necessarily the
number of copies of \(V_\nu\) in the tensor-product decomposition.  More
precisely, for an irreducible output,
\[
K_{\nu;\lambda\mu}
=
N_{\nu;\lambda\mu}
\dim_{\mathbb R}\operatorname{End}_{G}(V_\nu),
\]
where \(N_{\nu;\lambda\mu}\) is the representation-copy multiplicity of
\(V_\nu\) inside \(V_\lambda\otimes V_\mu\).  For a two-dimensional
complex-type real irrep of \(C_n\), the endomorphism factor is two.  Therefore
one representation copy gives two independent real neural-network coupling
weights.  The tensor-product modules use this complete real basis.  

Now suppose the left input contains \(U\) copies of \(V_\lambda\), the right
input contains \(V\) copies of \(V_\mu\), and the output contains \(M\) copies
of \(V_\nu\):
\[
    V_\lambda^{\oplus U},
    \qquad
    V_\mu^{\oplus V},
    \qquad
    V_\nu^{\oplus M}.
\]
For the default fully connected connection mode \texttt{uvw}, the layer learns
scalars
\[
    w_{muvp},
\]
where
\[
m=1,\ldots,M,\quad
u=1,\ldots,U,\quad
v=1,\ldots,V,\quad
p=1,\ldots,K_{\nu;\lambda\mu}.
\]
It computes
\begin{equation}
z_{m,o}
=
\sum_{u=1}^{U}
\sum_{v=1}^{V}
\sum_{p=1}^{K_{\nu;\lambda\mu}}
w_{muvp}
\sum_{i=1}^{d_\lambda}
\sum_{j=1}^{d_\mu}
(C_p)_{oij}
x_{u,i}
y_{v,j}.
\label{eq:tp-uvw}
\end{equation}
Here \(x_{u,i}\) is coordinate \(i\) of left copy \(u\),
\(y_{v,j}\) is coordinate \(j\) of right copy \(v\), and \(z_{m,o}\) is
coordinate \(o\) of output copy \(m\).  The trainable tensor for this
representation triple has shape
\[
    M\times U\times V\times K_{\nu;\lambda\mu}.
\]

The connection mode \texttt{uvu} identifies the output multiplicity index
with the left-input multiplicity index.  It therefore requires \(M=U\) for
the selected representation block and learns \(w_{uvp}\), with shape
\[
    U\times V\times K_{\nu;\lambda\mu}.
\]
The operation is
\begin{equation}
z_{u,o}
=
\sum_{v=1}^{V}
\sum_{p=1}^{K_{\nu;\lambda\mu}}
w_{uvp}
\sum_{i=1}^{d_\lambda}
\sum_{j=1}^{d_\mu}
(C_p)_{oij}
x_{u,i}
y_{v,j}.
\label{eq:tp-uvu}
\end{equation}
There is no sum over \(u\): output copy \(u\) is connected only to left-input
copy \(u\), while all right-input copies \(v\) may still be mixed.  In the
automatic compiler, the pairing is by occurrence order within the relevant multiplicity chunks.
\subsection{Tensor products involving regular representations}

When a regular representation occurs among the input or output
representations, \texttt{we3nn} uses analytic orbit constructions instead of
forming the potentially large generic rank-three coupling space.  If the
output is regular, the regular-output construction is selected even when one
or both inputs are also regular.  Otherwise, a regular left or right input is
handled by the corresponding analytic input construction.

\subsubsection{Arbitrary representations to regular output}

Consider a single-copy bilinear map
\[
    B:
    V_\lambda\times V_\mu
    \longrightarrow
    \mathbb R[G].
\]
Let
\[
    A\in
    \mathbb R^{d_\lambda\times d_\mu}
\]
be an arbitrary learned coefficient matrix.  Its entry \(A_{ij}\) couples
coordinate \(i\) of \(V_\lambda\) to coordinate \(j\) of \(V_\mu\) in the
reference frame.  For each \(q\in G\), define the inverse-transformed input
coordinates
\[
    x^{(q)}
    =
    \rho_\lambda(q)^{\mathsf T}x
    =
    \rho_\lambda(q^{-1})x,
    \qquad
    y^{(q)}
    =
    \rho_\mu(q)^{\mathsf T}y
    =
    \rho_\mu(q^{-1})y.
\]
The regular output \(z=\sum_{q\in G}z_q\delta_q\) is
\begin{equation}
z_q
=
\frac{1}{\sqrt{|G|}}
\sum_{i=1}^{d_\lambda}
\sum_{j=1}^{d_\mu}
A_{ij}
\left[
\rho_\lambda(q)^{\mathsf T}x
\right]_i
\left[
\rho_\mu(q)^{\mathsf T}y
\right]_j.
\label{eq:tp-output-reg}
\end{equation}
Thus the coordinate \(z_q\) stores the same learned reference-frame bilinear
form evaluated after both inputs have been moved to the frame associated with
\(q\).  The map is completely determined by this arbitrary bilinear form at
one reference group coordinate, and
\begin{equation}
\dim_{\mathbb R}
\operatorname{Hom}_{G}
(V_\lambda\otimes V_\mu,\mathbb R[G])
=
d_\lambda d_\mu.
\end{equation}
Accordingly, the matrix entries \(A_{ij}\) are a reshaping of the
\(K=d_\lambda d_\mu\) independent real coupling coefficients.

With \(U\), \(V\), and \(M\) left, right, and regular-output copies,
respectively, the \texttt{uvw} coefficient tensor can be written
\(A_{muvij}\) and has shape
\[
    M\times U\times V\times d_\lambda\times d_\mu.
\]
The implementation may store the last two axes as the analytic coupling shape
rather than flattening them to a single index \(p\).

\subsubsection{Regular left input}

Consider
\[
    B:
    \mathbb R[G]\times V_\mu
    \longrightarrow
    V_\nu.
\]
Write the regular input as
\(f=\sum_{q\in G}f_q\delta_q\), let \(y\in V_\mu\), and let
\[
    A:V_\mu\longrightarrow V_\nu,
    \qquad
    A\in\mathbb R^{d_\nu\times d_\mu},
\]
be an arbitrary learned reference-frame linear map.  The equivariant
bilinear operation is
\begin{equation}
z
=
\frac{1}{\sqrt{|G|}}
\sum_{q\in G}
f_q\,
\rho_\nu(q)
A
\rho_\mu(q)^{\mathsf T}y.
\label{eq:left-reg-tp}
\end{equation}
In coordinates, this is
\[
z_o
=
\frac{1}{\sqrt{|G|}}
\sum_{q\in G}
\sum_{i=1}^{d_\nu}
\sum_{j=1}^{d_\mu}
f_q
[\rho_\nu(q)]_{oi}
A_{ij}
[\rho_\mu(q)^{\mathsf T}y]_j.
\]
The regular scalar \(f_q\) selects the contribution from group coordinate
\(q\). The non-regular input is transformed into the corresponding frame,
the reference map \(A\) is applied, and the result is transformed into the
output frame.  The coupling-space dimension is
\begin{equation}
\dim_{\mathbb R}
\operatorname{Hom}_{G}
(\mathbb R[G]\otimes V_\mu,V_\nu)
=
d_\nu d_\mu.
\end{equation}

\subsubsection{Regular right input}

The right-regular case is analogous.  For
\[
    B:
    V_\lambda\times\mathbb R[G]
    \longrightarrow
    V_\nu,
\]
write the regular right input as
\(f=\sum_{q\in G}f_q\delta_q\), let \(x\in V_\lambda\), and let
\[
    A:V_\lambda\longrightarrow V_\nu,
    \qquad
    A\in\mathbb R^{d_\nu\times d_\lambda}.
\]
Then
\begin{equation}
z
=
\frac{1}{\sqrt{|G|}}
\sum_{q\in G}
f_q\,
\rho_\nu(q)
A
\rho_\lambda(q)^{\mathsf T}x,
\label{eq:right-reg-tp}
\end{equation}
or, in components,
\[
z_o
=
\frac{1}{\sqrt{|G|}}
\sum_{q\in G}
\sum_{i=1}^{d_\nu}
\sum_{j=1}^{d_\lambda}
f_q
[\rho_\nu(q)]_{oi}
A_{ij}
[\rho_\lambda(q)^{\mathsf T}x]_j.
\]
The coupling-space dimension is
\begin{equation}
\dim_{\mathbb R}
\operatorname{Hom}_{G}
(V_\lambda\otimes\mathbb R[G],V_\nu)
=
d_\nu d_\lambda.
\end{equation}

\subsubsection{Regular--regular to regular tensor product}

An important special case is
\[
    \mathbb R[G]\otimes\mathbb R[G]
    \longrightarrow
    \mathbb R[G].
\]
Since the output is regular, this case is an instance of
Equation~\eqref{eq:tp-output-reg}.  For a single left input copy, single right
input copy, and single output copy, the learned reference bilinear form is
\[
    A\in\mathbb R^{|G|\times |G|}
\]
and contains \(|G|^2\) independent coefficients.

Let \(x_a\) and \(y_b\) be the input coordinates associated with
\(a,b\in G\).  Then, for every output group coordinate \(q\in G\),
\begin{equation}
z_q
=
\frac{1}{\sqrt{|G|}}
\sum_{a\in G}
\sum_{b\in G}
A_{ab}
x_{qa}
y_{qb},
\label{eq:reg-reg-reg-tp}
\end{equation}
where \(qa\) and \(qb\) denote group multiplication.  Thus the same learned
interaction between relative group coordinates \(a\) and \(b\) is evaluated
around every output coordinate \(q\).

For the cyclic group \(C_n\), identify group elements with integers modulo
\(n\).  Group multiplication then becomes addition modulo \(n\), and
\begin{equation}
z_k
=
\frac{1}{\sqrt n}
\sum_{a=0}^{n-1}
\sum_{b=0}^{n-1}
A_{ab}
x_{k+a}
y_{k+b},
\end{equation}
where \(k+a\) and \(k+b\) are reduced modulo \(n\).

Coordinatewise multiplication,
\[
    z_k=x_k y_k,
\]
is contained in this family.  Under the orthonormal
\(1/\sqrt n\) convention used above, it is obtained by setting
\[
    A_{00}=\sqrt n
    \qquad\text{and}\qquad
    A_{ab}=0\ \text{for }(a,b)\ne(0,0).
\]
If \(A_{00}=1\), the result is instead
\(z_k=n^{-1/2}x_k y_k\).  The normalization does not affect expressivity
because it can be absorbed into the learned coefficient.  The full tensor
product is more expressive than coordinatewise multiplication because it can
couple arbitrary pairs of relative group orientations.

\subsection{Internal and externally supplied tensor-product weights}
\label{sec:we3nn-tp-weights}

The coupling tensors define the symmetry-allowed bilinear basis, but their
scalar coefficients may be supplied in two ways.

With internal weights, the tensor-product module owns trainable parameters that are shared across all samples. Different representation paths and
multiplicity combinations receive independent learned coefficients, but those
coefficients do not depend on the numerical value of a particular input
sample.

Alternatively, the tensor product can receive externally generated weights, for example, an invariant contextual feature. For geometric neural networks, a radial MLP may generate tensor-product coefficients from
\(\lVert r_{ij}\rVert\). Since the supplied coefficients are invariant
scalars, input-dependent weighting does not alter the equivariance of the
coupling basis.

\subsection{Bravais lattice}

A two-dimensional Bravais lattice is generated by two linearly independent primitive lattice vectors ${a}_1$ and ${a}_2$:
\begin{equation}
\Lambda = \left\{ n_1{a}_1 + n_2{a}_2 \;:\; n_1,n_2 \in \mathbb{Z} \right\}.
\label{bravais_lattice}
\end{equation}
\subsection{Voronoi boundaries}
\label{voronoi_boundaries}
The minimizer is unique except on the Voronoi boundaries of
\(\mathcal A\), which form a measure-zero subset of the plane.

{\secondreviewercolor
The hard nearest-anchor map is nevertheless discontinuous at these boundaries:
crossing a boundary changes the selected anchor, and an exact tie requires a
choice between two or more equally near anchors. A point sampled from a
continuous distribution is unlikely to lie exactly on a boundary, but
optimization trajectories can pass arbitrarily close to one, and
symmetry-constrained coordinates can produce exact ties. Moreover, an arbitrary
deterministic tie-breaking order need not be equivariant.

\paragraph{Boundary-swap sensitivity.}
To measure the resulting sensitivity of trained models, we performed a
controlled experiment on 100 randomly selected test structures from the
molecule--graphene dataset. In each structure, the molecular atom closest to
the Voronoi boundary between its nearest and second-nearest anchors was moved
exactly onto that boundary. We then evaluated the same boundary geometry twice,
assigning this atom first to its original anchor and then to the other tied
anchor. Atomic coordinates, graph edges, all other anchor assignments, and all
other WA vectors were identical in the two evaluations. This construction
isolates the effect of the discrete anchor reassignment from geometric changes.

Table~\ref{tab:wa-boundary-swap} reports the absolute difference between the
two predictions, as a mean and standard deviation over three trained seeds.
The energy difference is normalized by the number of full-cell atoms, force
metrics are evaluated over all molecular atoms in the 100 structures. As in
the main experiments, C denotes force-weighted training with energy-derived
forces and V denotes direct vector-head training.

\begin{table}[!t]
    \secondreviewercolor
    \centering
    \caption{Sensitivity to exchanging the two tied nearest-anchor assignments
    at a Voronoi boundary. Energy differences are in
    \(\mathrm{meV\,atom^{-1}}\), force-component differences in
    \(\mathrm{meV\,\mathring{A}^{-1}}\), and relative force \(L_2\) in percent.
    Values are mean \(\pm\) standard deviation over three seeds.}
    \label{tab:wa-boundary-swap}
    \small
    \resizebox{\textwidth}{!}{%
    \begin{tabular}{lccccc}
        \toprule
        Model &
        \shortstack{Energy swap\\MAE} &
        \shortstack{Relative force\\\(L_2\), C} &
        \shortstack{Relative force\\\(L_2\), V} &
        \shortstack{Force-component\\swap MAE, C} &
        \shortstack{Force-component\\swap MAE, V} \\
        \midrule
        WP-Inv
            & \(0.353\pm0.033\) & \(7.57\pm1.26\) & \(3.25\pm0.59\)
            & \(3.61\pm0.88\) & \(2.29\pm0.31\) \\
        WP-InvEq
            & \(\mathbf{0.247\pm0.110}\) & \(6.15\pm0.48\) & \(3.73\pm0.27\)
            & \(2.99\pm0.21\) & \(3.48\pm0.03\) \\
        WP-Equi
            & \(0.342\pm0.031\) & \(7.02\pm2.50\) & \(4.58\pm0.35\)
            & \(4.29\pm0.94\) & \(3.78\pm0.38\) \\
        WP-VecInv
            & \(0.653\pm0.090\) & \(9.74\pm1.06\) & \(6.96\pm0.19\)
            & \(9.86\pm0.82\) & \(9.20\pm0.19\) \\
        WP-Reg
            & \(0.454\pm0.091\) & \(1.33\pm0.25\) & \(2.05\pm0.18\)
            & \(1.84\pm0.50\) & \(2.00\pm0.08\) \\
        WP-RegTP
            & \(0.492\pm0.065\) & \(3.28\pm0.17\) & \(2.66\pm0.13\)
            & \(4.41\pm0.17\) & \(2.72\pm0.15\) \\
        WP-O3
            & \(0.426\pm0.078\) & \(\mathbf{1.24\pm0.58}\) & \(\mathbf{0.98\pm0.53}\)
            & \(\mathbf{1.48\pm0.33}\) & \(\mathbf{1.49\pm0.77}\) \\
        \bottomrule
    \end{tabular}%
    }
\end{table}

Relative to each model's reported test MAE, the force-component changes are
2.2--9.0\% in conservative mode and 3.1--20.5\% in direct-head mode. The
per-atom energy changes are 5.2--29.1\% of the corresponding test energy MAE.
Thus the discontinuity is measurable but smaller than the ordinary prediction
error for every tested model. WP-O3 is least sensitive in the force metrics,
while WP-InvEq has the smallest energy change. Models with larger test-force
errors also tend to exhibit larger force changes under reassignment in the
conservative setting, although this trend is not universal.

This experiment measures the finite prediction jump for representative trained
models, it does not make the hard assignment continuous. Randomizing between
tied assignments for boundary-adjacent atoms during training, or replacing the
hard nearest-anchor choice by a symmetry-compatible soft assignment, are
possible ways to regularize this sensitivity and are left for future work.
}

\subsection{WA covariance proof}
\label{sec:wa-covariance-proof}
\begin{proof}
Because $g$ is an isometry,
\begin{equation}
    \|ga-gx\|=\|a-x\|.
\end{equation}
Since $g$ is a bijection of $\mathcal A$, the unique nearest anchor to $gx$ is
$ga(x)$. Therefore
\begin{align}
    v_{\WA}(gx)
    &=a(gx)-gx\\
    &=ga(x)-gx\\
    &=(Q_ga(x)+\tau_g)-(Q_gx+\tau_g)\\
    &=Q_g(a(x)-x)\\
    &=Q_gv_{\WA}(x).
\end{align}
The translational part $\tau_g$, including the fractional translation in a
glide reflection, cancels exactly.
\end{proof}

\subsection{Translation-suborbit regular-representation implementation details}
\label{translation-suborbit-regular-representation}

Recall from Equation~\eqref{eq:joint-orbit-lift} that the raw tensor
\(\bar X_i=(\bar X_{i,p})_{p\in P}\) is obtained by stacking the blocks over
\(p\in P\). Under a wallpaper operation \(g\in G\),
the vector and member label transform as
\begin{equation}
    v_{\pi_g(i)}(gX)=Q_gv_i(X),
    \qquad
    \ell_{\pi_g(i)}(gX)
    =\rho_{\mathrm{mem}}(p_g)\ell_i(X),
    \qquad
    p_g=g\Lambda\in G/\Lambda\simeq P.
\end{equation}
For \(g(x)=Q_gx+\tau_g\), every lattice translation \(t\in\Lambda\) has
identity linear part \(Q_t=I\). Hence the linear-part homomorphism
\(g\mapsto Q_g\) is constant on cosets of \(\Lambda\) and descends to the
well-defined point-group representation \(Q:P\to O(2)\). In particular,
\(Q_{p_g}=Q_g\).
Here, \(p\) is the free index that ranges over all rows of the lifted field,
whereas \(p_g\) is fixed by the applied operation \(g\). Thus, \(Q_p\) is the
matrix used to construct row \(p\), while \(Q_{p_g}=Q_g\) is the matrix that
transforms the physical WA vector under \(g\), both are values of the same
representation \(Q\).

Because \(Q\) and \(\rho_{\mathrm{mem}}\) are representations of \(P\),
\(Q_pQ_{p_g}=Q_{pp_g}\) and
\(\rho_{\mathrm{mem}}(p)\rho_{\mathrm{mem}}(p_g)
=\rho_{\mathrm{mem}}(pp_g)\). Substitution into the three blocks of the lift gives
\begin{align}
    \bar X_{\pi_g(i),p}(gX)
    &={}
    \left[
        Q_pQ_{p_g}v_i,
        \rho_{\mathrm{mem}}(p)\rho_{\mathrm{mem}}(p_g)\ell_i,
        \bigl(\rho_{\mathrm{mem}}(p)\rho_{\mathrm{mem}}(p_g)\ell_i\bigr)
        \otimes(Q_pQ_{p_g}v_i)
    \right]
    \\
    &=\bar X_{i,pp_g}(X).
    \label{eq:raw-right-regular-lift}
\end{align}
In other words, row \(p\) after transforming the input contains exactly the
features previously stored in row \(pp_g\). The transformation therefore
permutes the group-indexed rows by \(p\mapsto pp_g\), which is right
multiplication by \(p_g\).

\paragraph{Dimensions of the lift.}
For a single selected WA family with \(m\) member channels, each group block
of the lift in Equation~\eqref{eq:joint-orbit-lift} contains two vector
coordinates, \(m\) member coordinates, and \(2m\) member-gated vector
coordinates. Hence
\begin{equation}
    C_{\mathrm{lift}}
    =
    2+m+2m
    =
    2+3m.
\end{equation}
For the selected nonsymmorphic anchor orbits, \(m=2\), and therefore
\begin{equation}
    C_{\mathrm{lift}}=8.
\end{equation}

\paragraph{Conversion to the left-regular convention.}
The regular representation used by \texttt{we3nn} follows the left-regular
convention \(\rho_{\mathrm{reg}}=\lambda\), where
\begin{equation}
    [\lambda(p_g)u](k)=u(p_g^{-1}k).
\end{equation}
We therefore apply the fixed inversion reindexing
\begin{equation}
    X^{\mathrm{reg}}_{i,k}
    =
    \bar X_{i,k^{-1}},
    \qquad k\in P.
    \label{eq:regular-inversion-reindexing}
\end{equation}
Using Equation~\eqref{eq:raw-right-regular-lift}, the transformed field obeys
\begin{align}
    X^{\mathrm{reg}}_{\pi_g(i),k}(gX)
    &=
    \bar X_{\pi_g(i),k^{-1}}(gX)
    \\
    &=
    \bar X_{i,k^{-1}p_g}(X)
    \\
    &=
    X^{\mathrm{reg}}_{i,p_g^{-1}k}(X).
    \label{eq:left-regular-lift}
\end{align}
Hence, inversion of the group index converts the deterministic orbit lift
from the right-regular convention to the left-regular convention used by
\texttt{we3nn}, so that \(X_i^{\mathrm{reg}}\) can be passed directly to
regular-representation layers. 

\paragraph{Layerwise regular encoder.}
For \(c\) feature channels, let
\(\mathcal R_c=(\mathbb R^{P})^c\) denote \(c\) copies of the regular
representation, with coordinates \(u_{a,k}\) for channel \(a\) and group
index \(k\). Thus \(X_i^{\mathrm{reg}}\in\mathcal R_{C_{\mathrm{lift}}}\).
Let the encoder have \(R\) layers and channel multiplicities
\(c_0=C_{\mathrm{lift}},c_1,\ldots,c_R=d_{\mathrm{reg}}\). At layer \(r\),
the linear map is an intertwiner
\begin{equation}
    W^{(r)}
    \in
    \operatorname{Hom}_P
    \left(\mathcal R_{c_r},\mathcal R_{c_{r+1}}\right),
    \qquad r=0,\ldots,R-1,
\end{equation}
and the bias \(b^{(r)}\) lies in the fixed subspace of
\(\mathcal R_{c_{r+1}}\), meaning that it is constant over the group index.
The same scalar nonlinearity \(\sigma_r\) is applied to every channel and
group index. Since the regular action only permutes the group index,
\begin{equation}
    \left[\sigma_r\!\left((I_c\otimes\rho_{\mathrm{reg}}(p))u\right)\right]_{a,k}
    =
    \sigma_r\!\left(u_{a,p^{-1}k}\right)
    =
    \left[(I_c\otimes\rho_{\mathrm{reg}}(p))\sigma_r(u)\right]_{a,k}.
\end{equation}
Thus each update
\begin{equation}
    u^{(r+1)}
    =
    \sigma_r\!\left(W^{(r)}u^{(r)}+b^{(r)}\right),
    \qquad
    u^{(0)}=X_i^{\mathrm{reg}},
\end{equation}
is equivariant, and their composition
\begin{equation}
    Z_i
    =
    f_{\mathrm{reg}}\!\left(X_i^{\mathrm{reg}}\right)
    \in \mathcal R_{d_{\mathrm{reg}}}
\end{equation}
satisfies
\begin{equation}
    Z_{\pi_g(i),k}(gX)
    =
    Z_{i,p_g^{-1}k}(X).
\end{equation}

\paragraph{Equivariant use of the registry feature.}
When the group axis of \(Z_i\) is retained, the registry feature keeps both its
trivial component and the nontrivial components that encode how the WA geometry
is associated with the permuting translation-suborbit member. When an invariant
architecture is required, averaging over the group axis gives
\begin{equation}
    L_i=\frac{1}{|P|}\sum_{k\in P}Z_{i,k}.
\end{equation}
The map \(k\mapsto p_g^{-1}k\) is a permutation of \(P\), so
\(L_{\pi_g(i)}(gX)=L_i(X)\).

\subsection{Action on translation-suborbit members}
\label{sec:translation-suborbit-action}
The decomposition
\begin{equation}
\mathcal A
=
\coprod_{\alpha=1}^{m}
\left(q_{\alpha}+\Lambda\right)
\end{equation}
is chosen such that the representatives \(q_{\alpha}\) are distinct modulo
the Bravais lattice. Every wallpaper operation maps each translation suborbit
to another translation suborbit in the same anchor family.
Indeed, let \(g(x)=Q_gx+\tau_g\). Since the linear part of a wallpaper
operation preserves the Bravais lattice,

\begin{equation}
Q_g \Lambda = \Lambda.
\end{equation}
It follows that
\begin{align}
g\bigl(q_{\alpha}+\Lambda\bigr)
&=
\left\{
g\bigl(q_{\alpha}+t\bigr)
\,\middle|\,
t\in\Lambda
\right\}
\\
&=
\left\{
gq_{\alpha}+Q_g t
\,\middle|\,
t\in\Lambda
\right\}
\\
&=
gq_{\alpha}+\Lambda.
\end{align}

Because \(\mathcal{A}\) is \(G\)-invariant,
\(gq_{\alpha}\in\mathcal{A}\). Hence, there exists a unique index
\(\beta\in\{1,\ldots,m\}\) such that
\begin{equation}
gq_{\alpha}
\equiv
q_{\beta}
\pmod{\Lambda}.
\end{equation}
We write \(\sigma_g(\alpha)=\beta\). Since \(g\) is invertible,
\(\sigma_g\) is a permutation of the translation-suborbit indices, and
\begin{equation}
\sigma_{g_1g_2}
=
\sigma_{g_1}\circ\sigma_{g_2}.
\end{equation}
A lattice translation \(t\in\Lambda\) maps every suborbit
\(q_\alpha+\Lambda\) to itself, so \(\sigma_t\) is the identity and
\(\sigma_g\) depends only on the coset \(p_g=g\Lambda\). The permutation
matrices of \(\sigma_g\) are therefore exactly the member permutation
representation of Equation~\eqref{eq:member-representation},
\begin{equation}
\rho_{\mathrm{mem}}(p_g)e_\alpha
=
e_{\sigma_g(\alpha)}.
\end{equation}

If \(\alpha(x)\) denotes the translation-suborbit member of the unique nearest
anchor to \(x\), nearest-anchor equivariance \(a(gx)=ga(x)\) implies
\begin{equation}
\alpha(gx)
=
\sigma_g\bigl(\alpha(x)\bigr).
\end{equation}
Consequently, the member one-hot vector
\(\ell(x)=\operatorname{onehot}(\alpha(x))\) transforms according to
\begin{equation}
\ell(gx)
=
\rho_{\mathrm{mem}}(p_g)\ell(x).
\end{equation}
For node \(i\), with \(\ell_i=\ell(x_{i,xy})\) and
\(v_i=v_{\WA}(x_{i,xy})\), this gives
\(\ell_{\pi_g(i)}(gX)=\rho_{\mathrm{mem}}(p_g)\ell_i(X)\), and
Equation~\eqref{eq:wa-covariance} gives
\(v_{\pi_g(i)}(gX)=Q_gv_i(X)\).
Thus, the member label is not generally an invariant scalar feature. Rather, it
transforms under the permutation representation induced by the wallpaper-group
action on the translation suborbits.

\subsection{Selected anchor orbits}
\label{sec:selected-anchor-orbits}

For each symmorphic wallpaper group, we choose a point whose site-symmetry
point group equals the full point group. Such an orbit has one
translation-suborbit member per primitive cell, $m=1$. For the four
nonsymmorphic wallpaper groups $pg$, $pmg$, $pgg$, and $p4g$, the selected
minimum-multiplicity orbit has $m=2$ in the implementation described in
Section~\ref{sec:nonsymmorphic}. Complete visualizations of the group actions
and the associated WA-vector transformations are provided in
Appendix~\ref{sec:wallpaper-action-visualizations}.

Table~\ref{tab:selected-wa-orbits} records the fixed anchor-orbit choices used
by the simulator. Coordinates are fractional coordinates in the basis and
origin convention used by the implementation. Labels such as C2-A, C3-A, and
C4-A are implementation labels rather than universal Wyckoff symbols.

The column ``N of anchor-orbit
choices'' reports how many inequivalent anchor orbits have the same highest site-symmetry type as the
chosen anchor. For Wyckoff positions with free coordinates, continuously many
such orbits exist. In those cases, the table gives the number and dimension of
the corresponding Wyckoff families. These choices are not used simultaneously.
The final column states whether the wallpaper group is symmorphic or
nonsymmorphic.

\begingroup
\renewcommand{\arraystretch}{1.15}
\setlength{\tabcolsep}{3pt}
\small
\begin{longtable}{@{}>{\raggedright\arraybackslash}p{1.0cm}
                      >{\raggedright\arraybackslash}p{1.75cm}
                      >{\raggedright\arraybackslash}p{4.8cm}
                      >{\centering\arraybackslash}p{0.55cm}
                      >{\centering\arraybackslash}p{2.7cm}
                      >{\centering\arraybackslash}p{2.1cm}@{}}
\caption{Selected Wyckoff-anchor orbits. The member multiplicity \(m\) is
counted modulo primitive translations. The fifth column reports the number of
inequivalent anchor-orbit choices. The subscript prim marks representatives
given modulo primitive translations: the conventional centered cells of $cm$ and
$cmm$ contain a centering-related copy that is equivalent modulo primitive
translations.}
\label{tab:selected-wa-orbits}\\
\toprule
Group & Selected anchor & Representatives \(A^0\) & \(m\) & N of anchor-orbit choices & Symmorphicity\\
\midrule
\endfirsthead
\toprule
Group & Selected anchor & Representatives \(A^0\) & \(m\) & N of anchor-orbit choices & Symmorphicity\\
\midrule
\endhead
\bottomrule
\endfoot

\(p1\)
& lattice anchor
& \(\{q_1=(0,0)\}\)
& 1 & continuum: one two-parameter family & symm.\\

\(p2\)
& C2-A
& \(\{q_1=(\tfrac12,\tfrac12)\}\)
& 1 & 4 discrete C2 orbits & symm.\\

\(pm\)
& lattice/\newline mirror anchor
& \(\{q_1=(0,0)\}\)
& 1 & continuum: two one-parameter mirror families & symm.\\

\(pg\)
& minimum orbit
& \(\{q_1=(0,0),\;q_2=(0,\tfrac12)\}\)
& 2 & continuum: one two-parameter family & non-symm.\\

\(cm\)
& lattice/\newline mirror anchor
& \(\{q_1=(0,0)\}_{\mathrm{prim}}\)
& 1 & continuum: one one-parameter mirror family & symm.\\

\(pmm\)
& C2-A
& \(\{q_1=(\tfrac12,\tfrac12)\}\)
& 1 & 4 discrete C2v orbits & symm.\\

\(pmg\)
& C2-A
& \(\{q_1=(\tfrac12,\tfrac12),\;q_2=(0,\tfrac12)\}\)
& 2 & 2 discrete C2 orbits & non-symm.\\

\(pgg\)
& C2-A
& \(\{q_1=(0,0),\;q_2=(\tfrac12,\tfrac12)\}\)
& 2 & 2 discrete C2 orbits & non-symm.\\

\(cmm\)
& C2-A
& \(\{q_1=(0,0)\}_{\mathrm{prim}}\)
& 1 & 2 discrete C2v orbits & symm.\\

\(p4\)
& C4-A
& \(\{q_1=(\tfrac12,\tfrac12)\}\)
& 1 & 2 discrete C4 orbits & symm.\\

\(p4m\)
& C4-A
& \(\{q_1=(\tfrac12,\tfrac12)\}\)
& 1 & 2 discrete C4v orbits & symm.\\

\(p4g\)
& C4
& \(\{q_1=(\tfrac12,\tfrac12),\;q_2=(0,0)\}\)
& 2 & 1 discrete C4 orbit & non-symm.\\

\(p3\)
& C3-A
& \(\{q_1=(\tfrac13,\tfrac23)\}\)
& 1 & 3 discrete C3 orbits & symm.\\

\(p3m1\)
& C3-A
& \(\{q_1=(\tfrac13,\tfrac23)\}\)
& 1 & 3 discrete C3v orbits & symm.\\

\(p31m\)
& C3-B
& \(\{q_1=(0,0)\}\)
& 1 & 1 discrete C3v orbit & symm.\\

\(p6\)
& C6
& \(\{q_1=(0,0)\}\)
& 1 & 1 discrete C6 orbit & symm.\\

\(p6m\)
& C6
& \(\{q_1=(0,0)\}\)
& 1 & 1 discrete C6v orbit & symm.\\

\end{longtable}
\endgroup

The number of anchor-orbit choices in the fifth column is tied to the highest site-symmetry type selected
for the anchor. It does not count orbits of the same multiplicity with a different site-symmetry type. The table records the fixed
implementation choice of the WallpaperNet framework, rather than every possible crystallographic origin convention.

\subsection{Proof of full wallpaper equivariance}
\label{sec:full-proof-appendix}

We prove Theorem~\ref{thm:wallpaper-equivariance} in the notation of
Section~\ref{sec:full-proof}.

\begin{proof}
For message-passing layer \(r\), let
\(z_{ij}^{(r)}\in V_{\mathrm{in},r}\) denote the complete
representation-valued input on the oriented edge \(j\to i\), and let
\(m_{ij}^{(r)}\in V_{m,r}\) denote the resulting message. Their
point-group representations are \(\rho_{\mathrm{in},r}\) and
\(\rho_{m,r}\), respectively. Write the message map as
\begin{equation}
    M_r:V_{\mathrm{in},r}\longrightarrow V_{m,r},
    \qquad
    m_{ij}^{(r)}=M_r\!\left(z_{ij}^{(r)}\right),
\end{equation}
and assume that it is equivariant:
\begin{equation}
    M_r\!\left(\rho_{\mathrm{in},r}(p)z\right)
    =
    \rho_{m,r}(p)M_r(z),
    \qquad p\in P.
\end{equation}
Let \(a_{ij}^{(r)}\) be an optional invariant scalar aggregation weight, and
define the aggregated message by
\begin{equation}
    \bar m_i^{(r)}
    =
    \sum_{j\in\mathcal N_X(i)}
    a_{ij}^{(r)}m_{ij}^{(r)}.
\end{equation}
Finally, let
\begin{equation}
    U_r:V_r\oplus V_{m,r}\longrightarrow V_{r+1},
    \qquad
    h_i^{(r+1)}
    =
    U_r\!\left(h_i^{(r)},\bar m_i^{(r)}\right),
\end{equation}
be an equivariant node update, meaning that
\begin{equation}
    U_r\!\left(\rho_r(p)h,\rho_{m,r}(p)\bar m\right)
    =
    \rho_{r+1}(p)U_r(h,\bar m),
    \qquad p\in P.
\end{equation}
These layer-indexed symbols are local to this proof and do not replace the
architecture-specific message notation used elsewhere in the appendix.

By assumption, initialization supplies the base case
\begin{equation}
    h_{\pi_g(i)}^{(0)}(gX)
    =
    \rho_0(p_g)h_i^{(0)}(X).
\end{equation}
For the induction step, suppose the incoming fields \(h_i^{(r)}\) satisfy their
assigned transformation laws.
Together with the geometric encodings, they form an edge input obeying
\begin{equation}
    z_{\pi_g(i)\pi_g(j)}^{(r)}(gX)
    =
    \rho_{\mathrm{in},r}(p_g)z_{ij}^{(r)}(X).
\end{equation}
Equivariance of the message function gives
\begin{equation}
    m_{\pi_g(i)\pi_g(j)}^{(r)}(gX)
    =
    \rho_{m,r}(p_g)m_{ij}^{(r)}(X).
\end{equation}

Because the graph construction is compatible with the wallpaper action,
the neighbor sets obey
\begin{equation}
    \mathcal N_{gX}\!\left(\pi_g(i)\right)
    =
    \left\{\pi_g(j):j\in\mathcal N_X(i)\right\}.
\end{equation}
Therefore, for
invariant scalar aggregation weights satisfying
\begin{equation}
    a_{\pi_g(i)\pi_g(j)}^{(r)}(gX)
    =
    a_{ij}^{(r)}(X),
\end{equation}
we may reindex the transformed aggregation to obtain
\begin{align}
    \bar m_{\pi_g(i)}^{(r)}(gX)
    &=
    \sum_{j\in\mathcal N_X(i)}
    a_{\pi_g(i)\pi_g(j)}^{(r)}(gX)\,
    m_{\pi_g(i)\pi_g(j)}^{(r)}(gX)
    \\
    &=
    \sum_{j\in\mathcal N_X(i)}
    a_{ij}^{(r)}(X)\,
    \rho_{m,r}(p_g)m_{ij}^{(r)}(X)
    \\
    &=
    \rho_{m,r}(p_g)
    \sum_{j\in\mathcal N_X(i)}
    a_{ij}^{(r)}(X)m_{ij}^{(r)}(X)
    \\
    &=
    \rho_{m,r}(p_g)\bar m_i^{(r)}(X).
\end{align}
Thus the aggregated message transforms in the representation
\(\rho_{m,r}\). Equivariance of the node update now gives
\begin{align}
    h_{\pi_g(i)}^{(r+1)}(gX)
    &=
    U_r\!\left(
        h_{\pi_g(i)}^{(r)}(gX),
        \bar m_{\pi_g(i)}^{(r)}(gX)
    \right)
    \\
    &=
    U_r\!\left(
        \rho_r(p_g)h_i^{(r)}(X),
        \rho_{m,r}(p_g)\bar m_i^{(r)}(X)
    \right)
    \\
    &=
    \rho_{r+1}(p_g)
    U_r\!\left(h_i^{(r)}(X),\bar m_i^{(r)}(X)\right)
    \\
    &=
    \rho_{r+1}(p_g)h_i^{(r+1)}(X).
\end{align}
Together with the base case, induction proves
Equation~\eqref{eq:all-field-wallpaper-equivariance} for every layer and
equivariant readout.
\end{proof}

In an
invariant-message model, hidden scalar fields carry the trivial
representation. Its planar scalar-weighted vector construction is
equivariant because scalar multiplication is the canonical intertwiner
\begin{equation}
    \mu\in\operatorname{Hom}_P
    \left(
        \mathbb R_{\mathrm{triv}}\otimes V_{\mathrm{vec}},
        V_{\mathrm{vec}}
    \right),
    \qquad
    \mu(a\otimes v)=av.
\end{equation}
Here \(\mathbb R_{\mathrm{triv}}\) is the one-dimensional trivial
representation and \(V_{\mathrm{vec}}=\mathbb R^2\) carries the standard
planar representation,
\begin{equation}
    \rho_{\mathrm{vec}}(p)=Q_p.
\end{equation}

\section{Wallpaper-group action visualizations}
\label{sec:wallpaper-action-visualizations}

Tables~\ref{tab:wa-actions-p1}--\ref{tab:wa-actions-p6m} visualize the
selected WA orbit and the action of the symmetry operations for all 17
wallpaper groups. Each table includes its own legend, and arrows show the
transformation of the molecular graph and the corresponding WA vectors.

\begingroup
\def\WallpaperTablesEmbedded{1}
\ifdefined\WallpaperTablesEmbedded
\else
\documentclass[10pt]{article}
\usepackage[margin=12mm]{geometry}
\usepackage{array}
\usepackage{graphicx}
\usepackage{longtable}
\usepackage{needspace}
\begin{document}
\fi
\setlength{\tabcolsep}{3pt}
\setlength{\LTcapwidth}{\linewidth}
\setlength{\LTpre}{6pt}
\setlength{\LTpost}{6pt}
\Needspace{200pt}
\begin{longtable}{@{}>{\centering\arraybackslash}p{0.475\linewidth}>{\centering\arraybackslash}p{0.475\linewidth}@{}}
\caption[Wallpaper group \texttt{p1}]{Wallpaper group \texttt{p1}, {\small WA is pointing to lattice anchor at $q_{1}=(0,0)$}}\label{tab:wa-actions-p1}\\[1mm]
\multicolumn{2}{c}{\includegraphics[width=0.96\linewidth]{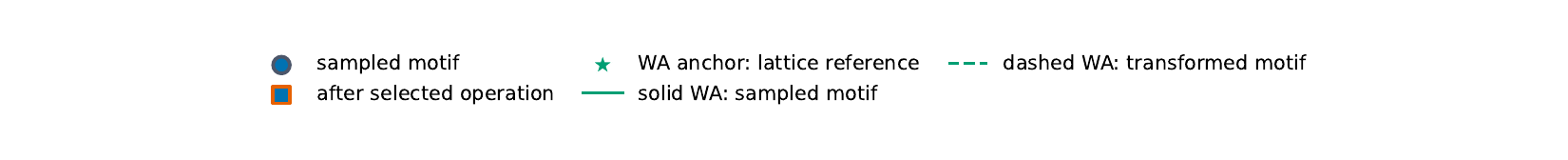}} \\[1mm]
\endfirsthead
\caption[]{Wallpaper group \texttt{p1} (continued)}\\[1mm]
\endhead
\includegraphics[width=\linewidth,height=0.22\textheight,keepaspectratio]{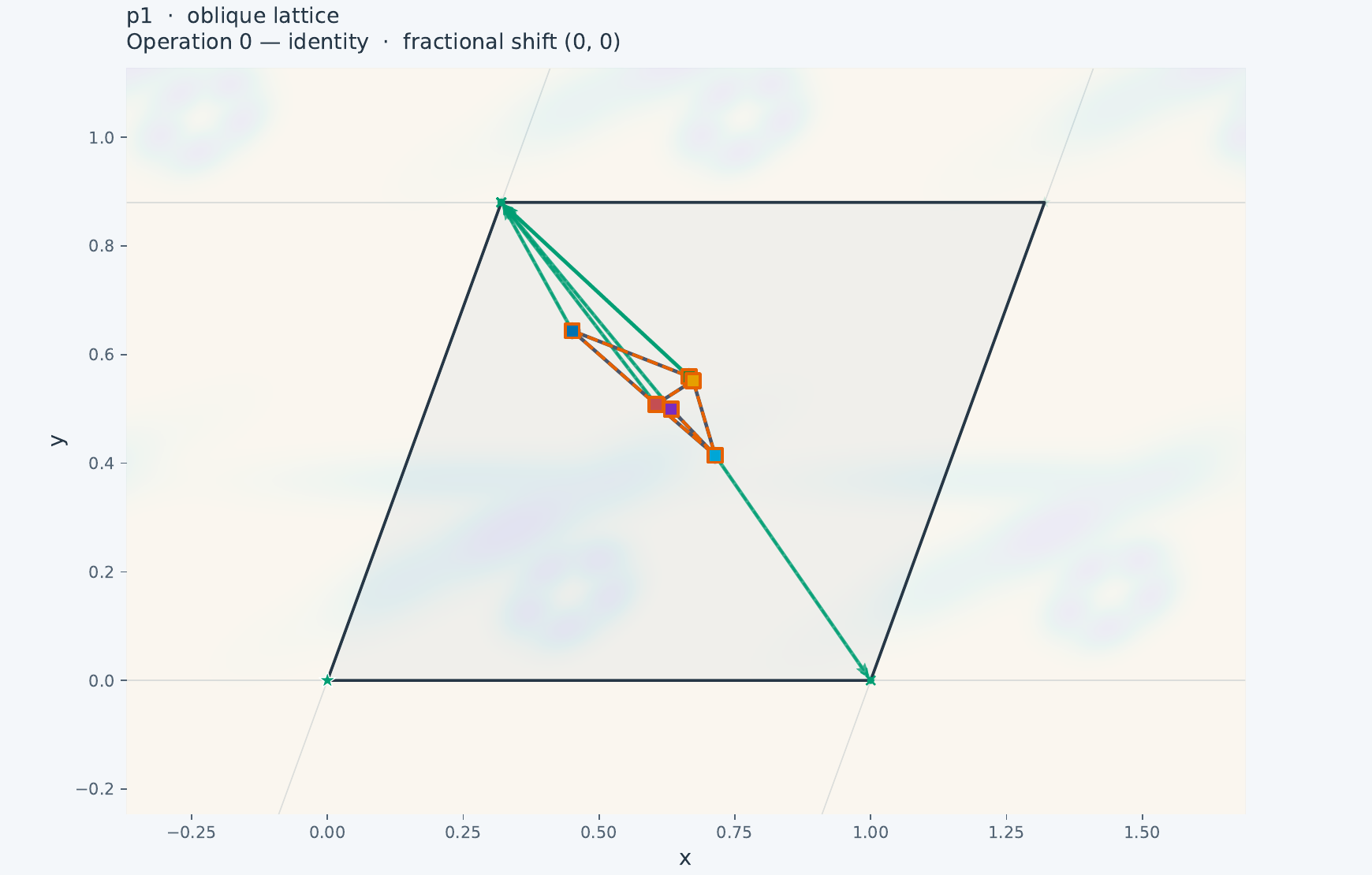} &  \\[2mm]
\end{longtable}
\Needspace{218pt}
\begin{longtable}{@{}>{\centering\arraybackslash}p{0.475\linewidth}>{\centering\arraybackslash}p{0.475\linewidth}@{}}
\caption[Wallpaper group \texttt{p2}]{Wallpaper group \texttt{p2}, {\small WA is pointing to C2-A at $q_{1}=(\frac{1}{2},\frac{1}{2})$}}\label{tab:wa-actions-p2}\\[1mm]
\multicolumn{2}{c}{\includegraphics[width=0.96\linewidth]{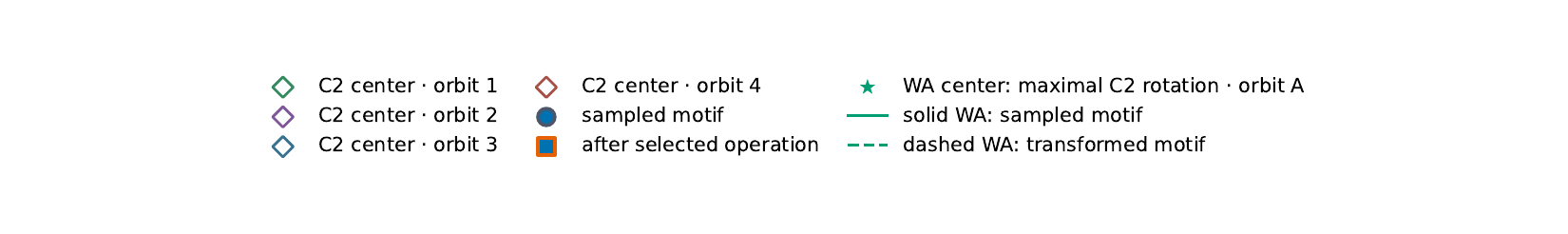}} \\[1mm]
\endfirsthead
\caption[]{Wallpaper group \texttt{p2} (continued)}\\[1mm]
\endhead
\includegraphics[width=\linewidth,height=0.22\textheight,keepaspectratio]{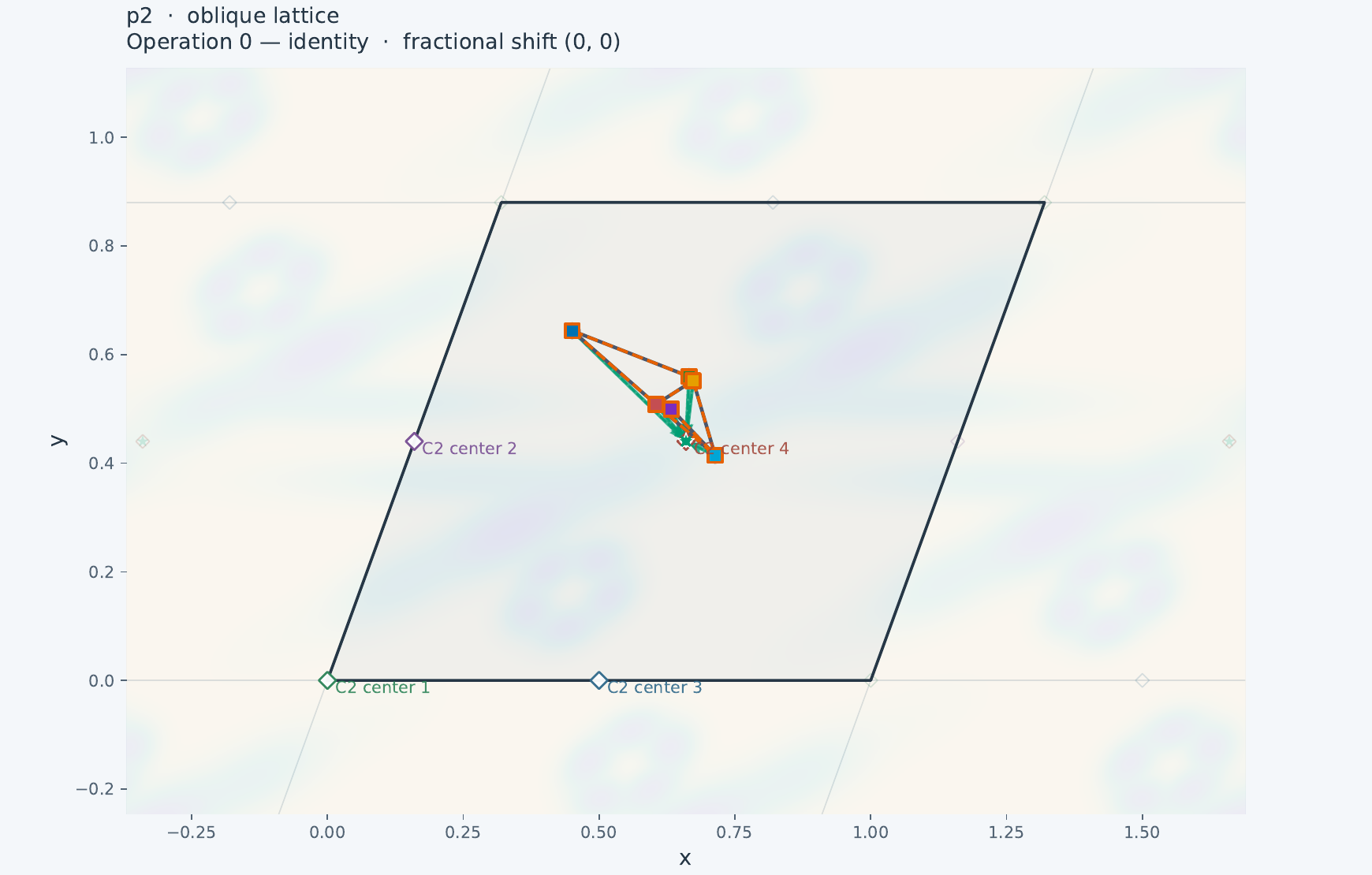} & \includegraphics[width=\linewidth,height=0.22\textheight,keepaspectratio]{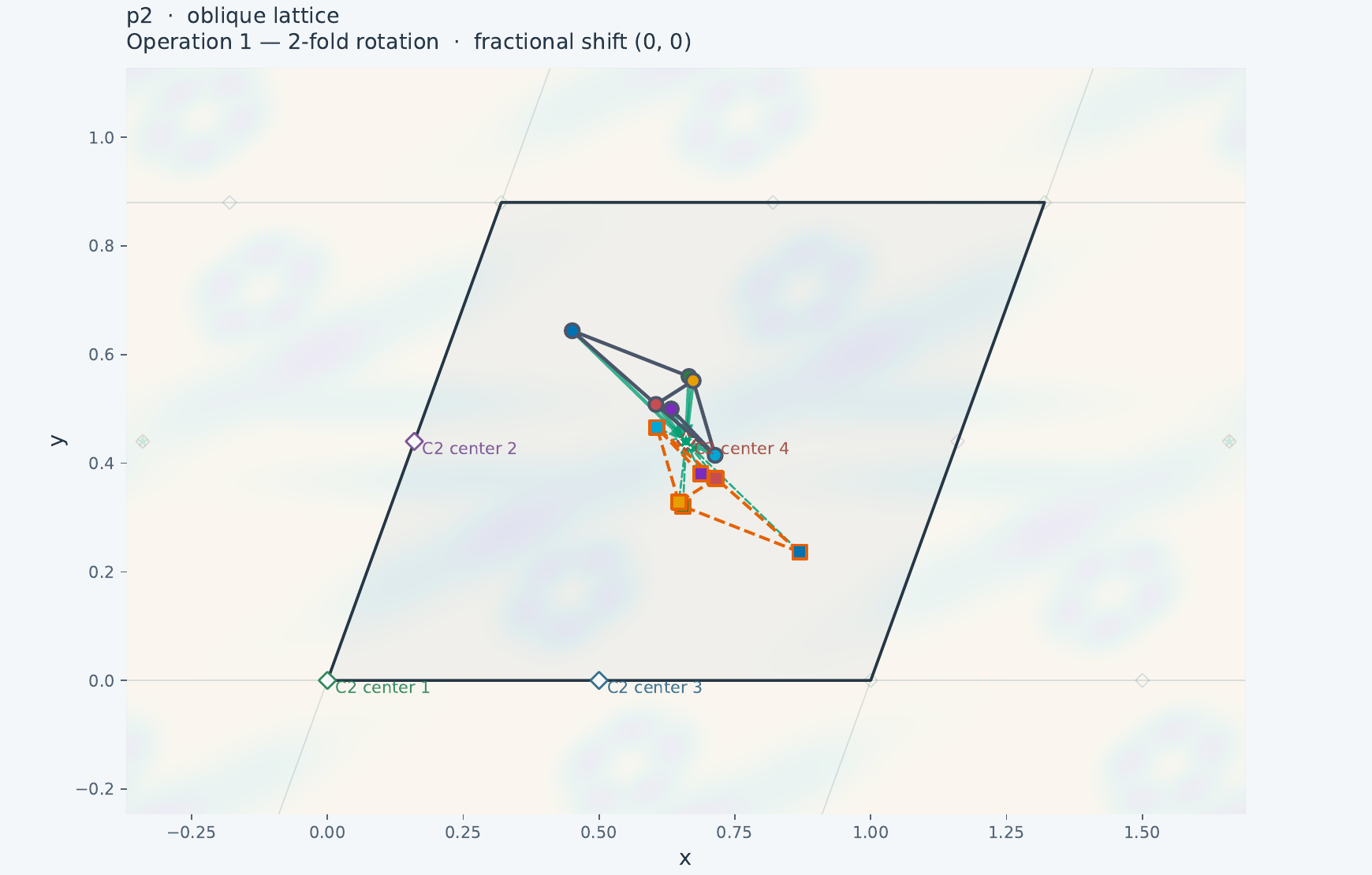} \\[2mm]
\includegraphics[width=\linewidth,height=0.22\textheight,keepaspectratio]{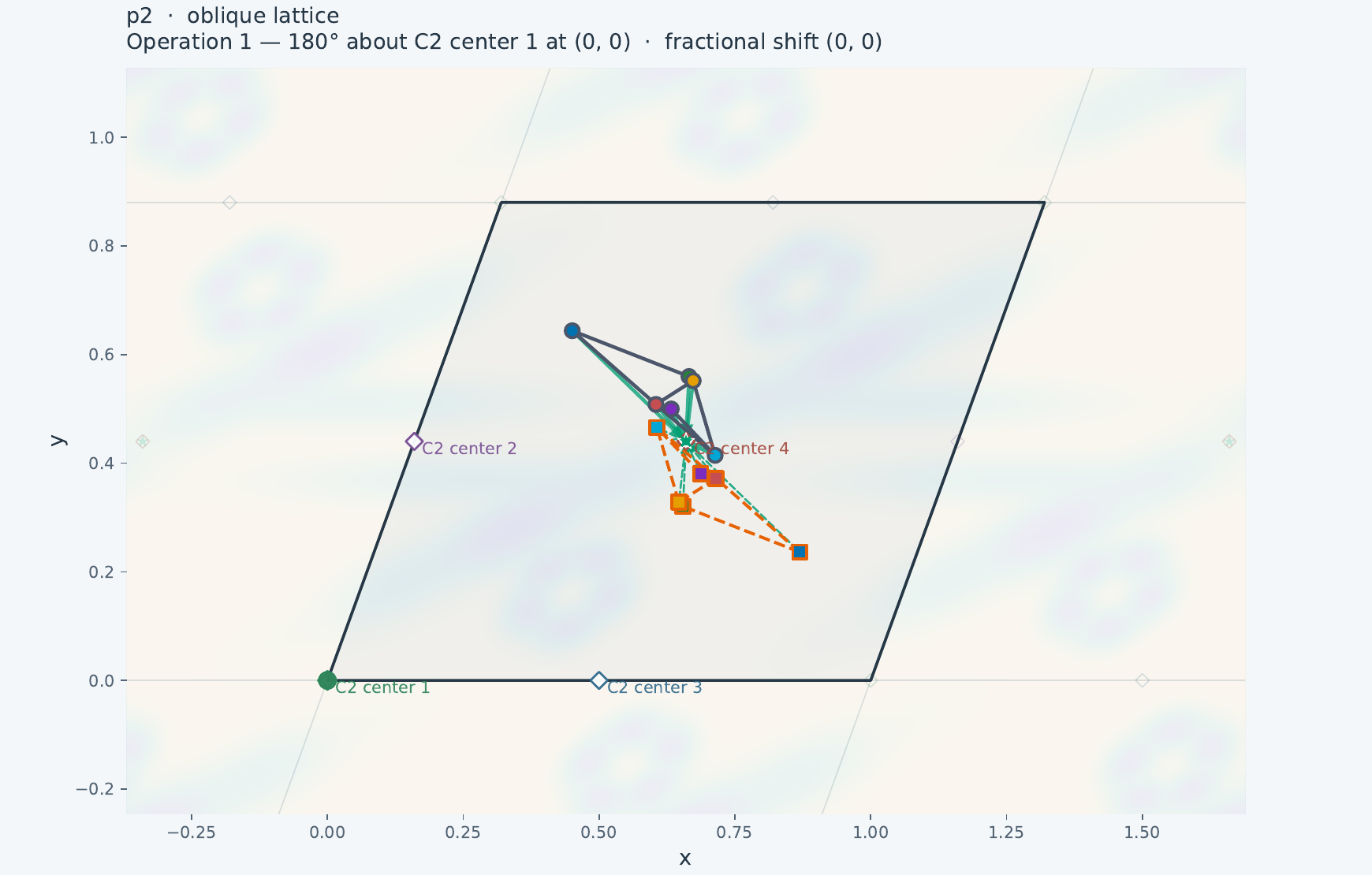} & \includegraphics[width=\linewidth,height=0.22\textheight,keepaspectratio]{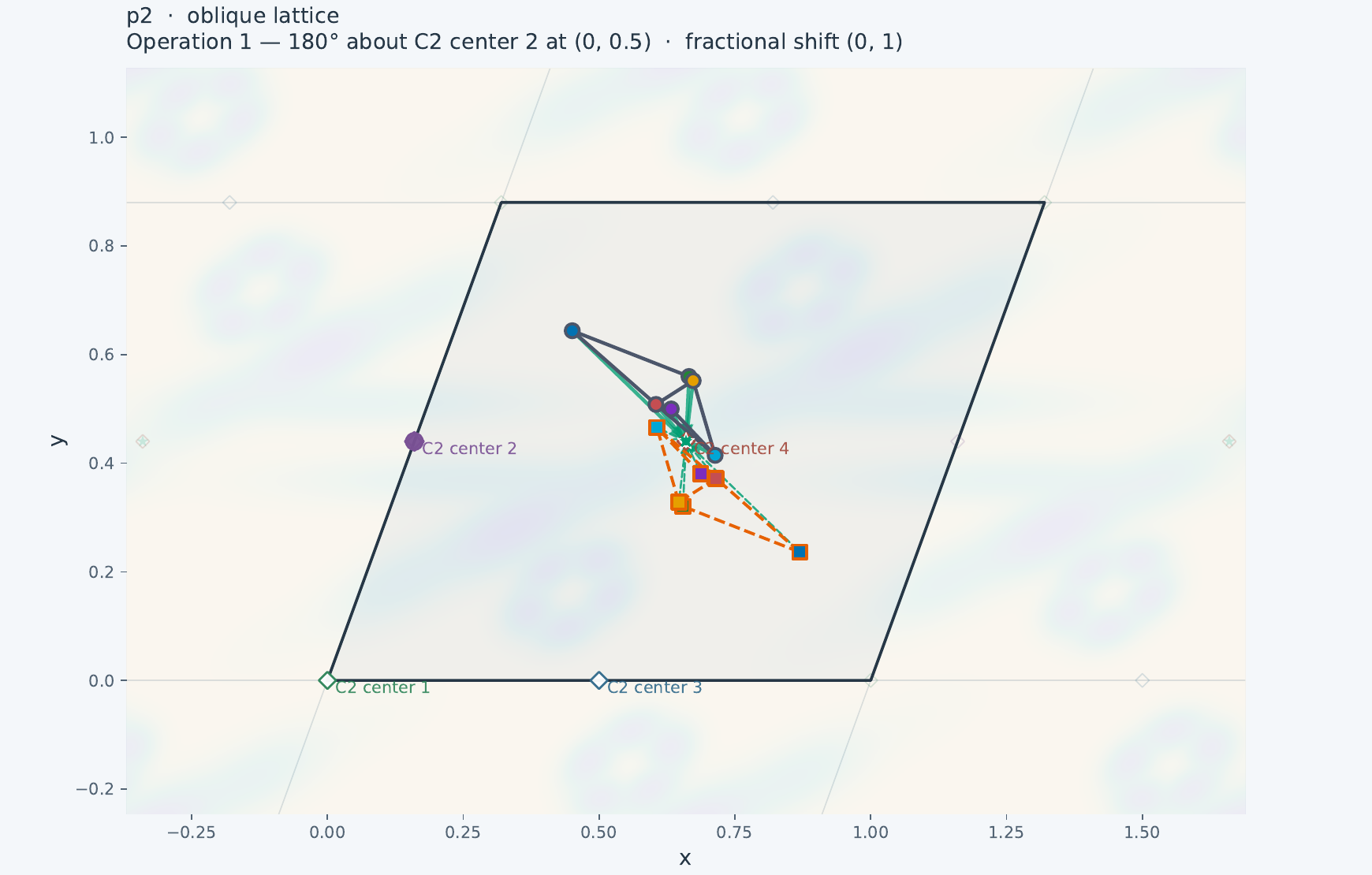} \\[2mm]
\includegraphics[width=\linewidth,height=0.22\textheight,keepaspectratio]{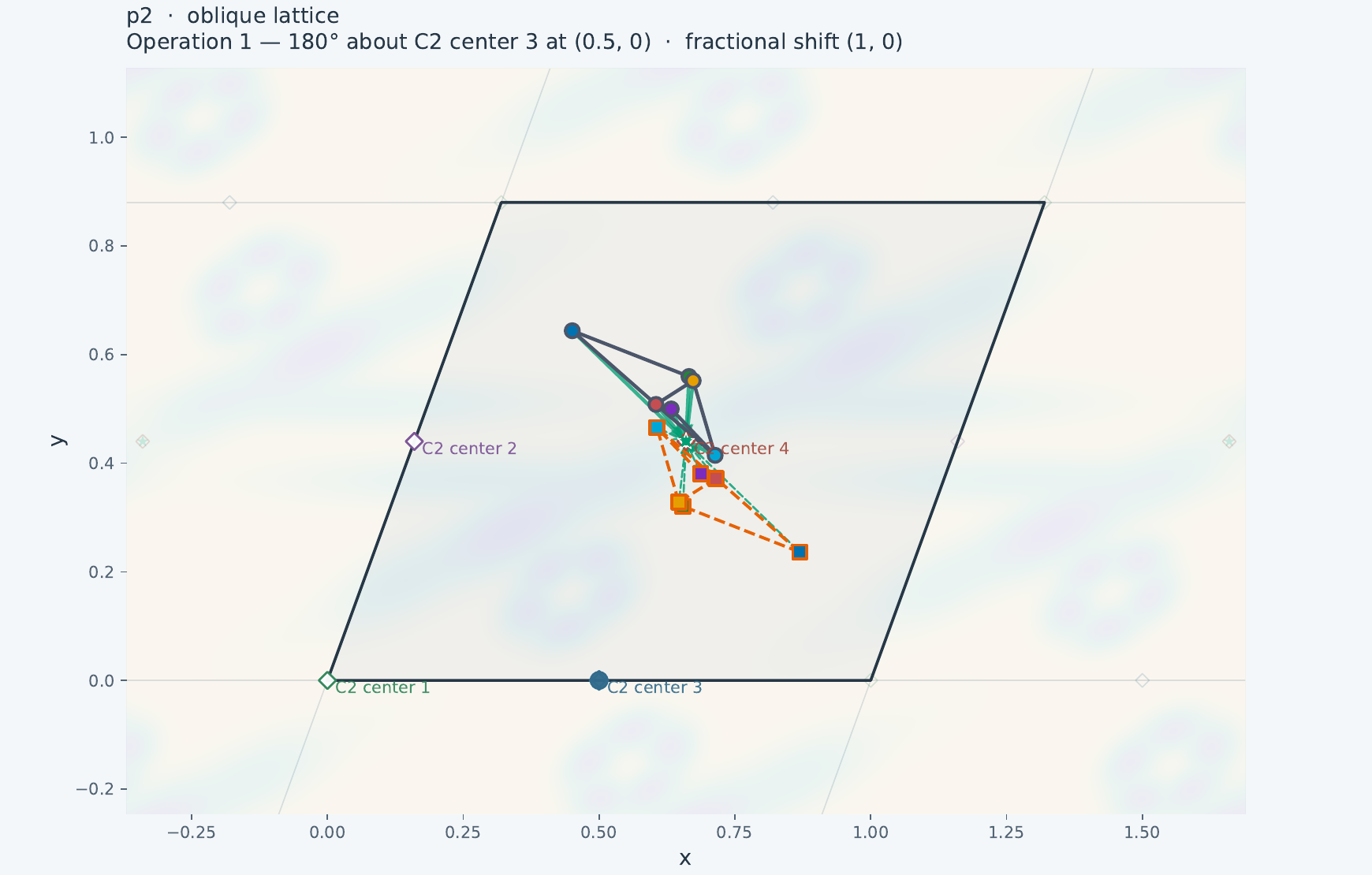} & \includegraphics[width=\linewidth,height=0.22\textheight,keepaspectratio]{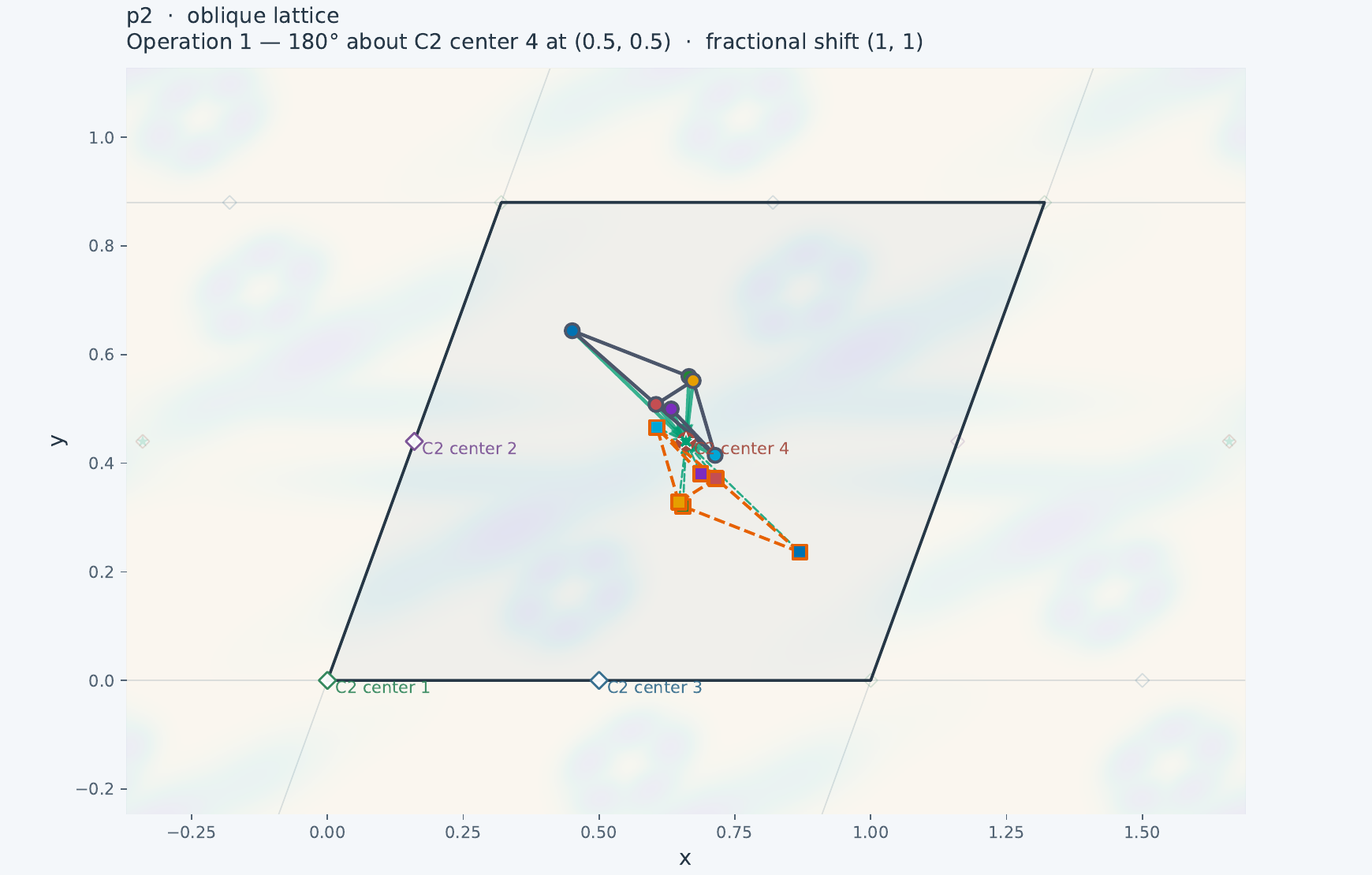} \\[2mm]
\end{longtable}
\Needspace{200pt}
\begin{longtable}{@{}>{\centering\arraybackslash}p{0.475\linewidth}>{\centering\arraybackslash}p{0.475\linewidth}@{}}
\caption[Wallpaper group \texttt{pm}]{Wallpaper group \texttt{pm}, {\small WA is pointing to lattice anchor at $q_{1}=(0,0)$}}\label{tab:wa-actions-pm}\\[1mm]
\multicolumn{2}{c}{\includegraphics[width=0.96\linewidth]{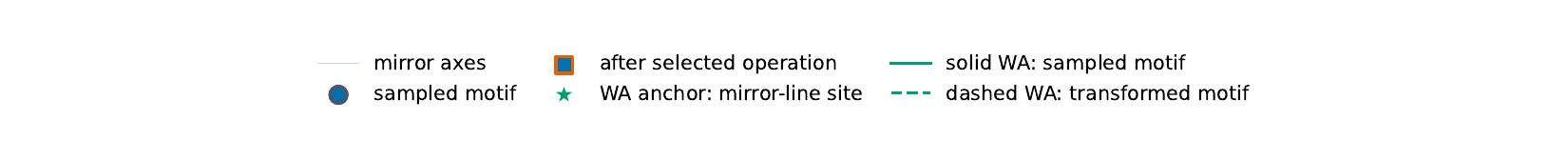}} \\[1mm]
\endfirsthead
\caption[]{Wallpaper group \texttt{pm} (continued)}\\[1mm]
\endhead
\includegraphics[width=\linewidth,height=0.22\textheight,keepaspectratio]{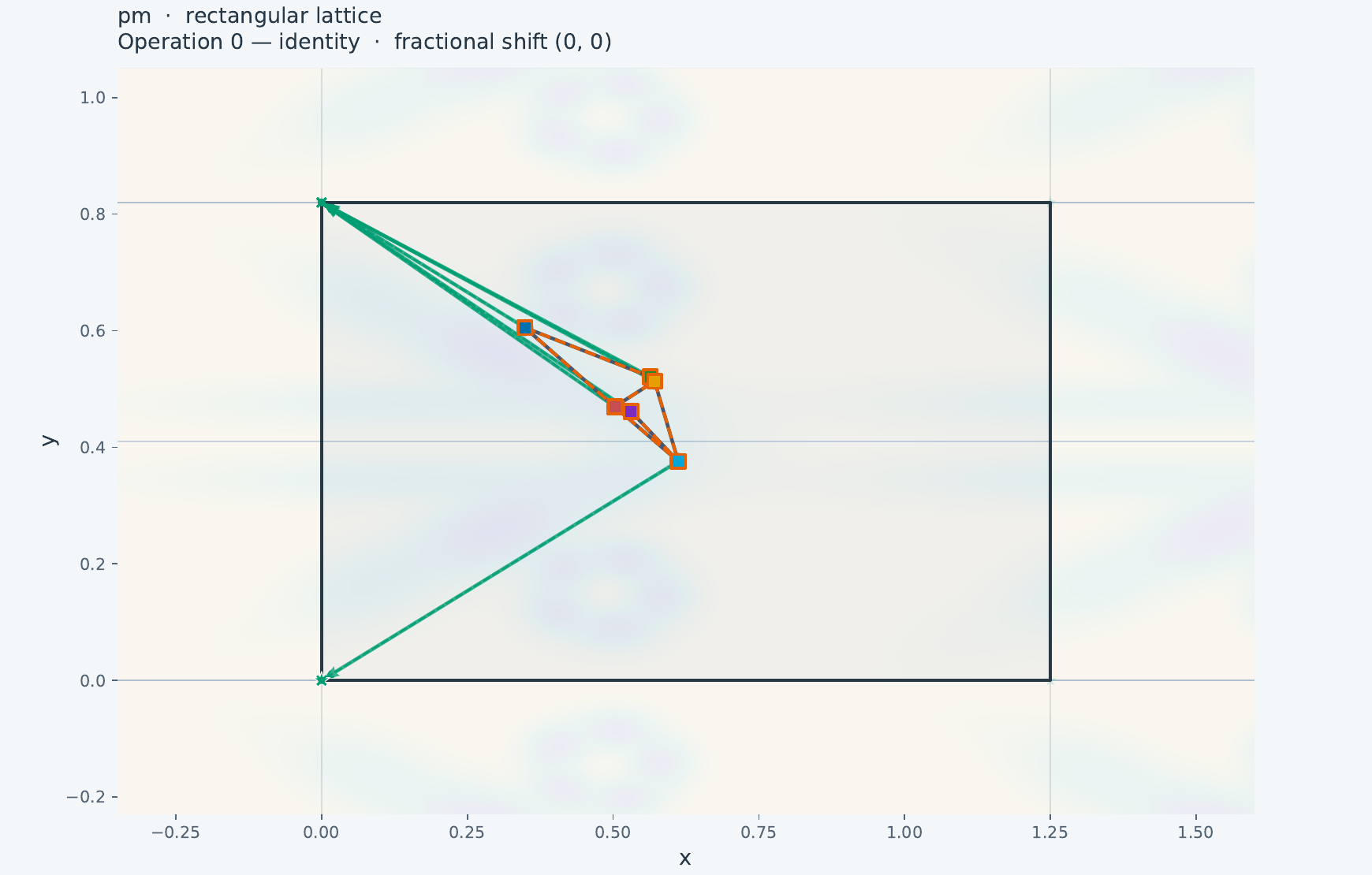} & \includegraphics[width=\linewidth,height=0.22\textheight,keepaspectratio]{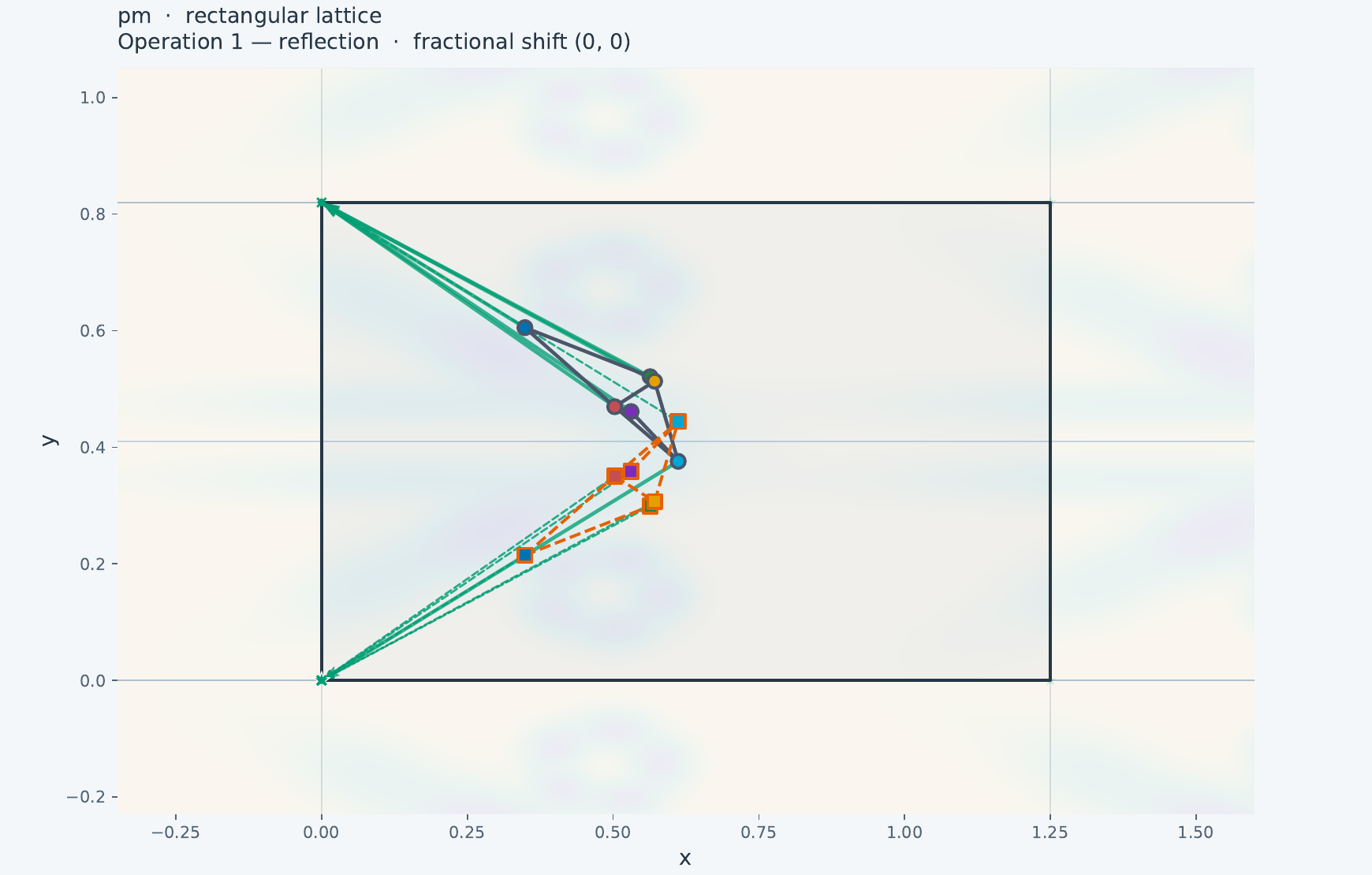} \\[2mm]
\end{longtable}
\Needspace{200pt}
\begin{longtable}{@{}>{\centering\arraybackslash}p{0.475\linewidth}>{\centering\arraybackslash}p{0.475\linewidth}@{}}
\caption[Wallpaper group \texttt{pg}]{Wallpaper group \texttt{pg}, {\small WA is pointing to lattice anchor at $q_{1}=(0,0),\; q_{2}=(0,\frac{1}{2})$}}\label{tab:wa-actions-pg}\\[1mm]
\multicolumn{2}{c}{\includegraphics[width=0.96\linewidth]{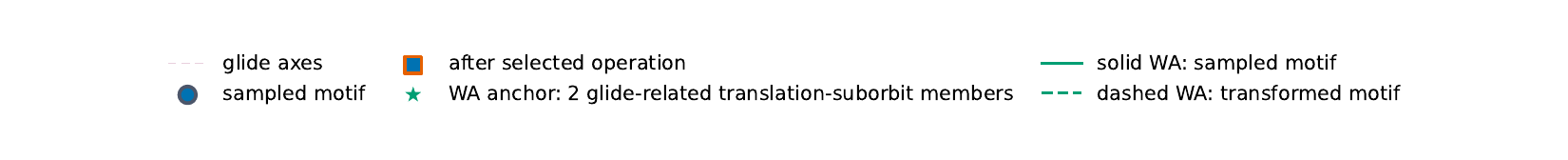}} \\[1mm]
\endfirsthead
\caption[]{Wallpaper group \texttt{pg} (continued)}\\[1mm]
\endhead
\includegraphics[width=\linewidth,height=0.22\textheight,keepaspectratio]{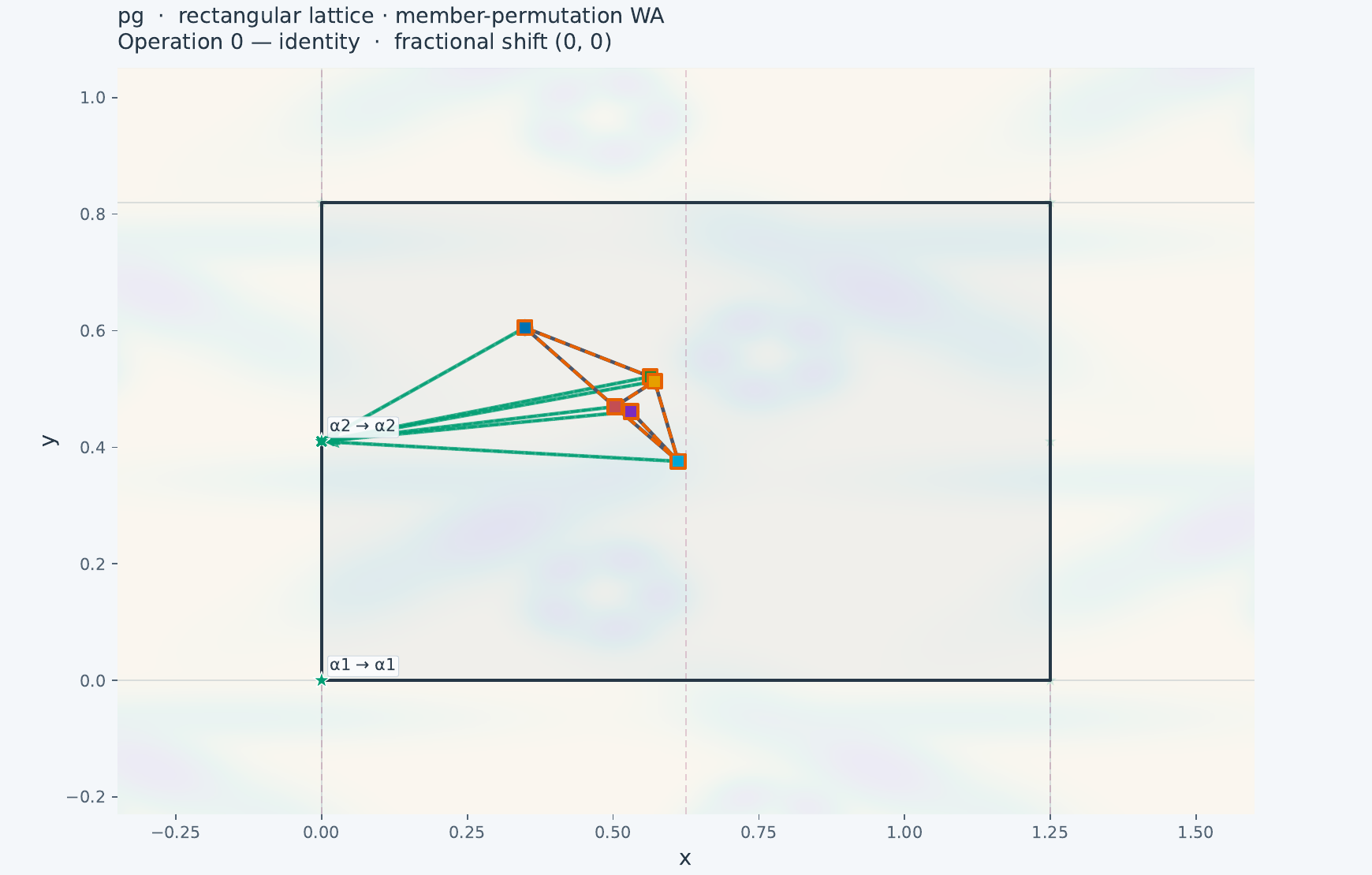} & \includegraphics[width=\linewidth,height=0.22\textheight,keepaspectratio]{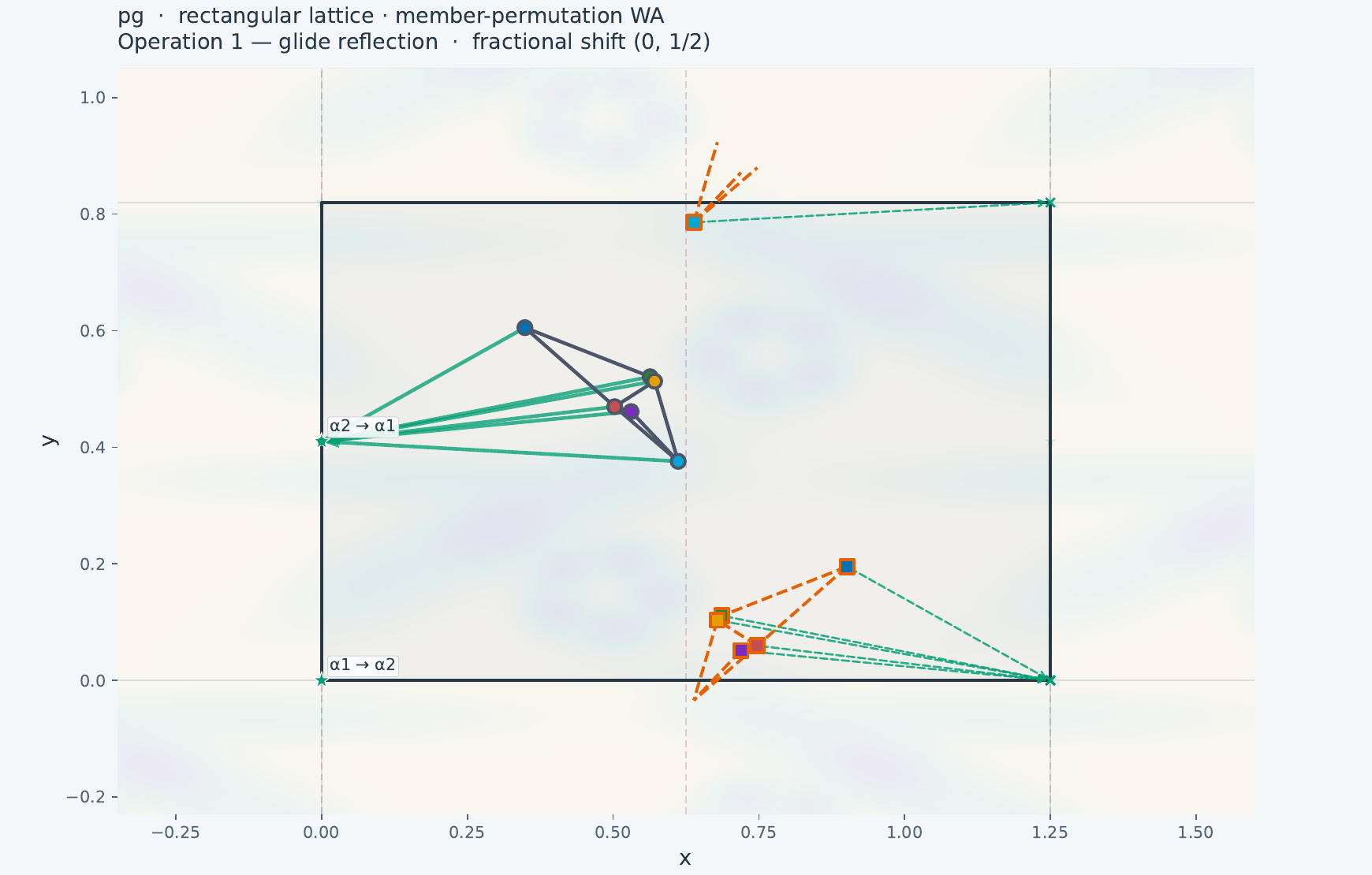} \\[2mm]
\end{longtable}
\Needspace{231pt}
\begin{longtable}{@{}>{\centering\arraybackslash}p{0.475\linewidth}>{\centering\arraybackslash}p{0.475\linewidth}@{}}
\caption[Wallpaper group \texttt{cm}]{Wallpaper group \texttt{cm}, {\small WA is pointing to lattice anchor at $q_{1}=(0,0)$}, {\small centered conventional cell shown, with the WA member multiplicity counted modulo primitive translations}}\label{tab:wa-actions-cm}\\[1mm]
\multicolumn{2}{c}{\includegraphics[width=0.96\linewidth]{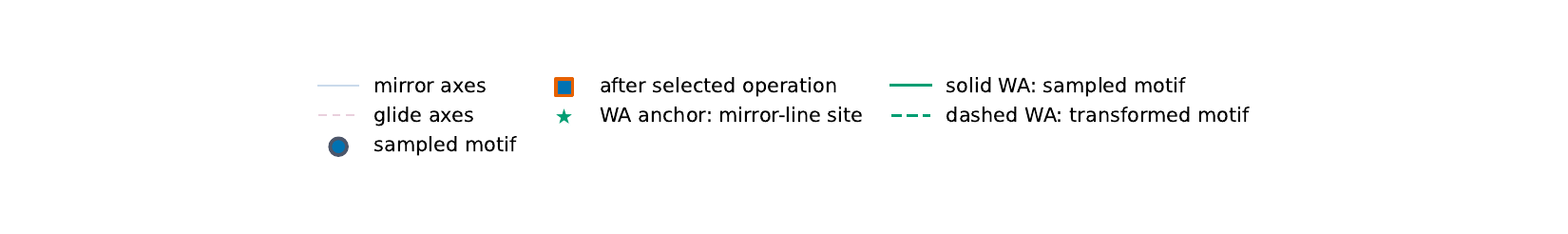}} \\[1mm]
\endfirsthead
\caption[]{Wallpaper group \texttt{cm} (continued)}\\[1mm]
\endhead
\includegraphics[width=\linewidth,height=0.22\textheight,keepaspectratio]{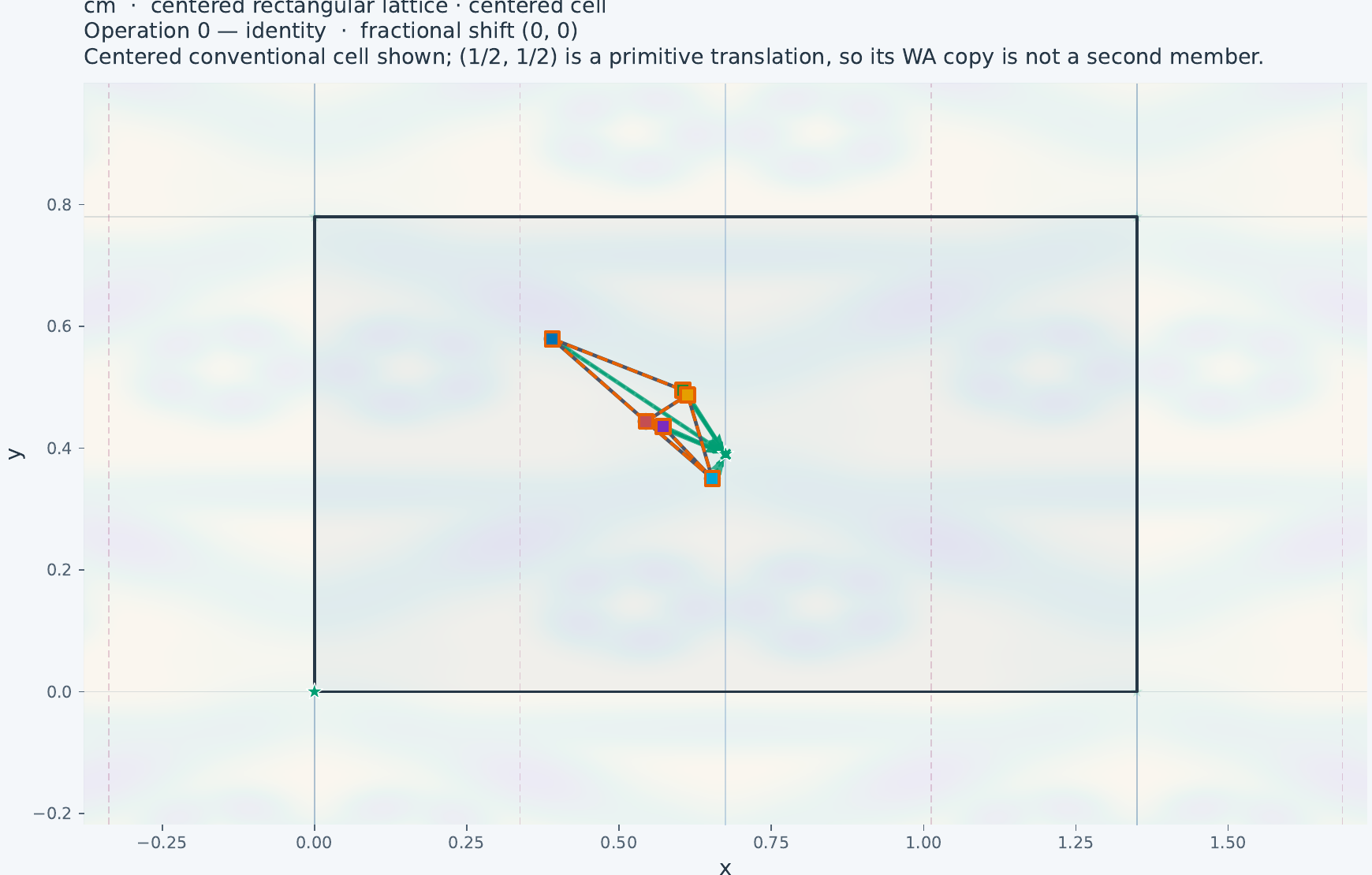} & \includegraphics[width=\linewidth,height=0.22\textheight,keepaspectratio]{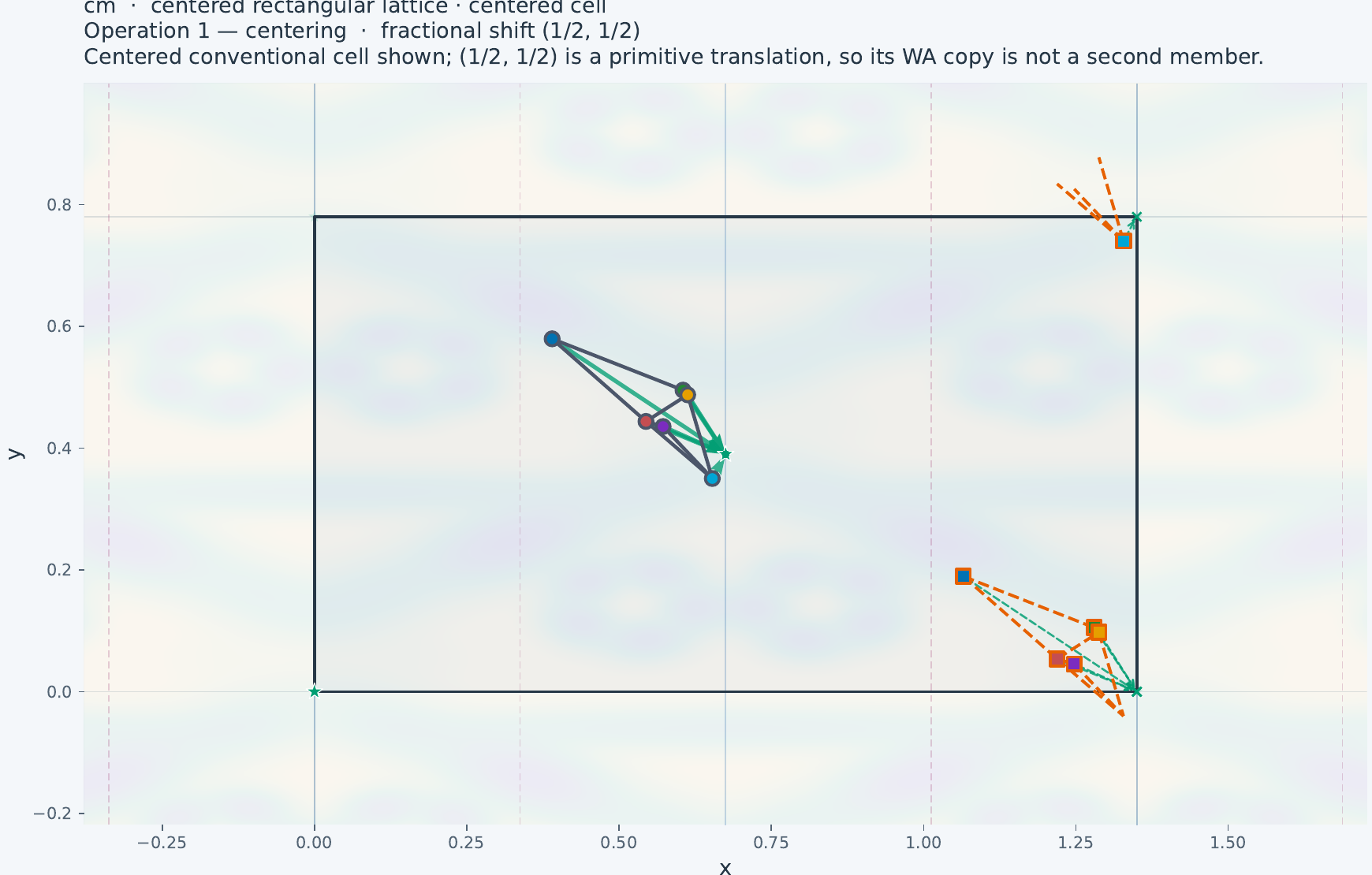} \\[2mm]
\includegraphics[width=\linewidth,height=0.22\textheight,keepaspectratio]{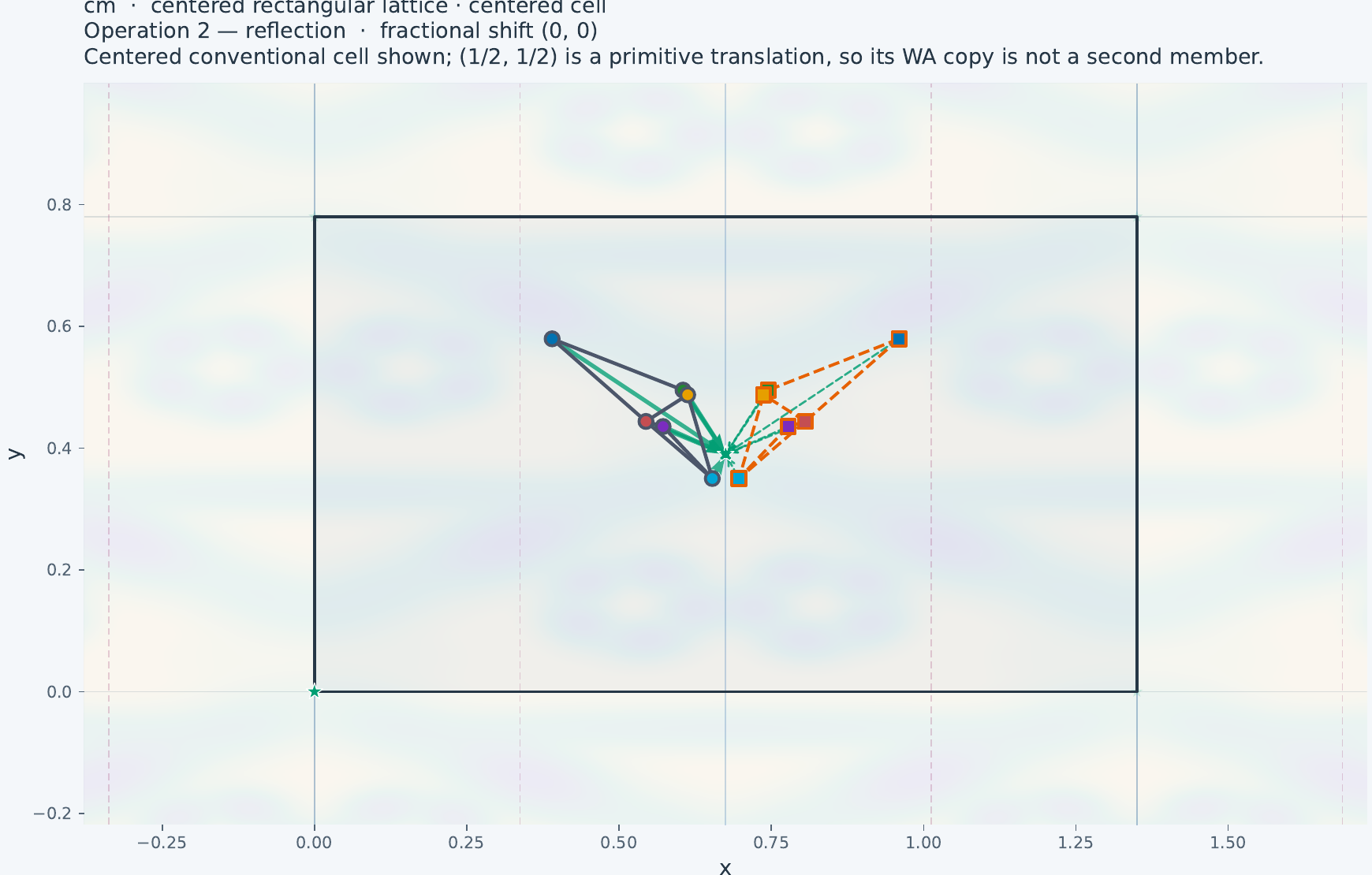} & \includegraphics[width=\linewidth,height=0.22\textheight,keepaspectratio]{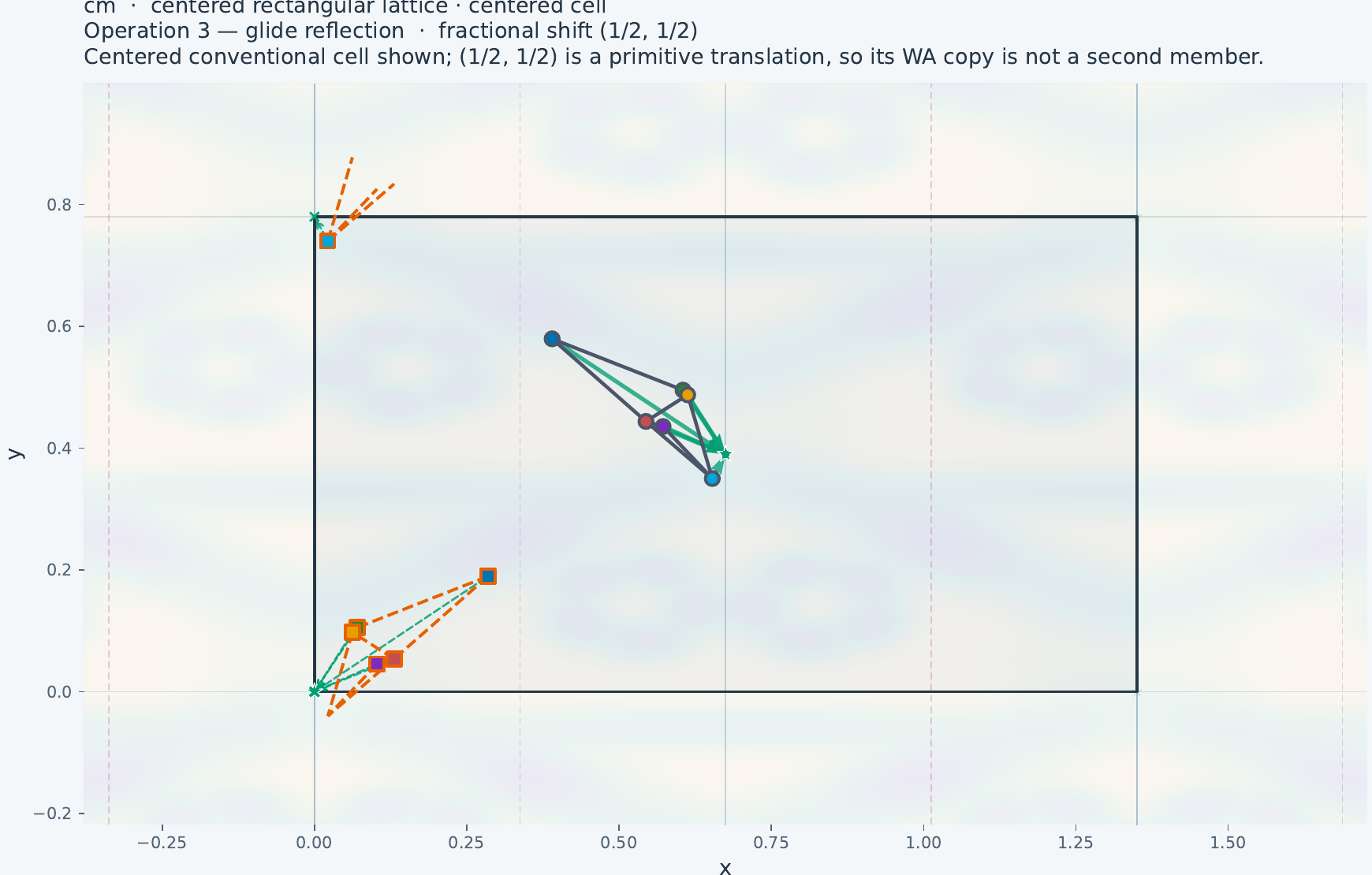} \\[2mm]
\end{longtable}
\Needspace{236pt}
\begin{longtable}{@{}>{\centering\arraybackslash}p{0.475\linewidth}>{\centering\arraybackslash}p{0.475\linewidth}@{}}
\caption[Wallpaper group \texttt{pmm}]{Wallpaper group \texttt{pmm}, {\small WA is pointing to C2-A at $q_{1}=(\frac{1}{2},\frac{1}{2})$}}\label{tab:wa-actions-pmm}\\[1mm]
\multicolumn{2}{c}{\includegraphics[width=0.96\linewidth]{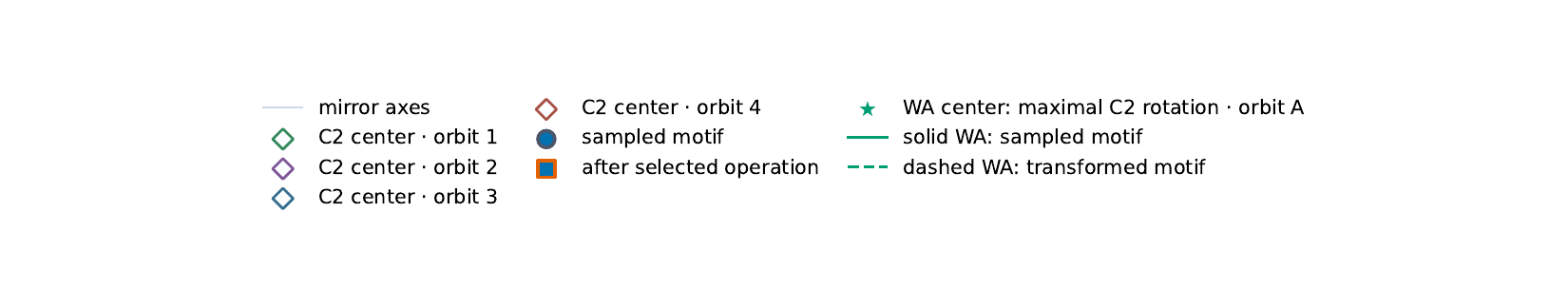}} \\[1mm]
\endfirsthead
\caption[]{Wallpaper group \texttt{pmm} (continued)}\\[1mm]
\endhead
\includegraphics[width=\linewidth,height=0.22\textheight,keepaspectratio]{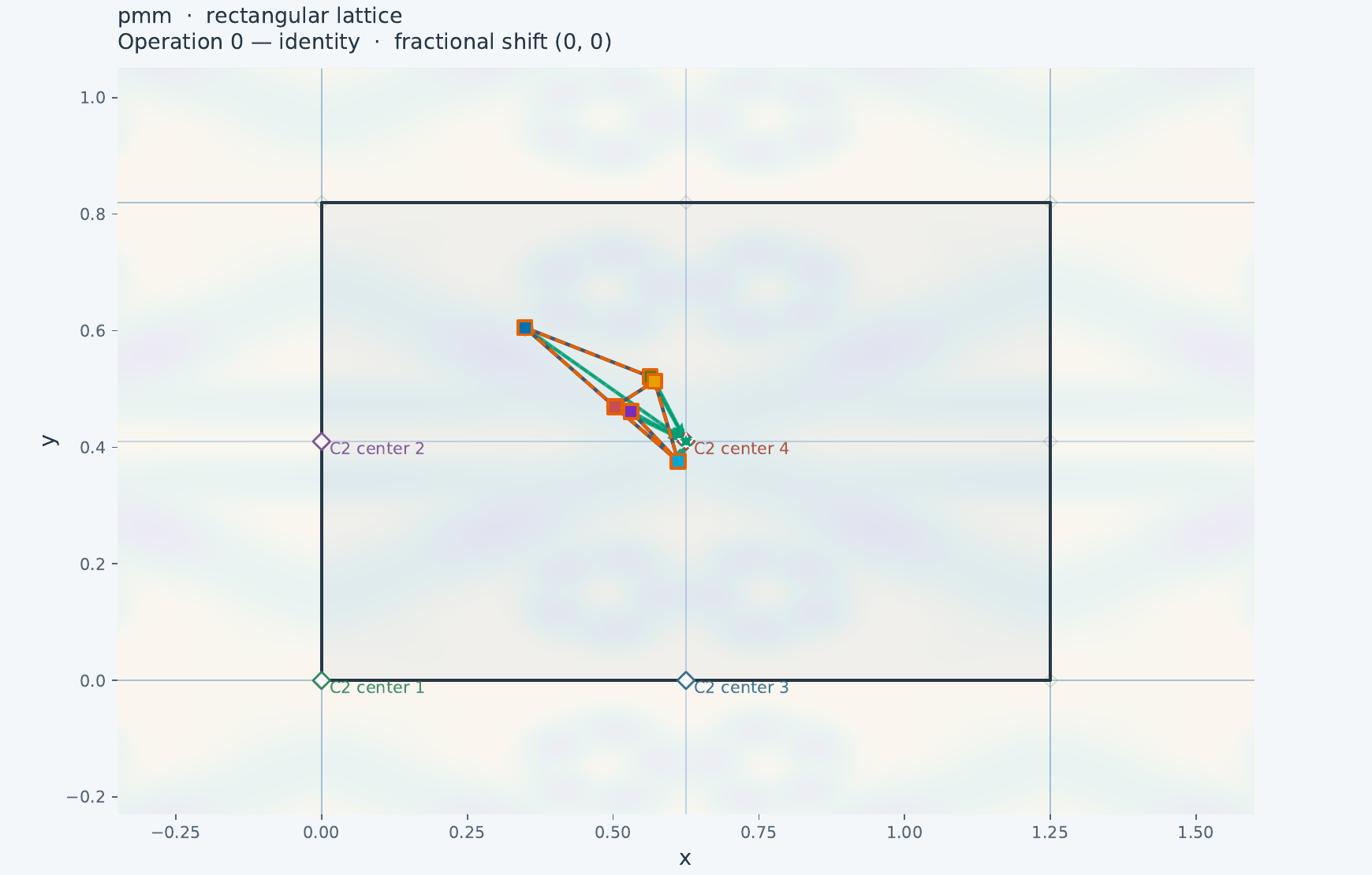} & \includegraphics[width=\linewidth,height=0.22\textheight,keepaspectratio]{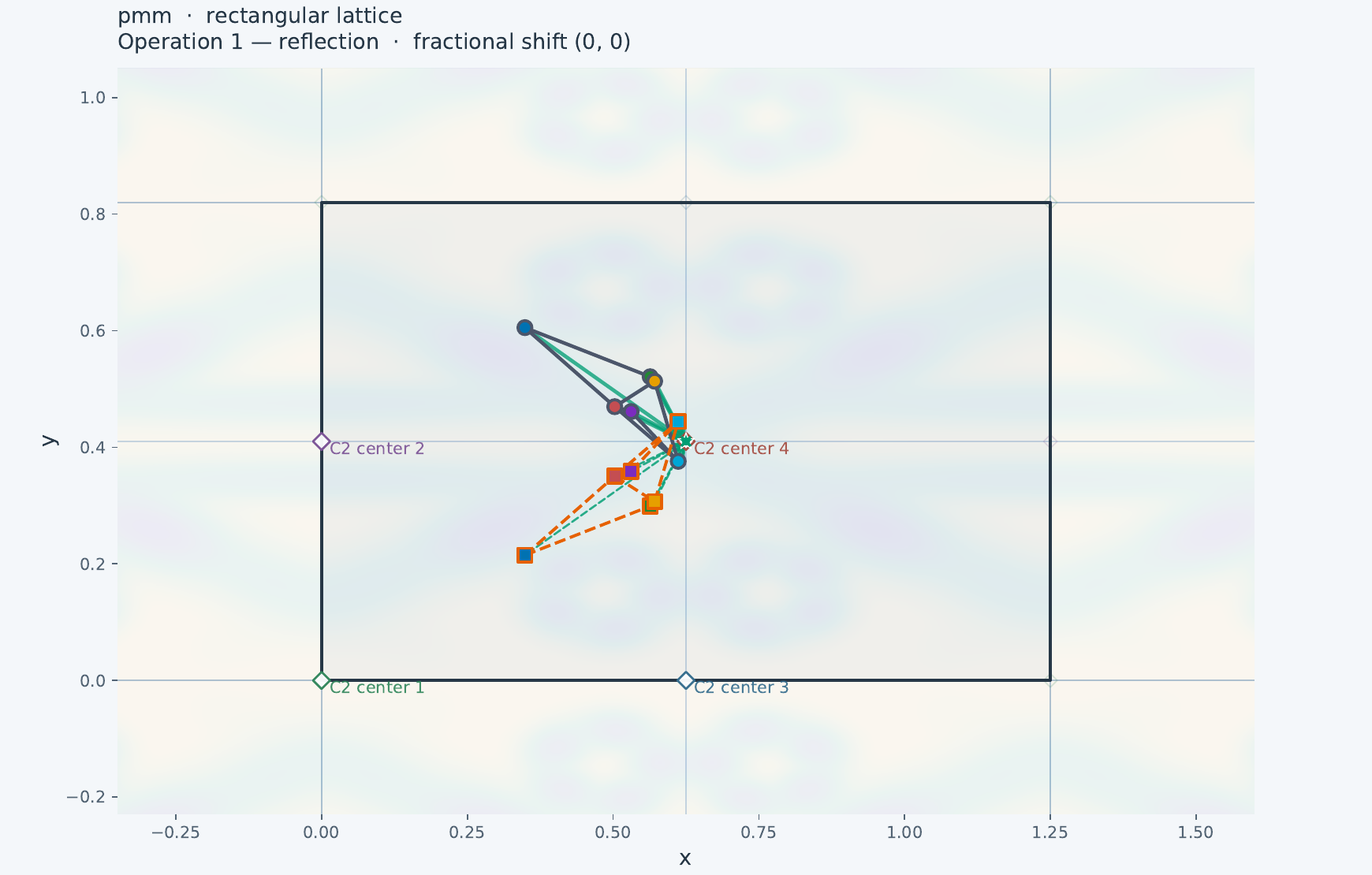} \\[2mm]
\includegraphics[width=\linewidth,height=0.22\textheight,keepaspectratio]{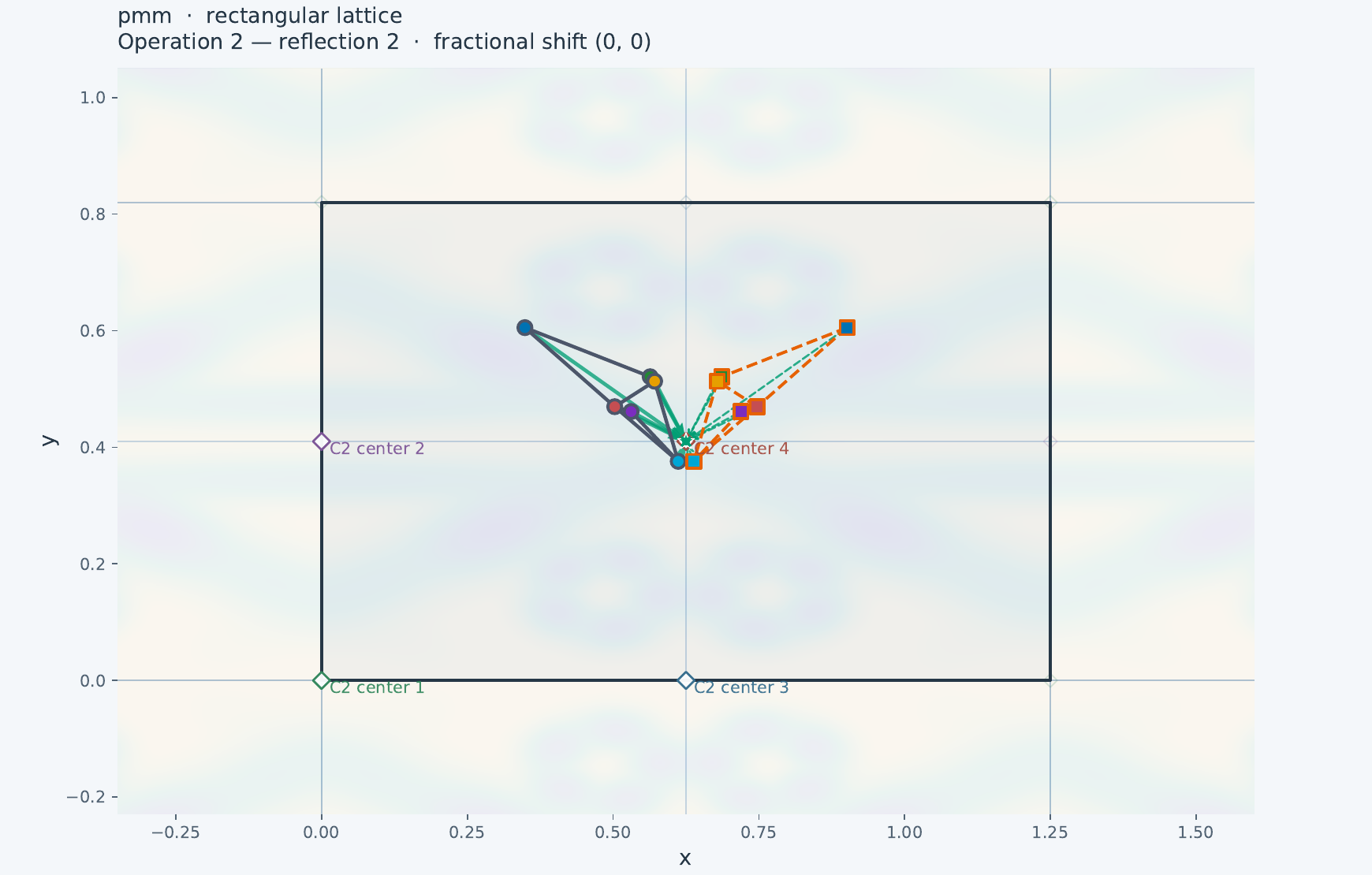} & \includegraphics[width=\linewidth,height=0.22\textheight,keepaspectratio]{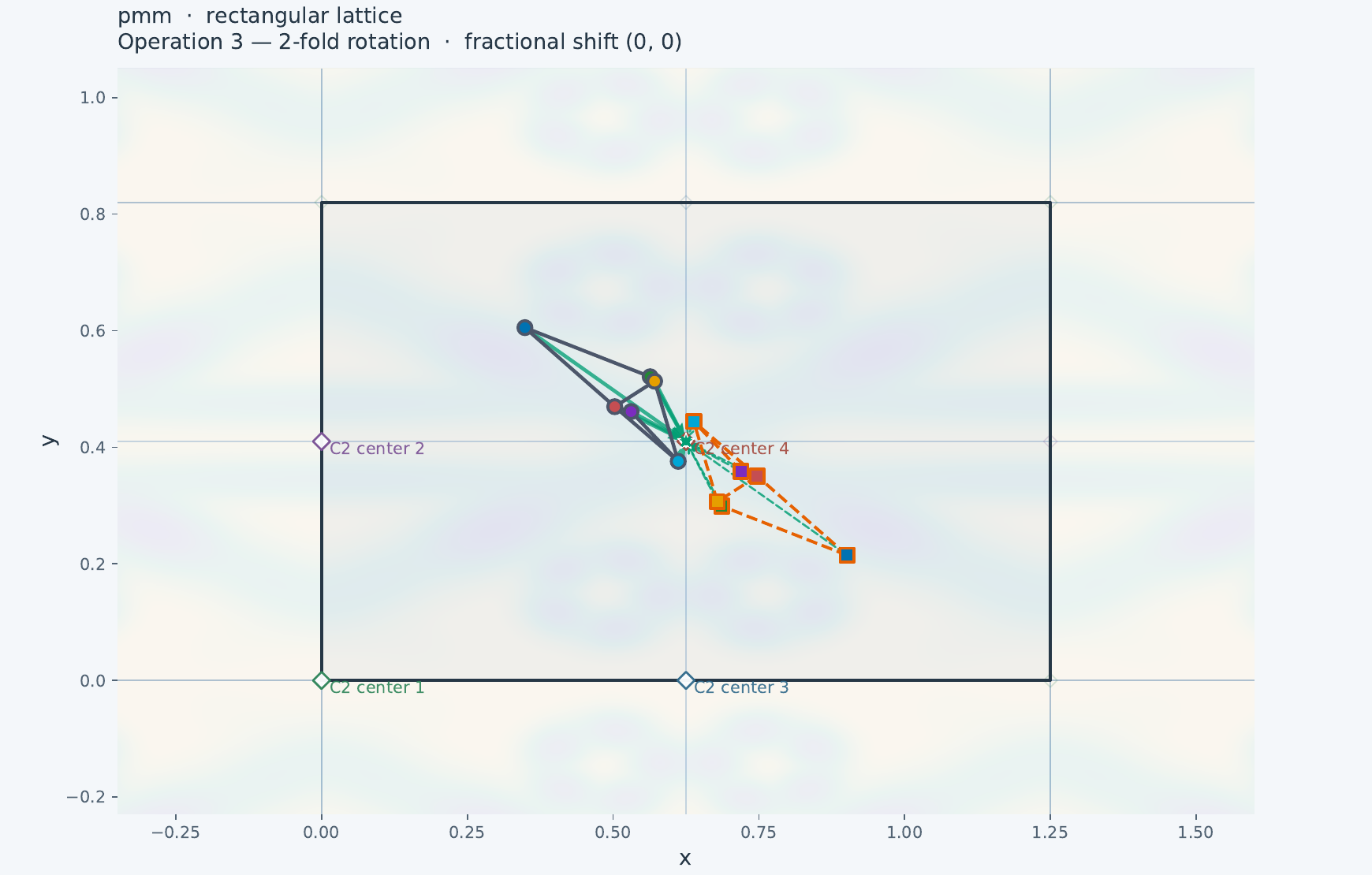} \\[2mm]
\includegraphics[width=\linewidth,height=0.22\textheight,keepaspectratio]{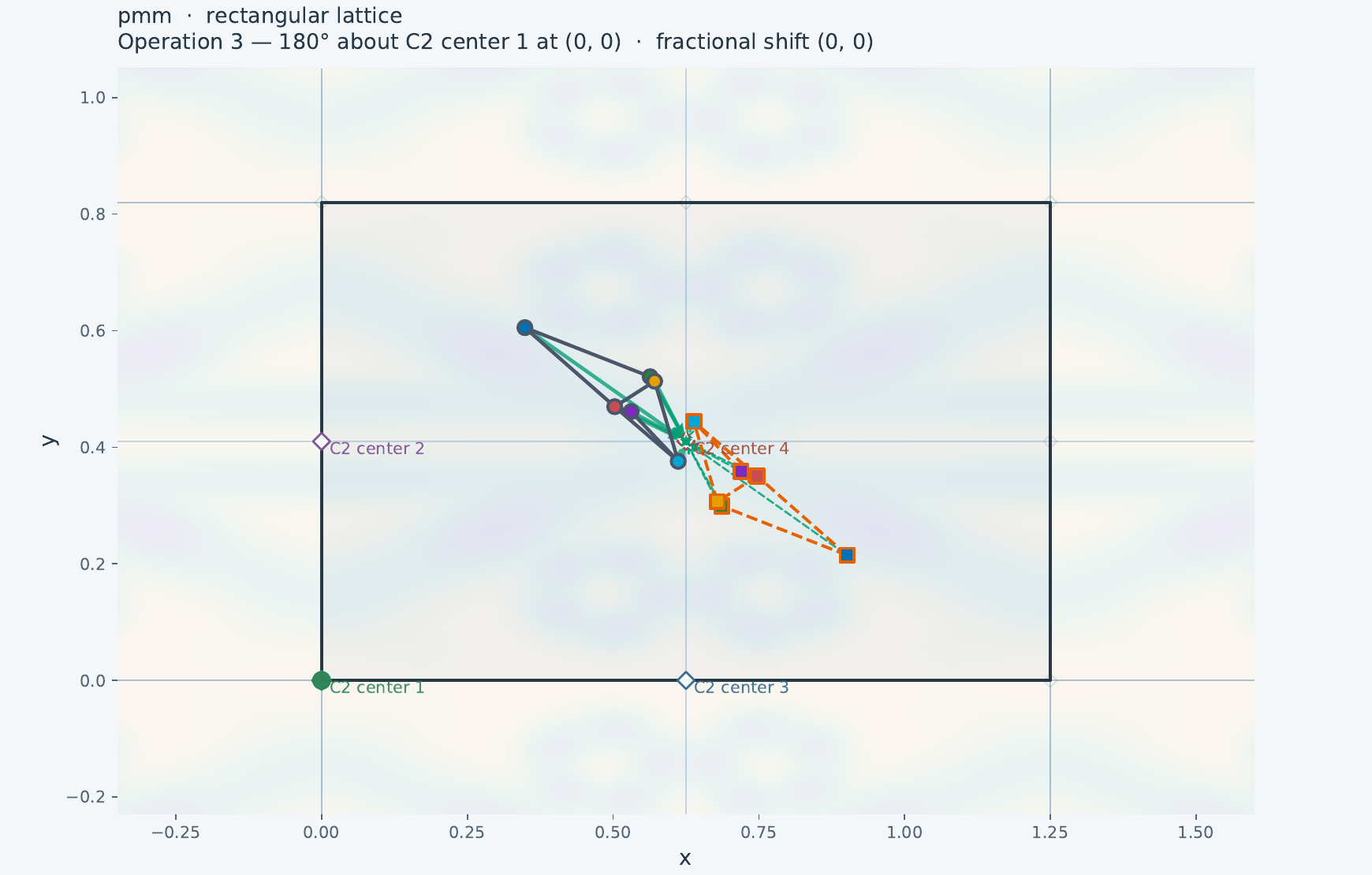} & \includegraphics[width=\linewidth,height=0.22\textheight,keepaspectratio]{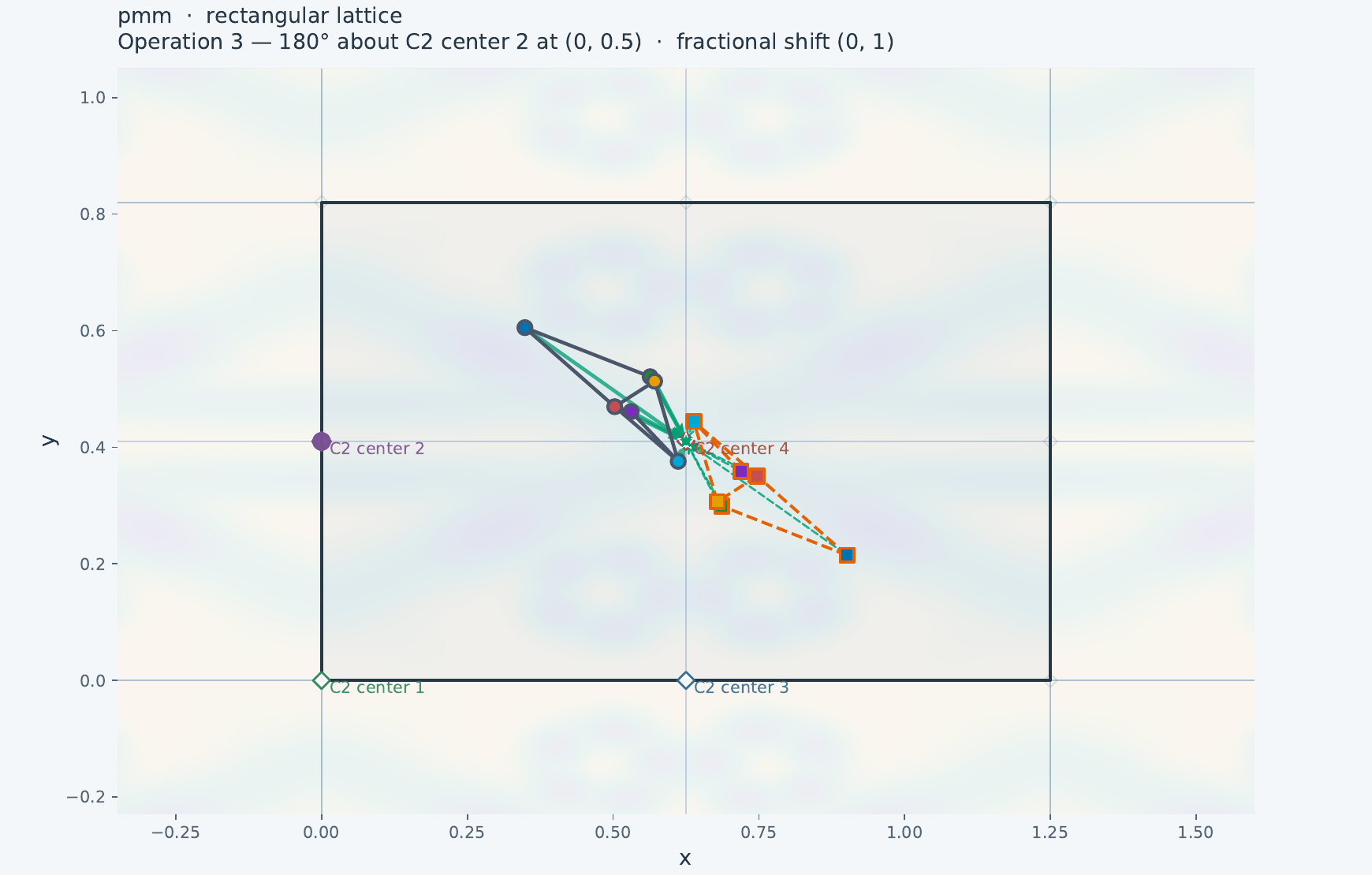} \\[2mm]
\includegraphics[width=\linewidth,height=0.22\textheight,keepaspectratio]{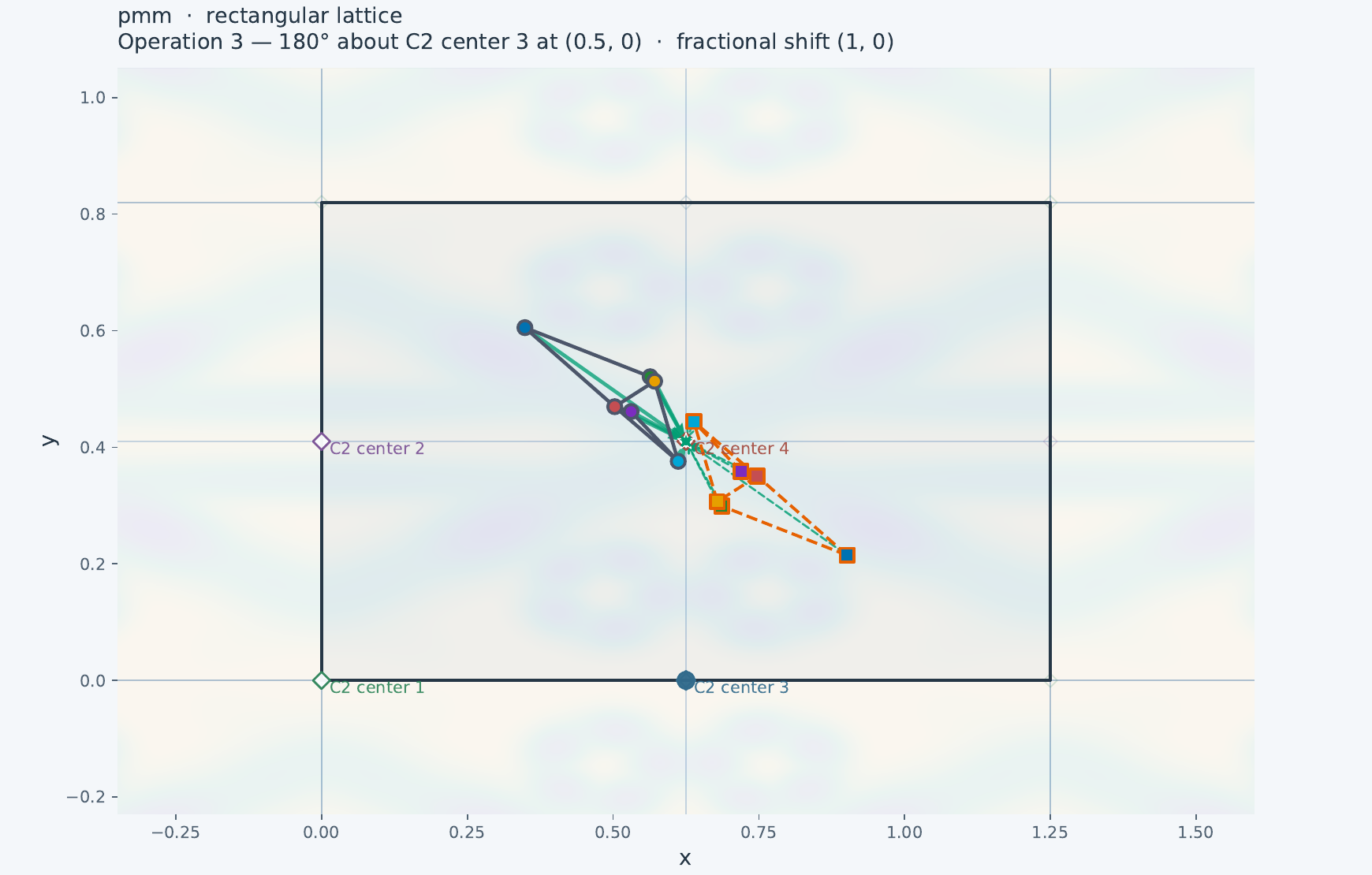} & \includegraphics[width=\linewidth,height=0.22\textheight,keepaspectratio]{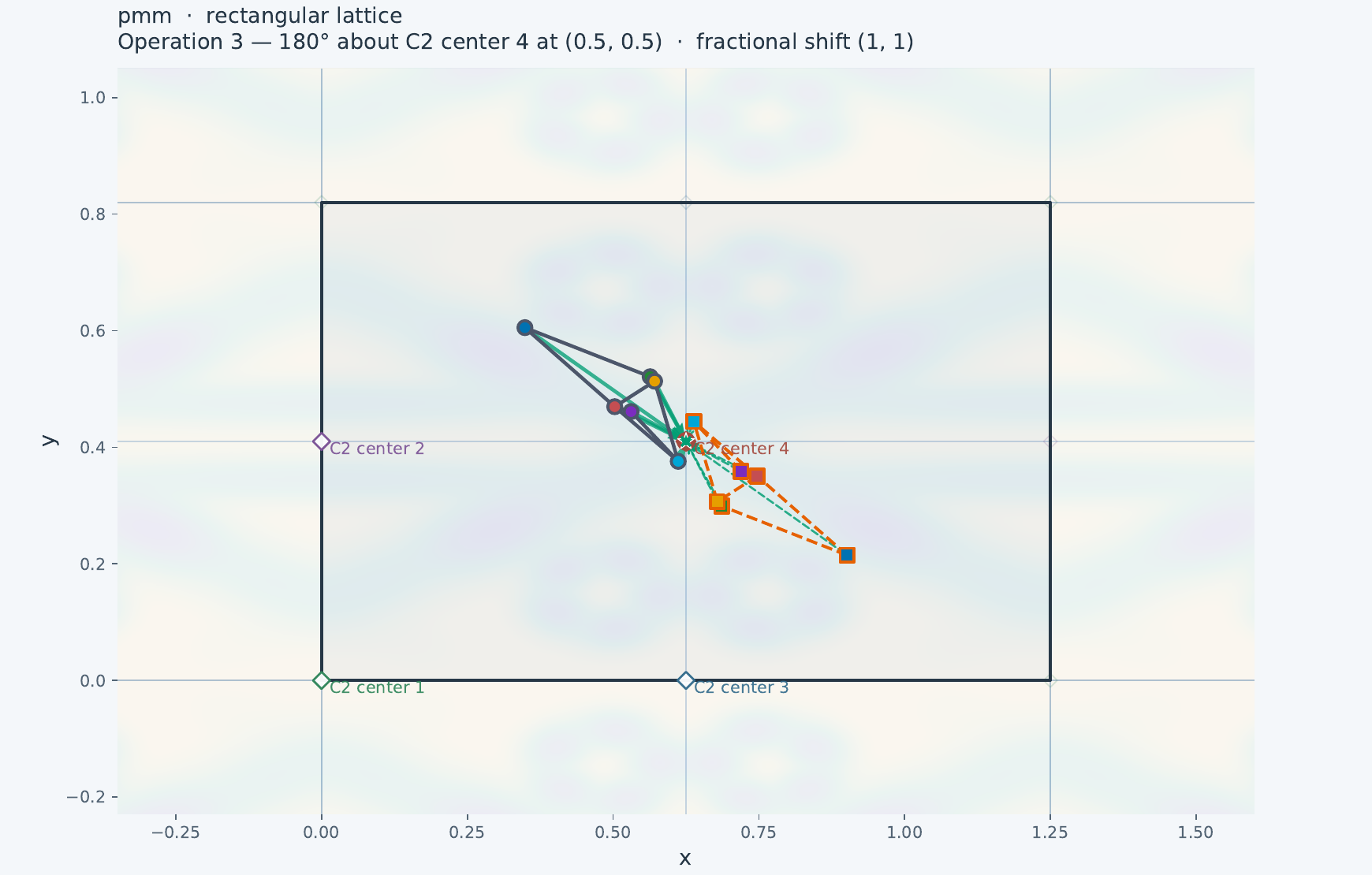} \\[2mm]
\end{longtable}
\Needspace{218pt}
\begin{longtable}{@{}>{\centering\arraybackslash}p{0.475\linewidth}>{\centering\arraybackslash}p{0.475\linewidth}@{}}
\caption[Wallpaper group \texttt{pmg}]{Wallpaper group \texttt{pmg}, {\small WA is pointing to C2-A at $q_{1}=(\frac{1}{2},\frac{1}{2}),\; q_{2}=(0,\frac{1}{2})$}}\label{tab:wa-actions-pmg}\\[1mm]
\multicolumn{2}{c}{\includegraphics[width=0.96\linewidth]{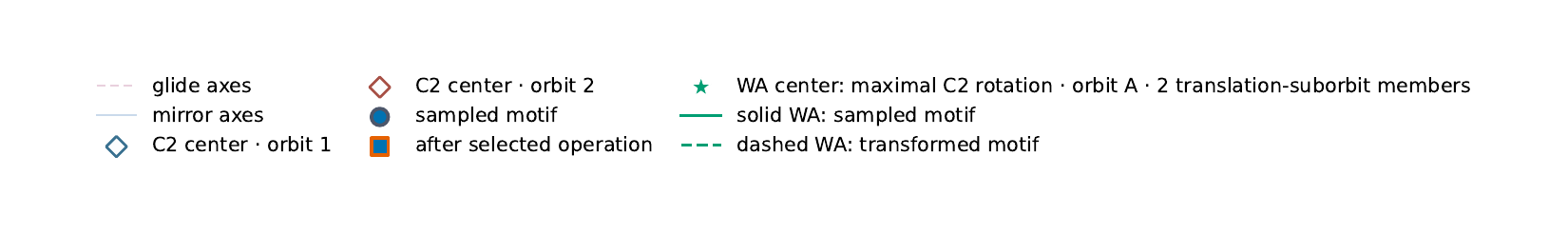}} \\[1mm]
\endfirsthead
\caption[]{Wallpaper group \texttt{pmg} (continued)}\\[1mm]
\endhead
\includegraphics[width=\linewidth,height=0.22\textheight,keepaspectratio]{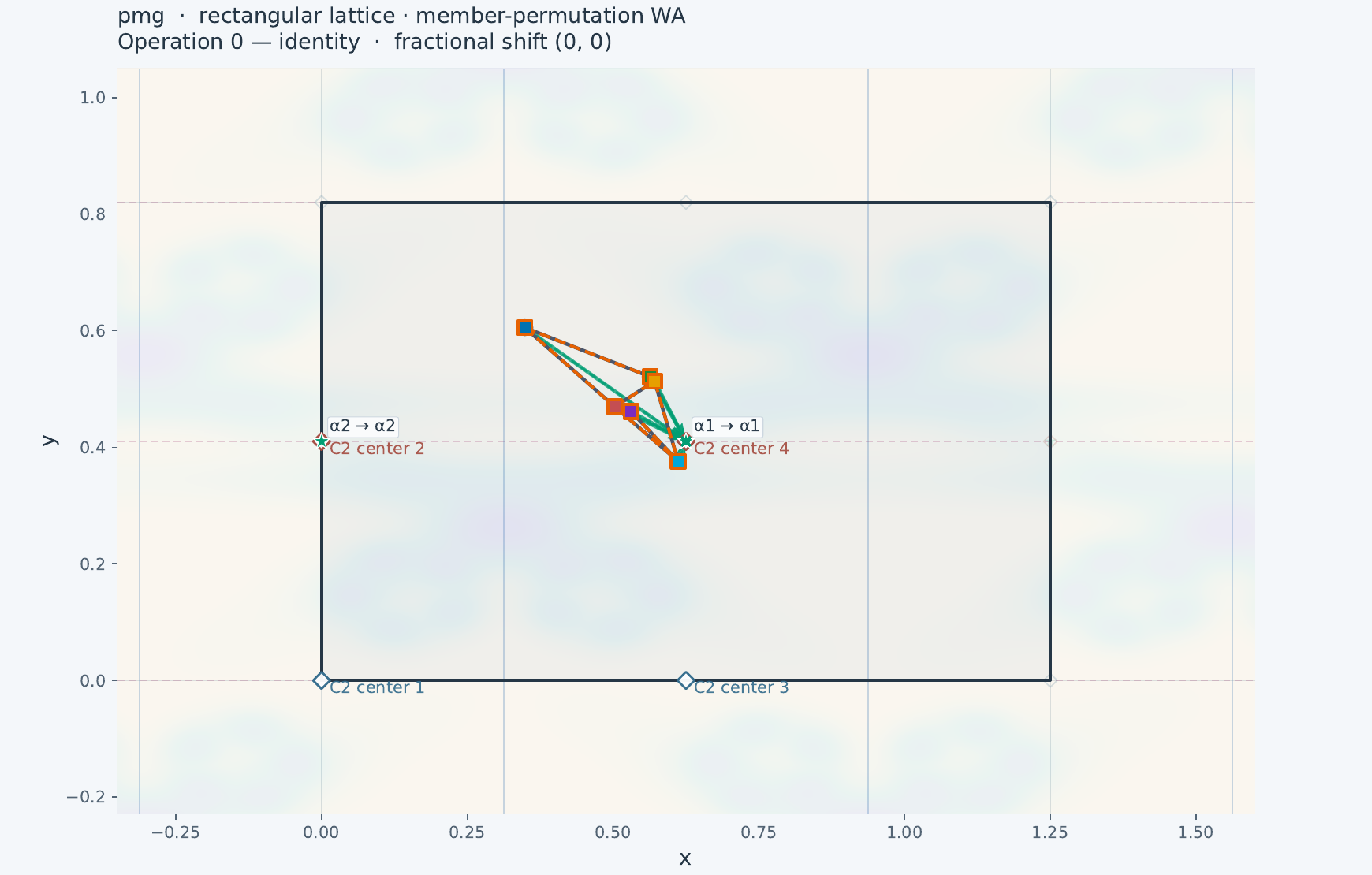} & \includegraphics[width=\linewidth,height=0.22\textheight,keepaspectratio]{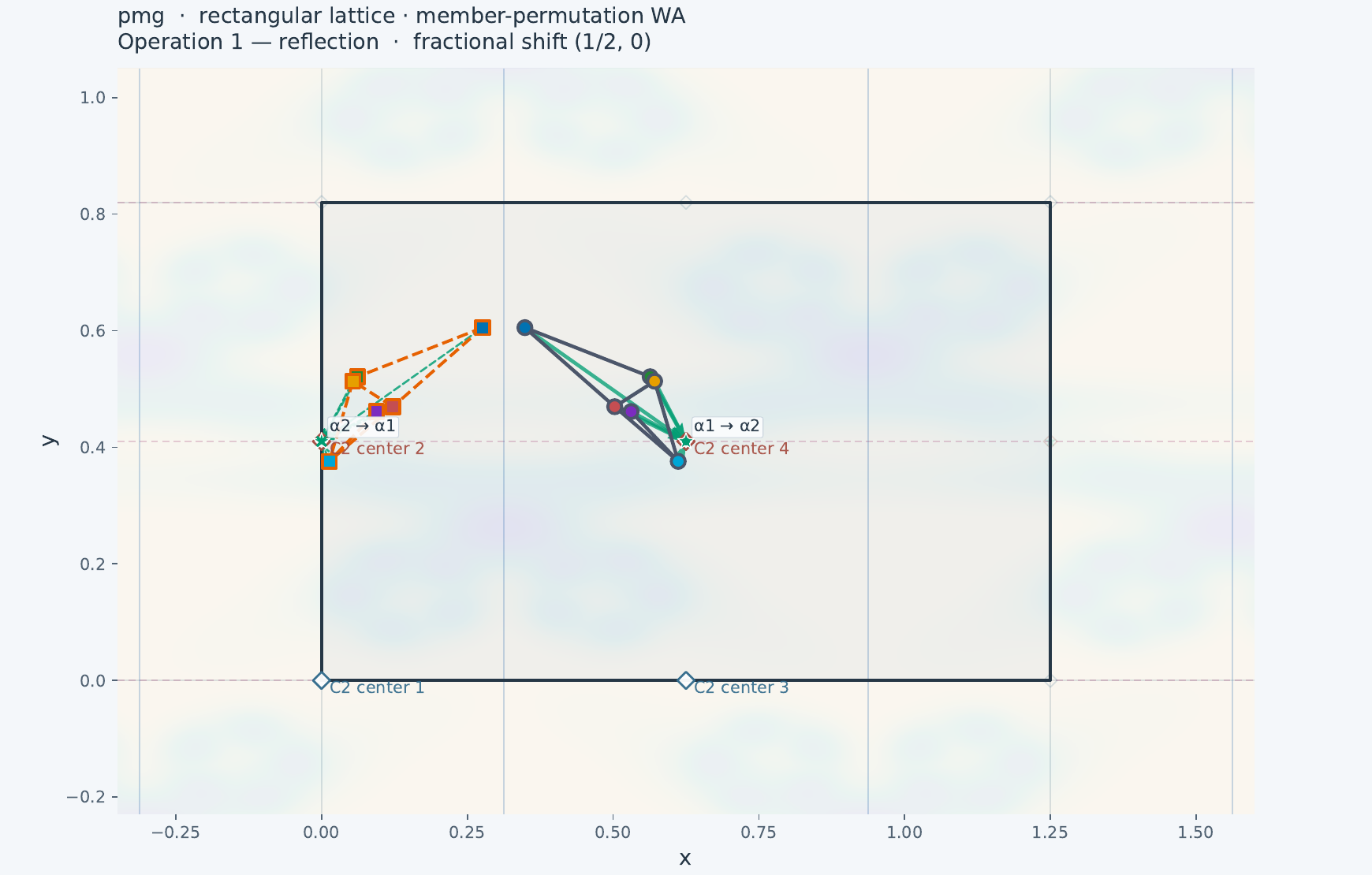} \\[2mm]
\includegraphics[width=\linewidth,height=0.22\textheight,keepaspectratio]{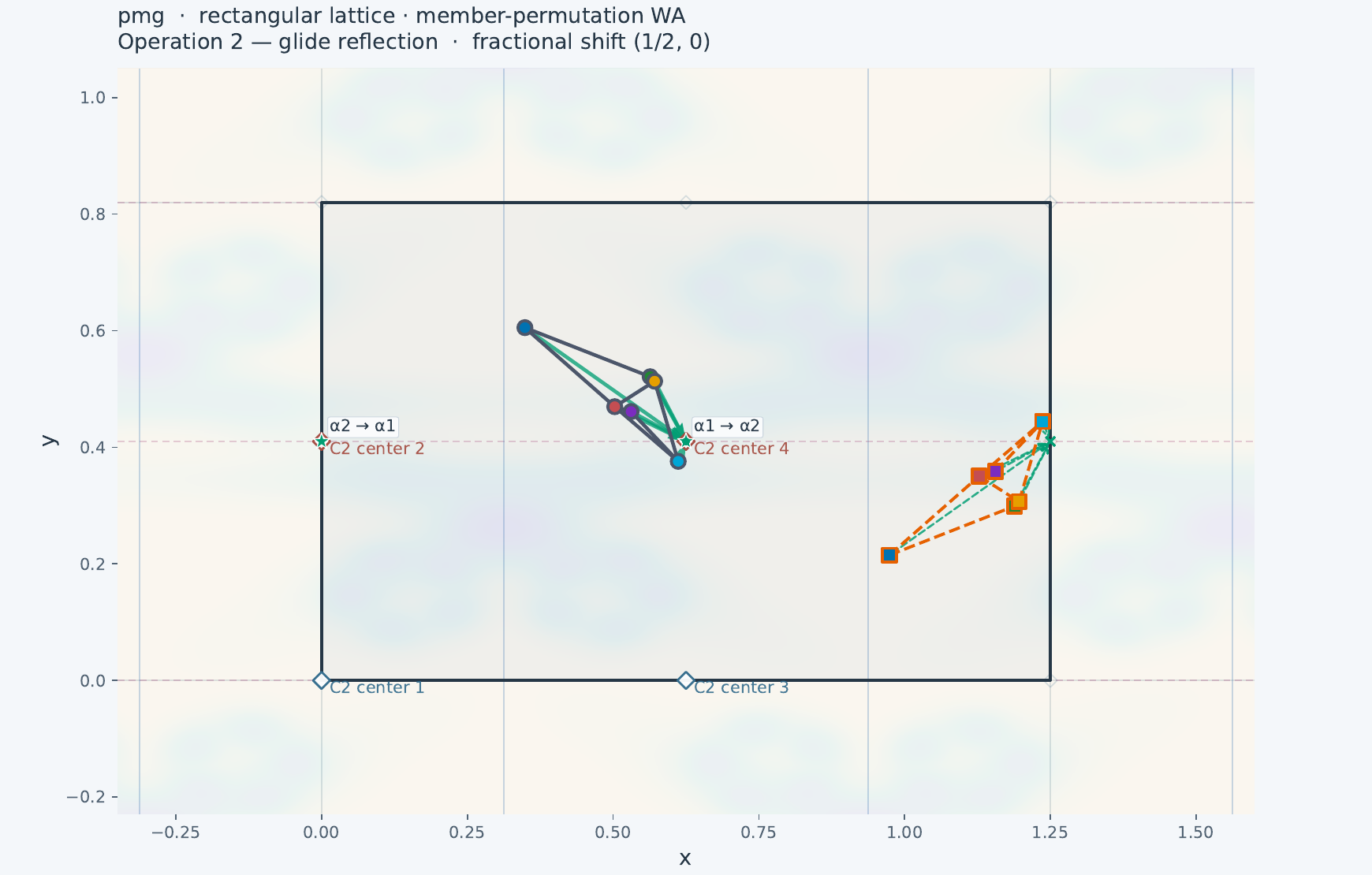} & \includegraphics[width=\linewidth,height=0.22\textheight,keepaspectratio]{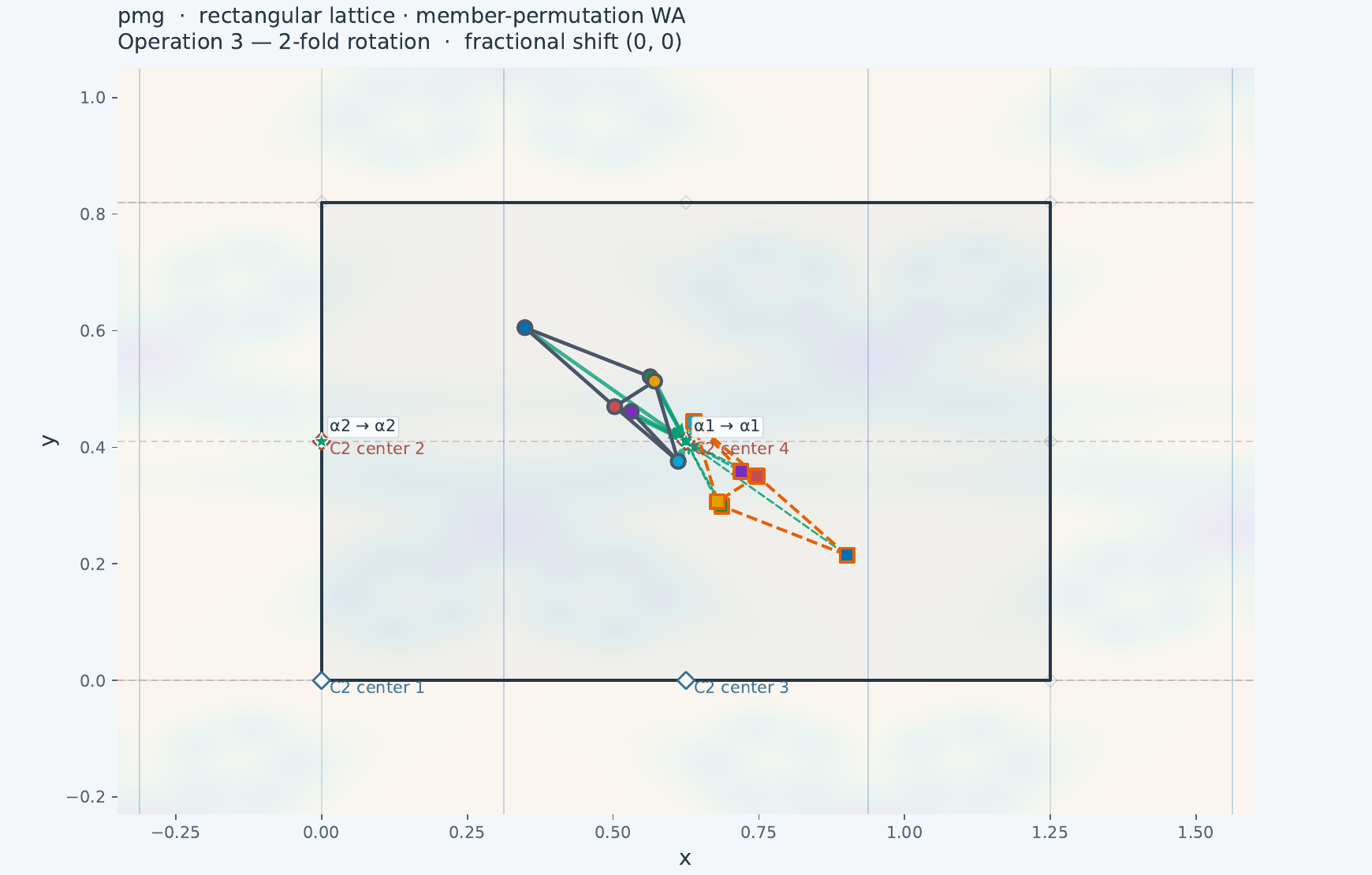} \\[2mm]
\includegraphics[width=\linewidth,height=0.22\textheight,keepaspectratio]{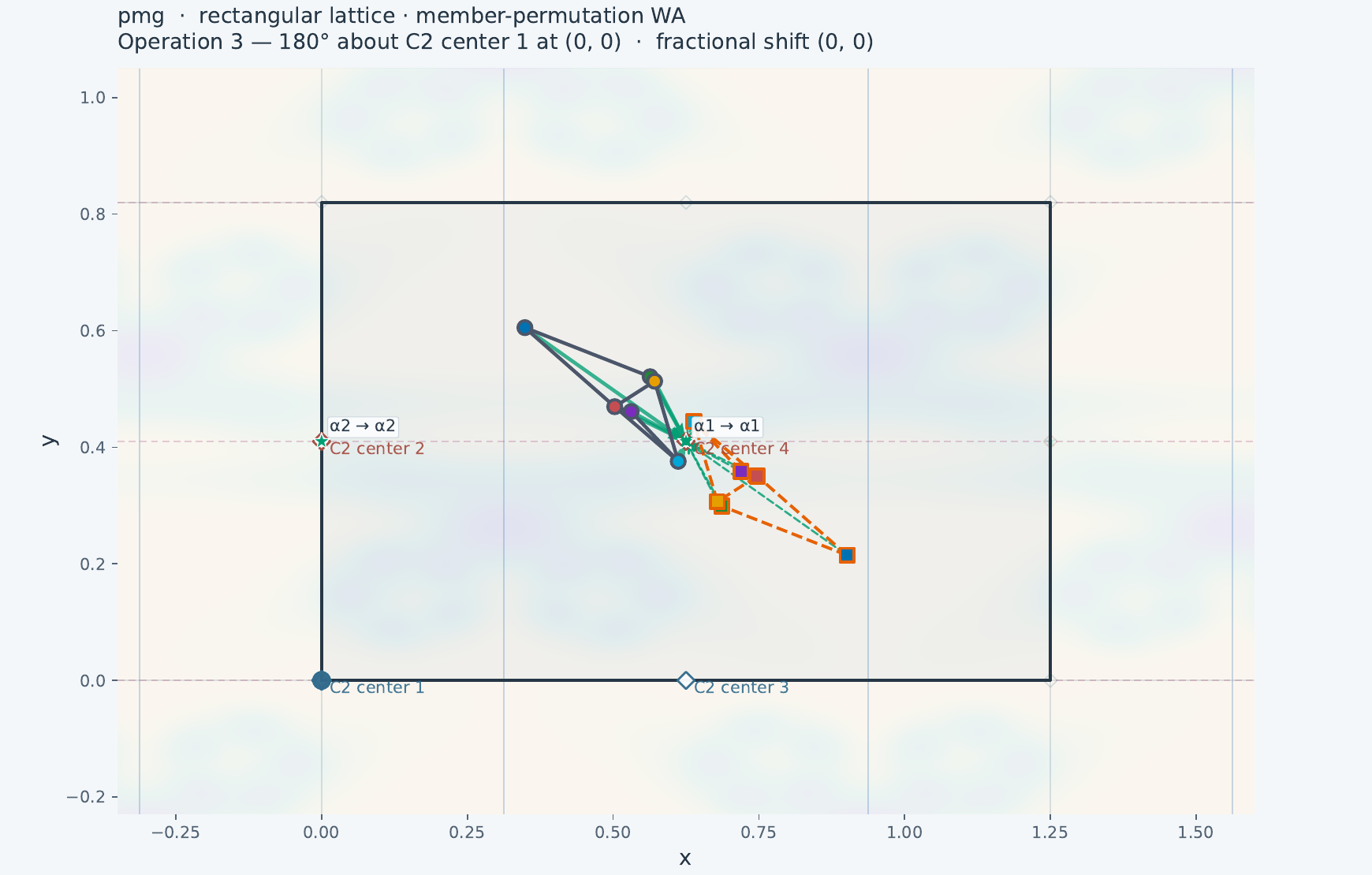} & \includegraphics[width=\linewidth,height=0.22\textheight,keepaspectratio]{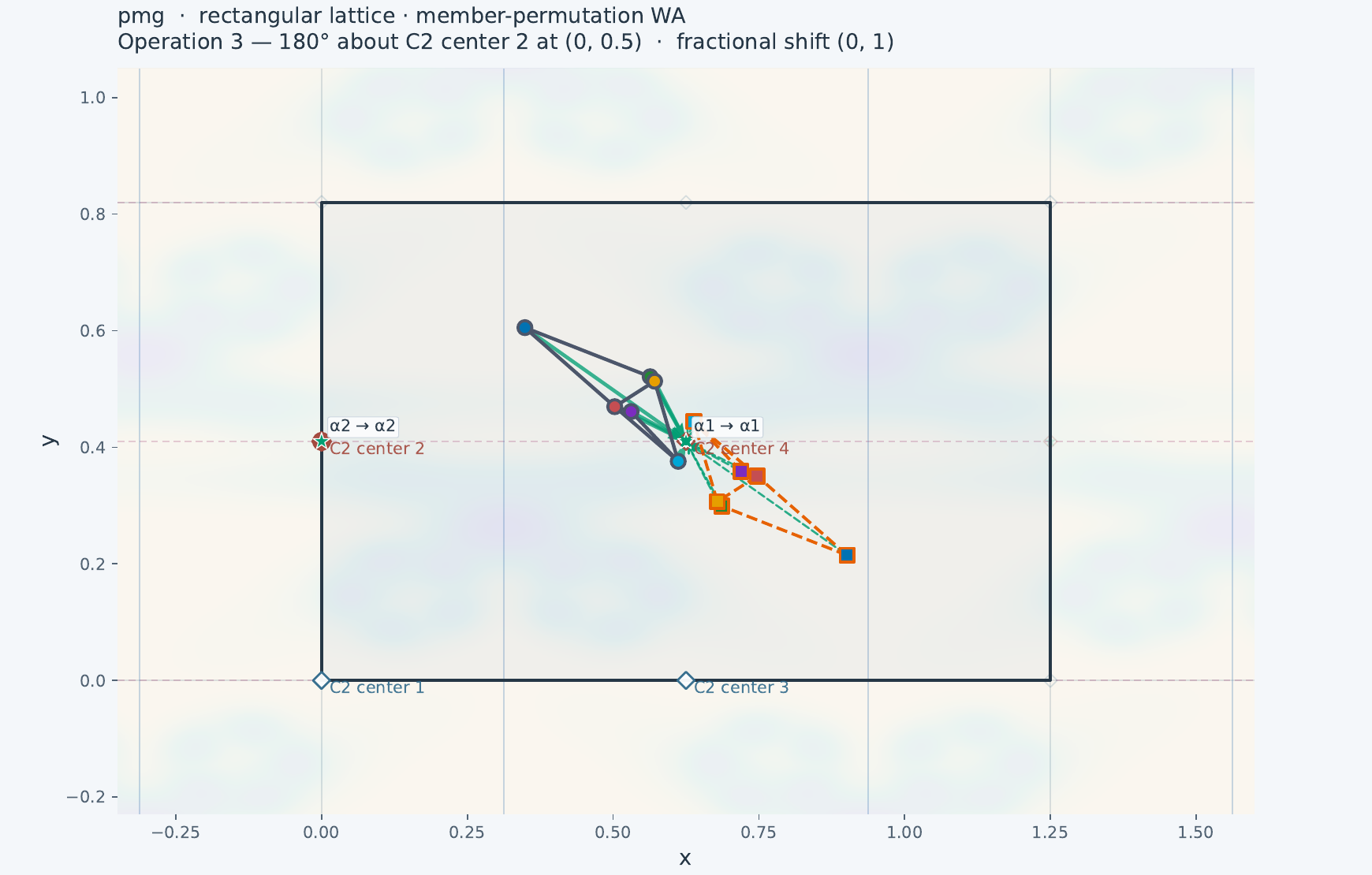} \\[2mm]
\includegraphics[width=\linewidth,height=0.22\textheight,keepaspectratio]{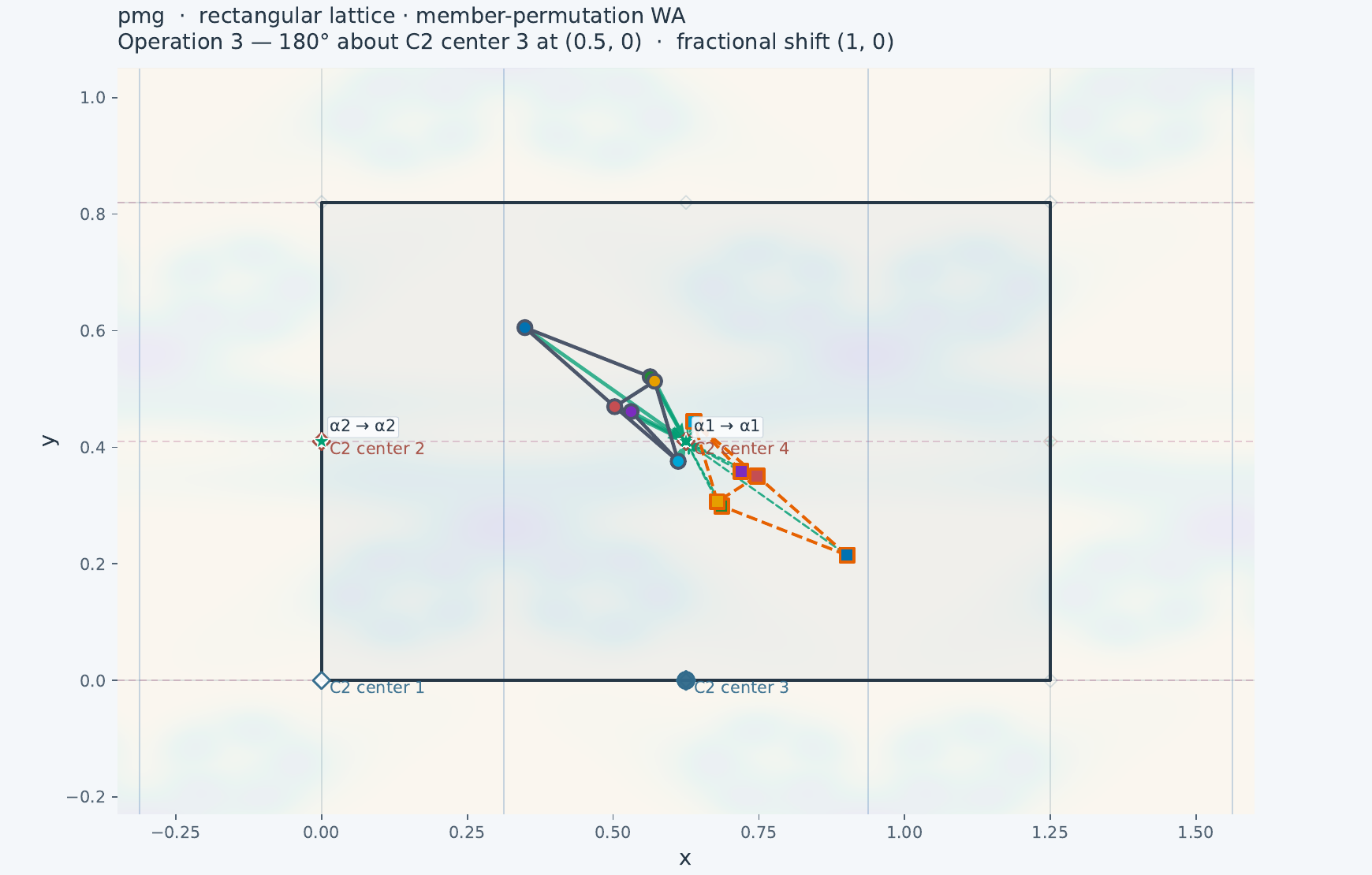} & \includegraphics[width=\linewidth,height=0.22\textheight,keepaspectratio]{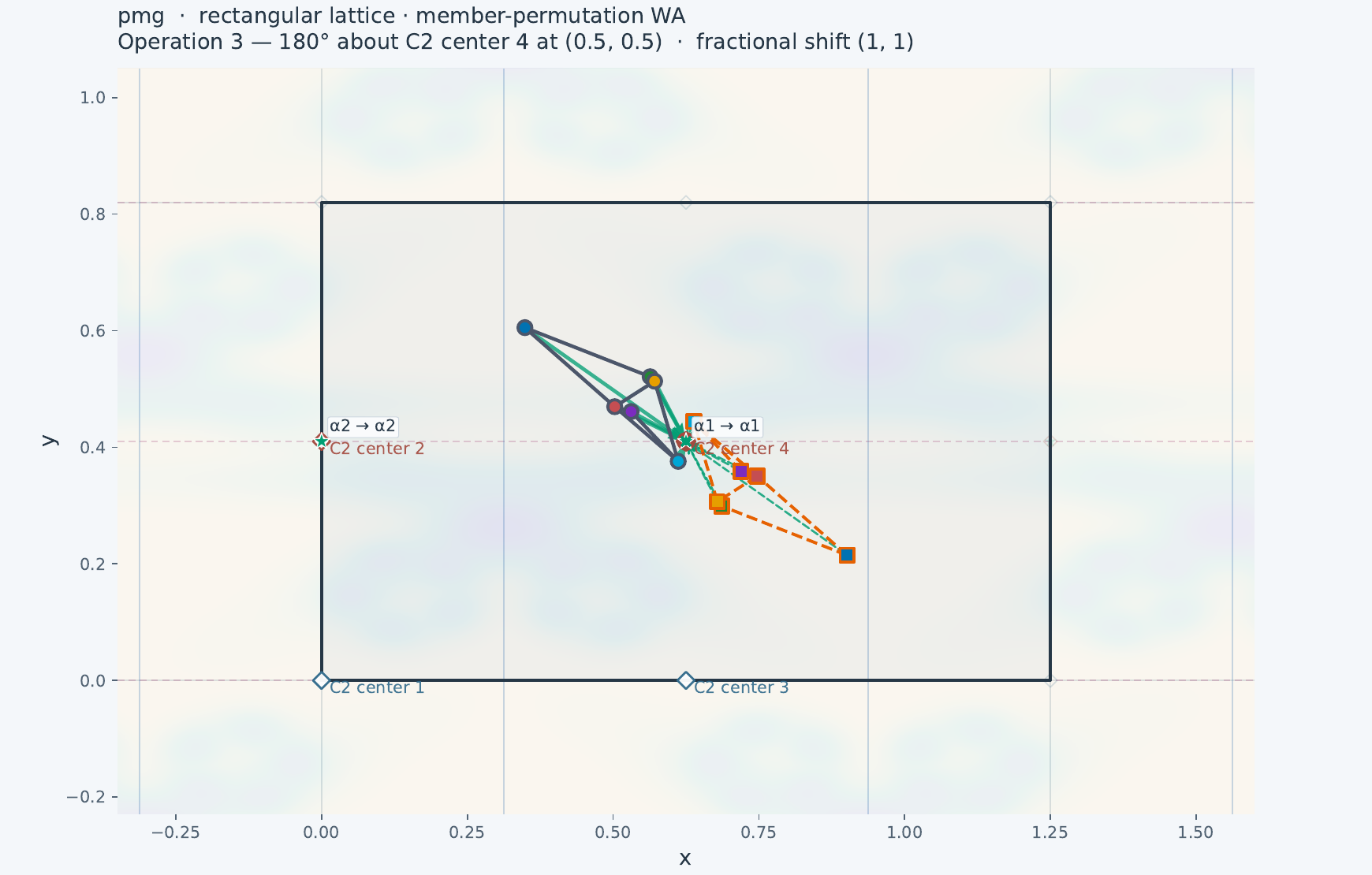} \\[2mm]
\end{longtable}
\Needspace{218pt}
\begin{longtable}{@{}>{\centering\arraybackslash}p{0.475\linewidth}>{\centering\arraybackslash}p{0.475\linewidth}@{}}
\caption[Wallpaper group \texttt{pgg}]{Wallpaper group \texttt{pgg}, {\small WA is pointing to C2-A at $q_{1}=(\frac{1}{2},\frac{1}{2}),\; q_{2}=(0,0)$}}\label{tab:wa-actions-pgg}\\[1mm]
\multicolumn{2}{c}{\includegraphics[width=0.96\linewidth]{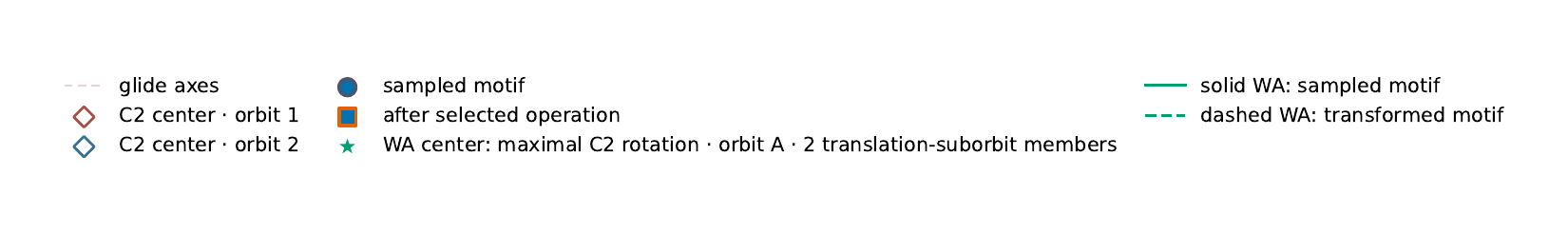}} \\[1mm]
\endfirsthead
\caption[]{Wallpaper group \texttt{pgg} (continued)}\\[1mm]
\endhead
\includegraphics[width=\linewidth,height=0.22\textheight,keepaspectratio]{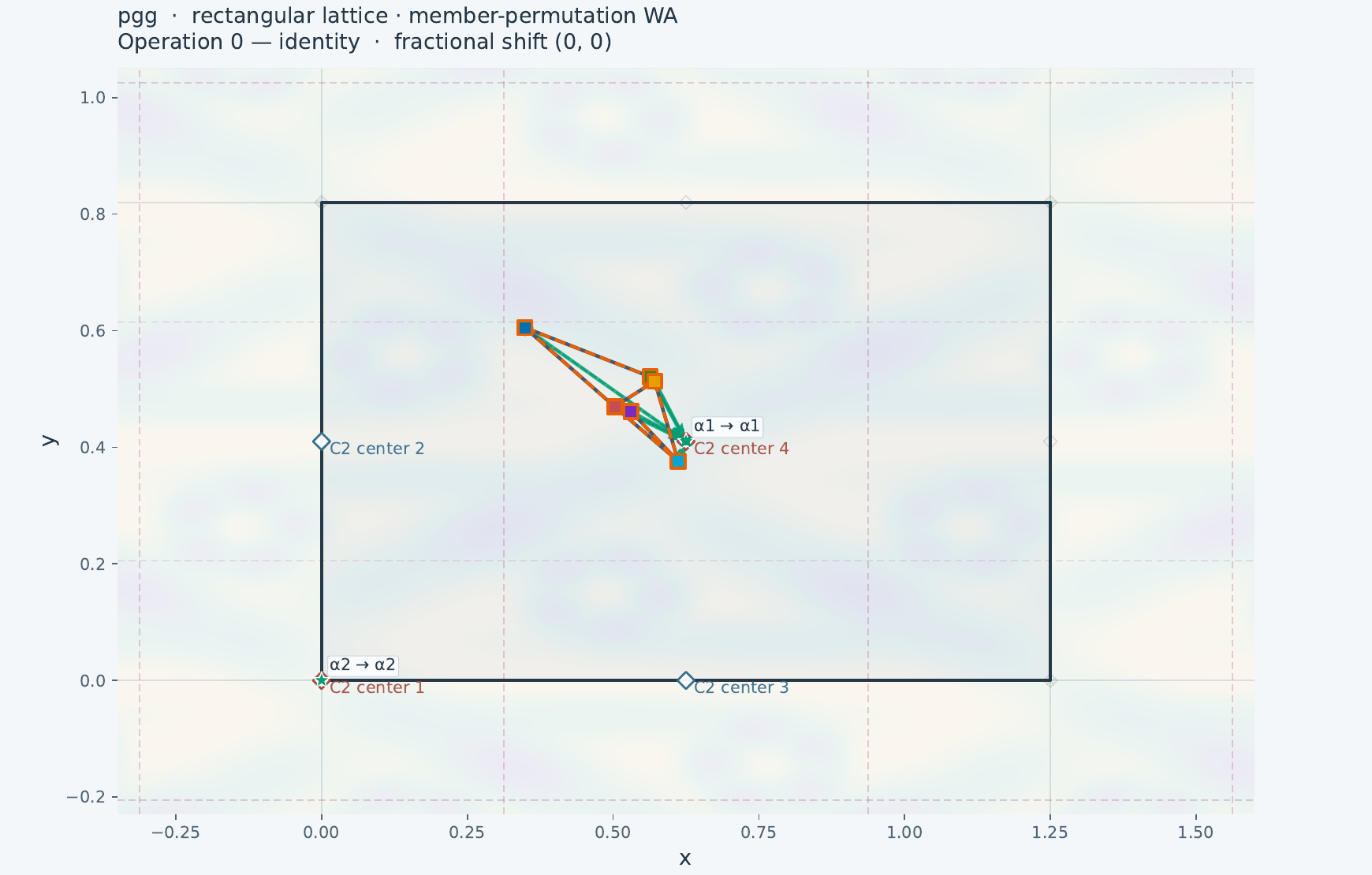} & \includegraphics[width=\linewidth,height=0.22\textheight,keepaspectratio]{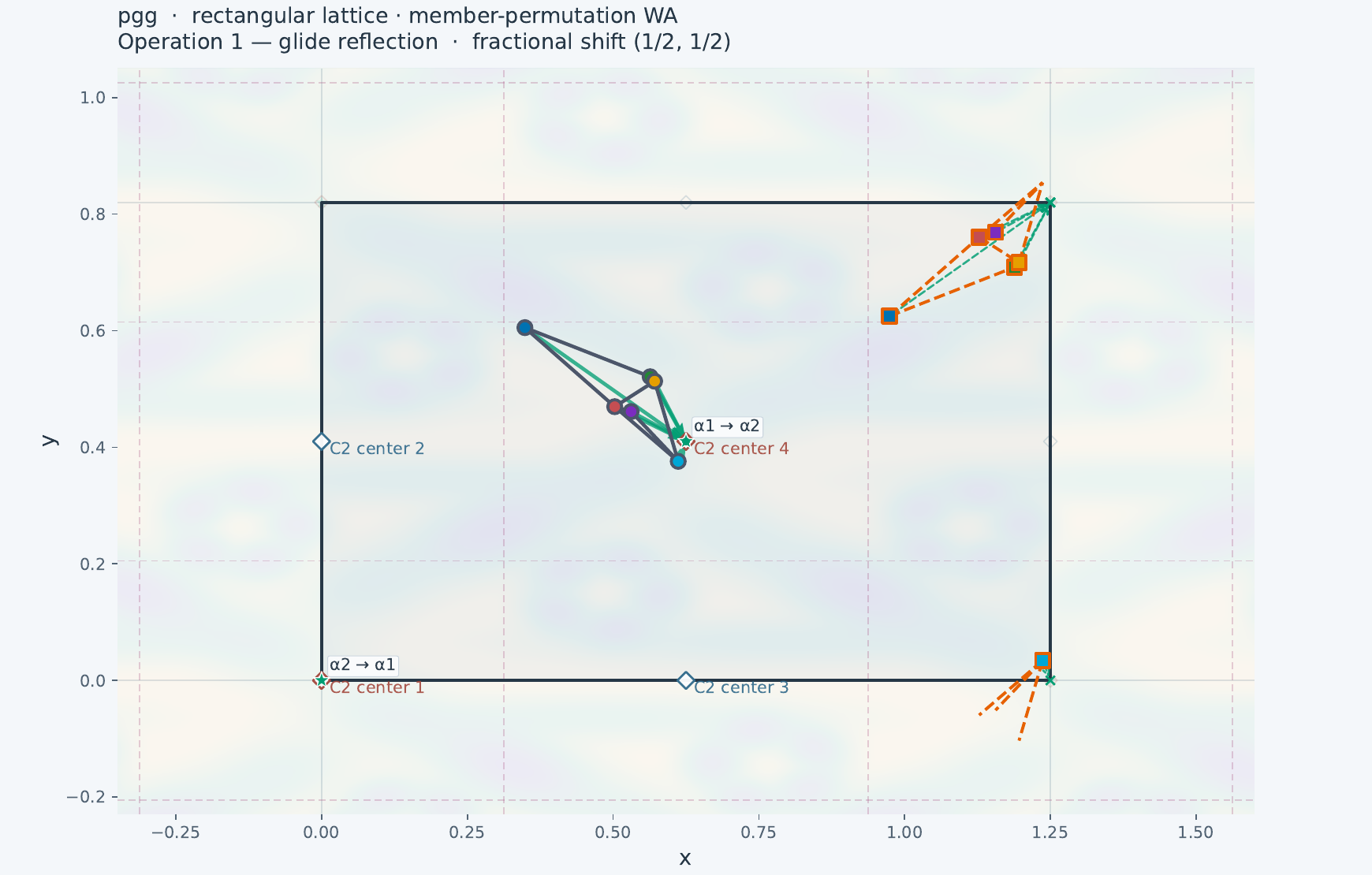} \\[2mm]
\includegraphics[width=\linewidth,height=0.22\textheight,keepaspectratio]{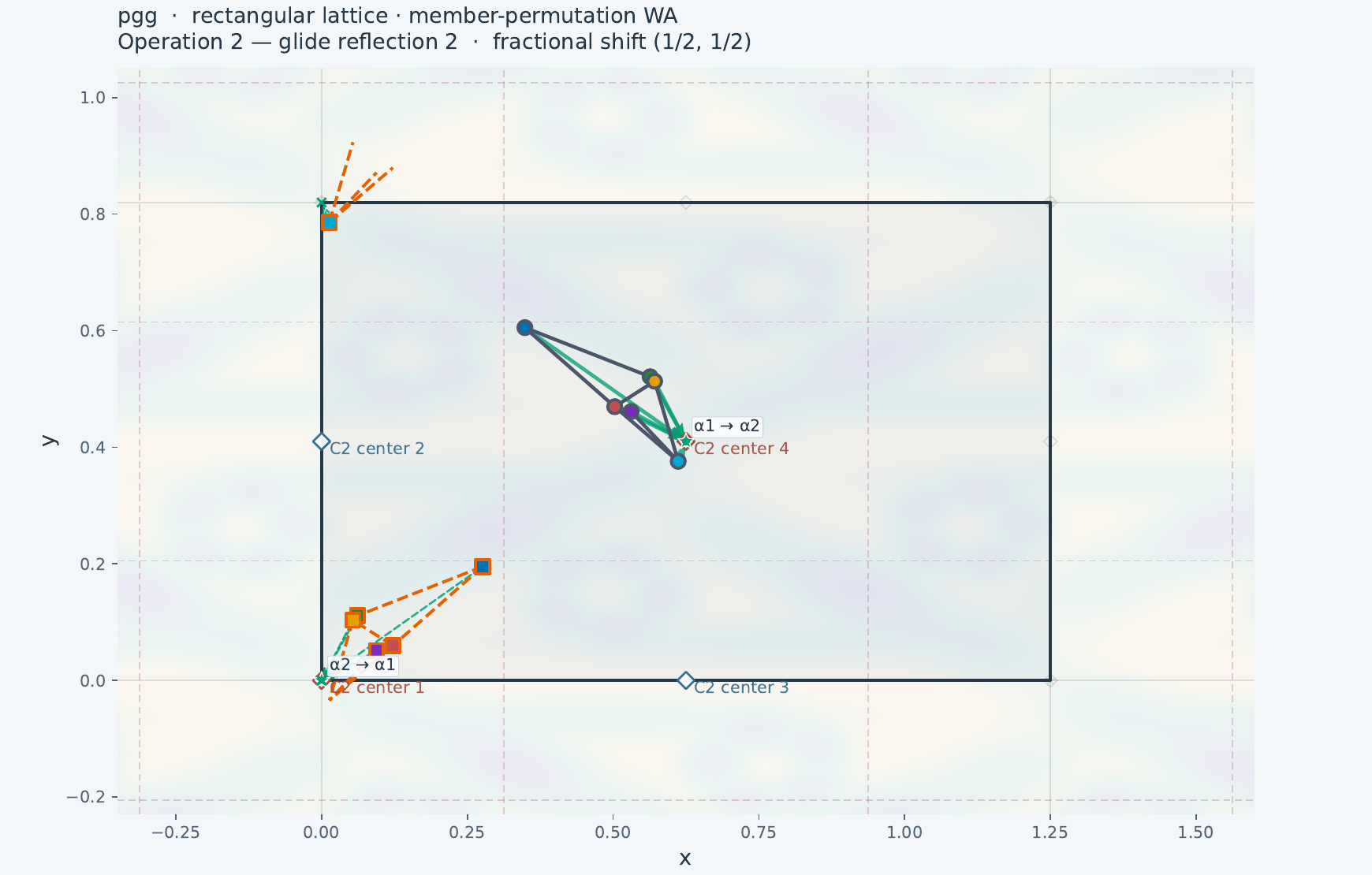} & \includegraphics[width=\linewidth,height=0.22\textheight,keepaspectratio]{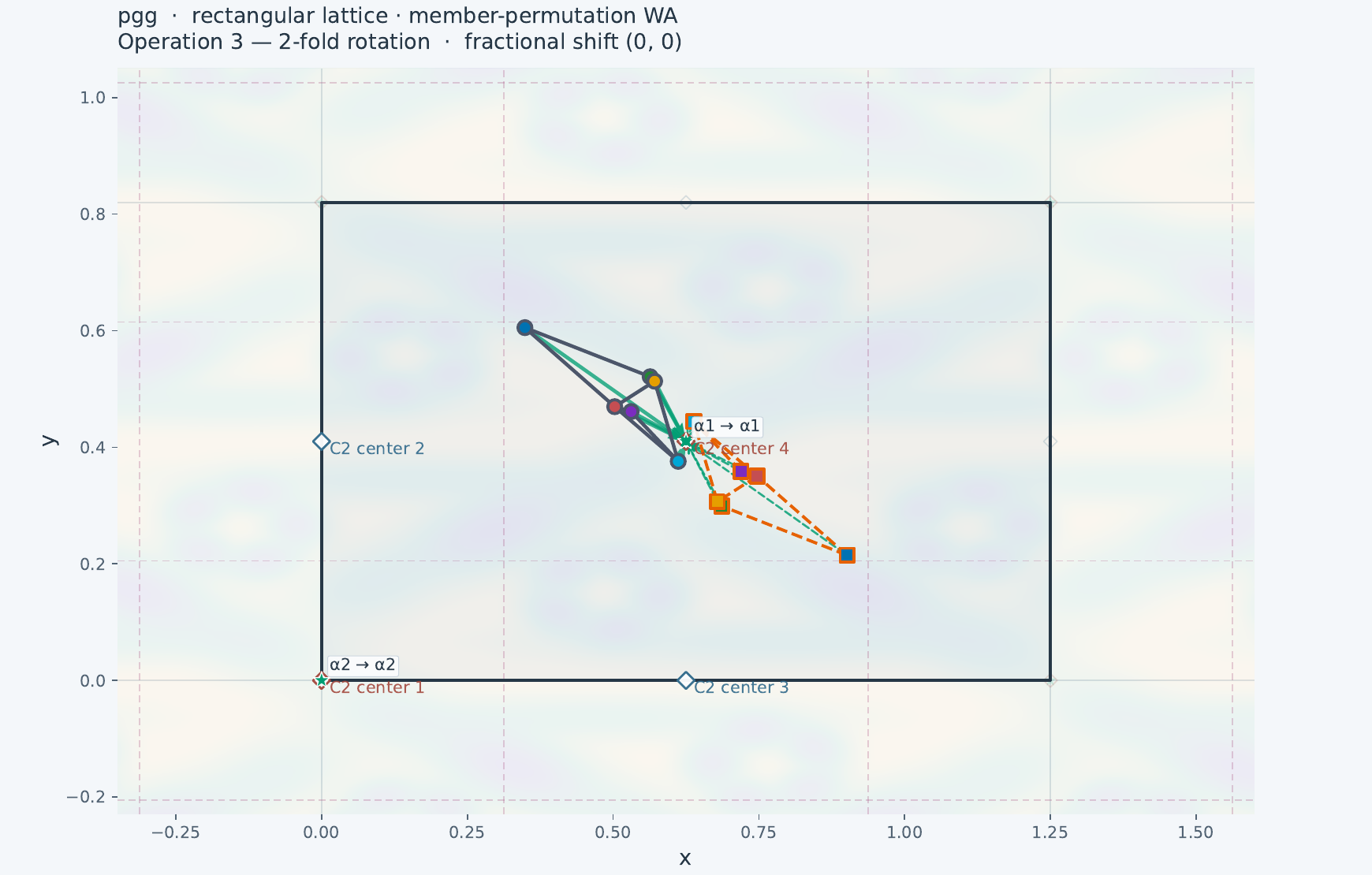} \\[2mm]
\includegraphics[width=\linewidth,height=0.22\textheight,keepaspectratio]{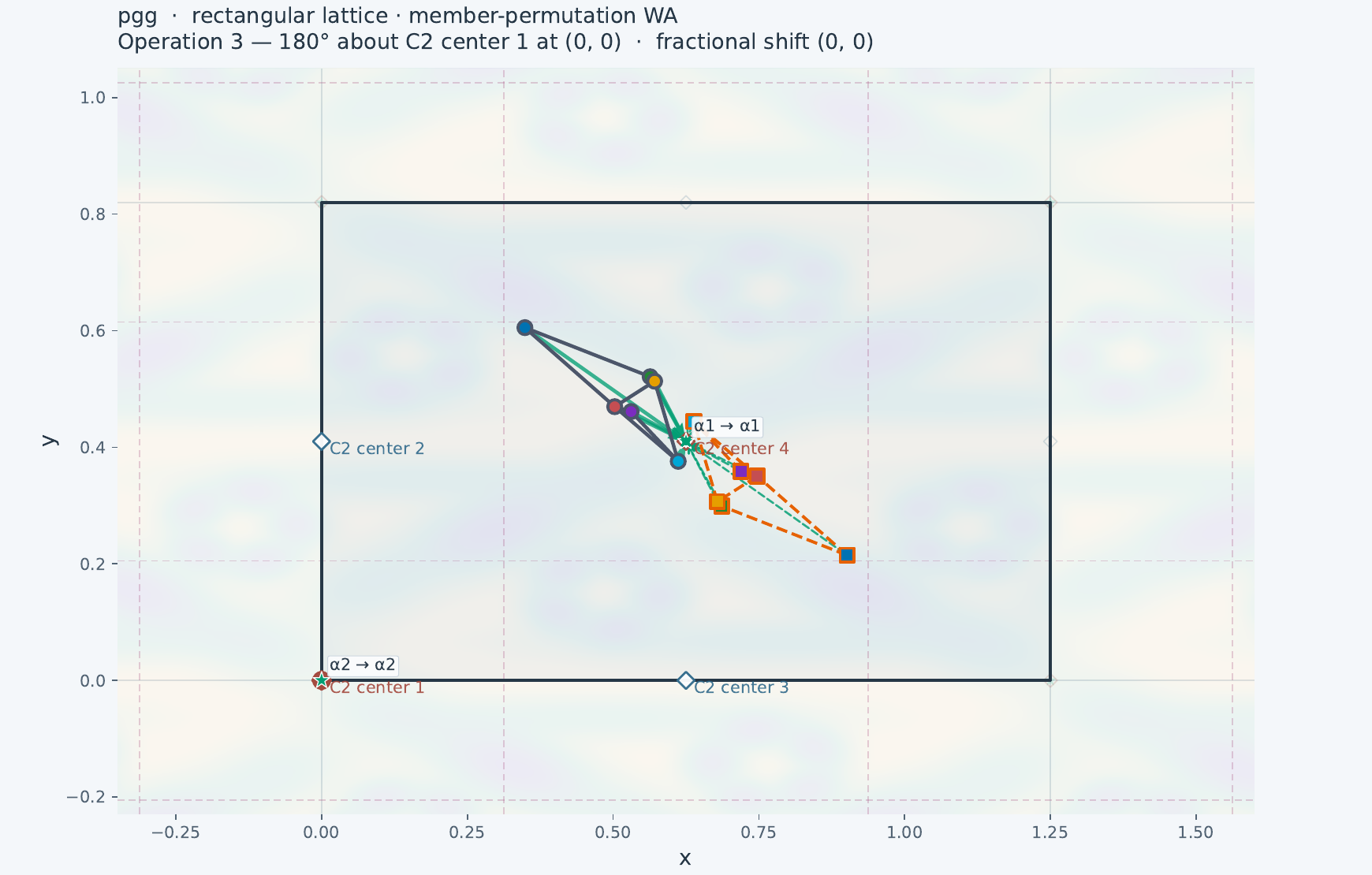} & \includegraphics[width=\linewidth,height=0.22\textheight,keepaspectratio]{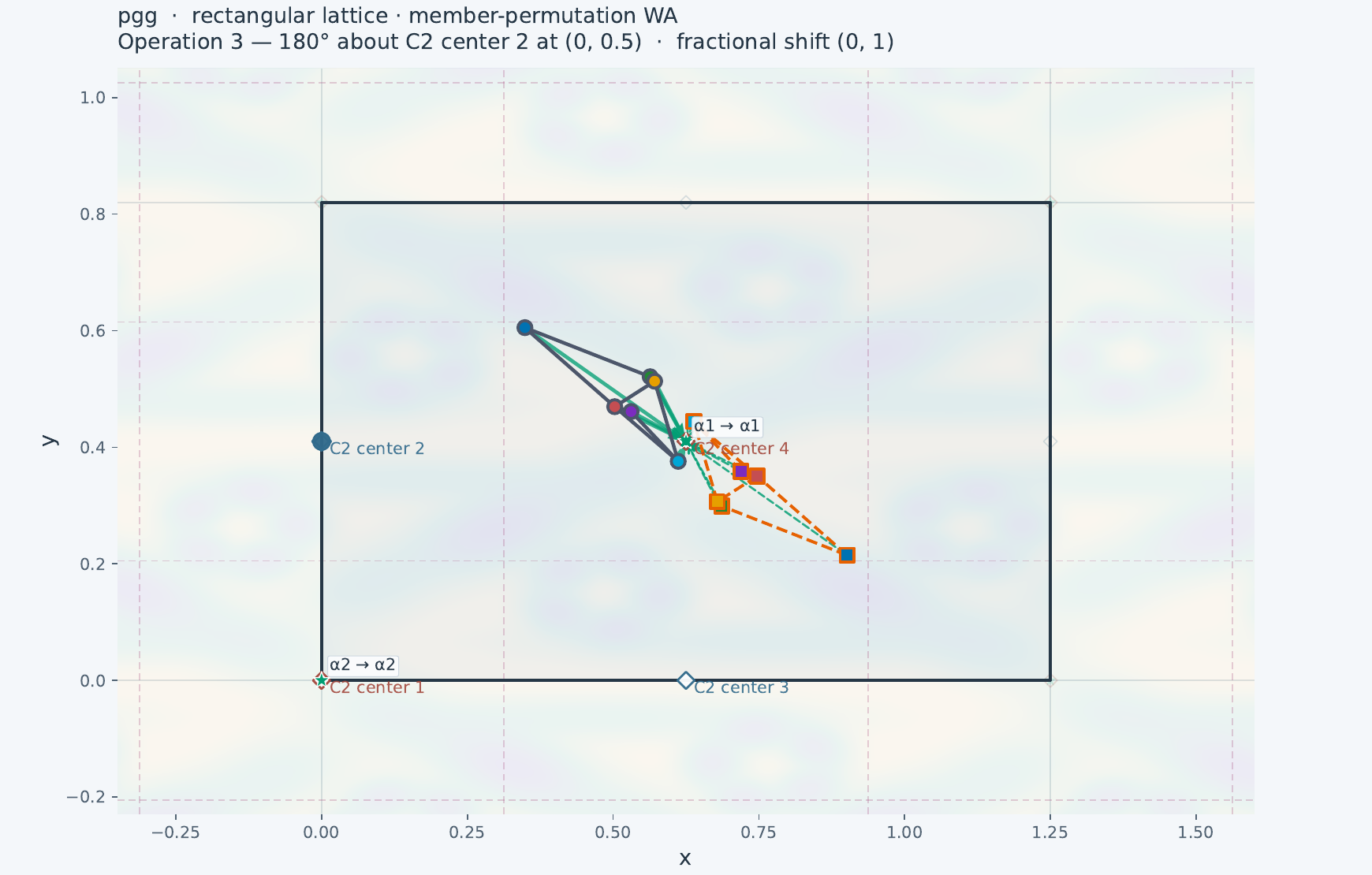} \\[2mm]
\includegraphics[width=\linewidth,height=0.22\textheight,keepaspectratio]{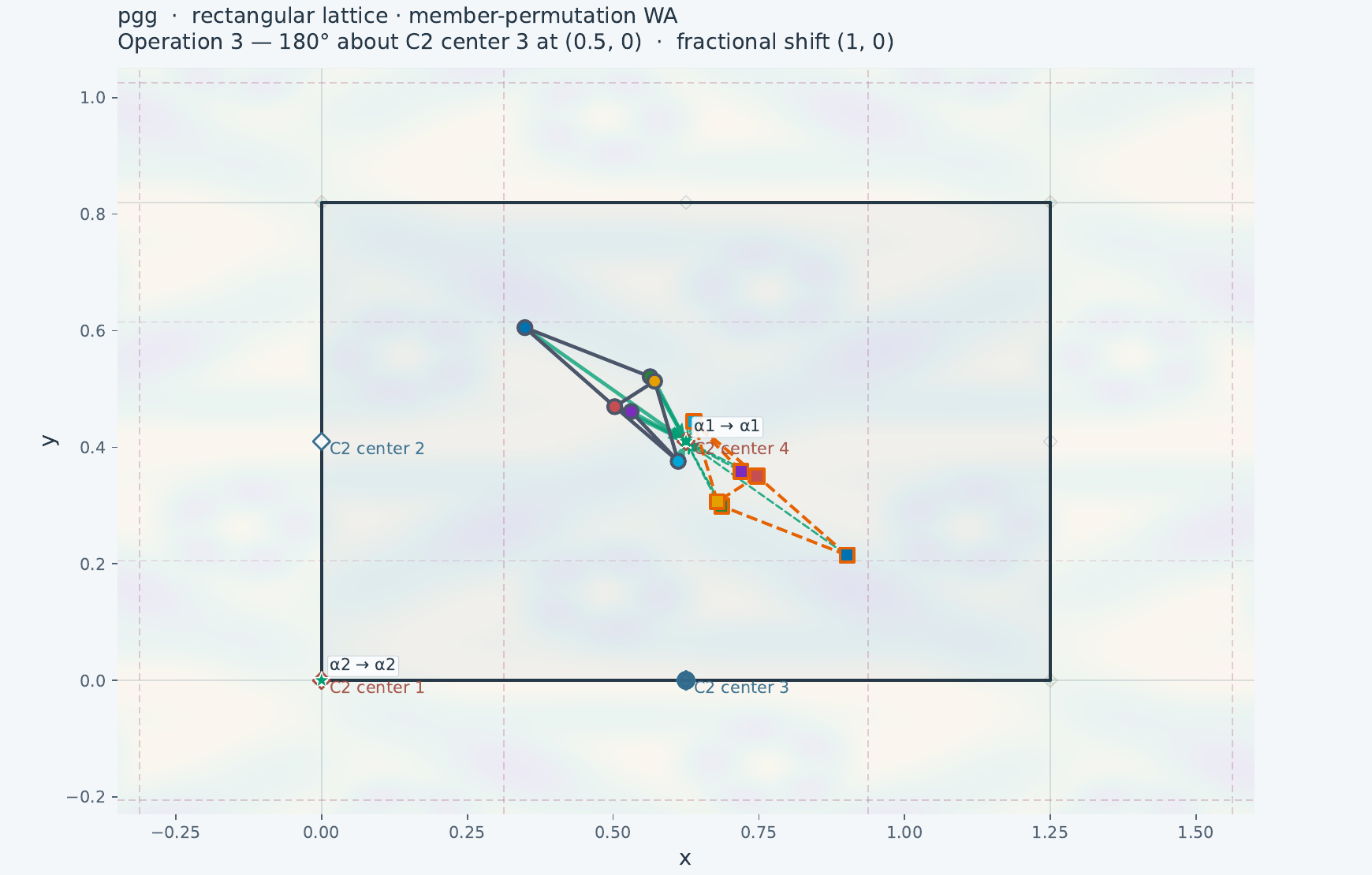} & \includegraphics[width=\linewidth,height=0.22\textheight,keepaspectratio]{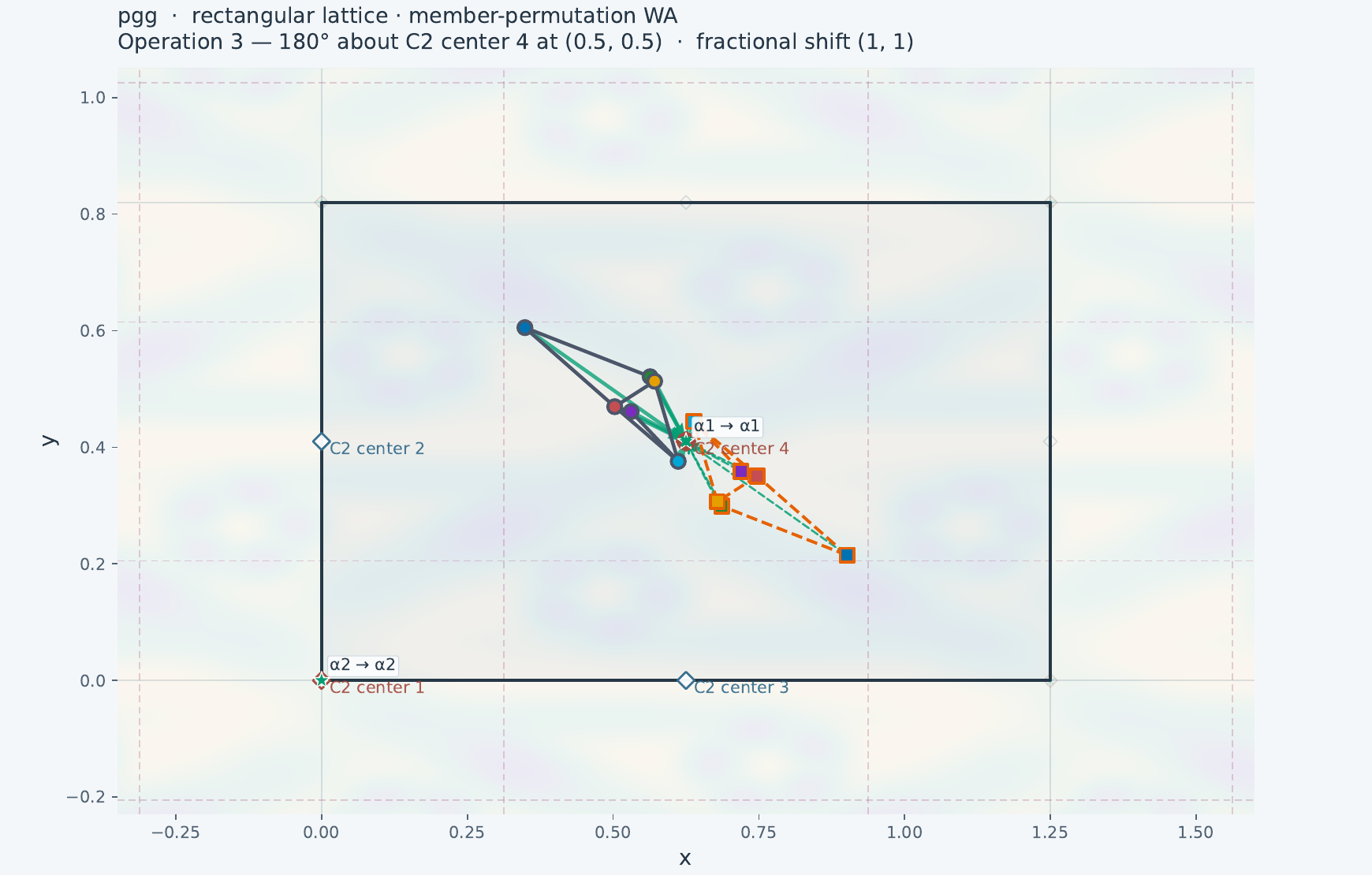} \\[2mm]
\end{longtable}
\Needspace{249pt}
\begin{longtable}{@{}>{\centering\arraybackslash}p{0.475\linewidth}>{\centering\arraybackslash}p{0.475\linewidth}@{}}
\caption[Wallpaper group \texttt{cmm}]{Wallpaper group \texttt{cmm}, {\small WA is pointing to C2-A at $q_{1}=(0,0)$}, {\small centered conventional cell shown, with the WA member multiplicity counted modulo primitive translations}}\label{tab:wa-actions-cmm}\\[1mm]
\multicolumn{2}{c}{\includegraphics[width=0.96\linewidth]{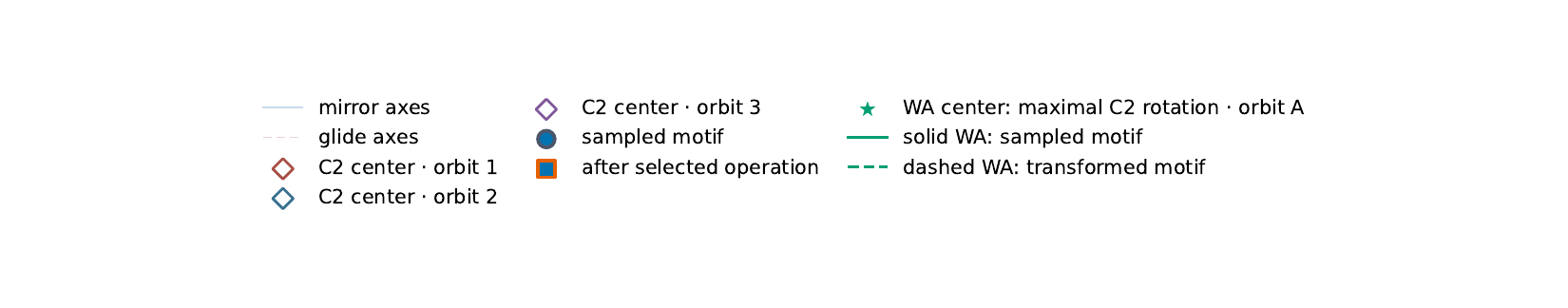}} \\[1mm]
\endfirsthead
\caption[]{Wallpaper group \texttt{cmm} (continued)}\\[1mm]
\endhead
\includegraphics[width=\linewidth,height=0.22\textheight,keepaspectratio]{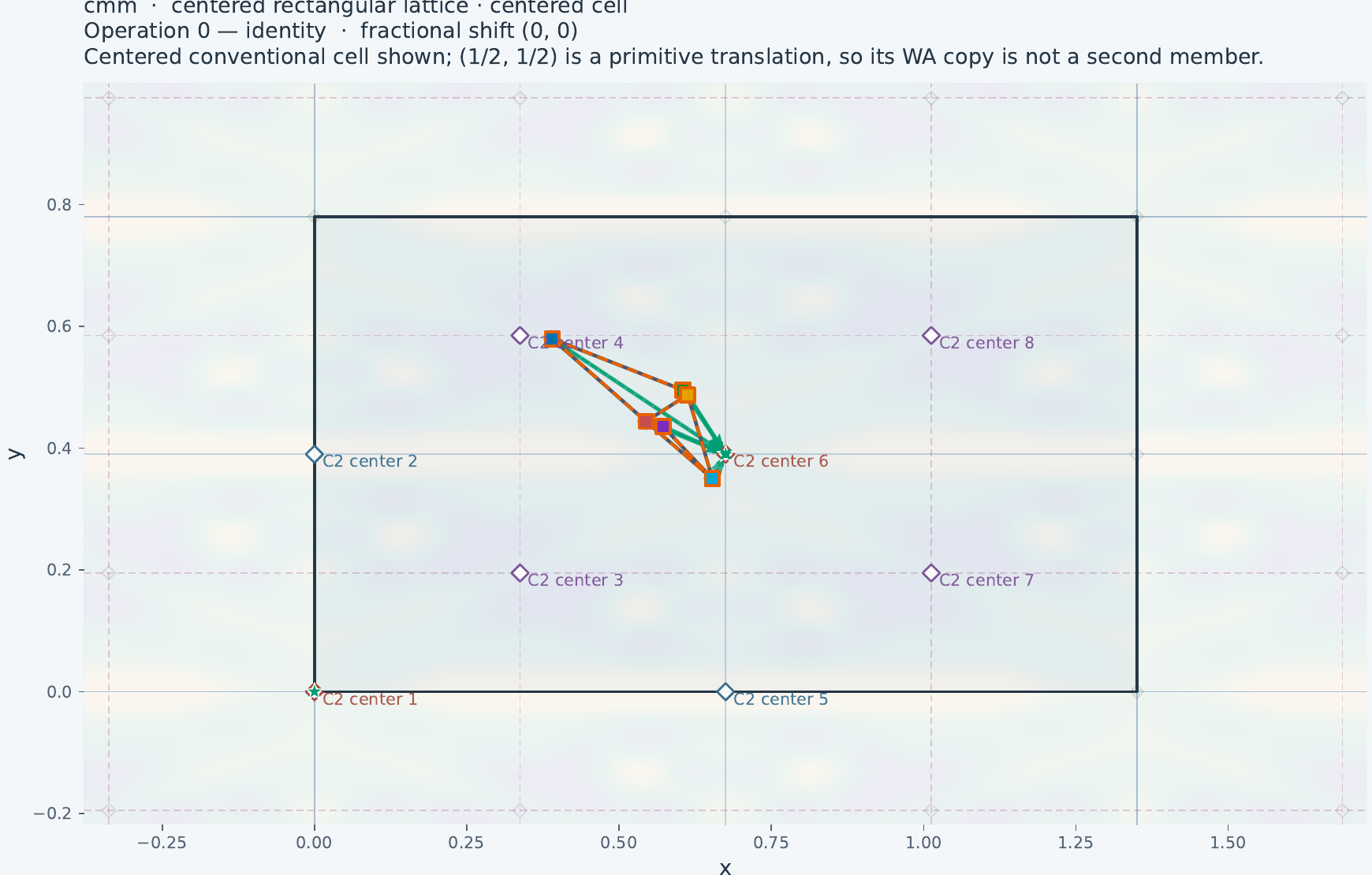} & \includegraphics[width=\linewidth,height=0.22\textheight,keepaspectratio]{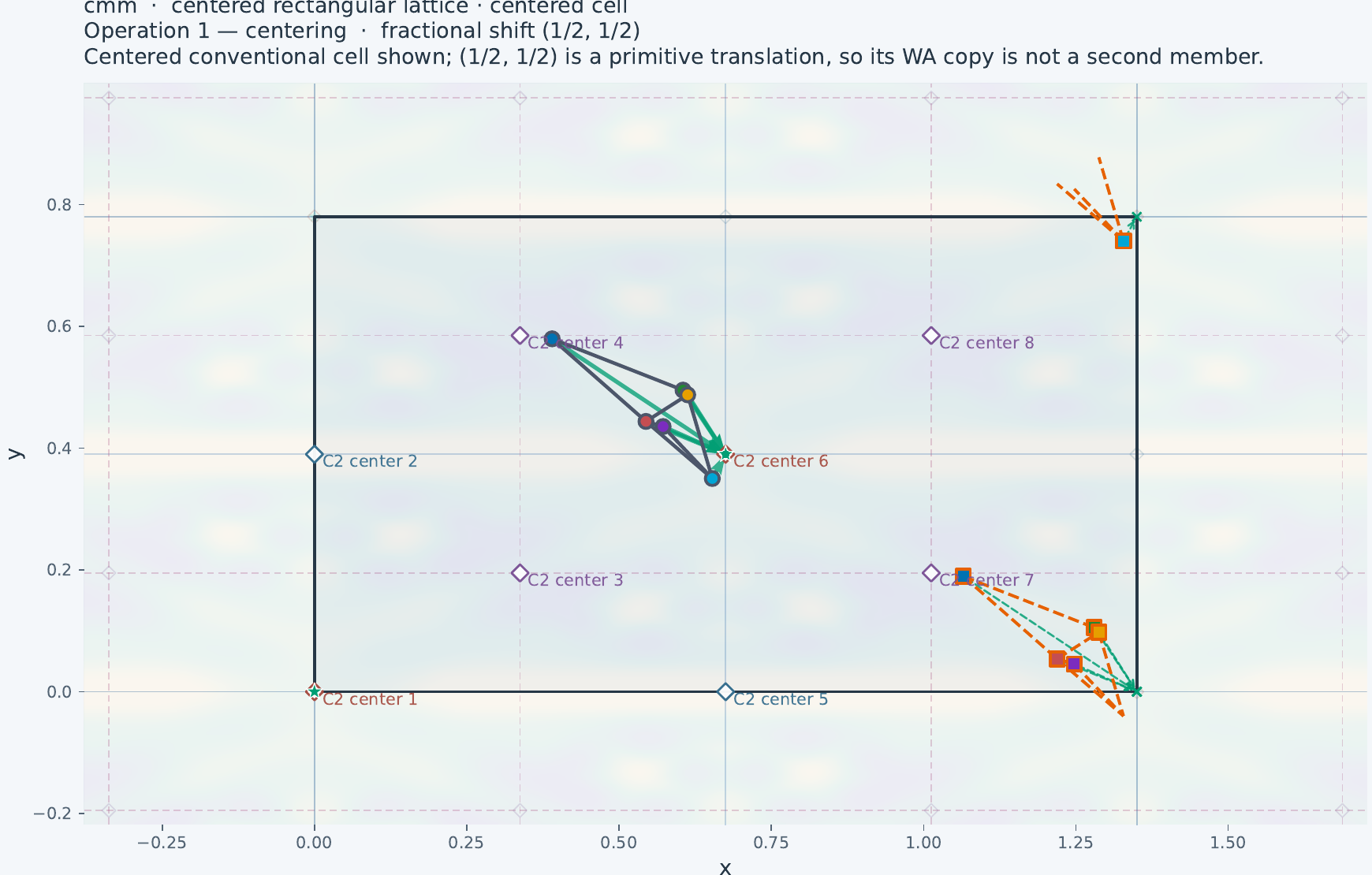} \\[2mm]
\includegraphics[width=\linewidth,height=0.22\textheight,keepaspectratio]{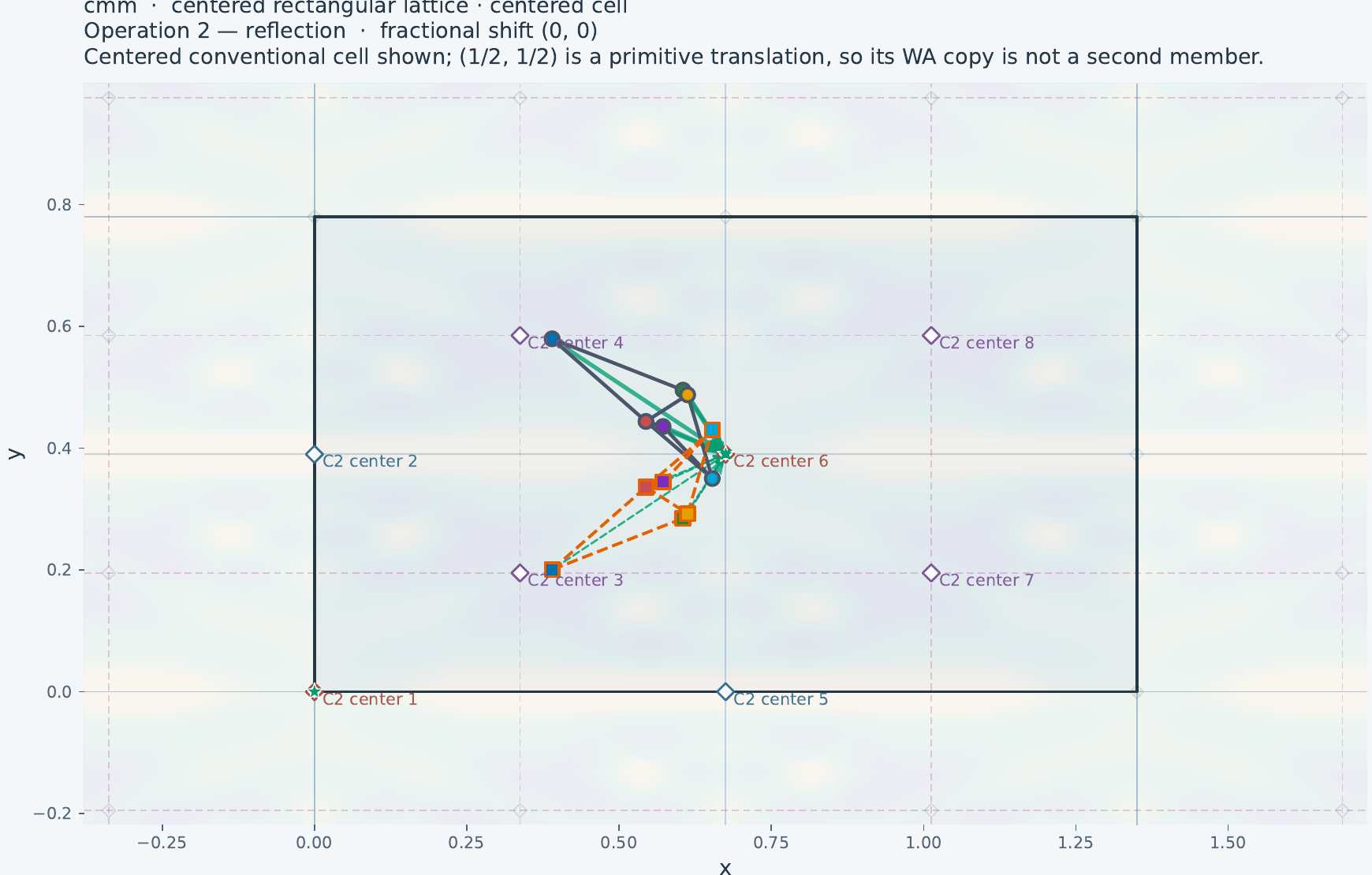} & \includegraphics[width=\linewidth,height=0.22\textheight,keepaspectratio]{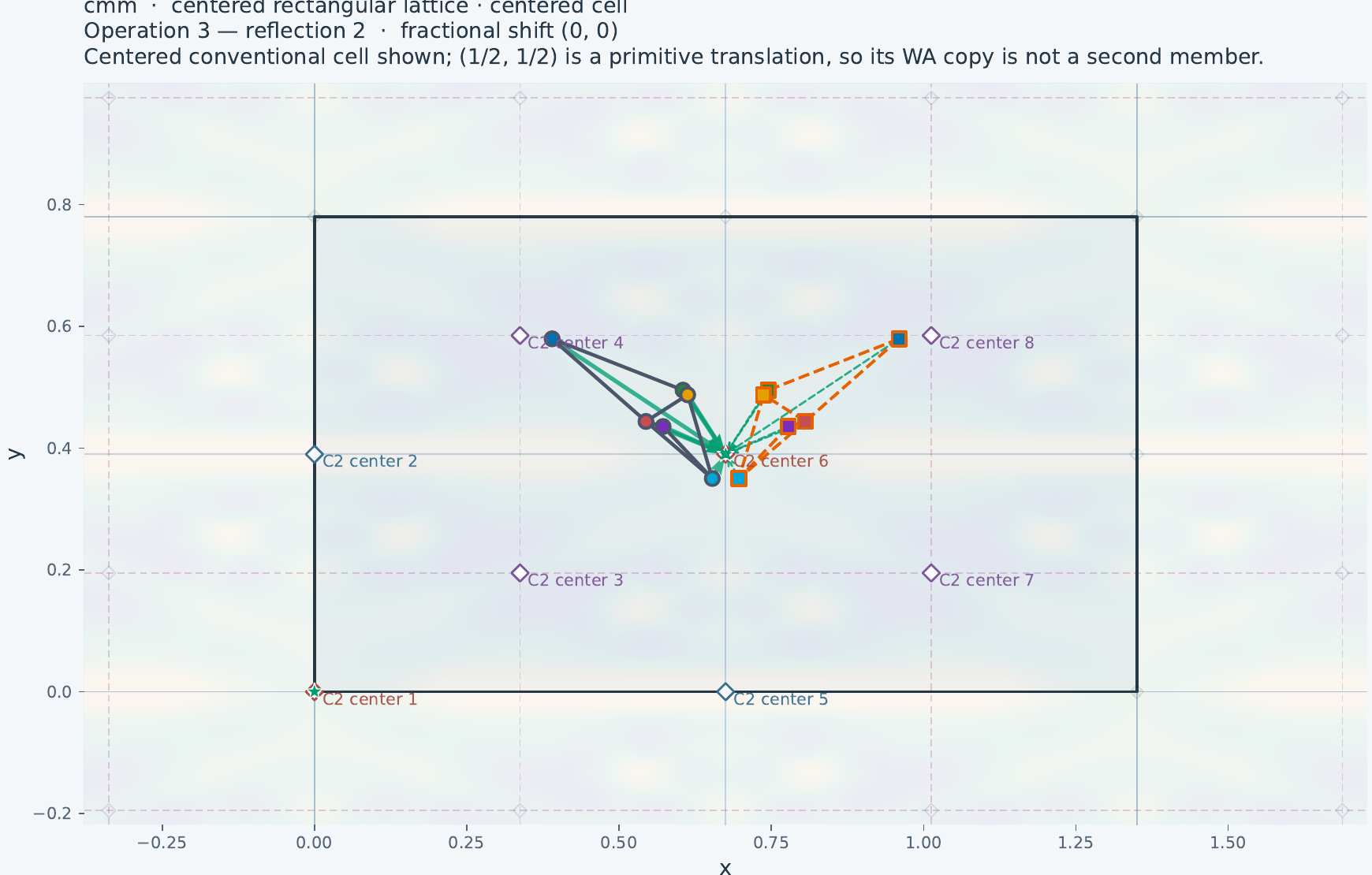} \\[2mm]
\includegraphics[width=\linewidth,height=0.22\textheight,keepaspectratio]{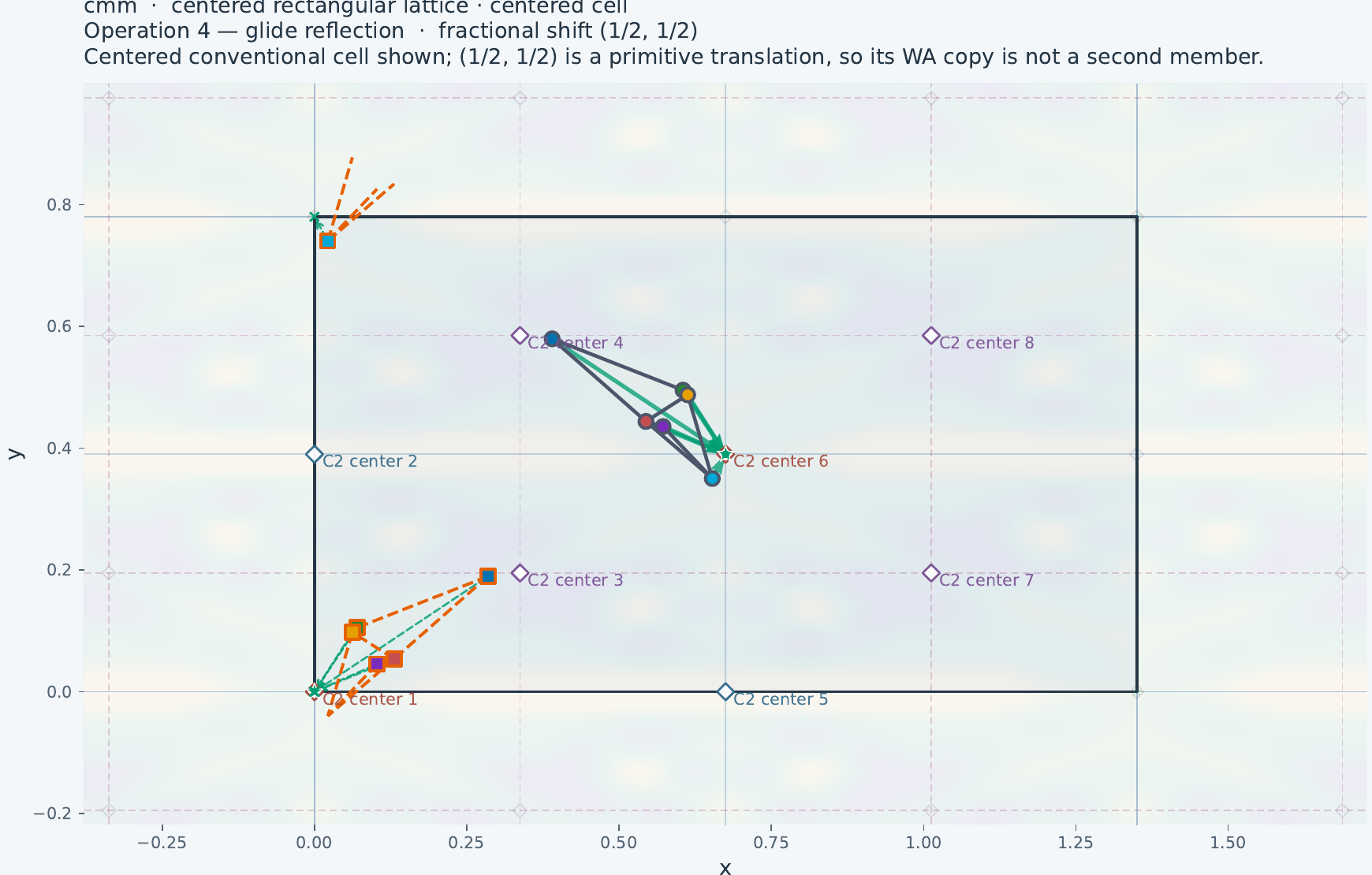} & \includegraphics[width=\linewidth,height=0.22\textheight,keepaspectratio]{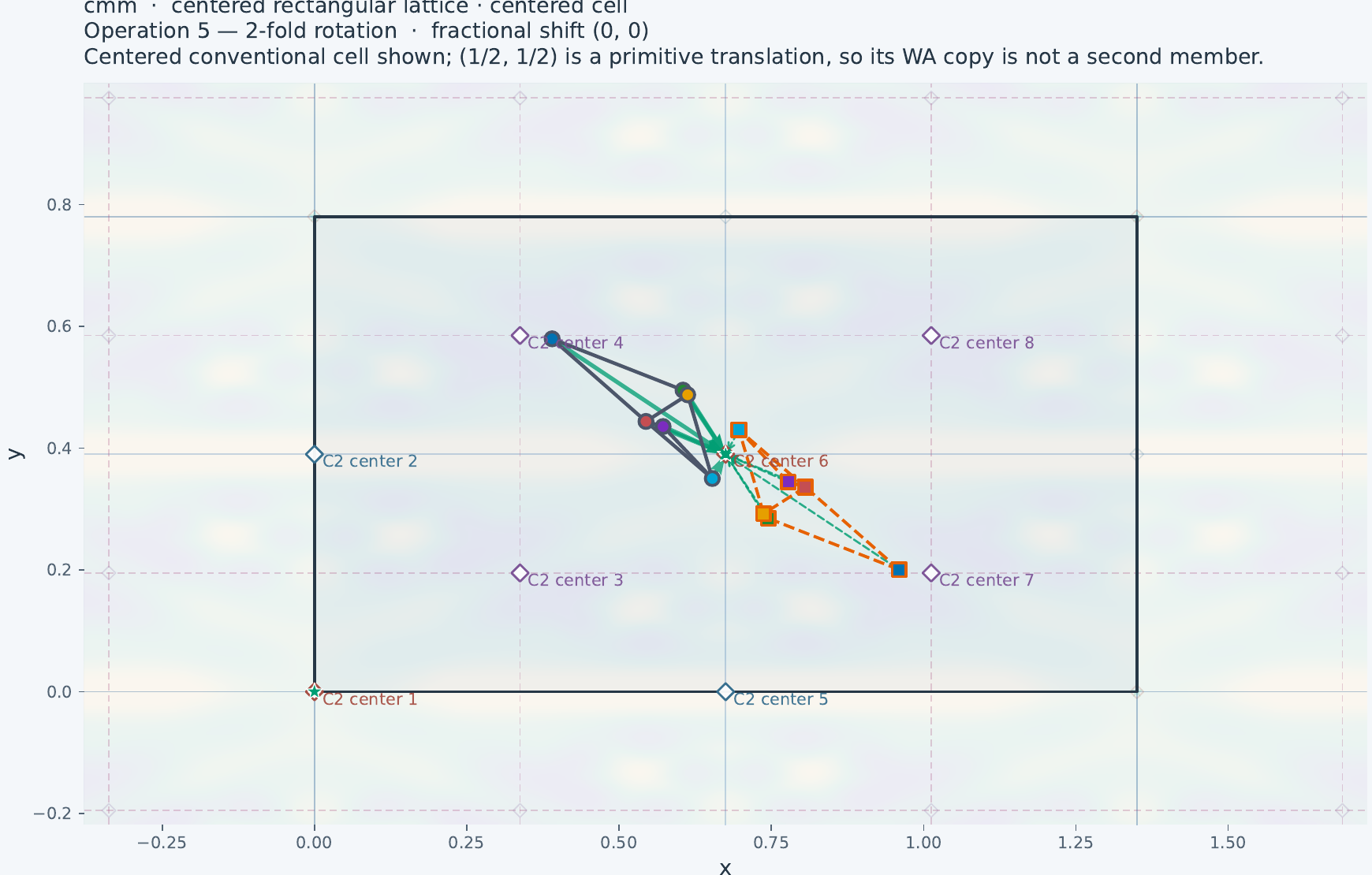} \\[2mm]
\includegraphics[width=\linewidth,height=0.22\textheight,keepaspectratio]{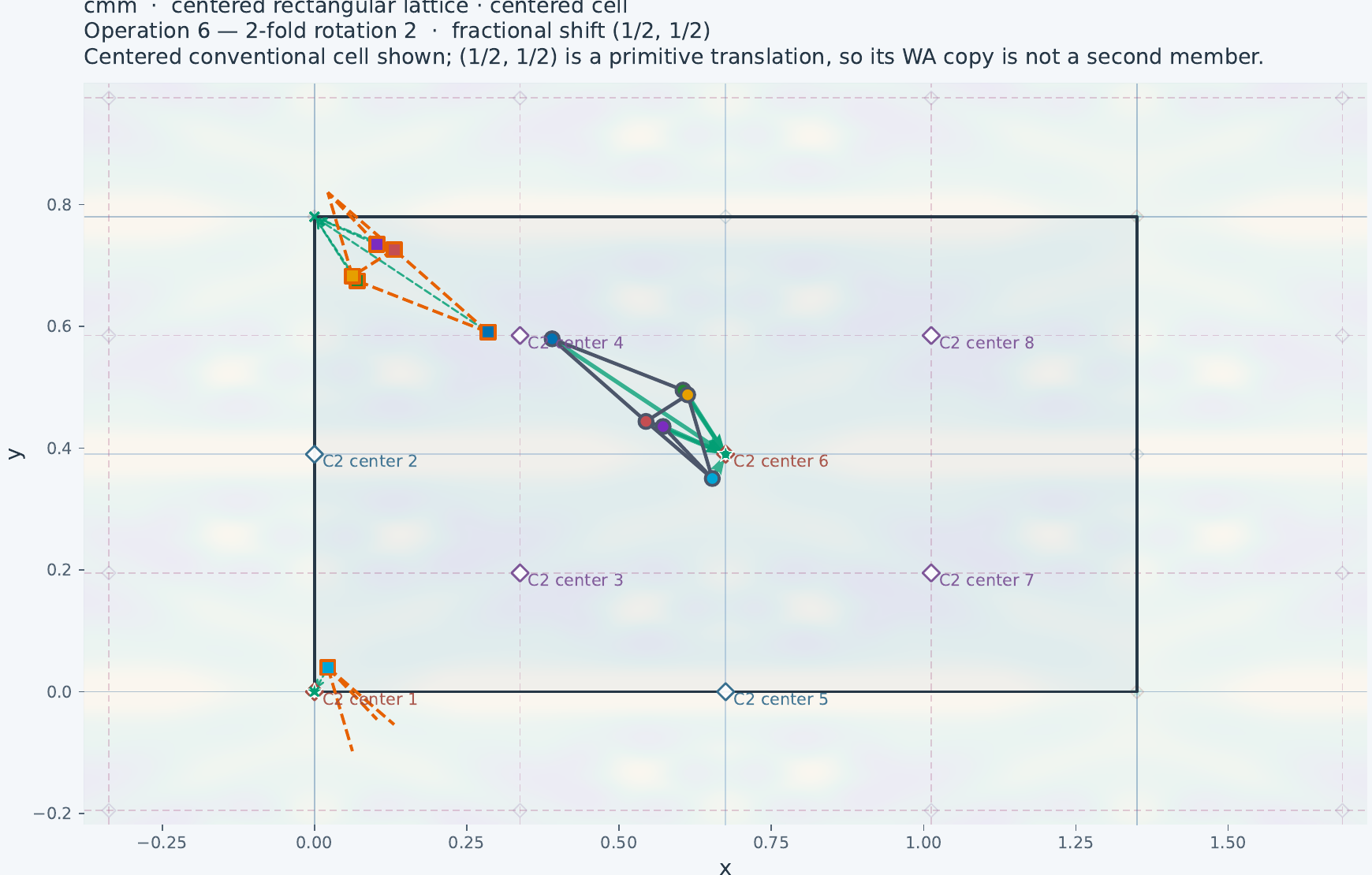} & \includegraphics[width=\linewidth,height=0.22\textheight,keepaspectratio]{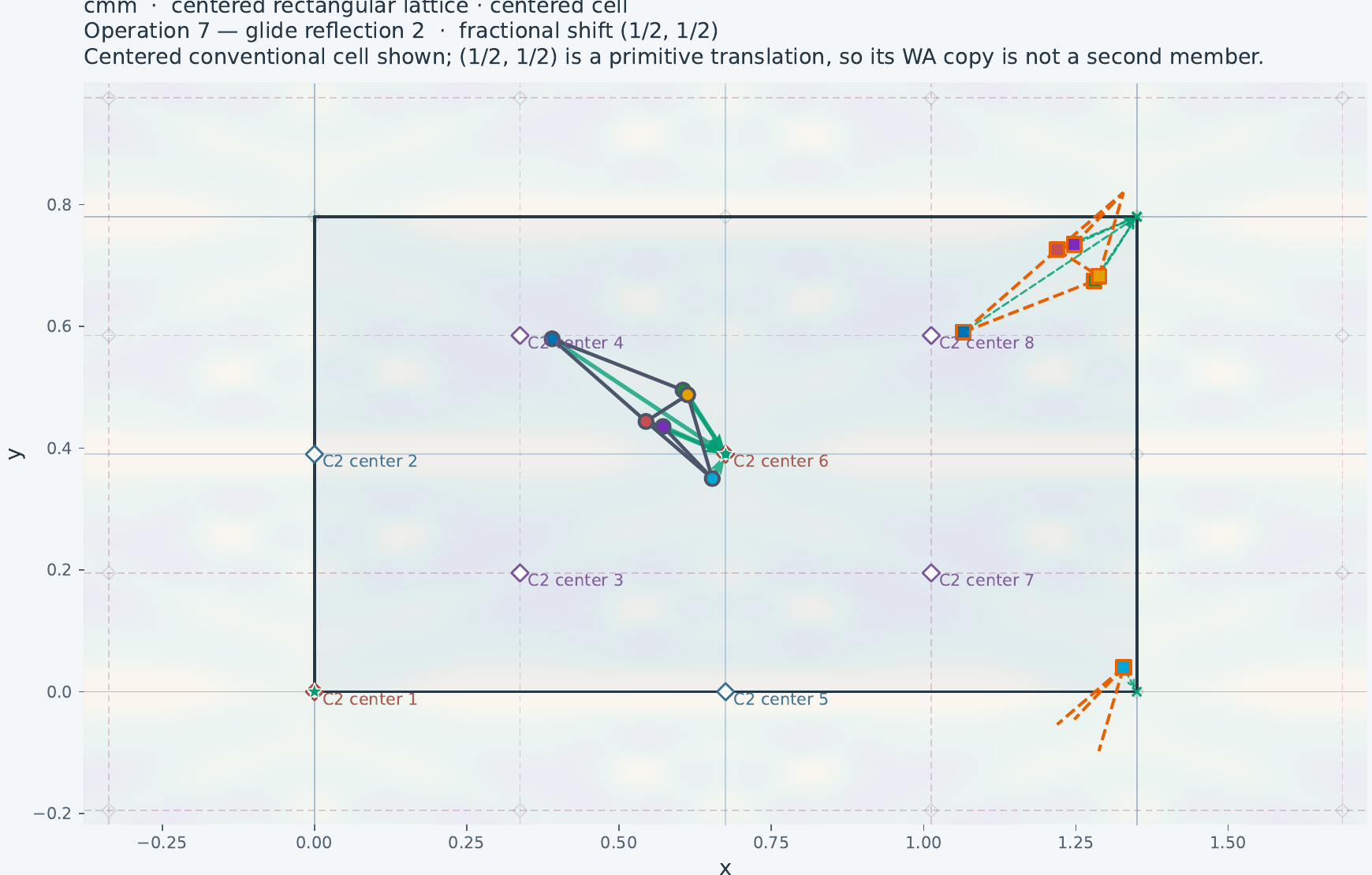} \\[2mm]
\includegraphics[width=\linewidth,height=0.22\textheight,keepaspectratio]{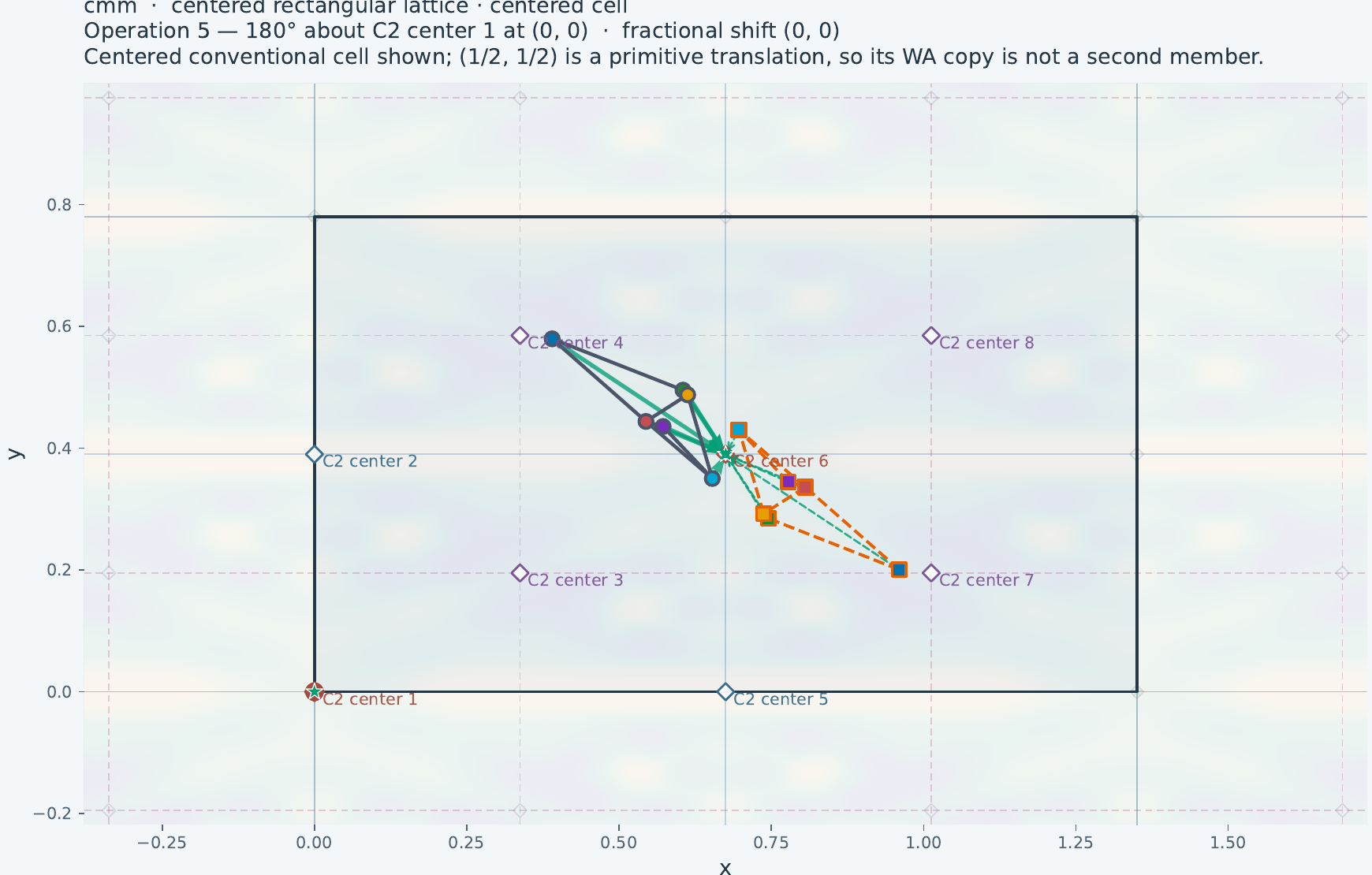} & \includegraphics[width=\linewidth,height=0.22\textheight,keepaspectratio]{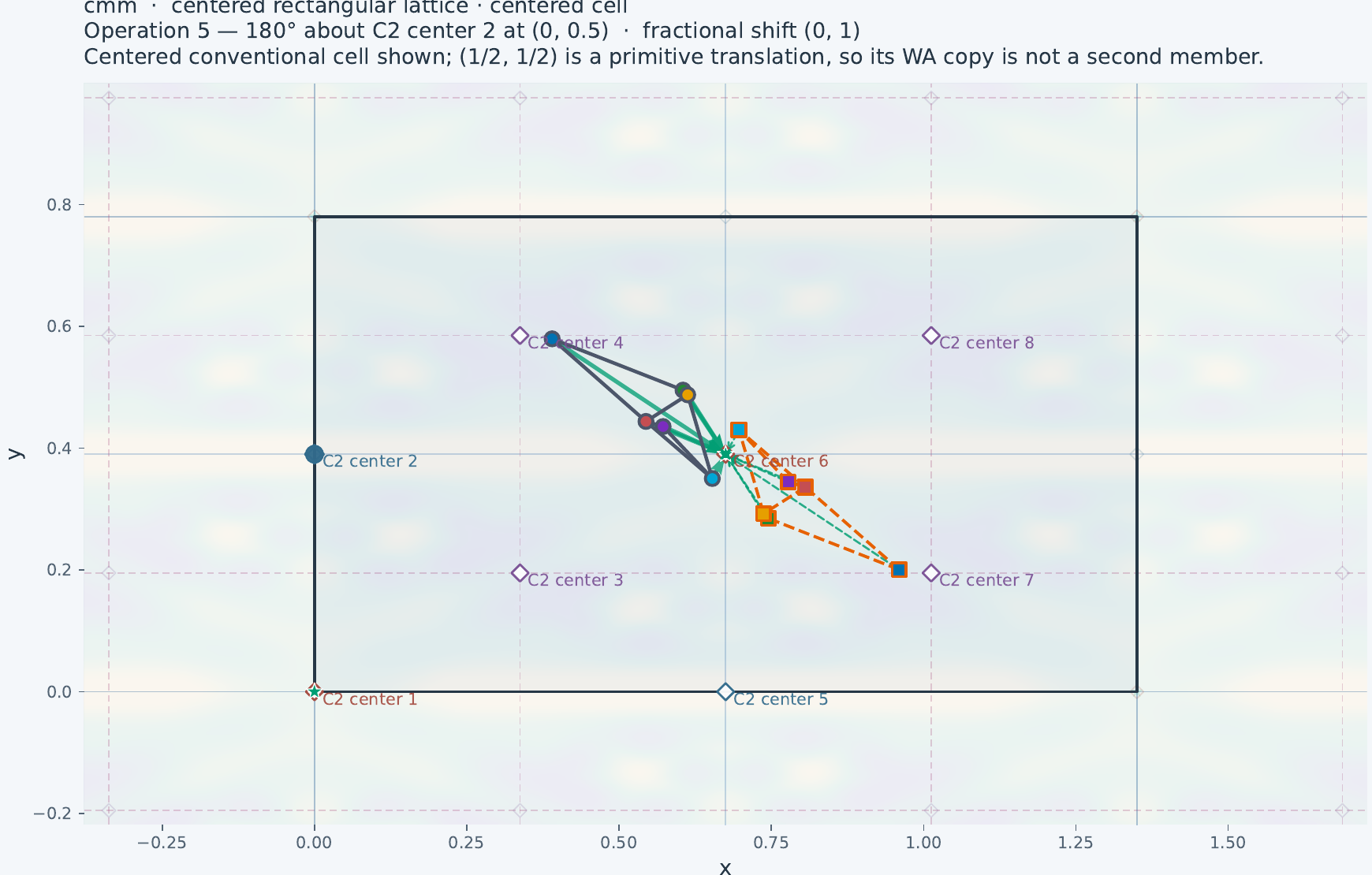} \\[2mm]
\includegraphics[width=\linewidth,height=0.22\textheight,keepaspectratio]{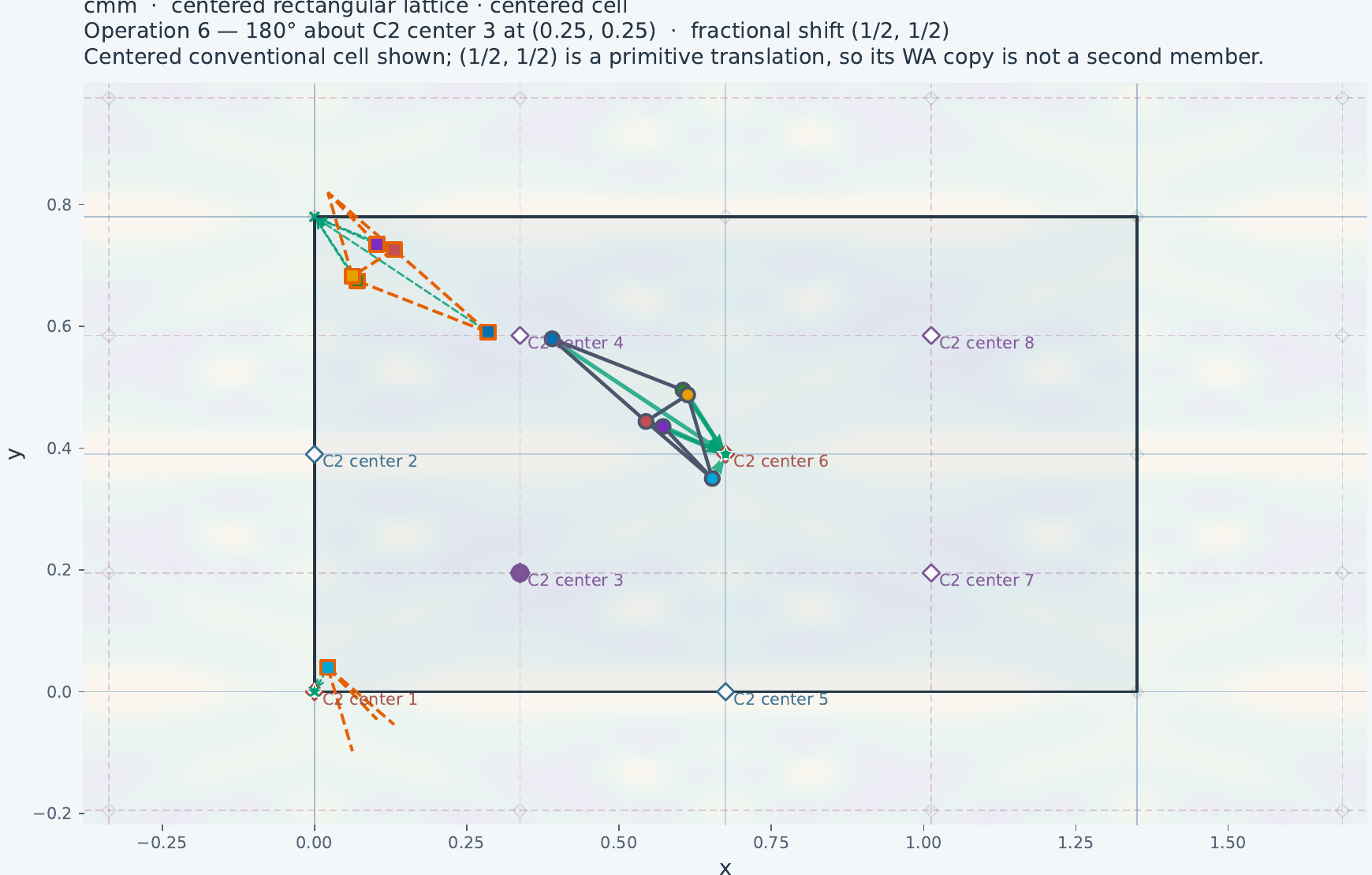} & \includegraphics[width=\linewidth,height=0.22\textheight,keepaspectratio]{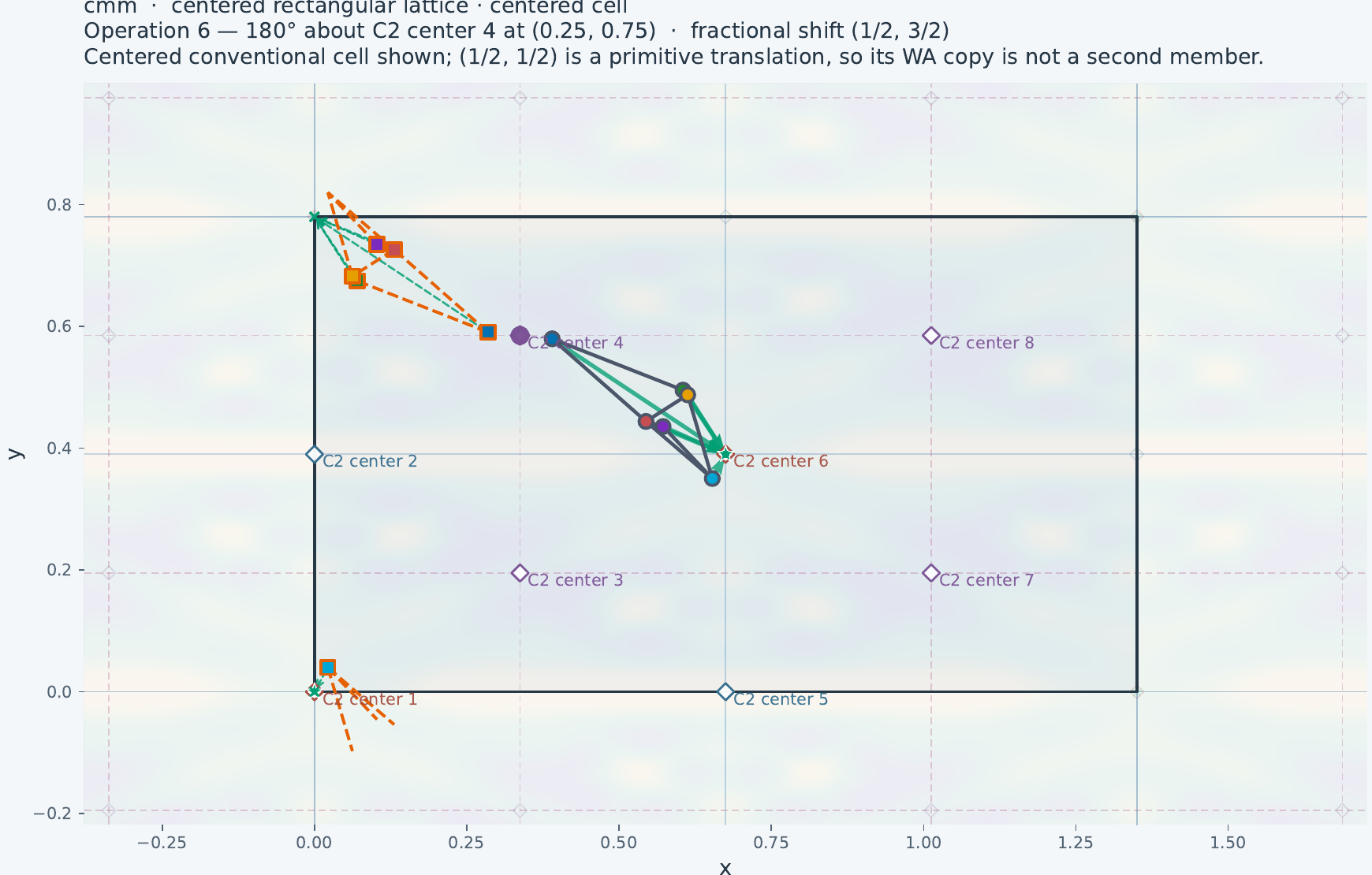} \\
\end{longtable}
\Needspace{218pt}
\begin{longtable}{@{}>{\centering\arraybackslash}p{0.475\linewidth}>{\centering\arraybackslash}p{0.475\linewidth}@{}}
\caption[Wallpaper group \texttt{p4}]{Wallpaper group \texttt{p4}, {\small WA is pointing to C4-A at $q_{1}=(\frac{1}{2},\frac{1}{2})$}}\label{tab:wa-actions-p4}\\[1mm]
\multicolumn{2}{c}{\includegraphics[width=0.96\linewidth]{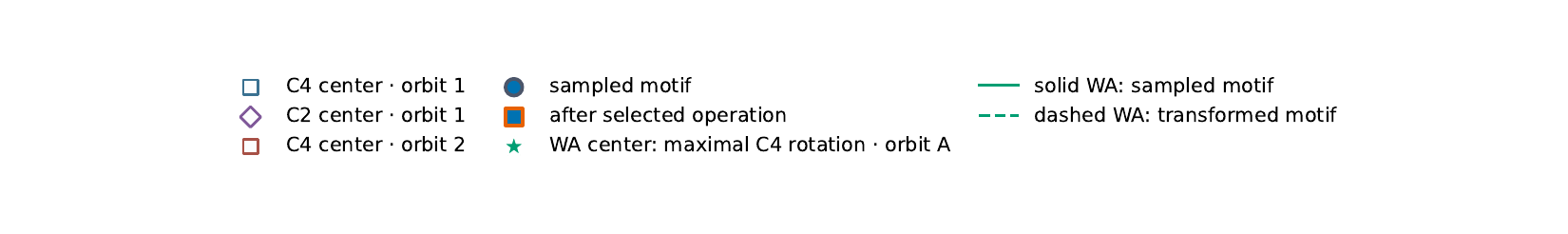}} \\[1mm]
\endfirsthead
\caption[]{Wallpaper group \texttt{p4} (continued)}\\[1mm]
\endhead
\includegraphics[width=\linewidth,height=0.22\textheight,keepaspectratio]{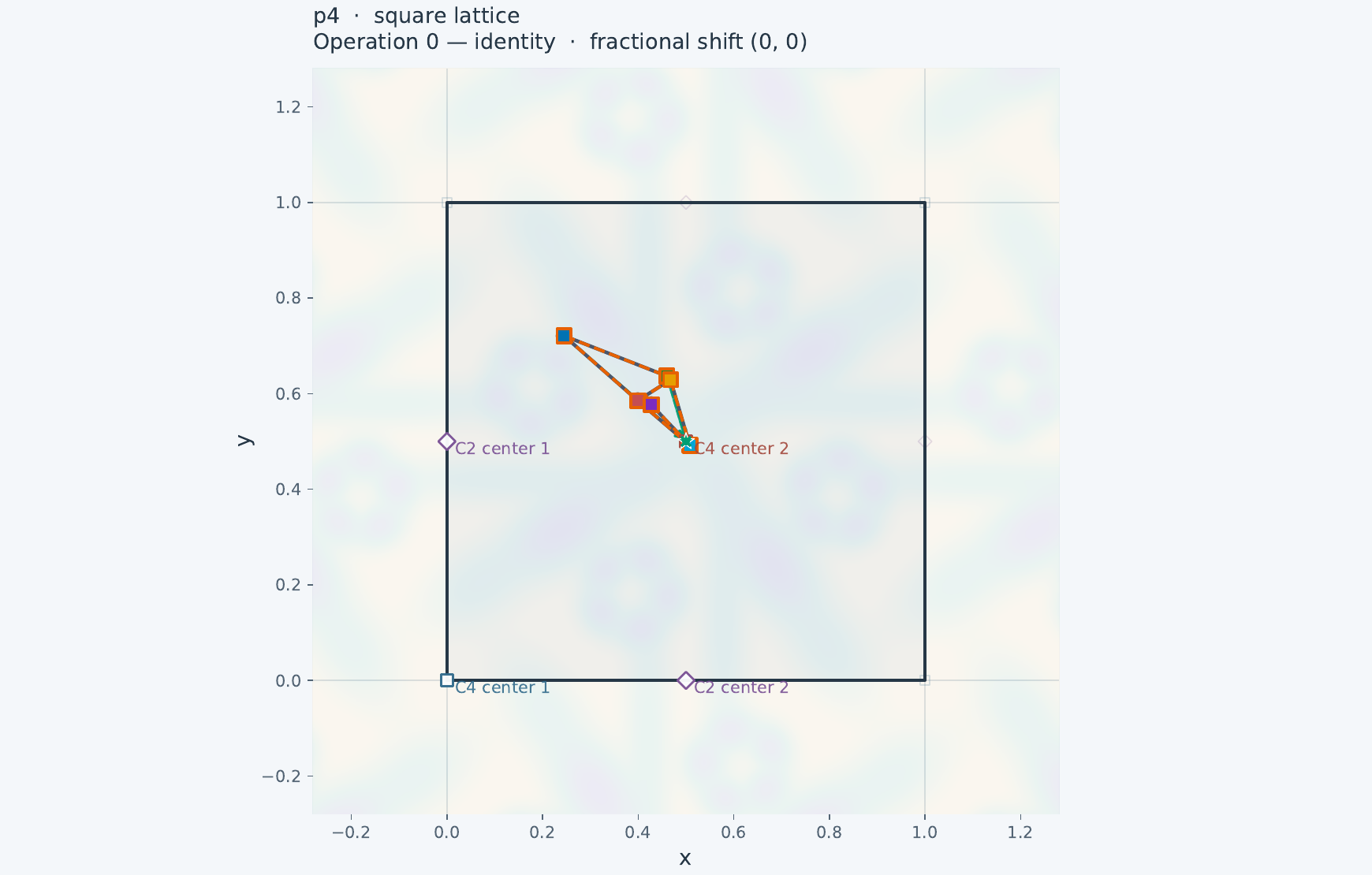} & \includegraphics[width=\linewidth,height=0.22\textheight,keepaspectratio]{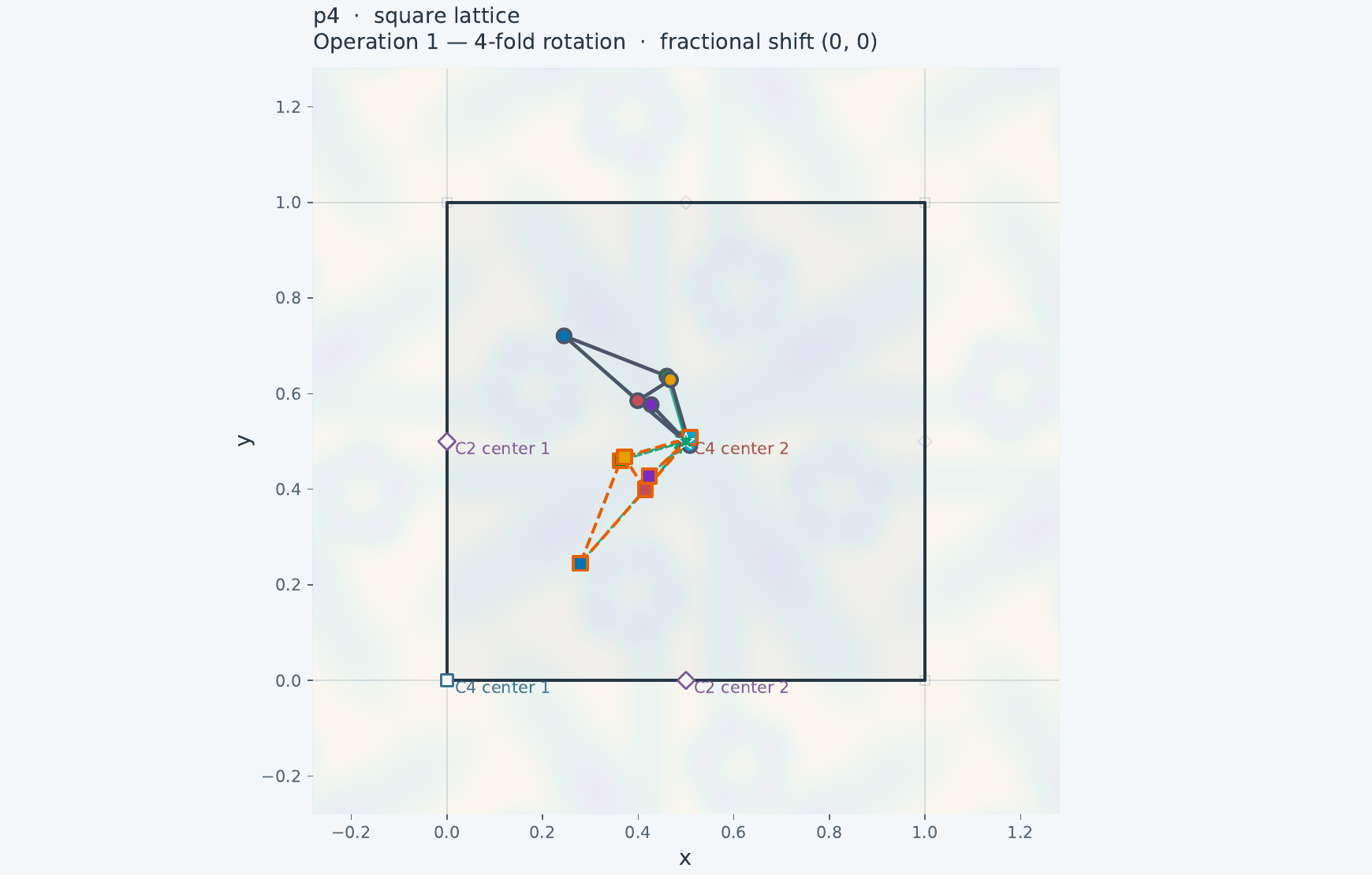} \\[2mm]
\includegraphics[width=\linewidth,height=0.22\textheight,keepaspectratio]{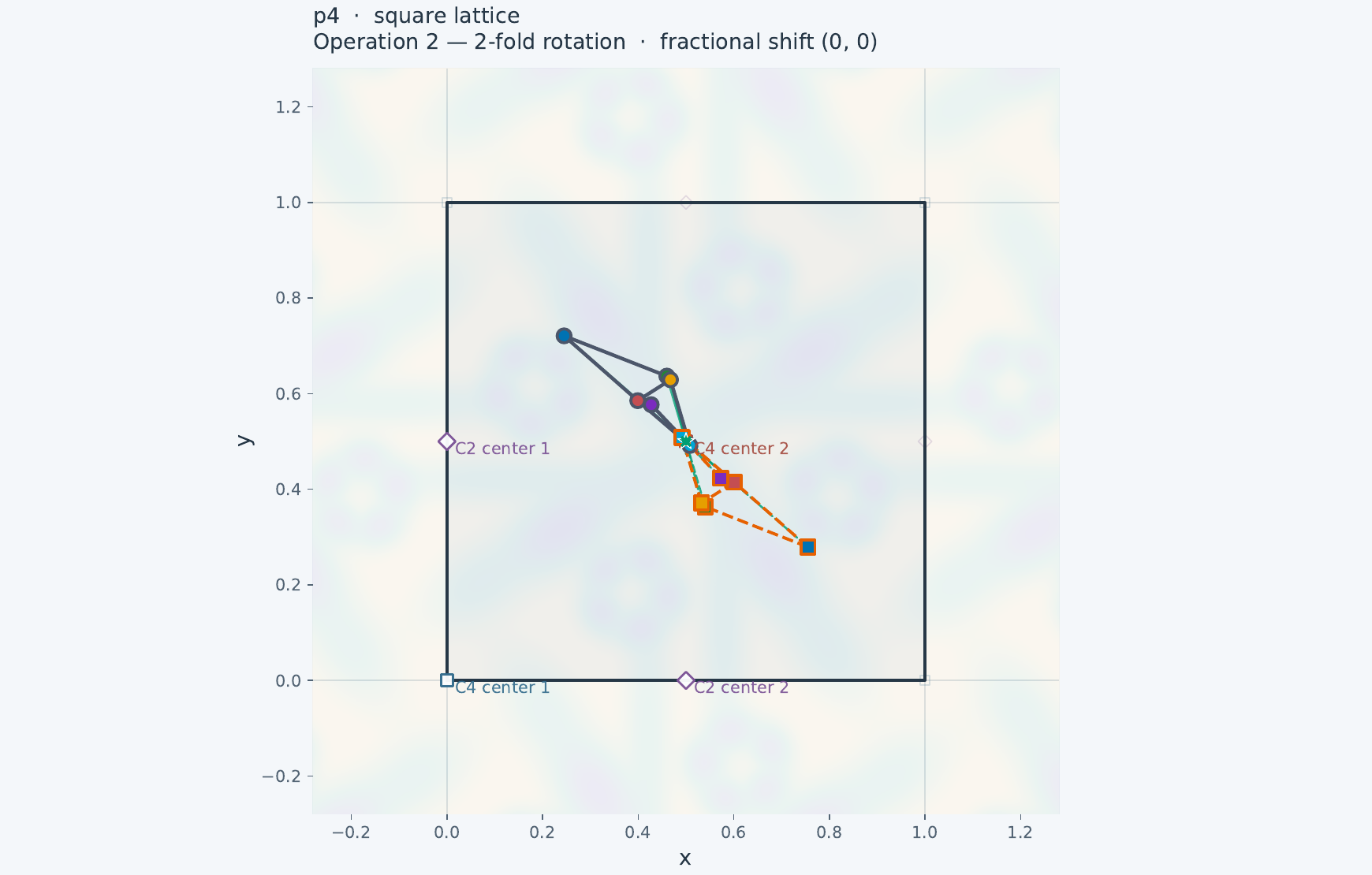} & \includegraphics[width=\linewidth,height=0.22\textheight,keepaspectratio]{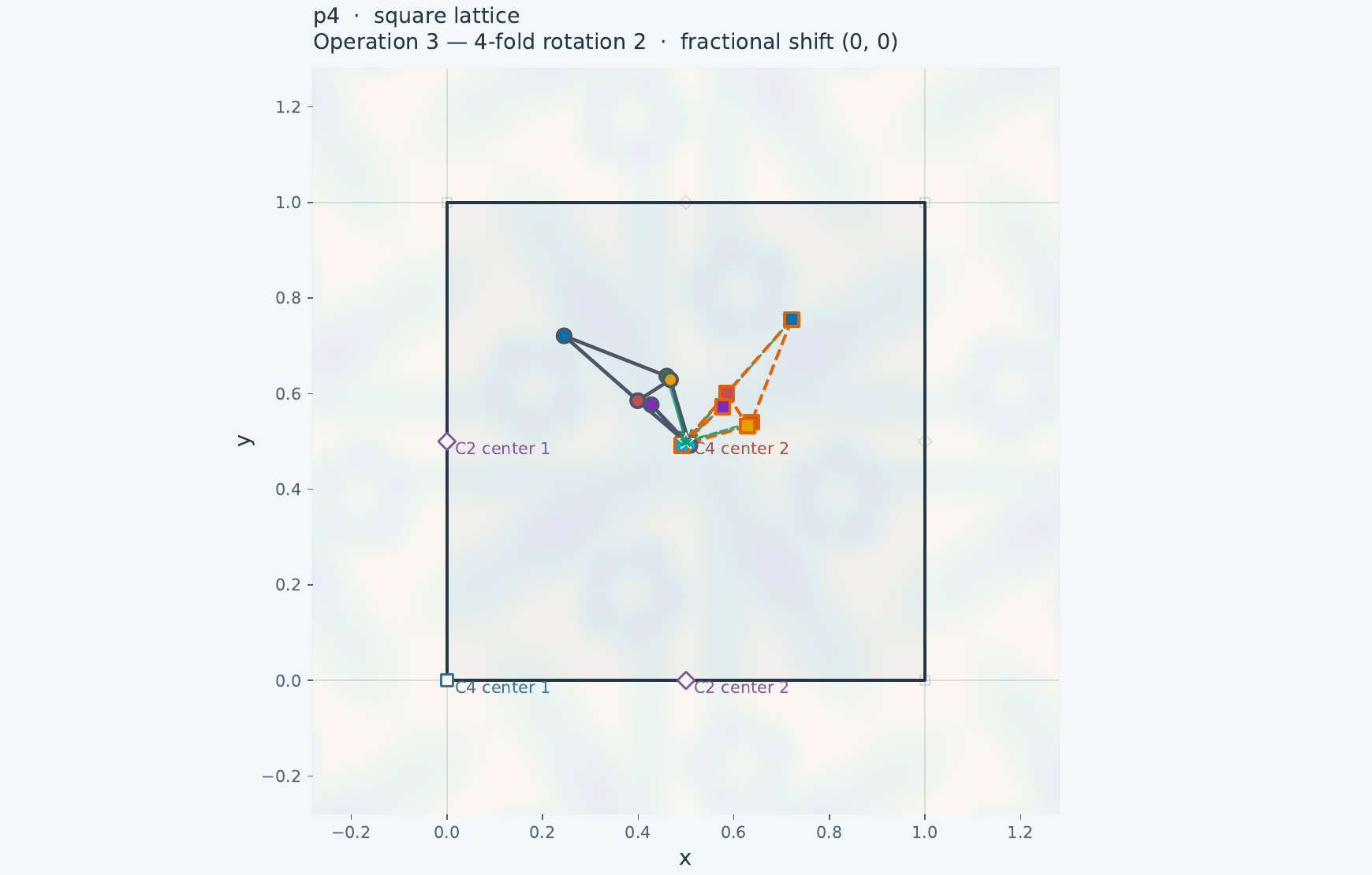} \\[2mm]
\includegraphics[width=\linewidth,height=0.22\textheight,keepaspectratio]{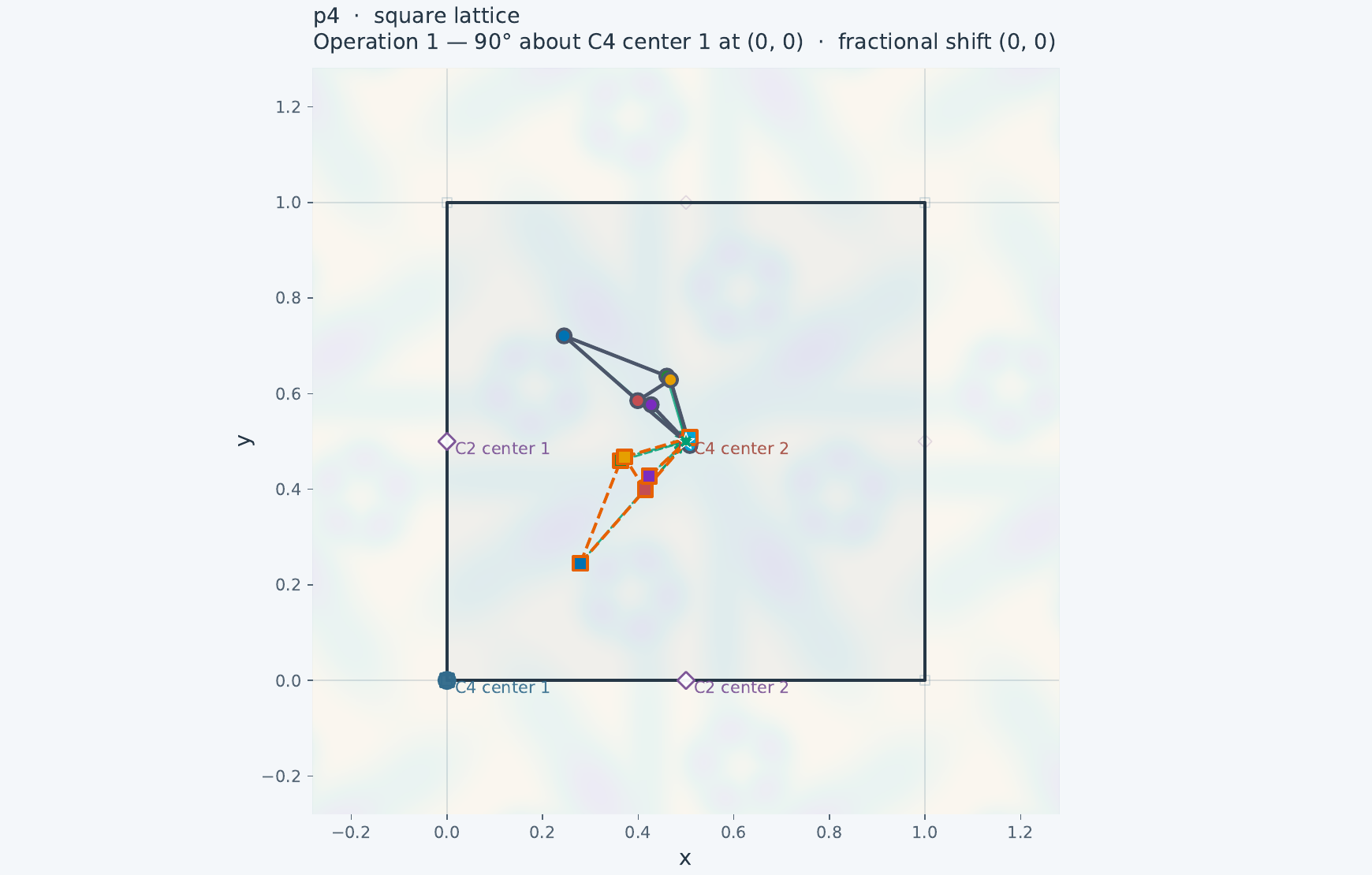} & \includegraphics[width=\linewidth,height=0.22\textheight,keepaspectratio]{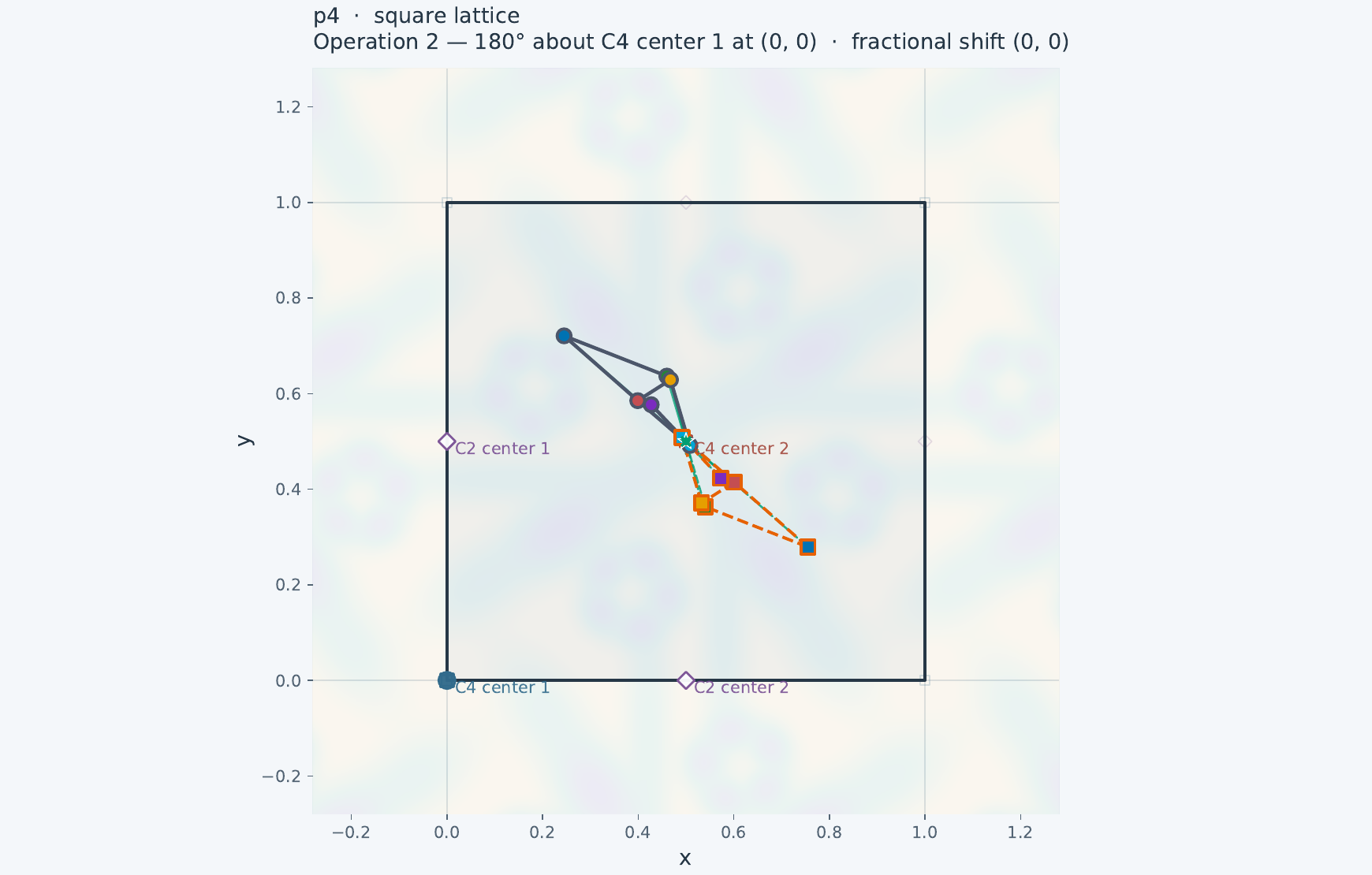} \\[2mm]
\includegraphics[width=\linewidth,height=0.22\textheight,keepaspectratio]{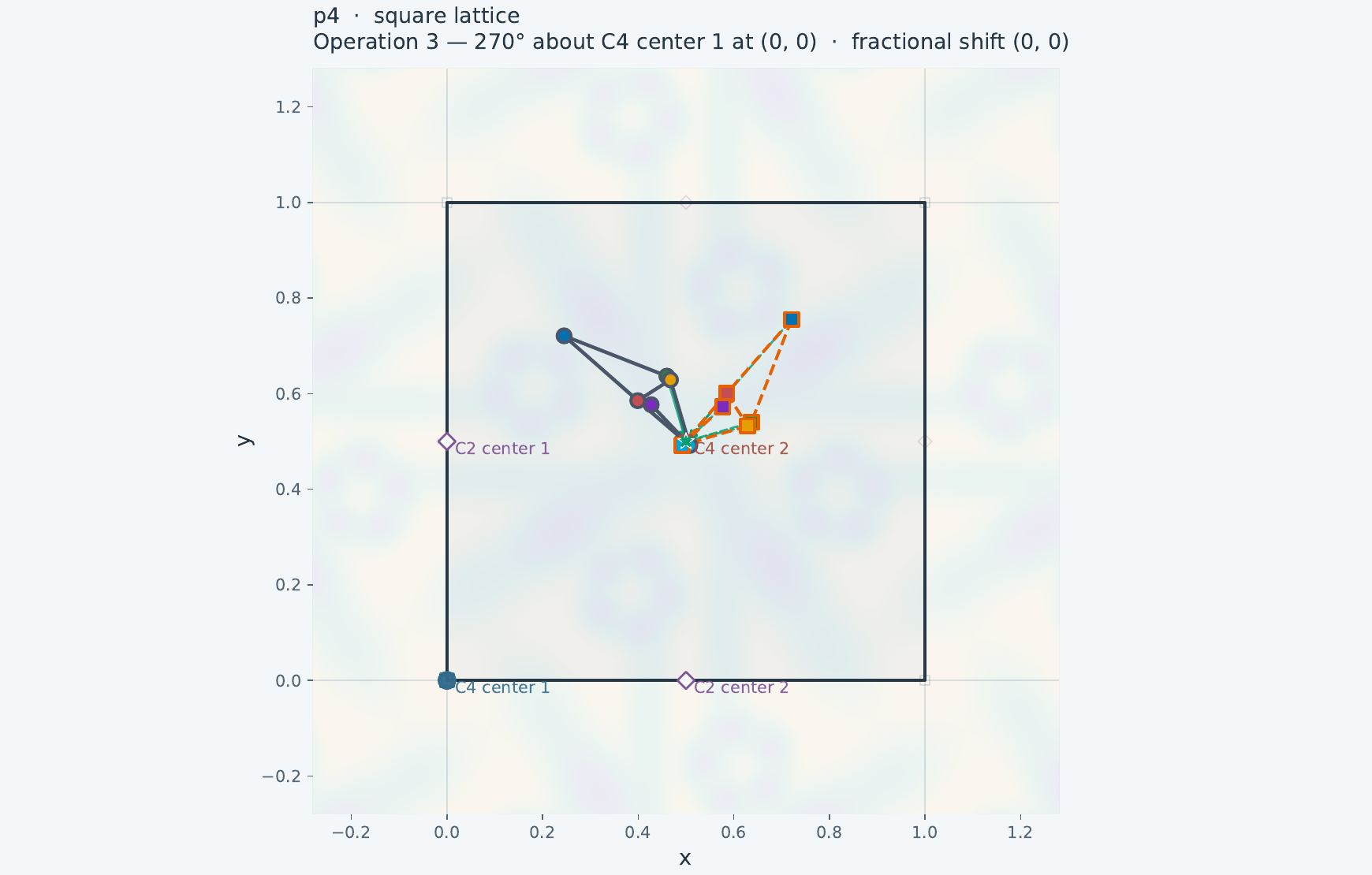} & \includegraphics[width=\linewidth,height=0.22\textheight,keepaspectratio]{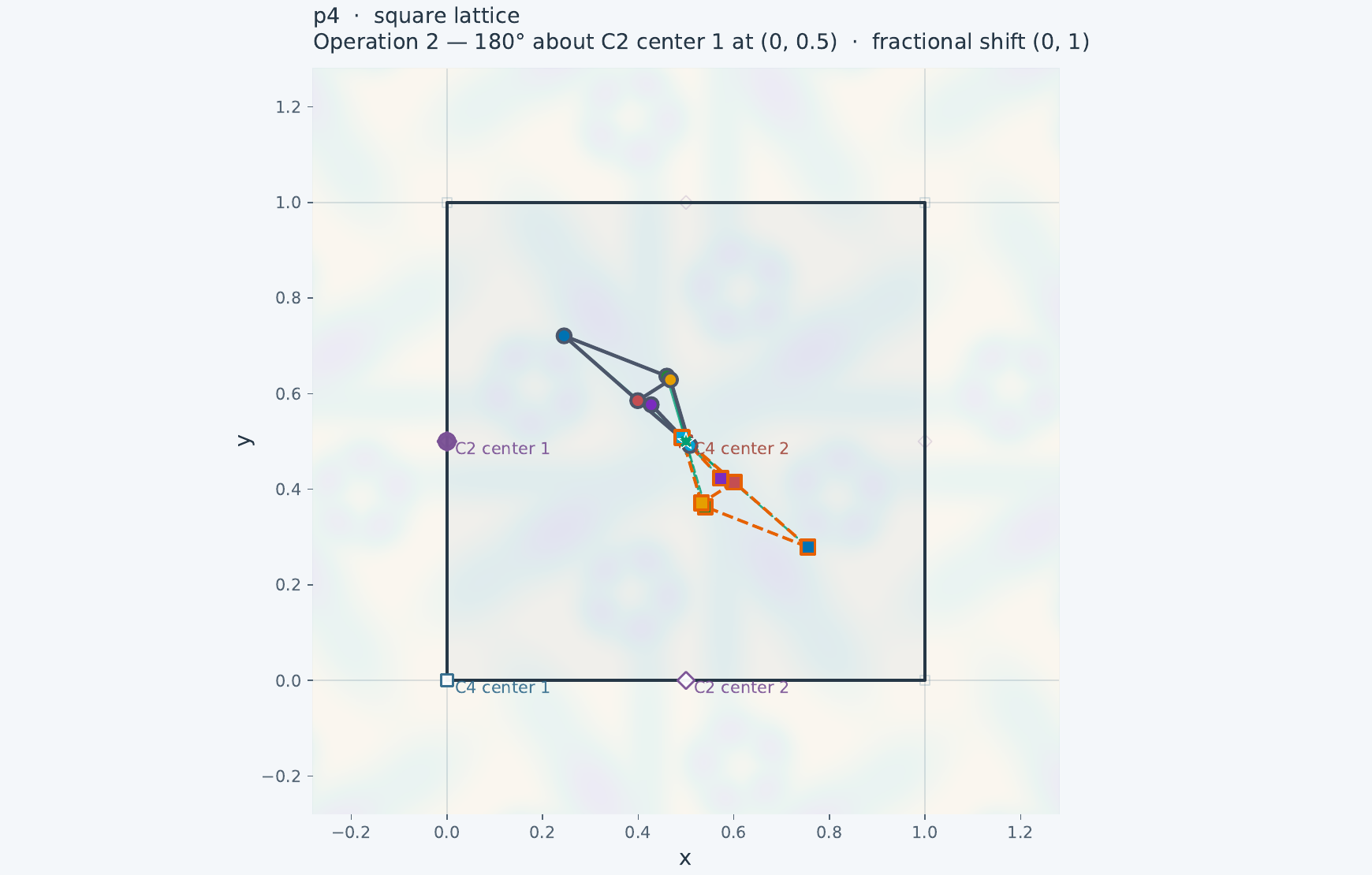} \\[2mm]
\includegraphics[width=\linewidth,height=0.22\textheight,keepaspectratio]{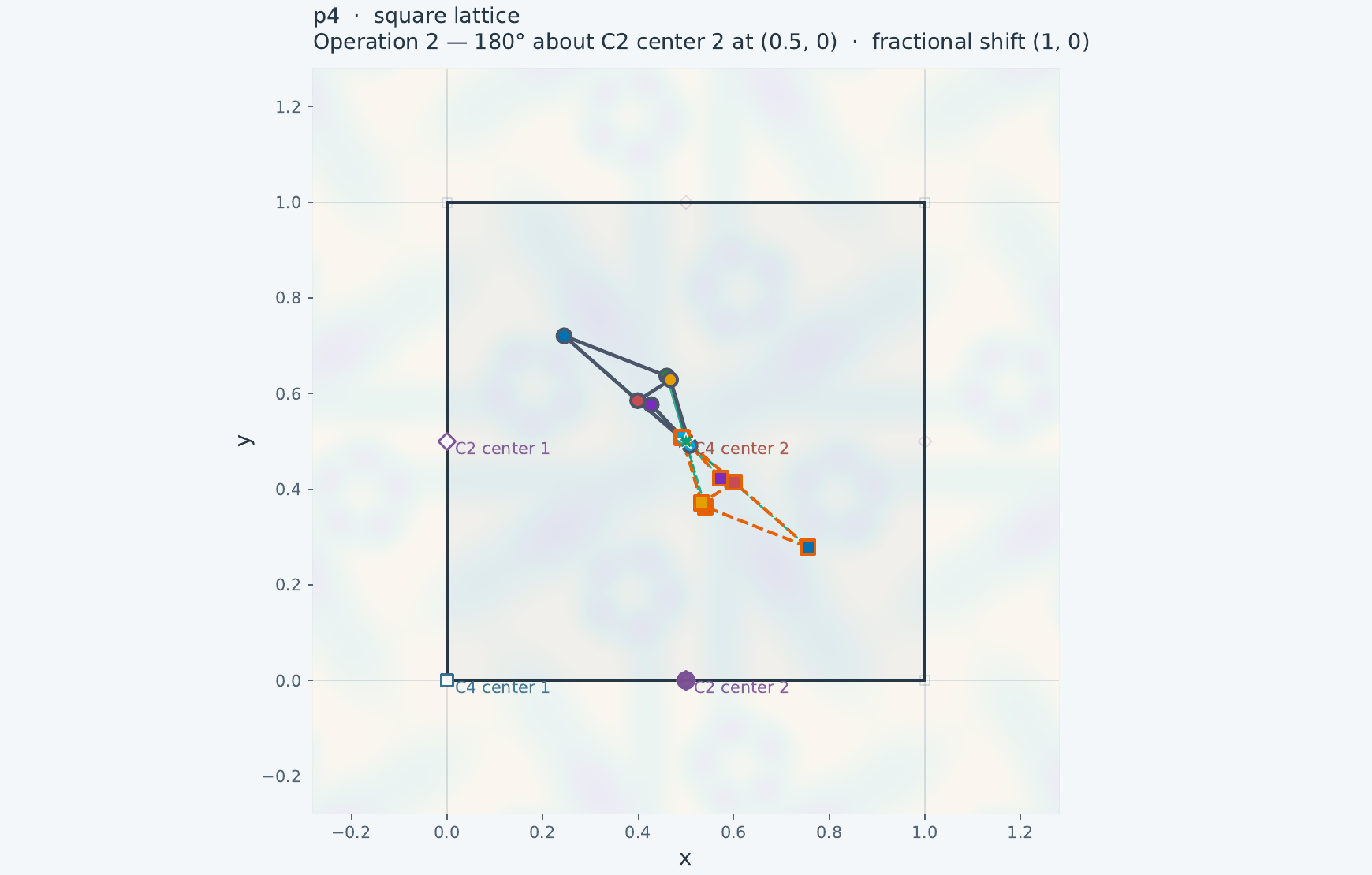} & \includegraphics[width=\linewidth,height=0.22\textheight,keepaspectratio]{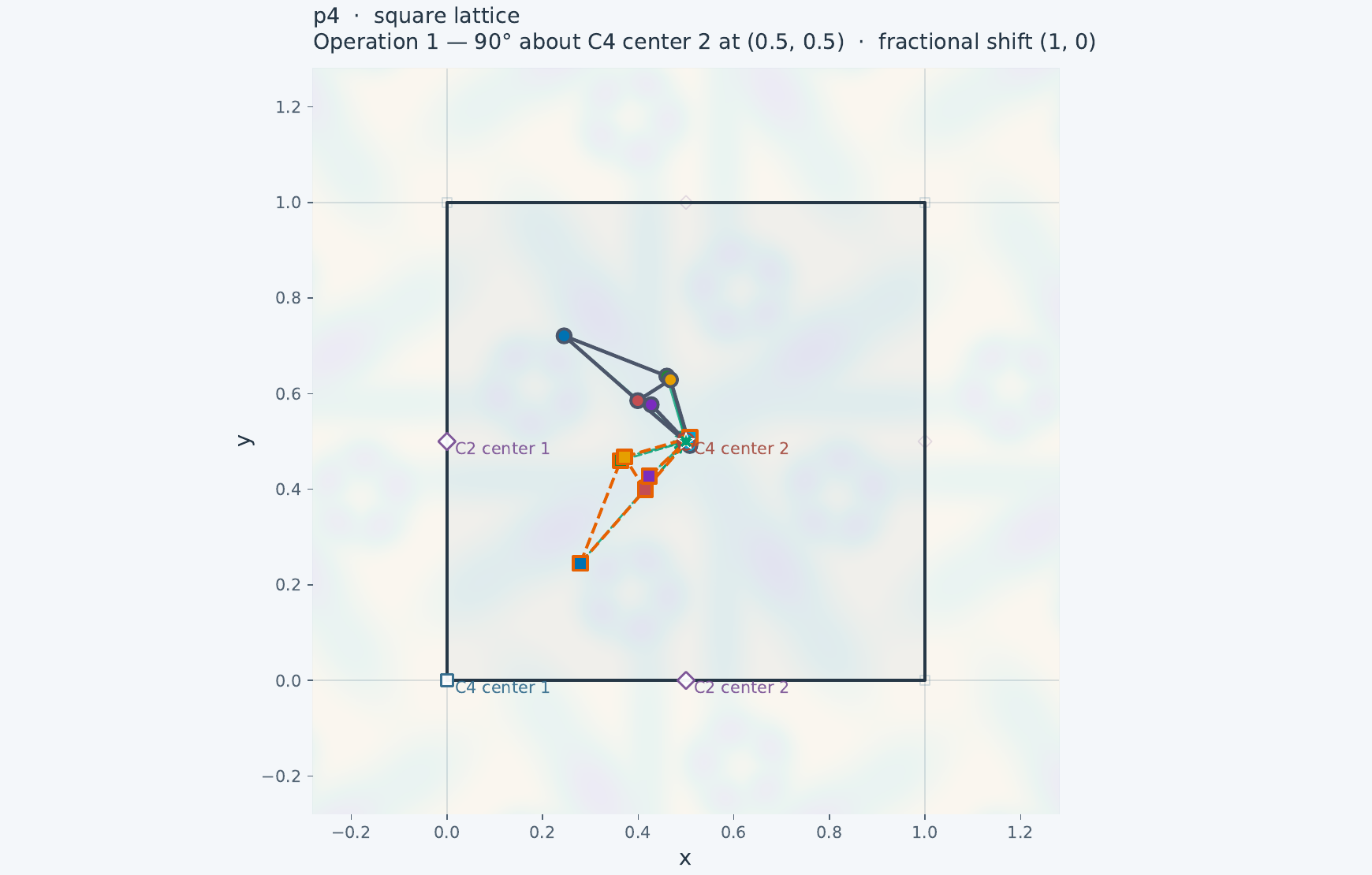} \\[2mm]
\includegraphics[width=\linewidth,height=0.22\textheight,keepaspectratio]{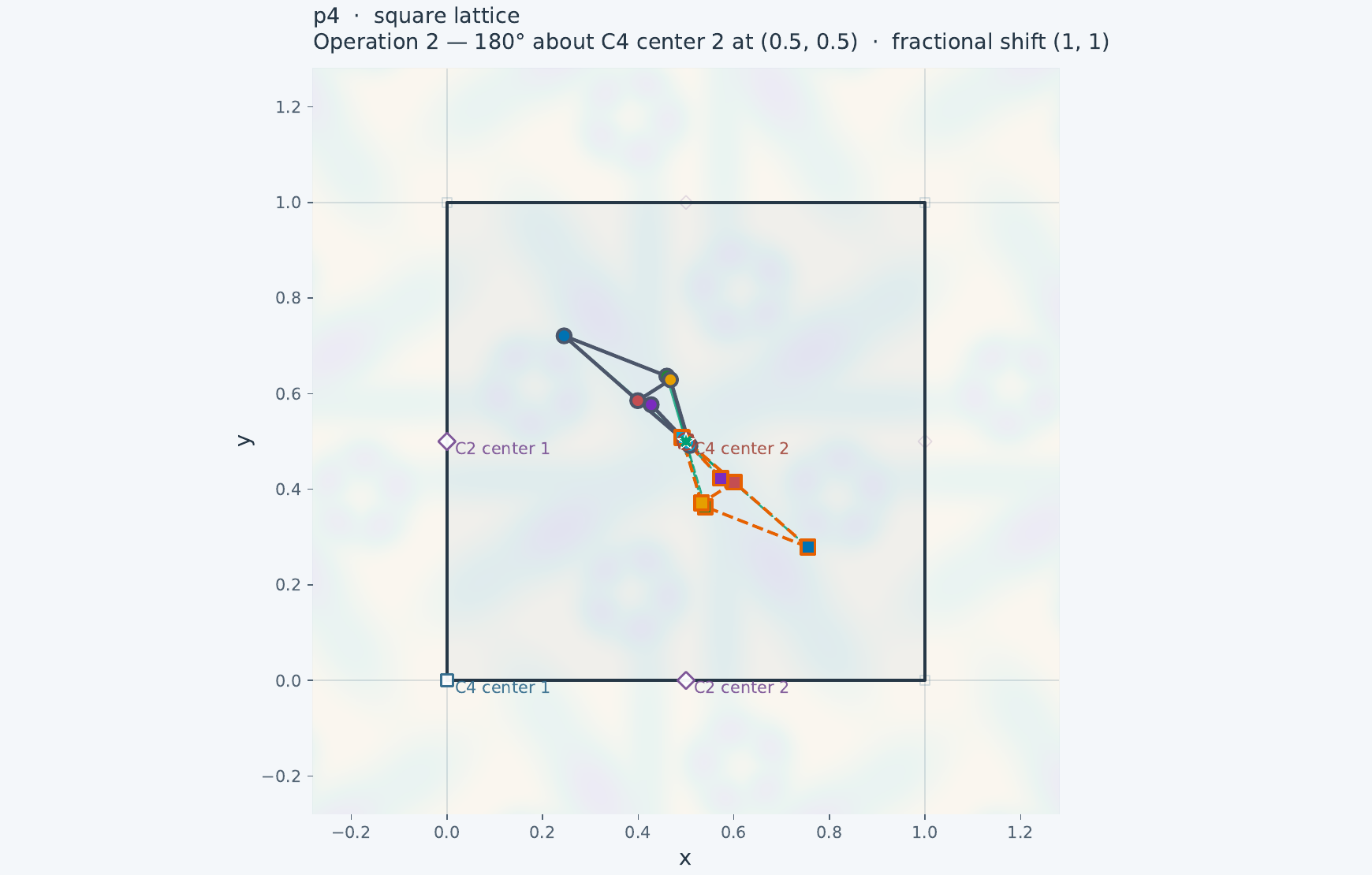} & \includegraphics[width=\linewidth,height=0.22\textheight,keepaspectratio]{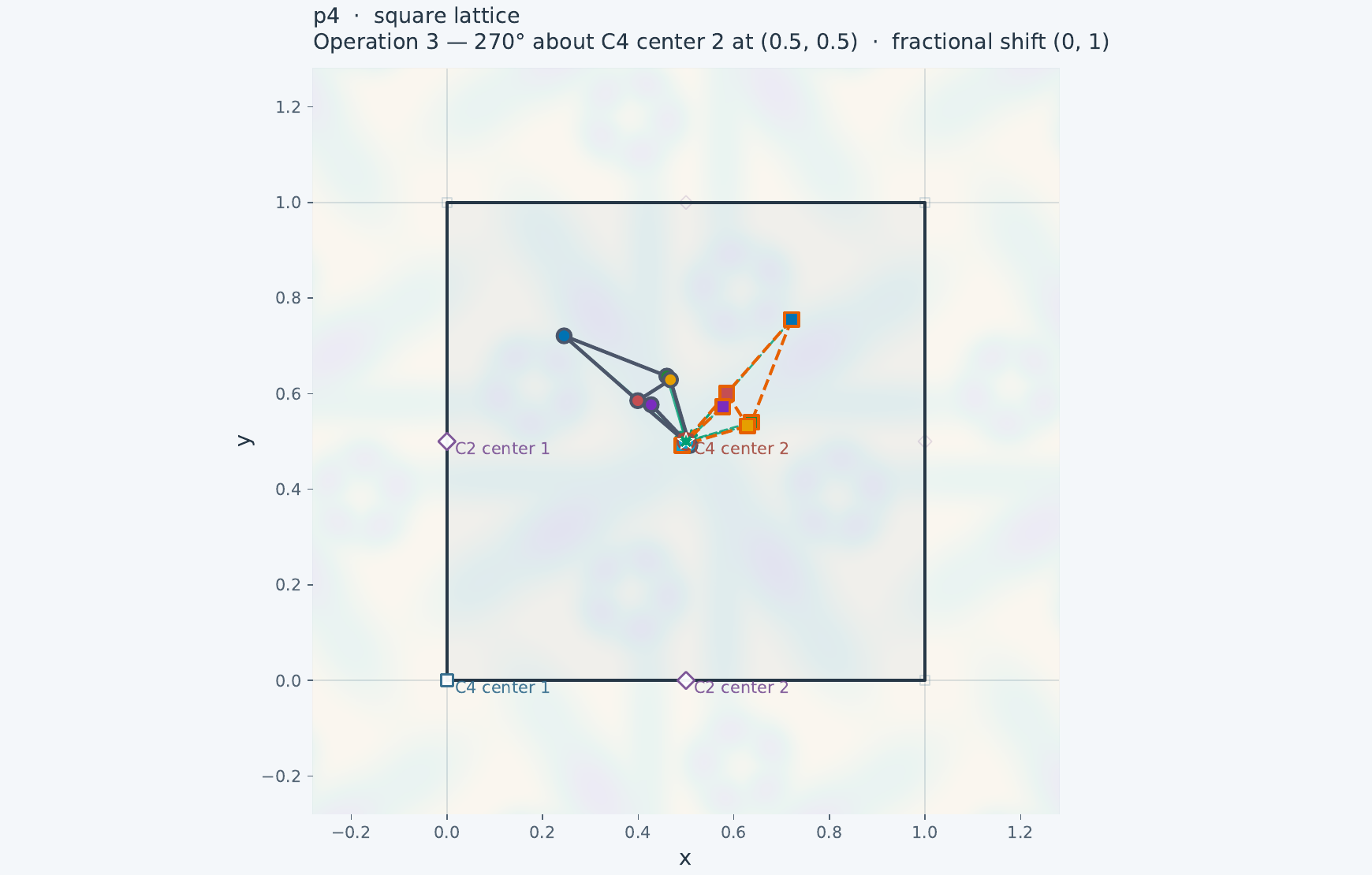} \\
\end{longtable}
\Needspace{236pt}
\begin{longtable}{@{}>{\centering\arraybackslash}p{0.475\linewidth}>{\centering\arraybackslash}p{0.475\linewidth}@{}}
\caption[Wallpaper group \texttt{p4m}]{Wallpaper group \texttt{p4m}, {\small WA is pointing to C4-A at $q_{1}=(\frac{1}{2},\frac{1}{2})$}}\label{tab:wa-actions-p4m}\\[1mm]
\multicolumn{2}{c}{\includegraphics[width=0.96\linewidth]{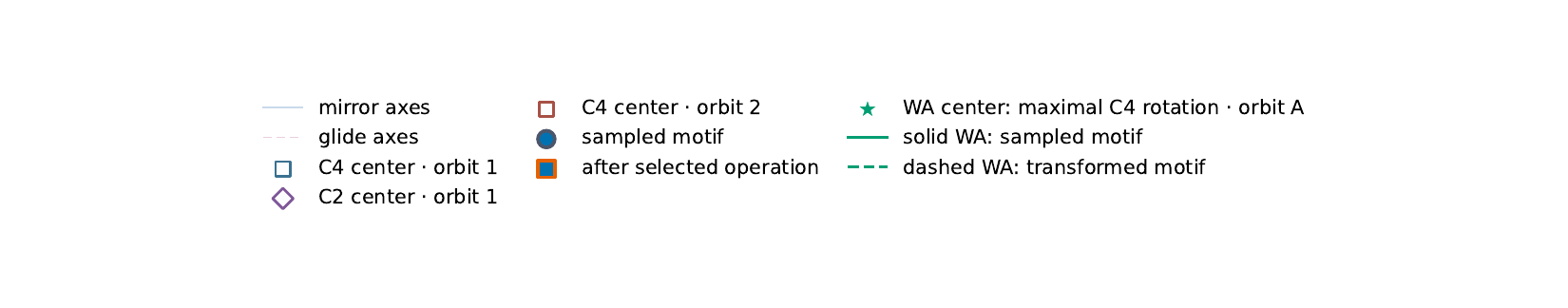}} \\[1mm]
\endfirsthead
\caption[]{Wallpaper group \texttt{p4m} (continued)}\\[1mm]
\endhead
\includegraphics[width=\linewidth,height=0.22\textheight,keepaspectratio]{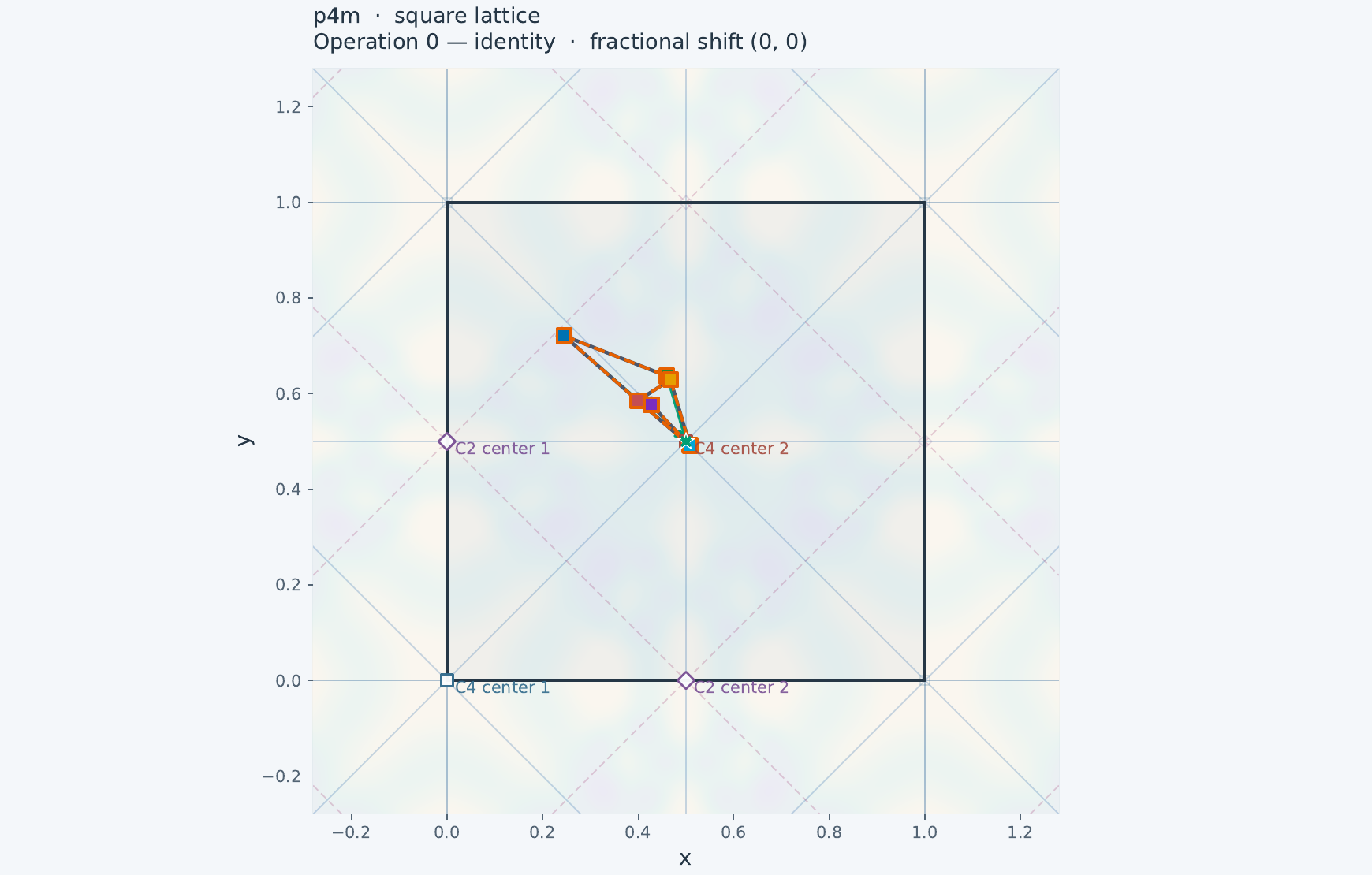} & \includegraphics[width=\linewidth,height=0.22\textheight,keepaspectratio]{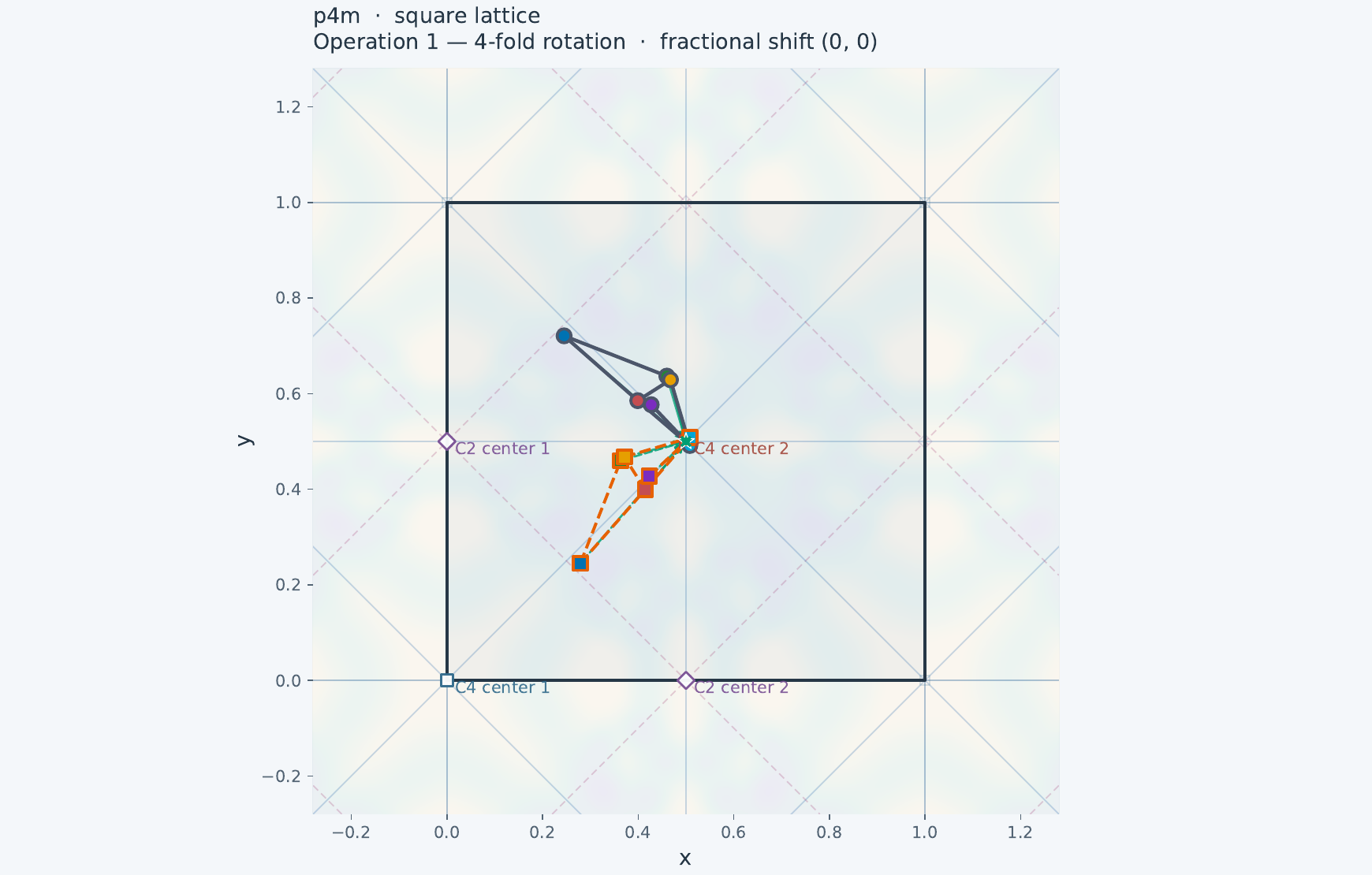} \\[2mm]
\includegraphics[width=\linewidth,height=0.22\textheight,keepaspectratio]{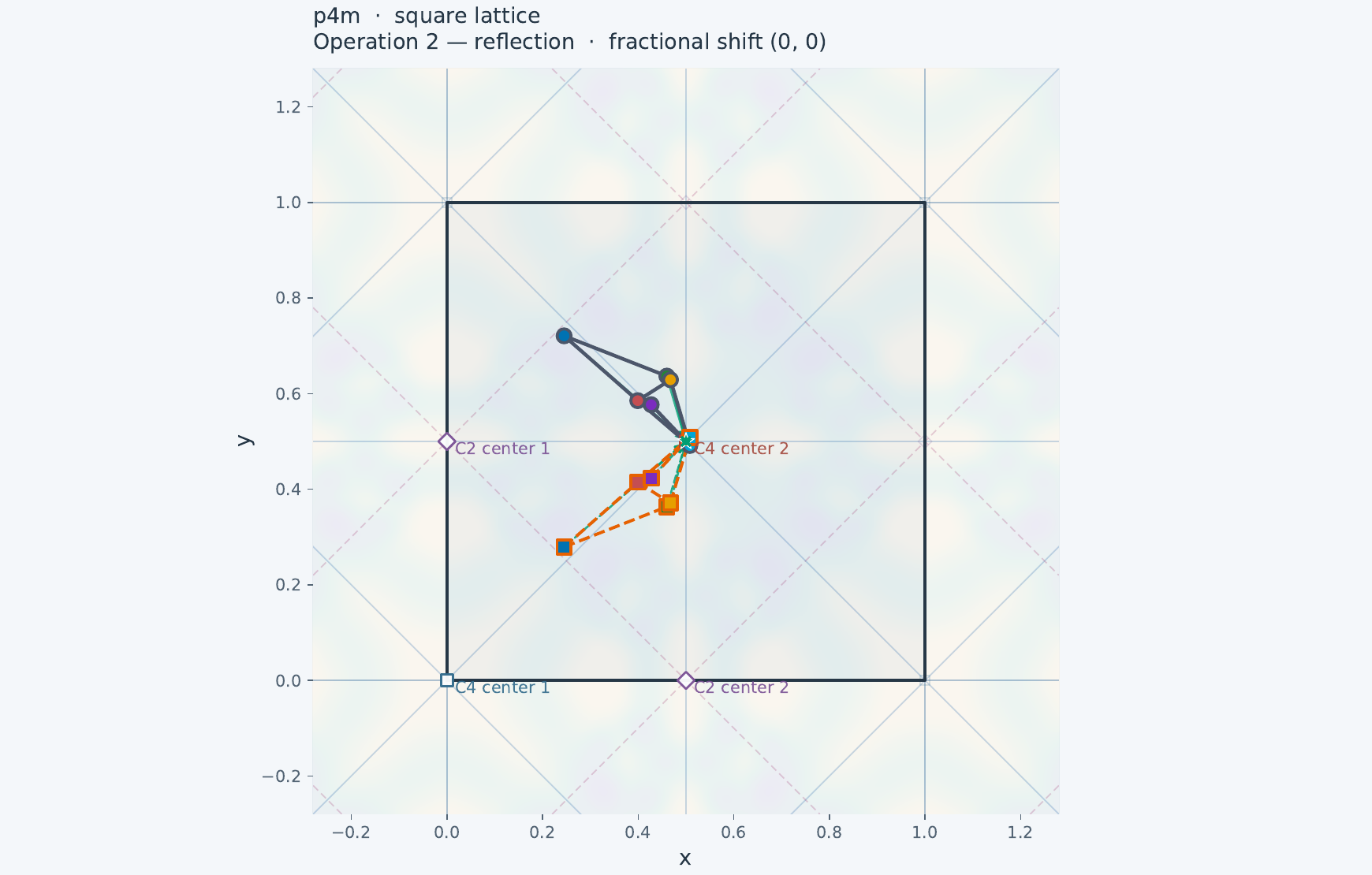} & \includegraphics[width=\linewidth,height=0.22\textheight,keepaspectratio]{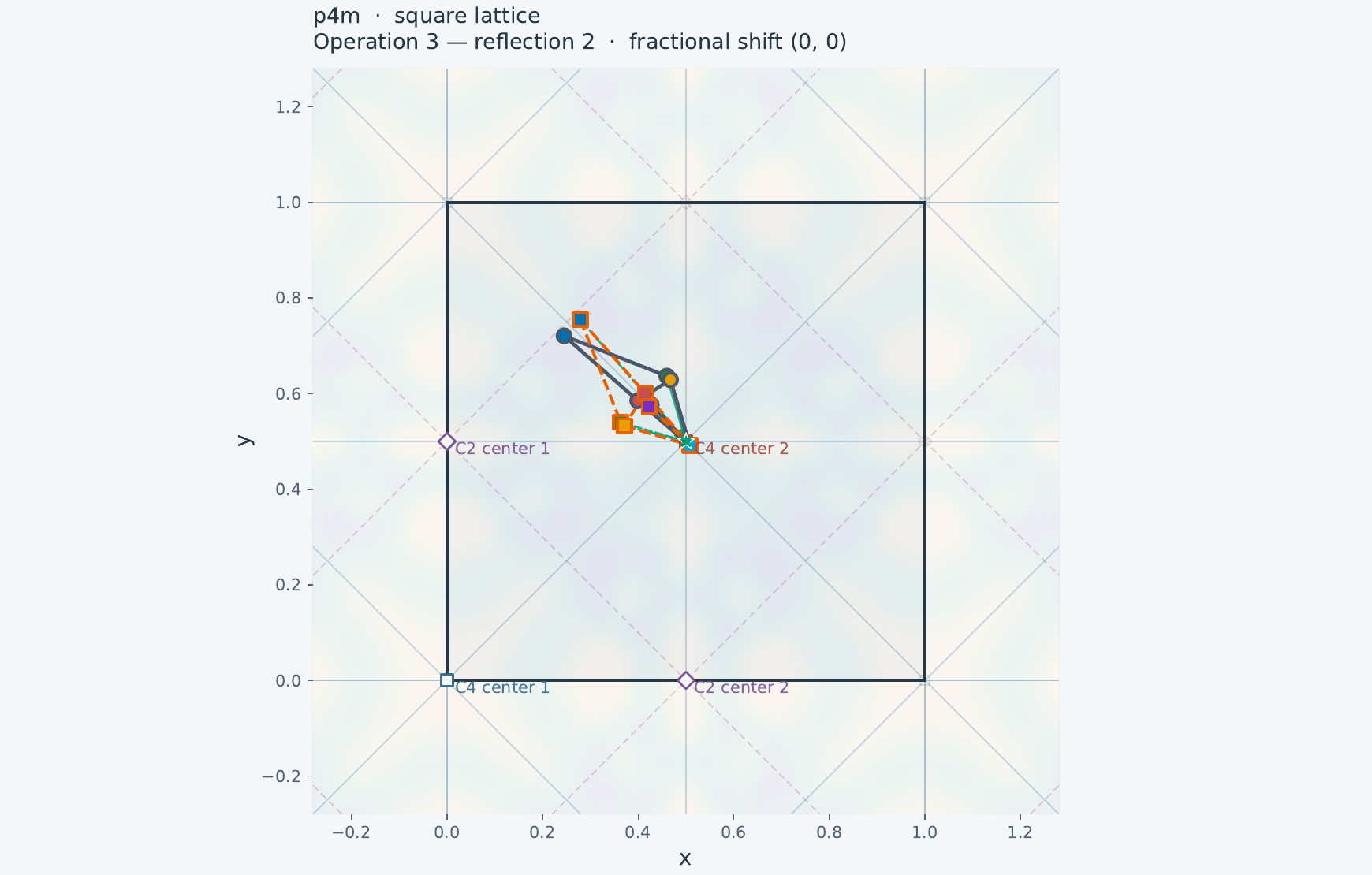} \\[2mm]
\includegraphics[width=\linewidth,height=0.22\textheight,keepaspectratio]{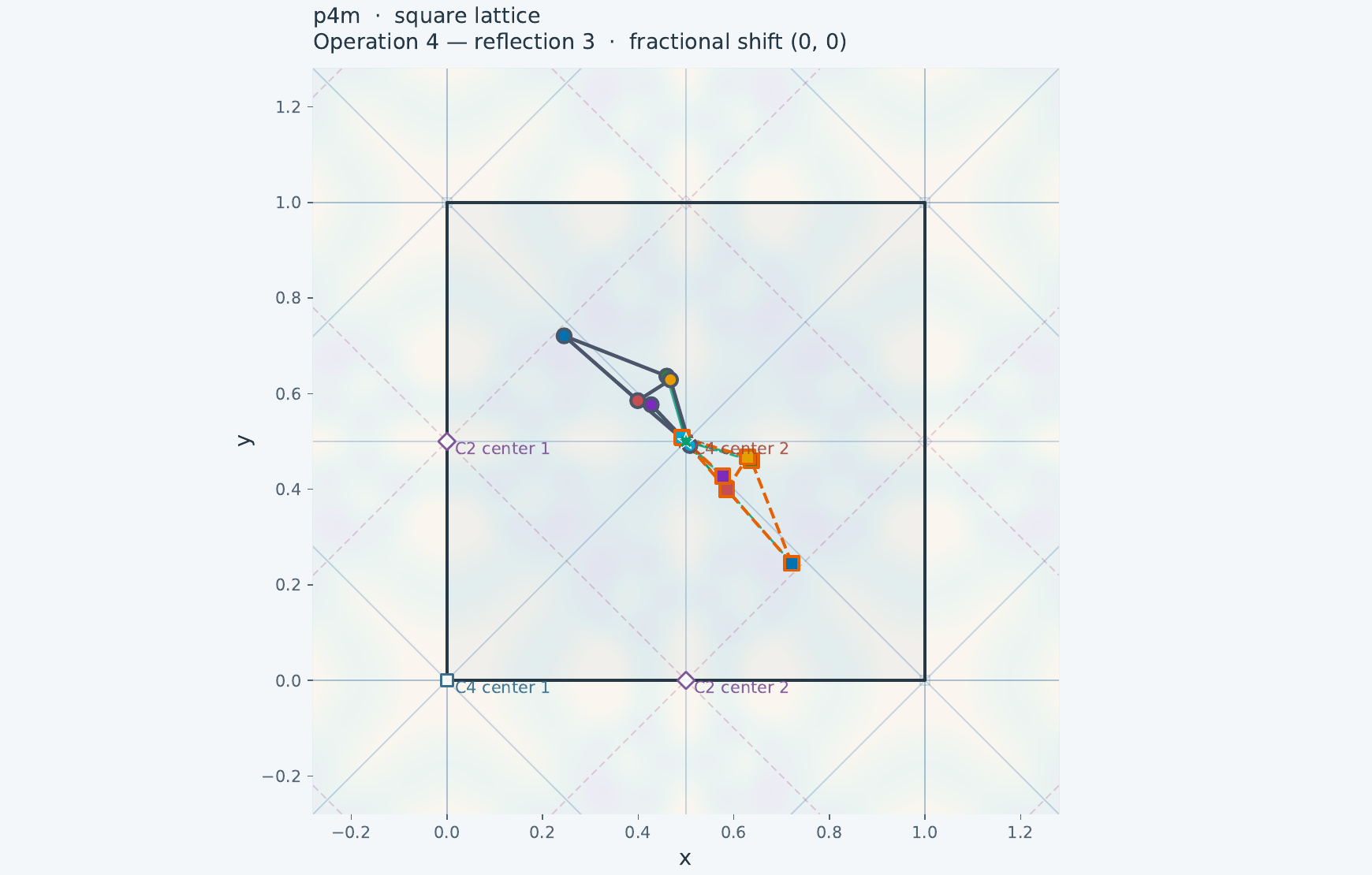} & \includegraphics[width=\linewidth,height=0.22\textheight,keepaspectratio]{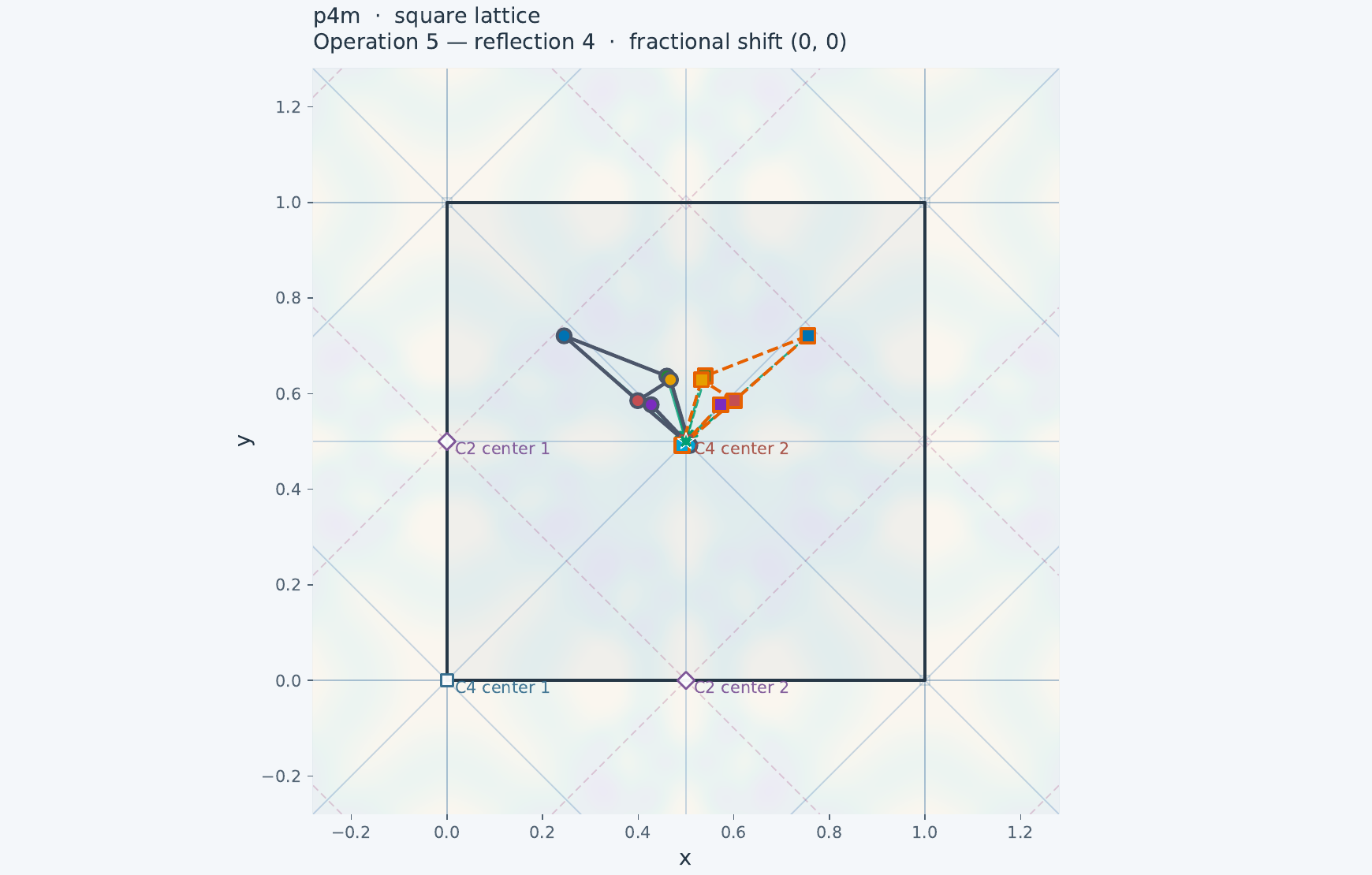} \\[2mm]
\includegraphics[width=\linewidth,height=0.22\textheight,keepaspectratio]{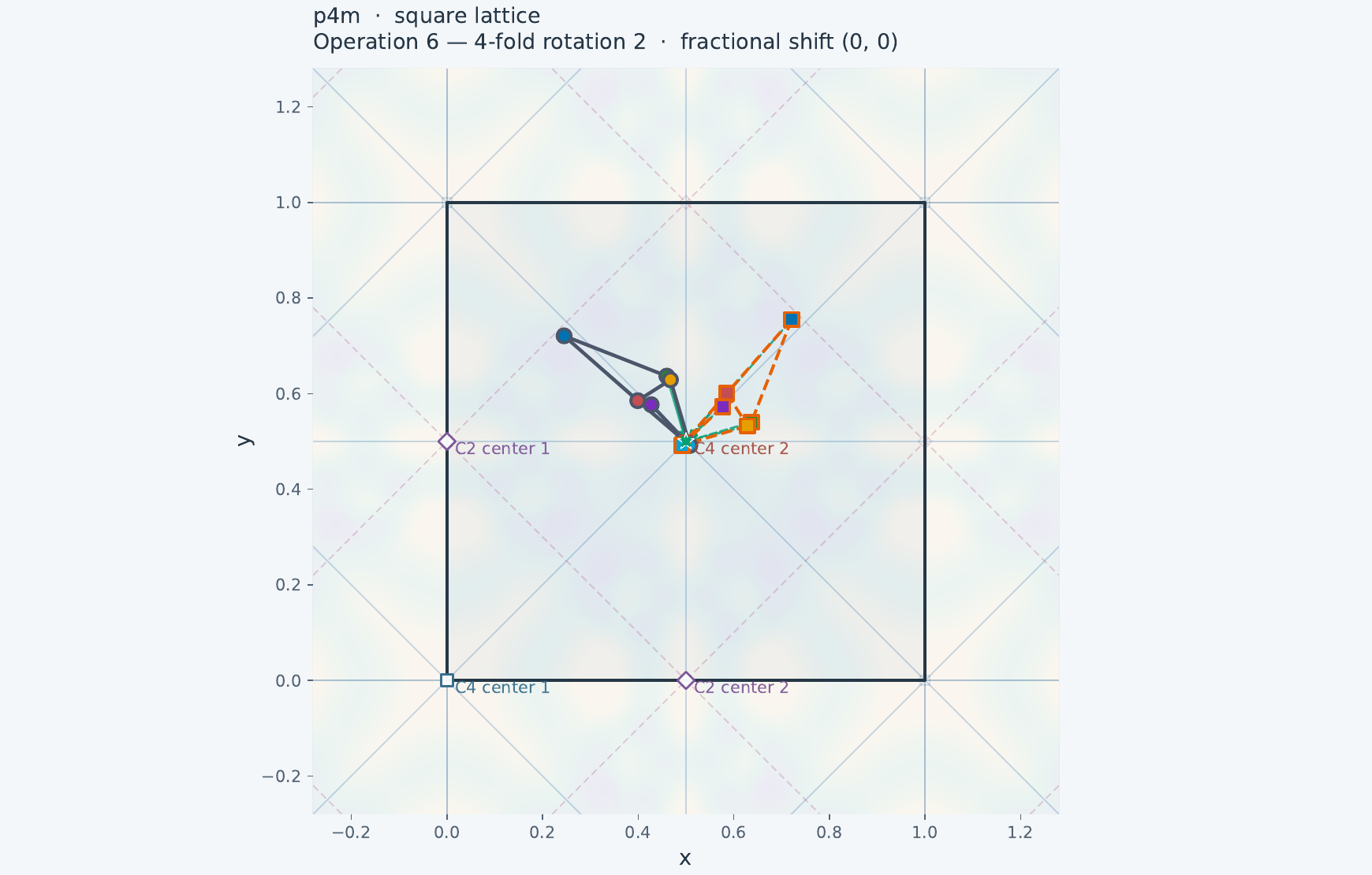} & \includegraphics[width=\linewidth,height=0.22\textheight,keepaspectratio]{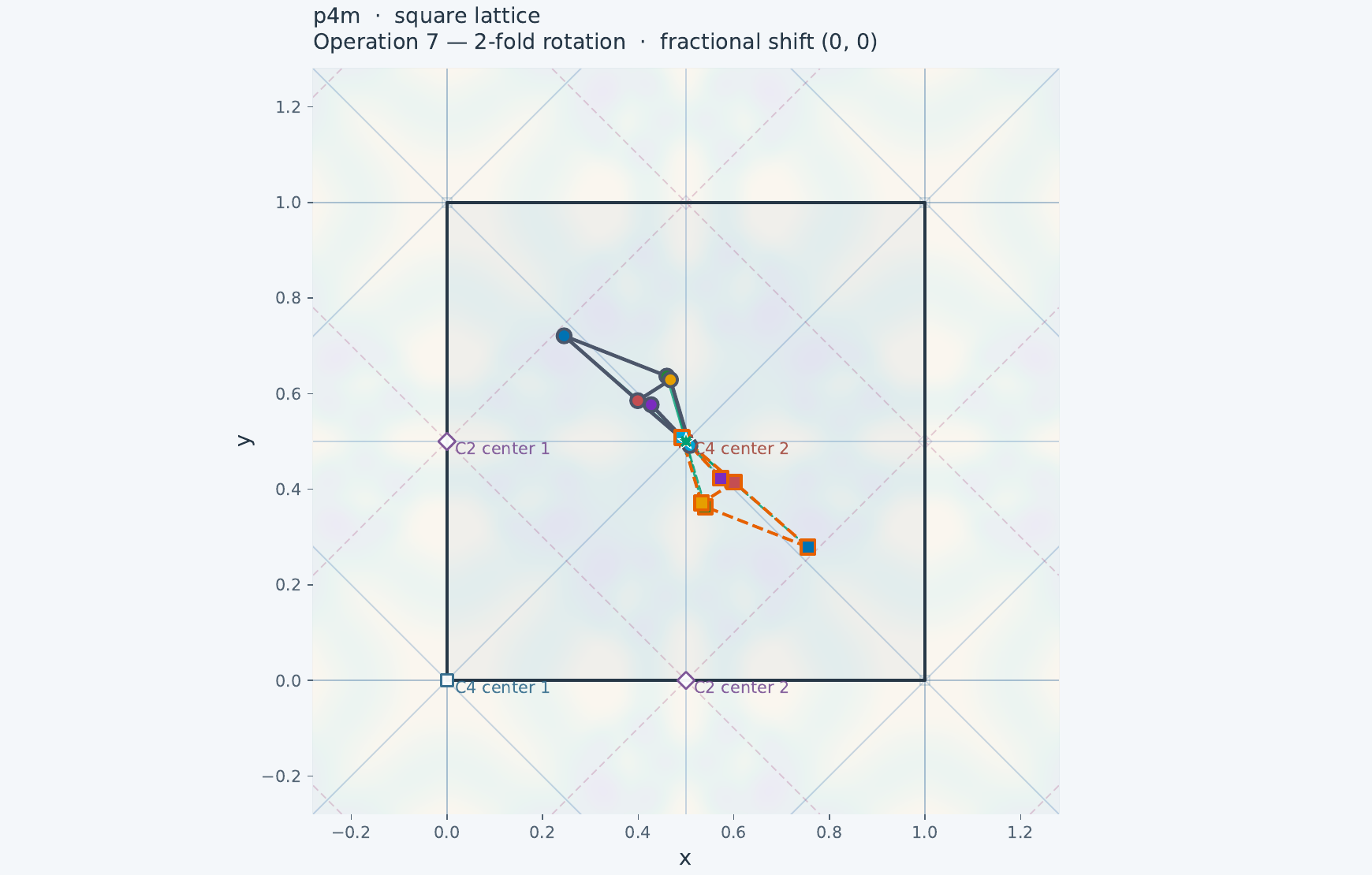} \\[2mm]
\includegraphics[width=\linewidth,height=0.22\textheight,keepaspectratio]{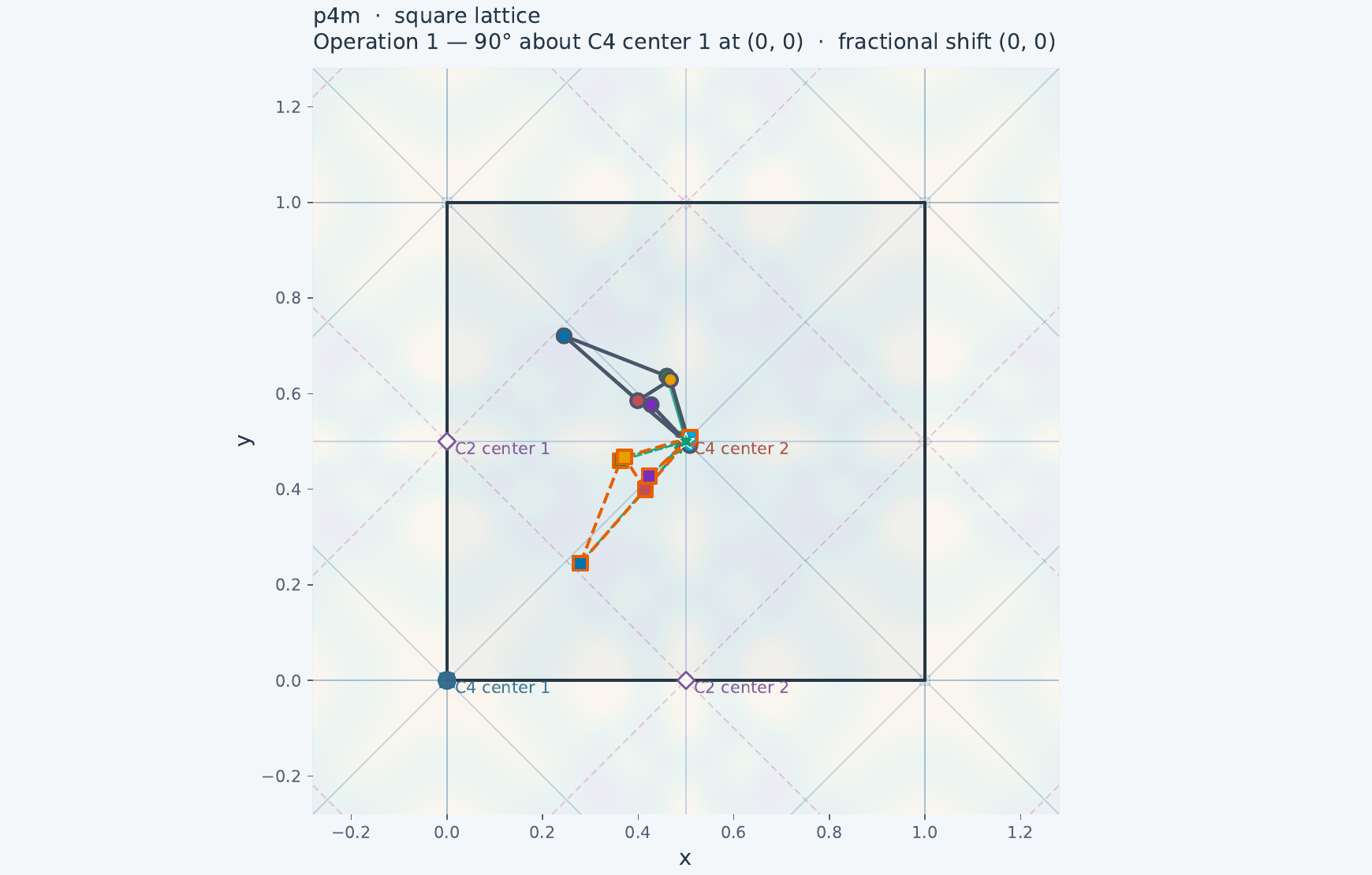} & \includegraphics[width=\linewidth,height=0.22\textheight,keepaspectratio]{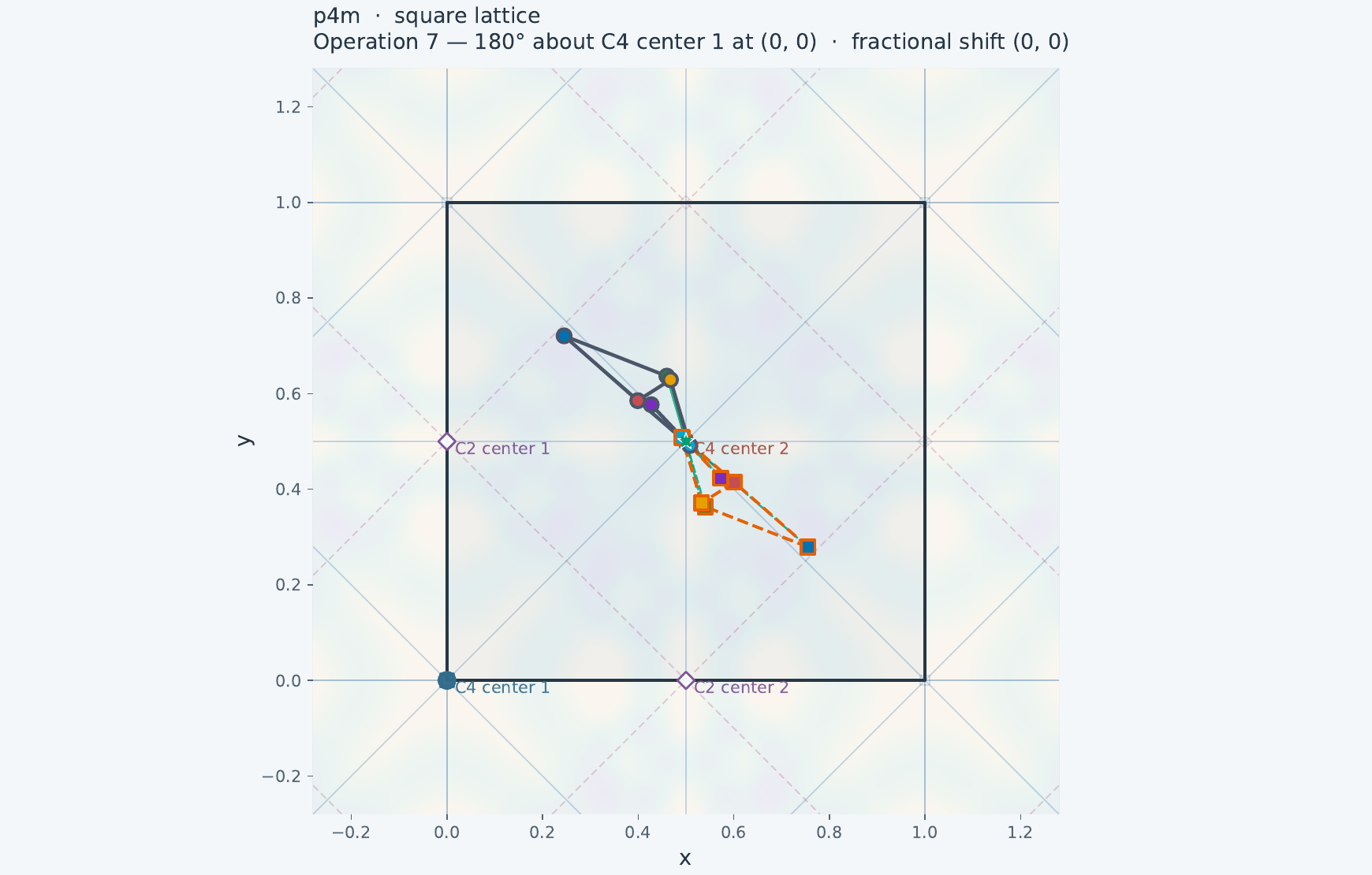} \\[2mm]
\includegraphics[width=\linewidth,height=0.22\textheight,keepaspectratio]{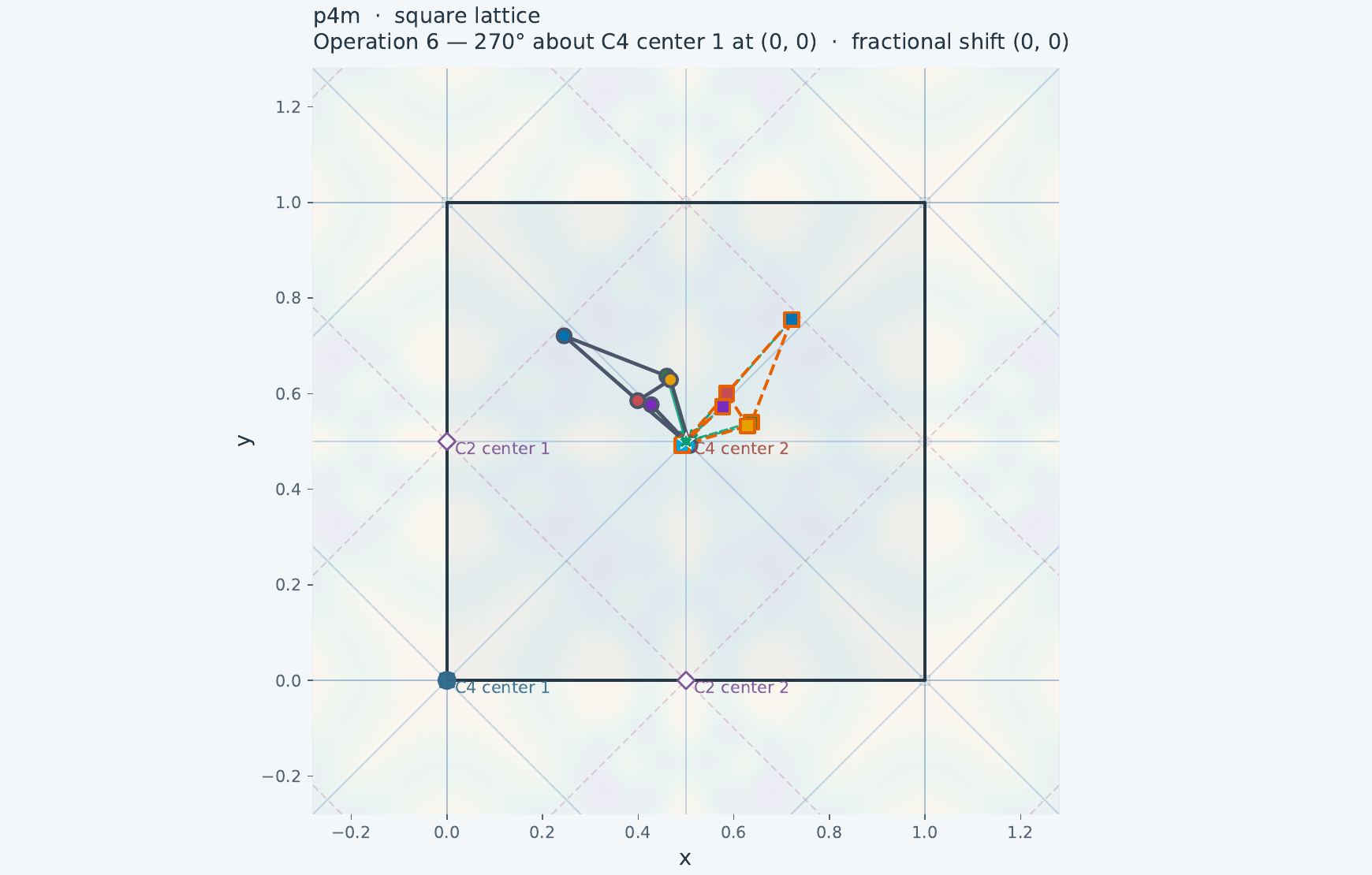} & \includegraphics[width=\linewidth,height=0.22\textheight,keepaspectratio]{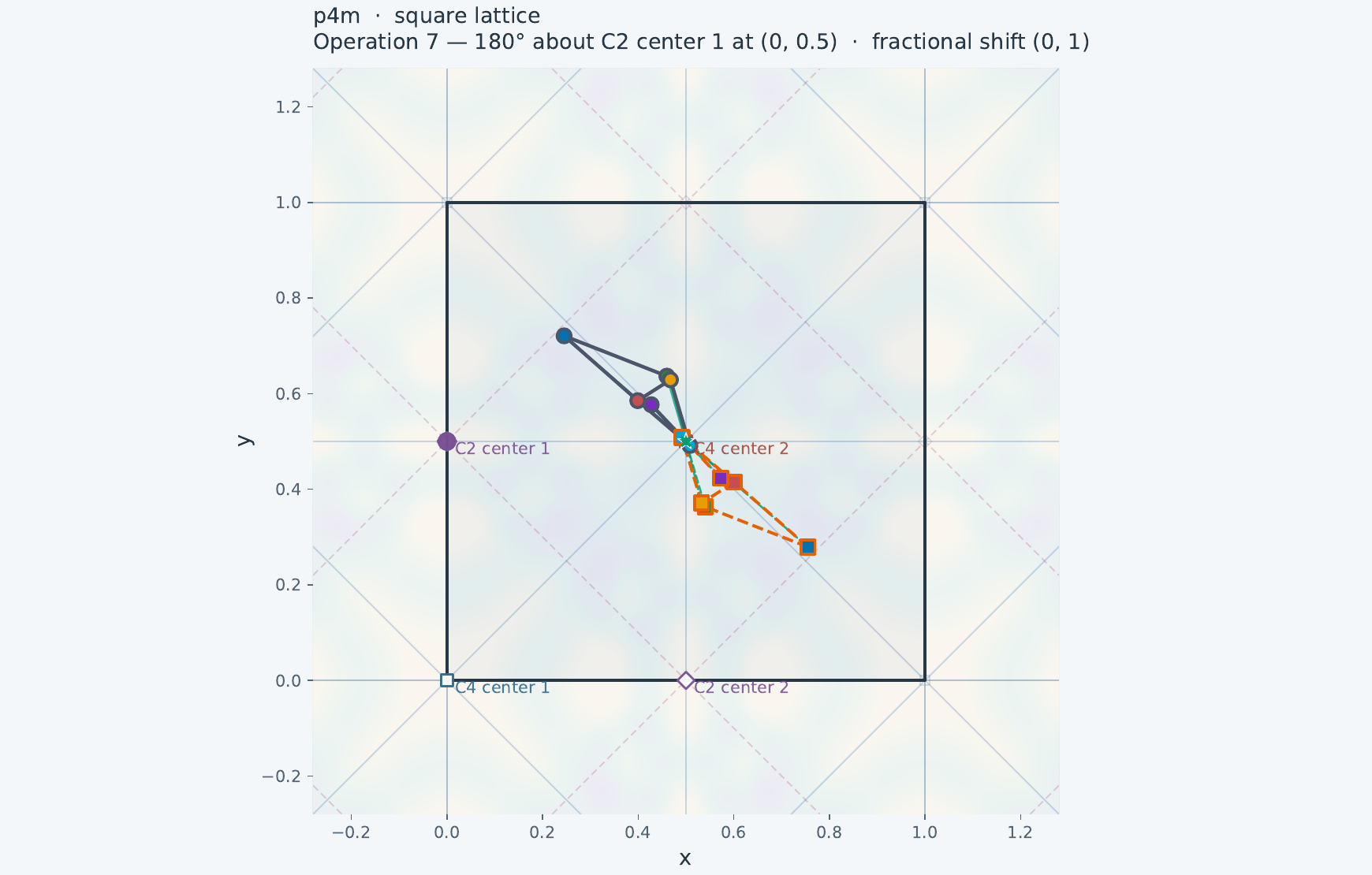} \\
\end{longtable}
\Needspace{218pt}
\begin{longtable}{@{}>{\centering\arraybackslash}p{0.475\linewidth}>{\centering\arraybackslash}p{0.475\linewidth}@{}}
\caption[Wallpaper group \texttt{p4g}]{Wallpaper group \texttt{p4g}, {\small WA is pointing to C4 at $q_{1}=(\frac{1}{2},\frac{1}{2}),\; q_{2}=(0,0)$}}\label{tab:wa-actions-p4g}\\[1mm]
\multicolumn{2}{c}{\includegraphics[width=0.96\linewidth]{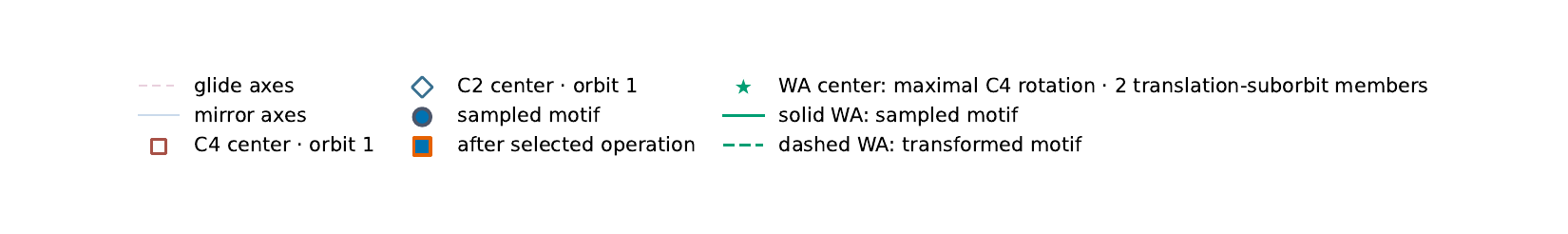}} \\[1mm]
\endfirsthead
\caption[]{Wallpaper group \texttt{p4g} (continued)}\\[1mm]
\endhead
\includegraphics[width=\linewidth,height=0.22\textheight,keepaspectratio]{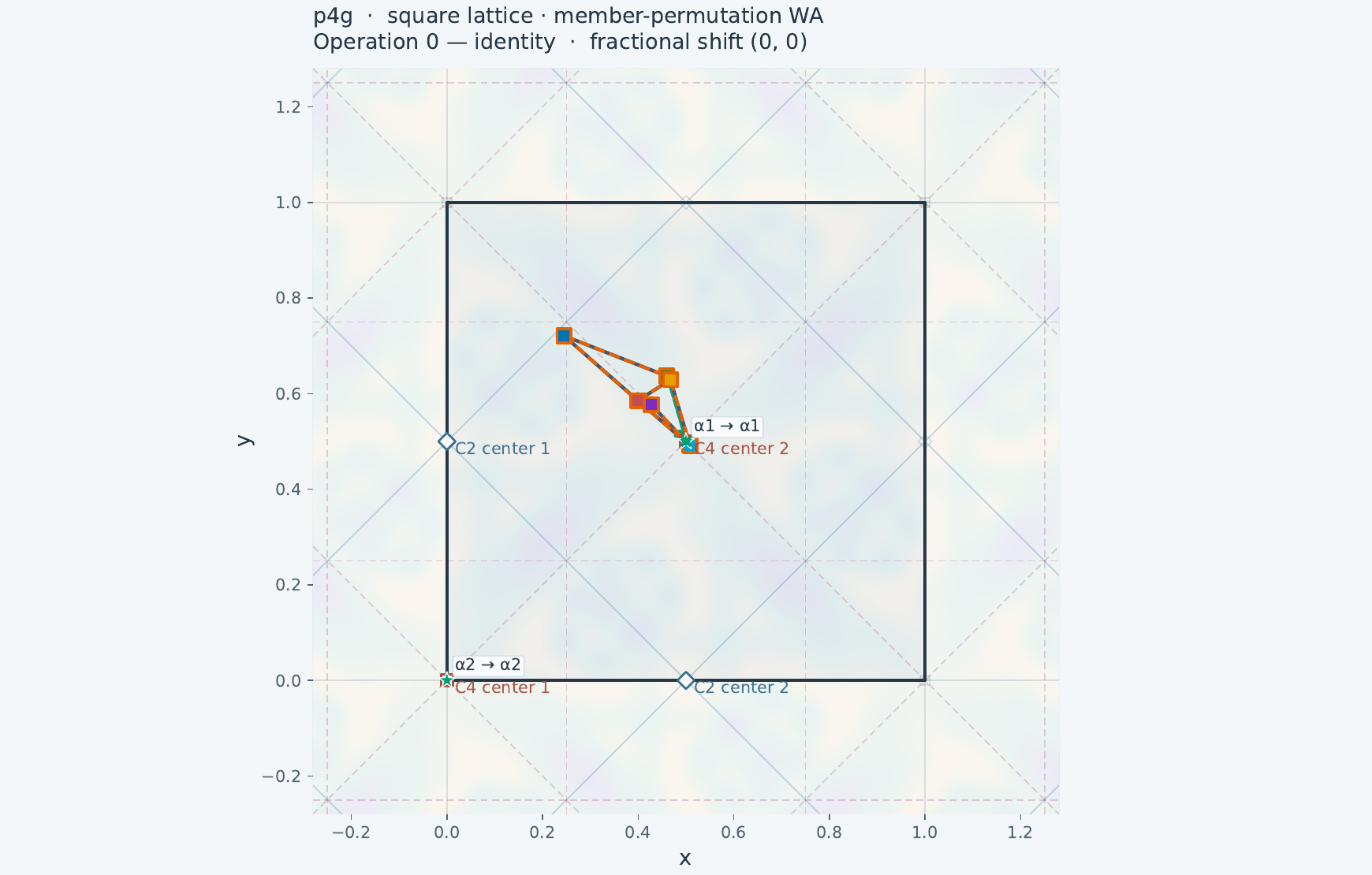} & \includegraphics[width=\linewidth,height=0.22\textheight,keepaspectratio]{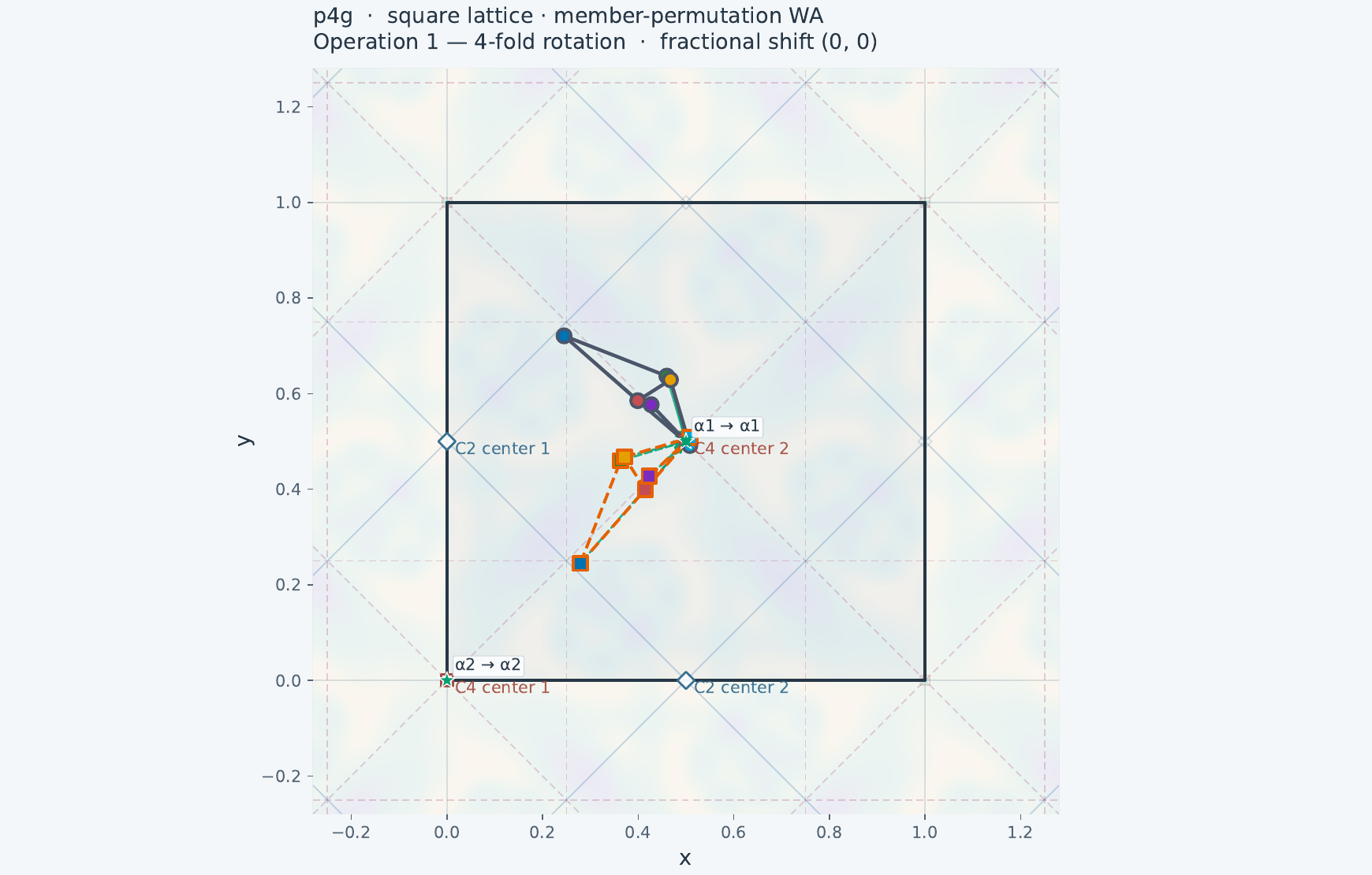} \\[2mm]
\includegraphics[width=\linewidth,height=0.22\textheight,keepaspectratio]{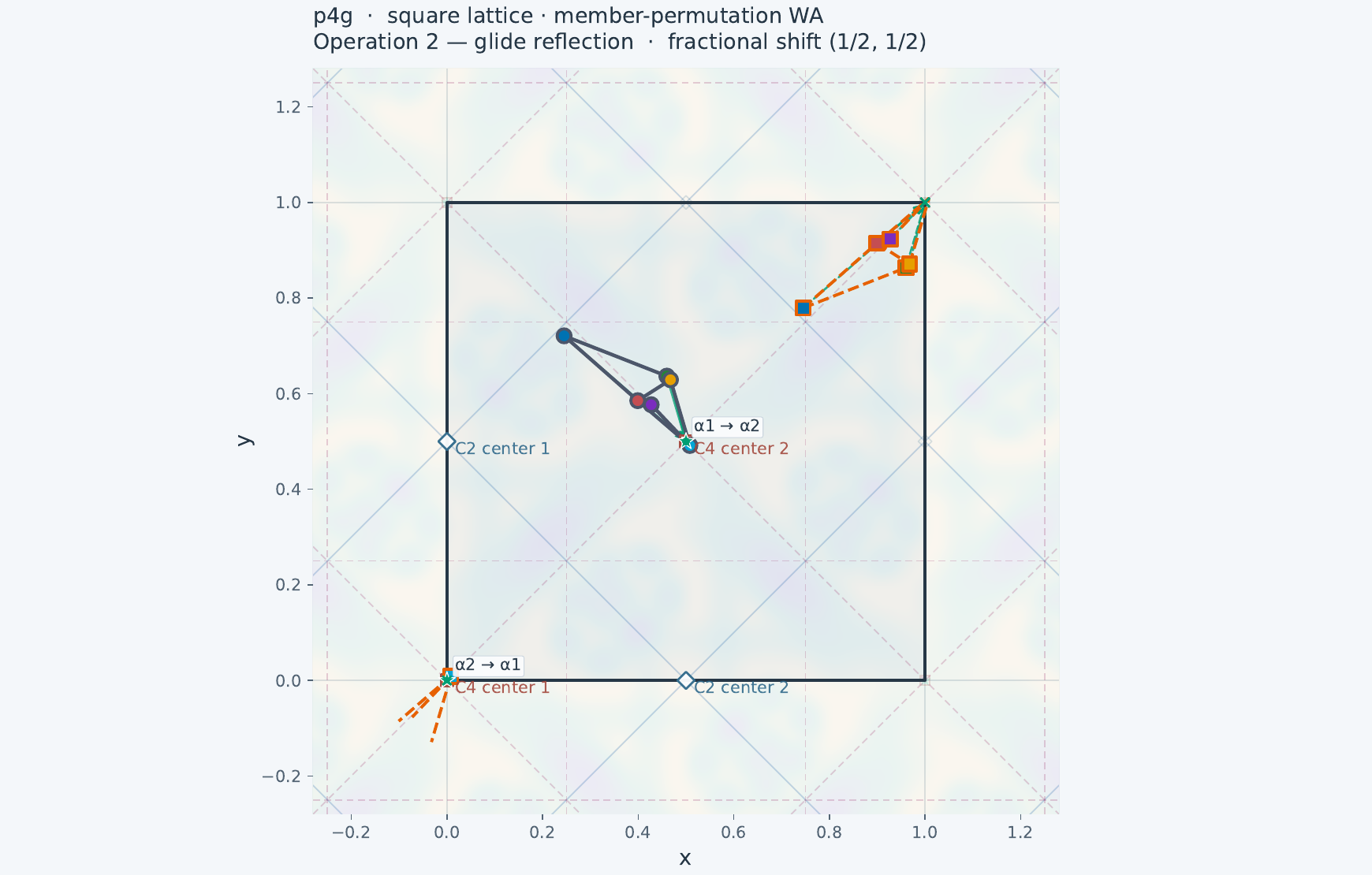} & \includegraphics[width=\linewidth,height=0.22\textheight,keepaspectratio]{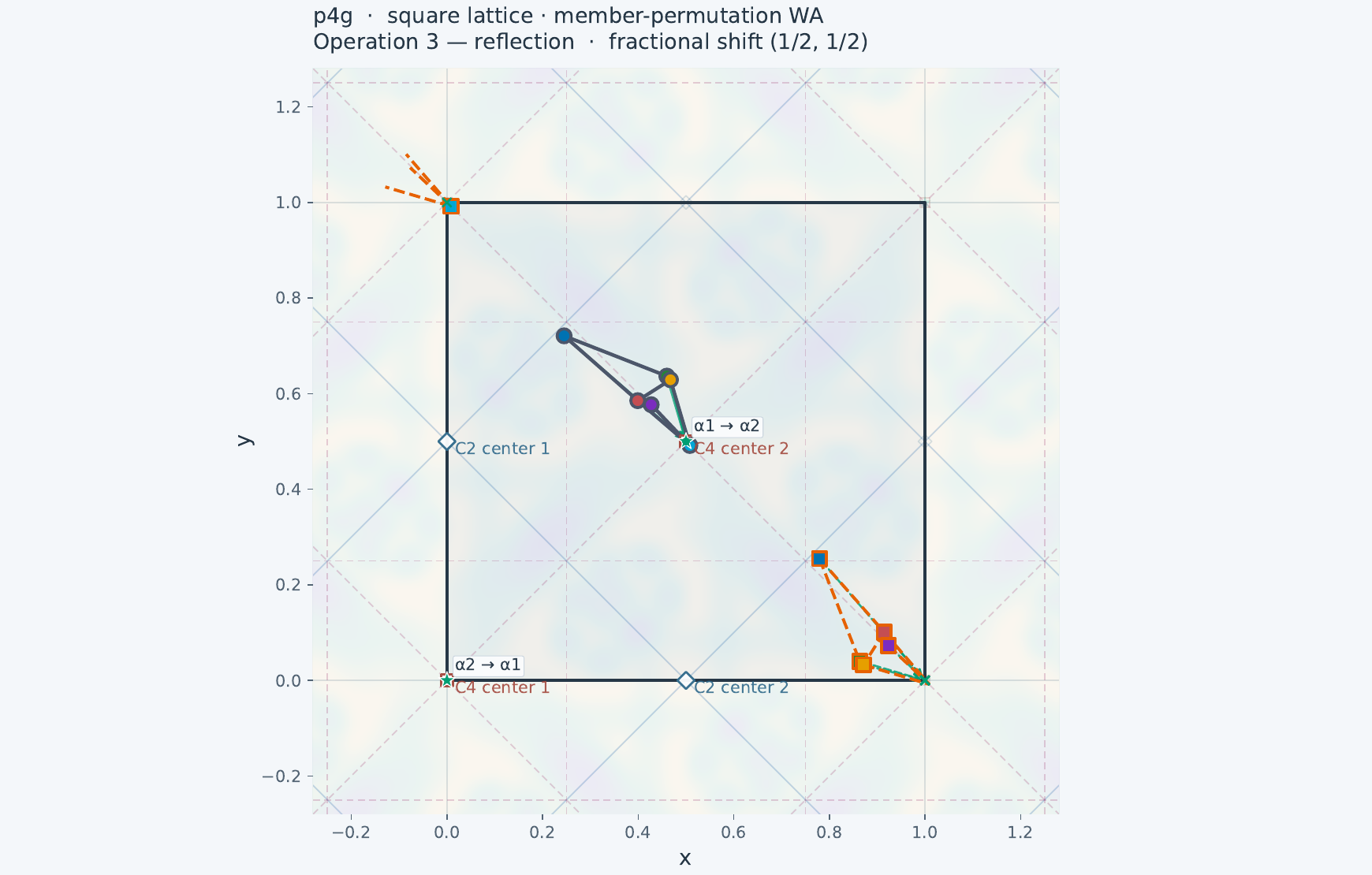} \\[2mm]
\includegraphics[width=\linewidth,height=0.22\textheight,keepaspectratio]{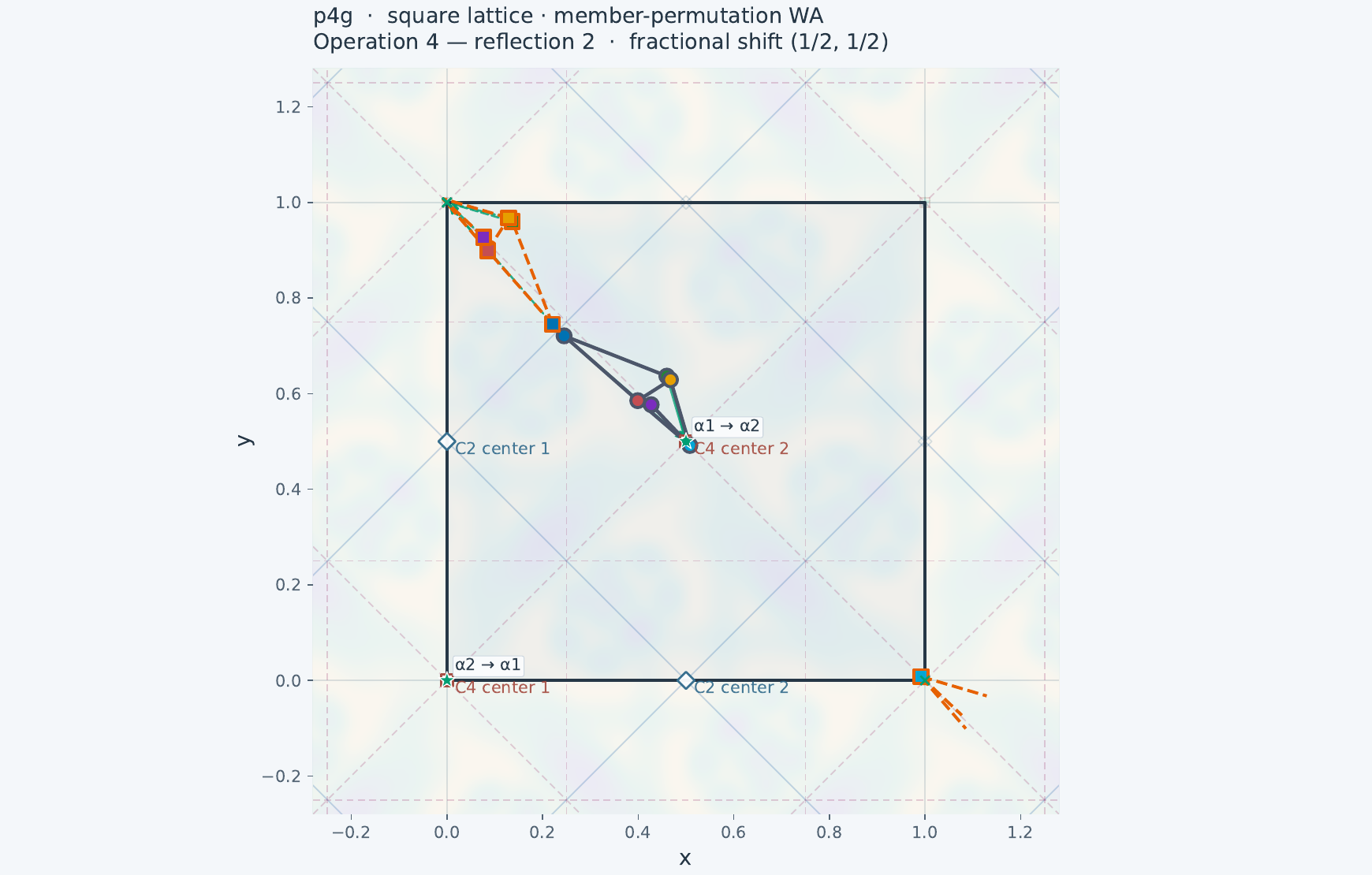} & \includegraphics[width=\linewidth,height=0.22\textheight,keepaspectratio]{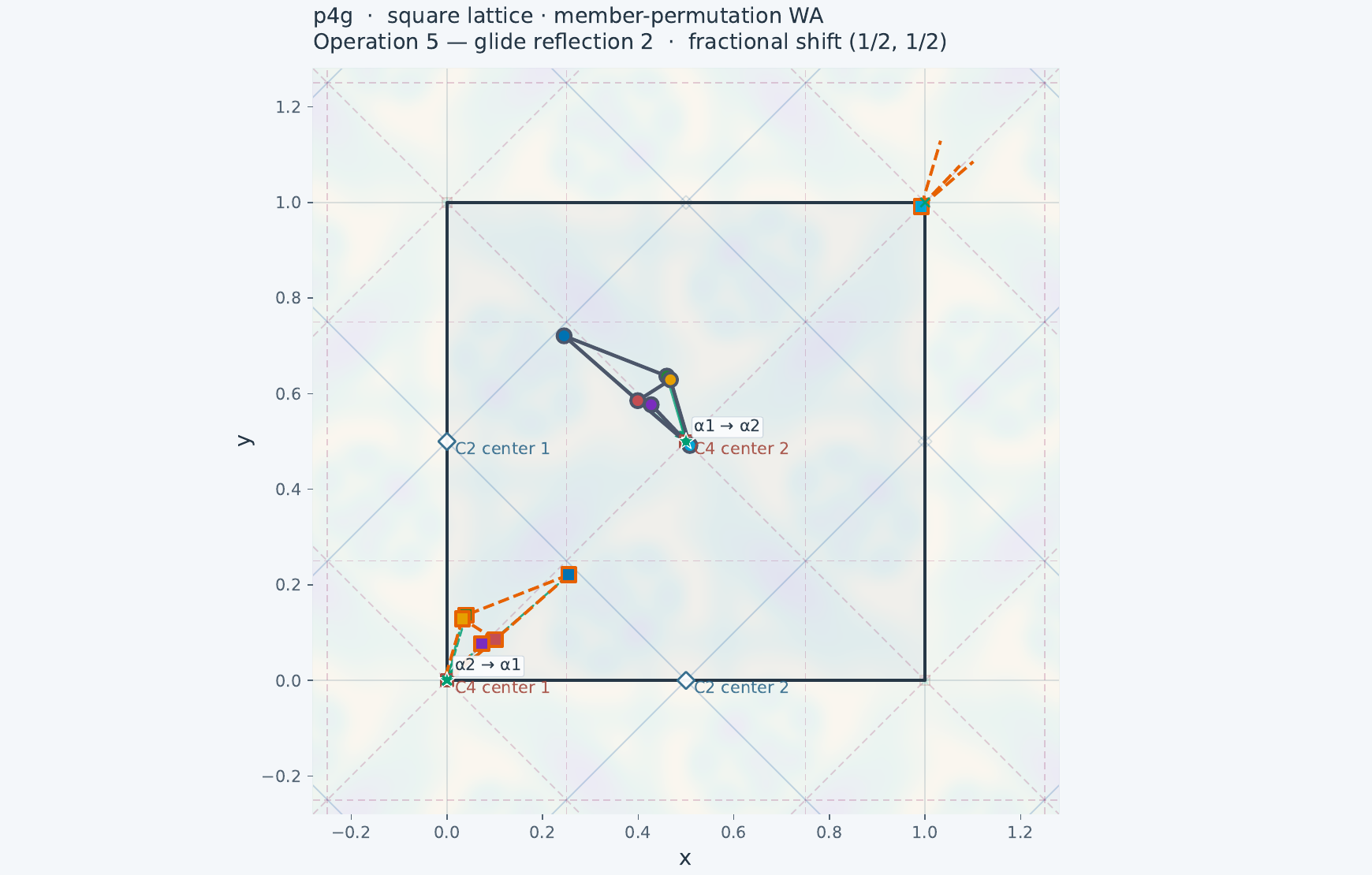} \\[2mm]
\includegraphics[width=\linewidth,height=0.22\textheight,keepaspectratio]{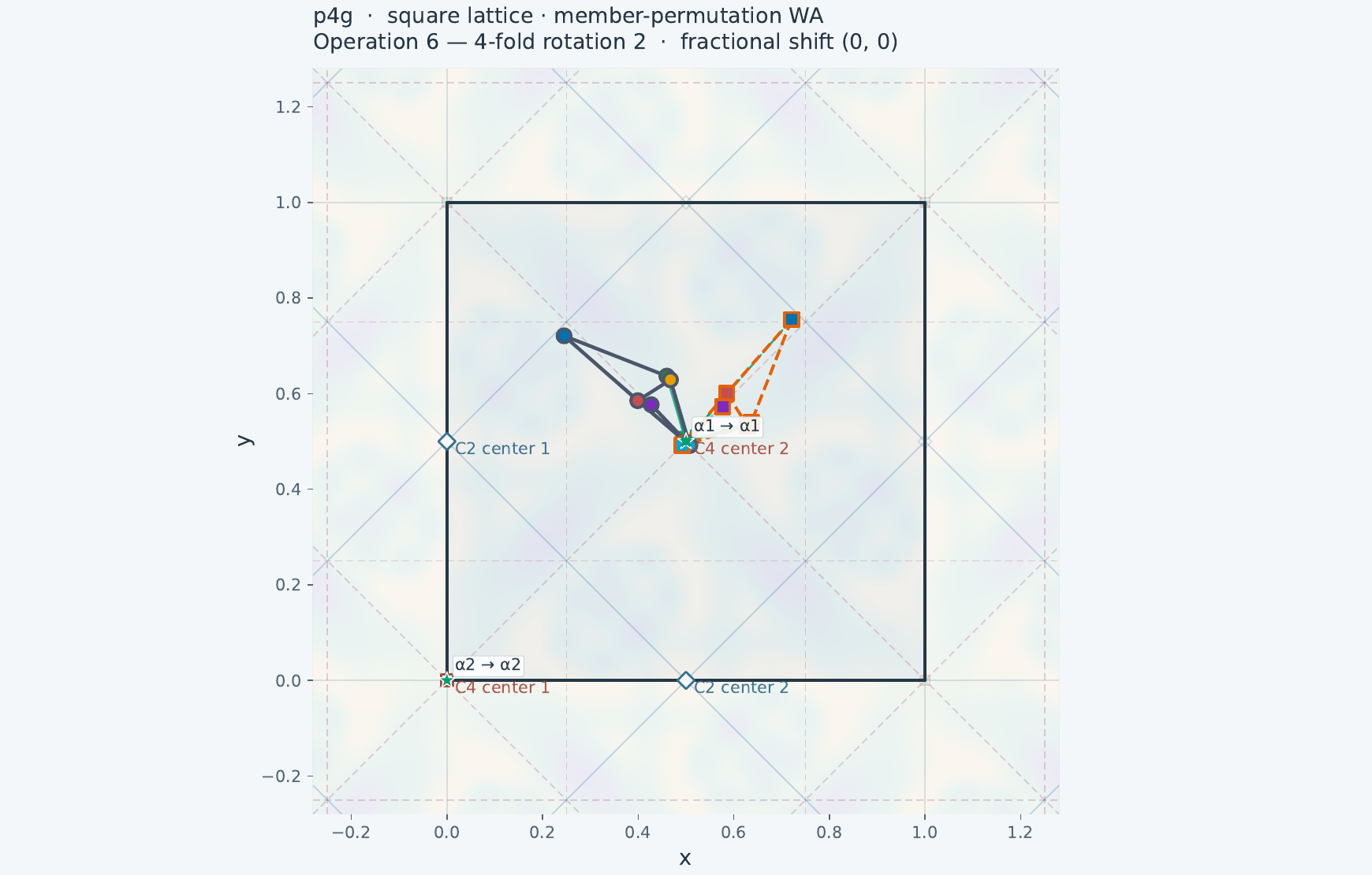} & \includegraphics[width=\linewidth,height=0.22\textheight,keepaspectratio]{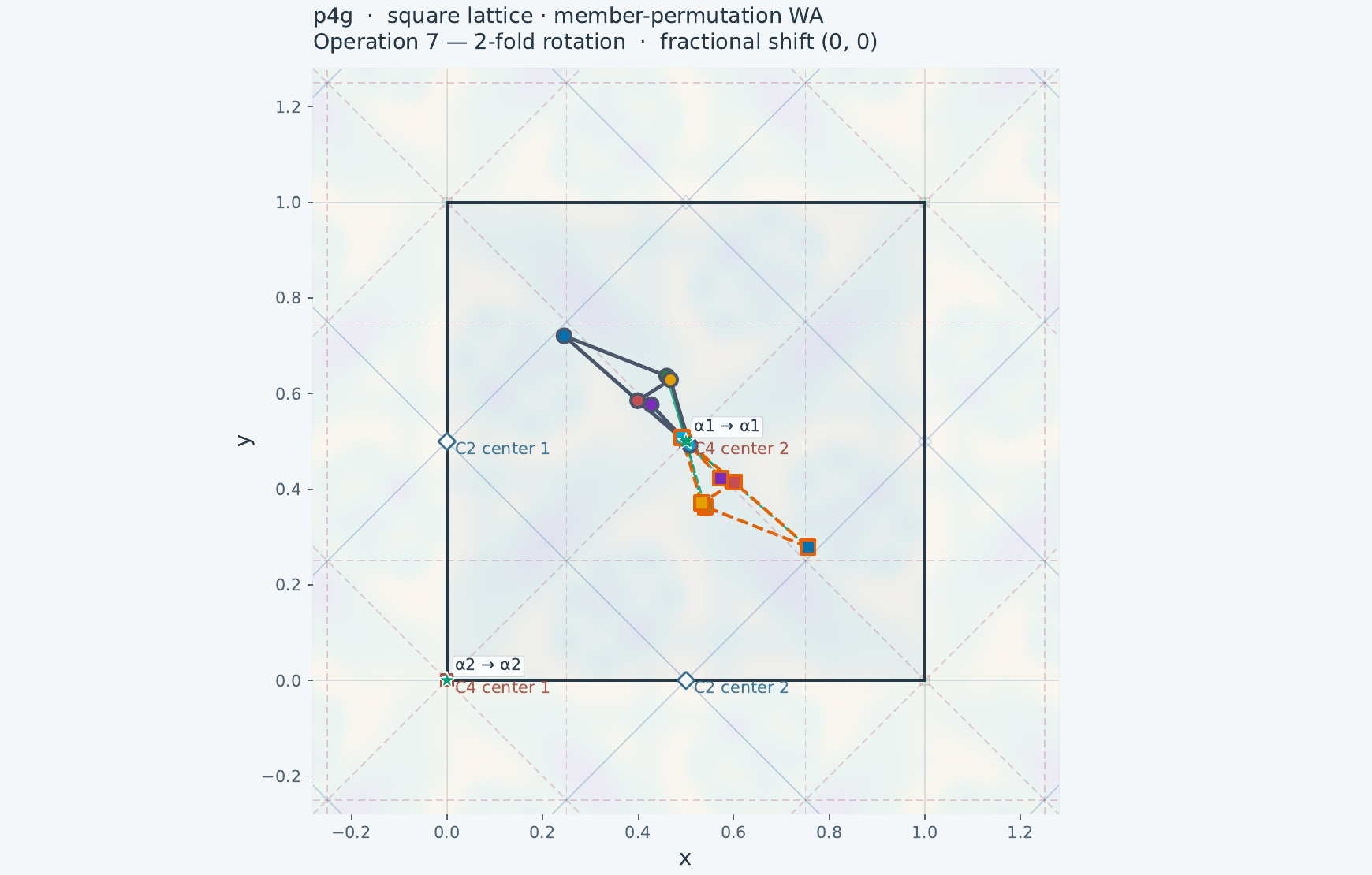} \\[2mm]
\includegraphics[width=\linewidth,height=0.22\textheight,keepaspectratio]{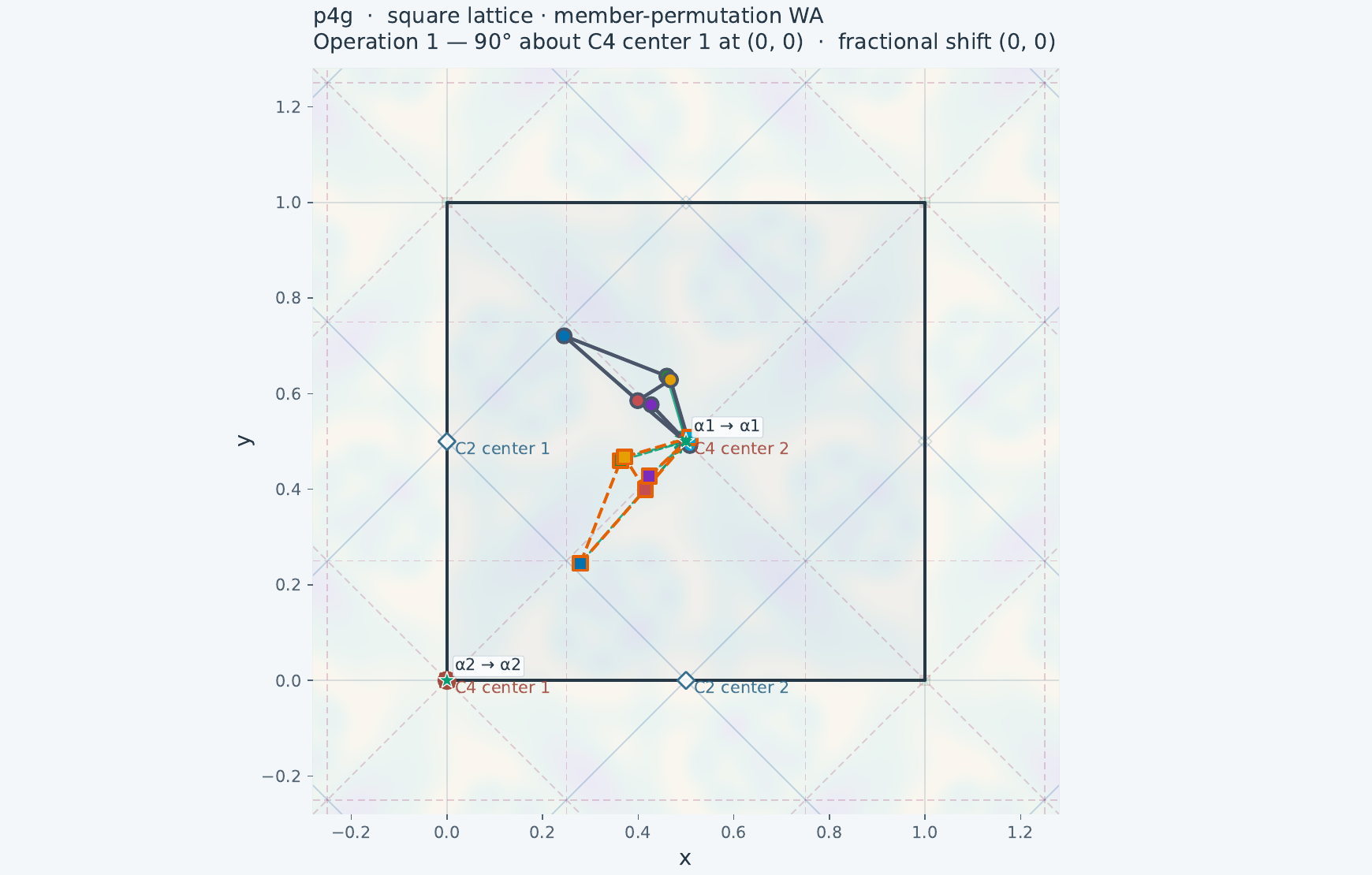} & \includegraphics[width=\linewidth,height=0.22\textheight,keepaspectratio]{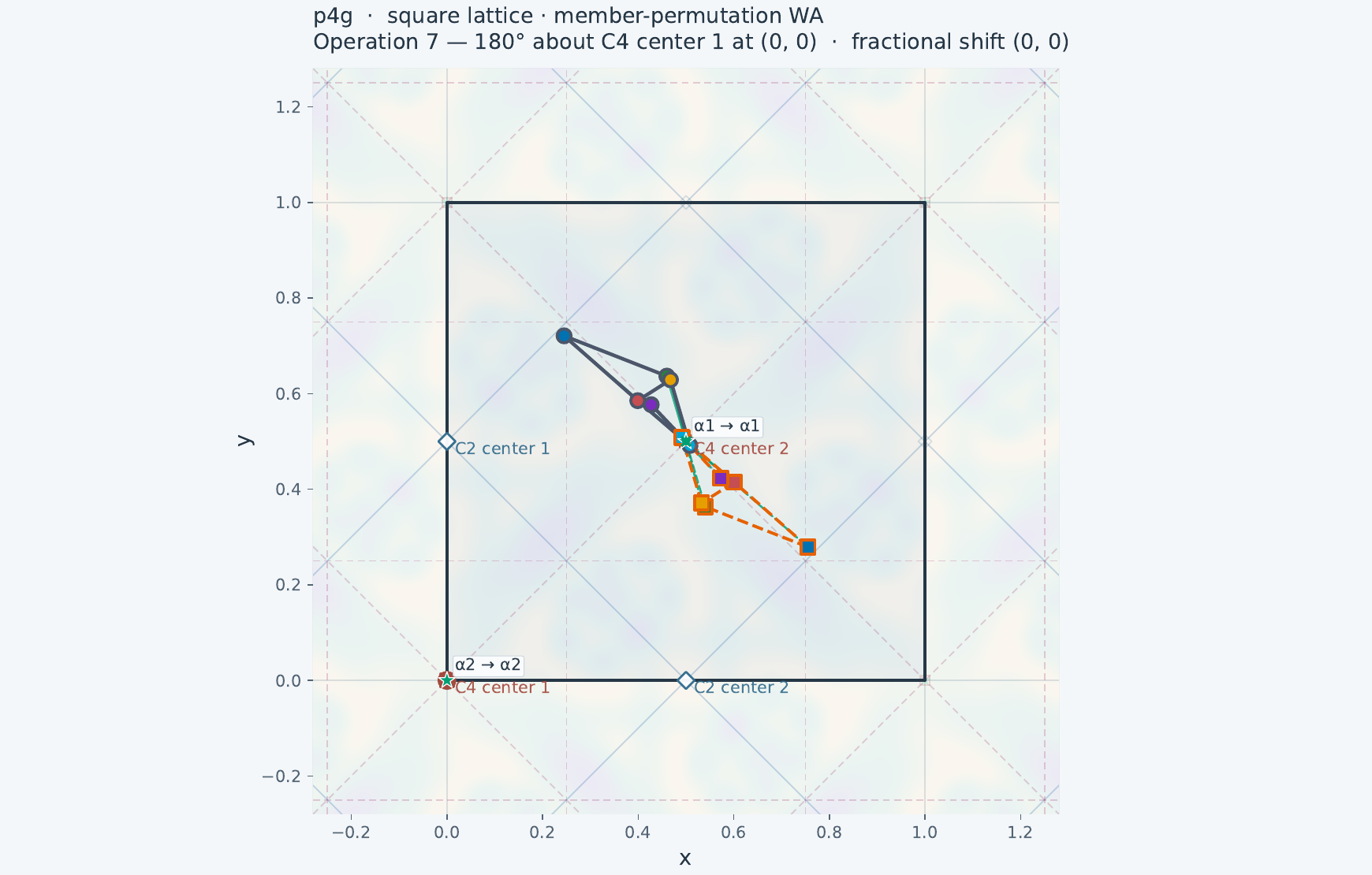} \\[2mm]
\includegraphics[width=\linewidth,height=0.22\textheight,keepaspectratio]{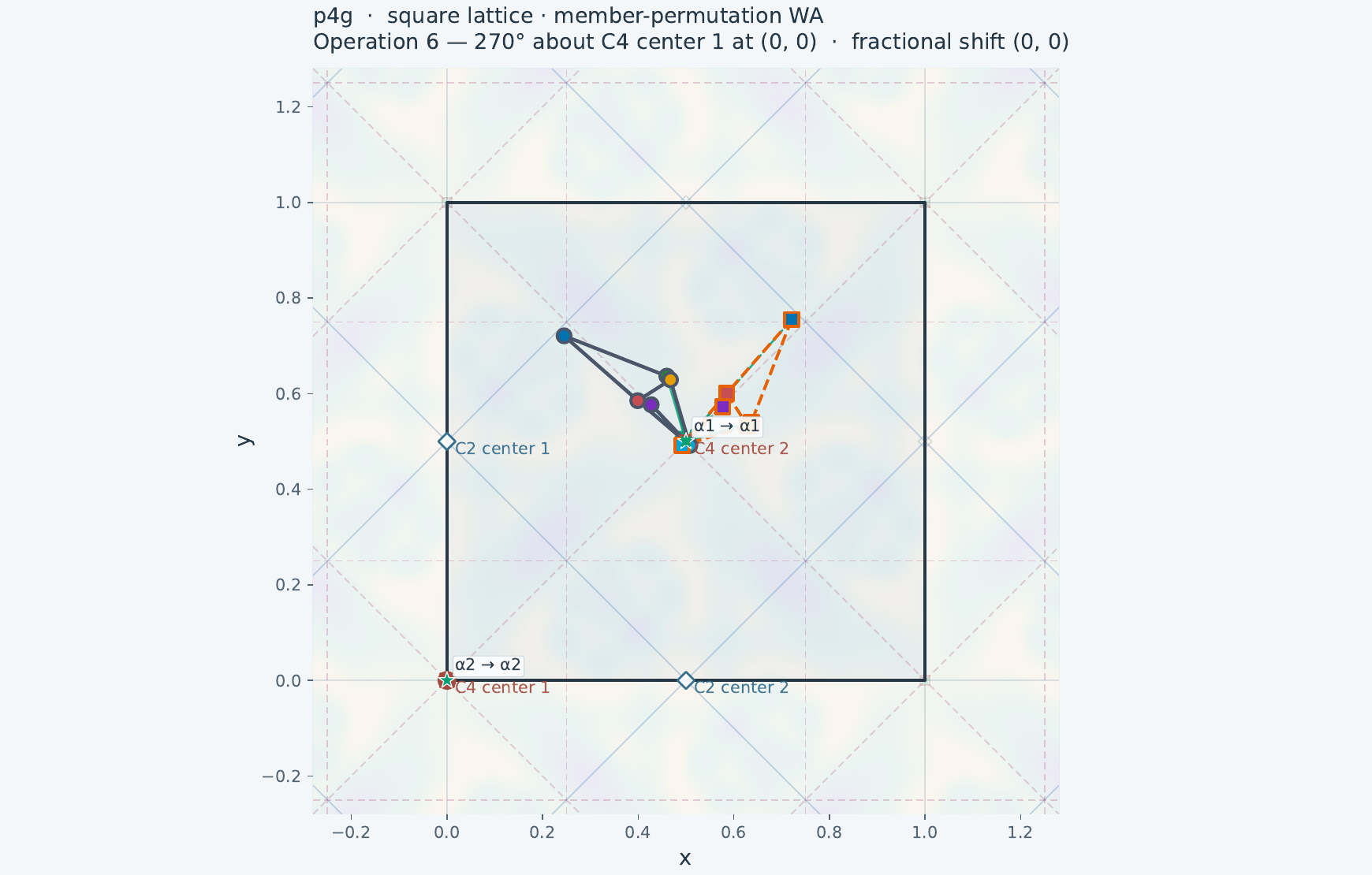} & \includegraphics[width=\linewidth,height=0.22\textheight,keepaspectratio]{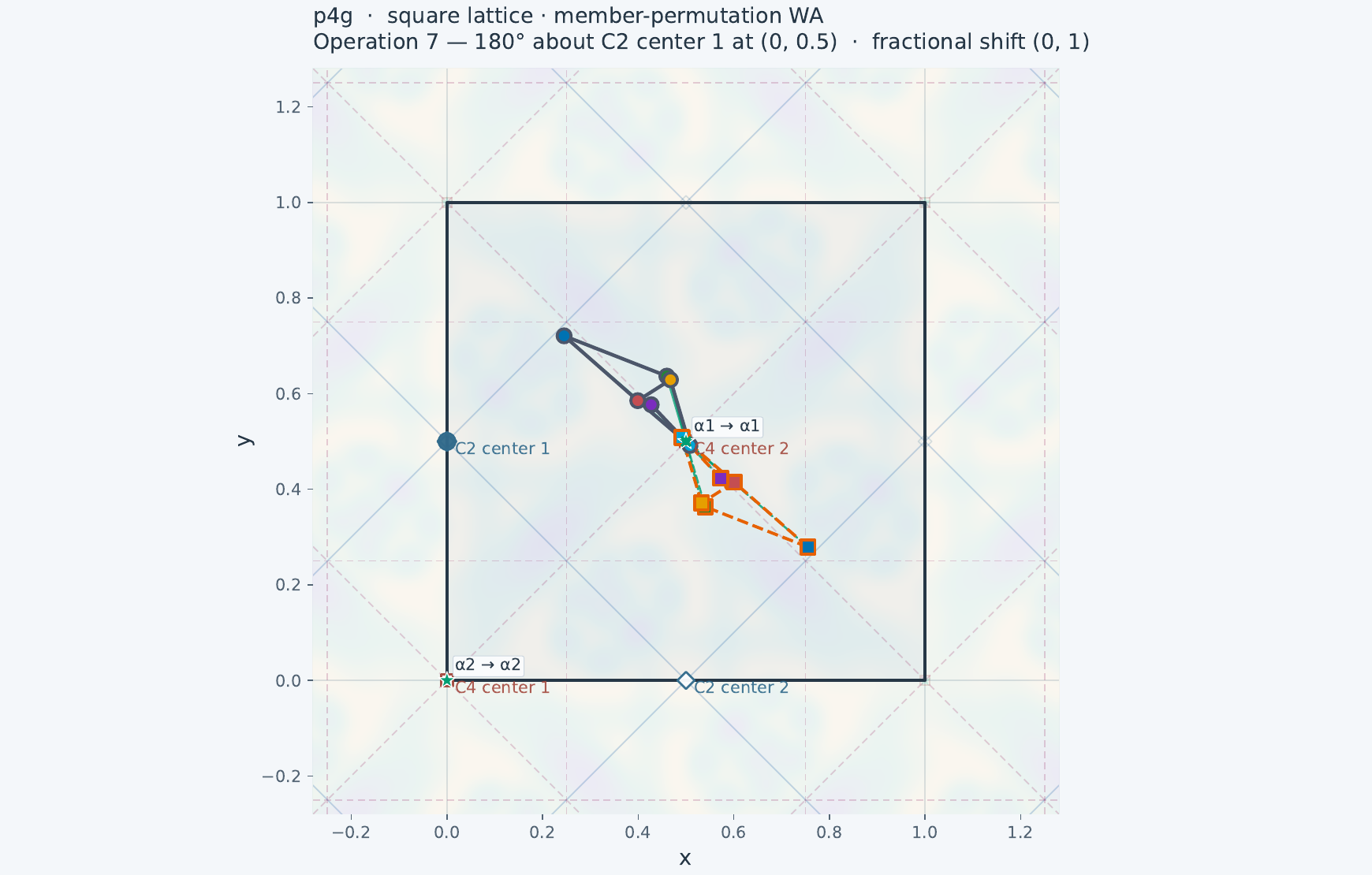} \\
\end{longtable}
\Needspace{218pt}
\begin{longtable}{@{}>{\centering\arraybackslash}p{0.475\linewidth}>{\centering\arraybackslash}p{0.475\linewidth}@{}}
\caption[Wallpaper group \texttt{p3}]{Wallpaper group \texttt{p3}, {\small WA is pointing to C3-A at $q_{1}=(\frac{1}{3},\frac{2}{3})$}}\label{tab:wa-actions-p3}\\[1mm]
\multicolumn{2}{c}{\includegraphics[width=0.96\linewidth]{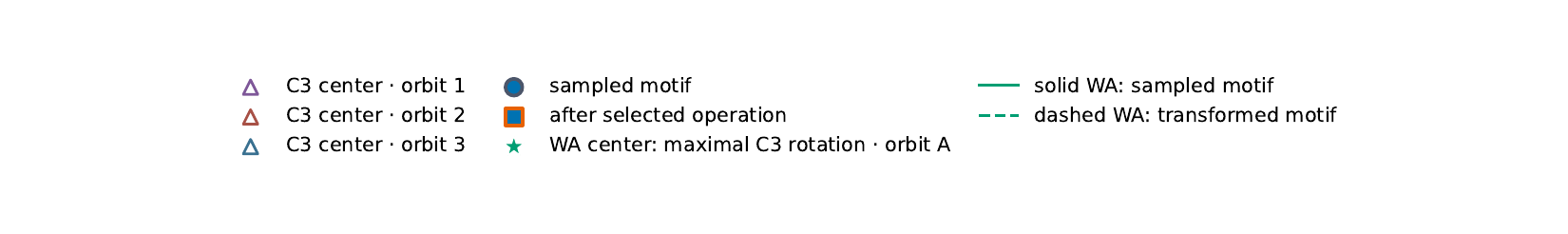}} \\[1mm]
\endfirsthead
\caption[]{Wallpaper group \texttt{p3} (continued)}\\[1mm]
\endhead
\includegraphics[width=\linewidth,height=0.22\textheight,keepaspectratio]{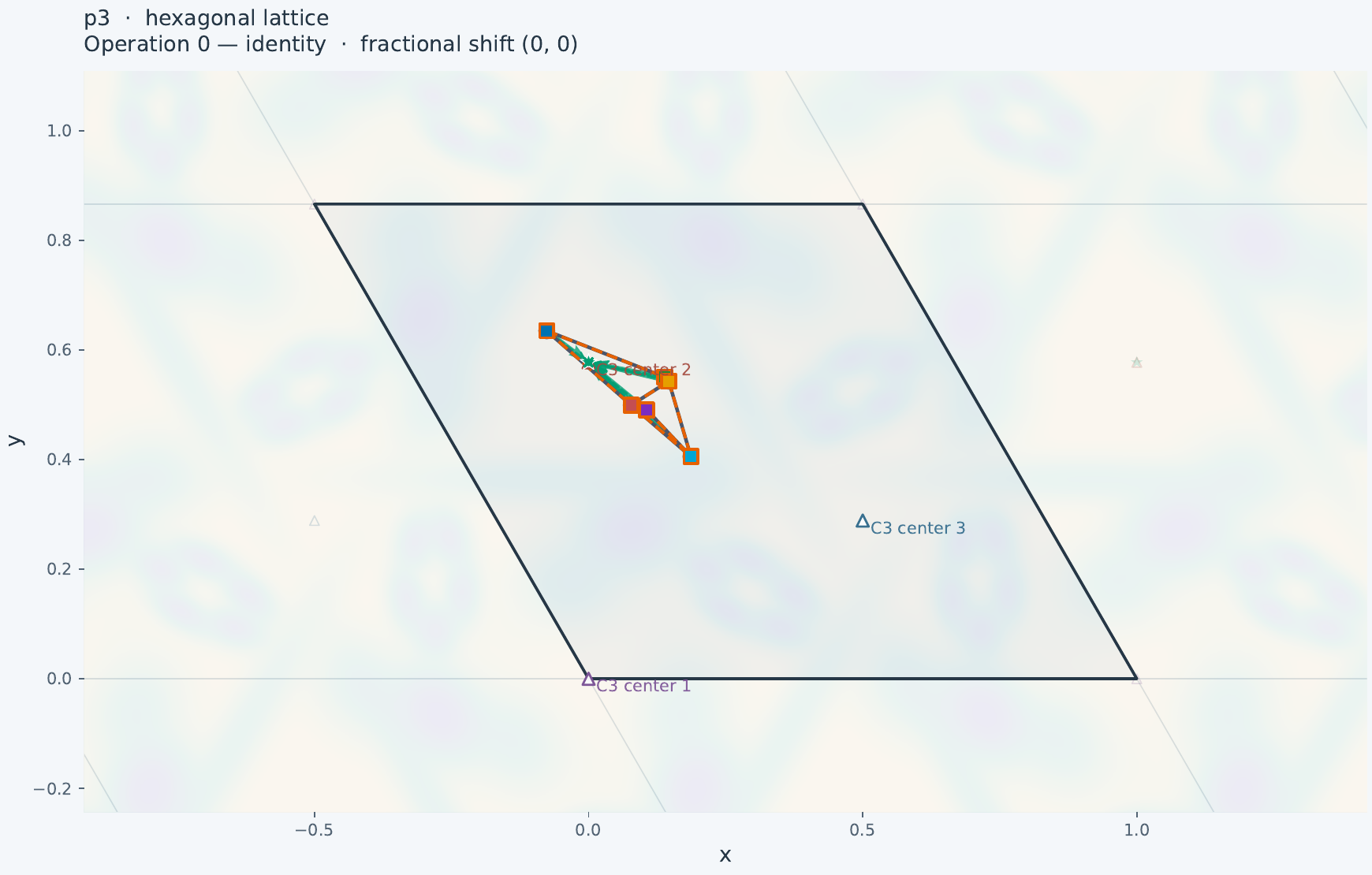} & \includegraphics[width=\linewidth,height=0.22\textheight,keepaspectratio]{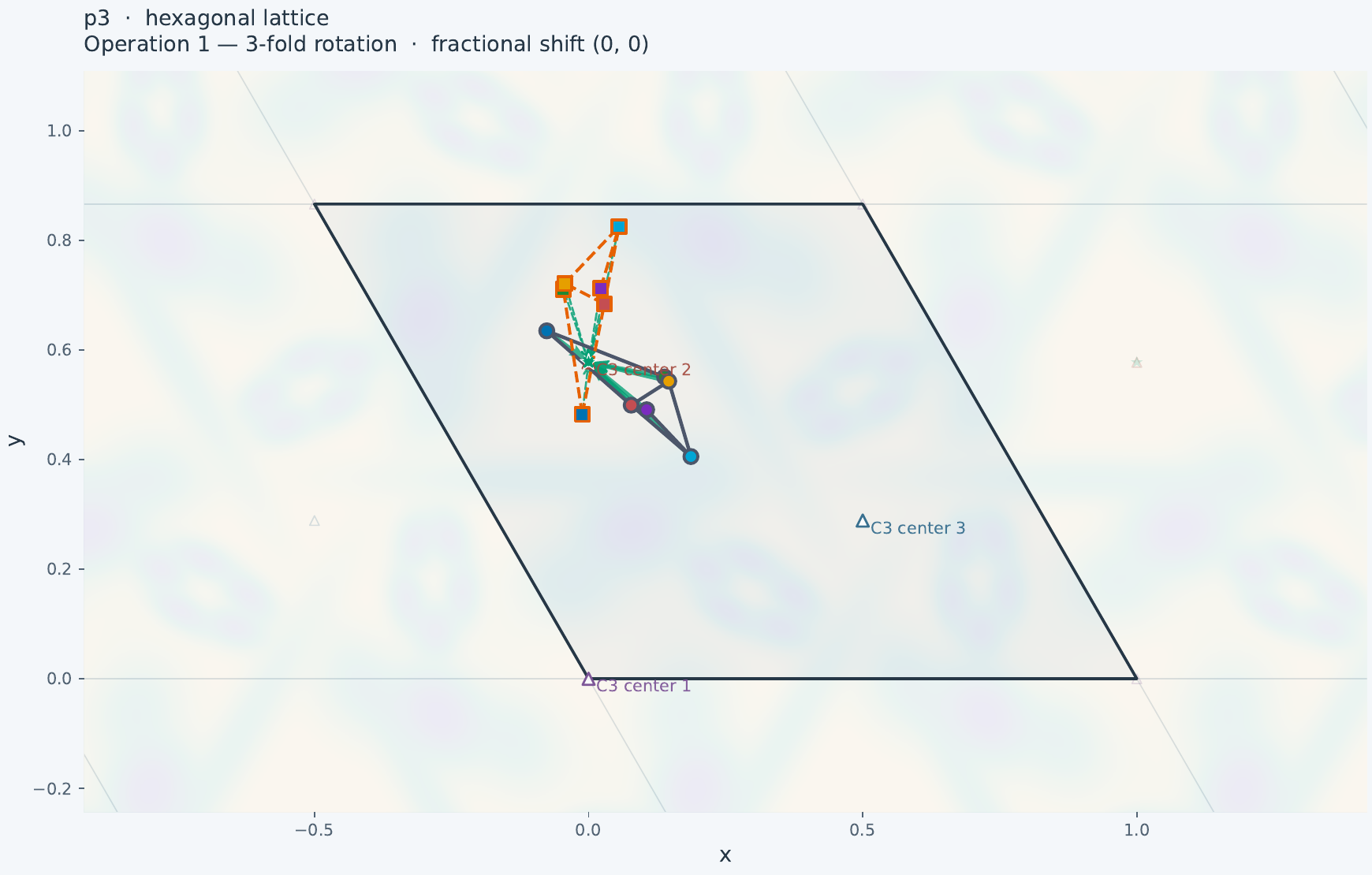} \\[2mm]
\includegraphics[width=\linewidth,height=0.22\textheight,keepaspectratio]{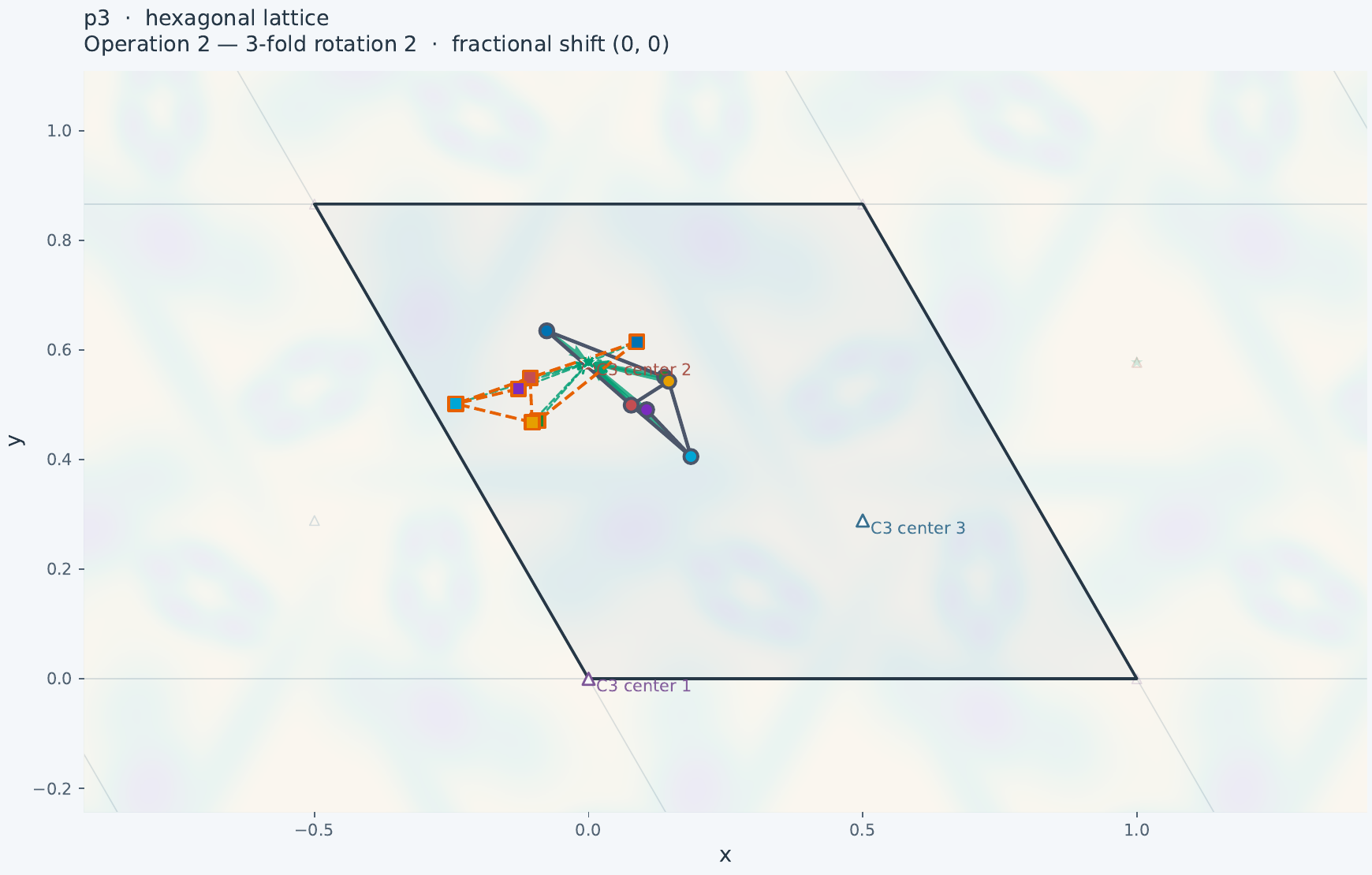} & \includegraphics[width=\linewidth,height=0.22\textheight,keepaspectratio]{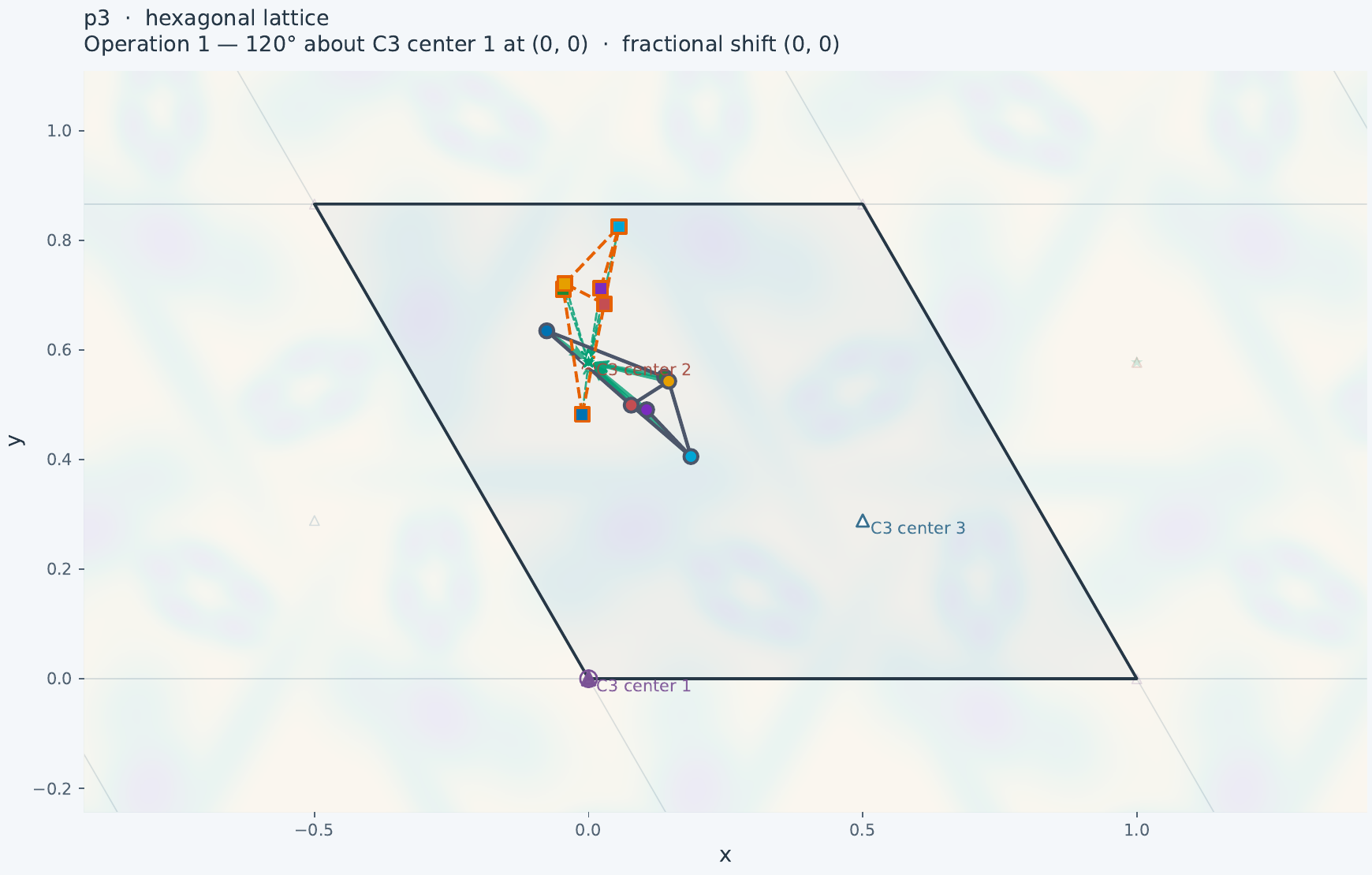} \\[2mm]
\includegraphics[width=\linewidth,height=0.22\textheight,keepaspectratio]{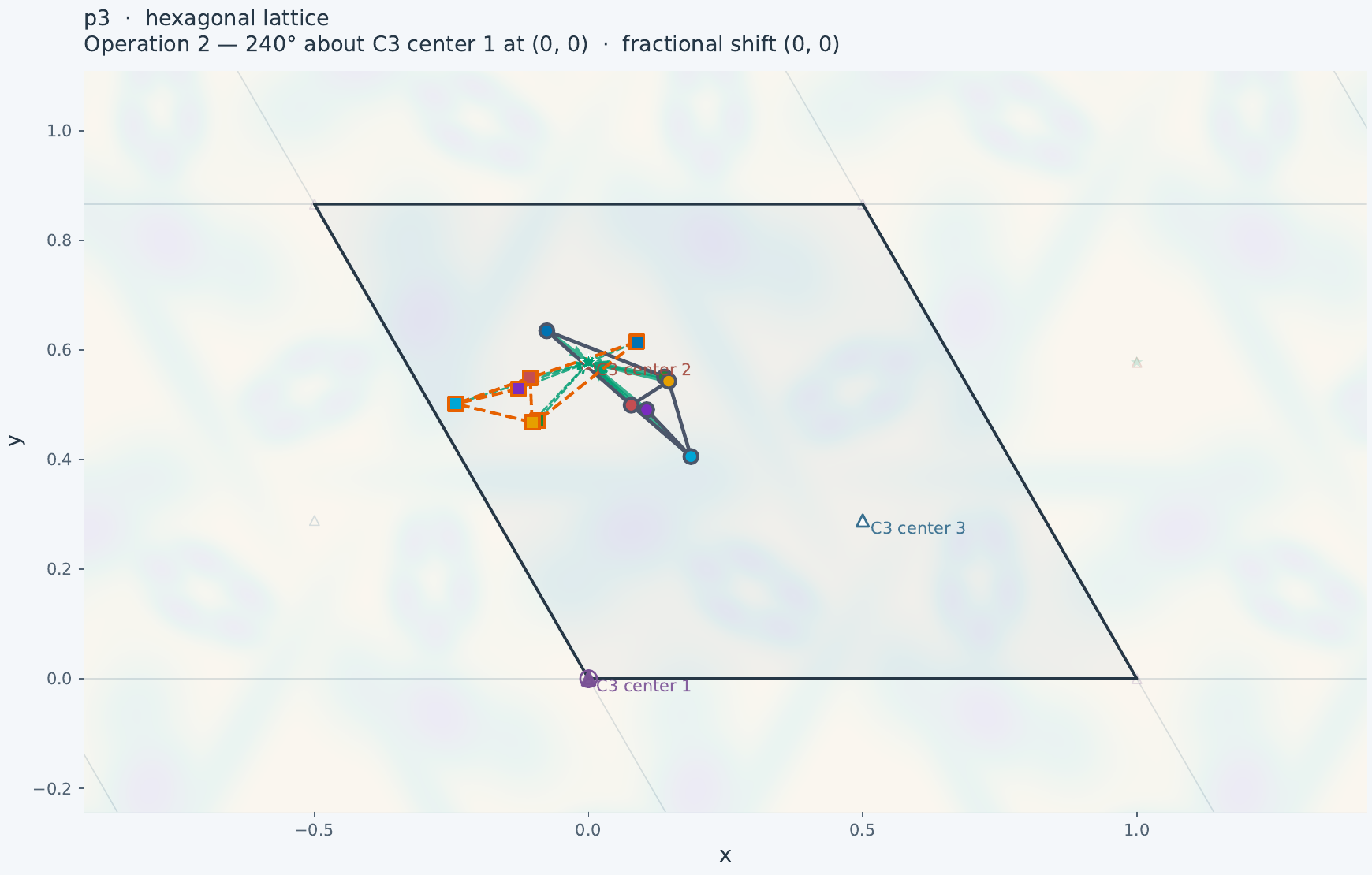} & \includegraphics[width=\linewidth,height=0.22\textheight,keepaspectratio]{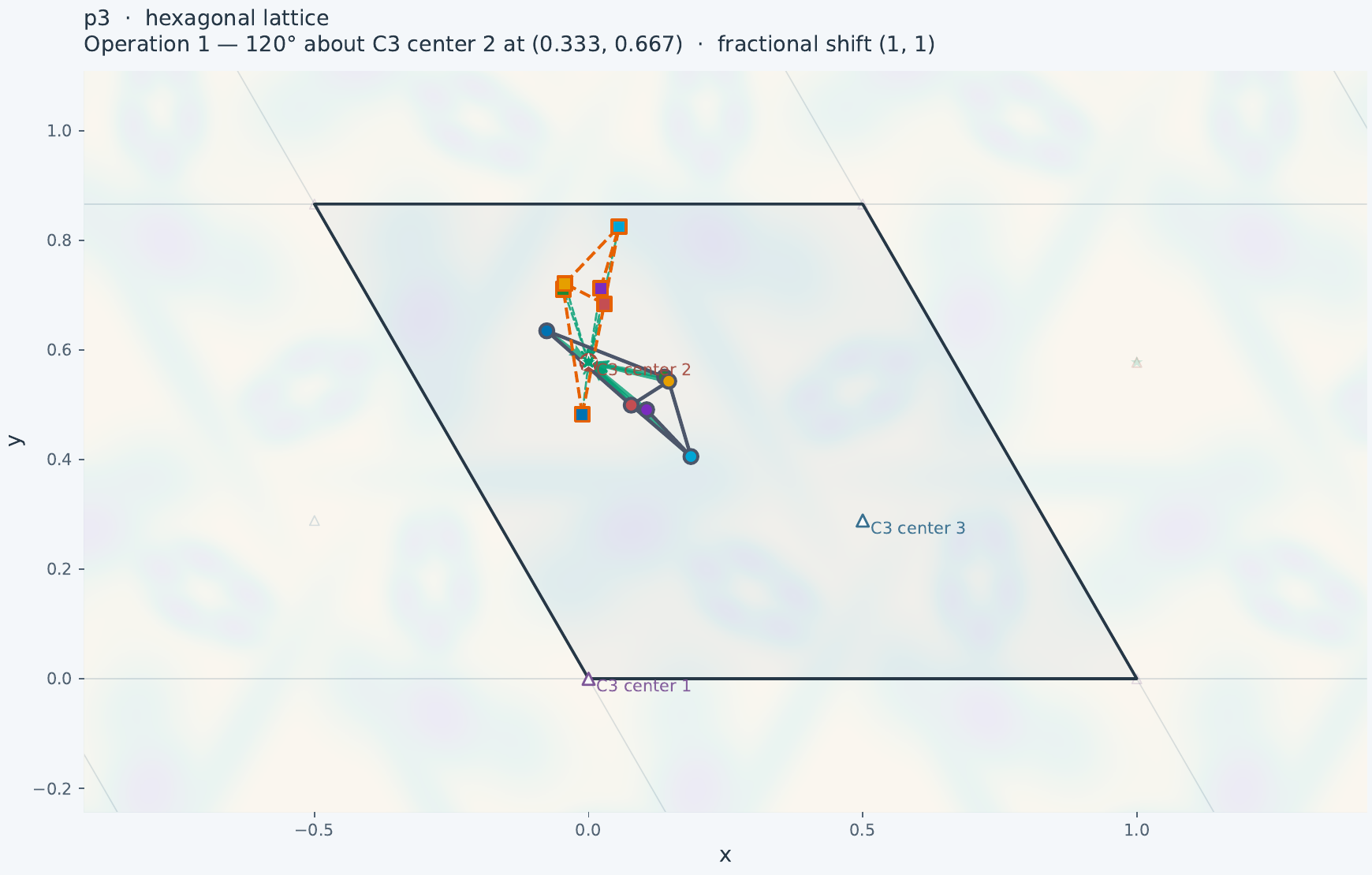} \\[2mm]
\includegraphics[width=\linewidth,height=0.22\textheight,keepaspectratio]{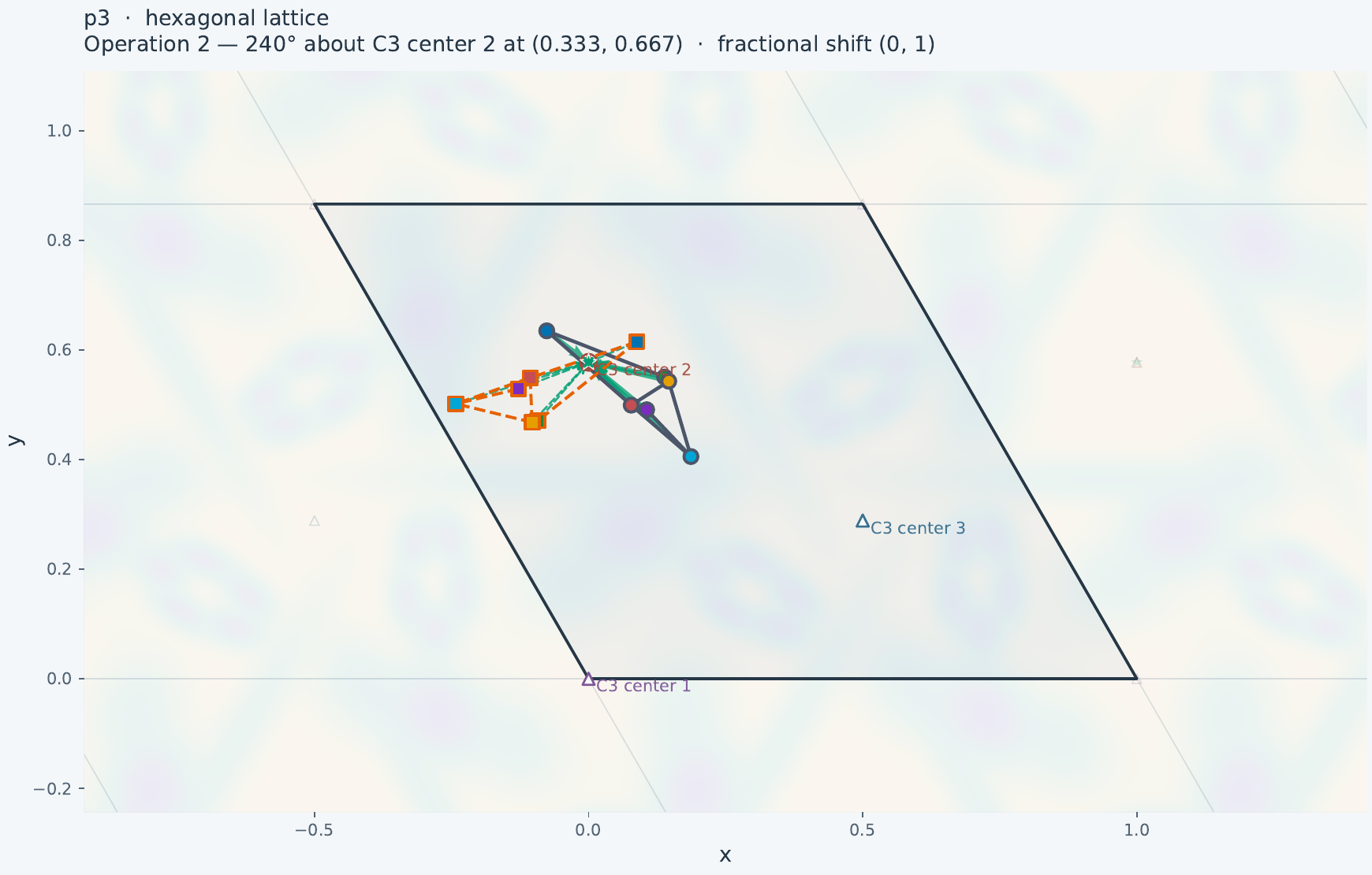} & \includegraphics[width=\linewidth,height=0.22\textheight,keepaspectratio]{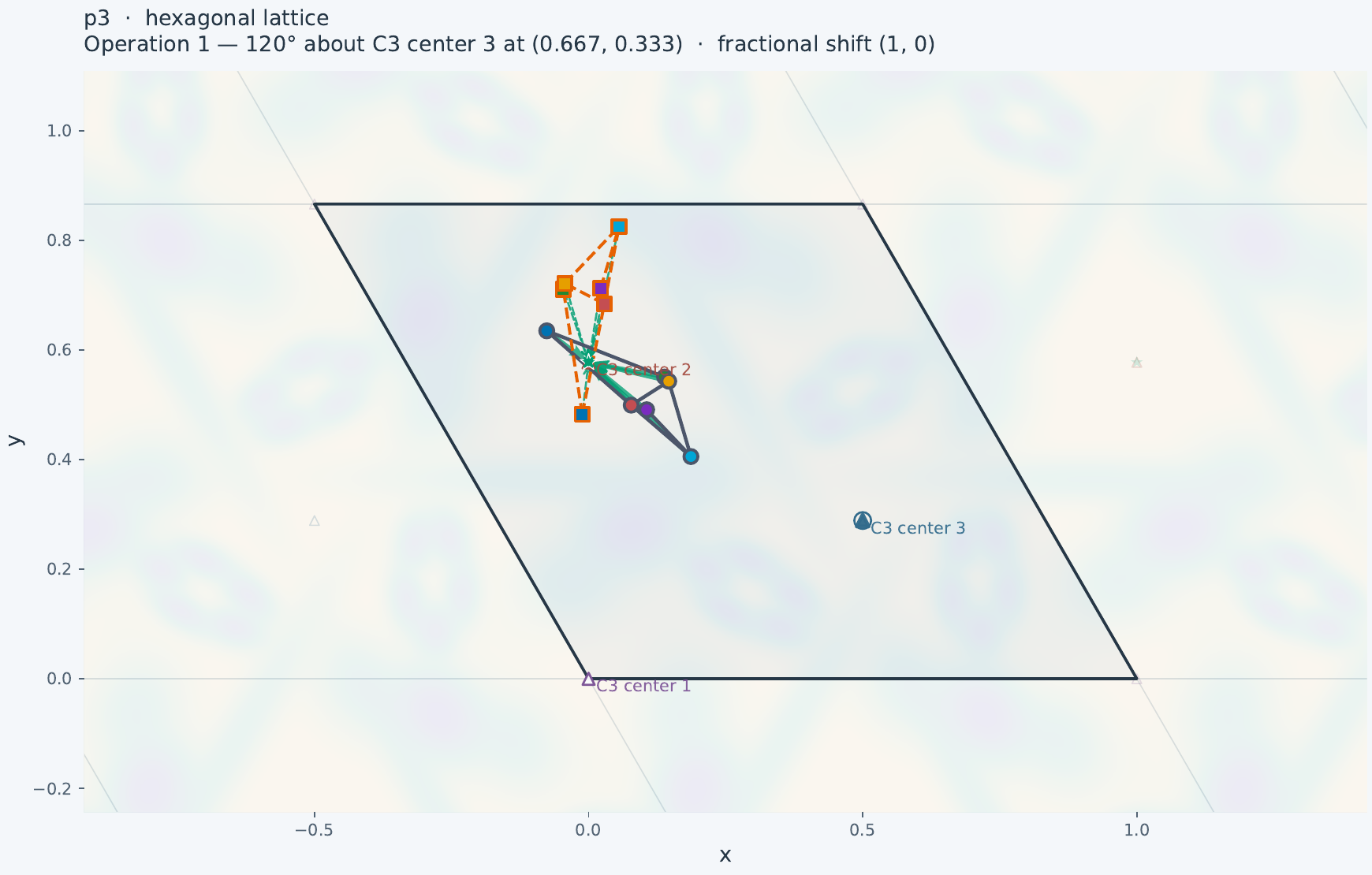} \\[2mm]
\includegraphics[width=\linewidth,height=0.22\textheight,keepaspectratio]{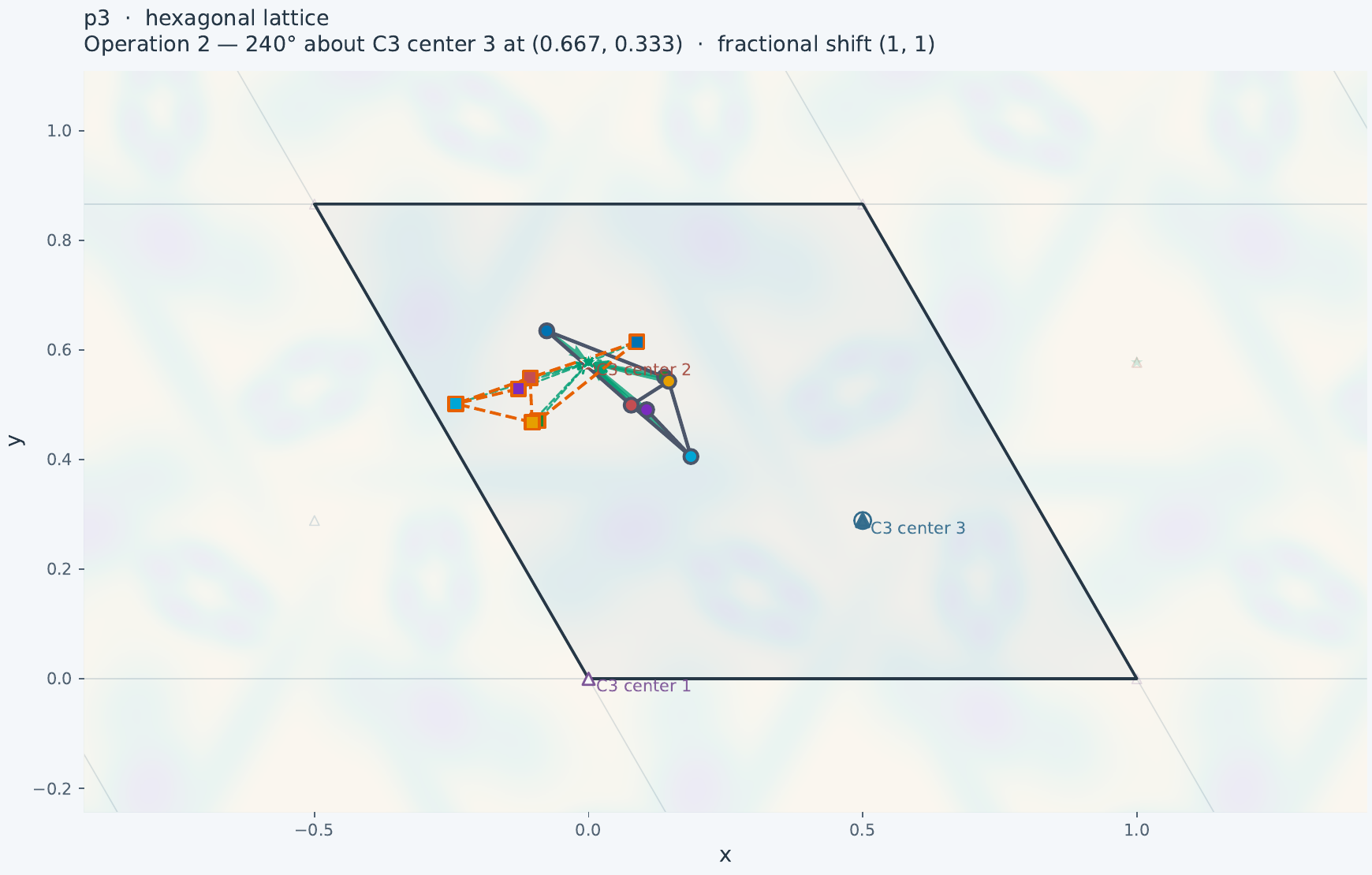} &  \\[2mm]
\end{longtable}
\Needspace{236pt}
\begin{longtable}{@{}>{\centering\arraybackslash}p{0.475\linewidth}>{\centering\arraybackslash}p{0.475\linewidth}@{}}
\caption[Wallpaper group \texttt{p3m1}]{Wallpaper group \texttt{p3m1}, {\small WA is pointing to C3-A at $q_{1}=(\frac{1}{3},\frac{2}{3})$}}\label{tab:wa-actions-p3m1}\\[1mm]
\multicolumn{2}{c}{\includegraphics[width=0.96\linewidth]{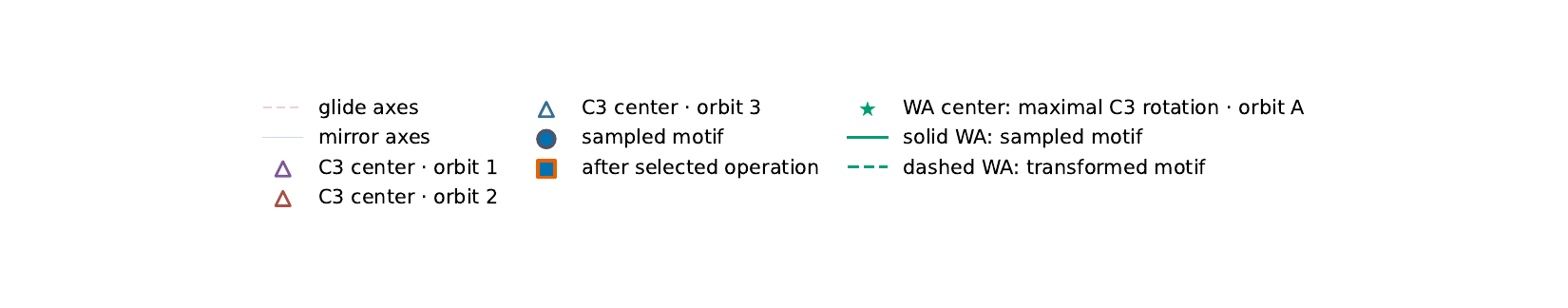}} \\[1mm]
\endfirsthead
\caption[]{Wallpaper group \texttt{p3m1} (continued)}\\[1mm]
\endhead
\includegraphics[width=\linewidth,height=0.22\textheight,keepaspectratio]{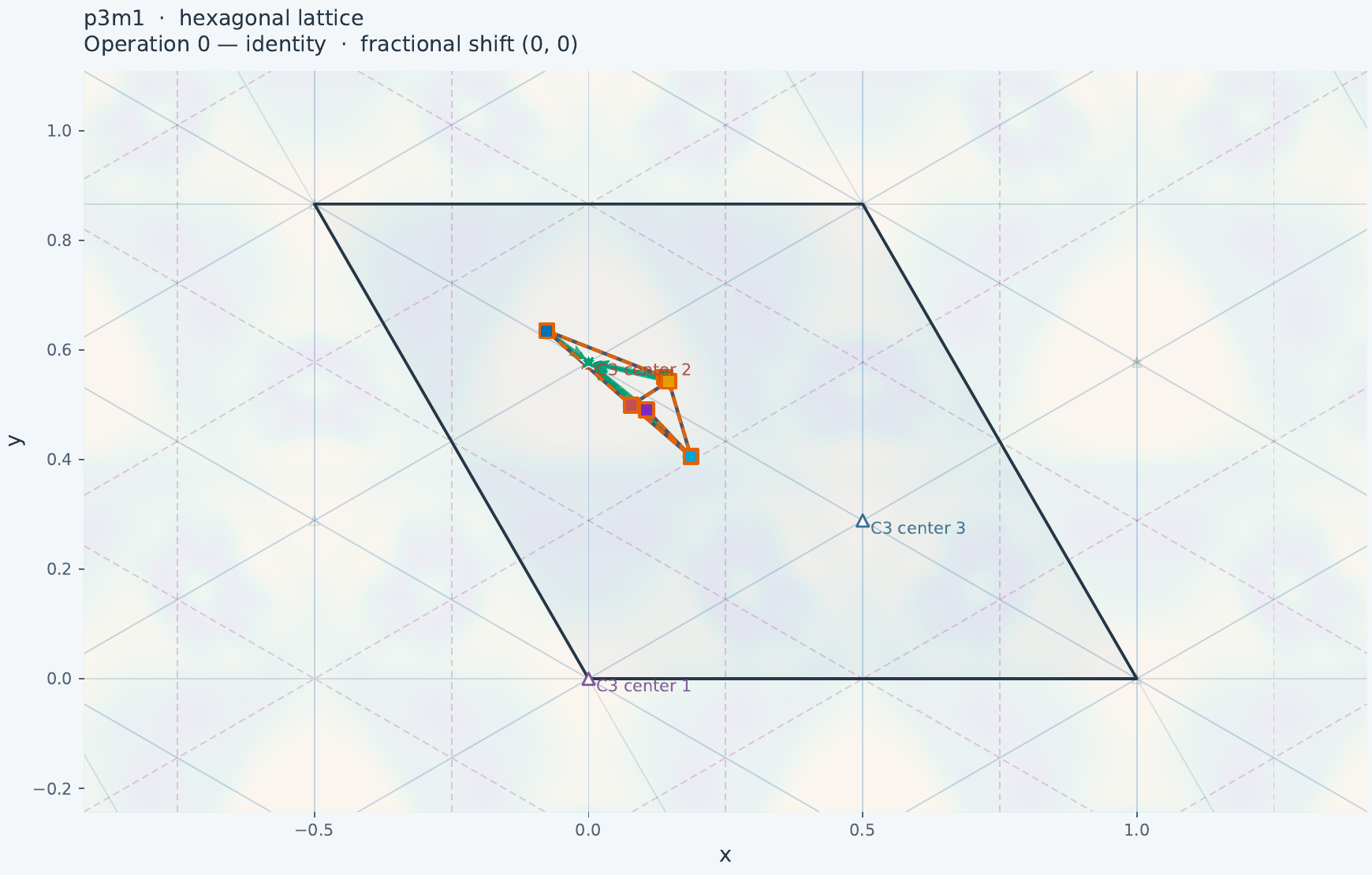} & \includegraphics[width=\linewidth,height=0.22\textheight,keepaspectratio]{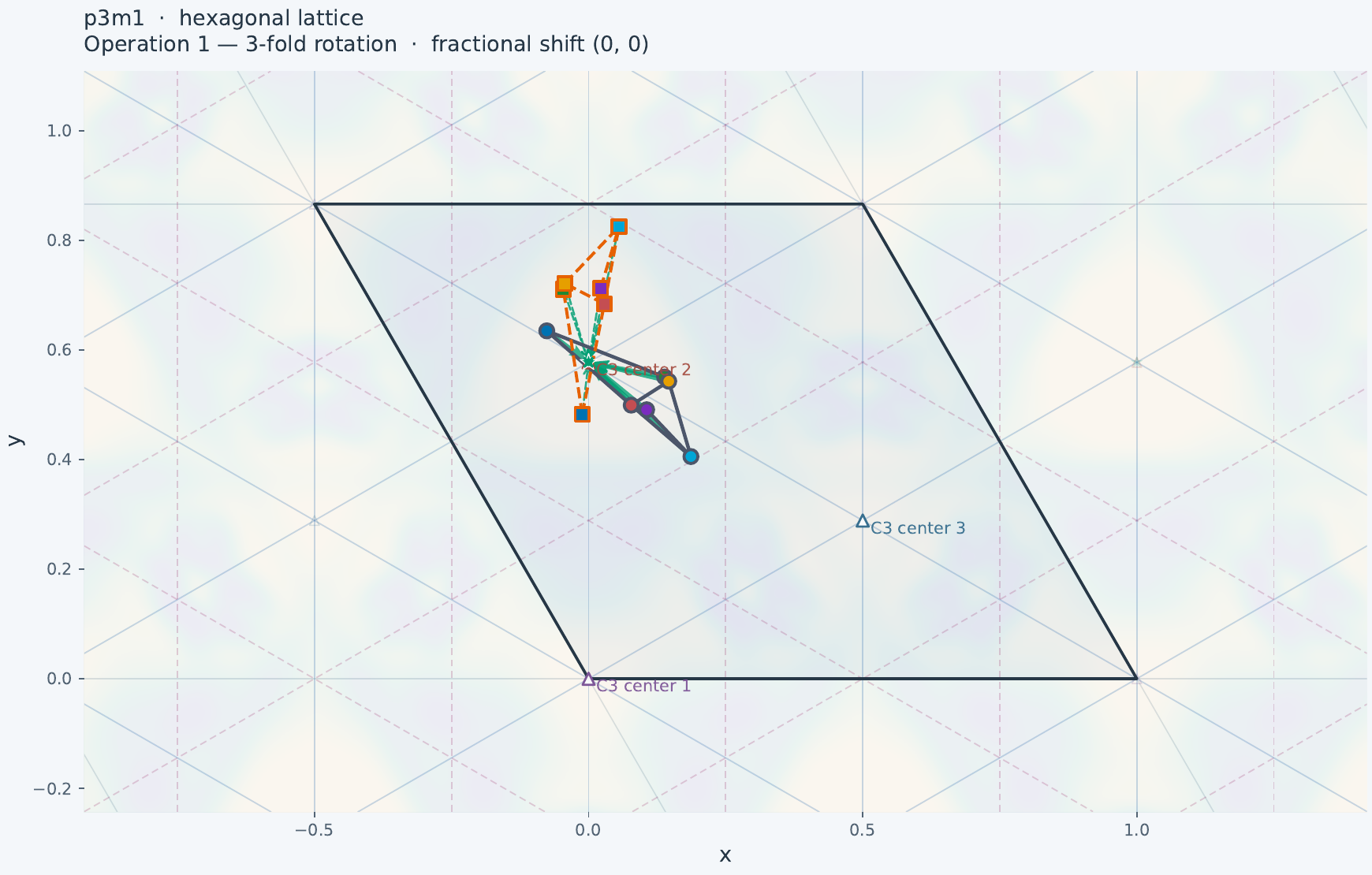} \\[2mm]
\includegraphics[width=\linewidth,height=0.22\textheight,keepaspectratio]{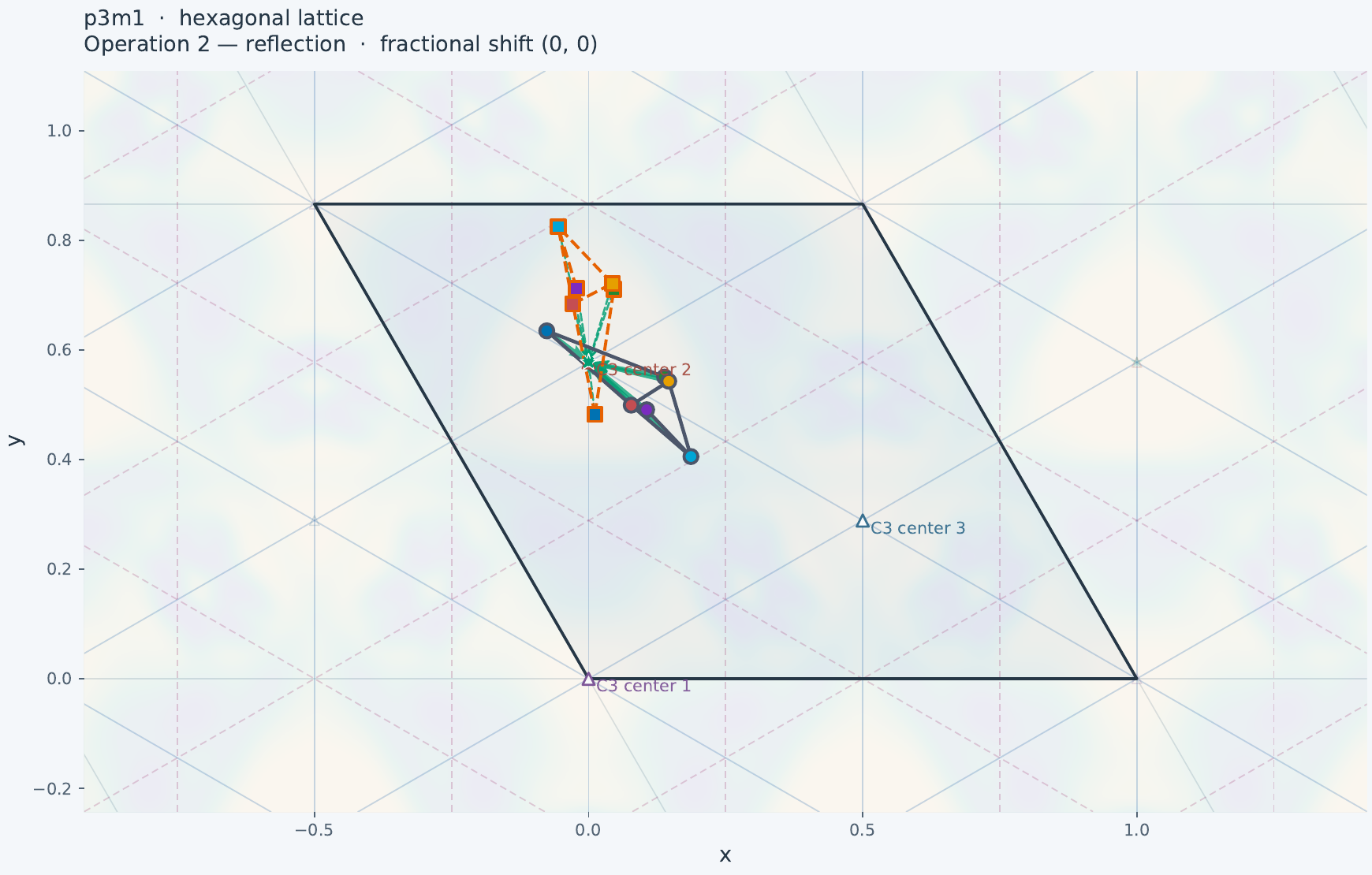} & \includegraphics[width=\linewidth,height=0.22\textheight,keepaspectratio]{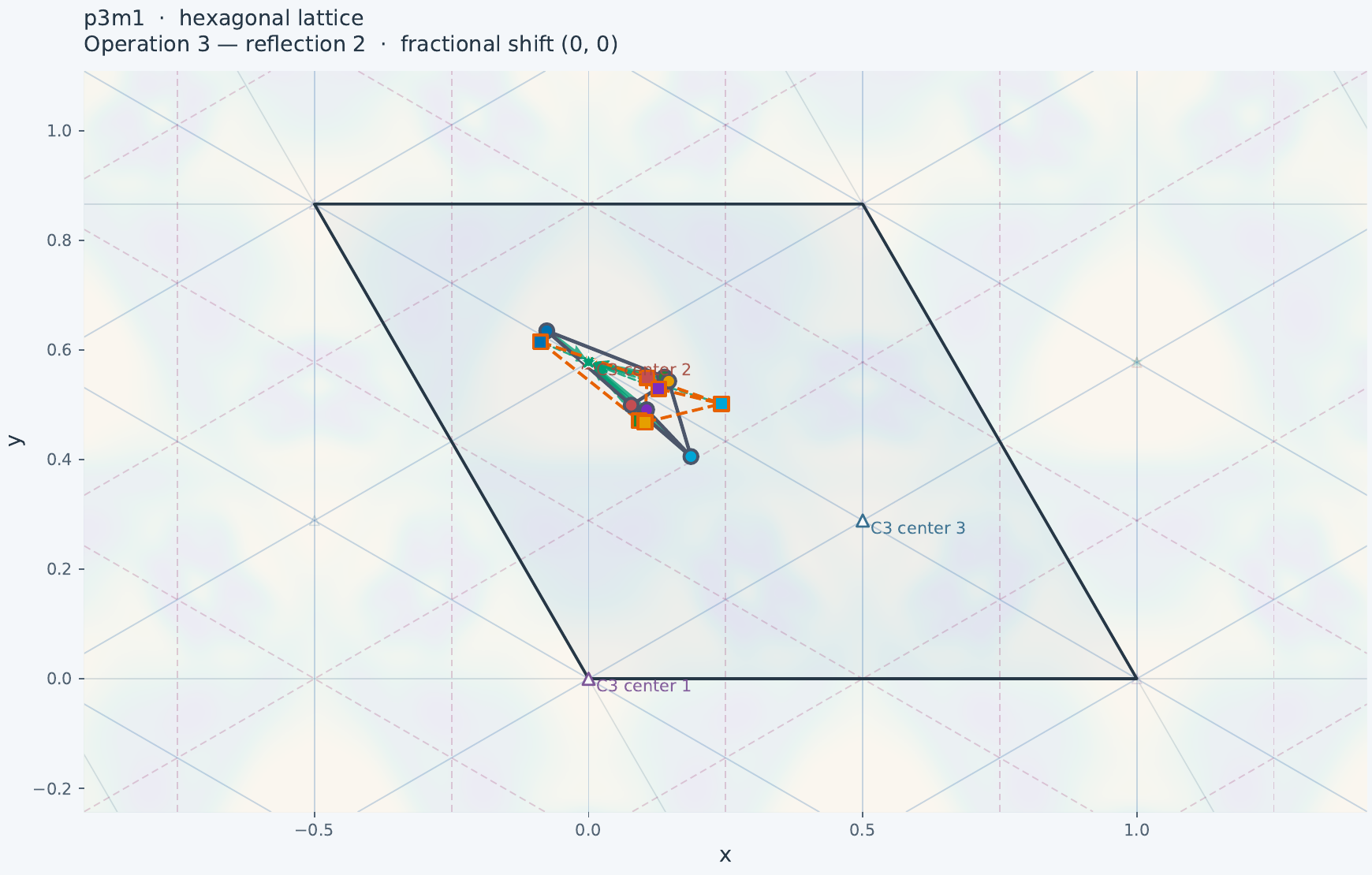} \\[2mm]
\includegraphics[width=\linewidth,height=0.22\textheight,keepaspectratio]{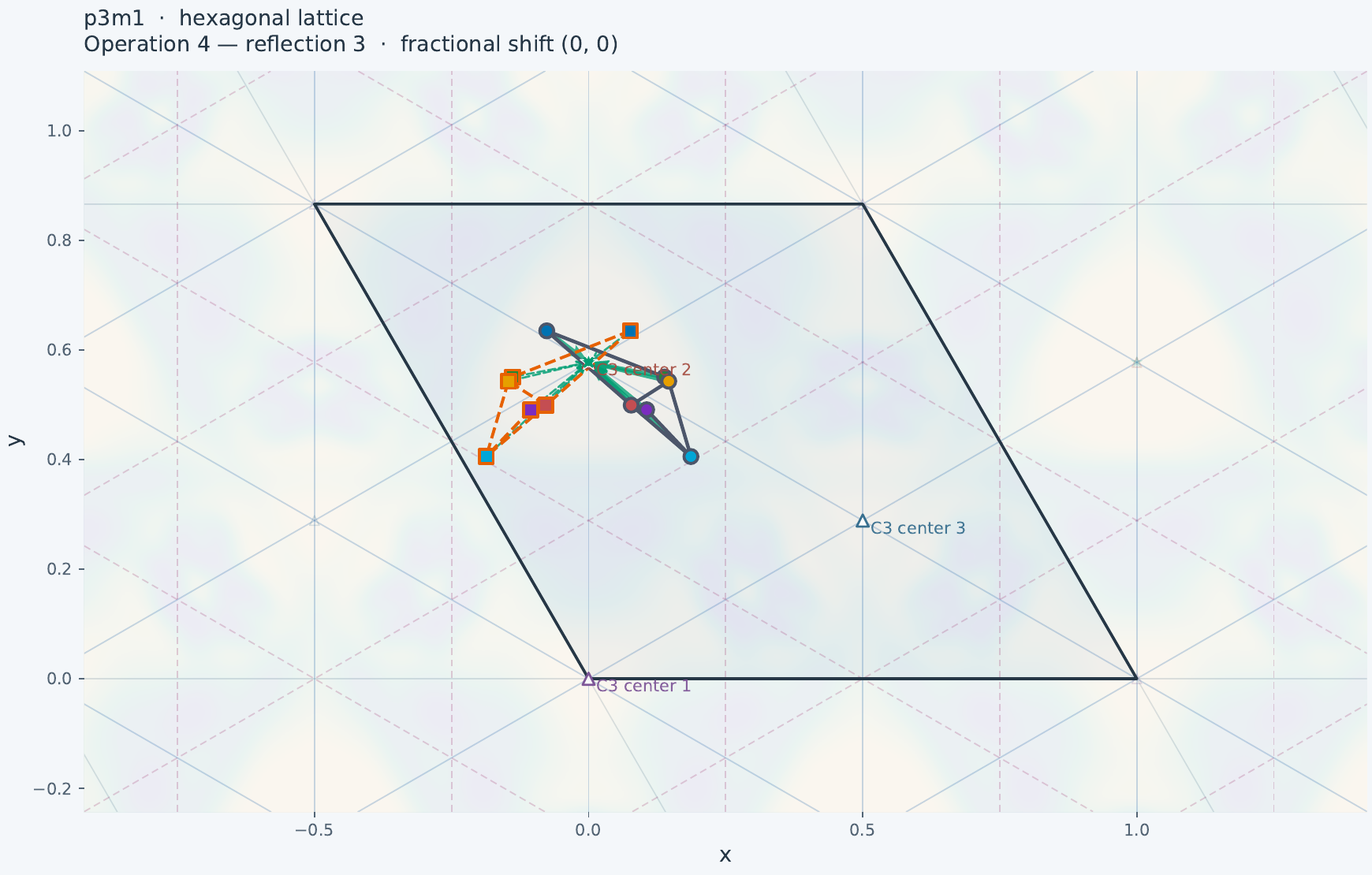} & \includegraphics[width=\linewidth,height=0.22\textheight,keepaspectratio]{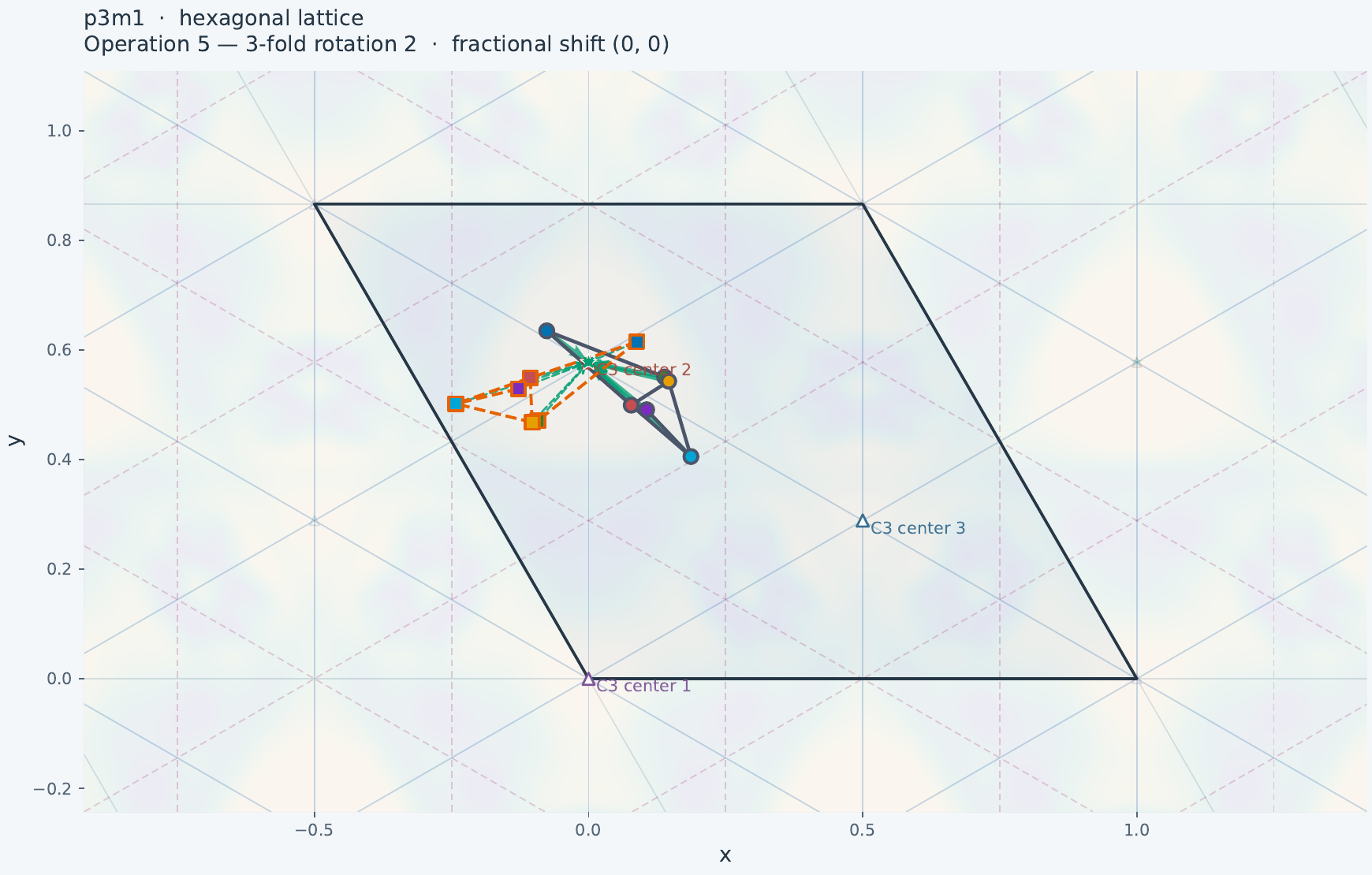} \\[2mm]
\includegraphics[width=\linewidth,height=0.22\textheight,keepaspectratio]{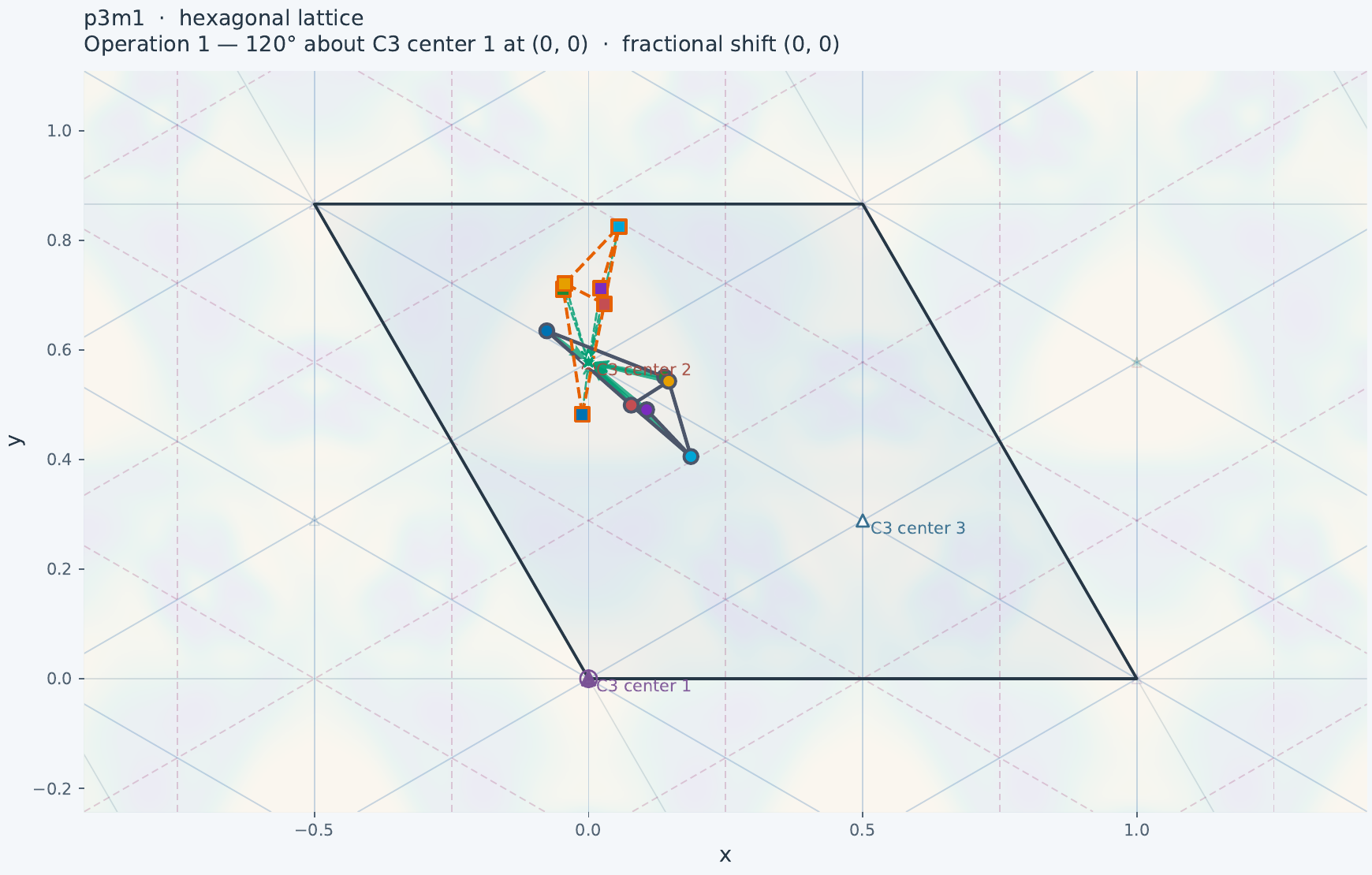} & \includegraphics[width=\linewidth,height=0.22\textheight,keepaspectratio]{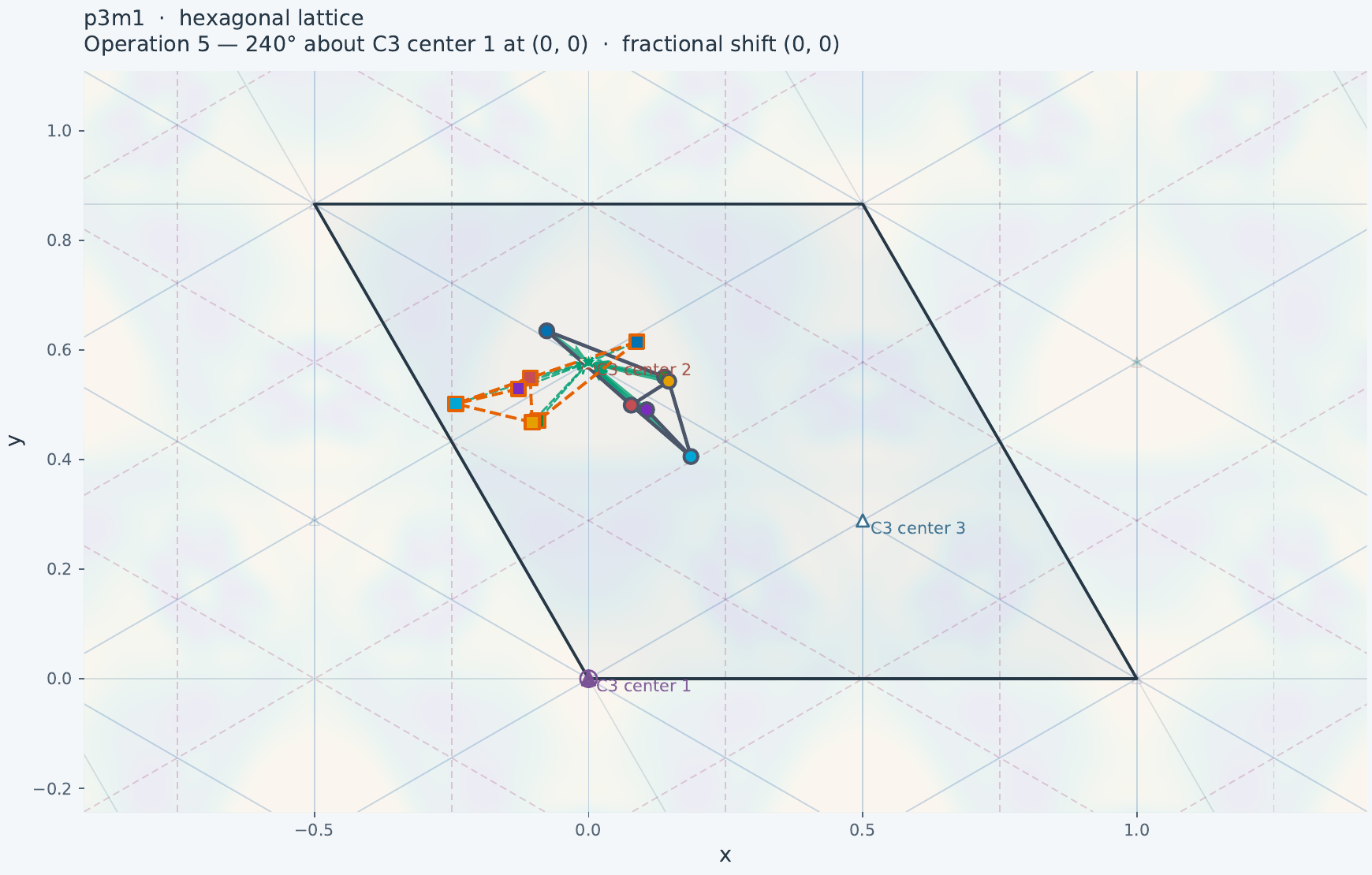} \\[2mm]
\includegraphics[width=\linewidth,height=0.22\textheight,keepaspectratio]{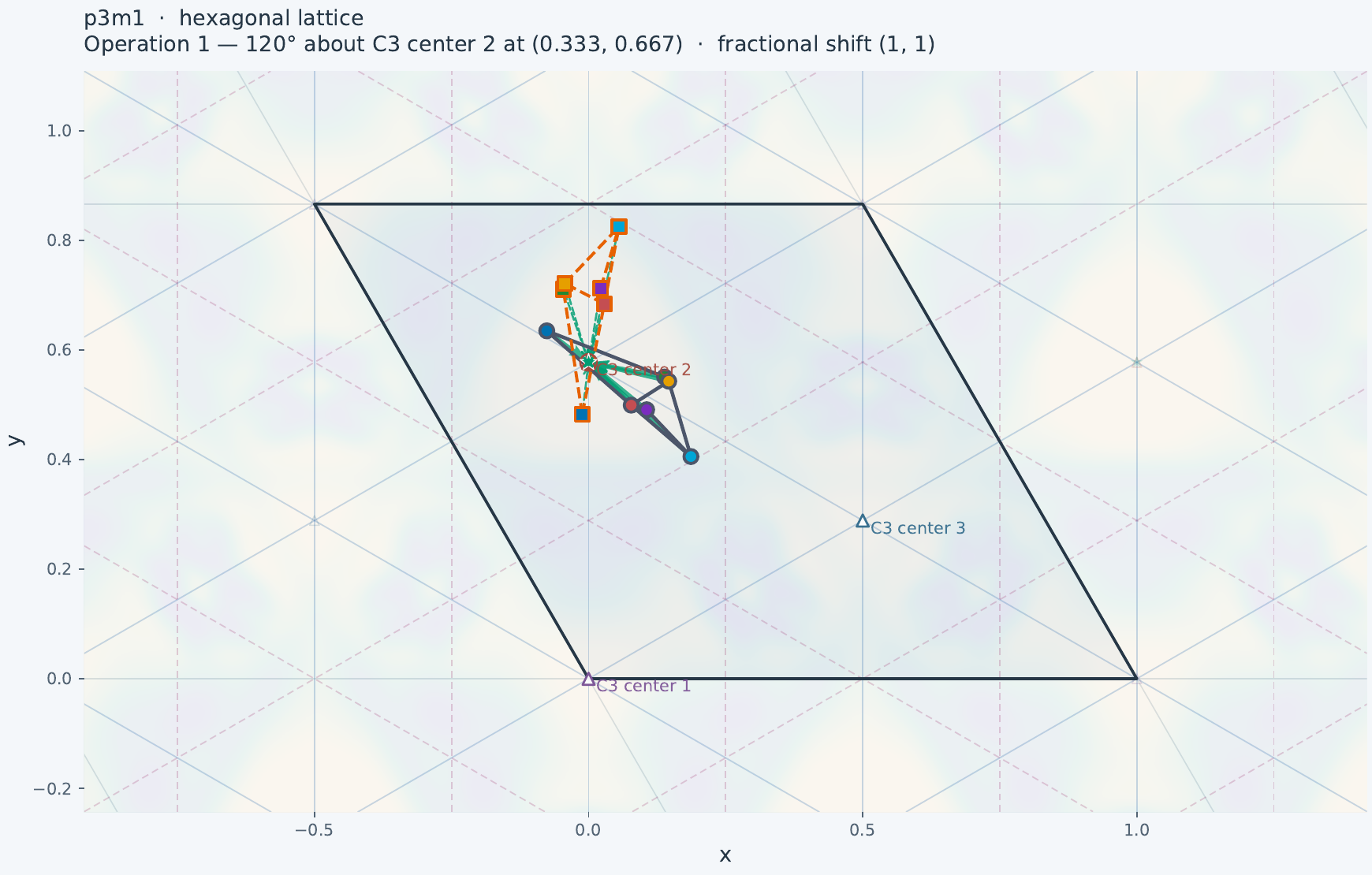} & \includegraphics[width=\linewidth,height=0.22\textheight,keepaspectratio]{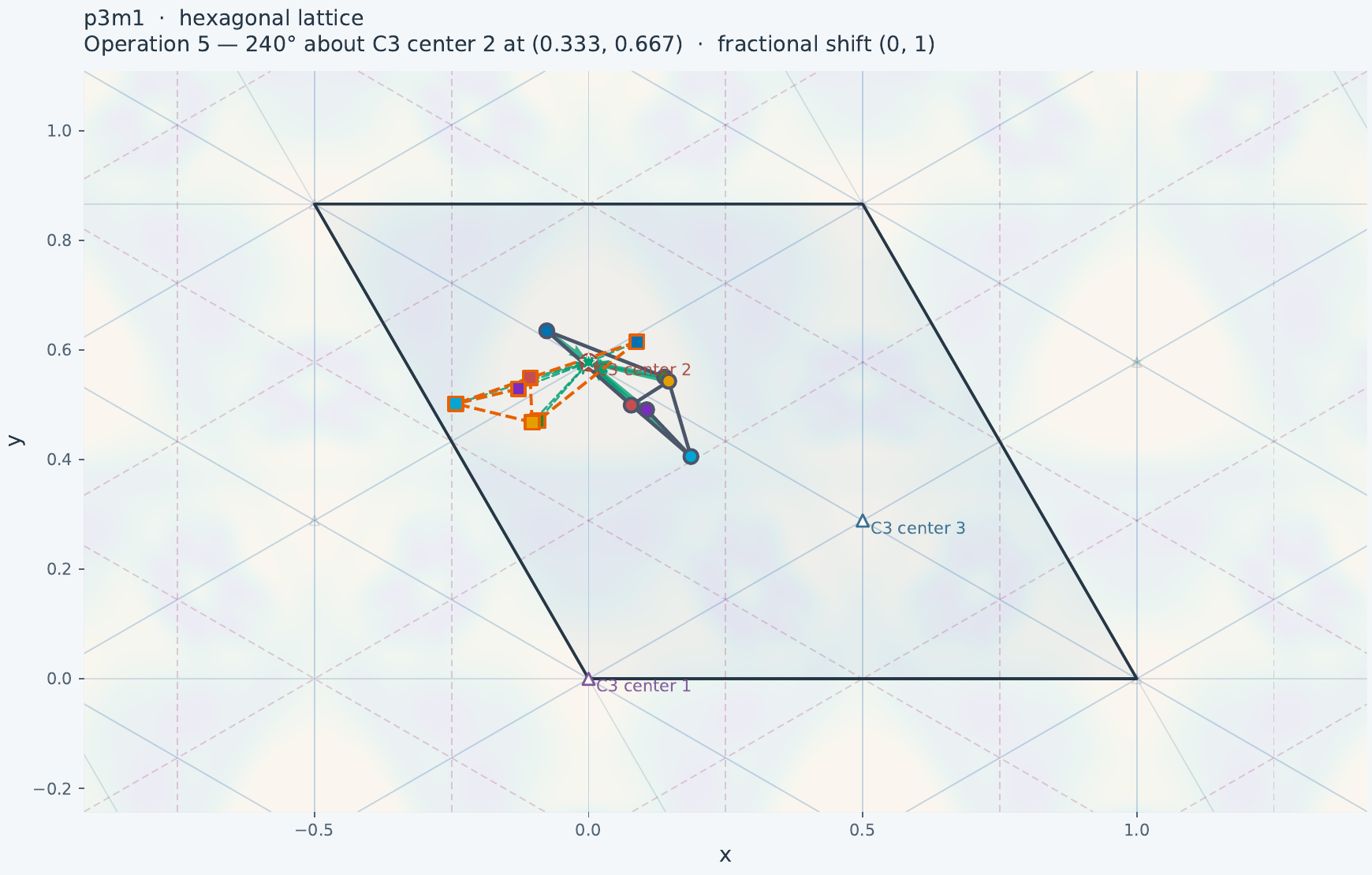} \\[2mm]
\includegraphics[width=\linewidth,height=0.22\textheight,keepaspectratio]{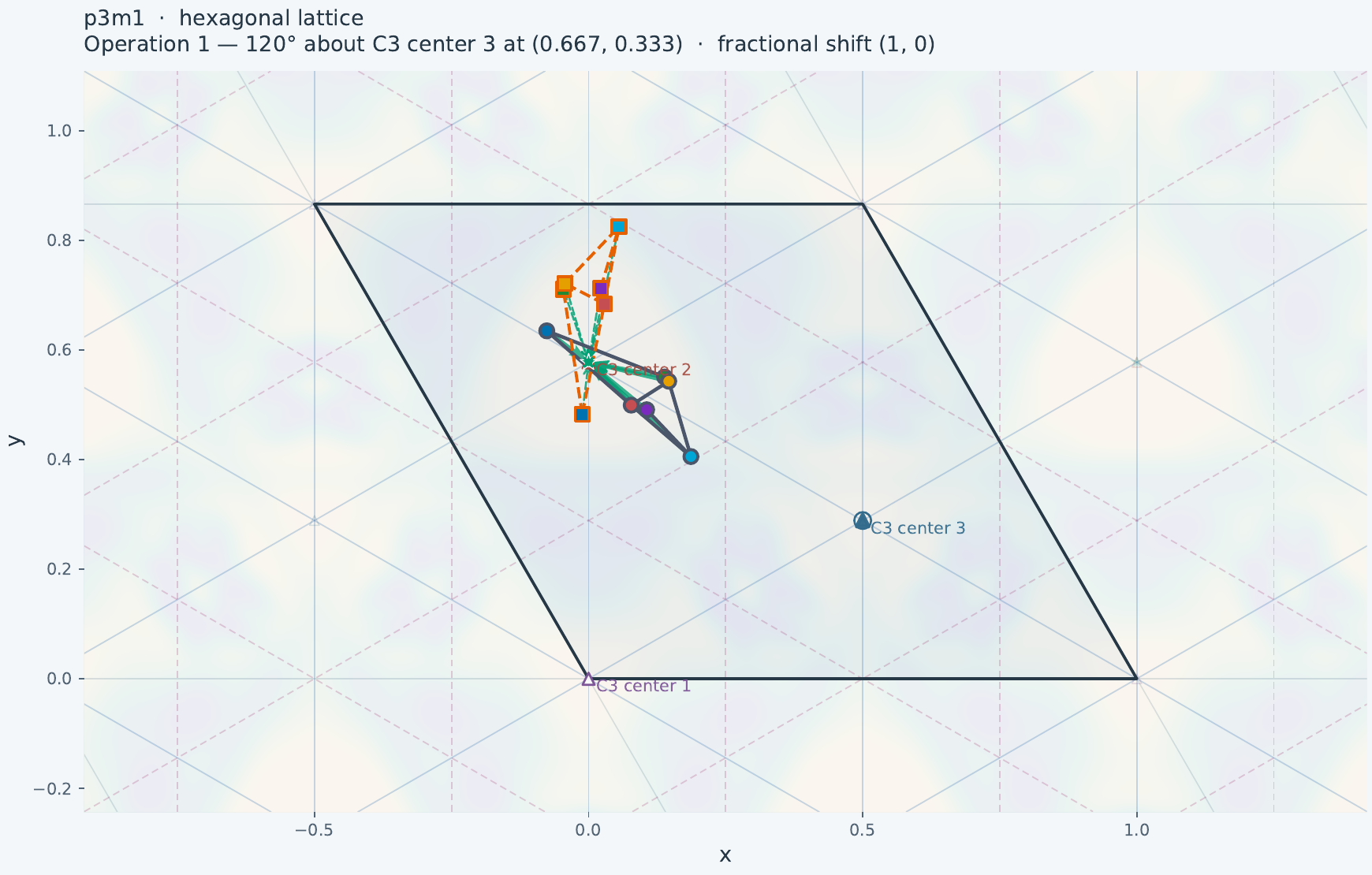} & \includegraphics[width=\linewidth,height=0.22\textheight,keepaspectratio]{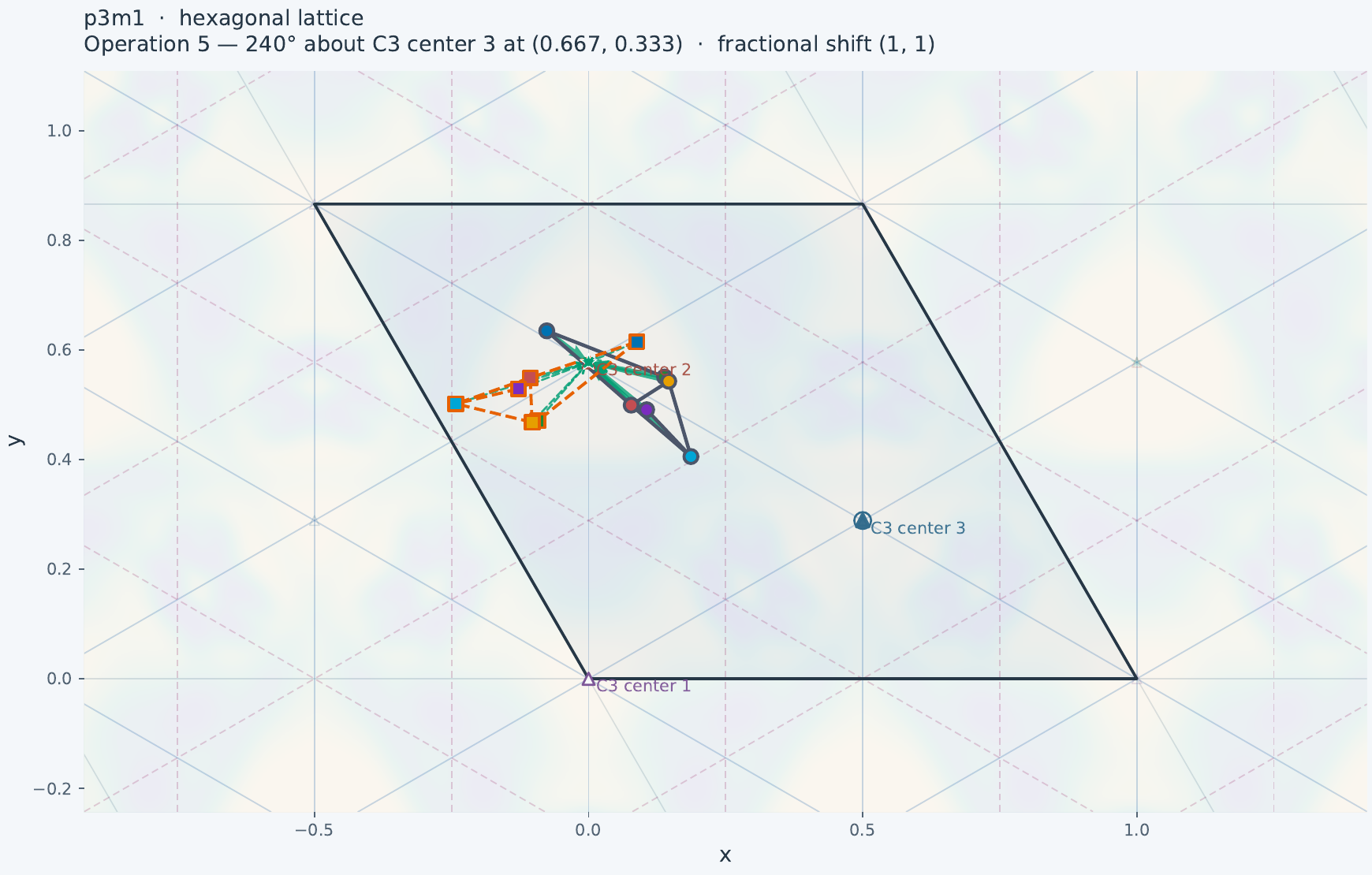} \\
\end{longtable}
\Needspace{218pt}
\begin{longtable}{@{}>{\centering\arraybackslash}p{0.475\linewidth}>{\centering\arraybackslash}p{0.475\linewidth}@{}}
\caption[Wallpaper group \texttt{p31m}]{Wallpaper group \texttt{p31m}, {\small WA is pointing to C3-B at $q_{1}=(0,0)$}}\label{tab:wa-actions-p31m}\\[1mm]
\multicolumn{2}{c}{\includegraphics[width=0.96\linewidth]{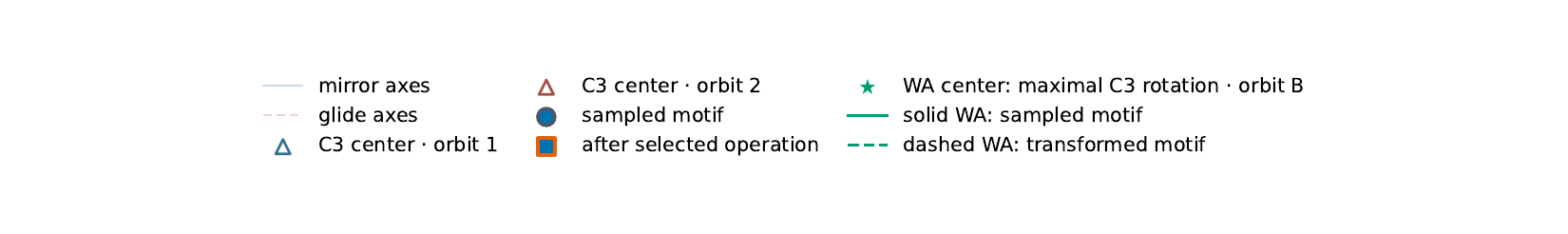}} \\[1mm]
\endfirsthead
\caption[]{Wallpaper group \texttt{p31m} (continued)}\\[1mm]
\endhead
\includegraphics[width=\linewidth,height=0.22\textheight,keepaspectratio]{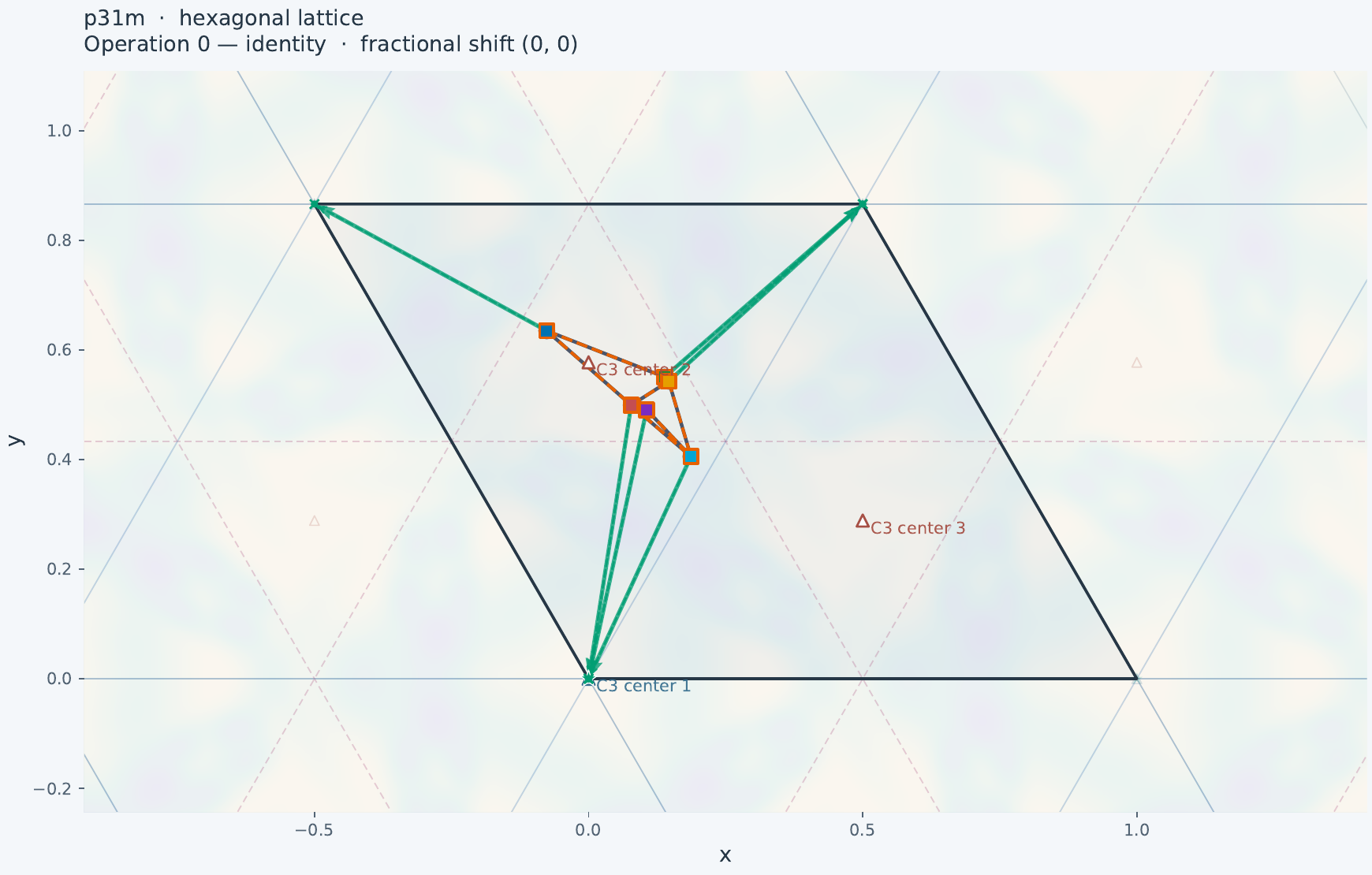} & \includegraphics[width=\linewidth,height=0.22\textheight,keepaspectratio]{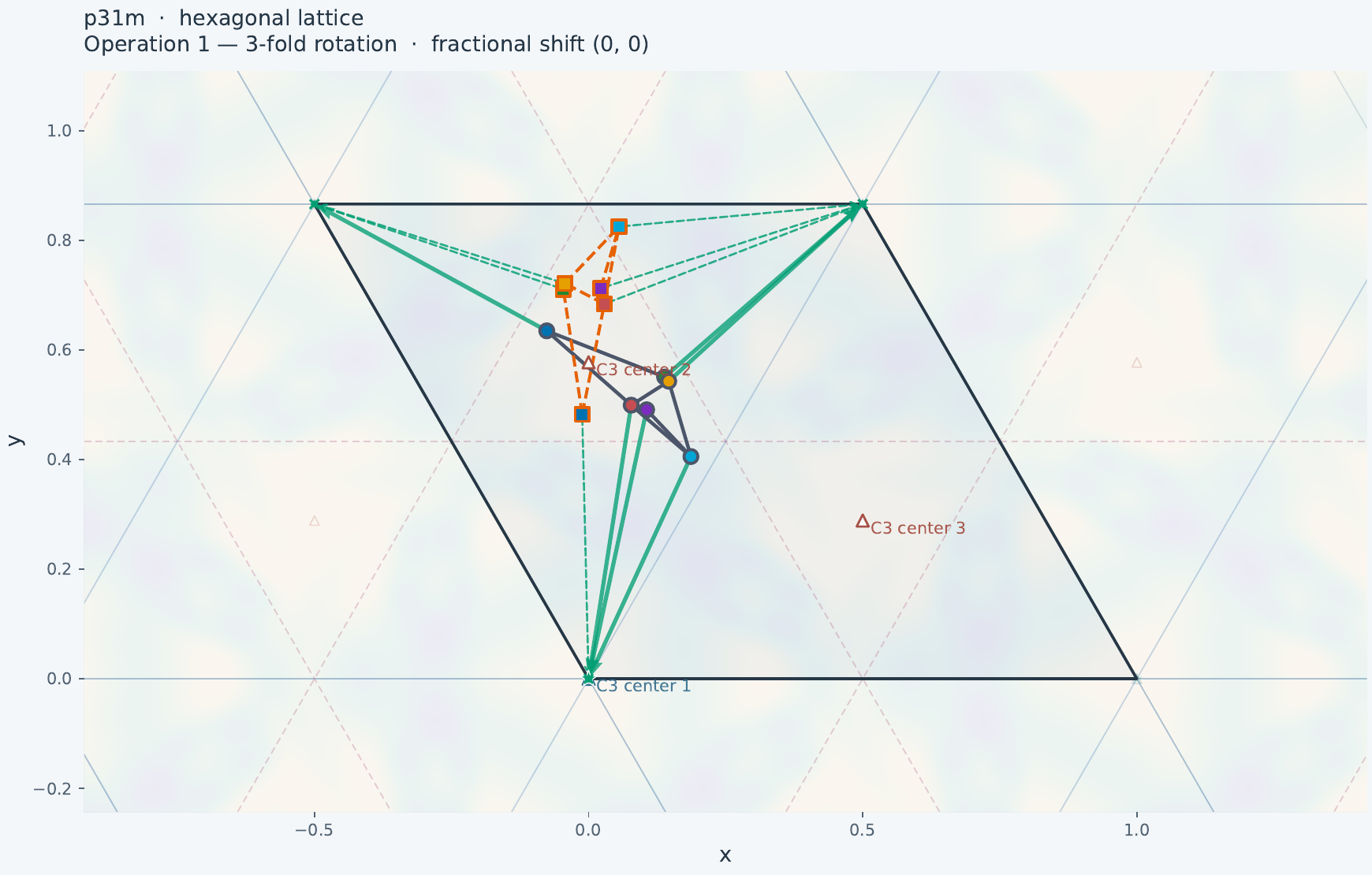} \\[2mm]
\includegraphics[width=\linewidth,height=0.22\textheight,keepaspectratio]{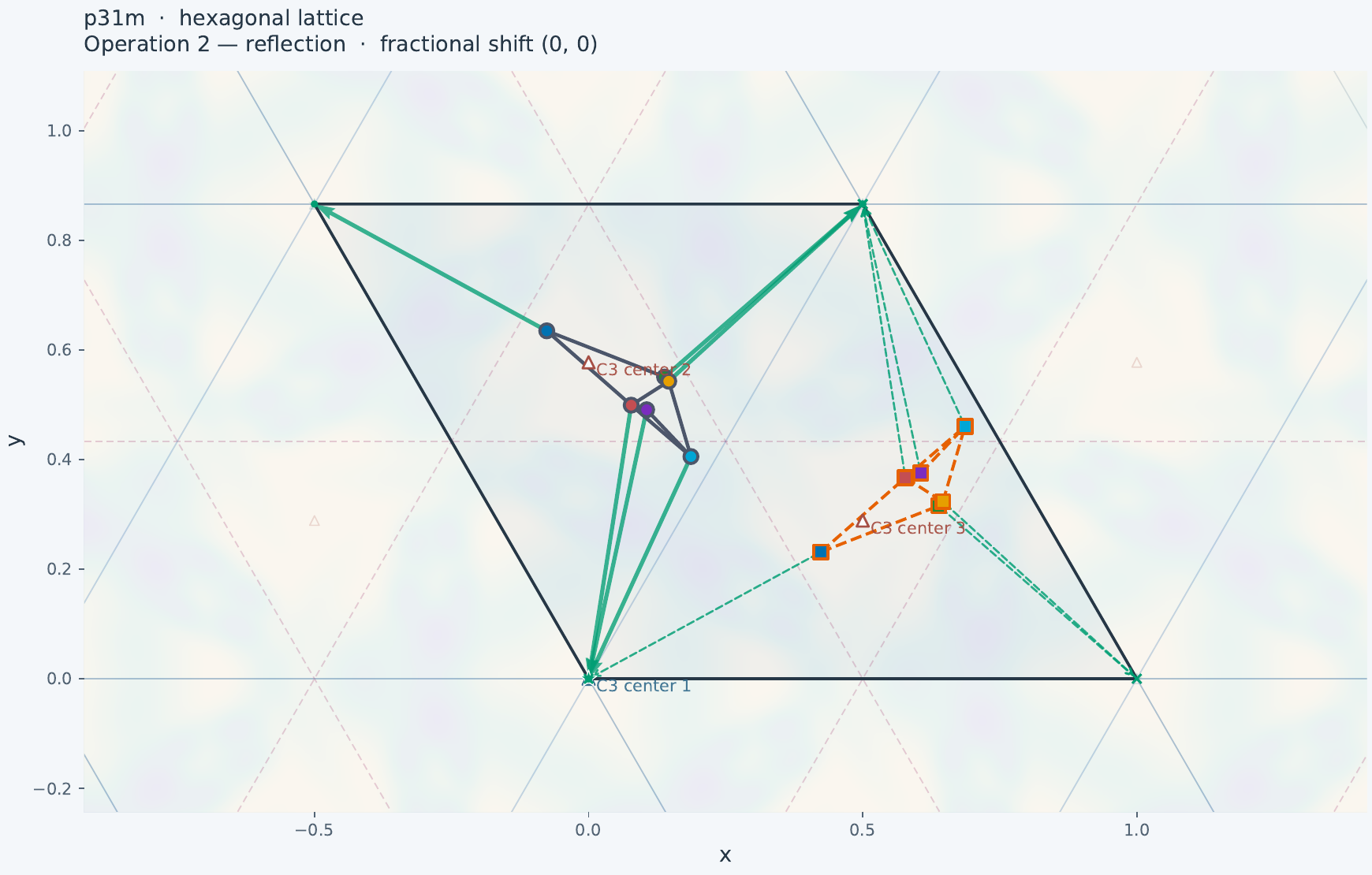} & \includegraphics[width=\linewidth,height=0.22\textheight,keepaspectratio]{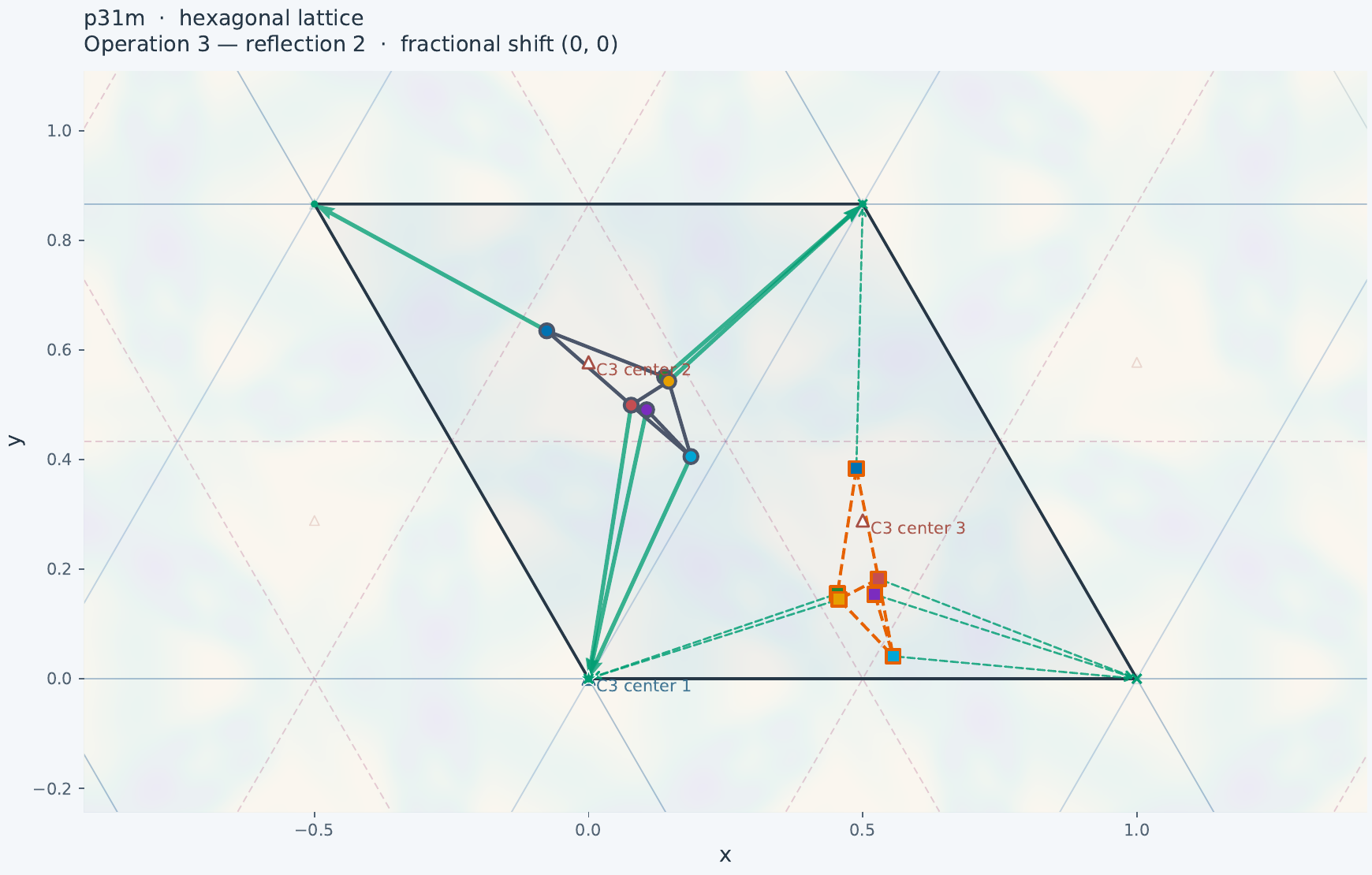} \\[2mm]
\includegraphics[width=\linewidth,height=0.22\textheight,keepaspectratio]{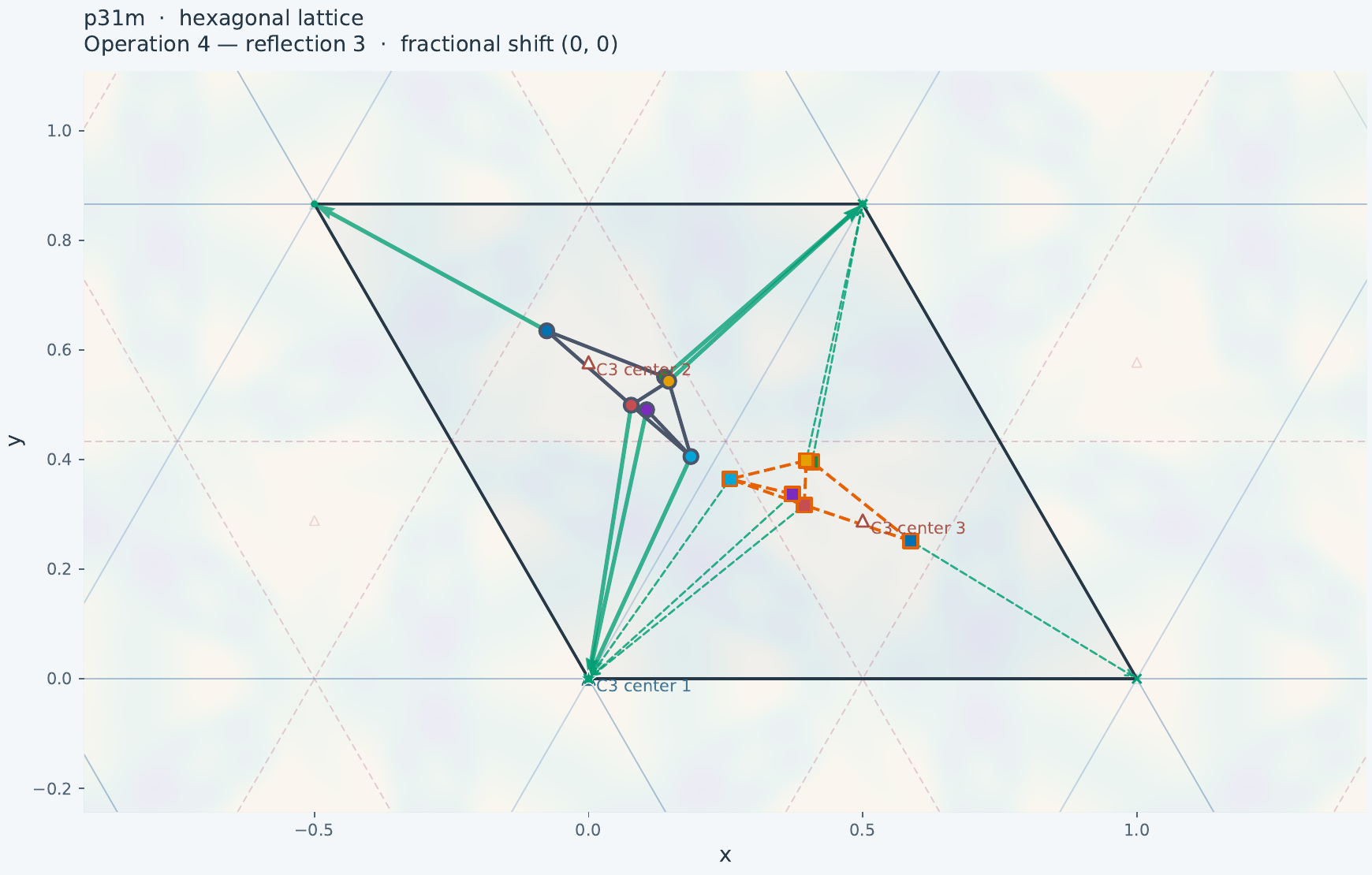} & \includegraphics[width=\linewidth,height=0.22\textheight,keepaspectratio]{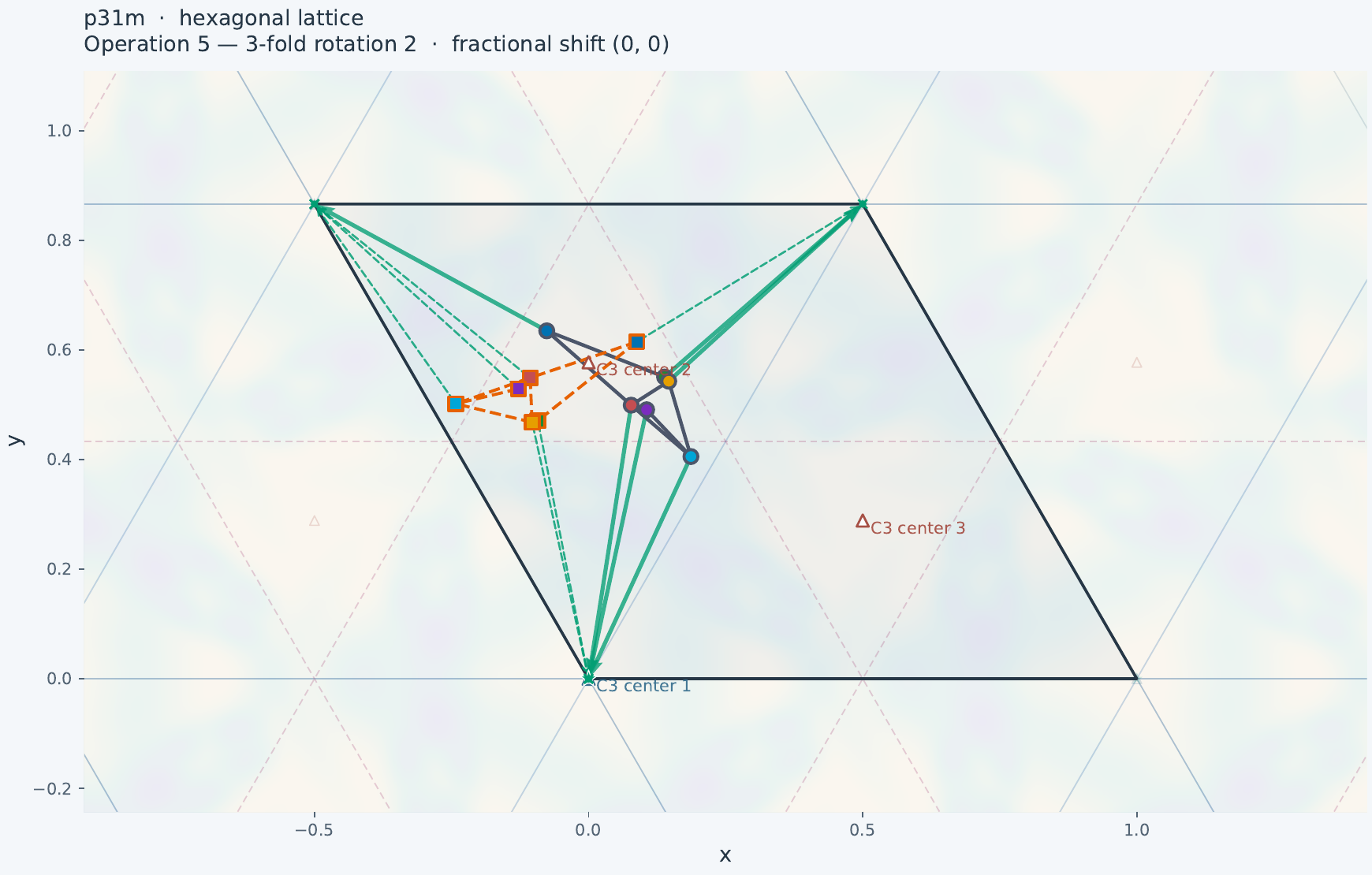} \\[2mm]
\includegraphics[width=\linewidth,height=0.22\textheight,keepaspectratio]{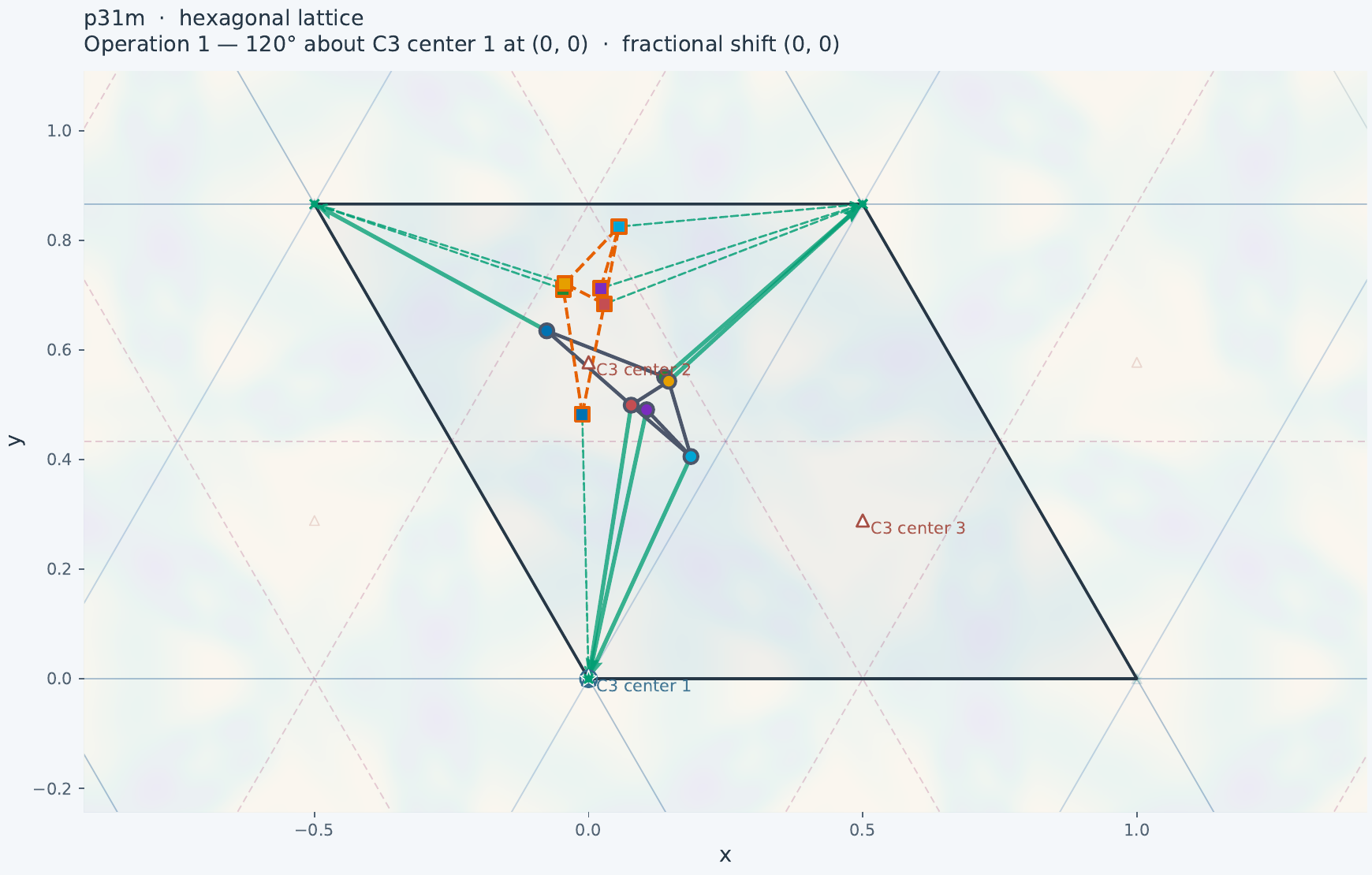} & \includegraphics[width=\linewidth,height=0.22\textheight,keepaspectratio]{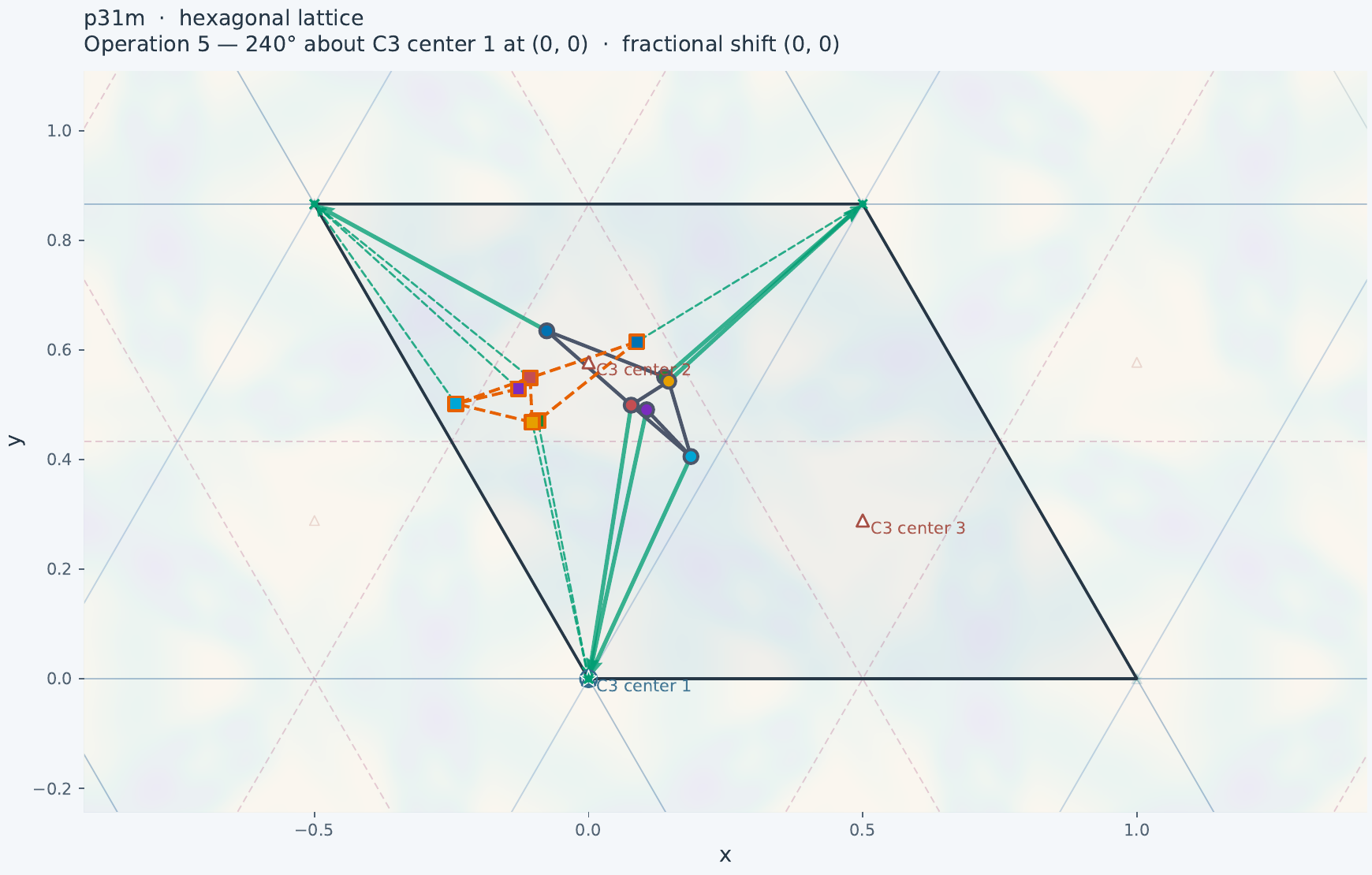} \\[2mm]
\includegraphics[width=\linewidth,height=0.22\textheight,keepaspectratio]{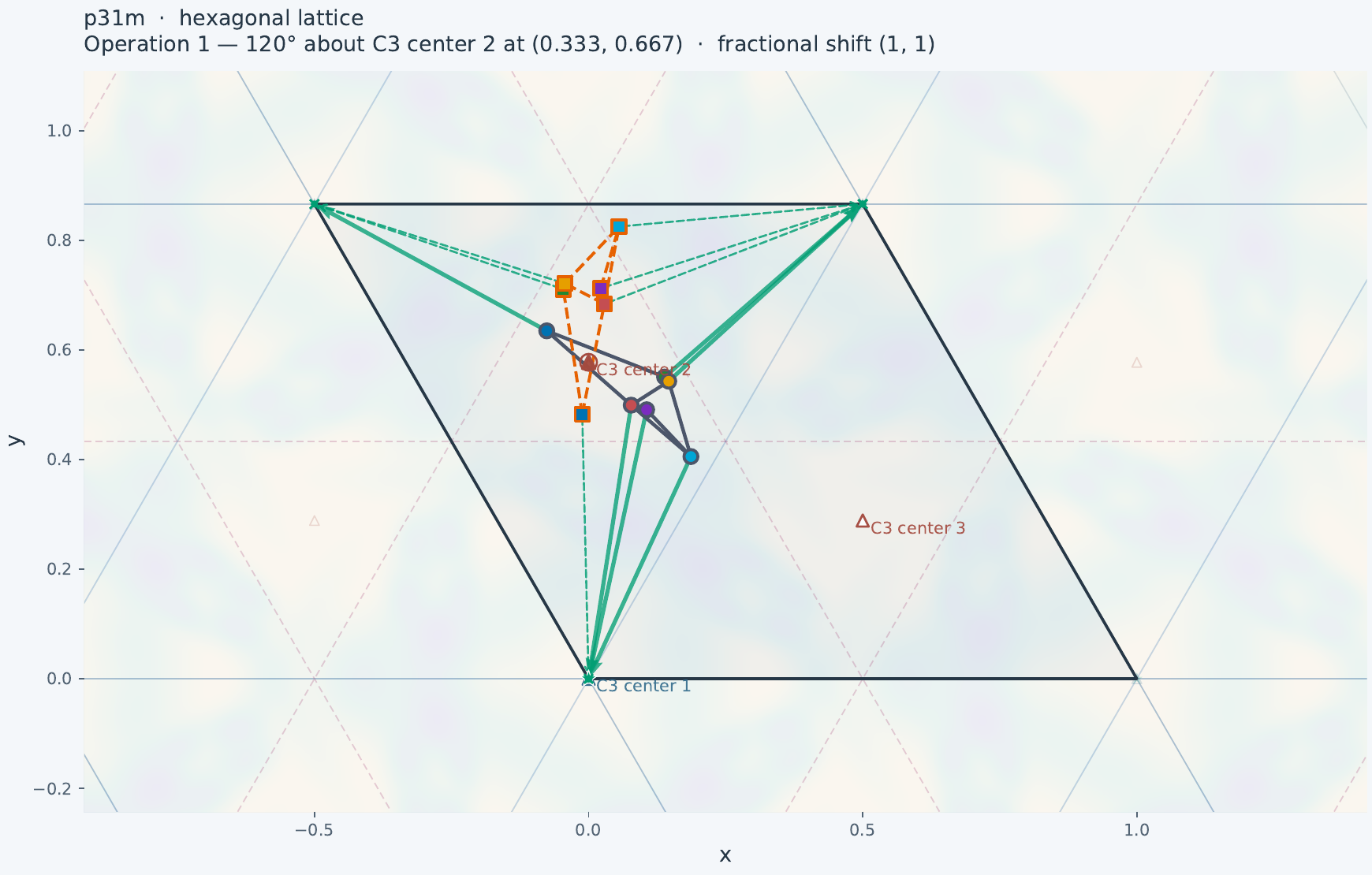} & \includegraphics[width=\linewidth,height=0.22\textheight,keepaspectratio]{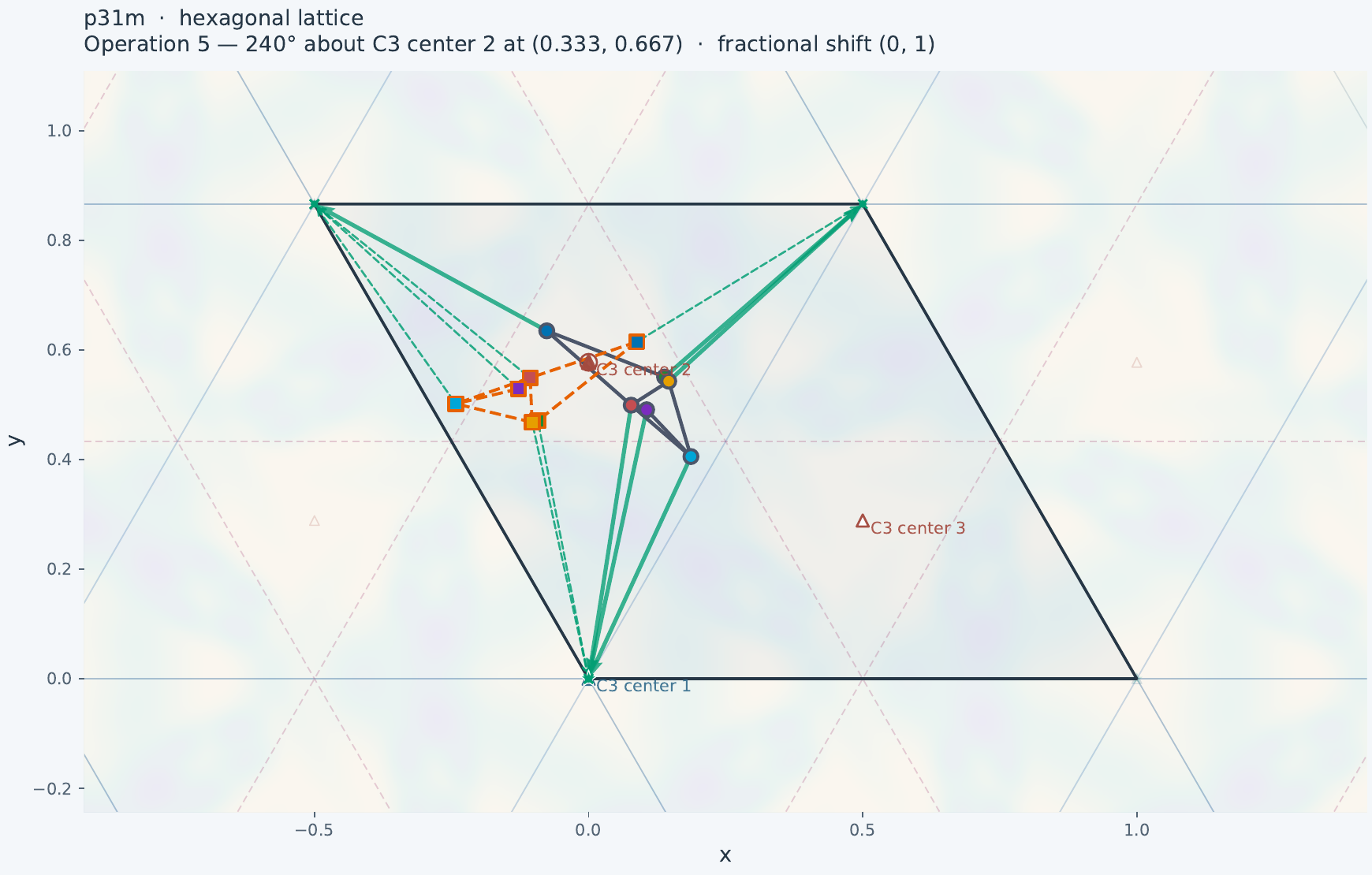} \\[2mm]
\includegraphics[width=\linewidth,height=0.22\textheight,keepaspectratio]{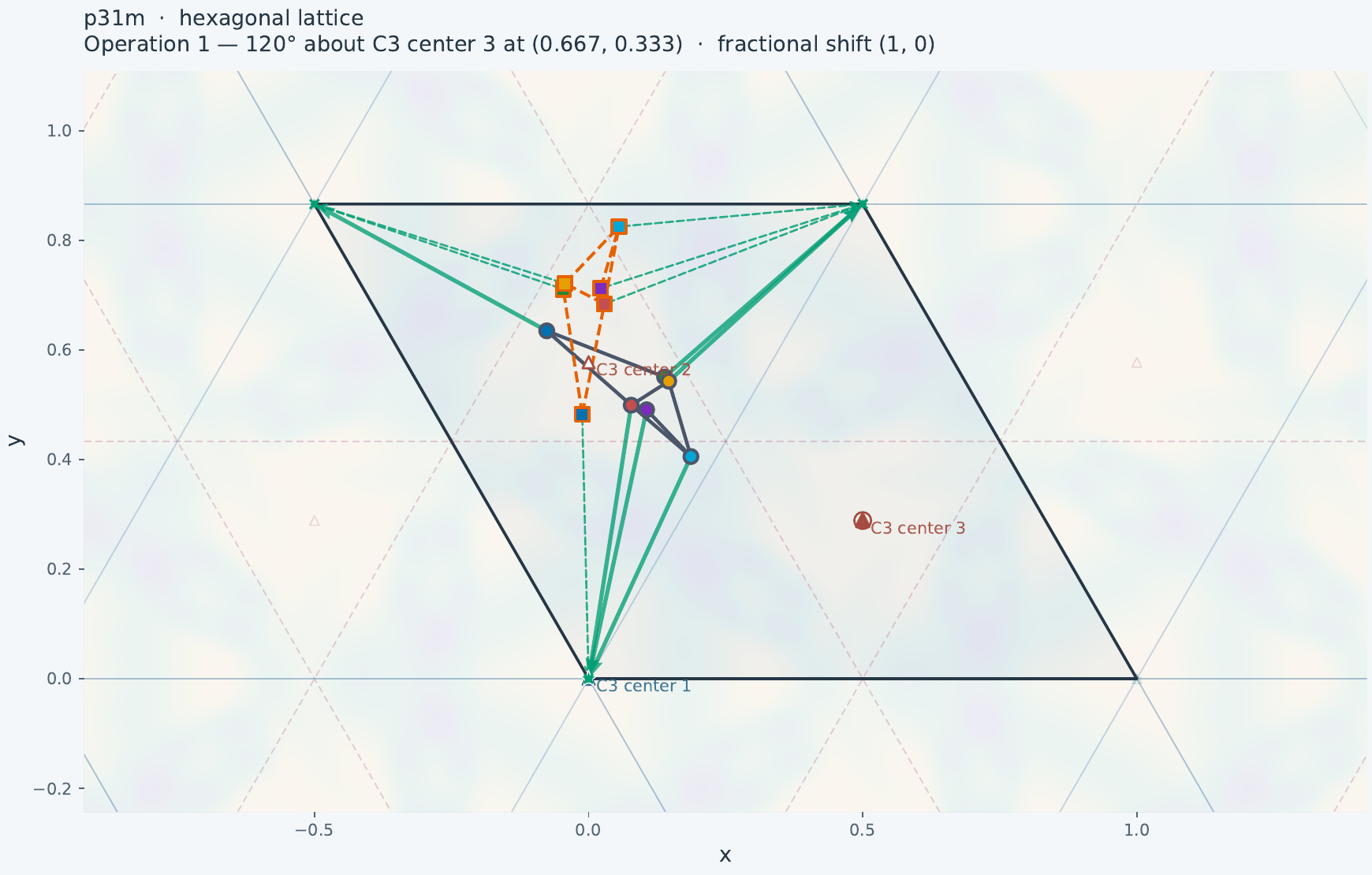} & \includegraphics[width=\linewidth,height=0.22\textheight,keepaspectratio]{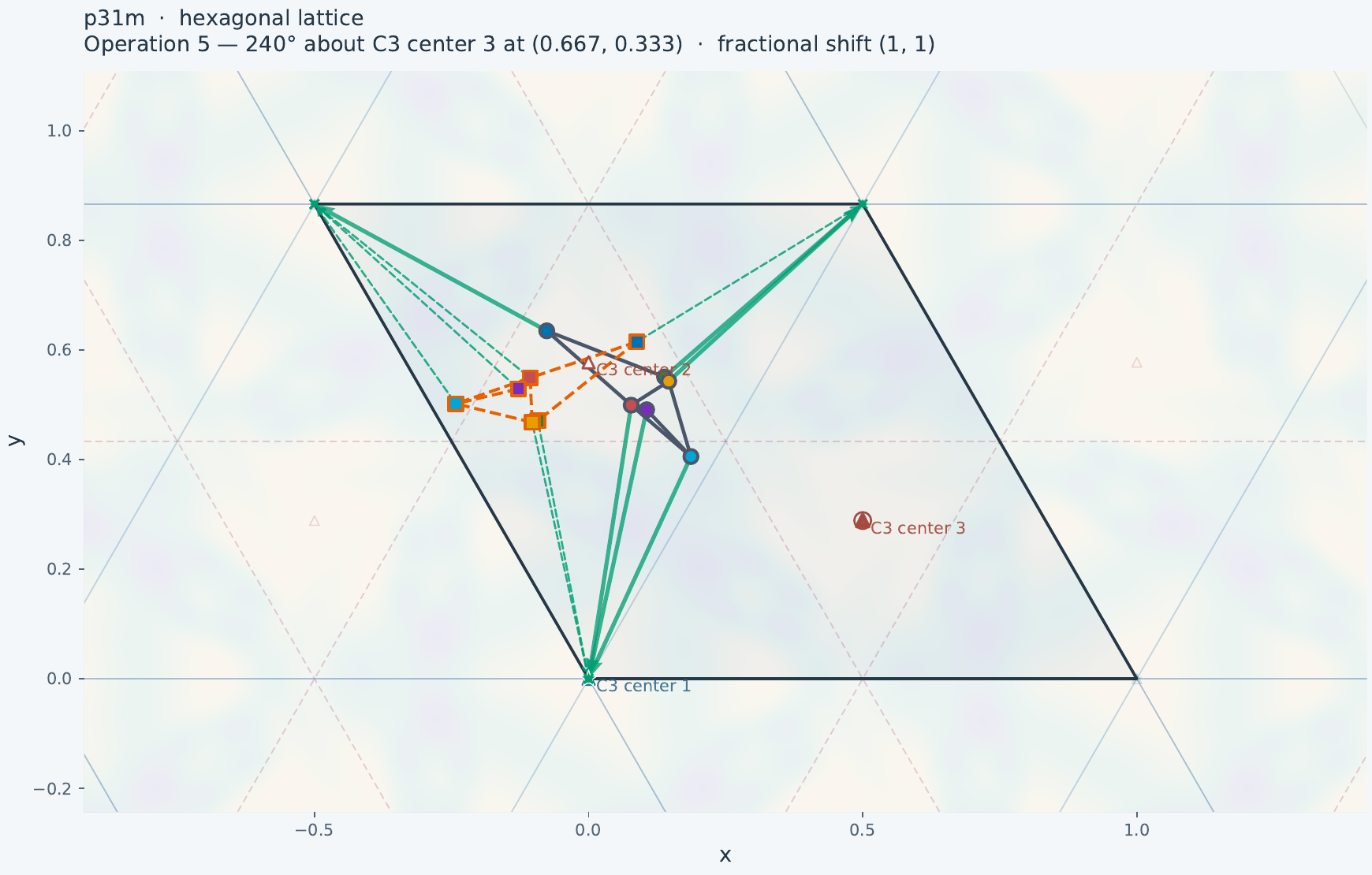} \\
\end{longtable}
\Needspace{218pt}
\begin{longtable}{@{}>{\centering\arraybackslash}p{0.475\linewidth}>{\centering\arraybackslash}p{0.475\linewidth}@{}}
\caption[Wallpaper group \texttt{p6}]{Wallpaper group \texttt{p6}, {\small WA is pointing to C6 at $q_{1}=(0,0)$}}\label{tab:wa-actions-p6}\\[1mm]
\multicolumn{2}{c}{\includegraphics[width=0.96\linewidth]{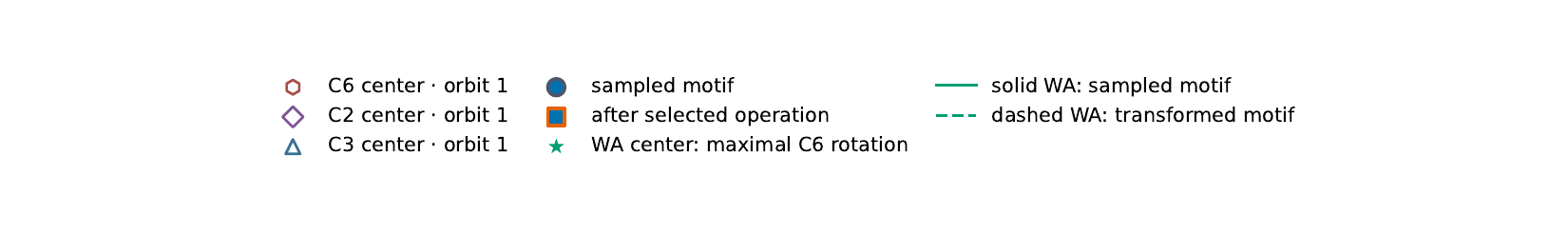}} \\[1mm]
\endfirsthead
\caption[]{Wallpaper group \texttt{p6} (continued)}\\[1mm]
\endhead
\includegraphics[width=\linewidth,height=0.22\textheight,keepaspectratio]{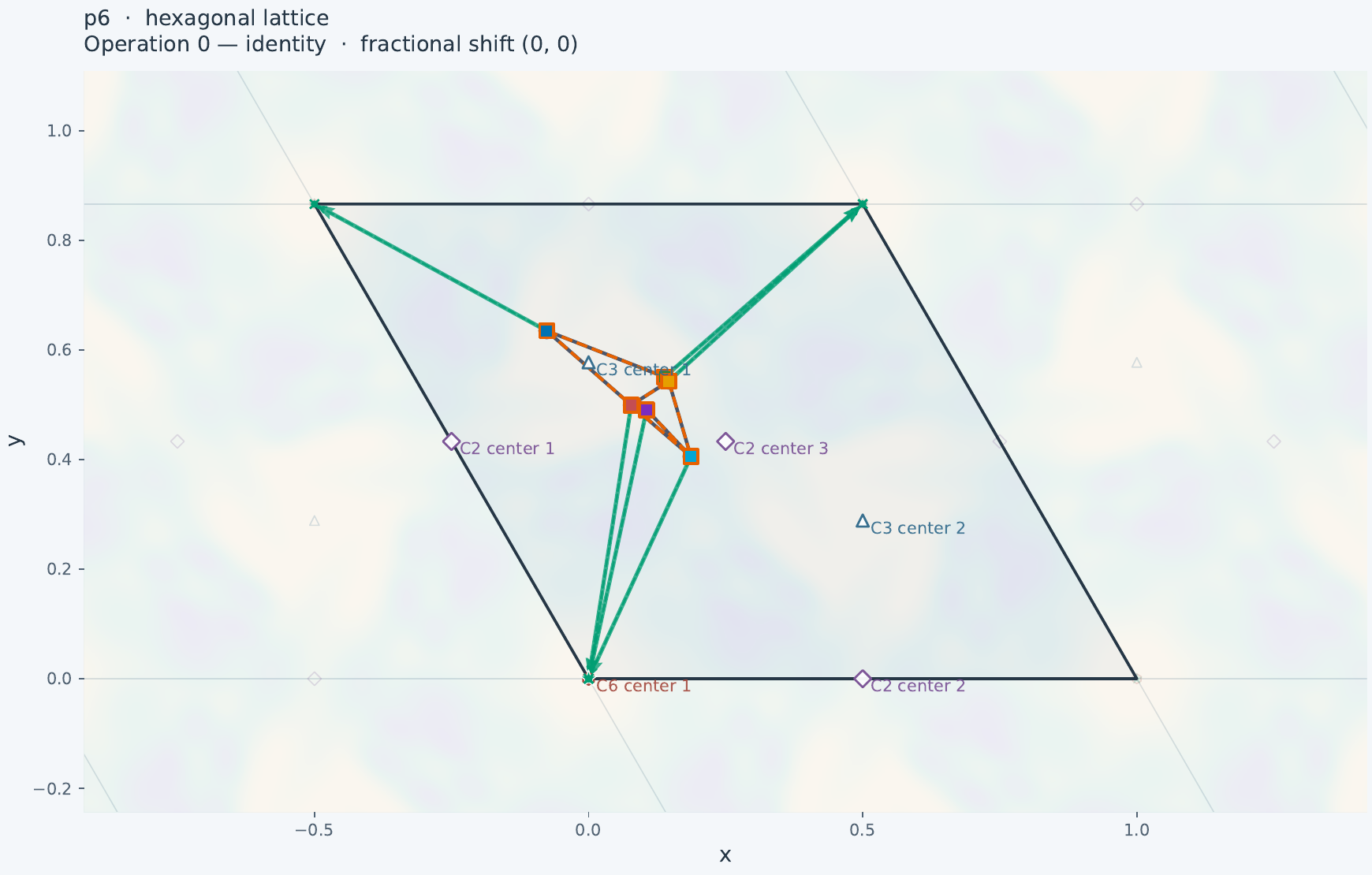} & \includegraphics[width=\linewidth,height=0.22\textheight,keepaspectratio]{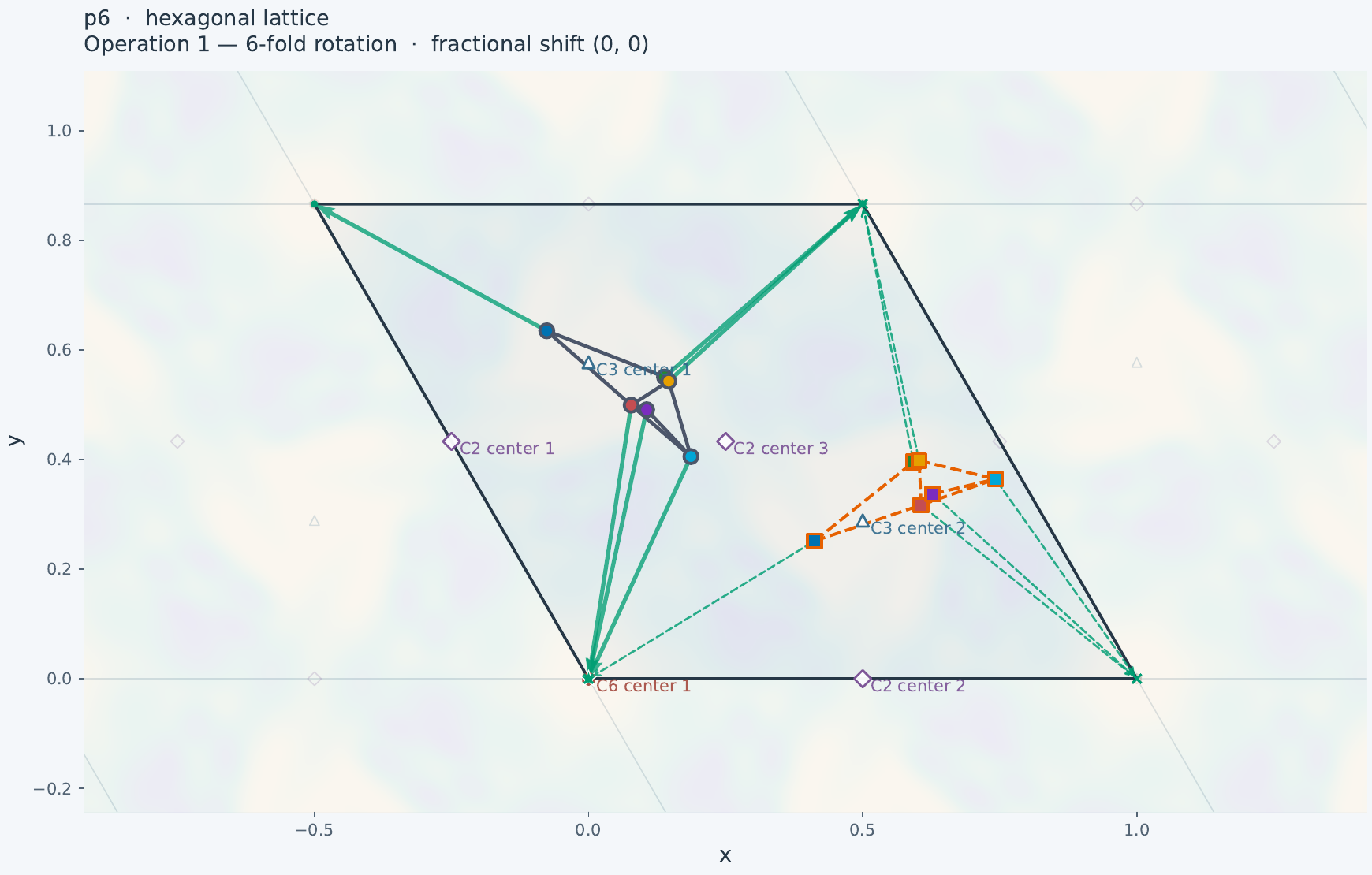} \\[2mm]
\includegraphics[width=\linewidth,height=0.22\textheight,keepaspectratio]{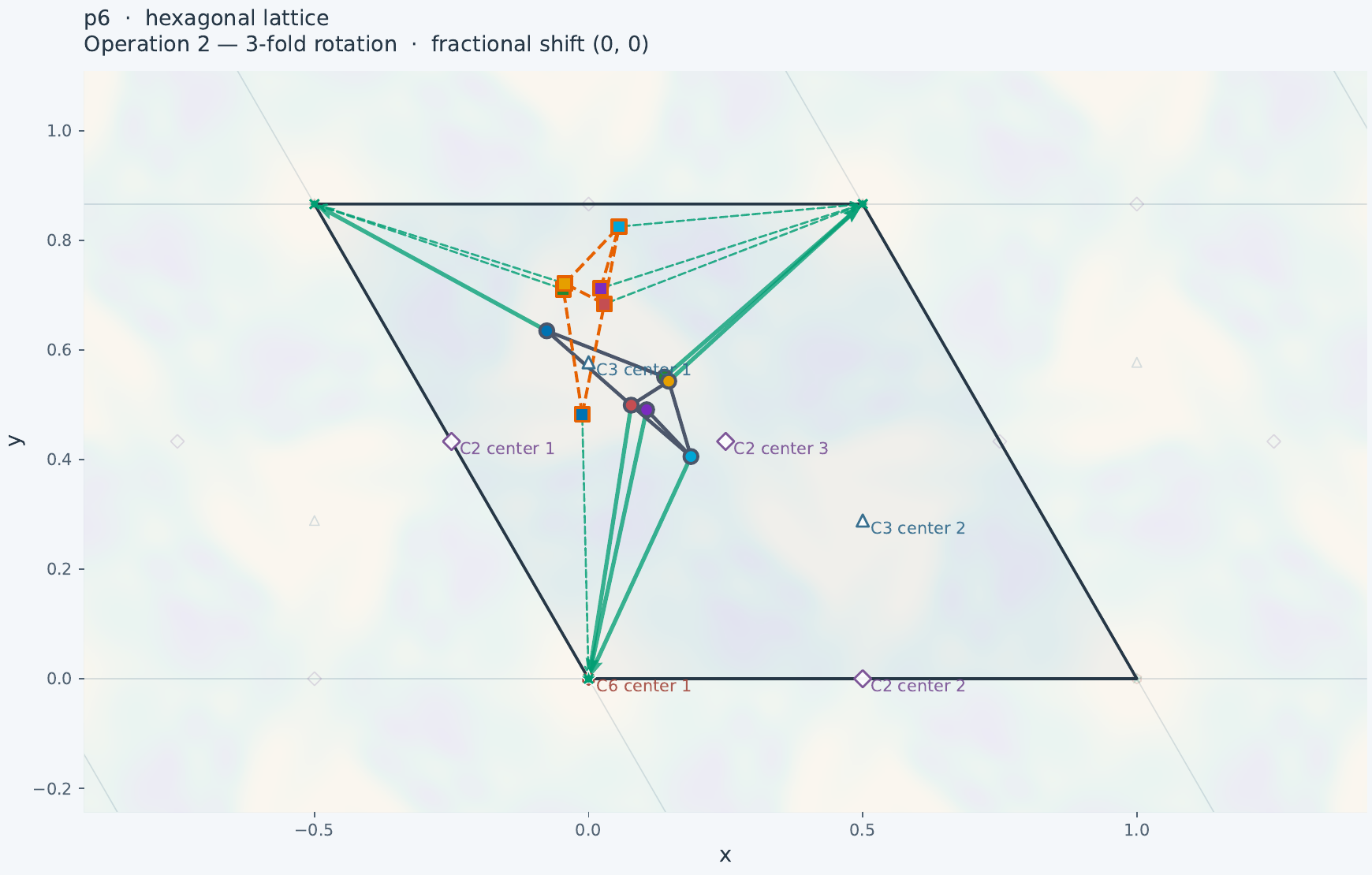} & \includegraphics[width=\linewidth,height=0.22\textheight,keepaspectratio]{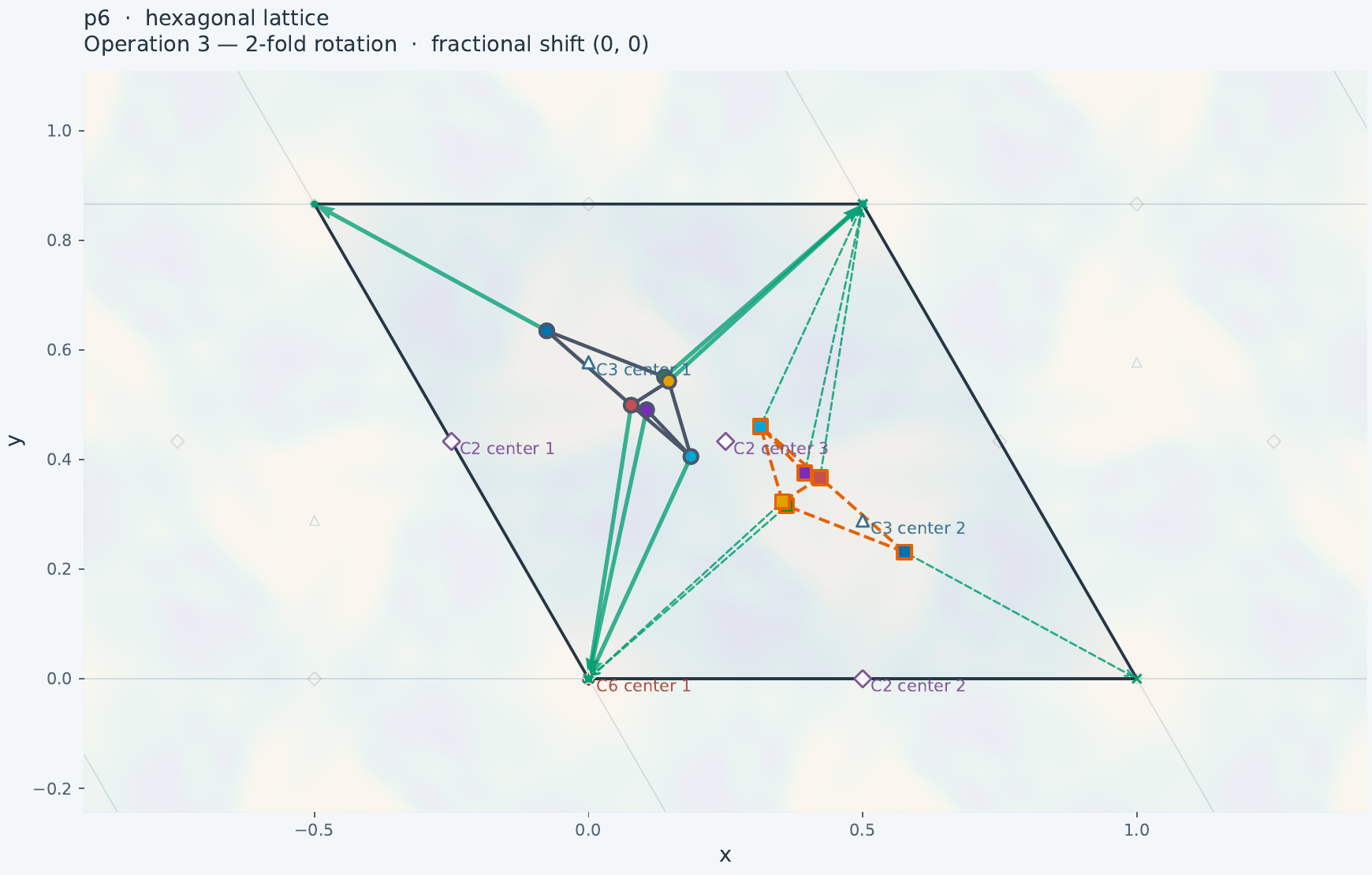} \\[2mm]
\includegraphics[width=\linewidth,height=0.22\textheight,keepaspectratio]{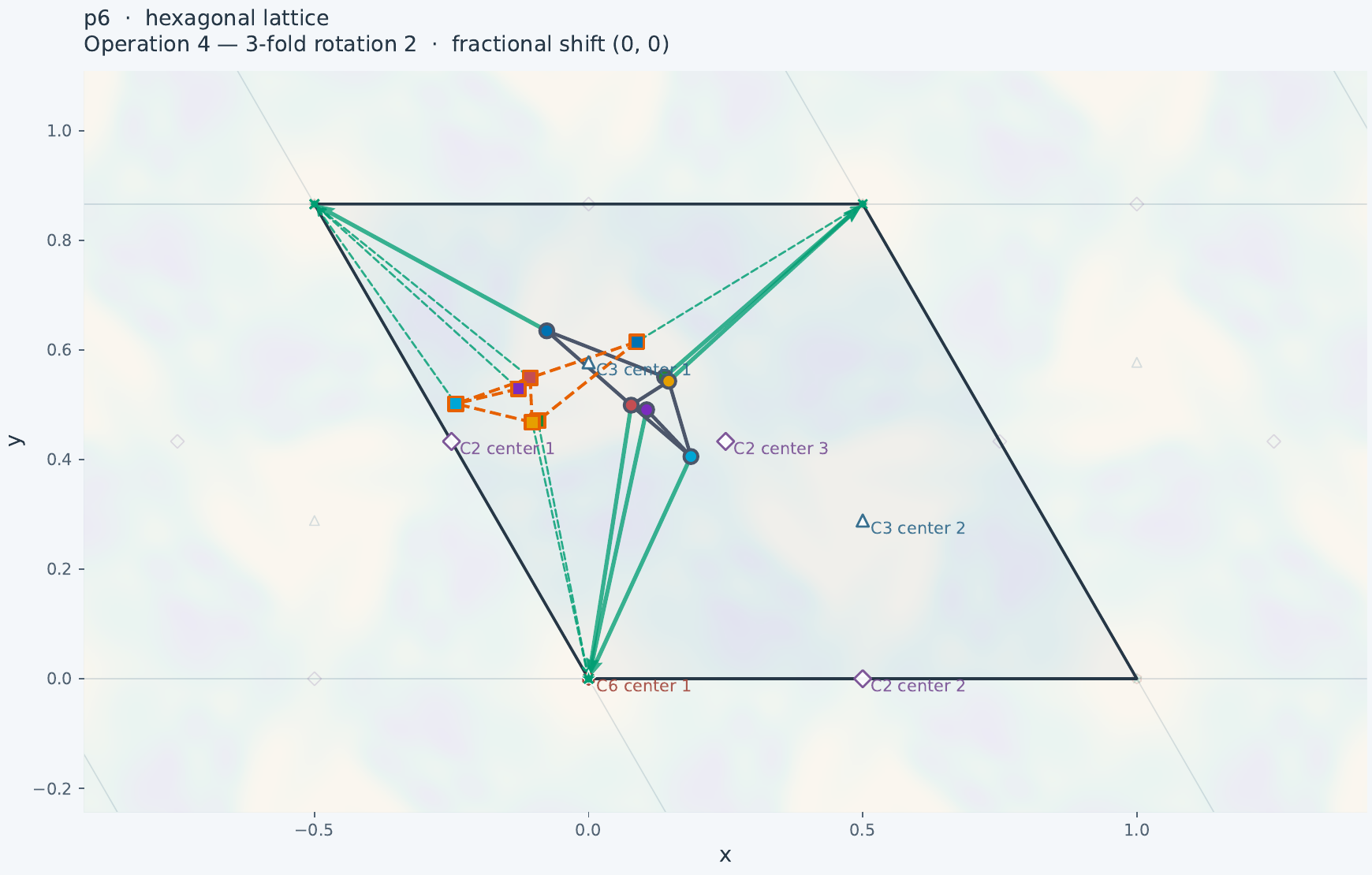} & \includegraphics[width=\linewidth,height=0.22\textheight,keepaspectratio]{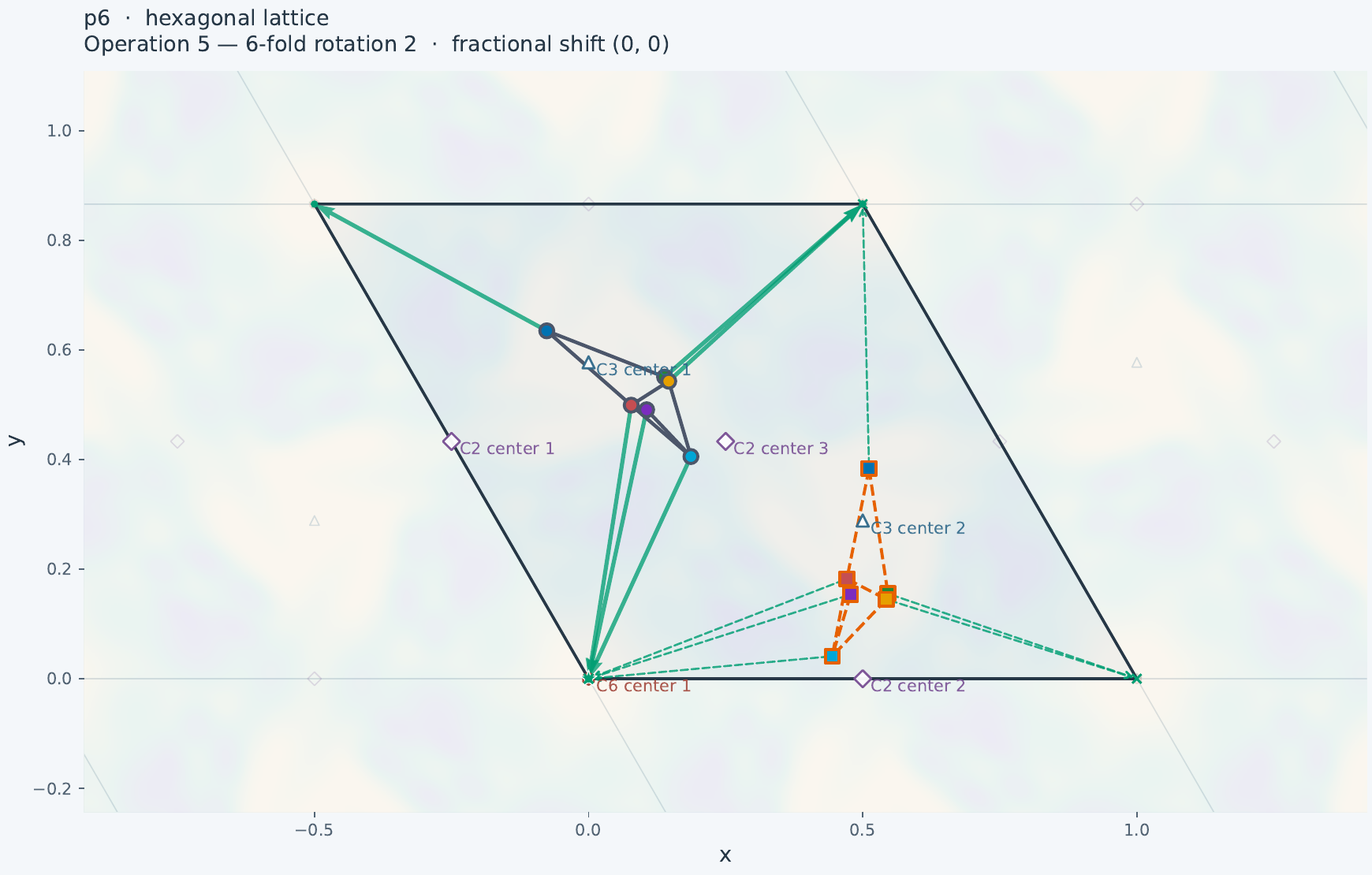} \\[2mm]
\includegraphics[width=\linewidth,height=0.22\textheight,keepaspectratio]{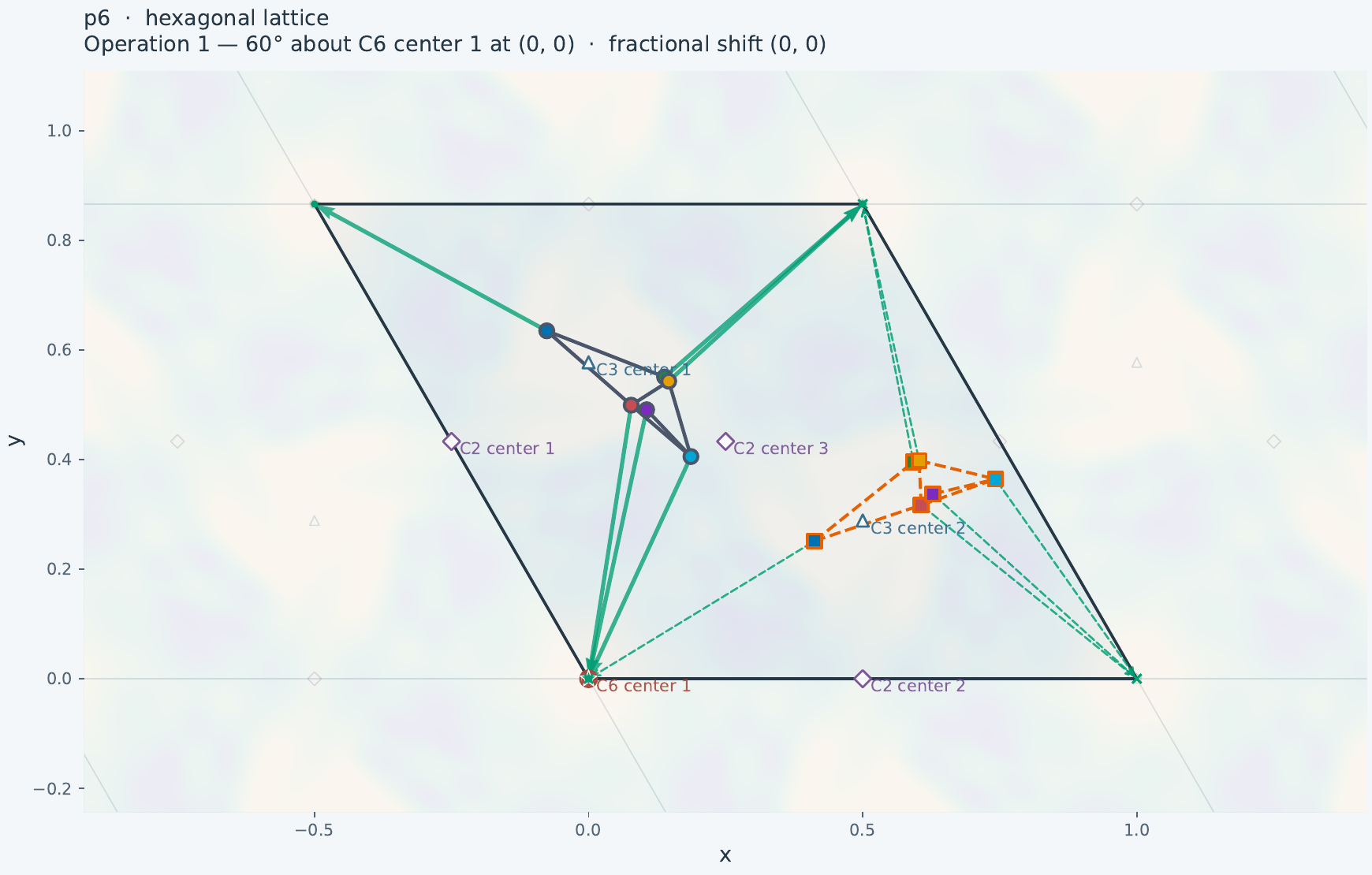} & \includegraphics[width=\linewidth,height=0.22\textheight,keepaspectratio]{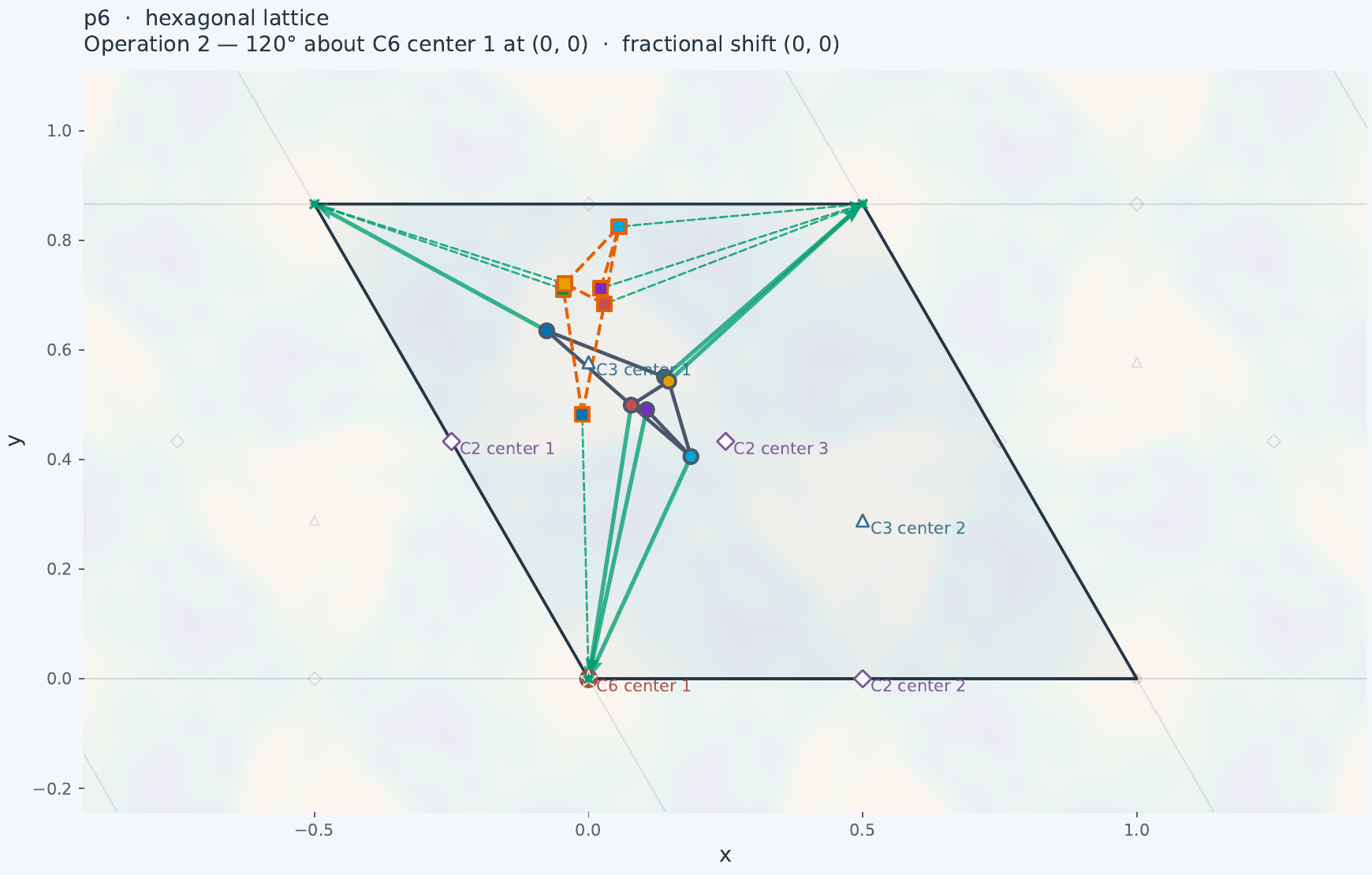} \\[2mm]
\includegraphics[width=\linewidth,height=0.22\textheight,keepaspectratio]{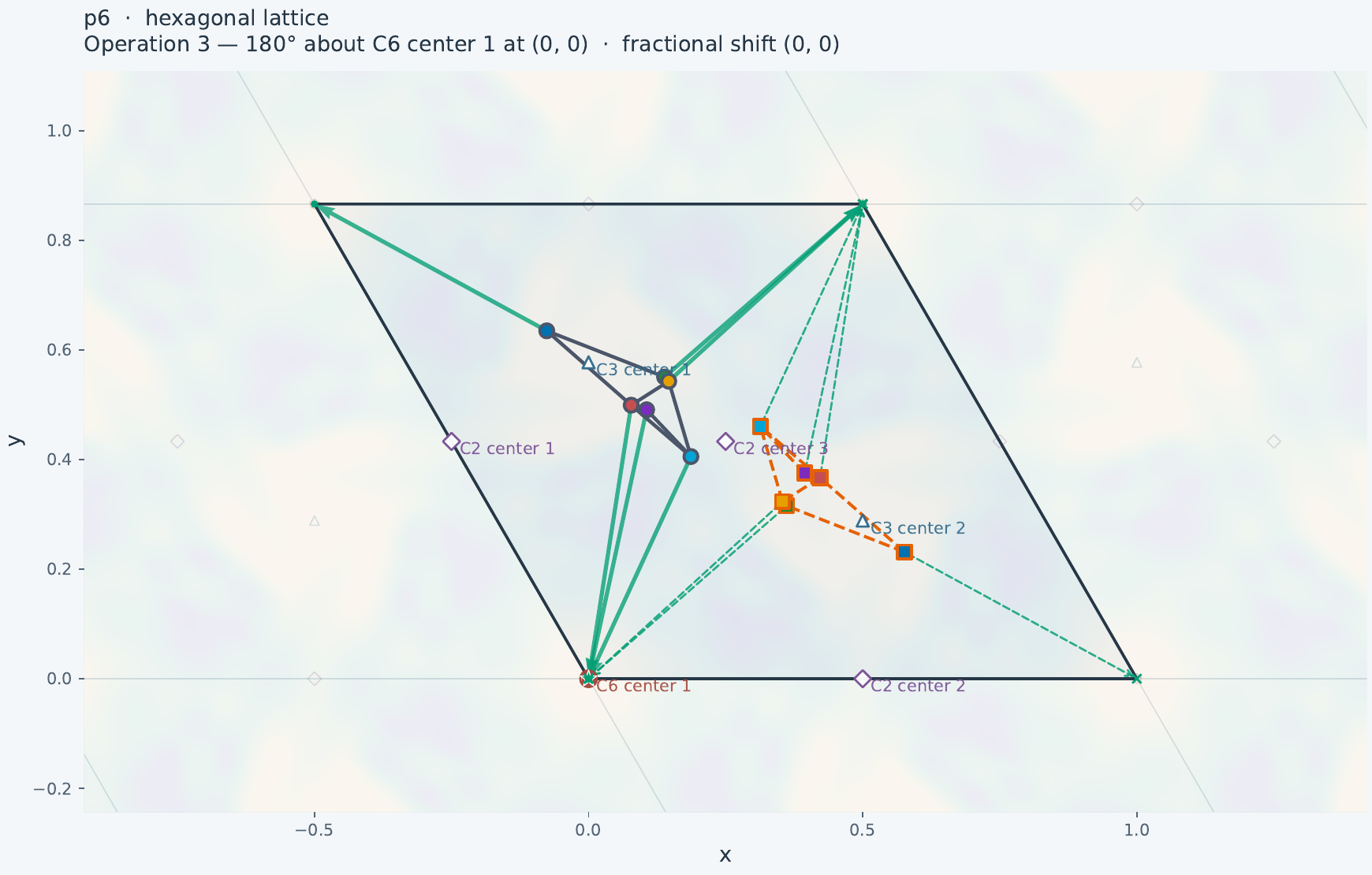} & \includegraphics[width=\linewidth,height=0.22\textheight,keepaspectratio]{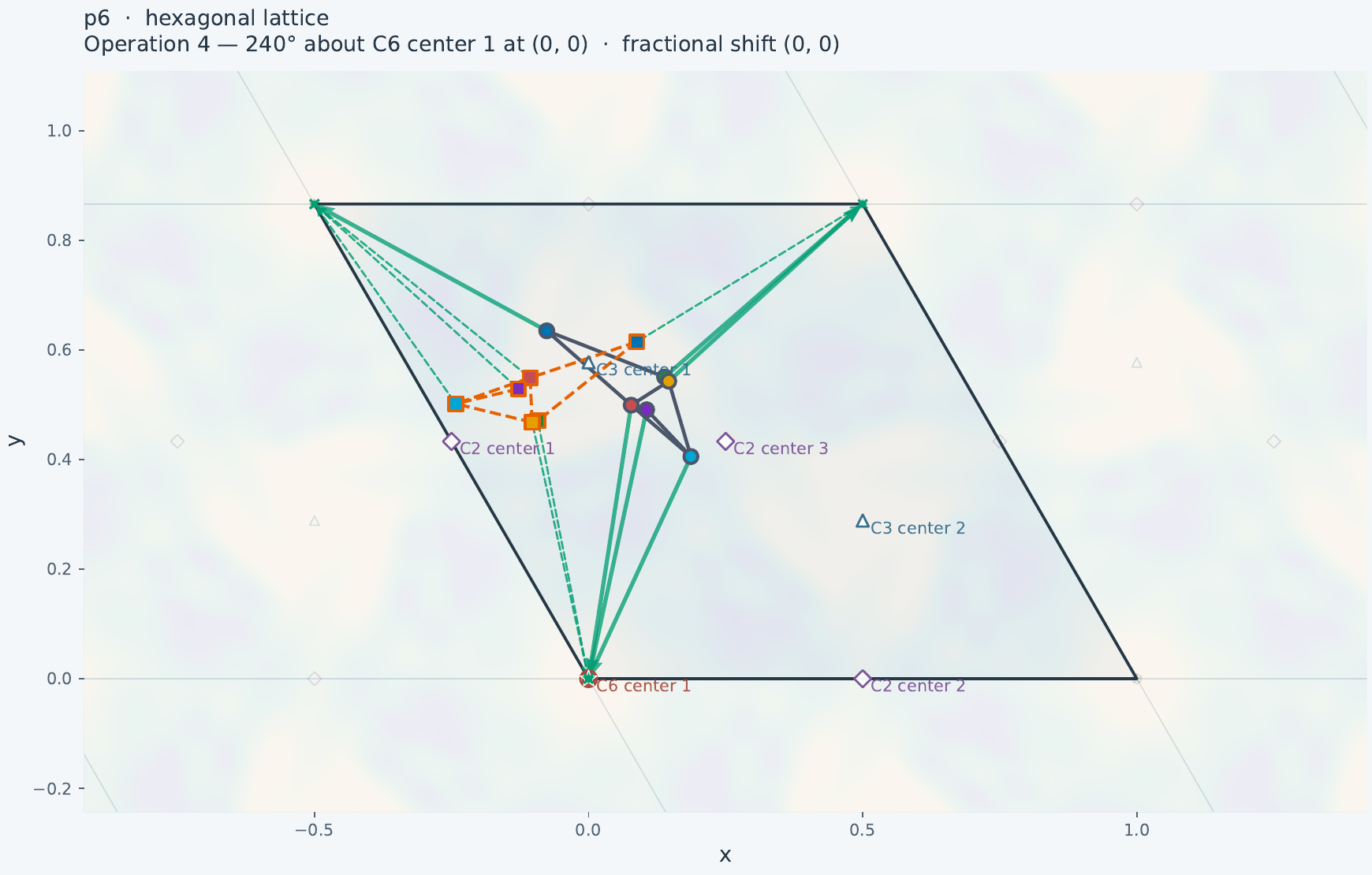} \\[2mm]
\includegraphics[width=\linewidth,height=0.22\textheight,keepaspectratio]{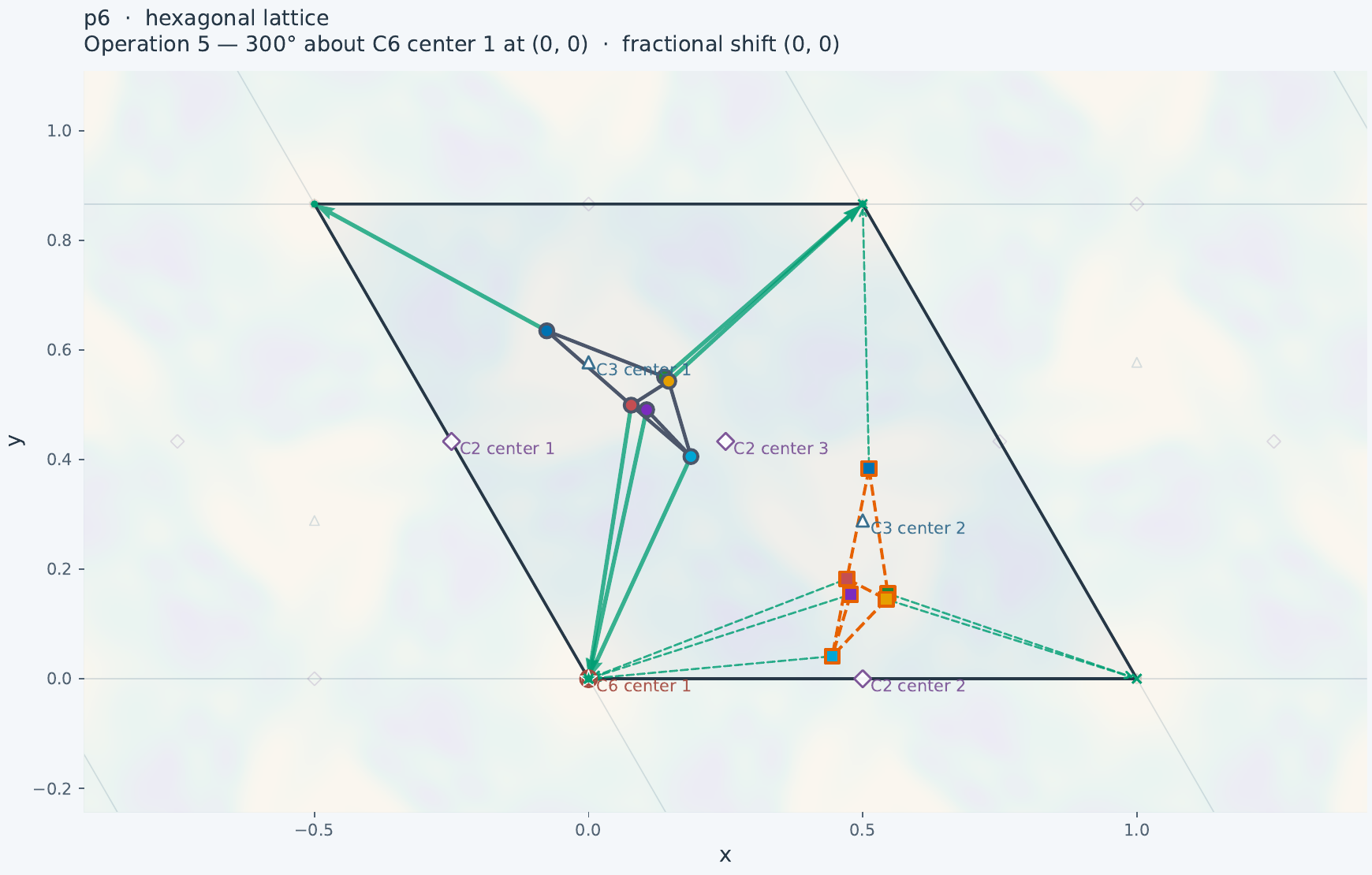} & \includegraphics[width=\linewidth,height=0.22\textheight,keepaspectratio]{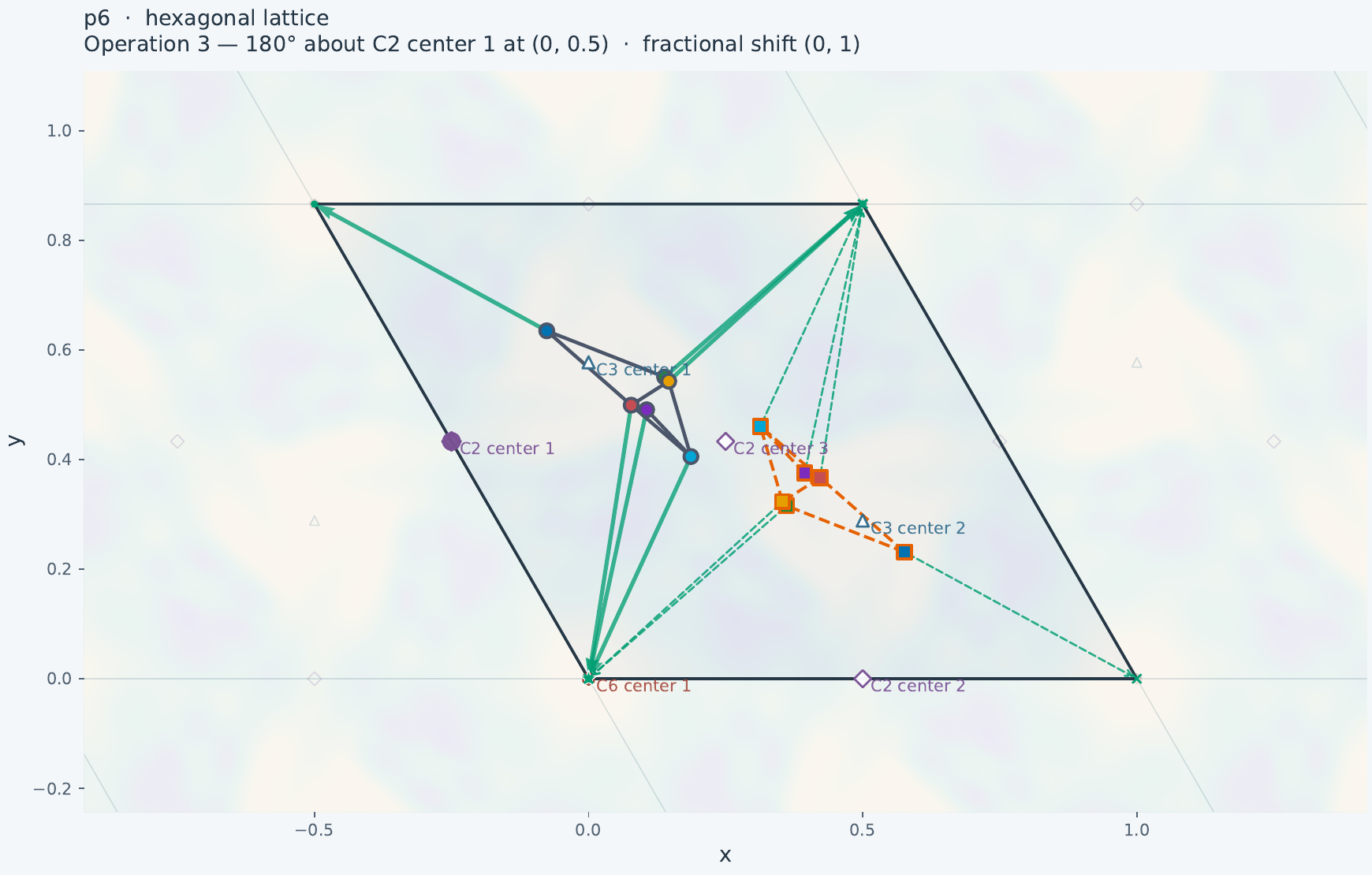} \\
\end{longtable}
\Needspace{236pt}
\begin{longtable}{@{}>{\centering\arraybackslash}p{0.475\linewidth}>{\centering\arraybackslash}p{0.475\linewidth}@{}}
\caption[Wallpaper group \texttt{p6m}]{Wallpaper group \texttt{p6m}, {\small WA is pointing to C6 at $q_{1}=(0,0)$}}\label{tab:wa-actions-p6m}\\[1mm]
\multicolumn{2}{c}{\includegraphics[width=0.96\linewidth]{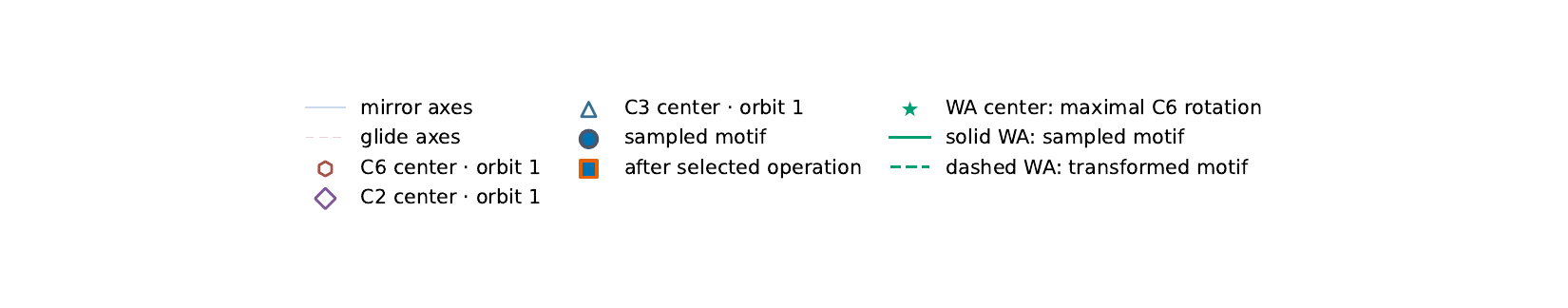}} \\[1mm]
\endfirsthead
\caption[]{Wallpaper group \texttt{p6m} (continued)}\\[1mm]
\endhead
\includegraphics[width=\linewidth,height=0.22\textheight,keepaspectratio]{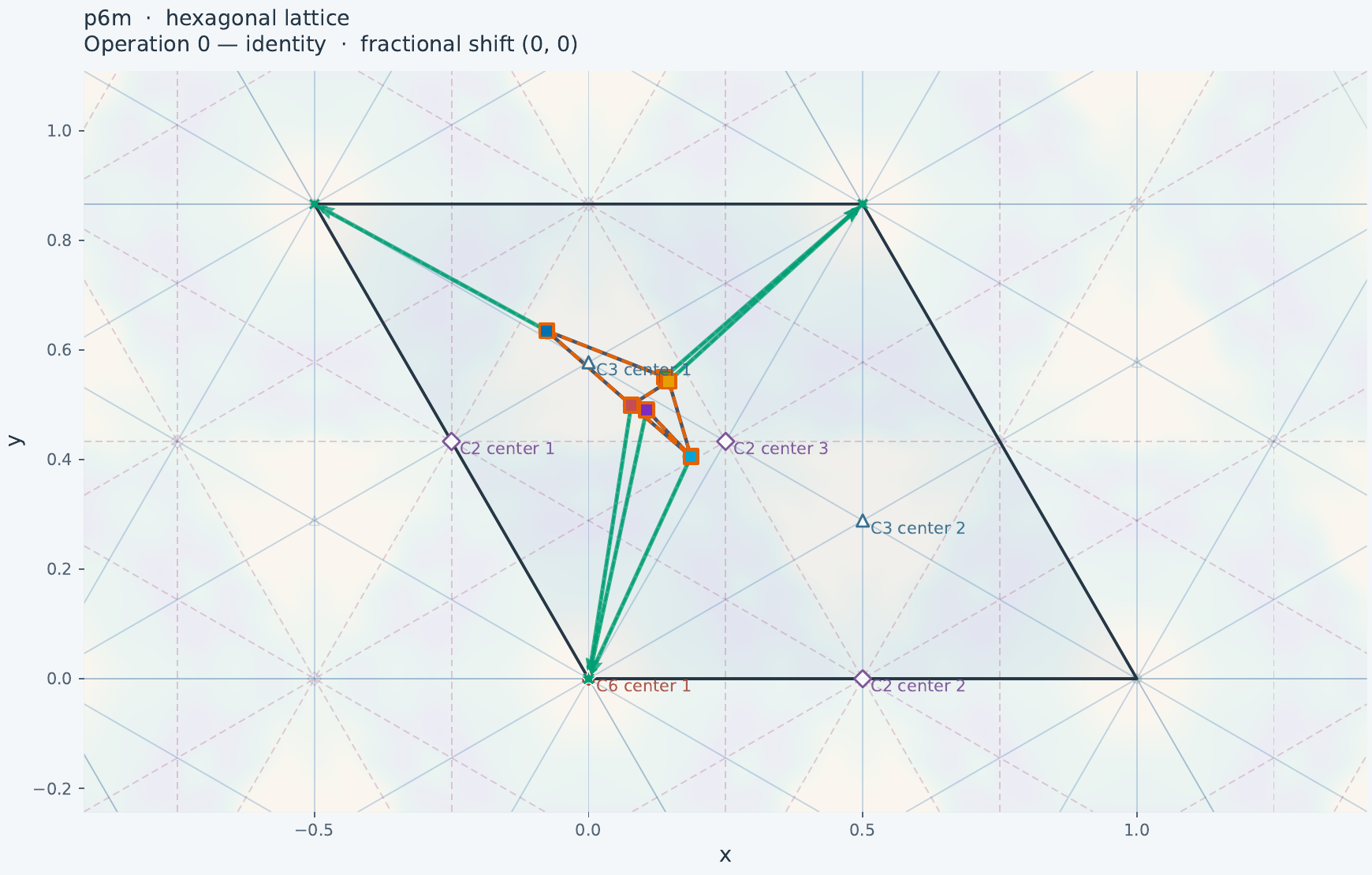} & \includegraphics[width=\linewidth,height=0.22\textheight,keepaspectratio]{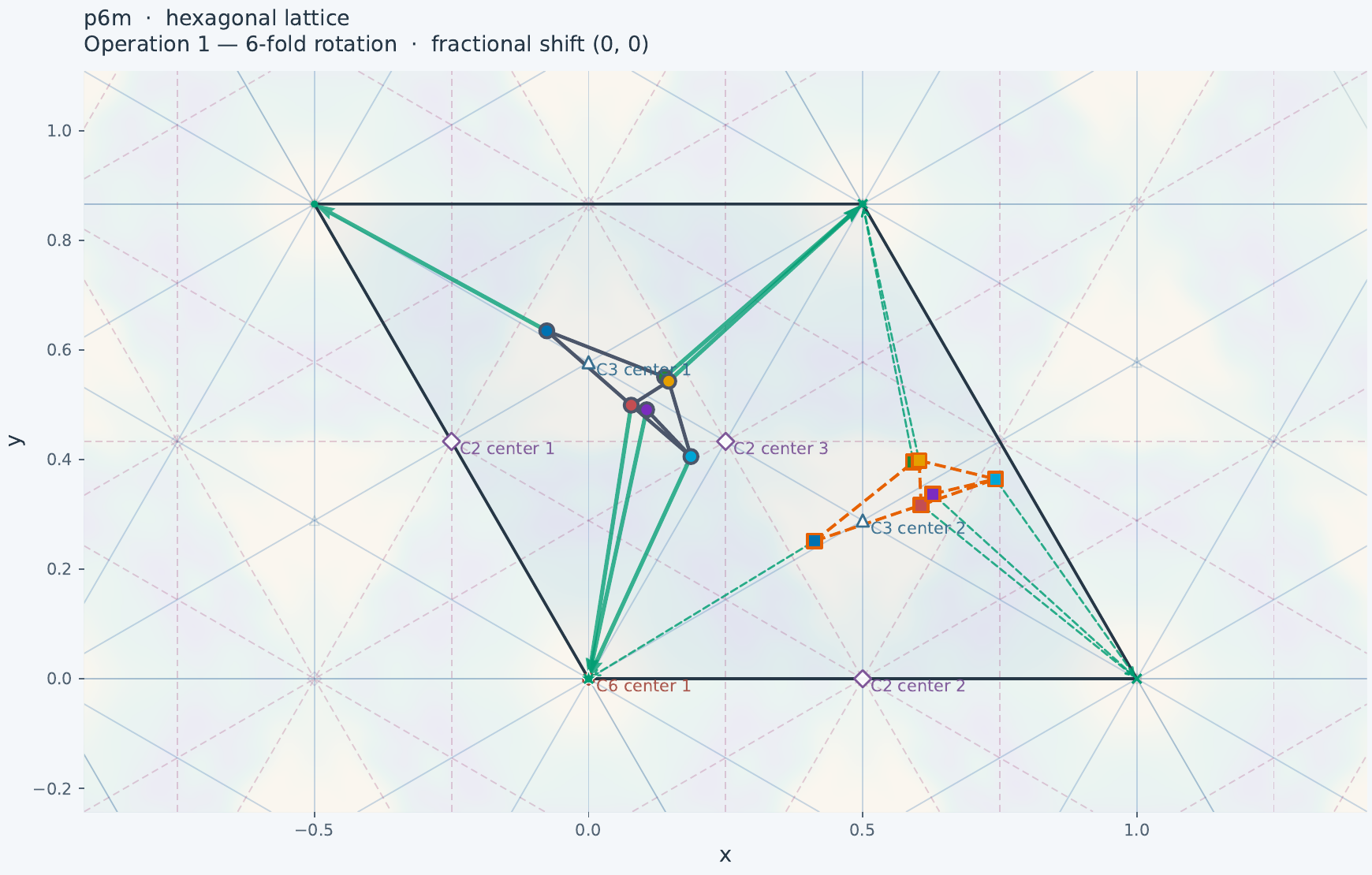} \\[2mm]
\includegraphics[width=\linewidth,height=0.22\textheight,keepaspectratio]{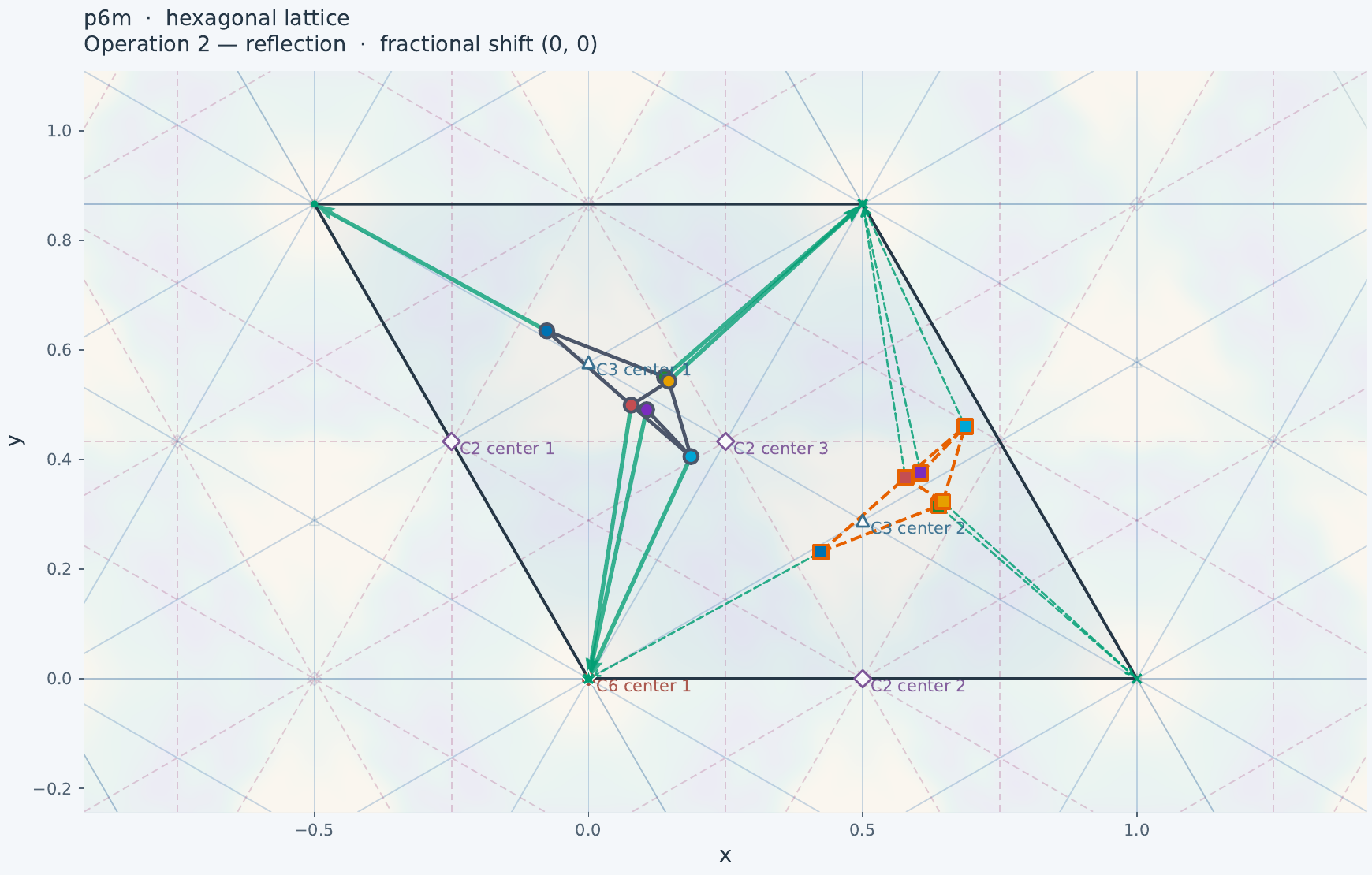} & \includegraphics[width=\linewidth,height=0.22\textheight,keepaspectratio]{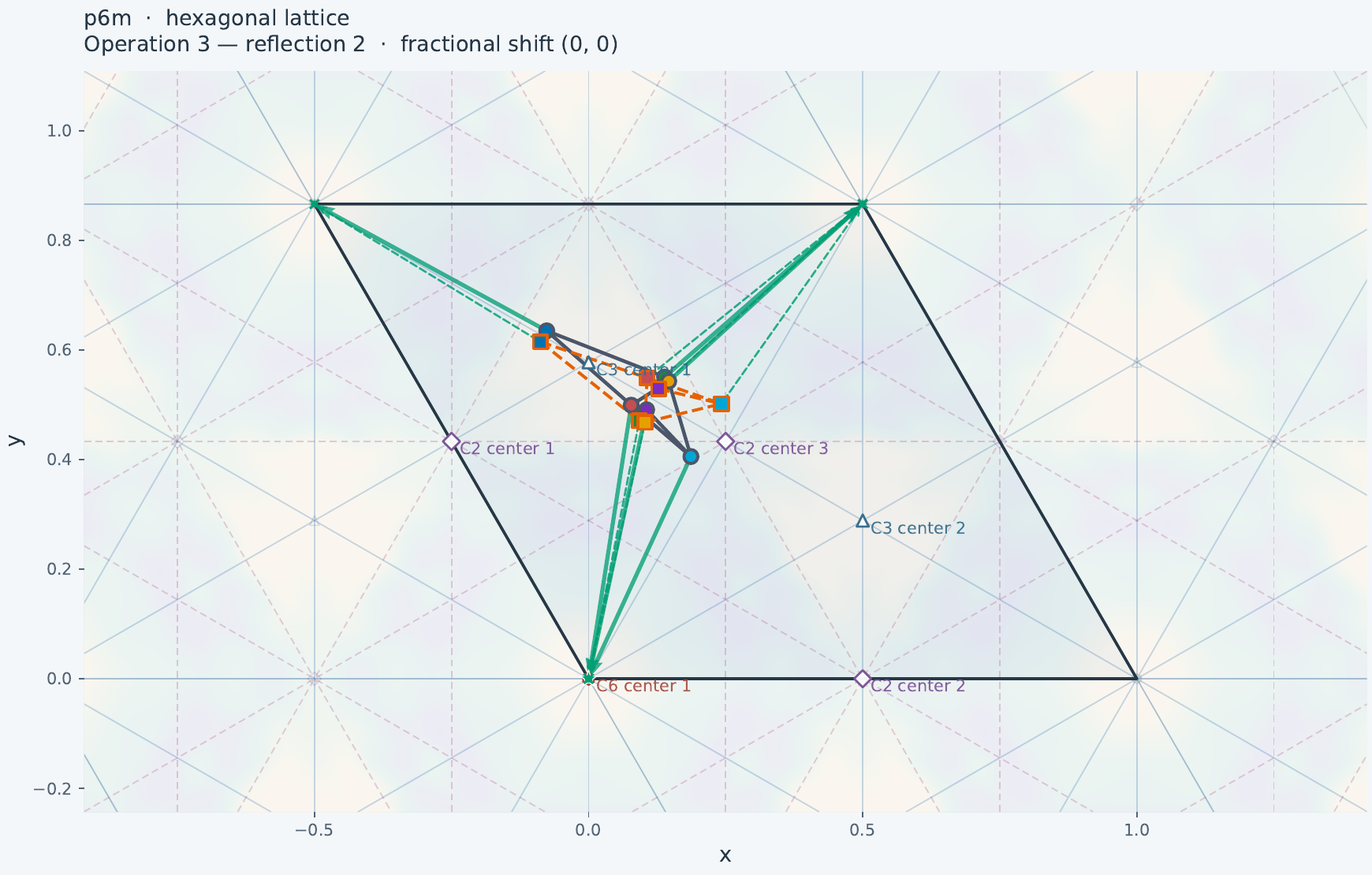} \\[2mm]
\includegraphics[width=\linewidth,height=0.22\textheight,keepaspectratio]{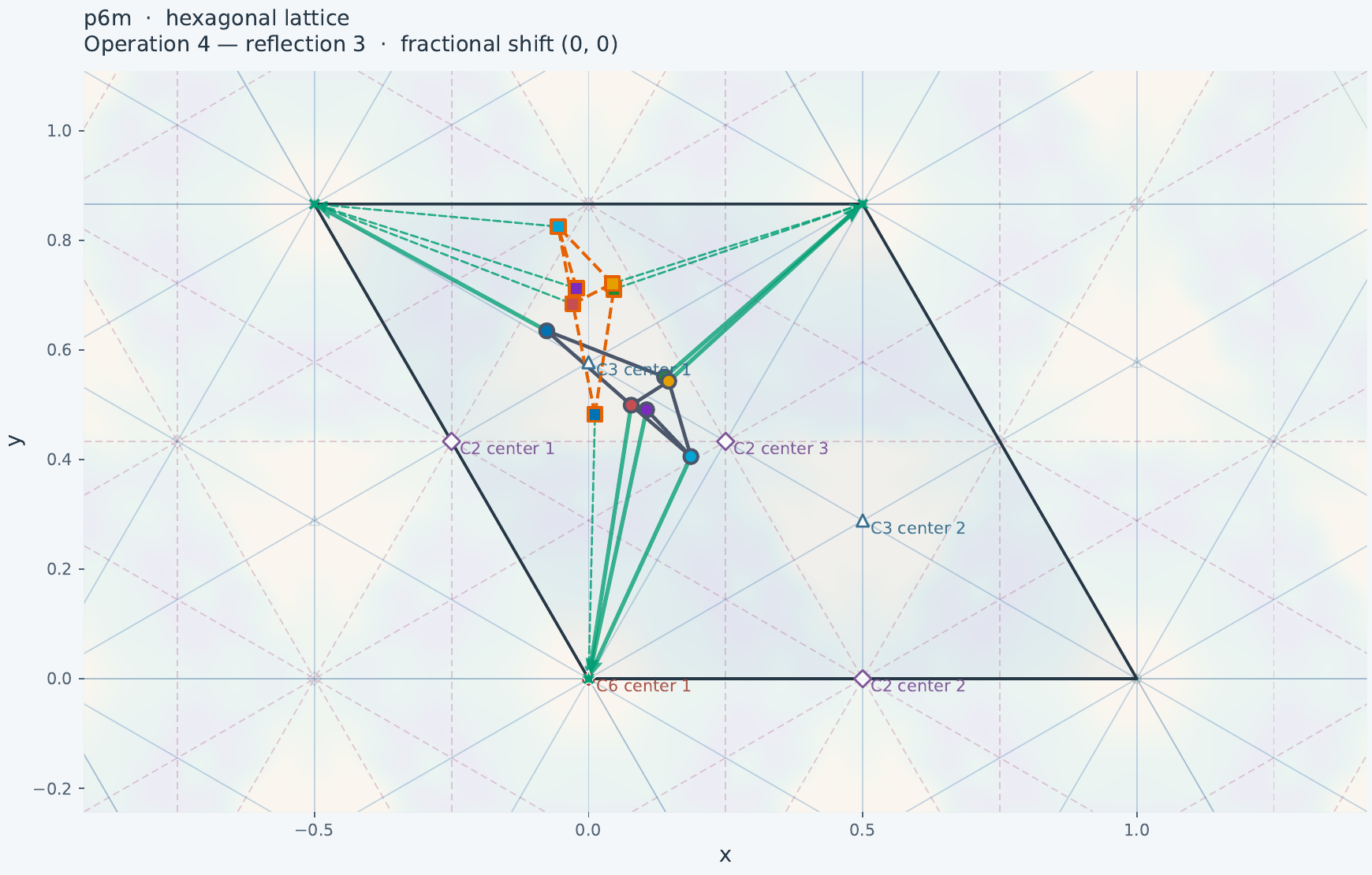} & \includegraphics[width=\linewidth,height=0.22\textheight,keepaspectratio]{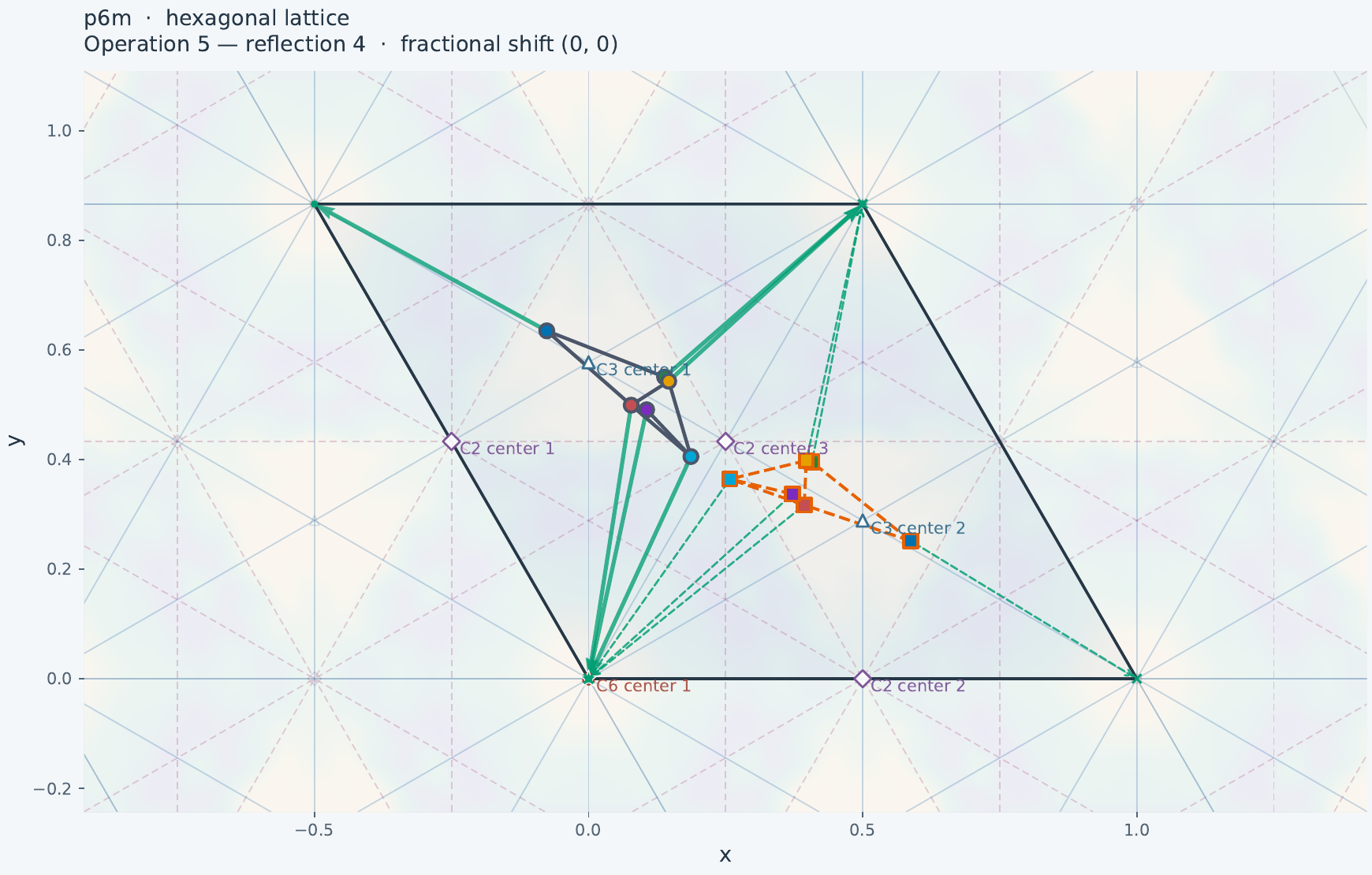} \\[2mm]
\includegraphics[width=\linewidth,height=0.22\textheight,keepaspectratio]{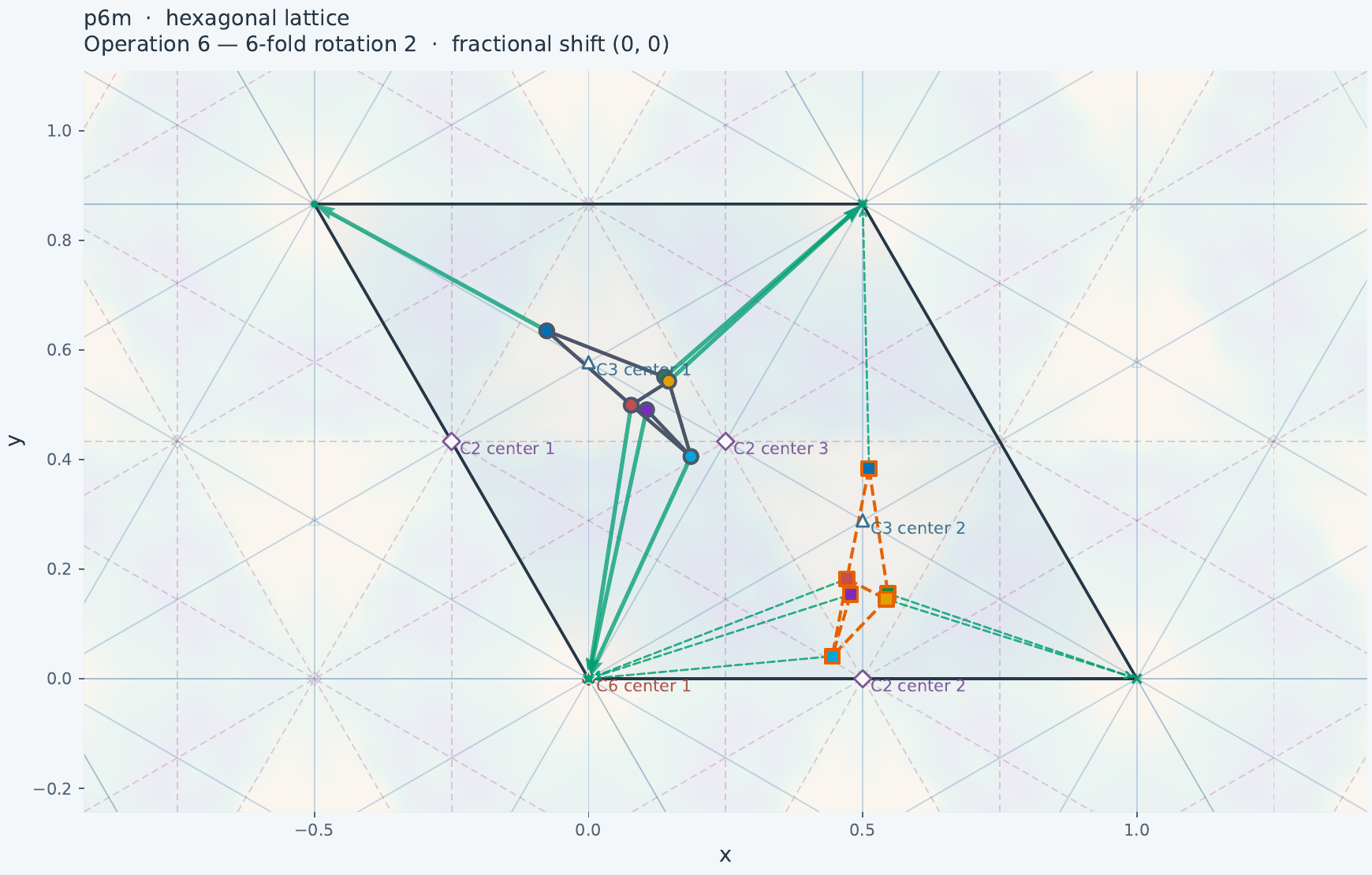} & \includegraphics[width=\linewidth,height=0.22\textheight,keepaspectratio]{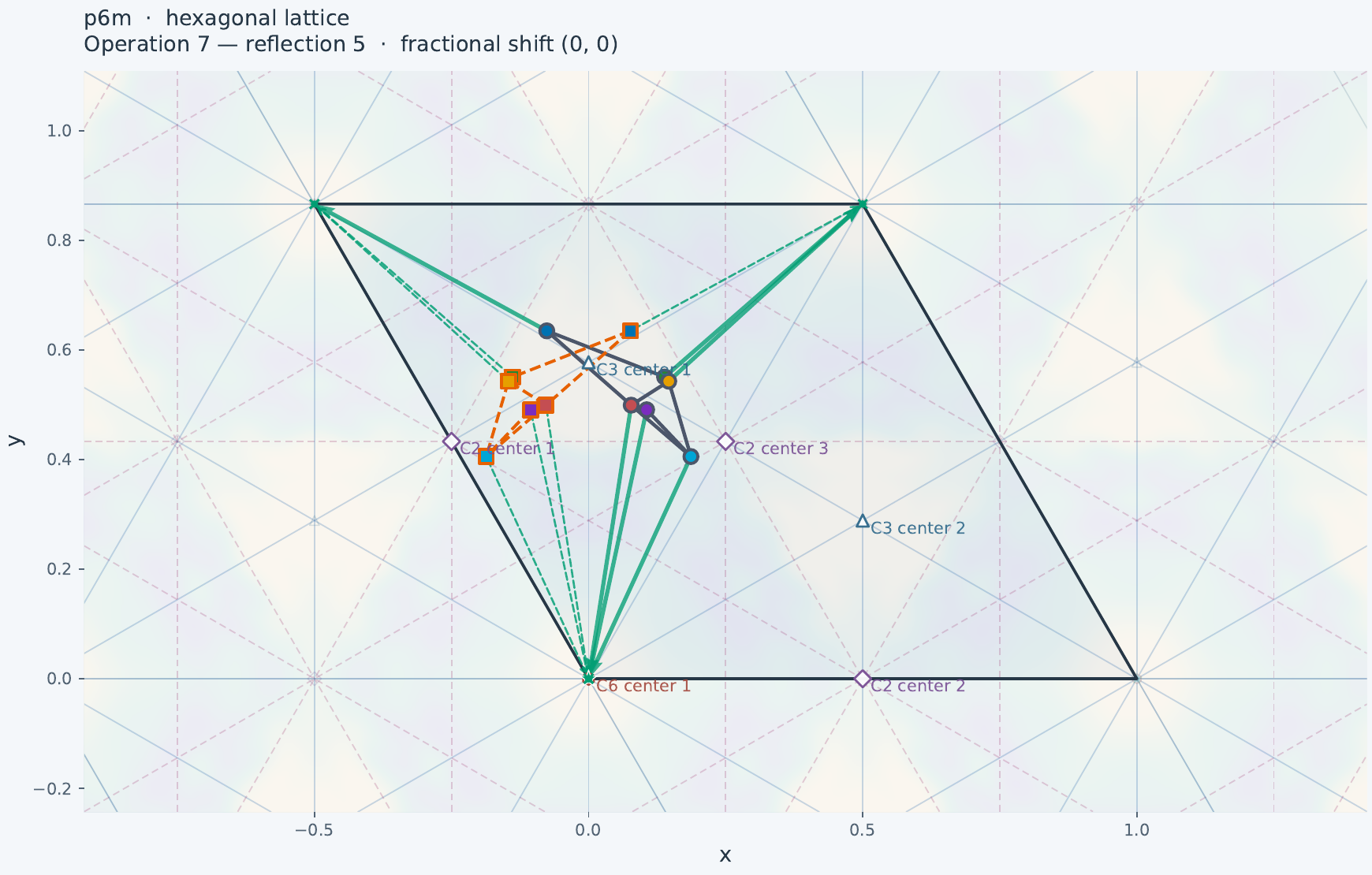} \\[2mm]
\includegraphics[width=\linewidth,height=0.22\textheight,keepaspectratio]{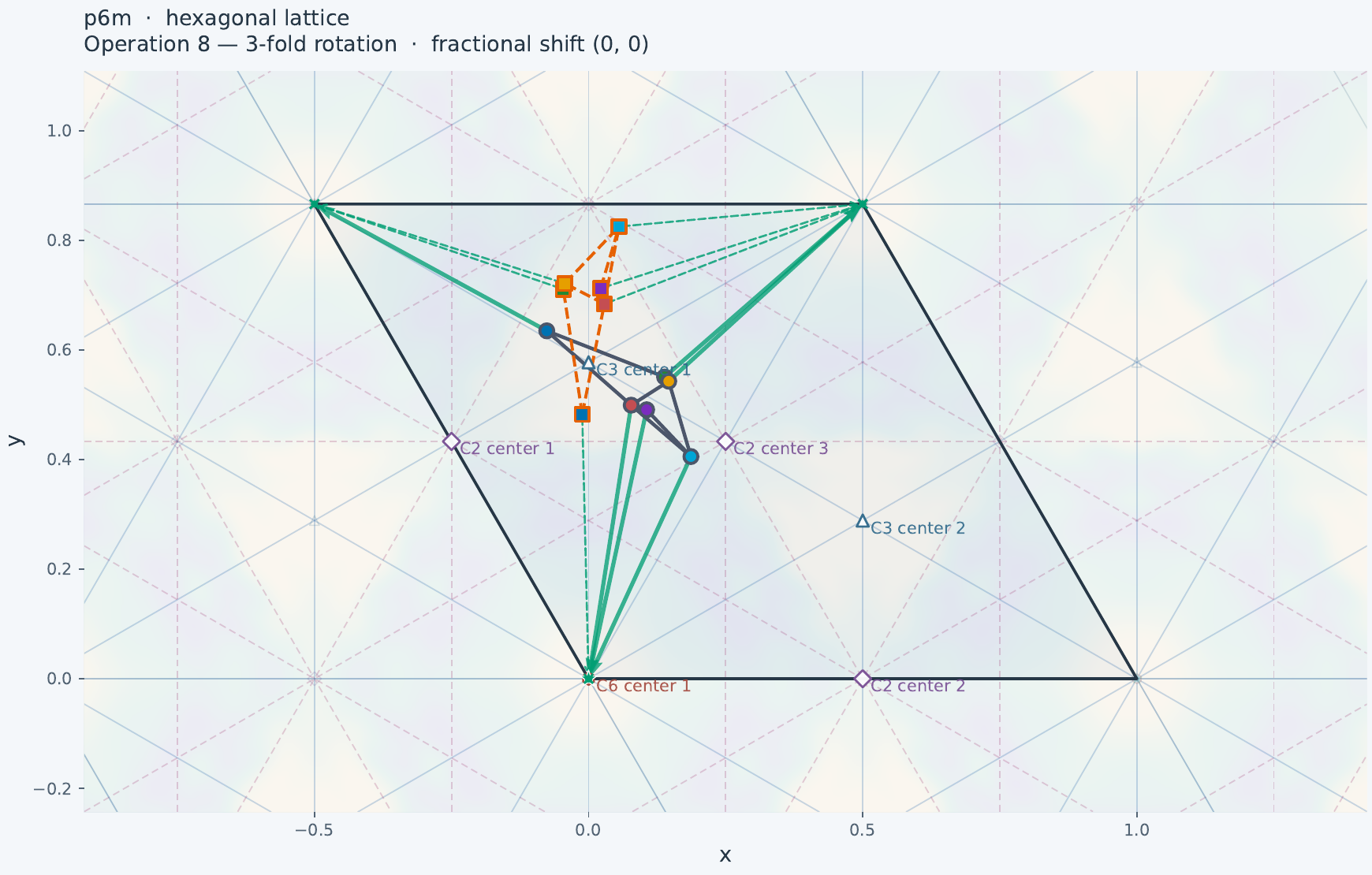} & \includegraphics[width=\linewidth,height=0.22\textheight,keepaspectratio]{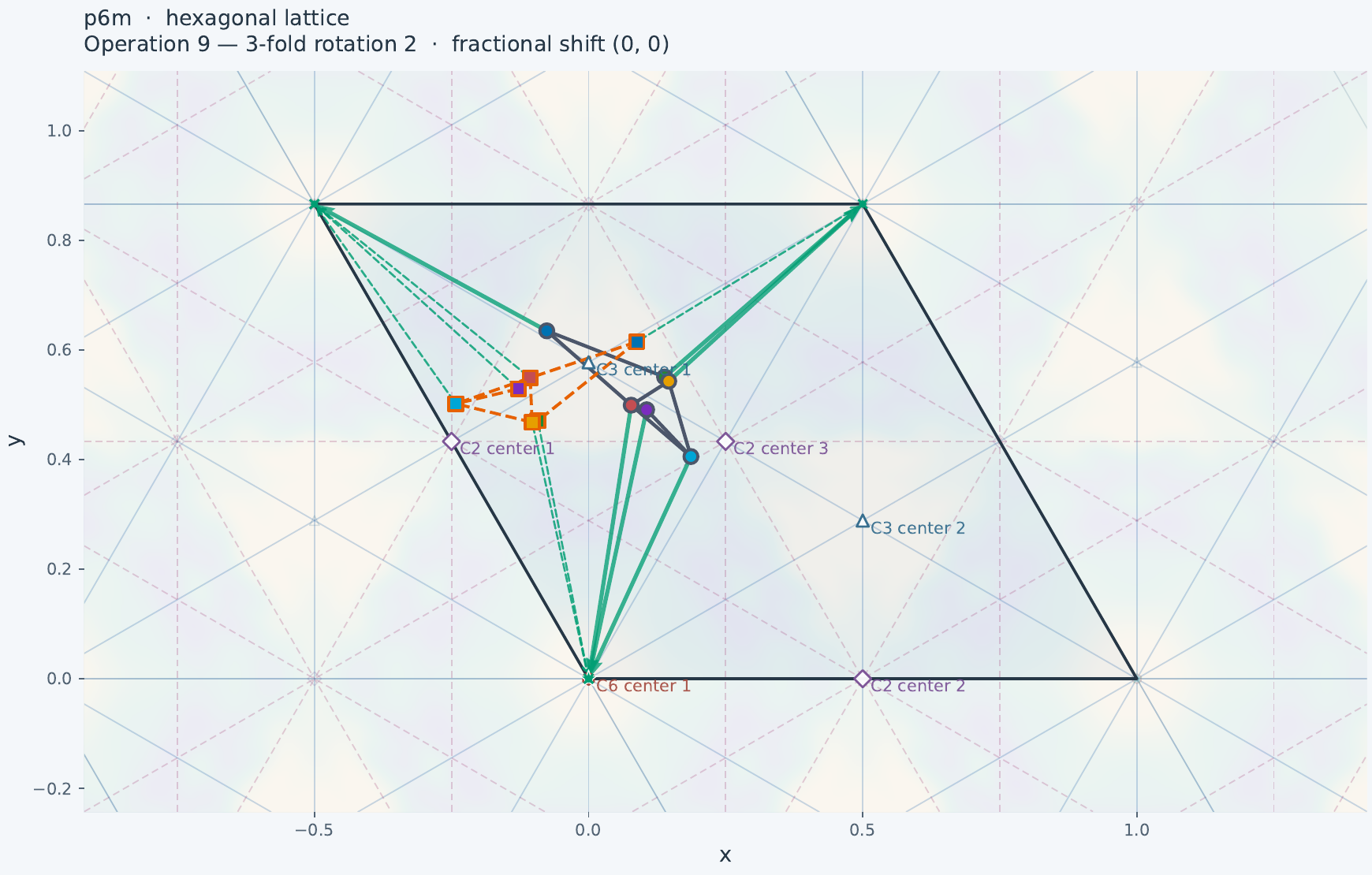} \\[2mm]
\includegraphics[width=\linewidth,height=0.22\textheight,keepaspectratio]{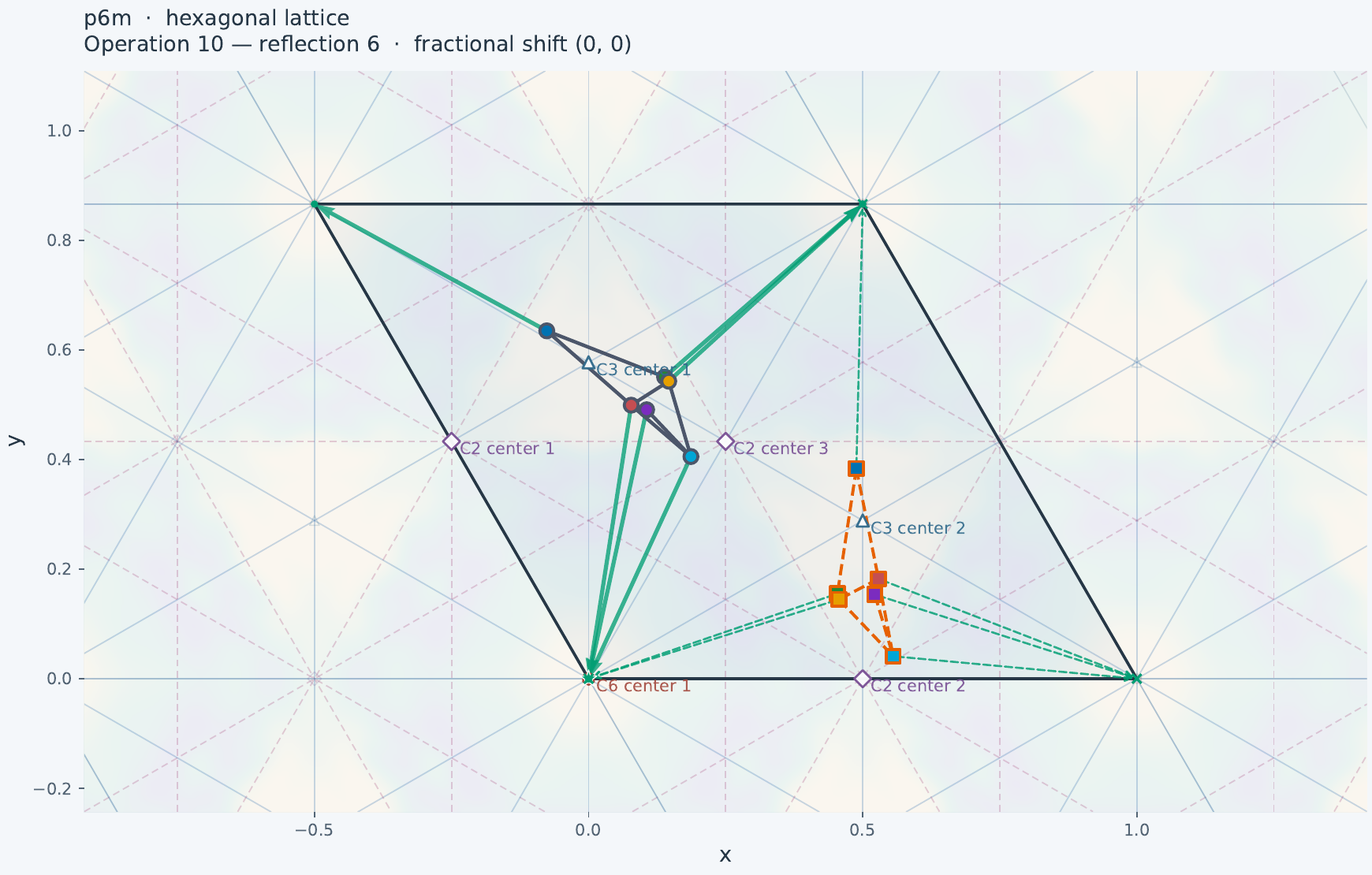} & \includegraphics[width=\linewidth,height=0.22\textheight,keepaspectratio]{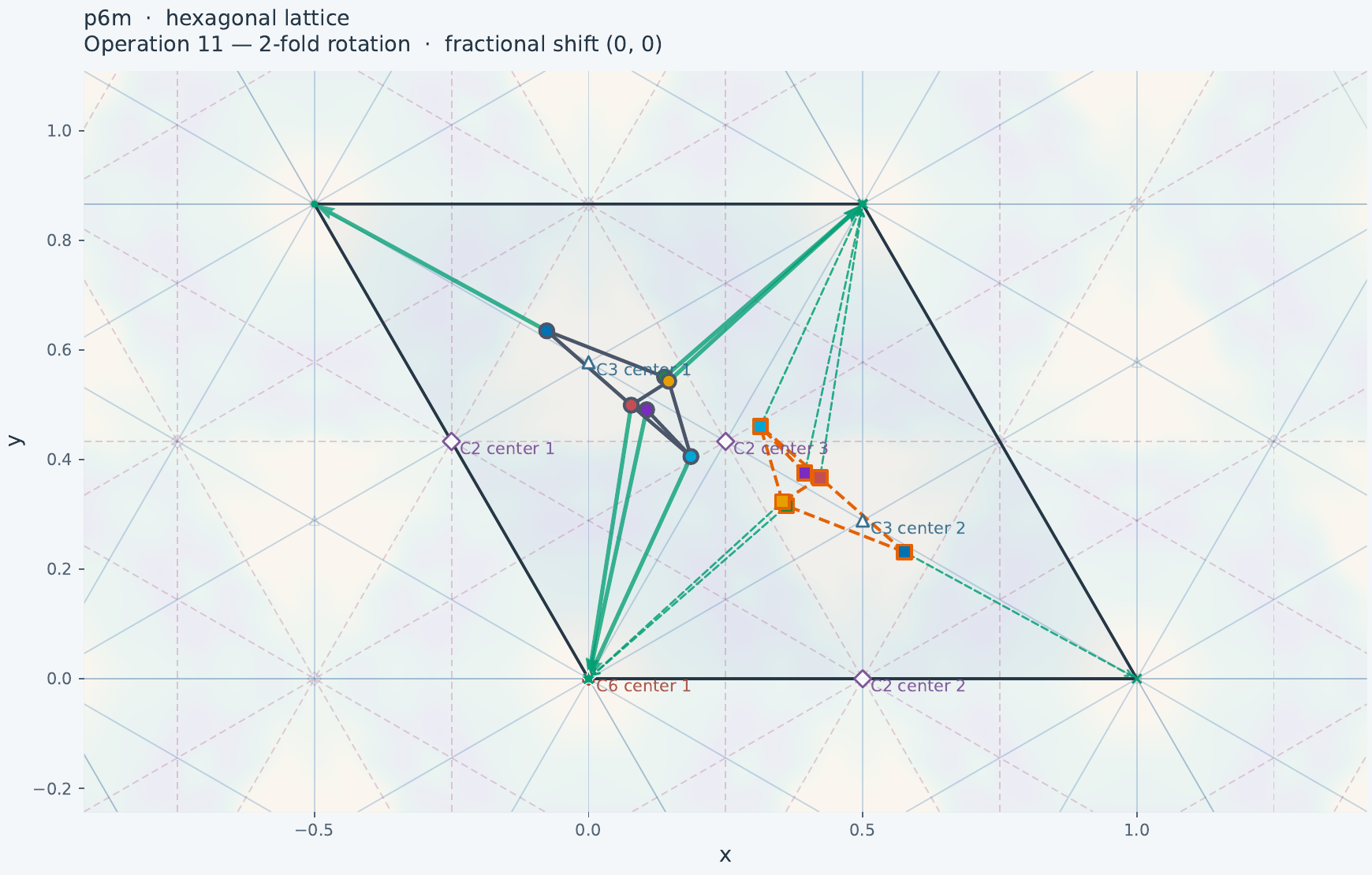} \\
\end{longtable}
\ifdefined\WallpaperTablesEmbedded
\else
\end{document}
\fi

\endgroup

\section{Wyckoff-anchor identifiability across all wallpaper groups}
\label{sec:wa_encoding_exp}

Wallpaper-equivariant message passing based only on relative edge geometry cannot, in general, identify the absolute registry of a graph within a periodic cell. We therefore construct a controlled experiment to isolate the information supplied by the Wyckoff-anchor, to numerically show its importance.

The experiment compares two parameter-matched models. The \emph{WA-aware} model receives the graph, its relative edge vectors, and the WA encoding. The \emph{WA-free} control uses exactly the same equivariant architecture and representation types, but every WA vector and, for nonsymmorphic groups, every translation-suborbit member field is replaced by zero. The comparison therefore isolates the information supplied by the WA representation rather than differences in architecture or parameter count.

This experiment is an identifiability test rather than an atomistic simulation. The graph geometry is fixed, the only variation between samples is its position within the periodic cell, and the prediction target is a known analytic wallpaper-periodic scalar field.

\paragraph{Fixed graph and registry sampling.}
All 17 wallpaper groups are considered. Each sample contains the same asymmetric four-node graph embedded in two dimensions. We treat fractional and Cartesian positions as column vectors. In fractional coordinates, the uncentered node offsets are
\[
\begin{aligned}
p_1&=(-0.040,-0.025), & p_2&=(0.038,-0.018),\\
p_3&=(0.018,0.043), & p_4&=(-0.016,0.011),
\end{aligned}
\]
and their mean is subtracted to obtain centered offsets \(q_i\).

For a sampled graph center \(c\), node \(i\) is positioned at
\[
f_i=c+q_i,
\qquad
x_i=C_Gf_i,
\]
where the columns of \(B_G\in\mathbb R^{2\times2}\) are the conventional-cell lattice vectors for wallpaper group \(G\), and \(C_G=5B_G\) is the cell used in the experiment. The scaling changes the Cartesian length scale but not the fractional symmetry operations.

The graph center is restricted to positions for which all four nodes remain inside the central cell, so no coordinate wrapping occurs. The graph is complete and directed without self-edges, giving 12 directed edges. All nodes use the same learned initial embedding and therefore carry no distinguishing node-type information.

Samples differ only by rigid translation of the complete graph. Consequently,
\[
r_{ij}=x_i-x_j
\]
is identical for every registry position. A graph network that receives only these relative edges therefore obtains exactly the same geometric input for configurations that can nevertheless have different target values. This creates the input collision that the experiment is designed to test.

Each wallpaper group contains 320 training registries and 256 independently sampled test registries. The embedded graph itself is identical in both splits, so the experiment measures generalization to unseen positions within the wallpaper cell rather than generalization to unseen graph geometries.

\paragraph{Analytic wallpaper-periodic target.}
We define a known analytic scalar target over the fractional coordinates of the graph. Starting from an integer reciprocal mode \(k=(k_u,k_v)\), the seed field is
\[
h_k(f)
=
\cos\!\left(2\pi k^{\mathsf T}f\right)
+
0.37\sin\!\left(2\pi k^{\mathsf T}f\right).
\]
Thus, before symmetrization, the target is a simple sinusoidal Fourier field over position in the periodic cell.

Let \(\Lambda_G^{\mathrm{cell}}=C_G\mathbb Z^2\) denote the displayed
conventional-cell translation lattice and let
\(\bar G=G/\Lambda_G^{\mathrm{cell}}\) be the corresponding finite affine
quotient. For each representative \(g\in\bar G\), the action in fractional
coordinates is
\[
g(f)=S_gf+s_g \pmod{\mathbb Z^2},
\qquad g\in\bar G,
\]
where \(S_g=C_G^{-1}Q_gC_G\) is the fractional-coordinate linear part and
\(s_g\) is the fractional translation. This quotient is taken modulo the
displayed conventional-cell translations, rather than the primitive lattice
\(\Lambda\) used in Section~\ref{sec:problem}. It therefore also retains any
centering translation inside the displayed cell. The corresponding
wallpaper-invariant node field is obtained by a finite Reynolds projection,
\[
V_{G,k}(f)
=
\frac{1}{|\bar G|}
\sum_{g\in\bar G}
h_k(S_gf+s_g).
\]
The scalar target for the complete graph is then obtained by averaging this field over its nodes,
\[
E_G(X)
=
\frac{1}{N}
\sum_{i=1}^{N}
V_{G,k_G}(f_i),
\qquad N=4.
\]
By closure of the affine quotient and periodicity of \(h_k\),
\[
E_G(gX)=E_G(X),
\qquad g\in G,
\]
so the resulting target is invariant under the prescribed wallpaper group. It is also invariant to permutations of the graph nodes because the node contributions are averaged.

The reciprocal mode \(k_G\) is chosen as a low-order mode that survives the corresponding wallpaper-group projection. We use \(k_G=(1,0)\) for
\(p1,p2,pm,pmm,p4,p4m,p3,p3m1,p31m,p6,\) and \(p6m\).
For \(cm,cmm,pg,\) and \(pmg\), we use \(k_G=(1,1)\), and for
\(pgg\) and \(p4g\), we use \(k_G=(2,1)\).

In addition to the scalar target, we define an equivariant vector target from its exact coordinate gradient,
\[
F_i=-\nabla_{x_i}E_G(X).
\]
These vector targets are therefore deterministic derivatives of the same analytic scalar field rather than independently generated labels. The learned model is treated analogously: its predicted vector field is obtained by differentiating its predicted scalar output, and no separate direct vector head is used.

\paragraph{Repeated target landscapes across wallpaper groups.}
The analytic target is constructed to be invariant under a prescribed wallpaper group, but it is not intended to uniquely identify that group. In particular, Reynolds projection guarantees (G)-invariance but does not require (G) to be the full symmetry group of the resulting scalar field. Because we use a single low-order reciprocal-lattice mode, the selected mode can already possess additional symmetries, and the projection for different wallpaper groups can therefore produce identical or visually equivalent periodic landscapes. This behavior is expected and does not affect the purpose of the experiment. The comparison is performed independently within each wallpaper group and tests whether the network can identify the registry of a rigidly translated graph. It does not test whether the analytic target itself distinguishes the 17 wallpaper groups.

\paragraph{Equivariant model.}
Both variants use the same single-message-passing-layer, finite-point-group-equivariant \texttt{WELinear} architecture. Every node starts from the same learned 24-dimensional scalar embedding. For each directed edge \(j\rightarrow i\), the WA-aware input contains the receiver and sender scalar embeddings, the squared edge distance, the planar relative edge vector, the planar WA vector, and, for nonsymmorphic groups, the translation-suborbit member representation. Because the four-node graph is complete and directed without self-edges, each node receives three messages, giving 12 directed edge messages per configuration. The aggregated 24-dimensional message is concatenated with the 24-dimensional node state and passed through a \(48\!\to\!24\!\to\!24\) residual node MLP. The final node readout has widths \(24\!\to\!24\!\to\!1\), and the four scalar node outputs are mean-pooled to obtain the graph-level prediction.

The typed edge representation is mapped by an equivariant linear layer to 16 copies of the regular representation of the corresponding finite point group. This is followed by a representation-compatible ELU nonlinearity and a second equivariant linear map to 24 trivial scalar channels,
\[
z_{ij}
\longrightarrow
16\,\rho_{\mathrm{reg}}
\longrightarrow
16\,\rho_{\mathrm{reg}}
\longrightarrow
m_{ij}\in24\,\rho_{\mathrm{triv}}.
\]
The scalar edge messages are summed at the receiver and used in a residual node update,
\[
h_i'
=
h_i+
\operatorname{MLP}_{\mathrm{node}}
\left(
[h_i,\sum_{j\neq i}m_{ij}]
\right).
\]
A scalar readout is subsequently applied to each node and mean-pooled over the graph.

The WA-free control uses the same architecture, representation types, widths, and number of trainable parameters within each wallpaper group. Its only modification is that WA features are replaced by 0s.

\paragraph{Optimization and evaluation.}
Only training-set statistics are used to normalize the scalar target,
\[
\widetilde E
=
\frac{E-\mu_E}{\sigma_E},
\qquad
\widetilde F
=
\frac{F}{\sigma_E}.
\]
The force normalization follows directly from differentiating the normalized
energy. In particular,
\[
\widehat{\widetilde F}_i
=
-\nabla_{x_i}\widehat{\widetilde E},
\qquad
\widehat F_i=\sigma_E\widehat{\widetilde F}_i.
\]
The training objective is
\[
\mathcal L
=
\operatorname{MSE}
(\widehat{\widetilde E},\widetilde E)
+
0.2\,
\operatorname{MSE}
(\widehat{\widetilde F},\widetilde F).
\]
Both models are trained for 500 epochs using AdamW with batch size 64, an initial learning rate of \(3\times10^{-3}\), weight decay \(10^{-6}\), cosine learning-rate decay to \(5\%\) of the initial value, and global gradient clipping at 10. No validation split or early stopping is used, and the final epoch is evaluated.

We report scalar-target MAE, RMSE, and \(R^2\), together with force-component
MAE and a dataset-level relative force \(L_2\) error. If \(s\) indexes test samples,
the latter is
\[
\varepsilon_{F,L_2}^{\mathrm{rel}}
=
\frac{
\left(\sum_s\sum_{i=1}^{N}
\lVert\widehat F_i^{(s)}-F_i^{(s)}\rVert_2^2\right)^{1/2}}
{
\left(\sum_s\sum_{i=1}^{N}
\lVert F_i^{(s)}\rVert_2^2\right)^{1/2}}.
\]

\begin{table*}[!t]
\centering
\caption{Held-out errors for the Wyckoff-anchor identifiability experiment.
All metrics are evaluated on 256 independently sampled test translations per
wallpaper group. Energy and force errors are reported in the arbitrary units
of the analytic target. Lower is better for all error metrics, and higher is
better for energy \(R^2\). ``WA'' denotes \texttt{WELinear} with the complete
WA encoding, and ``No WA'' denotes its parameter-matched WA-free control.}
\label{tab:wa-identifiability-errors}
\setlength{\tabcolsep}{3.5pt}
\small
\begin{tabular}{@{}llcccccc@{}}
\toprule
Group & Model & Parameters & \shortstack{Energy\\MAE} &
\shortstack{Energy\\RMSE} & \shortstack{Energy\\\(R^2\)} &
\shortstack{Force-component\\MAE} & \shortstack{Relative force\\\(L_2\)} \\
\midrule
\(p1\)   & WA    & 3697 & 0.002613 & 0.003625 & 0.999974 & 0.004610 & 0.04473 \\
           & No WA & 3697 & 0.634311 & 0.707385 & $-$0.008664 & 0.152430 & 1.00000 \\
\addlinespace[1pt]
\(p2\)   & WA    & 3697 & 0.002398 & 0.003072 & 0.999979 & 0.004576 & 0.03931 \\
           & No WA & 3697 & 0.593303 & 0.665364 & $-$0.005738 & 0.143588 & 1.00000 \\
\addlinespace[1pt]
\(pm\)   & WA    & 3697 & 0.002929 & 0.003475 & 0.999974 & 0.003460 & 0.04237 \\
           & No WA & 3697 & 0.607179 & 0.694349 & $-$0.027733 & 0.092653 & 0.99999 \\
\addlinespace[1pt]
\(pg\)   & WA    & 3761 & 0.005303 & 0.006673 & 0.999846 & 0.011865 & 0.10526 \\
           & No WA & 3761 & 0.450136 & 0.537128 & $-$0.001104 & 0.137822 & 1.00000 \\
\addlinespace[1pt]
\(cm\)   & WA    & 3697 & 0.004503 & 0.005898 & 0.999868 & 0.008377 & 0.07944 \\
           & No WA & 3697 & 0.419240 & 0.513149 & $-$0.000015 & 0.136741 & 1.00000 \\
\addlinespace[1pt]
\(pmm\)  & WA    & 3697 & 0.001333 & 0.001654 & 0.999993 & 0.002161 & 0.02753 \\
           & No WA & 3697 & 0.582893 & 0.660050 & $-$0.056104 & 0.085241 & 1.00000 \\
\addlinespace[1pt]
\(pmg\)  & WA    & 3825 & 0.004778 & 0.005853 & 0.999880 & 0.009671 & 0.08614 \\
           & No WA & 3825 & 0.464532 & 0.535519 & $-$0.001146 & 0.134341 & 1.00000 \\
\addlinespace[1pt]
\(pgg\)  & WA    & 3825 & 0.005683 & 0.006936 & 0.999819 & 0.014312 & 0.08448 \\
           & No WA & 3825 & 0.445993 & 0.515654 & $-$0.001492 & 0.181650 & 0.99997 \\
\addlinespace[1pt]
\(cmm\)  & WA    & 3697 & 0.014763 & 0.018893 & 0.998436 & 0.022645 & 0.20213 \\
           & No WA & 3697 & 0.380032 & 0.477817 & $-$0.000370 & 0.126929 & 1.00000 \\
\addlinespace[1pt]
\(p4\)   & WA    & 3697 & 0.015689 & 0.019680 & 0.998291 & 0.017312 & 0.20629 \\
           & No WA & 3697 & 0.391711 & 0.479993 & $-$0.016825 & 0.105393 & 0.99998 \\
\addlinespace[1pt]
\(p4m\)  & WA    & 3697 & 0.031993 & 0.049409 & 0.989661 & 0.026882 & 0.34761 \\
           & No WA & 3697 & 0.405150 & 0.486931 & $-$0.004129 & 0.105298 & 1.00000 \\
\addlinespace[1pt]
\(p4g\)  & WA    & 3953 & 0.010970 & 0.013299 & 0.998741 & 0.031948 & 0.22845 \\
           & No WA & 3953 & 0.304266 & 0.375349 & $-$0.002736 & 0.143413 & 0.99999 \\
\addlinespace[1pt]
\(p3\)   & WA    & 3697 & 0.010445 & 0.013420 & 0.998990 & 0.013996 & 0.18085 \\
           & No WA & 3697 & 0.328657 & 0.423490 & $-$0.005963 & 0.089013 & 0.99999 \\
\addlinespace[1pt]
\(p3m1\) & WA    & 3697 & 0.014300 & 0.017517 & 0.998098 & 0.018869 & 0.23138 \\
           & No WA & 3697 & 0.311802 & 0.401951 & $-$0.001774 & 0.090252 & 1.00000 \\
\addlinespace[1pt]
\(p31m\) & WA    & 3697 & 0.009437 & 0.013730 & 0.998681 & 0.009836 & 0.14281 \\
           & No WA & 3697 & 0.314269 & 0.378082 & $-$0.000078 & 0.085820 & 1.00000 \\
\addlinespace[1pt]
\(p6\)   & WA    & 3697 & 0.011164 & 0.016016 & 0.998123 & 0.010694 & 0.16201 \\
           & No WA & 3697 & 0.304423 & 0.370006 & $-$0.001695 & 0.083255 & 0.99999 \\
\addlinespace[1pt]
\(p6m\)  & WA    & 3697 & 0.009741 & 0.014438 & 0.998181 & 0.011095 & 0.16585 \\
           & No WA & 3697 & 0.278382 & 0.338782 & $-$0.001671 & 0.083724 & 1.00000 \\
\bottomrule
\end{tabular}
\end{table*}

\begin{figure*}[p]
    \centering
    \includegraphics[height=0.86\textheight,keepaspectratio]{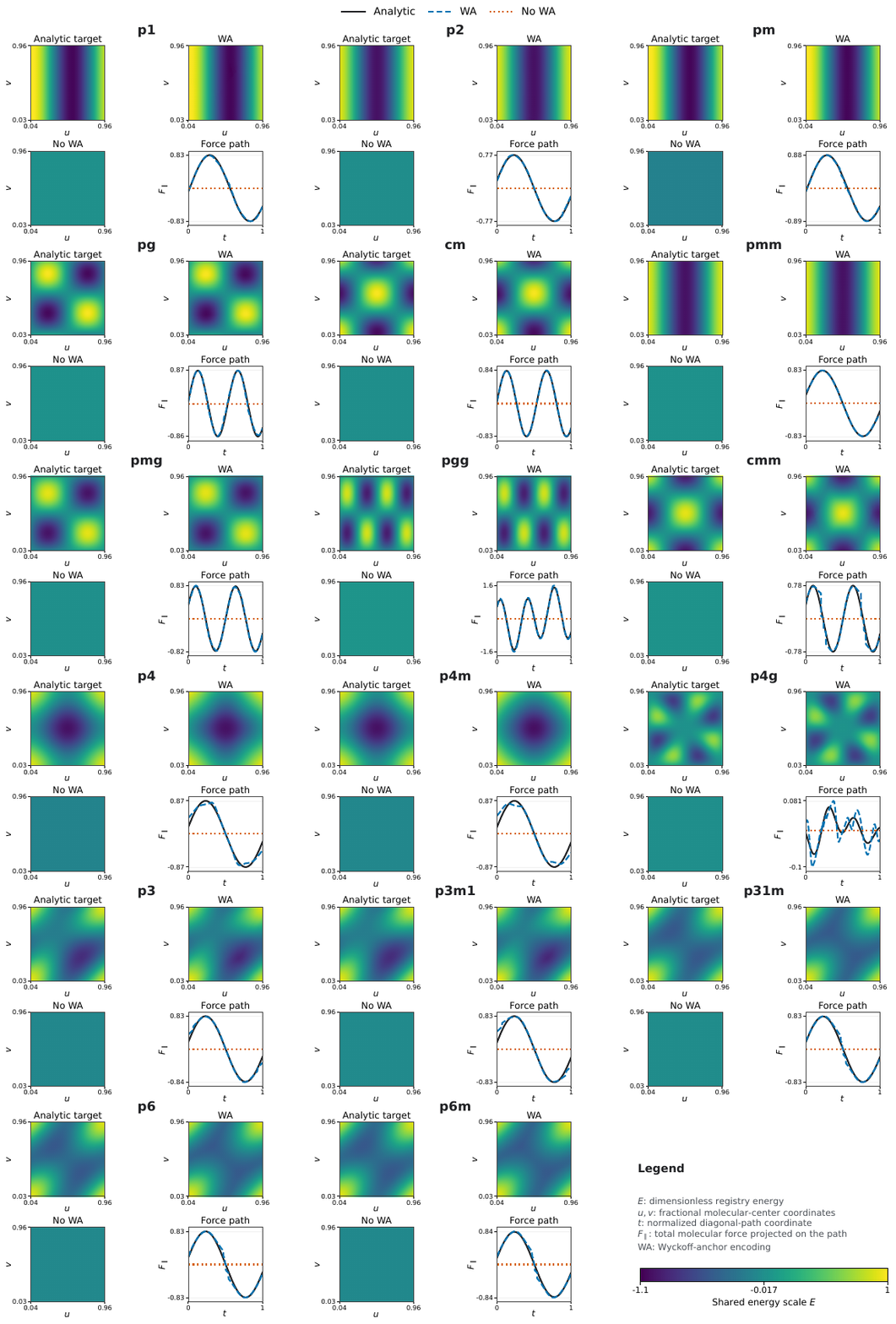}
    \caption{Wyckoff-anchor identifiability results for all 17 wallpaper
    groups. For each group, the panels compare the analytic periodic target,
    the WA-aware \texttt{WELinear} prediction, the parameter-matched WA-free
    prediction, and the force along the indicated reference path. The WA-aware
    model recovers both the registry-dependent scalar field and its gradient,
    whereas the WA-free control collapses to an approximately constant scalar
    prediction and a near-zero force.}
    \label{fig:wa-identifiability-all-groups}
\end{figure*}

\paragraph{Results.}
The numerical results are reported in
Table~\ref{tab:wa-identifiability-errors}, and the corresponding scalar fields
and force profiles are shown in
Figure~\ref{fig:wa-identifiability-all-groups}. The WA-aware model successfully
identifies the registry-dependent target for every wallpaper group. Its test
scalar-target \(R^2\) is at least \(0.9897\) for all 17 groups. The energy RMSE
ranges from \(1.65\times10^{-3}\) to \(4.94\times10^{-2}\), while the relative
force \(L_2\) error ranges from \(2.75\times10^{-2}\) to
\(3.48\times10^{-1}\), depending on the group.

In contrast, the WA-free control exhibits the behavior expected from its exact input collisions. Since all translated configurations provide the same relative graph geometry, its optimal scalar prediction is approximately a constant corresponding to the training-set mean. Its test \(R^2\) is consequently approximately zero or slightly negative. A registry-independent scalar prediction also has approximately zero coordinate gradient, yielding relative force \(L_2\) errors close to one for every wallpaper group.

\paragraph{Interpretation and limitations.}
This experiment establishes an information limitation for graph representations based only on relative coordinates. If a rigid translation leaves every model input unchanged while changing the prediction target, no deterministic model operating on those inputs can represent the target. Finite point-group equivariance alone does not provide the missing translation-registry information.

The WA representation resolves this ambiguity by specifying the graph's position relative to symmetry-defined anchors of the wallpaper group. For the constructed nonsymmorphic cases, this continuous displacement must additionally be supplemented by translation-suborbit membership.

The experiment is intentionally restricted to representational identifiability. It uses one fixed four-node graph, an analytically constructed wallpaper-periodic target, one train/test realization, and one optimization run per wallpaper group. It therefore does not test generalization to new graph geometries or establish predictive performance for a particular physical system. Its purpose is to illustrate, in isolation, that the WA representation supplies registry information that cannot be recovered from the relative geometry of a rigid graph alone.

\section{Synthetic wallpaper-group benchmark}
\label{sec:si_synthetic_dataset}

Although wallpaper symmetries arise in any two-dimensional periodic structure,
crystallographic systems provide one of their most direct physical
applications. The benchmark is deliberately designed as a toy problem: it
removes much of the chemical complexity of realistic materials while retaining
the geometric ingredients needed to study the effect of wallpaper symmetry in a
setting that resembles a surface--adsorbate system. The support has on average
3 atoms per unit cell.

We construct a synthetic atomistic benchmark to isolate the effect of wallpaper-group symmetry in a controlled surface--adsorbate setting. The dataset contains all 17 crystallographic wallpaper groups, with 1,000 independently generated molecular configurations for each group, resulting in 17,000 systems in total. Every atom is carbon. Each system consists of a wallpaper-symmetric support and a small, intentionally non-symmetric carbon fragment containing three, four, or five atoms. 

For each wallpaper group, a single support realization and cell are generated and subsequently held fixed across all 1,000 samples. Molecular geometry, atom count, vertical position, and lateral registry are sampled independently for each example.

\paragraph{Wallpaper-symmetric support generation.}
Support structures are constructed directly from the affine actions of the corresponding wallpaper group. Candidate seed positions are enumerated on rational grids with denominators 6, 8, and 12 and expanded under all group operations. The resulting symmetry-closed orbits are deduplicated before candidate support structures are assembled.

The target support size is three atoms per primitive cell. When the symmetry-orbit multiplicities do not permit exactly three atoms, nearby atom counts are considered. This occurs for \(pg\), \(pmg\), \(pgg\), and \(p4g\), for which the realized support contains four atoms per primitive cell. For the centered groups \(cm\) and \(cmm\), three atoms per primitive cell correspond to six atoms in the displayed conventional cell.

For each candidate support, the lattice metric is selected to produce carbon-like local distances while respecting the lattice constraints imposed by the wallpaper group. Square and hexagonal groups retain their required metric, whereas rectangular and oblique groups are searched over a set of aspect ratios. Candidate periodic neighbor shells are evaluated over translations in \(\{-1,0,1\}^2\). Each support atom is required to have at least two neighbors. The cell is then uniformly rescaled such that the median selected support bond is \(1.43\,\text{\AA}\), and all selected bonds are required to lie between \(1.28\) and \(1.60\,\text{\AA}\). Finally, the explicit \(3\times3\) support must form a connected graph at a \(1.60\,\text{\AA}\) cutoff. All support atoms are placed in a common plane at \(z=-1.0\,\text{\AA}\).

Table~\ref{tab:si_wallpaper_supports} summarizes the realized support geometry for all wallpaper groups.

\begin{table*}[t]
\centering
\caption{
Realized support geometries for the 17 wallpaper groups.
Support counts are given as unit-cell / explicit \(3\times3\)-cluster
counts. The bond interval reports the
periodic support bonds selected during construction. The coordination
histogram is given as \(\{\text{coordination}:\text{number of sites}\}\)
for one conventional cell.
}
\label{tab:si_wallpaper_supports}
\resizebox{\textwidth}{!}{%
\begin{tabular}{llrrrrrll}
\toprule
Group &
Lattice &
\shortstack{Point-group\\order} &
\shortstack{Support atoms\\unit cell / \(3\times3\)} &
\( |\mathbf a| \) [\AA] &
\( |\mathbf b| \) [\AA] &
\(\gamma\) &
\shortstack{Support-bond\\range [\AA]} &
Selected WA family \\
\midrule
\(p1\)   & oblique              & 1  & 3 / 27 & 2.265 & 3.485 & \(90^\circ\)  & 1.336--1.558 & lattice \\
\(p2\)   & oblique              & 2  & 3 / 27 & 3.701 & 1.851 & \(90^\circ\)  & 1.318--1.542 & max-rotation-A \\
\(pm\)   & rectangular          & 2  & 3 / 27 & 2.773 & 2.798 & \(90^\circ\)  & 1.386--1.474 & lattice \\
\(pg\)   & rectangular          & 2  & 4 / 36 & 4.867 & 2.060 & \(90^\circ\)  & 1.311--1.594 & lattice \\
\(cm\)   & centered rectangular & 2  & 3 / 54 & 5.935 & 2.743 & \(90^\circ\)  & 1.372--1.488 & lattice \\
\(pmm\)  & rectangular          & 4  & 3 / 27 & 2.847 & 2.873 & \(90^\circ\)  & 1.423--1.437 & max-rotation-A \\
\(pmg\)  & rectangular          & 4  & 4 / 36 & 2.735 & 3.588 & \(90^\circ\)  & 1.367--1.495 & max-rotation-A \\
\(pgg\)  & rectangular          & 4  & 4 / 36 & 3.887 & 2.550 & \(90^\circ\)  & 1.363--1.430 & max-rotation-A \\
\(cmm\)  & centered rectangular & 4  & 3 / 54 & 3.064 & 5.311 & \(90^\circ\)  & 1.328--1.532 & max-rotation-A \\
\(p4\)   & square               & 4  & 3 / 27 & 2.860 & 2.860 & \(90^\circ\)  & 1.430--1.430 & max-rotation-A \\
\(p4m\)  & square               & 8  & 3 / 27 & 2.860 & 2.860 & \(90^\circ\)  & 1.430--1.430 & max-rotation-A \\
\(p4g\)  & square               & 8  & 4 / 36 & 2.860 & 2.860 & \(90^\circ\)  & 1.430--1.430 & max-rotation \\
\(p3\)   & hexagonal            & 3  & 3 / 27 & 2.748 & 2.748 & \(120^\circ\) & 1.430--1.430 & max-rotation-A \\
\(p3m1\) & hexagonal            & 6  & 3 / 27 & 2.860 & 2.860 & \(120^\circ\) & 1.430--1.430 & max-rotation-A \\
\(p31m\) & hexagonal            & 6  & 3 / 27 & 2.625 & 2.625 & \(120^\circ\) & 1.430--1.430 & max-rotation-B \\
\(p6\)   & hexagonal            & 6  & 3 / 27 & 2.860 & 2.860 & \(120^\circ\) & 1.430--1.430 & max-rotation \\
\(p6m\)  & hexagonal            & 12 & 3 / 27 & 2.860 & 2.860 & \(120^\circ\) & 1.430--1.430 & max-rotation \\
\bottomrule
\end{tabular}%
}
\end{table*}

Although 17 wallpaper groups are represented, there are 15 geometrically distinct realized supports. In particular, \(p3m1\), \(p6\), and \(p6m\) share the same support coordinates and cell metric, but differ in the symmetry operations associated with the structure.

\paragraph{Molecular fragment generation.}
The molecular fragment contains
\begin{equation}
    N_{\mathrm{mol}}\in\{3,4,5\}
\end{equation}
carbon atoms, with the atom count sampled independently for every example. A connected path backbone is generated sequentially in the surface plane. Given the direction of the previous backbone edge, the direction of the next edge is changed by
\begin{equation}
    \Delta\theta \sim
    \mathcal U(0.55,2.45)\ \mathrm{rad},
\end{equation}
with the sign of the turn sampled independently. Thus, the magnitude of the proposed directional change covers approximately
\(31.5^\circ\)--\(140.4^\circ\).
Equivalently, when expressed as the conventional internal angle between
successive backbone bonds, the proposal range is approximately
\(39.6^\circ\)--\(148.5^\circ\).

The projected backbone distance is sampled independently as
\begin{equation}
    d_{i,i+1}^{xy}
    \sim
    \mathcal U(1.38,1.46)\,\text{\AA}.
\end{equation}

The angle and distance intervals above are proposal distributions. A proposed
fragment is retained only if it satisfies additional geometric constraints.
Specifically,
\begin{enumerate}
    \item every pair of molecular atoms must be separated by at least
    \(1.22\,\text{\AA}\) in the generated plane,
    \item the difference between the largest and smallest molecular pair
    distances must exceed \(0.12\,\text{\AA}\), suppressing highly degenerate
    regular point sets,
    \item a radius graph with cutoff \(6.0\,\text{\AA}\) must give every atom
    degree at least two, and
    \item every atom must have two neighbors whose normalized two-dimensional
    cross product exceeds \(0.20\), preventing locally collinear
    configurations.
\end{enumerate}
Up to 2,048 candidate conformations are attempted. After acceptance, the
planar molecular coordinates are translated such that their centroid is at
the origin.

Because these rejection criteria act after sampling, the realized bond-angle
distribution is not exactly uniform over the proposal interval.

\paragraph{Molecular registry and height.}
After generating the internal molecular geometry, the fragment is placed
relative to the support. In fractional coordinates of the conventional cell,
the molecular centroid is sampled uniformly from
\begin{equation}
    (u_c,v_c)
    \sim
    \mathcal U\!\left([0.44,0.56]^2\right).
\end{equation}
The molecule is therefore sampled near the center of the conventional cell. Individual molecular atoms
are not required to remain inside the central cell. In the realized dataset,
7,613 systems (44.78\%) have every molecular atom inside the central cell,
whereas 9,387 systems (55.22\%) have at least one molecular atom outside it.
Thus, centering the molecular centroid near the middle of the cell does not
constrain the complete fragment to a single cell.

Each molecular atom receives an independent vertical coordinate
\begin{equation}
    z_i = 1.80 + \epsilon_i\ \text{\AA},
    \qquad
    \epsilon_i\sim\mathcal U(-0.16,0.16),
\end{equation}
giving
\begin{equation}
    z_i\in[1.64,1.96]\,\text{\AA}.
\end{equation}
Since the support lies at \(z=-1.0\,\text{\AA}\), the purely vertical
molecule--support separation lies between \(2.64\) and \(2.96\,\text{\AA}\)
before accounting for lateral displacement.

The conventional support cell is tiled \(3\times3\), and the molecule is
translated by \(\mathbf a+\mathbf b\) so that its centroid lies above the
central tile.

\paragraph{Realized molecular geometry.}
The realized dataset is consistent with the prescribed proposal distributions
and rejection criteria. Projected backbone distances range from
\(1.380\) to \(1.460\,\text{\AA}\), with a mean and median of
\(1.420\,\text{\AA}\). Independent vertical displacements increase the
corresponding full three-dimensional backbone distances slightly, yielding a
range of \(1.380\)--\(1.492\,\text{\AA}\) and a mean of
\(1.426\,\text{\AA}\).

Considering all molecular atom pairs, 
the realized distances range from \(1.221\) to \(5.407\,\text{\AA}\), with
a median of \(1.473\,\text{\AA}\). Consequently, every pair of atoms in the
C$_3$--C$_5$ fragments lies within the \(6\,\text{\AA}\) molecular graph
cutoff used in the experiments. The resulting radius graphs are therefore
complete graphs \(K_3\), \(K_4\), or \(K_5\).

The closest support atom to each molecular atom lies between
\(2.642\) and \(3.261\,\text{\AA}\), with a mean of
\(2.879\,\text{\AA}\). The absolute vertical difference across molecular
backbone bonds has a mean of \(0.106\,\text{\AA}\), a 95th percentile of
\(0.248\,\text{\AA}\), and a maximum of \(0.318\,\text{\AA}\).

Across all 17,000 samples, the three molecular sizes occur with approximately
equal frequency: 5,627 systems contain C$_3$, 5,764 contain C$_4$, and 5,609
contain C$_5$.

\paragraph{Reference energies and forces.}
Each generated configuration is labeled using GFN2-xTB as implemented in xTB
version 6.7.1. All systems contain only carbon atoms and are treated as
neutral, closed-shell systems. No \texttt{--chrg} argument is supplied, so
xTB uses its default total charge of zero, while \texttt{--uhf 0} explicitly
sets the number of unpaired electrons to zero. The reference energy is the
final total energy parsed from a single-point calculation. The calculation uses the
default electronic temperature of \(300\,\mathrm{K}\), and the resulting
energy is stored in Hartree and converted to electronvolts. The calculation
logs report the default accuracy parameter of 1.0, a limit of 250 SCF
iterations, an SCF convergence threshold of \(10^{-6}\,\mathrm{Hartree}\),
and a wavefunction convergence threshold of \(10^{-3}\,e\).

Atomic forces are obtained from a separate gradient calculation. When available, this
calculation is initialized from the converged restart file produced by the
single-point calculation. The final Cartesian gradient is read in
\(\mathrm{Hartree}/a_0\), and the force on atom \(i\) is defined as the negative of the gradient of the energy.
The conversion to the force units used for learning is $1\,\mathrm{Hartree}/a_0 = 51.42206709\,\mathrm{eV}/\text{\AA}$.

Force labels are generated and retained for every atom, including the support
atoms, although a downstream model evaluation may select only the molecular
forces. Because the energy and gradient are produced by separate xTB
processes, the stored energy is not taken from the process that writes the
gradient. Both calculations nevertheless use the same geometry, method,
charge, spin state, and electronic temperature. For an inspected example,
their energies differ by approximately \(4\times10^{-10}\,\mathrm{Hartree}\),
which is numerically negligible but not bitwise zero.

If the initial single-point calculation fails because the SCF procedure does
not converge, one recovery attempt is made. The electronic structure is first
converged at an electronic temperature of \(1000\,\mathrm{K}\), and the
resulting restart density is then reused for a second single-point calculation
at \(300\,\mathrm{K}\). Only the energy recomputed at \(300\,\mathrm{K}\) is
retained. This recovery procedure was required for 10 of the 17,000
configurations, and all 17,000 configurations have complete reference
results.

The relatively
large forces in the dataset reflect the intentionally
non-equilibrium sampled geometries. The calculations are performed on the explicitly included
finite \(3\times3\) support cluster. Although the serialized extended-XYZ
structures contain a simulation cell and in-plane periodic-boundary metadata
for subsequent graph construction, no periodic lattice option is passed to
xTB.

\begin{figure}[!t]
    \centering
    \includegraphics[width=\textwidth,trim={127bp 0 0 50bp},clip]{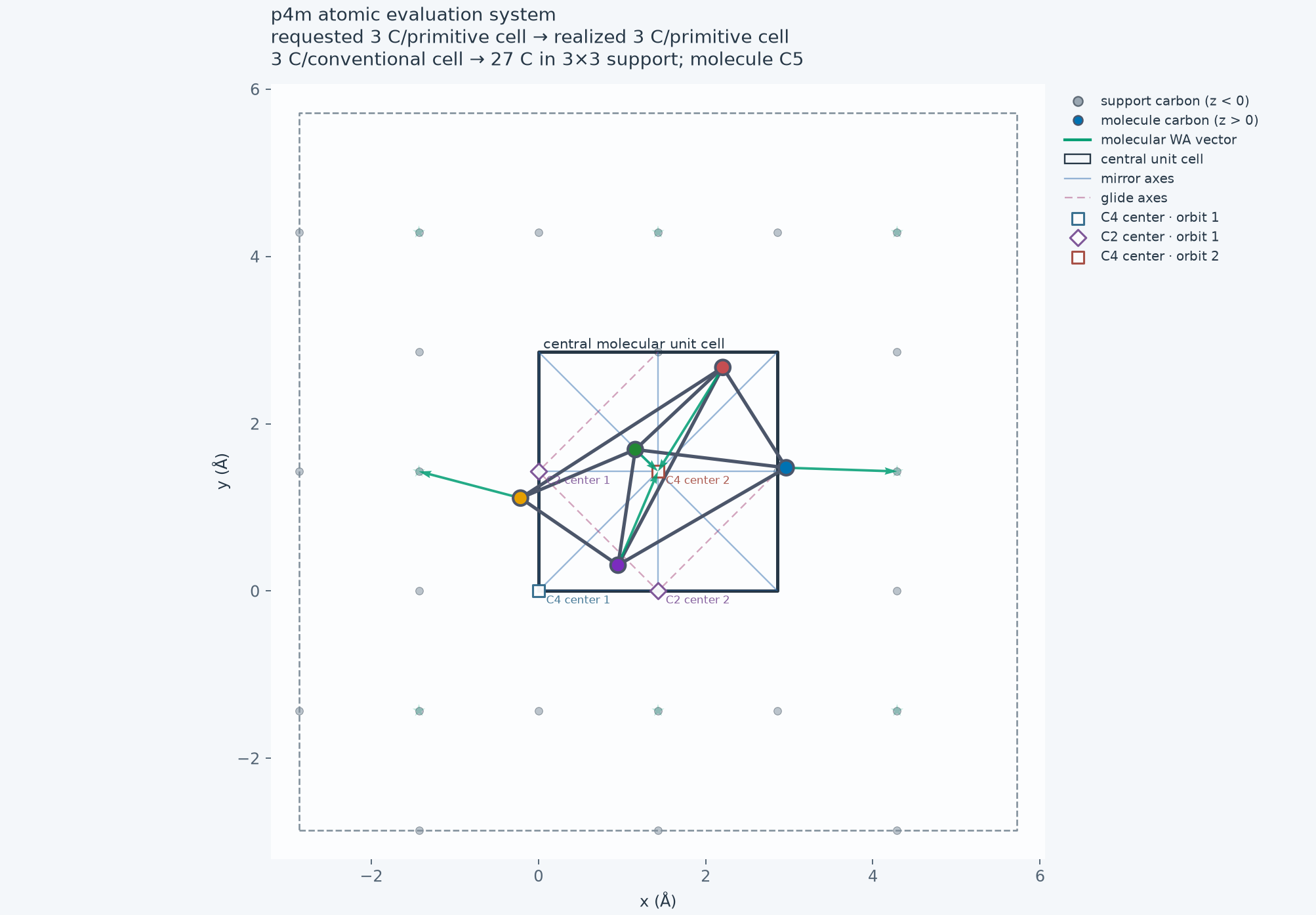}
    \caption{Top view of a synthetic \(p4m\) system with a five-atom carbon
    fragment. Gray points are the support atoms of the finite \(3\times3\)
    cluster, with three atoms per primitive cell, and the dashed line marks the
    cluster boundary. The fragment is placed above the central unit cell,
    shown as a solid square. Colored nodes and dark lines show the molecular
    graph, and green arrows show the molecular WA vectors. Mirror and glide
    axes and the \(C_4\) and \(C_2\) rotation centers of \(p4m\) are also
    shown.}
    \label{fig:toy-top-view}
\end{figure}

\paragraph{Finite-cluster limitation.}
Because the reference labels are computed for this explicit, nonperiodic
\(3\times3\) support cluster rather than for an infinite periodic support, they
are only approximately invariant and equivariant under the wallpaper group.
Symmetry-equivalent placements of the fragment can lie at different distances
from the cluster boundary, so their reference energies and forces can differ
slightly, whereas the WallpaperNet models treat them as equivalent by
construction. We reduce this finite-size effect by placing the molecular
centroid above the central cell of the cluster, as described above. One full
support cell then separates the central cell from the cluster edge in every
in-plane direction, which keeps the fragment away from the boundary.
Figure~\ref{fig:toy-top-view} shows an example five-atom fragment in the
synthetic \(p4m\) environment. The remaining boundary contribution acts as a
residual symmetry-breaking perturbation of the targets.

{\secondreviewercolor
\paragraph{Direct symmetry-consistency audit.}
We quantified this perturbation using configurations already present in the
benchmark. For each wallpaper group and molecular size (3, 4, or 5 atoms), we
selected one configuration and rigidly transformed the molecular fragment by
every symmetry representative used in the unit-cell construction, while the
explicit \(3\times3\) support cluster was left unchanged. A lattice translation
was then applied to the complete fragment to return its center of mass to the
central unit cell.
Relative to the identity calculation, we measured the absolute energy change
per full-system atom, the componentwise MAE between the new molecular forces and
the covariantly transformed identity forces, and their relative \(L_2\)
difference. The group \(p1\) was omitted because it has no nonidentity
representative.

Across the remaining 16 wallpaper groups, the mean discrepancies were
\(1.28\,\mathrm{meV\,atom^{-1}}\) for energy,
\(66.3\,\mathrm{meV\,\mathring{A}^{-1}}\) for force components, and
\(2.53\%\) for relative force \(L_2\). Table~\ref{tab:synthetic-label-symmetry-audit}
compares these values with the model errors. For each group, the lowest error
among the WallpaperNet variants was selected before averaging across groups.
Here, C denotes force-weighted training with energy-derived forces, and V
denotes direct vector-head training.

\begin{table}[!t]
    \secondreviewercolor
    \centering
    \caption{Scale of the finite-cluster symmetry discrepancy relative to the
    lowest WallpaperNet error for the synthetic benchmark, averaged over the
    16 wallpaper groups with nonidentity representatives.}
    \label{tab:synthetic-label-symmetry-audit}
    \small
    \resizebox{\textwidth}{!}{%
    \begin{tabular}{lrrr}
        \toprule
        Metric &
        \shortstack{Mean symmetry\\discrepancy} &
        \shortstack{Mean lowest\\model error} &
        \shortstack{Discrepancy /\\model error} \\
        \midrule
        Energy MAE (\(\mathrm{meV\,atom^{-1}}\))
            & 1.3 & 4.5 & 28.4\% \\
        Force-component MAE, C (\(\mathrm{meV\,\mathring{A}^{-1}}\))
            & 66.3 & 412.4 & 16.0\% \\
        Force-component MAE, V (\(\mathrm{meV\,\mathring{A}^{-1}}\))
            & 66.3 & 250.7 & 26.0\% \\
        Relative force \(L_2\), C (\%)
            & 2.5 & 18.3 & 14.0\% \\
        Relative force \(L_2\), V (\%)
            & 2.5 & 11.9 & 21.0\% \\
        \bottomrule
    \end{tabular}%
    }
\end{table}

The symmetry mismatch is therefore smaller than the lowest model error for
every reported metric: its mean magnitude is approximately 14--28\% of the
corresponding error. The benchmark should consequently be interpreted as
learning from approximately equivariant labels with a small but nonzero
finite-cluster perturbation, rather than from exactly wallpaper-equivariant
quantum-chemical targets.
}

\section{Dataset construction and simulation details}
\label{sec:dataset-construction}

{\explanationcolor
\subsection{Motivation for molecular adsorption on graphene}

Lithium-ion batteries underpin modern portable electronics, stationary energy
storage, and electric mobility. Their widespread use relies on the high
operating voltage and energy density of transition-metal-oxide cathodes paired
with graphite anodes, but the use of critical metals such as Co, Ni, and Mn
creates cost and sustainability concerns. This has motivated electrode
chemistries based on redox-active organic molecules and earth-abundant
elements~\citep{bhattacharyaRelativeStabilityNormal2013,
liangOrganicElectrodeMaterials2012,armandBuildingBetterBatteries2008,
moritaOrganicTailoredBatteries2011}. Organic electrode materials are chemically
tunable, but small redox-active molecules can dissolve in the electrolyte and
often have low intrinsic electronic conductivity, which limits cycle life and
rate capability~\citep{nishideFlexibleBatteries2008,
zhuReviewAdvancedCarbonSupported2015}. Immobilizing these molecules on a
conductive carbon support is therefore a promising strategy for simultaneously
reducing dissolution and improving charge transport.

Graphene is an attractive support because it combines high electrical
conductivity with a large accessible surface area. Experiments and calculations
have shown that graphene--organic composites can improve cycling stability and
charge/discharge performance, and that stronger molecule--graphene binding is
associated with reduced dissolution of the organic component~\citep{
kalinathanAnthraquinoneModifiedCarbon2008,
yuBindingEnergyWork2014,yuDispersioncorrectedDFTStudy2014,
rasheevFundamentalPromiseAnthraquinone2020}. At the molecular scale, adsorption
energy measures the thermodynamic stability of the composite, while the atomic
forces determine its relaxed adsorption geometry and orientation.
Frontier-orbital alignment also affects interfacial electron transfer and the resulting
electrochemical response~\citep{ma2021BatteryTechnology2021,
ehlertMetalFreeMolecularCatalysts2023}. Molecule--graphene interactions are
therefore relevant not only to organic electrodes, but also to sensing,
catalysis, and molecular electronics~\citep{kongMolecularAdsorptionGraphene2014}.

Predicting these interactions across chemical and configurational space is
computationally demanding. Adsorption on graphene is generally governed by weak,
dispersion-dominated physisorption, and the energy depends on molecular
conformation, orientation, adsorption site, and lattice registry~\citep{
kongMolecularAdsorptionGraphene2014,lazarAdsorptionSmallOrganic2013}. Treating
the graphene sheet as a fixed periodic substrate is a common and useful
approximation that substantially reduces this cost~\citep{
chenCrystalShapeControlled2017,londeroDesorptionNalkanesGraphene2012,
cunhaWaterDispersibleFewLayer2018,songInfluenceGrapheneElectronic2024,
carlsonModelingWaterInteractions2024}. It also creates precisely the symmetry
setting studied in this work: configurations related by the graphene
wallpaper group \(p6m\) must have equal energies and symmetry-related forces,
whereas a displacement that changes the molecular registry can change both.

The graphene experiment is consequently both an application-motivated test and
a stringent evaluation of the proposed architecture. The dataset contains
\(15{,}089\) chemically diverse molecule--graphene systems. The underlying
geometry optimizations required 201 steps on average, with a maximum of 714
steps. Because every optimization step requires another energy-and-force
evaluation, a fast learned interatomic potential could substantially reduce the
cost of exploring adsorption configurations across chemical space. WallpaperNet
provides such a route without explicitly representing the much larger substrate
graph in every message-passing step, while retaining the physically relevant
periodic symmetry.
}

\subsection{Motivation for the electronic-structure level}

The graphene adsorption dataset was designed to provide a large and chemically
diverse set of adsorption geometries, energies, and forces at a computational
cost compatible with training an interatomic potential. Because the dataset
contains  thousands of adsorbates and, for each adsorbate, an entire
geometry-optimization trajectory, direct density-functional-theory (DFT)
calculations for all configurations would be prohibitively expensive.
We therefore used the semiempirical GFN2-xTB method for dataset generation.

The suitability of this approximation was assessed on a chemically diverse
benchmark set of 300 organic molecules. The benchmark molecules were selected
from the underlying molecular database using molecular fingerprints and
Tanimoto distances, choosing mutually dissimilar structures in order to span
the available chemical space.
The GFN2-xTB protocol was selected based on comparison with periodic
PBE-D4/DZP calculations~\citep{perdewGeneralizedGradientApproximation1996,caldeweyherGenerallyApplicableAtomiccharge2019}. The simulation protocol was motivated by previous studies of molecular adsorption on graphene and verifed to experimental results.\cite{yuDispersioncorrectedDFTStudy2014,yuBindingEnergyWork2014}
Ordinary least-squares regressions of the GFN2-xTB results on the corresponding
PBE-D4/DZP reference values yielded $R^2=0.849$ and a residual MAE of
$0.058\,\mathrm{eV}$ for the binding energies ($N=300$), and $R^2=0.895$ and a
residual MAE of $0.196\,\mathrm{eV}$ for the unoccupied LUMO-like
adsorbate-projected peak positions ($N=299$). The adsorption energy governs the
stability of the adsorbate--graphene complex, whereas the LUMO frontier energy
governs the propensity for interfacial electron transfer and is therefore
important for molecular reduction, charge-transfer processes, and catalytic
activation~\citep{luderUnderstandingDopingStrategies2017,fedorovUnlockingPotentialPredicting2023,yuDispersioncorrectedDFTStudy2014,yuBindingEnergyWork2014,yuCanAllNitrogendoped2013,medfordSabatierPrinciplePredictive2015,mateoFundamentalsApplicationsPhotothermal2021,liuFrontierMolecularOrbital2020}.

\begin{figure}[h]
    \centering
    \includegraphics[width=0.9\textwidth]{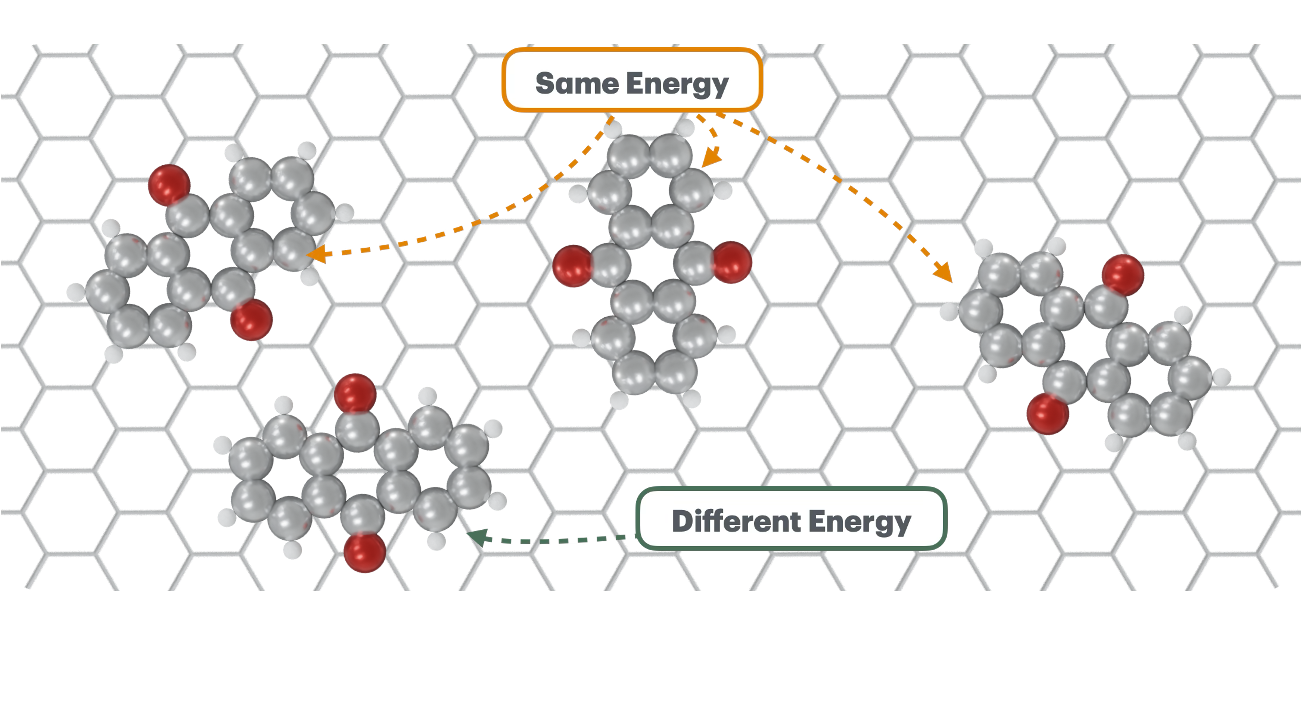}
    \caption{Four different adsorption configurations of a small molecule on a graphene substrate. The first three configurations are related by the action of the \(p6m\) wallpaper group, while the fourth configuration is not. The adsorption energies of the first, second and third configurations are identical, while the fourth configuration has a different adsorption energy.}
    \label{fig:there-graphene-adsorption-example}
\end{figure}

The importance of the molecules relative position to the periodic substrate is illustrated in Figure~\ref{fig:there-graphene-adsorption-example}. Here, three molecules, that are different only by the unitcell translation and rotation have the same (adsorption) energy. While a fourth molecule, which position can not be obtained by a group action of \(p6m\) yield a different adsorption energy.

\subsection{Adsorbate preparation and simulation-cell construction}

The adsorption structures were generated automatically to permit high-throughput
treatment of a chemically diverse molecular dataset. For each molecule,
aromatic-ring planes were identified from sets of three ring atoms. Their
surface-normal vectors were oriented consistently and summed to obtain an
overall molecular normal. The molecule was subsequently rotated such that this
normal was aligned with the surface-normal ($z$) direction. When necessary,
the molecule was additionally reflected with respect to the $xy$ plane to
prevent bulky substituents from initially pointing into the graphene surface.

The lateral dimensions of the graphene supercell were chosen separately for
each adsorbate. A molecular footprint was first constructed with an
approximately $2\,\text{\AA}$ margin around the molecule, and the graphene
unit cell was repeated until a suitable hexagonal supercell was obtained.
The surface coverage was kept approximately comparable between systems by
targeting a ratio of molecular footprint area to graphene area of approximately
0.4. The adsorbate was initially positioned $3\,\text{\AA}$ above the
graphene surface, measured from its lowest atom. Along the surface-normal
direction, at least $15\,\text{\AA}$ of vacuum was included above the
highest adsorbate atom.

\begin{figure}
    \centering
    \includegraphics[width=0.9\textwidth]{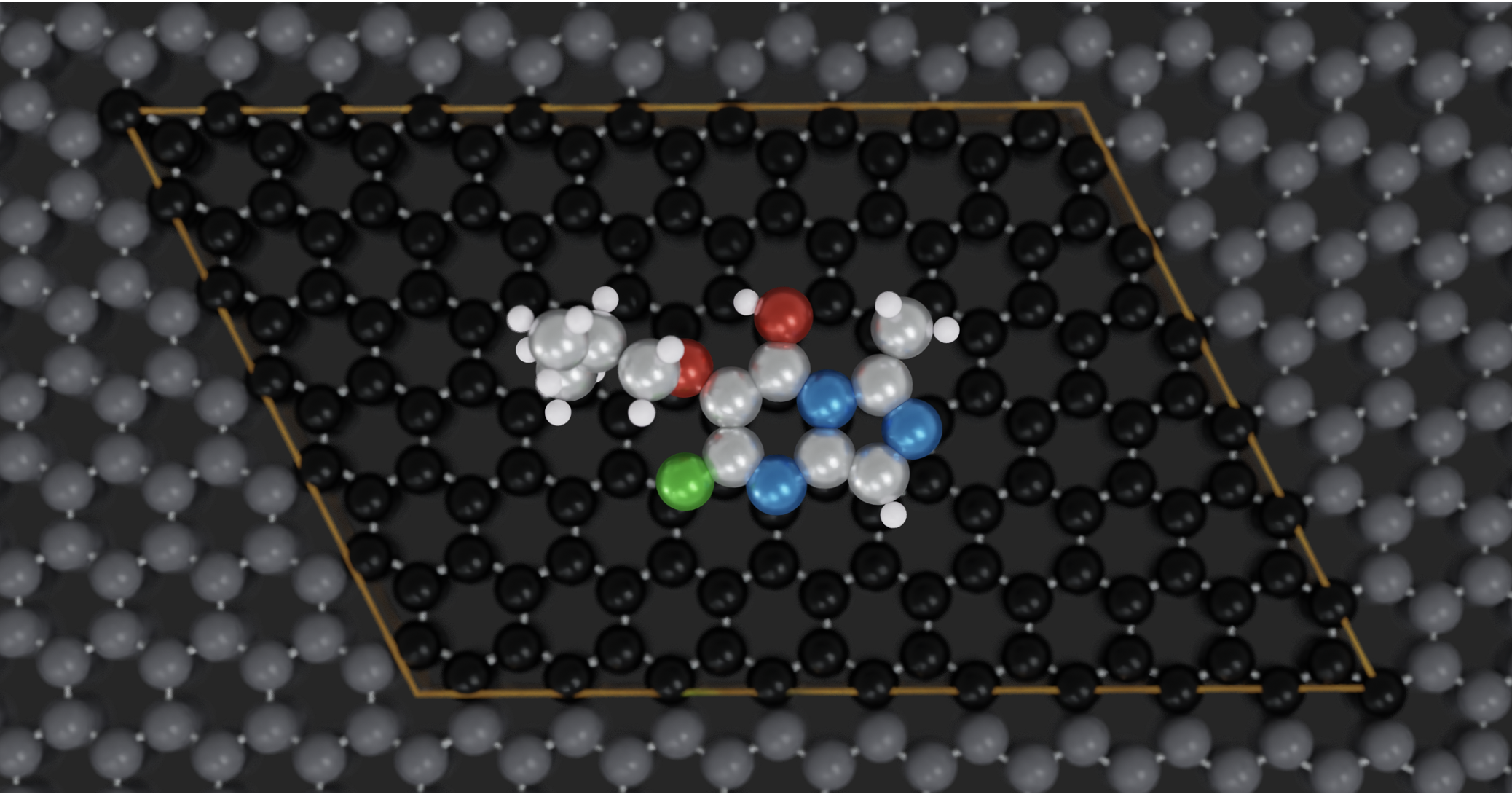}
    \caption{Example of a simulation cell for a small molecule adsorbed on a three-layer graphene substrate.atoms of the graphene substrate are shown in black, while grey atoms are there continuous extension.}
    \label{fig:graphene-adsorption-example}
\end{figure}

The substrate consisted of three graphene layers. The number of layers was
chosen based on a convergence test of the adsorption energy: increasing the
substrate beyond three layers produced only minor changes in the calculated
binding energy. All graphene atoms were kept fixed throughout the subsequent
geometry optimizations and property calculations. Figure \ref{fig:graphene-adsorption-example} shows an example of simulation cell, containing small molecule adsorbed on a three-layer graphene substrate.

\subsection{Validation against periodic DFT}
\label{sec:si_xtb_validation}

The accuracy required from the semiempirical reference level was assessed
using a benchmark set of 300 chemically diverse adsorbates. Rather than
selecting this set randomly, molecular structures were encoded using Morgan
fingerprints and compared using Tanimoto distances. Molecules were selected to
maximize their mutual dissimilarity, providing a benchmark intended to span
the chemical diversity of the full molecular library.

Binding energies were computed using the energies of the complete
graphene-adsorbate system, the graphene substrate, and the isolated
adsorbate in its adsorption geometry, with identical supercell dimensions and
boundary conditions. We use the positive binding-energy convention
\begin{equation}
    E_{\mathrm{B}}
    = E_{\mathrm{graphene}} + E_{\mathrm{adsorbate}}
    - E_{\mathrm{graphene+adsorbate}},
    \label{eq:binding-energy}
\end{equation}
where $E_{\mathrm{graphene+adsorbate}}$ is the energy of the relaxed combined
system, $E_{\mathrm{graphene}}$ is the energy of the isolated graphene
substrate, and $E_{\mathrm{adsorbate}}$ is the energy of the isolated
adsorbate in its adsorption geometry. With this convention, stronger
favorable adsorption corresponds to a larger positive $E_{\mathrm{B}}$.
The original benchmark compared GFN2-xTB against
periodic PBE-D4/DZP calculations. 

As a complementary electronic-structure test, projected densities of states
were calculated for the adsorbate and substrate. A $5\times5\times1$
Monkhorst-Pack mesh was used for these calculations, coarser meshes were
found to spuriously open a gap in graphene. Energies were referenced to the
Fermi energy, and the first maximum of the unoccupied adsorbate-projected
density of states was used as a practical definition of the adsorbate LUMO.

Taken together, the energetic and frontier-state benchmarks indicate that
GFN2-xTB retains the relevant trends of the periodic DFT description at a
fraction of its computational cost. This trade-off is essential for generating training data for many thousands of graphene-adsorbate
systems and was therefore used as the reference level for the machine-learning dataset.

Molecular preprocessing, periodic-cell construction, and geometry handling used ASE and
RDKit~\citep{hjorthlarsenAtomicSimulationEnvironment2017,
greglandrumRdkitRdkit2025_03_22025}. The initial molecular geometries were drawn from the chemical space described by
\citet{fedorovMultiSolventGraphNeural2026}. The adsorbates were relaxed with GFN2-xTB in DFTB+~\citep{hourahineDFTBSoftwarePackage2020}, and the DFT computation was carried out in GPAW~\citep{mortensenGPAWOpenPython2024}.

\subsection{Dataset statistics and force distribution}

For the WallpaperNet models, the graph representation contains only the
molecular adsorbate atoms. The graphene atoms are excluded from message
passing, although their number is retained as metadata. Each system contributes
one graph record. Energies are reported in \si{\electronvolt} per system and
forces in $\mathrm{eV\,\mathring{A}^{-1}}$.

Table~\ref{tab:graphene-dataset-stats} summarizes the complete dataset. All
reported standard deviations are population standard deviations.

\begin{table}[htbp]
    \centering
    \caption{Composition and force-magnitude statistics for the complete
    graphene adsorption dataset.}
    \label{tab:graphene-dataset-stats}
    \small
    \begin{tabular}{lrrrrrr}
        \toprule
        Quantity & $N$ & Mean & SD & Min. & Median & Max. \\
        \midrule
        Molecular atoms/system & $15{,}089$ & $24.926$ & $3.744$ & $8$ & $25$ & $44$ \\
        Graphene atoms/system  & $15{,}089$ & $361.900$ & $50.753$ & $150$ & $378$ & $660$ \\
        Full-cell atoms/system & $15{,}089$ & $386.825$ & $51.593$ & $158$ & $400$ & $699$ \\
        Force norm ($\mathrm{eV\,\mathring{A}^{-1}}$)
            & $376{,}104$ & $0.360$ & $0.290$ & $0.0005$ & $0.271$ & $4.545$ \\
        \bottomrule
    \end{tabular}
\end{table}

The total system energies span from $-40{,}589.4$ to $-9{,}197.05\,\si{\electronvolt}$, with a median of
$-23{,}471.2\,\si{\electronvolt}$. The 5th and 95th percentiles are
$-26{,}726.2$ and $-18{,}242.2\,\si{\electronvolt}$, respectively. The force
norm has a mean of $0.360\,\mathrm{eV\,\mathring{A}^{-1}}$, a median of
$0.271\,\mathrm{eV\,\mathring{A}^{-1}}$, and a 5th--95th percentile interval
of $0.063$--$0.940\,\mathrm{eV\,\mathring{A}^{-1}}$. The mean signed in-plane
components are effectively zero, while the mean surface-normal component is
$-0.00956\,\mathrm{eV\,\mathring{A}^{-1}}$. These force statistics include
one vector per molecular atom and exclude the frozen graphene atoms.

\section{Architectures for real graphene--adsorbate and synthetic experiments}
\label{sec:wallpapernet-architectures_graphene}
\label{architectures-used-in-the-real-system-experiments}

The synthetic experiments evaluate two training
scenarios, whereas the real-system study evaluates three.  The construction is
presented first in the
generic form used by the synthetic (toy) wallpaper-group experiments
and then specialized to the real graphene--adsorbate data.  The WP aliases
correspond to the implementation names as follows:
WP-Inv is \path{egnn_like}, WP-InvEq is
\path{mixed_egnn_like}, WP-Equi is \path{wa_linear}, WP-VecInv
is \path{mix_equivariant_advanced}, WP-Reg is
\path{wa_advanced_sr}, WP-RegTP is \path{wa_linear_mod}, and
WP-O3 is \path{wa_advanced_o3_feedback}.

Seven architectures operate on the adsorbate/molecule graph only. Three
architectures (\path{full_cell_egnn}, \path{full_cell_e3nn_lmax1}, and
\path{full_cell_e3nn_lmax2}) operate on the complete adsorbate--graphene cell.
For every architecture, forces are supervised and evaluated only on adsorbate
atoms. Graphene atoms therefore contribute context in full-cell models, but do
not contribute entries to the force loss.

\begin{sloppypar}
The corresponding toy model names are, in the same order,
\path{molecule_invariant}, \path{molecule_mixed_conservative} (combined mode)
or \path{molecule_mixed} (direct-force mode),
\path{p_molecule_full_equivariant}, \path{wa_advanced},
\path{wa_advanced_sr}, \path{wa_linear_mod},
\path{wa_advanced_o3_feedback}, \path{full_graph_invariant},
\path{full_graph_e3nn_lmax1_tensor_product}, and
\path{full_graph_e3nn_tensor_product}.  The toy and real-system models share
the same architectural blueprint, but are not always algebraically identical.
\end{sloppypar}

\subsection{Common notation and graph construction}\label{common-notation-and-graph-construction}

Node positions are $x_i\in\mathbb{R}^3$. The real experiments use a
cutoff of $6.0$ and at most 128 neighbors per target. For an edge, define

\[
\begin{aligned}
r_{ij} &= x_i-x_j,\\
d_{ij} &= \lVert r_{ij}\rVert,\\
\widehat{r}_{ij} &= r_{ij}/\max(d_{ij},\varepsilon),\\
r_{ij,xy} &= (r_{ij,x},r_{ij,y}),\\
r_{ij,z} &= (r_{ij})_z.
\end{aligned}
\]

Here $j\to i$ denotes a directed edge, $\mathcal N(i)$ is the set of senders
connected to receiver $i$, $\deg(i)=|\mathcal N(i)|$, and
$\varepsilon=10^{-8}$ prevents division by zero.  We embed a planar WA vector
in three dimensions as $\overline{w}_i=(w_{i,x},w_{i,y},0)$.
Indices $i$ and $j$ always label receiver and sender nodes, respectively.
$k$ labels a feature channel, $p$ labels a message path, and $\ell$ labels a
message-passing layer. Whether a symbol denotes a scalar, vector, feature
array, or representation-valued field is specified by its definition and
ambient space.

Each adsorbate node also has a two-dimensional Wyckoff-anchor vector
in the surface plane. To simplify indexing in this section, we write
$w_i$ for the Wyckoff-anchor vector $v_{\mathrm{WA},i}$ defined earlier. It
encodes the node's displacement relative to its associated crystallographic
anchor. WA vectors are used by the molecule-only wallpaper models and are
ignored by the full-cell baselines.

The toy implementation is group-generic.  A node has a single WA vector
$w_i$ and, for nonsymmorphic wallpaper groups,
a learned regular-representation encoding of the translation-coset/member
label.  The real data use the fixed $p6m$ point group ($D_6$ point symmetry),
one two-dimensional WA vector per adsorbate atom, and no nonsymmorphic member
feature.

\paragraph{Initial scalar node features.}\label{initial-scalar-node-features}

In the real-system models, all Wallpaper architectures begin with the same three scalar
node attributes:

\[
q_i = [Z_i,\,m_i,\,z_i],
\]

where $Z_i$ is the atomic number, $m_i$ is the atomic mass, and $z_i$ is the
$z$ coordinate of the node. (It's important to note that for the full-cell architectures, the $z$ coordinate is not used as an additional node feature.) For the full-cell architectures, the $z$ coordinate is not used in the message passing. The full-cell architectures don't receive the $z$ coordinate as input. A learned affine map $L_{\rm in}$ produces the
initial scalar hidden state of width $H$:

\[
h_i^{(0)}=L_{\rm in}(q_i)\in\mathbb R^H.
\]

All models use
four message-passing layers and ReLU in the ordinary scalar MLPs unless a layer
explicitly uses a symmetry-aware SiLU or ELU activation.

The toy models instead use a simplified schema: a linear map of $Z_i$, an MLP
of $z_i$, and, for full-cell models, a linear map of the binary node-type
indicator.

\paragraph{Radial basis.}\label{radial-basis}

Every real-system architecture expands $d_{ij}$ in 16 smooth finite-support
radial basis functions on $[0,6]$ as implemented in \texttt{e3nn}. The basis is multiplied by $\sqrt{16}$. Denoting
this expansion by $B(d_{ij})$, scalar edge networks receive both
$\lVert r_{ij}\rVert^2$ and $B(d_{ij})$, except for the e3nn UVU models,
whose tensor-product weights are generated from $B(d_{ij})$.

The toy models use the same cutoff and 16 channels, but their common
radial map, denoted by $B_{\rm toy}(d_{ij})\in\mathbb R^{16}$, is a Gaussian basis multiplied by a cosine cutoff.

\paragraph{Compact \(D_6\) invariants.}\label{compact-d6-invariants}

For a planar vector $v=(v_x,v_y)$, let
$\zeta(v)=(v_x+\mathrm{i}v_y)/\max(\lVert v\rVert,\varepsilon)$,
where $\mathrm i^2=-1$. The compact $D_6$ feature map used
by the scalar $D_6$ layers is

\[
\phi_{D_6}(v)=\left[\operatorname{Re}(\zeta(v)^6),\,
\operatorname{Re}(\zeta(v)^6),\,\lVert v\rVert\right].
\]

The repeated first component is intentional in the current implementation: a
reflection changes $\operatorname{Im}(\zeta^6)$, so only
$\operatorname{Re}(\zeta^6)$ is invariant under $D_6$. Repeating it
retains a three-channel interface without introducing a reflection-odd scalar.
The clamped denominator gives $\zeta(0)=0$.

For models with joint edge--WA features enabled, the \(D_6\)-invariant contraction

\[
J_{D_6}(r_{ij,xy},w_i)=\left[r_{ij,xy}\!\cdot\!w_i,\,r_{ij,xy}\!\cdot\!w_i\right].
\]

is appended. The duplicate is again intentional and keeps two joint channels.
Unlike the \(C_6\) version, no signed planar cross product is used because it is odd
under reflection.

\paragraph{Aggregation and residual updates.}\label{aggregation-and-residual-updates}

For compactness, define the two aggregators used below:

\[
\mathcal A_i^{\rm sum}(u)=\sum_{j\in\mathcal N(i)}u_{ij},
\qquad
\mathcal A_i^{\rm dn}(u)=
\frac{1}{\sqrt{\max(1,\deg i)}}
\sum_{j\in\mathcal N(i)}u_{ij}.
\]

The scalar and simple equivariant layers use $\mathcal A^{\rm sum}$.  The
advanced regular-state WA layers use $\mathcal A^{\rm dn}$.

Scalar hidden updates use a learned residual projection $W_{\rm res}$ so that input and
output widths may differ. In the real-system matrix all four layers have equal
input and output width $H$.

In later equations, $1$ denotes the trivial representation, $V$ the
two-dimensional planar vector representation, and ${\rm Reg}(G)$ the regular
representation of the finite point group $G$, where $G$ denotes the wallpaper
point group relevant to the current model. The notation
$n\,\rho$ means $n$ copies of representation $\rho$.
$\mathcal L_{\rho\to\rho'}$ denotes a learned equivariant linear map,
$\operatorname{TP}_{\rm UVU}$ a factorized equivariant tensor product,
$\operatorname{InvNorm}$ the vector of smooth norms of non-scalar irrep
fields, and $\sigma_{\rm Reg}$ a representation-compatible nonlinearity.
$\phi_G$ denotes the invariant single-vector feature map for group $G$, and
$J_G$ the corresponding invariant joint feature of an edge and a WA vector.
Brackets denote feature concatenation and $\odot$ denotes channel-wise
multiplication.  Symbols such as $U$, $M$, $E$, $F$, $G$, and $L$ with
subscripts denote independently learned maps unless stated otherwise.
The symbol $S$ in a map subscript denotes the persistent-state representation
of the architecture under discussion. A prime marks the output of one
message-passing layer, and a tilde marks an intermediate quantity.
For \(O(3)\) irreducible representations, $\ell$ is the angular degree and the
suffixes $e$ and $o$ denote even and odd inversion parity, respectively. For
example, $1o$ is a polar vector and $1e$ is an axial vector.

\subsection{Training scenarios}\label{training-scenarios}

\paragraph{Shared conservative formulation.}\label{combined-energy-and-conservative-force-model-combined}

The final scalar node states are sum-pooled over all graph nodes:

\[
g=\sum_ih_i^{(4)}.
\]

The pooled vector is concatenated with ten global quantities.  With $N_{\rm
sup}$ the number of support atoms, $a,b,c\in\mathbb{R}^3$
the cell vectors, and $\operatorname{vec}$ denoting concatenation of
Cartesian components, define

\[
c_{\rm cell}=
[N_{\rm sup},\operatorname{vec}(a),\operatorname{vec}(b),
\operatorname{vec}(c)]\in\mathbb R^{10},\qquad
E_{\rm pred}=E_{\rm ref}+M_{\rm graph}([g,c_{\rm cell}]).
\]

Here $E_{\rm ref}$ is the appropriate sum of atomic reference energies and
$M_{\rm graph}$ is a four-layer MLP with widths
$H+10\to H\to H\to H\to1$.  The toy implementation deliberately matches
this topology: it supplies the synthetic support-atom count and cell, and adds
the reference energy of the complete synthetic system even when message
passing is restricted to the adsorbate.  For molecule-only real-system models,
the prediction instead adds the adsorbate atom-reference energies and the
separately supplied sum of the atom-reference energies of the graphene. For full-cell real models, it
adds the atom-reference energies of all included atoms.  Thus every
conservative run predicts the full system's total energy.

Forces are obtained from the predicted total energy:

\[
F_i^{\rm pred}=-\frac{\partial E_{\mathrm{pred}}}{\partial x_i}.
\]

Training therefore differentiates through this coordinate gradient.  Let
$E$ and $F_i$ denote the reference energy and force, respectively,
and let $\mathcal M$ be the set of adsorbate atoms.  The graph-energy L1 loss,
component-wise force MSE, and weighted objective are
\[
L_E=|E_{\rm pred}-E|,\qquad
L_F=\frac{1}{3|\mathcal M|}\sum_{i\in\mathcal M}
\lVert F_i^{\rm pred}-F_i\rVert_2^2,\qquad
\mathcal L=\lambda_E L_E+\lambda_F L_F.
\]
Here $\lambda_E,\lambda_F\geq0$ are the energy- and force-loss weights.
The force loss is masked to molecule atoms even for full-cell graphs.  Both
conservative scenarios enforce energy conservation by construction, up to
ordinary numerical differentiation error.

Each message-passing layer also returns a direct force proposal because the
same layer class is shared with force-head training. That returned proposal is
discarded in combined mode and is not part of the conservative force
prediction. Internally, some hybrid architectures can still use equivariant
vector intermediates to update their scalar or persistent states.

\paragraph{Toy conservative training.}\label{toy-conservative-training}

The toy experiment has one conservative joint energy--force condition, with
$(\lambda_E,\lambda_F)=(1,1)$.  Its other condition is the direct force head
described below. 

\paragraph{Real-system conservative training.}\label{real-conservative-training}

The real-system experiment has two conservative conditions.  The standard
combined condition uses $(\lambda_E,\lambda_F)=(1,1)$ and is the condition used
to assess prediction of the full total energy.  The
force-weighted conservative condition keeps the identical energy model and
gradient-derived force definition but weights the energy term by a
coefficient determined from the training split, $\lambda_E\approx1.22\times10^{-6}$
with $\lambda_F=1$ (Appendix~\ref{sec:mol-grap-results}). The coefficient
matches the numerical scales of the energy-L1 and force-MSE terms and accounts
for the ratio of adsorbate to full-system atoms, so both terms contribute to
the objective.

\paragraph{Direct force model.}\label{direct-force-model-force_head}

The direct-head scenario, used in both the toy and real-system studies, does
not predict or supervise energy. Each of the four
message-passing layers emits an equivariant node-vector proposal
$f_i^{(\ell)}$, and
the network output is their sum:

\[
F_i^{\rm pred}=\sum_{\ell=1}^{4}f_i^{(\ell)}.
\]

The objective is component-wise MSE over the \(x\), \(y\), and \(z\) components of
adsorbate atoms. Full-cell models still propagate messages through graphene
nodes, but graphene force entries are removed by~\texttt{mol\_mask}~before loss and
metric computation. Direct forces transform according to the representation
implemented by each layer, but are not constrained to be the gradient of a
scalar energy and are therefore not guaranteed to be conservative.

The graph-energy readout remains instantiated by the shared base model but is
not called in this scenario.

\subsection{Registered and loss-connected parameter counts}

Here, registered means every parameter with
\texttt{requires\_grad=True}.  A parameter is counted as \emph{active} for a
training mode when its gradient is not \texttt{None} after backpropagation
from that mode's objective. The models' capacity was selected to place every model near one million registered trainable parameters for the generated real-system configurations and around 150,000 registered trainable parameters for the generated toy-model configurations, to achieve a fair comparison between the models. 

For a conservative model, forces are obtained
as the negative gradient of the energy with respect to the molecular atom positions. The real-system standard
combined loss, $(\lambda_E,\lambda_F)=(1,1)$, and the force-weighted
conservative loss, $(\lambda_E,\lambda_F)\approx(1.22\times10^{-6},1)$, have the same active-parameter count:
their energy coefficient is nonzero and both use the same energy graph.
The direct-head count instead follows only the explicitly predicted force. The parameter count for every model for real system and toy models for a p6m wallpaper group are given in Table~\ref{tab:real_system_parameter_counts} and Table~\ref{tab:toy_model_parameter_counts} respectively.  

\paragraph{Real-system models.}

These counts use the ten generated real-system architectures.  The registered
counts reproduce the values stored in their configurations.  The conservative
column applies to both real conservative scenarios described above. The abbreviations $\ConservativeMode$ and
$\DirectHeadMode$ denote conservative combined training and direct-force-head training,
respectively.

\begingroup
\setlength{\tabcolsep}{3pt}
\scriptsize
\begin{longtable}{@{}
  >{\raggedright\arraybackslash}p{0.16\linewidth}
  >{\raggedright\arraybackslash}p{0.29\linewidth}
  >{\centering\arraybackslash}p{0.09\linewidth}
  >{\centering\arraybackslash}p{0.055\linewidth}
  >{\centering\arraybackslash}p{0.10\linewidth}
  >{\centering\arraybackslash}p{0.10\linewidth}
  >{\centering\arraybackslash}p{0.10\linewidth}@{}}
\caption{Parameter counts for the real-system models.}
\label{tab:real_system_parameter_counts}\\
\toprule
Architecture &
MP implementation &
Scope &
$H$ &
Registered &
Active $\ConservativeMode$ &
Active $\DirectHeadMode$ \\
\midrule
\endfirsthead
\toprule
Architecture &
MP implementation &
Scope &
$H$ &
Registered &
Active $\ConservativeMode$ &
Active $\DirectHeadMode$ \\
\midrule
\endhead
\bottomrule
\endlastfoot
WP-Inv &
\path{D6_invariant} &
molecule only & 144 & 997,929 & 829,729 & 829,592 \\
WP-InvEq &
\path{D6_p_mix_equivariant} &
molecule only & 138 & 996,373 & 839,869 & 822,423 \\
WP-Equi &
\path{D6_p_full_equivariant} &
molecule only & 116 & 992,501 & 668,393 & 930,208 \\
WP-VecInv &
\path{D6_p_mix_equivariant_advanced} &
molecule only & 104 & 1,004,661 & 936,420 & 764,210 \\
WP-Reg &
\path{D6_wa_advanced_sr} &
molecule only & 103 & 999,746 & 663,073 & 636,638 \\
WP-RegTP &
\path{D6_wa_linear_mod} &
molecule only & 94 & 995,897 & 994,933 & 827,733 \\
WP-O3 &
\path{D6_wa_advanced_o3_feedback} &
molecule only & 70 & 994,449 & 836,836 & 816,169 \\
Full cell invariant GNN &
\path{mp_full_invariant} &
full cell & 158 & 991,613 & 990,977 & 789,214 \\
Full cell e3nn $\ell_{\max}=1$ &
\path{mp_full_equivariant_tensor_state_l1_uvu} &
full cell & 120 & 997,429 & 943,729 & 750,732 \\
Full cell e3nn $\ell_{\max}=2$ &
\path{mp_full_equivariant_tensor_state_uvu} &
full cell & 112 & 1,000,457 & 946,177 & 762,424 \\
\end{longtable}
\endgroup

The difference between registered and active parameters is intentional.
Conservative training does not use parameters exclusive to a returned direct
force proposal.  Direct-head training does not call the graph-energy readout.
Parameters shared by message passing, persistent states, and feedback paths
are counted whenever the selected loss remains connected to them.

\paragraph{Toy-model architectures for \(p6m\).}

The toy architecture reproduce the core ideas of the real-system architectures, but with simplified implementations in most cases. The only difference  
is WP-InvExt and real system WP-InvEq, toy
WP-VecEqui and real WP-VecInv have different implementations.  The toy architectures
are group dependent, so a single group must be fixed
before quoting one width and parameter count.  This table uses the complete \(p6m\) configuration matrix, the seed affects data splitting and
initialization but not architecture size. The abbreviations $\ConservativeMode$ and
$\DirectHeadMode$ denote conservative combined training and direct-force-head training,
respectively.

Where the paired configurations use the same implementation, width, or
registered count, that value is shown once.  A slash gives
$\ConservativeMode/\DirectHeadMode$.  In particular, WP-InvExt and full-cell e3nn
$\ell_{\max}=1$ use independently calibrated model variants or widths.

\begingroup
\setlength{\tabcolsep}{3pt}
\scriptsize
\begin{longtable}{@{}
  >{\raggedright\arraybackslash}p{0.13\linewidth}
  >{\raggedright\arraybackslash}p{0.275\linewidth}
  >{\centering\arraybackslash}p{0.085\linewidth}
  >{\centering\arraybackslash}p{0.08\linewidth}
  >{\centering\arraybackslash}p{0.13\linewidth}
  >{\centering\arraybackslash}p{0.09\linewidth}
  >{\centering\arraybackslash}p{0.09\linewidth}@{}}
\caption{Parameter counts for the \(p6m\) toy models.}
\label{tab:toy_model_parameter_counts}\\
\toprule
Architecture &
Toy implementation ($\ConservativeMode/\DirectHeadMode$ where different) &
Scope &
$H$ ($\ConservativeMode/\DirectHeadMode$) &
Registered ($\ConservativeMode/\DirectHeadMode$) &
Active $\ConservativeMode$ &
Active $\DirectHeadMode$ \\
\midrule
\endfirsthead
\toprule
Architecture &
Toy implementation ($\ConservativeMode/\DirectHeadMode$ where different) &
Scope &
$H$ ($\ConservativeMode/\DirectHeadMode$) &
Registered ($\ConservativeMode/\DirectHeadMode$) &
Active $\ConservativeMode$ &
Active $\DirectHeadMode$ \\
\midrule
\endhead
\bottomrule
\endlastfoot
WP-Inv &
\path{molecule_invariant} &
molecule only & 58 & 149,997 & 149,525 & 122,098 \\
WP-InvExt &
\path{molecule_mixed_conservative}/\path{molecule_mixed} &
molecule only & 40/58 & 151,925/150,233 & 148,644 & 122,334 \\
WP-Equi &
\path{p_molecule_full_equivariant} &
molecule only & 46 & 149,413 & 123,649 & 130,598 \\
WP-VecEqui &
\path{wa_advanced} &
molecule only & 50 & 143,885 & 143,028 & 120,902 \\
WP-Reg &
\path{wa_advanced_sr} &
molecule only & 27 & 149,335 & 140,347 & 139,706 \\
WP-RegTP &
\path{wa_linear_mod} &
molecule only & 44 & 149,917 & 149,733 & 113,707 \\
WP-O3 &
\path{wa_advanced_o3_feedback} &
molecule only & 22 & 148,957 & 148,655 & 143,297 \\
Full cell invariant GNN &
\path{full_graph_invariant} &
full cell & 58 & 148,137 & 147,901 & 120,238 \\
Full cell e3nn $\ell_{\max}=1$ &
\path{full_graph_e3nn_lmax1_tensor_product} &
full cell & 20/29 & 152,765/195,359 & 150,861 & 184,895 \\
Full cell e3nn $\ell_{\max}=2$ &
\path{full_graph_e3nn_tensor_product} &
full cell & 22 & 144,109 & 142,620 & 137,563 \\
\end{longtable}
\endgroup

\subsection{Architecture descriptions}\label{architecture-descriptions}

\paragraph{Overview.}
The toy benchmark uses seven WallpaperNet architectures of increasing
complexity. WP-Inv is a distance-based model with angular invariants and an
EGNN-style vector head for force prediction. WP-InvExt introduces more
geometry features while remaining scalar only: it extends invariant scalar
messages with covariant edge and Wyckoff-anchor directions, with an additional
vector state used during the conservative training objective. WP-Equi replaces
invariant message construction with a nonlinear wallpaper-group-equivariant
edge network whose hidden fields transform as regular representations.
WP-VecEqui couples edge and Wyckoff-anchor regular fields through a single
explicit tensor product and uses the resulting field to update a persistent
regular-representation node state. WP-Reg extends this approach through
separately gated regular-representation paths, including receiver- and
sender-anchor fields and state-conditioned tensor-product interactions.
WP-RegTP explicitly couples direct, edge, and Wyckoff-anchor regular fields
through successive tensor products. WP-O3 combines three-dimensional $O(3)$
tensor features with wallpaper-group-equivariant processing through
bidirectional feedback. The full-cell invariant GNN applies invariant
EGNN-style message passing to the complete molecule--support graph. The
full-cell e3nn $\ell_{\max}=1$ model uses $O(3)$-equivariant tensor products
with scalar and vector features, and the $\ell_{\max}=2$ model additionally
propagates rank-two features to capture quadrupolar angular structure.

For the real-system experiments, we reuse the WP-Inv, WP-Equi, WP-Reg,
WP-RegTP, and WP-O3 designs. Because WP-InvExt outperformed most explicitly
equivariant toy models despite relying primarily on scalar message passing, its
design choices are transferred into two real-system variants. WP-InvEq keeps
essentially the WP-Inv invariant core but sends the scalar and force outputs
through explicit \(D_6\)-equivariant nonlinear regular-representation
post-maps and has no persistent non-scalar state. WP-VecInv instead keeps
invariant scalar messages while introducing persistent polar vectors whose
invariant contractions influence subsequent scalar messages. Its direct force
is formed from edge/WA covariant bases and then passed through a
\(D_6\)-equivariant output network. The toy WP-VecEqui architecture is
omitted because it was the least accurate toy model and substantially overlaps
with the more effective WP-Reg and WP-RegTP designs.

\paragraph{WP-Inv: scalar \(D_6\)-invariant molecule model.}\label{wp-inv-scalar-d6-invariant-molecule-model}

This is the simplest molecule-only reference. It follows the scalar-state
pattern of an EGNN while augmenting the scalar edge message with \(D_6\) invariants
and WA information. For each edge, define the invariant input and message

\[
\begin{aligned}
u_{ij}^{D_6}={}&[h_i,h_j,
\lVert r_{ij}\rVert^2,B(d_{ij}),
\phi_{D_6}(w_i),\phi_{D_6}(r_{ij,xy}),r_{ij,z},
J_{D_6}(r_{ij,xy},w_i)],\\
m_{ij}={}&M_{\rm inv}^{D_6}(u_{ij}^{D_6})\in\mathbb R^H.
\end{aligned}
\]

The joint term is enabled in the real-system configuration. The message MLP
has three affine layers of width \(H\) with two ReLUs. Messages are summed at the
receiver, and a second three-layer MLP updates the scalar state:

\[
h_i'=W_{\mathrm{res}}h_i+
U_{\mathrm{node}}\!\left([h_i,
\mathcal A_i^{\rm sum}(m)]\right).
\]

There is no persistent non-scalar hidden state. The direct planar force is an
EGNN-like scalar-weighted edge vector, while the vertical component is an
independent scalar MLP:

\[
\begin{aligned}
f_{ij,xy}&=\alpha(m_{ij})r_{ij,xy},\\
f_{ij,z}&=\beta(m_{ij}),\\
f_i&=\mathcal A_i^{\rm sum}
([f_{ij,xy},f_{ij,z}]).
\end{aligned}
\]

Here $M_{\rm inv}^{D_6}$ and $U_{\rm node}$ are scalar MLPs, while
$\alpha:\mathbb R^H\to\mathbb R$ and
$\beta:\mathbb R^H\to\mathbb R$ are learned scalar readouts.

The scalar stream is \(D_6\) invariant, and multiplication of the planar vector by
an invariant coefficient gives a \(D_6\)-covariant planar output. The \(z\) component
is a scalar under in-plane \(D_6\) operations.

The toy \path{molecule_invariant} uses the same
message-sum-node-update template, with $\phi_G$ and $J_G$ constructed for
the selected wallpaper group. It additionally can incorporate a nonsymmorphic member encoding.

\paragraph{WP-InvExt: invariant messages extended by covariant WA bases.}
\label{wp-invext-toy}

This is a toy-only architecture that uses \path{molecule_mixed_conservative} for conservative training
and \path{molecule_mixed} for direct-force training. The name
indicates that an invariant scalar core is extended by covariant edge and
Wyckoff-anchor directions. The core idea of this implementation was for the model to stay essentially invariant, but use more provided geometric information. Since the conservative training does not use a force head constructed from the edge and WA directions, we added a persistent vector-state hidden state for the node.

For a selected wallpaper group $G$, both variants form
\[
\begin{aligned}
u_{ij}^{G}={}&[
  h_i,h_j,
  \lVert r_{ij}\rVert^2,
  B_{\rm toy}(d_{ij}),
  \phi_G(r_{ij,xy}),
  \phi_G(w_i),
  r_{ij,z},
  J_G(r_{ij,xy},w_i),
  \rho_i
],\\
m_{ij}={}&M_{\rm inv}^{G}(u_{ij}^{G})\in\mathbb R^H,
\end{aligned}
\]
where $\rho_i$ is present only for nonsymmorphic groups and encodes
the selected translation-suborbit member. The direct \path{molecule_mixed} variant has only scalar node states:
\[
h_i'
=W_{\rm res}h_i+
U_{\rm inv}\!\left(
  [h_i,\mathcal A_i^{\rm sum}(m)]
\right).
\]
Its planar force spans two covariant bases rather than only the molecular edge:
\[
\begin{aligned}
f_{ij,xy}
 &=a_{ij}^{(r)}r_{ij,xy}
  +a_{ij}^{(w)}w_i,\\
f_{ij,z}&=b_{ij},\\
f_i&=\mathcal A_i^{\rm sum}
  ([f_{ij,xy},f_{ij,z}]),
\end{aligned}
\qquad
[a_{ij}^{(r)},a_{ij}^{(w)},b_{ij}]
=F_{\rm inv}(m_{ij}).
\]
Invariant coefficients multiplying covariant bases give a $G$-equivariant
planar force without introducing a regular-representation hidden field. The conservative \path{molecule_mixed_conservative} variant adds
$H$ persistent polar three-vectors
$V_i\in\mathbb R^{H\times3}$.  For every channel $k$, it constructs
\[
\begin{aligned}
c_{ij,k}=[
&\lVert V_{i,k}\rVert^2,
\lVert V_{j,k}\rVert^2,
V_{i,k}\!\cdot\!V_{j,k},
V_{i,k}\!\cdot\!\widehat{r}_{ij},
V_{j,k}\!\cdot\!\widehat{r}_{ij},\\
&V_{i,k}\!\cdot\!\overline{w}_i,
V_{j,k}\!\cdot\!\overline{w}_i
],
\qquad
s_{ij}=S_{\rm inv}(\{c_{ij,k}\}_{k=1}^{H})\in\mathbb R^H.
\end{aligned}
\]
All entries of $c_{ij,k}$ are invariant scalars.  The scalar node
update is
\[
h_i'
=W_{\rm res}h_i+
U_{\rm mix}\!\left(
[h_i,
  \mathcal A_i^{\rm sum}(m),
  \mathcal A_i^{\rm sum}(s)]
\right).
\]
Here $S_{\rm inv}$ is an invariant MLP that maps the complete set of
channel-wise contractions to the length-$H$ message $s_{ij}$.
An invariant coefficient map $G_V$ produces
$[(a_{ij,k})_{k=1}^H,(b_{ij,k})_{k=1}^H]=G_V(m_{ij})$.  These
coefficients update the polar memory as
\[
V_{i,k}'
=V_{i,k}
+\gamma\sum_{j\in\mathcal N(i)}
\left[
a_{ij,k}r_{ij}
+b_{ij,k}\overline{w}_i
\right].
\]
Here $\gamma$ is a learned scalar gain.
This variant returns a zero direct-force proposal.  Its force is defined only
by differentiating the scalar energy.
Since this design is similar to WP-Inv and but it's design improved the toy
results notably. We therefore used its invariant-core idea in the real system experiments, resulting in the
WP-InvEq and WP-VecInv architectures, together with
equivariant output maps.

\paragraph{WP-InvEq: invariant core with \(D_6\)-equivariant output maps.}\label{wp-inveq-invariant-core-with-d6-equivariant-output-maps}

This model uses the same compact invariant edge message and scalar node update
as~WP-Inv, including the enabled joint edge--WA dot features. Its defining
difference is the addition of symmetry-aware \texttt{we3nn} maps after the ordinary
scalar/EGNN computation.

Writing the invariant message as $m_{ij}$, the provisional update and force
are
\[
\widetilde{h}_i=W_{\rm res}h_i+
 U\!\left([h_i,\mathcal A_i^{\rm sum}(m)]\right),\qquad
\widetilde{f}_i=\mathcal A_i^{\rm sum}
\left([\alpha(m_{ij})r_{ij,xy},\,
\beta(m_{ij})]\right).
\]
The \(H\) scalar channels are interpreted as \(H\) copies of the \(D_6\) trivial
representation. They are mapped to copies of the \(D_6\) regular representation,
passed through a representation-compatible ReLU, and mapped back to trivial
scalars. Likewise, the provisional force $[f_x,f_y,f_z]$ is interpreted as the
direct sum of the planar standard representation and one trivial \(z\) scalar,
mapped through a regular-representation hidden field, activated, and mapped
back to vector plus scalar output.
\[
\begin{aligned}
h_i'&=\mathcal L_{n_h{\rm Reg}(D_6)\to1^H}
 \!\left(\sigma_{\rm Reg}\!\left(
 \mathcal L_{1^H\to n_h{\rm Reg}(D_6)}
 (\widetilde{h}_i)\right)\right),\\
f_i&=\mathcal L_{n_f{\rm Reg}(D_6)\to(V\oplus1)}
 \!\left(\sigma_{\rm Reg}\!\left(
 \mathcal L_{(V\oplus1)\to n_f{\rm Reg}(D_6)}
 (\widetilde{f}_i)\right)\right).
\end{aligned}
\]

The integers $n_h$ and $n_f$ are the hidden regular-field multiplicities of
the scalar and force post-maps, respectively.

Consequently, most feature extraction remains an inexpensive invariant MLP,
while the final scalar and vector transformations are explicitly \(D_6\)
equivariant. No non-scalar state persists from one layer to the next.

\paragraph{WP-VecInv: invariant messages with persistent polar vectors.}
\label{wp-vecinv-real}

WP-VecInv is the alias for the
\path{D6_p_mix_equivariant_advanced} model. It uses a persistent
bank of polar vectors, while operating on scalar coefficients.

Its ordinary invariant message is
\[
m_{ij}=M_{\rm inv}(u_{ij}^{D_6}),
\]
with the same $D_6$ input $u_{ij}^{D_6}$ defined for WP-Inv.  Let
$V_i\in\mathbb R^{H\times3}$ be the persistent vector state.  The
second message uses the seven contractions
\[
\begin{aligned}
c_{ij,k}=[
&\lVert V_{i,k}\rVert^2,
\lVert V_{j,k}\rVert^2,
V_{i,k}\!\cdot\!V_{j,k},
V_{i,k}\!\cdot\!\widehat{r}_{ij},
V_{j,k}\!\cdot\!\widehat{r}_{ij},\\
&V_{i,k}\!\cdot\!\overline{w}_i,
V_{j,k}\!\cdot\!\overline{w}_i
],
\qquad
s_{ij}=S_{\rm inv}(\{c_{ij,k}\}_{k=1}^{H})\in\mathbb R^H.
\end{aligned}
\]
Message aggregation and the scalar node update are
\[
h_i'
=W_{\rm res}h_i+
U_{\rm mix}\!\left(
[h_i,
  \mathcal A_i^{\rm sum}(m),
  \mathcal A_i^{\rm sum}(s)]
\right).
\]
An invariant map $G_V$ produces the two length-$H$ coefficient vectors
$[(a_{ij,k})_{k=1}^H,(b_{ij,k})_{k=1}^H]=G_V(m_{ij})$.  The
persistent vector update is
\[
V_{i,k}'
=V_{i,k}
+\gamma\sum_{j\in\mathcal N(i)}
\left[
a_{ij,k}r_{ij}
+b_{ij,k}\overline{w}_i
\right],
\qquad \gamma(0)=0.1.
\]
Here $\gamma$ is a learned scalar gain initialized to $0.1$. Because
$a_{ij,k}$ and $b_{ij,k}$ are invariant scalars, this update is
covariant and preserves directional information between message-passing
layers.

For the direct head, the learned invariant readout
$F_{\rm inv}(m_{ij})=[a_{ij}^{(r)},a_{ij}^{(w)},
\beta_{ij}]$ forms the provisional output
\[
\widetilde{f}_i
=\mathcal A_i^{\rm sum}\!\left(
[
a_{ij}^{(r)}r_{ij,xy}
+a_{ij}^{(w)}w_i,
 \ \beta_{ij}
]
\right).
\]
A final real-system \texttt{we3nn} map enforces the typed $D_6$ output:
\[
f_i
=\mathcal L_{{\rm Reg}\to(V\oplus1)}
\!\left[
\sigma_{\rm Reg}\!\left(
\mathcal L_{(V\oplus1)\to{\rm Reg}}
(\widetilde{f}_i)
\right)
\right].
\]
In conservative training, $f_i$ is discarded but the vector state
remains active because its invariant contractions modify $h_i'$ and
hence the energy.  In direct training, the force proposals are summed across
layers.  Unlike WP-InvEq, WP-VecInv does not pass the scalar node output
through a regular-representation post-map.

\paragraph{WP-VecEqui: persistent regular fields with WA--edge tensor products.}
\label{wp-vecequi-toy}

WP-VecEqui  is the toy only architecture, denoted as \path{wa_advanced} model.  Its persistent
state is a finite-group regular
field
\[
S_i\in n_S\,{\rm Reg}(G),
\]
and its edge computation remains equivariant before being projected to scalar
or vector outputs. Here $n_S$ is the persistent-state multiplicity.
The WA encoder receives the  WA vector, its circular harmonics, its
squared magnitude, and, for nonsymmorphic groups, the regular member encoding:
\[
W_{ij}=E_{\rm WA}^{G}
\left(w_i,\mathcal Y_G(w_i),
\lVert w_i\rVert^2,\rho_i\right)
\in n_R\,{\rm Reg}(G).
\]
Here $\mathcal Y_G$ is the group-adapted circular-harmonic feature map,
$E_{\rm WA}^G$ is a learned equivariant encoder, and $n_R$ is the regular-field
multiplicity used by the edge paths.
The edge encoder receives typed planar geometry together with invariant
chemistry/radial/vertical features and the receiver/sender persistent states:
\[
R_{ij}=E_{\rm edge}^{G}
\left(
h_i,h_j,
r_{ij,xy},\mathcal Y_G(r_{ij,xy}),
\lVert r_{ij}\rVert^2,B_{\rm toy}(d_{ij}),r_{ij,z},
S_i,S_j
\right)
\in n_R\,{\rm Reg}(G).
\]
The map $E_{\rm edge}^G$ is the corresponding learned equivariant edge
encoder.
The two regular fields interact through an explicit UVU tensor product:
\[
C_{ij}
=\sigma_{\rm Reg}\!\left(
\operatorname{TP}_{\rm UVU}(R_{ij},W_{ij})
\right)
\in n_R\,{\rm Reg}(G).
\]

Three equivariant paths contribute to the scalar edge message:
\[
z_{ij}
=\mathcal L_{W\to1^{H}}(W_{ij})
+\mathcal L_{R\to1^{H}}(R_{ij})
+\gamma_C\mathcal L_{C\to1^{H}}(C_{ij}).
\]
The field $z_{ij}\in\mathbb R^H$ is the scalar edge message. The
subscripts $W$, $R$, and $C$ identify independent projections of the WA,
edge, and coupled fields.  The scalar $\gamma_C$ is a learned coupling gain.
Degree-normalized aggregation is followed by an ordinary scalar node update:
\[
h_i'
=W_{\rm res}h_i+
U\!\left(
[h_i,\mathcal A_i^{\rm dn}(z)]
\right).
\]

The persistent regular state receives a residual equivariant update:
\[
\begin{aligned}
\bar R_i
 &=\mathcal A_i^{\rm dn}
 \left(W_{ij}+R_{ij}+\gamma_C C_{ij}\right),\\
S_i'
 &=S_i+
 \eta_S\,
 \sigma_S\!\left(
 \mathcal L_{{\rm Reg}\to S}(\bar R_i)
 \right),
\qquad \eta_S(0)=0.1.
\end{aligned}
\]
Here $\eta_S$ is a learned residual-state gain initialized to $0.1$, and
$\sigma_S$ is the representation-compatible state activation.
Finally, regular-to-vector and regular-to-trivial projections produce the
direct force:
\[
\begin{aligned}
f_{ij,xy}
={}&
\mathcal L_{W\to V}(W_{ij})
+\mathcal L_{R\to V}(R_{ij})
+\gamma_F\mathcal L_{C\to V}(C_{ij}),\\
f_{ij,z}
={}&
\mathcal L_{W\to1}(W_{ij})
+\mathcal L_{R\to1}(R_{ij})
+\gamma_F\mathcal L_{C\to1}(C_{ij}),\\
f_i
={}&\mathcal A_i^{\rm dn}\!\left(
[f_{ij,xy},f_{ij,z}]
\right).
\end{aligned}
\]

The scalar $\gamma_F$ is a learned force-coupling gain.

This architecture shares many design choices with WP-Reg and
WP-RegTP, while performing consistently worse, so we did not use it in the real-system experiments.

\paragraph{WP-Equi: fully \(D_6\)-equivariant edge-message model.}\label{wp-equi-fully-d6-equivariant-edge-message-model}

WP-Equi~moves the edge computation itself into \texttt{we3nn}. Its input field is a
direct sum of:

\begin{itemize}
\item
  trivial representations for $h_i$, $h_j$,
  $\lVert r_{ij}\rVert^2$,
  the 16 radial features, $r_{ij,z}$, and the two enabled joint edge--WA
  contractions.
\item
  one $D_6$ standard representation for $w_i$.
\item
  one $D_6$ standard representation for $r_{ij,xy}$.
\end{itemize}

This typed edge field passes through four \(D_6\)-equivariant linear maps. The first
three are followed by equivariant ReLU activations, and the hidden fields use
copies of the \(D_6\) regular representation. Separate equivariant projections
produce $H$ trivial scalar message channels, one planar standard vector, and one
trivial \(z\) scalar. Scalar messages are aggregated with a residual scalar node
projection, vector outputs are summed at the receiver.

More explicitly, if $q_{ij}$ is the typed direct sum listed above,
\[
\begin{aligned}
R_{ij}
 &=\mathcal M_{D_6}(q_{ij})\in n_{\rm reg}{\rm Reg}(D_6),\\
m_{ij}&=\mathcal L_{{\rm Reg}\to1^{H}}(R_{ij}),\\
h_i'&=W_{\rm res}h_i+
 \sum_{j\in\mathcal N(i)}m_{ij},\\
f_i&=\mathcal A_i^{\rm sum}\!\left(
 [\mathcal L_{{\rm Reg}\to V}(R_{ij}),
  \mathcal L_{{\rm Reg}\to1}(R_{ij})]\right).
\end{aligned}
\]
Here $\mathcal M_{D_6}$ denotes the four-layer equivariant edge network and
$n_{\rm reg}$ its hidden regular-representation multiplicity. In
contrast to WP-Inv, there is no separate nonlinear receiver MLP:
the nonlinear computation occurs before aggregation in the typed edge
network.
Unlike the advanced WA models,~WP-Equi~has no persistent regular or tensor
state. Its advantage over~WP-InvEq~is that WA and planar edge vectors
enter a fully typed \(D_6\)-equivariant message network rather than first being
reduced to compact scalar invariants. Toy \path{p_molecule_full_equivariant} applies the
same idea for an arbitrary wallpaper point group, with all WA/member fields
and periodic geometry.  Its scalar receiver update additionally uses a
nonlinear node MLP,
$h_i'=W_{\rm res}h_i+
U([h_i,\mathcal A_i^{\rm sum}(m)])$, whereas the real $D_6$
implementation adds the projected scalar message directly as written above.
The input and radial featurization also differ as described in the
initial-feature and radial-basis paragraphs above.

\paragraph{WP-Reg: regular-representation WA state with gated paths.}\label{wp-reg-regular-representation-wa-state-with-gated-paths}

WP-Reg~uses explicit \(D_6\) regular representations for edge messages
and for a persistent node state
$S_i\in n_S{\rm Reg}(D_6)$, where $n_S$ is its multiplicity.

Circular harmonics of planar edge and WA vectors are evaluated using a smooth
zero-vector prescription. Nontrivial angular components are multiplied by
$\lVert v\rVert^2/(\lVert v\rVert^2+10^{-4})$, which suppresses
singular angular derivatives near the
origin. Separate equivariant encoders map receiver WA, sender WA, and edge
features to regular fields. UVU tensor products form receiver-edge,
sender-edge, and state-conditioned couplings.

Six regular-message paths are retained: receiver WA, sender WA, encoded edge,
lifted state, receiver coupling, and sender/state coupling. An invariant MLP
produces channel-wise sigmoid gates for these paths. The gates are scaled by two
and initialized to one. The gated paths are summed with a $1/\sqrt{6}$ normalization.

In equations, let
\[
\begin{aligned}
R_i&=E_{\rm WA}(w_i),\quad
R_j=E_{\rm WA}(w_j),\quad
E_{ij}=E_{\rm edge}(h_i,h_j,r_{ij},B(d_{ij}),
                    S_i,S_j),\\
S_{ij}&=\mathcal L_i(S_i)+\mathcal L_j(S_j),\\
C_i&=\operatorname{TP}_{\rm UVU}(E_{ij},R_i),\quad
C_j=\operatorname{TP}_{\rm UVU}(E_{ij},R_j),\\
C_{S,ij}&=\operatorname{TP}_{\rm UVU}(C_i+C_j,S_{ij}).
\end{aligned}
\]
Here $E_{\rm WA}$ and $E_{\rm edge}$ are equivariant encoders,
$\mathcal L_i$ and $\mathcal L_j$ are learned state lifts, and all quantities
in this display are regular fields. If
$P_{ij}^{(p)}\in\{R_i,R_j,E_{ij},S_{ij},C_i,C_j+C_{S,ij}\}$ and the invariant
context produces gates $g_{ij}^{(p)}$, the edge message is
\[
M_{ij}=\frac{1}{\sqrt6}\sum_{p=1}^{6}
2\sigma(g_{ij}^{(p)})\odot P_{ij}^{(p)}.
\]
The index $p$ enumerates the six paths, $\sigma$ is the logistic sigmoid, and
each gate has the channel multiplicity of its associated regular field.

The regular message is projected to trivial scalar channels for the node
update. The persistent regular state receives a residual update with learned
per-state-channel gains initialized to 0.1. Direct planar force prediction uses
three covariant bases (edge, receiver WA, sender WA), invariant basis weights,
and a final \(D_6\)-equivariant linear readout. The vertical component combines the
regular message with an invariant edge-z basis and terminates in a trivial
representation.
Let $\delta_{ij}\in\mathbb R^H$ denote the direct invariant scalar
edge path and let $\gamma_S$ be the learned vector of residual-state
gains, initialized component-wise to $0.1$. Then
\[
\begin{aligned}
h_i'&=W_{\rm res}h_i+
U\!\left([h_i,\mathcal A_i^{\rm dn}
(\mathcal L_{{\rm Reg}\to1^H}(M_{ij})+
\delta_{ij})]\right),\\
S_i'&=S_i+
\gamma_S\odot
\mathcal L_{{\rm Reg}\to S}\!\left(\mathcal A_i^{\rm dn}(M_{ij})\right),\\
f_i&=\mathcal A_i^{\rm dn}
\left(\mathcal F(M_{ij},r_{ij,xy},w_i,w_j,
r_{ij,z})\right).
\end{aligned}
\]
Here $\mathcal F$ denotes the learned typed readout that returns a planar
standard vector and a trivial $z$ scalar.

The real-system configuration leaves the optional compact joint edge--WA invariant off.
edge--WA interactions are instead represented by the explicit equivariant
coupling paths.

Toy \path{wa_advanced_sr} implements the same
six-path, gated regular-state construction for each selected wallpaper group.
It generalizes the representation types, number of WA vectors, and optional
nonsymmorphic member field, it also uses periodic edges and the toy radial and
node encodings.

\paragraph{WP-RegTP: regular fields with explicit tensor products.}\label{wp-regtp-modified-finite-group-wa-linear-model}

This is a denser finite-group architecture. 
The model constructs three principal regular fields:

\begin{enumerate}
\def\labelenumi{\arabic{enumi}.}
\item
  A direct field from scalar invariants, receiver/sender persistent states,
  $r_{ij,xy}$, and $w_i$.
\item
  An edge field from the planar edge vector and smooth circular harmonics.
\item
  A WA field from the WA vector and smooth circular harmonics.
\end{enumerate}

An explicit \(D_6\) tensor product couples the edge and WA fields. A second tensor
product couples this result to the lifted persistent state. The direct, edge,
WA, edge--WA, and state-coupled paths are concatenated and mixed by two
equivariant regular-to-regular layers with SiLU activations.

Let $D_{ij}$, $E_{ij}$, and $W_i$ denote the direct, edge, and WA regular
fields, and let $S_i$ be the persistent regular state. The complete
message is
\[
\begin{aligned}
C_{ij}&=\operatorname{TP}_{D_6}(E_{ij},W_i),\\
Q_{ij}&=\operatorname{TP}_{D_6}
\!\left(D_{ij}+C_{ij},\mathcal L_S(S_i+S_j)\right),\\
M_{ij}&=\operatorname{Mix}_{D_6}
\left([D_{ij},E_{ij},W_i,C_{ij},Q_{ij}]\right).
\end{aligned}
\]
Here $\mathcal L_S$ lifts the receiver and sender states into the tensor-product
field, $\operatorname{TP}_{D_6}$ is the learned $D_6$-equivariant tensor
product, and $\operatorname{Mix}_{D_6}$ is the subsequent equivariant channel
mixing network.

The scalar update has a strong invariant baseline. It combines the old scalar
state, a direct invariant edge aggregation, and invariant norms obtained by
decomposing the regular field into \(D_6\) irreducible representations.  A separate
regular-derived scalar correction is multiplied by a learned gain initialized
to zero. This initialization makes early optimization behave like the stable
invariant baseline and allows the symmetry-rich correction to enter gradually.
Let $\delta_{ij}\in\mathbb R^H$ be the direct invariant scalar
edge path and define its receiver aggregation
$\delta_i=\mathcal A_i^{\rm sum}(\delta)$. Then
\[
\begin{aligned}
\bar M_i&=\mathcal A_i^{\rm dn}(M_{ij}),\\
h_i^{\rm base}
 &=W_{\rm base}h_i\\
 &\quad+U_{\rm base}\!\left([h_i,\delta_i,
 E_{\rm norm}(\operatorname{InvNorm}(\bar M_i))]\right),\\
h_i'&=h_i^{\rm base}+
\eta\,U_{\rm reg}\!\left(
[h_i,\mathcal A_i^{\rm dn}
(\mathcal L_{{\rm Reg}\to1^H}M_{ij})]\right),\\
\eta(0)&=0,\qquad
S_i'=\mathcal L_{{\rm Reg}\to S}(\bar M_i),\\
f_i&=\mathcal A_i^{\rm dn}
\left(\mathcal L_{{\rm Reg}\to(V\oplus1)}(M_{ij})\right).
\end{aligned}
\]

The maps $W_{\rm base}$, $U_{\rm base}$, $E_{\rm norm}$, and $U_{\rm reg}$
are learned scalar maps.  The scalar $\eta$ is the regular correction gain,
whose initialization is denoted by $\eta(0)=0$.
$\mathcal L_{{\rm Reg}\to S}$ replaces
rather than residually updates the persistent state.

The persistent state is replaced after every message. Planar and vertical force
components are equivariant projections of the regular edge message. Compact
\(D_6\) edge and WA features and the duplicated edge--WA dot product are always
included by this implementation, regardless of the external joint-feature
flag. Toy \path{wa_linear_mod} uses this same direct
field--geometry field--WA field--tensor-product design and the same
zero-initialized regular correction, but instantiates it for arbitrary
wallpaper groups with the toy periodic graph and group-dependent registry
inputs. 

\paragraph{WP-O3: coupled \(O(3)\) and \(D_6\) processing.}\label{wp-o3-coupled-o3-and-d6-processing}

This hybrid model combines a full \(O(3)\)-equivariant UVU tensor-state layer with
the~WP-Reg~\(D_6\) wallpaper layer. Its purpose is to preserve general 3D
geometric interactions while injecting the known planar \(D_6\) structure and
WA conditioning.

The \(O(3)\) branch uses $1o$, $1e$, and $2e$
families, the \(D_6\) wallpaper sublayer
uses regular hidden fields and regular state fields. A separate three-component
polar feedback vector is retained between layers.

Within one layer:

\begin{enumerate}
\def\labelenumi{\arabic{enumi}.}
\item
  The previous \(D_6\) feedback vector is embedded into the \(O(3)\) tensor state with
  a learned equivariant map and gain initialized to 0.02.
\item
  The \(O(3)\) UVU layer updates scalar and tensor states and proposes an \(O(3)\)
  polar force.
\item
  The \(O(3)\) tensor irreps and force are restricted to the planar \(D_6\) subgroup
  and equivariantly encoded as the initial condition for the wallpaper state.
\item
  The \(D_6\) advanced scalar/regular layer performs WA-conditioned message
  passing.
\item
  The difference between the resulting wallpaper state and its condition is
  projected to a new polar feedback vector for the next layer.
\item
  \(D_6\)-equivariant maps combine the two scalar outputs and the two force
  proposals.
\end{enumerate}

Writing the incoming state as $(T_i,p_i)$, where $T_i$
is the \(O(3)\) tensor state and $p_i$ the polar feedback vector, one layer
can be summarized as
\[
\begin{aligned}
\widetilde{T}_i
 &=T_i+\gamma_{\rm fb}\mathcal L_{1o\to T}(p_i),\\
(h_i^{O3},T_i',f_i^{O3})
 &=\operatorname{UVU}_{O(3)}
   (h_i,\widetilde{T}_i,\{r_{ij}\}_j),\\
C_i&=\operatorname{Cond}_{D_6}
  \!\left(\operatorname{Res}_{O(3)\downarrow D_6}
  [T_i',f_i^{O3}]\right),\\
(h_i^{W},S_i^{W},f_i^{W})
 &=\operatorname{WPReg}
  (h_i^{O3},\{r_{ij},w_i\}_j,C_i),\\
p_i'&=\mathcal L_{S\to1o}(S_i^{W}-C_i),\\
h_i'&=\mathcal L_{\rm scal}
 [h_i^{O3},h_i^{W}],\qquad
f_i'=\mathcal L_{\rm polar}
 [f_i^{O3},f_i^{W}].
\end{aligned}
\]
Here $\gamma_{\rm fb}$ is the learned feedback gain initialized to $0.02$.
$\mathcal L_{1o\to T}$ embeds a polar vector into the \(O(3)\) tensor-state type.
$\operatorname{UVU}_{O(3)}$ denotes the complete \(O(3)\) message-passing
sublayer. The map $\operatorname{Res}_{O(3)\downarrow D_6}$ restricts \(O(3)\)
irreps to the planar subgroup. The map $\operatorname{Cond}_{D_6}$ encodes the restricted
field as the initial wallpaper-state condition $C_i$. The maps
$\mathcal L_{S\to1o}$, $\mathcal L_{\rm scal}$, and
$\mathcal L_{\rm polar}$ are learned equivariant output maps.
The wallpaper condition $C_i$ and state $S_i^W$ have the same
representation type, so their difference is well defined.
The \(O(3)\) operation in the second line is itself edge message passing.  With
$a_{ij}=Y_{\ell\le2}(\widehat{r}_{ij})$,
\[
q_{ij}=\operatorname{TP}_{\rm UVU}
\!\left(P[h_j,\widetilde{T}_j],
a_{ij},W(B(d_{ij}))\right),\qquad
q_i=\mathcal A_i^{\rm sum}(q_{ij}).
\]
Here $Y_{\ell\le2}$ is the spherical-harmonic edge embedding, $P$ is a learned
equivariant projection to the tensor-product input type, and $W$ is the radial
MLP that generates tensor-product weights. The scalar and tensor parts of
$q_i$ produce the \(O(3)\) state update shown above.

The model therefore realizes bidirectional interaction: 3D \(O(3)\) features
condition \(D_6\) processing, and a \(D_6\) polar summary feeds back into the next \(O(3)\)
layer. 

The toy architecture \path{wa_advanced_o3_feedback} has the same
\(O(3)\)-to-wallpaper conditioning and wallpaper-to-\(O(3)\) feedback topology.
Its \(O(3)\) branch, however, uses a fully connected tensor product rather than
the real model's factorized UVU product.  It is also wallpaper-group generic,
periodic, and can include registry fields. 

\paragraph{Scalar full-cell GNN baseline with EGNN-style force head.}\label{full_cell_egnn-scalar-full-cell-egnn-baseline}

This model operates on all adsorbate and graphene atoms and deliberately uses
neither \(D_6\) features nor WA encoding. Its edge message is

\[
m_{ij}=\operatorname{MLP}\!\left([h_i,h_j,\lVert r_{ij}\rVert^2,B(d_{ij})]\right),
\]

followed by summed aggregation and the residual scalar update
\[
h_i'=W_{\rm res}h_i+
U\!\left([h_i,\mathcal A_i^{\rm sum}(m)]\right).
\]
The direct vector proposal is the standard EGNN form

\[
f_i=\sum_{j\in\mathcal N(i)}\alpha(m_{ij})r_{ij}.
\]

Because $r_{ij}$ is a full 3D relative vector and $\alpha$ is invariant,
the proposal
is \(O(3)\) equivariant. The scalar energy stream is invariant to global rotations,
reflections, and translations (apart from the separately concatenated global
cell representation in the common readout). Toy \path{full_graph_invariant} has the same
EGNN-style edge message, scalar receiver update, and edge-weighted 3D force.
It differs in the common toy input, radial, and periodic-edge construction.
Here $\alpha:\mathbb R^H\to\mathbb R$ is a learned invariant scalar readout.

\paragraph{E3nn full-cell model with $\ell_{\max}=1$ geometric features.}\label{full_cell_e3nn_lmax1-o3-uvu-model-with-vector-states}

This full-cell e3nn model carries scalar $0e$ channels and persistent $1o$ and
$1e$ representations. Edge attributes are normalized spherical harmonics through
$\ell=1$, i.e., $0e+1o$. Distances are represented by the common 16-function smooth
radial basis.

To control memory and arithmetic cost, wide node states are projected once per
node to a narrow tensor-product state. All symmetry-allowed products between this
state and the edge spherical harmonics are built with UVU tensor products. In
UVU mode the output multiplicity is tied to the input multiplicity, avoiding
the cubic multiplicity scaling of a fully connected UVW product. A radial MLP
produces edge-specific tensor-product weights.

The narrow equivariant edge products are summed before a wide equivariant
channel-mixing map. Scalar updates use the old scalar state, scalar messages,
and smooth norms of tensor messages. Tensor updates use sigmoid gates generated
from scalar features and a residual equivariant self-map. A separate UVU path
selects a polar $1o$ output for the direct force proposal.

Let $T_i$ denote the concatenated $1o+1e$ state,
$a_{ij}=Y_{\ell\le1}(\widehat{r}_{ij})$, and
$n_j=P[h_j,T_j]$ the narrow projected node field.
Here $Y_{\ell\le1}$ is the normalized spherical-harmonic embedding and $P$ is
a learned equivariant projection. The message and force radial networks
$W_m$ and $W_F$ generate the weights for their respective tensor products.
The message and receiver updates are
\[
\begin{aligned}
q_{ij}&=\operatorname{TP}_{\rm UVU}
(n_j,a_{ij},W_m(B(d_{ij}))),\\
q_i&=\mathcal A_i^{\rm sum}(q_{ij}),\qquad
(q_i^0,q_i^T)=\operatorname{Mix}(q_i),\\
h_i'&=L_0\!\left(L_{\rm self}^0h_i+
U([h_i,q_i^0,\operatorname{InvNorm}(q_i^T)])\right),\\
T_i'&=L_T\!\left(L_{\rm self}^TT_i+
\sigma(G([h_i,q_i^0]))\odot q_i^T\right),\\
f_i&=L_F\mathcal A_i^{\rm sum}\!\left(
\operatorname{TP}_{\rm UVU}
(n_j,a_{ij},W_F(B(d_{ij})))\right).
\end{aligned}
\]
In this display, $\operatorname{Mix}$ is the post-aggregation equivariant
channel mixer. The fields $q_i^0$ and $q_i^T$ are its scalar and non-scalar
outputs. The maps $L_0$, $L_T$, and $L_F$ are learned equivariant output maps.
The maps $L_{\rm self}^0$ and $L_{\rm self}^T$ are residual self-maps. Here
$U$ is the scalar update MLP, $G$ is the scalar gate network, $\sigma$ is the
logistic sigmoid, and $\odot$ applies the resulting gates channel-wise to
$q_i^T$.
The last projection selects a polar $1o$ vector.  The UVU intermediate is
aggregated before the wide channel-mixing map, which is algebraically valid
because that map is linear and avoids constructing a wide field per edge.

Including both polar $1o$ and axial $1e$ vector fields lets the model distinguish
their parity under inversion. With $\ell_{\max}=1$ edge harmonics, higher-order
directional structure must emerge through stacked products and persistent node
states rather than entering directly as $\ell=2$ edge features. Toy
\path{full_graph_e3nn_lmax1_tensor_product} uses the same $0e+1o$ edge
harmonics, persistent $1o+1e$ state families, gated tensor update, and polar
force type, but its edge tensor product is fully connected rather than the
real model's narrow UVU factorization. 

\paragraph{E3nn full-cell model with $\ell_{\max}=2$ geometric features.}\label{full_cell_e3nn_lmax2-o3-uvu-model-with-vector-and-quadrupole-states}

This architecture has the same factorized UVU design as the $\ell_{\max}=1$
model but extends both edge geometry and persistent state through $\ell=2$.
It uses $0e$, $1o$, $1e$, and $2e$ components for the node state. Edge
attributes contain spherical harmonics $0e+1o+2e$.

Radial weights, pre-aggregation UVU products,
post-aggregation equivariant channel mixing, smooth tensor norms, scalar gates,
and the polar $1o$ force path otherwise follow the $\ell_{\max}=1$ architecture.

With $T_i$ now containing $1o+1e+2e$ and
$a_{ij}=Y_{\ell\le2}(\widehat{r}_{ij})$, the symbols
$P$, $W_m$, $W_F$, $\operatorname{Mix}$, $L_0$, $L_T$, $L_F$,
$L_{\rm self}^0$, $L_{\rm self}^T$, $U$, and $G$
retain the definitions given for the $\ell_{\max}=1$ model. Its equations are
\[
\begin{aligned}
q_{ij}&=\operatorname{TP}_{\rm UVU}
(P[h_j,T_j],a_{ij},W_m(B(d_{ij}))),\quad
(q_i^0,q_i^T)=\operatorname{Mix}
(\mathcal A_i^{\rm sum}(q_{ij})),\\
h_i'&=L_0\!\left(L_{\rm self}^0h_i+
U([h_i,q_i^0,\operatorname{InvNorm}(q_i^T)])\right),\\
T_i'&=L_T\!\left(L_{\rm self}^TT_i+
\sigma(G([h_i,q_i^0]))\odot q_i^T\right),\\
f_i&=L_F\mathcal A_i^{\rm sum}\!\left(
\operatorname{TP}_{\rm UVU}
(P[h_j,T_j],a_{ij},W_F(B(d_{ij})))\right).
\end{aligned}
\]

The $2e$ channels represent rank-2 symmetric traceless information and permit
quadrupolar angular structure to be communicated directly in one layer. This
increases angular expressivity and state memory relative to $\ell_{\max}=1$, while the
smaller scalar width keeps the total registered parameter count near one
million. Toy
\path{full_graph_e3nn_tensor_product} carries the corresponding
$1o+1e+2e$ state and $\ell\le2$ edge harmonics, but again uses a fully
connected tensor product rather than the real narrow UVU schedule.  Its
geometry and initial features follow the common toy construction, whereas the real
model uses the full cached cell and the real feature/radial definitions.

\subsection{Symmetry summary}\label{symmetry-summary}

Table~\ref{tab:symmetry-summary} summarizes how each architecture handles the
symmetry. Here $G$ denotes the selected wallpaper point group. For the real-system
molecule-only models, $G=D_6$. ``Persistent state'' means a non-scalar node
state passed between message-passing layers, rather than a temporary
equivariant edge field. The final column describes the explicit force proposal
used in direct-force-head training. In conservative training this proposal is
discarded and forces are obtained as
$F=-\nabla_{x}E$.

\begingroup
\setlength{\tabcolsep}{3pt}
\renewcommand{\arraystretch}{1.14}
\setlength{\LTcapwidth}{\linewidth}
\small
\begin{longtable}{@{}
  >{\scriptsize\raggedright\arraybackslash}m{0.135\linewidth}
  >{\scriptsize\centering\arraybackslash}m{0.095\linewidth}
  >{\scriptsize\centering\arraybackslash}m{0.085\linewidth}
  >{\scriptsize\raggedright\arraybackslash}m{0.205\linewidth}
  >{\scriptsize\raggedright\arraybackslash}m{0.185\linewidth}
  >{\scriptsize\raggedright\arraybackslash}m{0.19\linewidth}@{}}
\caption{Symmetry summary of the WallpaperNet architectures and the full-cell
baselines. For each model, the columns give the message-passing graph, the
experiments in which it is evaluated, the scalar-message and update mechanism,
the persistent equivariant node state, and the explicit force proposal used in
direct-force-head training.}
\label{tab:symmetry-summary}\\
\toprule
\textbf{Architecture} &
\textbf{Graph} &
\textbf{Evaluated on} &
\textbf{Scalar-message and update mechanism} &
\textbf{Persistent equivariant node state} &
\textbf{Direct-force construction} \\
\midrule
\endfirsthead
\caption[]{Symmetry summary of the WallpaperNet architectures and the full-cell
baselines (continued).}\\
\toprule
\textbf{Architecture} &
\textbf{Graph} &
\textbf{Evaluated on} &
\textbf{Scalar-message and update mechanism} &
\textbf{Persistent equivariant node state} &
\textbf{Direct-force construction} \\
\midrule
\endhead
\bottomrule
\endlastfoot
\multicolumn{6}{@{}l}{\scriptsize\textit{Molecule-only architectures}} \\
\addlinespace[1pt]

WP-Inv &
molecule &
toy and real &
Invariant MLP messages built from compact $G$-invariants, invariant receiver
update &
none &
Invariant coefficient multiplying the planar edge vector, together with a
trivial-representation $z$ scalar \\

WP-InvExt &
molecule &
toy only &
Invariant scalar MLP, the conservative variant additionally feeds invariant
contractions of its vector memory into the scalar update &
conservative: $H$ polar three-vectors, direct: none &
direct variant: invariant coefficients multiplying the planar edge and WA
vectors, plus a trivial $z$ scalar, conservative variant: no direct proposal \\

WP-InvEq &
molecule &
real only &
Invariant $D_6$ core followed by an explicit
$1\!\to\!\mathrm{Reg}\!\to\!1$ scalar post-map &
none &
Provisional edge-weighted force followed by
$(V\oplus1)\!\to\!\mathrm{Reg}\!\to\!(V\oplus1)$
equivariant post-maps \\

WP-Equi &
molecule &
toy and real &
Fully typed $G$-equivariant edge network, trivial-representation projections
provide scalar messages &
none &
Standard-vector $V$ and trivial-$z$ projections of the equivariant edge field \\

WP-VecInv &
molecule &
real only &
Invariant $D_6$ messages augmented by invariant contractions of persistent
polar vectors &
$H$ polar three-vectors with a residual covariant update &
Invariant edge/WA basis coefficients followed by a
$(V\oplus1)\!\to\!\mathrm{Reg}\!\to\!(V\oplus1)$ output network \\

WP-VecEqui &
molecule &
toy only &
Separate regular edge and WA fields coupled by an explicit UVU tensor product,
trivial projections update the scalar stream &
residually updated $G$-regular state &
Regular-to-standard-vector and regular-to-trivial projections of the WA, edge,
and coupled fields \\

WP-Reg &
molecule &
toy and real &
Six gated regular-field paths with explicit UVU couplings, followed by trivial
projections and an invariant scalar update &
residually updated $G$-regular state &
Equivariant readout using the planar edge, receiver-WA, and sender-WA covariant
bases, plus a trivial $z$ output \\

WP-RegTP &
molecule &
toy and real &
Direct, edge, and WA regular fields coupled by tensor products, irrep norms and
a zero-initialized regular correction feed the scalar update &
$G$-regular state replaced by a projection of the aggregated message &
Equivariant $G$-regular-to-$(V\oplus1)$ projection of the edge message \\

WP-O3 &
molecule &
toy and real &
Coupled $O(3)$ tensor and wallpaper-group regular pathways, with bidirectional
conditioning and scalar fusion &
$O(3)$ $1o+1e+2e$ tensor state plus a wallpaper-derived polar feedback vector &
Equivariant fusion of the $O(3)$ polar proposal and the wallpaper-group force
proposal \\

\midrule
\multicolumn{6}{@{}l}{\scriptsize\textit{Full-cell baselines}} \\
\addlinespace[1pt]

Full-cell invariant GNN &
full cell &
toy and real &
$O(3)$-invariant scalar EGNN messages and an invariant receiver update &
none &
Invariant coefficient multiplying the full three-dimensional edge vector \\

Full-cell e3nn ($\ell_{\max}=1$) &
full cell &
toy and real &
$O(3)$-equivariant tensor products with $0e+1o$ edge harmonics, $0e$
projections and invariant tensor norms update the scalar stream &
persistent $1o+1e$ state &
Equivariant projection to a polar $1o$ vector \\

Full-cell e3nn ($\ell_{\max}=2$) &
full cell &
toy and real &
$O(3)$-equivariant tensor products with $0e+1o+2e$ edge harmonics, $0e$
projections and invariant tensor norms update the scalar stream &
persistent $1o+1e+2e$ state &
Equivariant projection to a polar $1o$ vector \\
\end{longtable}
\endgroup

\subsection{Experimental comparability and interpretation}\label{experimental-comparability-and-interpretation}

All models use four message-passing layers, the same initial atomic attributes,
the same radial cutoff and basis size, and approximately one million registered
parameters. Nevertheless, the comparison intentionally includes two distinct
information regimes. Molecule-only models receive WA vectors and an explicit
graphene reference energy but do not propagate through graphene atoms.
Full-cell models receive all graphene coordinates and species directly and do
not use WA encoding or a separate graphene graph reference. Their graph sizes
and edge counts are therefore much larger.

The three real-system scenarios answer different questions. Standard combined
training jointly tests full-total-energy prediction and conservative forces.
Because the real system has a varying number of atoms in each cell, its total
energy varies substantially across samples. We therefore add a force-weighted conservative
experiment in which the energy loss is weighted by a data-derived coefficient of
approximately $1.22\times10^{-6}$ relative to a unit force weight
(Appendix~\ref{sec:mol-grap-results}), which balances the energy and force terms
while retaining
$F=-\nabla E$. Force-head training tests direct equivariant vector
regression, which avoids second derivatives during training and can be
substantially faster, but does not enforce a globally integrable energy/force
field. 

Finally, the graph-level energy readout concatenates raw Cartesian cell-vector
components embedded as raw features. Those components are  frame-aligned in the present
dataset and held fixed in the present dataset convention. Since every cell has its own size, the total energy of the cell is dependent on the cell size as well. Therefore the cell vectors are included in the energy readout to allow the model to learn the energy dependence on the cell size. 

\subsection{Training metrics}
\label{sec:toy-metrics}

Let \(\mathcal M_{\rm eval}\) denote the set of all molecular atoms in the
evaluated split, \(F_i\in\mathbb R^3\) the reference force on atom
\(i\), and \(F_i^{\rm pred}\) its predicted force. Support-atom forces
are excluded from all toy force metrics.

The reported relative force \(L_2\) error is the norm over all molecular atoms
and Cartesian components:
\[
L_{2,\rm rel}^{F}
=
\frac{
\left[
\displaystyle\sum_{i\in\mathcal M_{\rm eval}}
\left\lVert
F_i^{\rm pred}-F_i
\right\rVert_2^2
\right]^{1/2}
}{
\left[
\displaystyle\sum_{i\in\mathcal M_{\rm eval}}
\left\lVert F_i\right\rVert_2^2
\right]^{1/2}
}.
\]
Equivalently, after flattening all force components into the vectors
\(f^{\rm pred}\) and \(f\), respectively,
\[
L_{2,\rm rel}^{F}
=
\frac{\lVert f^{\rm pred}-f\rVert_2}
     {\lVert f\rVert_2}.
\]

For \(K\) scalar targets \(y_k\) and predictions \(y_k^{\rm pred}\), the
coefficient of determination is
\[
R^2
=
1-
\frac{
\displaystyle\sum_{k=1}^{K}(y_k^{\rm pred}-y_k)^2
}{
\displaystyle\sum_{k=1}^{K}(y_k-\overline y)^2
},
\qquad
\overline y=\frac{1}{K}\sum_{k=1}^{K}y_k.
\]
For the reported force-component \(R^2\), the index \(k\) runs over every
\(x\)-, \(y\)-, and \(z\)-component of every molecular atom in the evaluated
split. For an energy \(R^2\), \(k\) runs over systems, using either the total
energy or the specified atom-normalized energy.

The zero-force score compares the model with the baseline
\(F_i^{\rm pred}=0\):
\[
S_{\rm zero}
=
1-
\frac{
\displaystyle\sum_{i\in\mathcal M_{\rm eval}}
\left\lVert
F_i^{\rm pred}-F_i
\right\rVert_2^2
}{
\displaystyle\sum_{i\in\mathcal M_{\rm eval}}
\left\lVert F_i\right\rVert_2^2
}.
\]
Consequently,
\[
S_{\rm zero}
=
1-\left(L_{2,\rm rel}^{F}\right)^2.
\]
A perfect prediction gives \(S_{\rm zero}=1\), predicting zero force gives
\(S_{\rm zero}=0\), and a prediction worse than the zero-force baseline gives
\(S_{\rm zero}<0\).

\subsection{Toy-model training objectives}
\label{sec:toy-objectives}
\label{sec:toy-training-objectives}

Let \(\mathcal B\) be a mini-batch of \(B=|\mathcal B|\) systems. The energy
loss is the mean absolute error of the unnormalized full-system total energies:
\[
L_E
=
\frac{1}{B}
\sum_{b\in\mathcal B}
\left|
E_b^{\rm pred}-E_b
\right|.
\]

Let \(\mathcal M_{\mathcal B}\) be the set of molecular atoms in the
mini-batch. The force loss is the mean squared error over molecular atoms and
Cartesian components:
\[
L_F
=
\frac{1}{3|\mathcal M_{\mathcal B}|}
\sum_{i\in\mathcal M_{\mathcal B}}
\sum_{\alpha\in\{x,y,z\}}
\left(
F_{i\alpha}^{\rm pred}-F_{i\alpha}
\right)^2
=
\frac{1}{3|\mathcal M_{\mathcal B}|}
\sum_{i\in\mathcal M_{\mathcal B}}
\left\lVert
F_i^{\rm pred}-F_i
\right\rVert_2^2.
\]

In conservative combined training, forces are obtained from the predicted
total energy:
\[
F_i^{\rm pred}
=
-\frac{\partial}{\partial x_i}
\sum_{b\in\mathcal B}E_b^{\rm pred}.
\]
Because the systems in a batch are disconnected, this is equivalent to
differentiating the energy of the system containing atom \(i\). Periodic edge
displacements and WA vectors are reconnected to the atomic coordinates before
taking this derivative.

The conservative combined objective is
\[
\mathcal L_{\rm combined}
=
\lambda_E L_E+\lambda_F L_F,
\qquad
\lambda_E=\lambda_F=1,
\]
and therefore
\[
\mathcal L_{\rm combined}=L_E+L_F.
\]

In direct-force-head training, the four layer-wise equivariant force proposals
are summed, as defined in the direct-force scenario above:
\[
F_i^{\rm pred}=\sum_{\ell=1}^{4}f_i^{(\ell)}.
\]
The energy term is deliberately omitted. The objective evaluated by the training code is
\[
\mathcal L_{\rm head}
=
\lambda_F L_F
=
L_F.
\]
Thus the force-head experiments neither optimize nor report an energy loss.

\subsection{Toy-model optimization and training conditions}
\label{sec:toy-optimization}

Every toy model was trained for a maximum of \(1000\) epochs. No early
stopping was used, so an uninterrupted run executes all \(1000\) epochs. The
checkpoint with the lowest validation loss was retained and used for the
reported training, validation, and test metrics.

Optimization used AdamW~\citep{loshchilovDecoupledWeightDecay2019} with initial learning rate
\[
\eta_0=10^{-3}.
\]
Only the learning rate was explicitly supplied to the optimizer. With the
PyTorch version used for these experiments, the remaining AdamW parameters
were
\[
(\beta_1,\beta_2)=(0.9,0.999),\qquad
\epsilon_{\rm Adam}=10^{-8},\qquad
\lambda_{\rm wd}=10^{-2},
\]
with AMSGrad disabled. Here \(\epsilon_{\rm Adam}\) is the numerical-stability
constant and \(\lambda_{\rm wd}\) is the weight-decay coefficient.

The learning rate was controlled by a
\texttt{ReduceLROnPlateau} scheduler monitoring the epoch-level validation
loss:
\[
\eta\leftarrow 0.5\,\eta
\]
after a plateau with patience \(20\) epochs. The scheduler used minimization
mode, an absolute improvement threshold of \(10^{-12}\), zero cooldown, a
minimum learning rate of \(0\), and a numerical update threshold of
\(\epsilon_{\rm sched}=10^{-8}\). It was evaluated once per epoch.

Each toy-model run used one GPU, a batch size of \(64\) systems, four
message-passing layers, eight data-loader workers, and an edge cutoff of
\(6\,\text{\AA}\). The data were split into training, validation, and test
subsets in the ratio
\[
0.8:0.1:0.1.
\]
The training loader was shuffled, whereas the validation and test loaders were
not. The experiment seed was used for both the data split and Lightning's model
and worker seeding. The toy-model experiments used three independent seeds (24, 42, 67) for both the data split and model initialization.

\subsection{Real-system optimization and training conditions}
\label{sec:real-system-optimization}

Similar training scenario was applied to the real-system models, unless explicitly stated otherwise. The real-system model was trained for a maximum of \(600\) epochs. Runs
that reached the cluster wall-time limit were resumed from the most recent
checkpoint, which was saved every five epochs, until all \(600\) epochs had
been completed. The checkpoint with the lowest validation loss was retained
and used for the reported training, validation, and test metrics.

Each real-system run used four GPUs with distributed data-parallel training,
a global batch size of \(64\) systems, and a per-GPU batch size of \(16\).
Gradient accumulation was not used. Every model used four message-passing
layers, eight data-loader workers, an edge cutoff of \(6\,\text{\AA}\), and at
most \(128\) neighbors per atom.

The \(15089\) systems were split into \(12071\) training, \(1509\) validation,
and \(1509\) test systems, corresponding to the ratio \(
0.8:0.1:0.1.
\) The training loader was shuffled, whereas the validation and test loaders were
not. Three independently seeded data partitions initial model seed where used(seeds: 1, 24, 42). For the some of the models we observed unstable gradient behavior during the first 40 epochs of the training. For four force-head models, the
global \(L_2\) norm of the gradients over all trainable parameters was clipped
at every optimizer step. The clipping thresholds were \(1\) for
WP-Equi, \(20\) for WP-Inv, \(30\) for
WP-O3, and \(50\) for
WP-Reg. All combined-loss models and all remaining
force-head models were trained without
gradient clipping. The global gradient norm before and after clipping was
logged every \(100\) optimization steps.

\section{Toy benchmark results for all wallpaper groups}
\label{sec:all_toys}

\paragraph{Remaining observations on the averaged results.}
Among the WallpaperNet models in Table~\ref{tab:toy-model-all-groups}, WP-InvExt
is distinctly the second best, with the remaining equivariant architectures
performing very similarly to each other. Direct force-head learning is also the
only target for which the full-cell e3nn models outperform all WallpaperNet
models. Increasing the angular resolution of the full-cell e3nn baseline from
\(\ell_{\max}=1\) to \(\ell_{\max}=2\) provides no systematic improvement in
the averaged results, and the two models have essentially identical force
accuracy. Rank-two features therefore do not appear to be necessary for
achieving the accuracy level reached by the full-cell baselines on this
benchmark.

\begin{figure}[H]
  \centering
  \includegraphics[width=\textwidth]{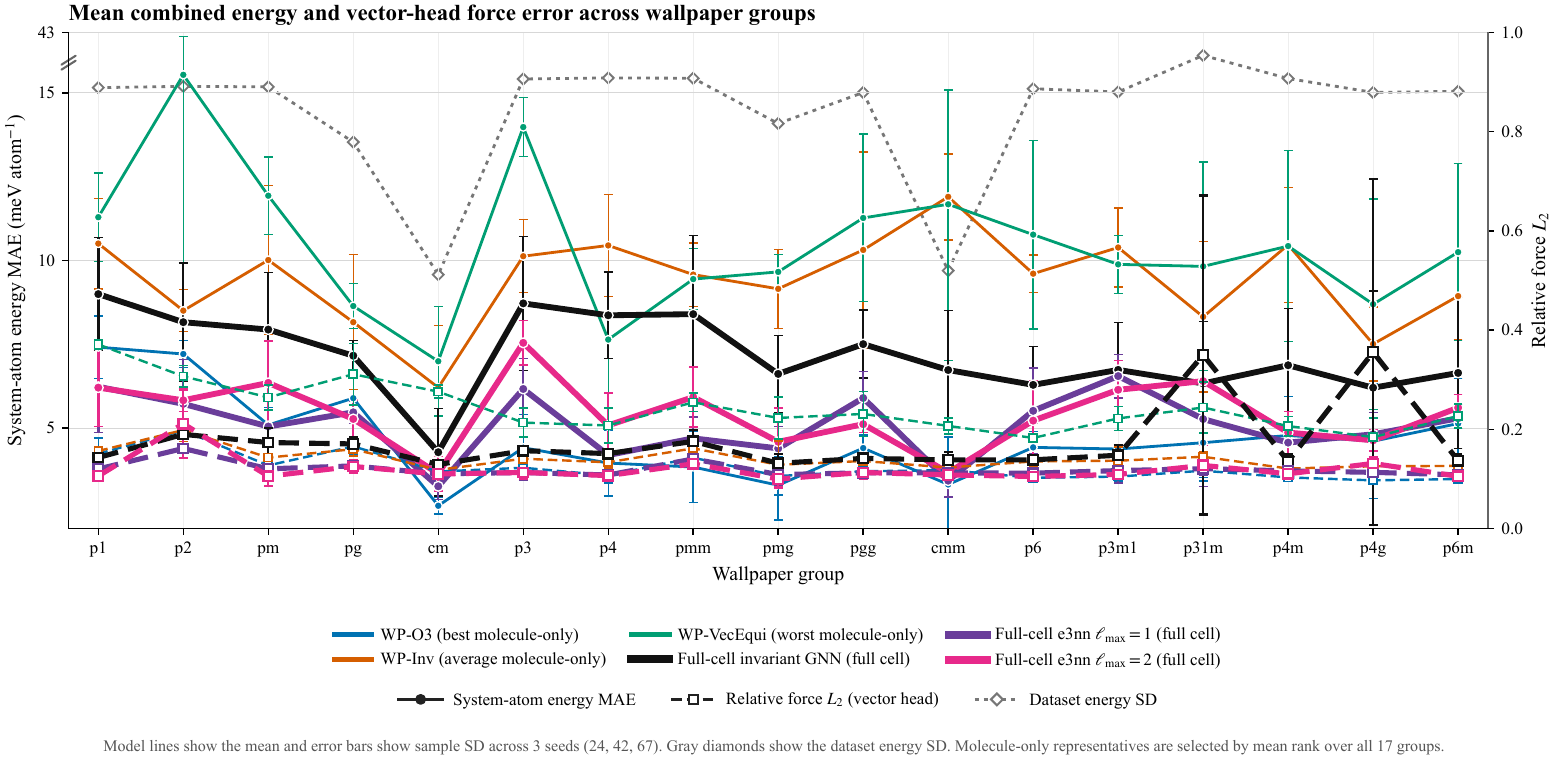}
  \caption{Group-resolved toy-benchmark performance. Solid curves show the
  mean system-atom energy MAE from combined training (left axis), and dashed
  curves show the mean relative force \(L_2\) error from vector-head training
  (right axis). Error bars denote the sample standard deviation across three
  seeds. WP-O3, WP-Inv, and WP-VecEqui are respectively the best-, average-,
  and worst-ranked molecule-only representatives across all groups, while all
  three full-cell baselines are shown. Gray diamonds indicate the standard
  deviation of the mean residual energy per atom within each group.}
  \label{fig:toy-model-energy-force-by-group}
\end{figure}

The group-resolved results largely support the trends obtained from averaging over all 17 wallpaper groups, but they also reveal a pronounced dependence on the symmetry of the individual system. Figure~\ref{fig:toy-model-energy-force-by-group} summarizes the energy and force errors across groups, and Tables~\ref{tab:toy-model-p1}--\ref{tab:toy-model-p6m} report the corresponding numerical results. Across the WallpaperNet architectures, WP-O3 is the most consistently accurate model. It gives the lowest energy MAE among the WallpaperNet models for 16 of the 17 groups and the best combined-training force metrics for all 17 groups. The only exception for energy is \(cmm\) (Table~\ref{tab:toy-model-cmm}), where WP-InvExt gives a slightly lower error. WP-InvExt is otherwise remarkably consistent: it is typically the second-best WallpaperNet model for energy and for forces obtained from conservative training. In contrast, increasing the complexity of the regular-representation pathways from WP-Equi through WP-VecEqui, WP-Reg, and WP-RegTP does not lead to a systematic improvement. This indicates that explicit wallpaper-group equivariance alone is not sufficient to guarantee higher accuracy. Rather, the combination of wallpaper-group processing with \(O(3)\) tensor features used by WP-O3 appears to provide the most effective representation of the geometric interactions in this benchmark.

For the lowest-symmetry groups, the advantage of exploiting wallpaper symmetry is less pronounced. For \(p1\) (Table~\ref{tab:toy-model-p1}), the full-cell e3nn models provide the lowest errors: the \(\ell_{\max}=2\) model gives the lowest energy MAE and the best vector-head force results, while the \(\ell_{\max}=1\) model performs best for forces from combined training. WP-O3 is nevertheless the strongest WallpaperNet architecture for this group, substantially improving over the other WP models. A similar behavior is observed for \(p2\) (Table~\ref{tab:toy-model-p2}), where the full-cell e3nn \(\ell_{\max}=1\) model gives the best result for every reported accuracy metric. The \(p2\) results also show comparatively large seed-to-seed variation for some of the regular-representation models, particularly WP-VecEqui and WP-Reg, suggesting less robust optimization for these architectures in this group.

The mirror and glide groups \(pm\) (Table~\ref{tab:toy-model-pm}) and \(pg\) (Table~\ref{tab:toy-model-pg}) show two somewhat different regimes. For \(pm\), WP-O3 gives the best combined-training force prediction, with a relative force error of \(18.8\%\) and \(R^2=0.964\), while its energy error of \(5.09\) meV atom\(^{-1}\) is essentially comparable to the \(5.05\) meV atom\(^{-1}\) obtained by the full-cell e3nn \(\ell_{\max}=1\) model. For vector-head training, however, the full-cell e3nn models retain an advantage. In \(pg\), the full-cell e3nn \(\ell_{\max}=2\) model performs best across the principal metrics, although WP-O3 remains the best WallpaperNet architecture for combined training. Interestingly, WP-Inv gives the best vector-head force result among the WallpaperNet models for \(pg\), marginally outperforming WP-O3. Thus, in these low-order reflection/glide groups, exploiting the wallpaper representation does not consistently outperform a full-cell \(O(3)\)-equivariant treatment.

The picture begins to change for \(cm\) (Table~\ref{tab:toy-model-cm}) and for the groups containing higher-order rotations. For \(cm\), WP-O3 already gives the lowest energy MAE and the best conservative-force prediction, whereas the vector-head results of the full-cell e3nn models remain slightly better. For \(p3\) (Table~\ref{tab:toy-model-p3}), WP-O3 reduces the energy MAE to \(4.42\) meV atom\(^{-1}\) and the combined relative force error to \(15.4\%\), outperforming all full-cell models under combined training. The full-cell e3nn models retain a small advantage for the vector head. The same pattern occurs for \(p4\) (Table~\ref{tab:toy-model-p4}): WP-O3 gives the best energy and combined-force performance and reaches a vector-head relative force error of \(10.7\%\), equal to the best full-cell value, although the full-cell \(\ell_{\max}=2\) model has a slightly smaller component MAE. For \(p6\) (Table~\ref{tab:toy-model-p6}), the difference narrows further: WP-O3 gives the best energy, combined-force, and vector-head relative-force errors, while its vector-head component MAE of \(229.2\) meV \AA\(^{-1}\) is essentially comparable to the \(227.5\) meV \AA\(^{-1}\) of the full-cell \(\ell_{\max}=2\) model. These groups therefore show that the advantage of the symmetry-reduced representation becomes increasingly apparent once nontrivial rotational symmetry is present.

A similar trend is found for the twofold groups \(pmm\) (Table~\ref{tab:toy-model-pmm}), \(pmg\) (Table~\ref{tab:toy-model-pmg}), \(pgg\) (Table~\ref{tab:toy-model-pgg}), and \(cmm\) (Table~\ref{tab:toy-model-cmm}). WP-O3 gives the best combined-training force performance in all four cases and the lowest energy error for \(pmm\), \(pmg\), and \(pgg\). For \(cmm\), WP-InvExt instead gives the lowest energy MAE, \(2.98\) meV atom\(^{-1}\), compared with \(3.32\) meV atom\(^{-1}\) for WP-O3. This is the only wallpaper group for which another WallpaperNet architecture outperforms WP-O3 in energy. Under vector-head training, the full-cell e3nn models remain slightly more accurate for these four groups. The differences are particularly small for \(pmg\) and \(pgg\): for example, in \(pmg\), WP-O3 reaches a relative vector-force error of \(10.6\%\), compared with \(10.0\%\) for the full-cell e3nn \(\ell_{\max}=2\) model. Thus, even where the full-cell baseline remains numerically best for the direct force head, the symmetry-reduced WP-O3 model is already operating in essentially the same accuracy regime.

The strongest results for WallpaperNet are obtained for the higher-symmetry groups \(p3m1\) (Table~\ref{tab:toy-model-p3m1}), \(p31m\) (Table~\ref{tab:toy-model-p31m}), \(p4m\) (Table~\ref{tab:toy-model-p4m}), \(p4g\) (Table~\ref{tab:toy-model-p4g}), and \(p6m\) (Table~\ref{tab:toy-model-p6m}). WP-O3 gives the best force metrics under both training modes for all five groups. For \(p3m1\) and \(p31m\), it also gives the lowest energy error, meaning that WP-O3 outperforms all three full-cell baselines simultaneously for energy, conservative forces, and direct vector-force prediction. The same complete dominance is observed for \(p4g\) and \(p6m\). In \(p4g\), for example, WP-O3 reaches a vector-head relative force error of \(9.7\%\), a component MAE of \(207.0\) meV \AA\(^{-1}\), and \(R^2=0.990\), while its energy MAE of \(4.61\) meV atom\(^{-1}\) is also marginally below that of the full-cell e3nn \(\ell_{\max}=2\) model. For \(p6m\), WP-O3 similarly gives \(10.0\%\), \(212.4\) meV \AA\(^{-1}\), and \(R^2=0.990\) for the vector head. The only exception within this set is the \(p4m\) energy prediction, where full-cell e3nn \(\ell_{\max}=1\) gives \(4.57\) meV atom\(^{-1}\) compared with \(4.78\) meV atom\(^{-1}\) for WP-O3. The reported seed variations substantially overlap, while WP-O3 remains the best model for every force metric.

The group-resolved results also refine the comparison between conservative forces and the explicitly learned vector force head. Direct vector-force training improves the force metrics in the overwhelming majority of model--group combinations, but the improvement is not literally universal. For WP-VecEqui and WP-RegTP on \(p1\) (Table~\ref{tab:toy-model-p1}), vector-head training produces larger force errors than combined training, and WP-RegTP shows a small degradation in relative force error and \(R^2\) for \(pmm\) (Table~\ref{tab:toy-model-pmm}). More pronounced failures occur for the full-cell invariant GNN on \(p31m\) (Table~\ref{tab:toy-model-p31m}) and \(p4g\) (Table~\ref{tab:toy-model-p4g}), where the vector-head results have both substantially larger mean errors and extremely large seed-to-seed variations. Apart from these isolated cases, the vector head consistently reduces force errors, confirming that direct force supervision is substantially easier for these strongly nonequilibrium synthetic structures than recovering forces through the energy gradient.

Taken together, the individual-group results therefore strengthen rather than merely reproduce the averaged comparison. The benefit of wallpaper-aware processing is smallest for groups with little nontrivial point-group symmetry, where full-cell \(O(3)\)-equivariant models frequently remain more accurate. As the amount of rotational, reflection, and glide symmetry increases, WP-O3 becomes increasingly competitive and, for several of the most symmetric groups, surpasses the full-cell baselines even for direct vector-force prediction. At the same time, the strong and consistent performance of WP-InvExt shows that much of the gain over the basic invariant model can already be obtained by enriching scalar message passing with covariant geometric directions. The poorer and more variable results of the intermediate regular-representation architectures further show that increasing equivariant architectural complexity by itself does not guarantee better prediction. Instead, the results favor architectures in which the reduced wallpaper representation is supplemented by geometrically expressive features, most clearly realized by WP-O3.

In Tables~\ref{tab:toy-model-p1}--\ref{tab:toy-model-p6m}, entries report the
mean \(\pm\) sample standard deviation across three seeds, C and V denote
combined and vector-head training, and the energy MAE is taken from combined
training. Bold marks the best mean for each metric and training mode. Best
epochs are not ranked.

\begin{table*}[p]
\centering
\caption{Toy-model performance for wallpaper group $\mathrm{p1}$ (1,000 structures, mean $\Delta E=-7.12$ eV atom$^{-1}$, energy SD 17.42 meV atom$^{-1}$).}
\label{tab:toy-model-p1}
\fontsize{8}{9.5}\selectfont
\setlength{\tabcolsep}{4pt}
\begin{tabular}{@{}lcccccc@{}}
\toprule
Model &
\shortstack{Energy\\MAE $\downarrow$\\(meV atom$^{-1}$)} &
Mode &
\shortstack{Rel.\ force\\$L_2$ $\downarrow$ (\%)} &
\shortstack{Force comp.\\MAE $\downarrow$\\(meV \AA$^{-1}$)} &
\shortstack{Force comp.\\$R^2$ $\uparrow$} &
\shortstack{Best\\epoch} \\
\midrule
\multicolumn{7}{l}{\textit{WallpaperNet models}} \\
\multirow{2}{*}{WP-Inv} & \multirow{2}{*}{\(10.50 \pm 1.35\)} & C & \(32.6 \pm 2.3\) & \(751.1 \pm 58.6\) & \(0.893 \pm 0.015\) & \(739.7 \pm 231.2\) \\
 & & V & \(15.6 \pm 0.6\) & \(341.6 \pm 9.3\) & \(0.975 \pm 0.002\) & \(304.7 \pm 47.0\) \\
\addlinespace[3pt]
\multirow{2}{*}{WP-InvExt} & \multirow{2}{*}{\(9.28 \pm 2.37\)} & C & \(28.2 \pm 5.6\) & \(635.0 \pm 123.5\) & \(0.918 \pm 0.033\) & \(511.7 \pm 121.5\) \\
 & & V & \(16.7 \pm 1.1\) & \(363.0 \pm 33.3\) & \(0.972 \pm 0.004\) & \(320.7 \pm 94.2\) \\
\addlinespace[3pt]
\multirow{2}{*}{WP-Equi} & \multirow{2}{*}{\(12.45 \pm 6.14\)} & C & \(34.9 \pm 4.9\) & \(798.3 \pm 136.3\) & \(0.877 \pm 0.033\) & \(510.7 \pm 345.5\) \\
 & & V & \(26.8 \pm 0.9\) & \(616.5 \pm 29.1\) & \(0.928 \pm 0.005\) & \(233.3 \pm 25.1\) \\
\bottomrule
\end{tabular}
\end{table*}

\begin{table*}[p]
\centering
\caption{Toy-model performance for wallpaper group $\mathrm{p2}$ (1,000 structures, mean $\Delta E=-7.15$ eV atom$^{-1}$, energy SD 18.03 meV atom$^{-1}$).}
\label{tab:toy-model-p2}
\fontsize{8}{9.5}\selectfont
\setlength{\tabcolsep}{4pt}
\begin{tabular}{@{}lcccccc@{}}
\toprule
Model &
\shortstack{Energy\\MAE $\downarrow$\\(meV atom$^{-1}$)} &
Mode &
\shortstack{Rel.\ force\\$L_2$ $\downarrow$ (\%)} &
\shortstack{Force comp.\\MAE $\downarrow$\\(meV \AA$^{-1}$)} &
\shortstack{Force comp.\\$R^2$ $\uparrow$} &
\shortstack{Best\\epoch} \\
\midrule
\multicolumn{7}{l}{\textit{WallpaperNet models}} \\
\multirow{2}{*}{WP-Inv} & \multirow{2}{*}{\(8.50 \pm 0.63\)} & C & \(35.9 \pm 4.3\) & \(741.0 \pm 84.1\) & \(0.870 \pm 0.030\) & \(558.7 \pm 149.1\) \\
 & & V & \(19.9 \pm 2.3\) & \(400.9 \pm 50.4\) & \(0.960 \pm 0.009\) & \(210.7 \pm 38.5\) \\
\addlinespace[3pt]
\multirow{2}{*}{WP-InvExt} & \multirow{2}{*}{\(11.61 \pm 3.68\)} & C & \(31.5 \pm 3.2\) & \(672.2 \pm 63.2\) & \(0.900 \pm 0.020\) & \(343.0 \pm 133.6\) \\
 & & V & \(19.1 \pm 1.6\) & \(381.7 \pm 35.3\) & \(0.963 \pm 0.006\) & \(291.7 \pm 37.8\) \\
\addlinespace[3pt]
\multirow{2}{*}{WP-Equi} & \multirow{2}{*}{\(12.78 \pm 4.05\)} & C & \(39.8 \pm 2.0\) & \(820.9 \pm 42.7\) & \(0.841 \pm 0.015\) & \(361.3 \pm 39.6\) \\
 & & V & \(23.1 \pm 3.2\) & \(472.2 \pm 45.1\) & \(0.946 \pm 0.015\) & \(256.3 \pm 38.9\) \\
\bottomrule
\end{tabular}
\end{table*}

\begin{table*}[p]
\centering
\caption{Toy-model performance for wallpaper group $\mathrm{pm}$ (1,000 structures, mean $\Delta E=-7.11$ eV atom$^{-1}$, energy SD 17.79 meV atom$^{-1}$).}
\label{tab:toy-model-pm}
\fontsize{8}{9.5}\selectfont
\setlength{\tabcolsep}{4pt}
\begin{tabular}{@{}lcccccc@{}}
\toprule
Model &
\shortstack{Energy\\MAE $\downarrow$\\(meV atom$^{-1}$)} &
Mode &
\shortstack{Rel.\ force\\$L_2$ $\downarrow$ (\%)} &
\shortstack{Force comp.\\MAE $\downarrow$\\(meV \AA$^{-1}$)} &
\shortstack{Force comp.\\$R^2$ $\uparrow$} &
\shortstack{Best\\epoch} \\
\midrule
\multicolumn{7}{l}{\textit{WallpaperNet models}} \\
\multirow{2}{*}{WP-Inv} & \multirow{2}{*}{\(10.01 \pm 2.22\)} & C & \(30.7 \pm 2.8\) & \(696.5 \pm 62.6\) & \(0.905 \pm 0.017\) & \(807.7 \pm 160.8\) \\
 & & V & \(14.3 \pm 2.6\) & \(306.6 \pm 47.2\) & \(0.979 \pm 0.007\) & \(416.0 \pm 177.8\) \\
\addlinespace[3pt]
\multirow{2}{*}{WP-InvExt} & \multirow{2}{*}{\(6.48 \pm 1.66\)} & C & \(23.6 \pm 1.7\) & \(537.9 \pm 39.4\) & \(0.944 \pm 0.008\) & \(469.7 \pm 115.5\) \\
 & & V & \(13.9 \pm 2.0\) & \(298.7 \pm 44.2\) & \(0.980 \pm 0.005\) & \(413.7 \pm 42.9\) \\
\addlinespace[3pt]
\multirow{2}{*}{WP-Equi} & \multirow{2}{*}{\(11.87 \pm 2.49\)} & C & \(31.9 \pm 0.9\) & \(723.2 \pm 36.3\) & \(0.898 \pm 0.006\) & \(383.0 \pm 88.5\) \\
 & & V & \(21.5 \pm 2.0\) & \(484.5 \pm 44.6\) & \(0.954 \pm 0.008\) & \(182.7 \pm 45.9\) \\
\bottomrule
\end{tabular}
\end{table*}

\begin{table*}[p]
\centering
\caption{Toy-model performance for wallpaper group $\mathrm{pg}$ (1,000 structures, mean $\Delta E=-7.49$ eV atom$^{-1}$, energy SD 13.53 meV atom$^{-1}$).}
\label{tab:toy-model-pg}
\fontsize{8}{9.5}\selectfont
\setlength{\tabcolsep}{4pt}
\begin{tabular}{@{}lcccccc@{}}
\toprule
Model &
\shortstack{Energy\\MAE $\downarrow$\\(meV atom$^{-1}$)} &
Mode &
\shortstack{Rel.\ force\\$L_2$ $\downarrow$ (\%)} &
\shortstack{Force comp.\\MAE $\downarrow$\\(meV \AA$^{-1}$)} &
\shortstack{Force comp.\\$R^2$ $\uparrow$} &
\shortstack{Best\\epoch} \\
\midrule
\multicolumn{7}{l}{\textit{WallpaperNet models}} \\
\multirow{2}{*}{WP-Inv} & \multirow{2}{*}{\(8.16 \pm 2.01\)} & C & \(32.7 \pm 1.3\) & \(739.6 \pm 29.8\) & \(0.893 \pm 0.008\) & \(573.3 \pm 183.5\) \\
 & & V & \(16.0 \pm 1.4\) & \(340.1 \pm 16.9\) & \(0.974 \pm 0.005\) & \(354.7 \pm 41.9\) \\
\addlinespace[3pt]
\multirow{2}{*}{WP-InvExt} & \multirow{2}{*}{\(6.68 \pm 1.41\)} & C & \(30.3 \pm 2.2\) & \(669.1 \pm 70.8\) & \(0.908 \pm 0.013\) & \(496.3 \pm 177.0\) \\
 & & V & \(16.3 \pm 0.7\) & \(356.6 \pm 9.2\) & \(0.973 \pm 0.002\) & \(352.0 \pm 92.0\) \\
\addlinespace[3pt]
\multirow{2}{*}{WP-Equi} & \multirow{2}{*}{\(7.74 \pm 0.98\)} & C & \(33.7 \pm 4.0\) & \(768.8 \pm 65.3\) & \(0.885 \pm 0.026\) & \(537.7 \pm 42.2\) \\
 & & V & \(24.6 \pm 2.8\) & \(540.9 \pm 59.0\) & \(0.939 \pm 0.014\) & \(296.3 \pm 31.5\) \\
\bottomrule
\end{tabular}
\end{table*}

\begin{table*}[p]
\centering
\caption{Toy-model performance for wallpaper group $\mathrm{cm}$ (1,000 structures, mean $\Delta E=-7.01$ eV atom$^{-1}$, energy SD 9.58 meV atom$^{-1}$).}
\label{tab:toy-model-cm}
\fontsize{8}{9.5}\selectfont
\setlength{\tabcolsep}{4pt}
\begin{tabular}{@{}lcccccc@{}}
\toprule
Model &
\shortstack{Energy\\MAE $\downarrow$\\(meV atom$^{-1}$)} &
Mode &
\shortstack{Rel.\ force\\$L_2$ $\downarrow$ (\%)} &
\shortstack{Force comp.\\MAE $\downarrow$\\(meV \AA$^{-1}$)} &
\shortstack{Force comp.\\$R^2$ $\uparrow$} &
\shortstack{Best\\epoch} \\
\midrule
\multicolumn{7}{l}{\textit{WallpaperNet models}} \\
\multirow{2}{*}{WP-Inv} & \multirow{2}{*}{\(6.21 \pm 1.86\)} & C & \(30.7 \pm 1.1\) & \(680.5 \pm 34.4\) & \(0.906 \pm 0.007\) & \(549.7 \pm 34.9\) \\
 & & V & \(11.8 \pm 1.1\) & \(247.0 \pm 18.1\) & \(0.986 \pm 0.003\) & \(495.7 \pm 130.2\) \\
\addlinespace[3pt]
\multirow{2}{*}{WP-InvExt} & \multirow{2}{*}{\(3.40 \pm 0.55\)} & C & \(24.6 \pm 0.8\) & \(535.2 \pm 10.7\) & \(0.940 \pm 0.004\) & \(541.7 \pm 143.2\) \\
 & & V & \(12.5 \pm 1.0\) & \(263.7 \pm 24.8\) & \(0.984 \pm 0.002\) & \(579.7 \pm 77.2\) \\
\addlinespace[3pt]
\multirow{2}{*}{WP-Equi} & \multirow{2}{*}{\(8.29 \pm 6.39\)} & C & \(33.3 \pm 0.6\) & \(731.2 \pm 28.2\) & \(0.889 \pm 0.004\) & \(488.3 \pm 24.4\) \\
 & & V & \(20.5 \pm 0.7\) & \(439.5 \pm 4.9\) & \(0.958 \pm 0.003\) & \(247.3 \pm 41.5\) \\
\bottomrule
\end{tabular}
\end{table*}

\begin{table*}[p]
\centering
\caption{Toy-model performance for wallpaper group $\mathrm{p3}$ (1,000 structures, mean $\Delta E=-6.78$ eV atom$^{-1}$, energy SD 21.43 meV atom$^{-1}$).}
\label{tab:toy-model-p3}
\fontsize{8}{9.5}\selectfont
\setlength{\tabcolsep}{4pt}
\begin{tabular}{@{}lcccccc@{}}
\toprule
Model &
\shortstack{Energy\\MAE $\downarrow$\\(meV atom$^{-1}$)} &
Mode &
\shortstack{Rel.\ force\\$L_2$ $\downarrow$ (\%)} &
\shortstack{Force comp.\\MAE $\downarrow$\\(meV \AA$^{-1}$)} &
\shortstack{Force comp.\\$R^2$ $\uparrow$} &
\shortstack{Best\\epoch} \\
\midrule
\multicolumn{7}{l}{\textit{WallpaperNet models}} \\
\multirow{2}{*}{WP-Inv} & \multirow{2}{*}{\(10.13 \pm 1.09\)} & C & \(30.8 \pm 1.0\) & \(725.9 \pm 22.5\) & \(0.905 \pm 0.006\) & \(910.7 \pm 131.8\) \\
 & & V & \(14.1 \pm 0.7\) & \(315.9 \pm 13.3\) & \(0.980 \pm 0.002\) & \(358.7 \pm 45.6\) \\
\addlinespace[3pt]
\multirow{2}{*}{WP-InvExt} & \multirow{2}{*}{\(7.73 \pm 1.22\)} & C & \(25.6 \pm 4.0\) & \(564.9 \pm 67.8\) & \(0.933 \pm 0.020\) & \(476.7 \pm 336.2\) \\
 & & V & \(14.1 \pm 1.5\) & \(311.9 \pm 26.3\) & \(0.980 \pm 0.004\) & \(468.3 \pm 156.5\) \\
\addlinespace[3pt]
\multirow{2}{*}{WP-Equi} & \multirow{2}{*}{\(11.05 \pm 2.28\)} & C & \(30.6 \pm 0.7\) & \(715.3 \pm 2.7\) & \(0.906 \pm 0.004\) & \(878.7 \pm 94.1\) \\
 & & V & \(16.1 \pm 1.5\) & \(367.6 \pm 25.9\) & \(0.974 \pm 0.005\) & \(246.3 \pm 24.8\) \\
\bottomrule
\end{tabular}
\end{table*}

\begin{table*}[p]
\centering
\caption{Toy-model performance for wallpaper group $\mathrm{p4}$ (1,000 structures, mean $\Delta E=-6.77$ eV atom$^{-1}$, energy SD 21.95 meV atom$^{-1}$).}
\label{tab:toy-model-p4}
\fontsize{8}{9.5}\selectfont
\setlength{\tabcolsep}{4pt}
\begin{tabular}{@{}lcccccc@{}}
\toprule
Model &
\shortstack{Energy\\MAE $\downarrow$\\(meV atom$^{-1}$)} &
Mode &
\shortstack{Rel.\ force\\$L_2$ $\downarrow$ (\%)} &
\shortstack{Force comp.\\MAE $\downarrow$\\(meV \AA$^{-1}$)} &
\shortstack{Force comp.\\$R^2$ $\uparrow$} &
\shortstack{Best\\epoch} \\
\midrule
\multicolumn{7}{l}{\textit{WallpaperNet models}} \\
\multirow{2}{*}{WP-Inv} & \multirow{2}{*}{\(10.45 \pm 1.52\)} & C & \(33.7 \pm 1.4\) & \(778.9 \pm 34.3\) & \(0.886 \pm 0.010\) & \(541.3 \pm 110.9\) \\
 & & V & \(13.4 \pm 1.1\) & \(280.8 \pm 29.9\) & \(0.982 \pm 0.003\) & \(419.3 \pm 56.6\) \\
\addlinespace[3pt]
\multirow{2}{*}{WP-InvExt} & \multirow{2}{*}{\(4.41 \pm 0.41\)} & C & \(21.3 \pm 0.5\) & \(466.1 \pm 22.7\) & \(0.955 \pm 0.002\) & \(726.7 \pm 263.4\) \\
 & & V & \(14.4 \pm 1.8\) & \(310.9 \pm 45.7\) & \(0.979 \pm 0.005\) & \(375.3 \pm 26.6\) \\
\addlinespace[3pt]
\multirow{2}{*}{WP-Equi} & \multirow{2}{*}{\(12.66 \pm 4.98\)} & C & \(32.3 \pm 3.0\) & \(718.7 \pm 78.5\) & \(0.895 \pm 0.019\) & \(547.7 \pm 213.3\) \\
 & & V & \(15.2 \pm 0.9\) & \(333.1 \pm 20.0\) & \(0.977 \pm 0.003\) & \(337.7 \pm 29.1\) \\
\bottomrule
\end{tabular}
\end{table*}

\begin{table*}[p]
\centering
\caption{Toy-model performance for wallpaper group $\mathrm{pmm}$ (1,000 structures, mean $\Delta E=-6.77$ eV atom$^{-1}$, energy SD 21.78 meV atom$^{-1}$).}
\label{tab:toy-model-pmm}
\fontsize{8}{9.5}\selectfont
\setlength{\tabcolsep}{4pt}
\begin{tabular}{@{}lcccccc@{}}
\toprule
Model &
\shortstack{Energy\\MAE $\downarrow$\\(meV atom$^{-1}$)} &
Mode &
\shortstack{Rel.\ force\\$L_2$ $\downarrow$ (\%)} &
\shortstack{Force comp.\\MAE $\downarrow$\\(meV \AA$^{-1}$)} &
\shortstack{Force comp.\\$R^2$ $\uparrow$} &
\shortstack{Best\\epoch} \\
\midrule
\multicolumn{7}{l}{\textit{WallpaperNet models}} \\
\multirow{2}{*}{WP-Inv} & \multirow{2}{*}{\(9.58 \pm 0.94\)} & C & \(30.7 \pm 0.5\) & \(698.0 \pm 23.1\) & \(0.906 \pm 0.003\) & \(695.7 \pm 266.6\) \\
 & & V & \(16.2 \pm 0.7\) & \(332.0 \pm 12.7\) & \(0.974 \pm 0.002\) & \(593.3 \pm 261.2\) \\
\addlinespace[3pt]
\multirow{2}{*}{WP-InvExt} & \multirow{2}{*}{\(5.81 \pm 1.04\)} & C & \(24.5 \pm 4.3\) & \(520.4 \pm 84.3\) & \(0.939 \pm 0.022\) & \(713.7 \pm 267.5\) \\
 & & V & \(15.8 \pm 0.4\) & \(320.8 \pm 7.9\) & \(0.975 \pm 0.001\) & \(463.3 \pm 89.8\) \\
\addlinespace[3pt]
\multirow{2}{*}{WP-Equi} & \multirow{2}{*}{\(15.18 \pm 1.77\)} & C & \(46.7 \pm 19.6\) & \(1032.9 \pm 384.1\) & \(0.756 \pm 0.205\) & \(531.7 \pm 220.9\) \\
 & & V & \(18.4 \pm 0.8\) & \(382.2 \pm 12.9\) & \(0.966 \pm 0.003\) & \(306.0 \pm 60.9\) \\
\bottomrule
\end{tabular}
\end{table*}

\begin{table*}[p]
\centering
\caption{Toy-model performance for wallpaper group $\mathrm{pmg}$ (1,000 structures, mean $\Delta E=-7.10$ eV atom$^{-1}$, energy SD 14.09 meV atom$^{-1}$).}
\label{tab:toy-model-pmg}
\fontsize{8}{9.5}\selectfont
\setlength{\tabcolsep}{4pt}
\begin{tabular}{@{}lcccccc@{}}
\toprule
Model &
\shortstack{Energy\\MAE $\downarrow$\\(meV atom$^{-1}$)} &
Mode &
\shortstack{Rel.\ force\\$L_2$ $\downarrow$ (\%)} &
\shortstack{Force comp.\\MAE $\downarrow$\\(meV \AA$^{-1}$)} &
\shortstack{Force comp.\\$R^2$ $\uparrow$} &
\shortstack{Best\\epoch} \\
\midrule
\multicolumn{7}{l}{\textit{WallpaperNet models}} \\
\multirow{2}{*}{WP-Inv} & \multirow{2}{*}{\(9.15 \pm 1.18\)} & C & \(31.3 \pm 2.1\) & \(717.3 \pm 46.3\) & \(0.902 \pm 0.013\) & \(553.7 \pm 381.7\) \\
 & & V & \(12.9 \pm 4.8\) & \(264.4 \pm 83.4\) & \(0.982 \pm 0.014\) & \(466.0 \pm 187.3\) \\
\addlinespace[3pt]
\multirow{2}{*}{WP-InvExt} & \multirow{2}{*}{\(4.43 \pm 1.32\)} & C & \(22.8 \pm 3.1\) & \(495.8 \pm 65.7\) & \(0.948 \pm 0.015\) & \(571.7 \pm 123.7\) \\
 & & V & \(12.8 \pm 1.8\) & \(260.8 \pm 25.2\) & \(0.983 \pm 0.005\) & \(535.7 \pm 196.1\) \\
\addlinespace[3pt]
\multirow{2}{*}{WP-Equi} & \multirow{2}{*}{\(8.93 \pm 1.21\)} & C & \(33.8 \pm 5.0\) & \(764.5 \pm 128.4\) & \(0.884 \pm 0.032\) & \(586.3 \pm 186.6\) \\
 & & V & \(16.4 \pm 3.4\) & \(340.9 \pm 48.5\) & \(0.972 \pm 0.012\) & \(398.0 \pm 64.7\) \\
\bottomrule
\end{tabular}
\end{table*}

\begin{table*}[p]
\centering
\caption{Toy-model performance for wallpaper group $\mathrm{pgg}$ (1,000 structures, mean $\Delta E=-7.71$ eV atom$^{-1}$, energy SD 15.18 meV atom$^{-1}$).}
\label{tab:toy-model-pgg}
\fontsize{8}{9.5}\selectfont
\setlength{\tabcolsep}{4pt}
\begin{tabular}{@{}lcccccc@{}}
\toprule
Model &
\shortstack{Energy\\MAE $\downarrow$\\(meV atom$^{-1}$)} &
Mode &
\shortstack{Rel.\ force\\$L_2$ $\downarrow$ (\%)} &
\shortstack{Force comp.\\MAE $\downarrow$\\(meV \AA$^{-1}$)} &
\shortstack{Force comp.\\$R^2$ $\uparrow$} &
\shortstack{Best\\epoch} \\
\midrule
\multicolumn{7}{l}{\textit{WallpaperNet models}} \\
\multirow{2}{*}{WP-Inv} & \multirow{2}{*}{\(10.31 \pm 2.92\)} & C & \(32.7 \pm 5.6\) & \(754.7 \pm 134.1\) & \(0.891 \pm 0.035\) & \(423.3 \pm 19.6\) \\
 & & V & \(13.6 \pm 1.5\) & \(290.1 \pm 33.9\) & \(0.981 \pm 0.004\) & \(368.7 \pm 31.0\) \\
\addlinespace[3pt]
\multirow{2}{*}{WP-InvExt} & \multirow{2}{*}{\(6.62 \pm 0.24\)} & C & \(24.1 \pm 1.2\) & \(529.6 \pm 35.4\) & \(0.942 \pm 0.006\) & \(595.0 \pm 317.8\) \\
 & & V & \(13.2 \pm 2.9\) & \(285.3 \pm 59.0\) & \(0.982 \pm 0.007\) & \(456.0 \pm 46.5\) \\
\addlinespace[3pt]
\multirow{2}{*}{WP-Equi} & \multirow{2}{*}{\(9.39 \pm 1.96\)} & C & \(33.8 \pm 2.8\) & \(779.0 \pm 64.8\) & \(0.885 \pm 0.018\) & \(571.0 \pm 76.0\) \\
 & & V & \(17.7 \pm 2.6\) & \(385.1 \pm 56.2\) & \(0.968 \pm 0.009\) & \(271.7 \pm 50.5\) \\
\bottomrule
\end{tabular}
\end{table*}

\begin{table*}[p]
\centering
\caption{Toy-model performance for wallpaper group $\mathrm{cmm}$ (1,000 structures, mean $\Delta E=-7.07$ eV atom$^{-1}$, energy SD 9.70 meV atom$^{-1}$).}
\label{tab:toy-model-cmm}
\fontsize{8}{9.5}\selectfont
\setlength{\tabcolsep}{4pt}
\begin{tabular}{@{}lcccccc@{}}
\toprule
Model &
\shortstack{Energy\\MAE $\downarrow$\\(meV atom$^{-1}$)} &
Mode &
\shortstack{Rel.\ force\\$L_2$ $\downarrow$ (\%)} &
\shortstack{Force comp.\\MAE $\downarrow$\\(meV \AA$^{-1}$)} &
\shortstack{Force comp.\\$R^2$ $\uparrow$} &
\shortstack{Best\\epoch} \\
\midrule
\multicolumn{7}{l}{\textit{WallpaperNet models}} \\
\multirow{2}{*}{WP-Inv} & \multirow{2}{*}{\(11.90 \pm 1.28\)} & C & \(32.3 \pm 1.1\) & \(723.7 \pm 10.1\) & \(0.896 \pm 0.007\) & \(797.7 \pm 219.8\) \\
 & & V & \(12.3 \pm 0.5\) & \(261.9 \pm 11.6\) & \(0.985 \pm 0.001\) & \(569.7 \pm 191.3\) \\
\addlinespace[3pt]
\multirow{2}{*}{WP-InvExt} & \multirow{2}{*}{\(\mathbf{2.98 \pm 0.59}\)} & C & \(21.7 \pm 2.5\) & \(480.8 \pm 59.1\) & \(0.952 \pm 0.011\) & \(428.0 \pm 50.4\) \\
 & & V & \(12.1 \pm 0.8\) & \(257.0 \pm 25.3\) & \(0.985 \pm 0.002\) & \(515.7 \pm 151.3\) \\
\addlinespace[3pt]
\multirow{2}{*}{WP-Equi} & \multirow{2}{*}{\(13.38 \pm 3.62\)} & C & \(33.1 \pm 0.4\) & \(758.1 \pm 13.5\) & \(0.890 \pm 0.003\) & \(604.0 \pm 126.2\) \\
 & & V & \(15.2 \pm 1.6\) & \(334.3 \pm 43.2\) & \(0.977 \pm 0.005\) & \(327.0 \pm 35.5\) \\
\bottomrule
\end{tabular}
\end{table*}

\begin{table*}[p]
\centering
\caption{Toy-model performance for wallpaper group $\mathrm{p6}$ (1,000 structures, mean $\Delta E=-7.34$ eV atom$^{-1}$, energy SD 16.91 meV atom$^{-1}$).}
\label{tab:toy-model-p6}
\fontsize{8}{9.5}\selectfont
\setlength{\tabcolsep}{4pt}
\begin{tabular}{@{}lcccccc@{}}
\toprule
Model &
\shortstack{Energy\\MAE $\downarrow$\\(meV atom$^{-1}$)} &
Mode &
\shortstack{Rel.\ force\\$L_2$ $\downarrow$ (\%)} &
\shortstack{Force comp.\\MAE $\downarrow$\\(meV \AA$^{-1}$)} &
\shortstack{Force comp.\\$R^2$ $\uparrow$} &
\shortstack{Best\\epoch} \\
\midrule
\multicolumn{7}{l}{\textit{WallpaperNet models}} \\
\multirow{2}{*}{WP-Inv} & \multirow{2}{*}{\(9.61 \pm 0.55\)} & C & \(31.2 \pm 0.9\) & \(709.5 \pm 9.0\) & \(0.903 \pm 0.006\) & \(883.3 \pm 21.1\) \\
 & & V & \(13.5 \pm 0.9\) & \(294.0 \pm 4.7\) & \(0.982 \pm 0.002\) & \(409.0 \pm 91.0\) \\
\addlinespace[3pt]
\multirow{2}{*}{WP-InvExt} & \multirow{2}{*}{\(7.31 \pm 2.11\)} & C & \(23.5 \pm 1.0\) & \(525.2 \pm 27.2\) & \(0.945 \pm 0.005\) & \(554.3 \pm 110.6\) \\
 & & V & \(13.7 \pm 2.0\) & \(301.2 \pm 24.6\) & \(0.981 \pm 0.006\) & \(414.0 \pm 116.0\) \\
\addlinespace[3pt]
\multirow{2}{*}{WP-Equi} & \multirow{2}{*}{\(12.36 \pm 1.58\)} & C & \(33.8 \pm 4.9\) & \(772.5 \pm 123.7\) & \(0.884 \pm 0.034\) & \(543.7 \pm 253.1\) \\
 & & V & \(14.1 \pm 2.7\) & \(306.5 \pm 47.4\) & \(0.980 \pm 0.008\) & \(336.7 \pm 59.1\) \\
\bottomrule
\end{tabular}
\end{table*}

\begin{table*}[p]
\centering
\caption{Toy-model performance for wallpaper group $\mathrm{p3m1}$ (1,000 structures, mean $\Delta E=-7.34$ eV atom$^{-1}$, energy SD 15.43 meV atom$^{-1}$).}
\label{tab:toy-model-p3m1}
\fontsize{8}{9.5}\selectfont
\setlength{\tabcolsep}{4pt}
\begin{tabular}{@{}lcccccc@{}}
\toprule
Model &
\shortstack{Energy\\MAE $\downarrow$\\(meV atom$^{-1}$)} &
Mode &
\shortstack{Rel.\ force\\$L_2$ $\downarrow$ (\%)} &
\shortstack{Force comp.\\MAE $\downarrow$\\(meV \AA$^{-1}$)} &
\shortstack{Force comp.\\$R^2$ $\uparrow$} &
\shortstack{Best\\epoch} \\
\midrule
\multicolumn{7}{l}{\textit{WallpaperNet models}} \\
\multirow{2}{*}{WP-Inv} & \multirow{2}{*}{\(10.39 \pm 1.18\)} & C & \(29.8 \pm 4.5\) & \(688.5 \pm 112.8\) & \(0.910 \pm 0.027\) & \(751.7 \pm 211.9\) \\
 & & V & \(13.7 \pm 2.8\) & \(286.9 \pm 53.8\) & \(0.981 \pm 0.008\) & \(449.0 \pm 65.3\) \\
\addlinespace[3pt]
\multirow{2}{*}{WP-InvExt} & \multirow{2}{*}{\(5.57 \pm 0.56\)} & C & \(22.8 \pm 2.1\) & \(510.4 \pm 49.5\) & \(0.948 \pm 0.010\) & \(659.0 \pm 315.1\) \\
 & & V & \(13.6 \pm 2.3\) & \(292.9 \pm 37.5\) & \(0.981 \pm 0.007\) & \(519.0 \pm 155.9\) \\
\addlinespace[3pt]
\multirow{2}{*}{WP-Equi} & \multirow{2}{*}{\(10.71 \pm 2.33\)} & C & \(35.3 \pm 4.5\) & \(830.0 \pm 97.6\) & \(0.874 \pm 0.032\) & \(560.0 \pm 268.3\) \\
 & & V & \(14.9 \pm 2.0\) & \(329.1 \pm 28.5\) & \(0.978 \pm 0.006\) & \(351.3 \pm 33.9\) \\
\bottomrule
\end{tabular}
\end{table*}

\begin{table*}[p]
\centering
\caption{Toy-model performance for wallpaper group $\mathrm{p31m}$ (1,000 structures, mean $\Delta E=-6.22$ eV atom$^{-1}$, energy SD 32.38 meV atom$^{-1}$).}
\label{tab:toy-model-p31m}
\fontsize{8}{9.5}\selectfont
\setlength{\tabcolsep}{4pt}
\begin{tabular}{@{}lcccccc@{}}
\toprule
Model &
\shortstack{Energy\\MAE $\downarrow$\\(meV atom$^{-1}$)} &
Mode &
\shortstack{Rel.\ force\\$L_2$ $\downarrow$ (\%)} &
\shortstack{Force comp.\\MAE $\downarrow$\\(meV \AA$^{-1}$)} &
\shortstack{Force comp.\\$R^2$ $\uparrow$} &
\shortstack{Best\\epoch} \\
\midrule
\multicolumn{7}{l}{\textit{WallpaperNet models}} \\
\multirow{2}{*}{WP-Inv} & \multirow{2}{*}{\(8.32 \pm 2.25\)} & C & \(27.9 \pm 3.2\) & \(613.7 \pm 78.3\) & \(0.921 \pm 0.017\) & \(796.7 \pm 175.0\) \\
 & & V & \(14.5 \pm 1.0\) & \(305.2 \pm 19.8\) & \(0.979 \pm 0.003\) & \(337.7 \pm 47.6\) \\
\addlinespace[3pt]
\multirow{2}{*}{WP-InvExt} & \multirow{2}{*}{\(7.68 \pm 1.36\)} & C & \(27.4 \pm 3.6\) & \(568.1 \pm 48.5\) & \(0.924 \pm 0.019\) & \(556.3 \pm 193.3\) \\
 & & V & \(15.2 \pm 1.7\) & \(309.9 \pm 40.0\) & \(0.977 \pm 0.005\) & \(407.0 \pm 56.0\) \\
\addlinespace[3pt]
\multirow{2}{*}{WP-Equi} & \multirow{2}{*}{\(9.47 \pm 0.92\)} & C & \(32.6 \pm 9.0\) & \(697.6 \pm 159.9\) & \(0.888 \pm 0.061\) & \(507.7 \pm 30.9\) \\
 & & V & \(16.1 \pm 1.0\) & \(335.6 \pm 23.7\) & \(0.974 \pm 0.003\) & \(336.3 \pm 39.3\) \\
\bottomrule
\end{tabular}
\end{table*}

\begin{table*}[p]
\centering
\caption{Toy-model performance for wallpaper group $\mathrm{p4m}$ (1,000 structures, mean $\Delta E=-6.77$ eV atom$^{-1}$, energy SD 21.66 meV atom$^{-1}$).}
\label{tab:toy-model-p4m}
\fontsize{8}{9.5}\selectfont
\setlength{\tabcolsep}{4pt}
\begin{tabular}{@{}lcccccc@{}}
\toprule
Model &
\shortstack{Energy\\MAE $\downarrow$\\(meV atom$^{-1}$)} &
Mode &
\shortstack{Rel.\ force\\$L_2$ $\downarrow$ (\%)} &
\shortstack{Force comp.\\MAE $\downarrow$\\(meV \AA$^{-1}$)} &
\shortstack{Force comp.\\$R^2$ $\uparrow$} &
\shortstack{Best\\epoch} \\
\midrule
\multicolumn{7}{l}{\textit{WallpaperNet models}} \\
\multirow{2}{*}{WP-Inv} & \multirow{2}{*}{\(10.47 \pm 1.72\)} & C & \(30.1 \pm 2.6\) & \(708.2 \pm 55.7\) & \(0.909 \pm 0.015\) & \(618.7 \pm 169.2\) \\
 & & V & \(12.1 \pm 1.6\) & \(272.2 \pm 30.3\) & \(0.985 \pm 0.004\) & \(415.7 \pm 58.6\) \\
\addlinespace[3pt]
\multirow{2}{*}{WP-InvExt} & \multirow{2}{*}{\(5.70 \pm 1.63\)} & C & \(21.4 \pm 2.9\) & \(475.9 \pm 58.7\) & \(0.953 \pm 0.013\) & \(429.7 \pm 44.6\) \\
 & & V & \(13.2 \pm 1.6\) & \(291.0 \pm 26.5\) & \(0.982 \pm 0.004\) & \(686.0 \pm 110.1\) \\
\addlinespace[3pt]
\multirow{2}{*}{WP-Equi} & \multirow{2}{*}{\(10.88 \pm 0.81\)} & C & \(31.6 \pm 2.7\) & \(716.5 \pm 74.9\) & \(0.900 \pm 0.017\) & \(723.7 \pm 253.0\) \\
 & & V & \(13.8 \pm 1.1\) & \(311.5 \pm 10.5\) & \(0.981 \pm 0.003\) & \(489.7 \pm 159.8\) \\
\bottomrule
\end{tabular}
\end{table*}

\begin{table*}[p]
\centering
\caption{Toy-model performance for wallpaper group $\mathrm{p4g}$ (1,000 structures, mean $\Delta E=-6.92$ eV atom$^{-1}$, energy SD 15.21 meV atom$^{-1}$).}
\label{tab:toy-model-p4g}
\fontsize{8}{9.5}\selectfont
\setlength{\tabcolsep}{4pt}
%
\end{table*}

\begin{table*}[p]
\centering
\caption{Toy-model performance for wallpaper group $\mathrm{p6m}$ (1,000 structures, mean $\Delta E=-7.34$ eV atom$^{-1}$, energy SD 15.73 meV atom$^{-1}$).}
\label{tab:toy-model-p6m}
\fontsize{8}{9.5}\selectfont
\setlength{\tabcolsep}{4pt}
%
\end{table*}

\clearpage
\section{Molecule--graphene system results}
\label{sec:mol-grap-results}
Table~\ref{tab:real_system_test_full} reports the complete test-set results
averaged over the three random seeds, including the zero-force score, with the
C and V training modes in separate rows.

\begin{table*}[htbp]
\centering
\caption{
Complete test-set results of the seven wallpaper-symmetry models and three
full-cell baselines on graphene, including the zero-force score. Energy metrics are taken from the combined experiments,
while force metrics are reported in separate C and V rows, where C denotes force-weighted
conservative training and V denotes direct vector-head training.
Values are means and sample standard deviations over three random seeds.
Lower is better for the error metrics, and higher is better for the zero-force
score and $R^2$.
}
\label{tab:real_system_test_full}
\small
\setlength{\tabcolsep}{3pt}
\begin{tabular}{@{}lcccccc@{}}
\toprule
Model &
\shortstack{Energy\\MAE $\downarrow$\\(meV atom$^{-1}$)} &
Mode &
\shortstack{Rel.\ force\\$L_2$ $\downarrow$ (\%)} &
\shortstack{Zero-force\\score $\uparrow$ (\%)} &
\shortstack{Force comp.\\MAE $\downarrow$\\(meV\,$\mathring{\mathrm A}^{-1}$)} &
\shortstack{Force comp.\\$R^2$ $\uparrow$} \\
\midrule
\multicolumn{7}{l}{\textit{Wallpaper-symmetry models}} \\
\multirow{2}{*}{WP-Inv} & \multirow{2}{*}{$\mathbf{1.73 \pm 0.02}$} & C & $80.79 \pm 1.35$ & $34.72 \pm 2.19$ & $139.9 \pm 0.6$ & $0.35 \pm 0.02$ \\
 & & V & $41.35 \pm 7.79$ & $82.50 \pm 6.64$ & $75.0 \pm 16.3$ & $0.82 \pm 0.07$ \\
\addlinespace[3pt]
\multirow{2}{*}{WP-InvEq} & \multirow{2}{*}{$4.75 \pm 5.09$} & C & $78.68 \pm 1.15$ & $38.08 \pm 1.81$ & $137.1 \pm 2.0$ & $0.38 \pm 0.02$ \\
 & & V & $25.67 \pm 0.63$ & $93.41 \pm 0.33$ & $43.5 \pm 1.0$ & $0.93 \pm 0.00$ \\
\addlinespace[3pt]
\multirow{2}{*}{WP-Equi} & \multirow{2}{*}{$1.90 \pm 0.17$} & C & $80.59 \pm 0.94$ & $35.05 \pm 1.51$ & $140.7 \pm 1.4$ & $0.35 \pm 0.02$ \\
 & & V & $27.95 \pm 0.80$ & $92.18 \pm 0.44$ & $48.8 \pm 0.9$ & $0.92 \pm 0.00$ \\
\addlinespace[3pt]
\multirow{2}{*}{WP-VecInv} & \multirow{2}{*}{$2.24 \pm 0.06$} & C & $62.49 \pm 15.95$ & $59.26 \pm 21.39$ & $109.5 \pm 25.4$ & $0.59 \pm 0.21$ \\
 & & V & $26.62 \pm 0.93$ & $92.91 \pm 0.50$ & $44.8 \pm 0.8$ & $0.93 \pm 0.01$ \\
\addlinespace[3pt]
\multirow{2}{*}{WP-Reg} & \multirow{2}{*}{$2.09 \pm 0.12$} & C & $30.70 \pm 1.90$ & $90.55 \pm 1.19$ & $54.0 \pm 2.7$ & $0.91 \pm 0.01$ \\
 & & V & $27.15 \pm 1.07$ & $92.62 \pm 0.58$ & $45.7 \pm 1.8$ & $0.93 \pm 0.01$ \\
\addlinespace[3pt]
\multirow{2}{*}{WP-RegTP} & \multirow{2}{*}{$1.86 \pm 0.24$} & C & $30.49 \pm 1.33$ & $90.69 \pm 0.82$ & $52.7 \pm 2.3$ & $0.91 \pm 0.01$ \\
 & & V & $26.21 \pm 1.14$ & $93.12 \pm 0.60$ & $44.6 \pm 1.1$ & $0.93 \pm 0.01$ \\
\addlinespace[3pt]
\multirow{2}{*}{WP-O3} & \multirow{2}{*}{$1.80 \pm 0.08$} & C & $\mathbf{19.78 \pm 0.70}$ & $\mathbf{96.09 \pm 0.28}$ & $\mathbf{32.9 \pm 0.9}$ & $\mathbf{0.96 \pm 0.00}$ \\
 & & V & $\mathbf{19.22 \pm 0.75}$ & $\mathbf{96.30 \pm 0.29}$ & $\mathbf{31.3 \pm 0.2}$ & $\mathbf{0.96 \pm 0.00}$ \\
\midrule
\multicolumn{7}{l}{\textit{Full-cell baselines}} \\
Full cell & \multirow{2}{*}{$2.58 \pm 0.23$} & C & $67.51 \pm 1.03$ & $54.42 \pm 1.39$ & $119.9 \pm 2.5$ & $0.54 \pm 0.01$ \\
invariant GNN & & V & $47.22 \pm 6.46$ & $77.42 \pm 5.93$ & $83.6 \pm 12.2$ & $0.77 \pm 0.06$ \\
\addlinespace[3pt]
Full cell e3nn & \multirow{2}{*}{$1.92 \pm 0.07$} & C & $25.89 \pm 0.89$ & $93.29 \pm 0.46$ & $44.7 \pm 2.1$ & $0.93 \pm 0.00$ \\
$\ell_{\max}=1$ & & V & $25.31 \pm 3.50$ & $93.51 \pm 1.83$ & $43.9 \pm 7.2$ & $0.94 \pm 0.02$ \\
\addlinespace[3pt]
Full cell e3nn & \multirow{2}{*}{$2.04 \pm 0.22$} & C & $21.43 \pm 0.75$ & $95.40 \pm 0.32$ & $36.2 \pm 1.2$ & $0.95 \pm 0.00$ \\
$\ell_{\max}=2$ & & V & $25.55 \pm 5.82$ & $93.24 \pm 2.81$ & $45.0 \pm 11.1$ & $0.93 \pm 0.03$ \\
\bottomrule
\end{tabular}
\end{table*}

\subsection{Discussion of the real-system results}
\label{sec:real-system-discussion}

The real-system experiments address three related questions, summarized in Table~\ref{tab:real_system_test}: how accurately the models predict the total energy, how accurately they predict forces directly, and whether comparable forces can be recovered conservatively as derivatives of a learned energy. The energy values, conservative-force metrics, and direct-force metrics therefore correspond to three different training objectives. The force-weighted conservative objective and the per-seed results are given in Appendix~\ref{sec:real-system-per-seed}.

For the first task, training is dominated by the total-energy objective, with forces included as an auxiliary target. Under this setting, all WallpaperNet models remain far below the commonly used chemical-accuracy threshold of \(43\,\mathrm{meV\,atom^{-1}}\). WP-Inv gives the lowest energy MAE at \(1.73\,\mathrm{meV\,atom^{-1}}\), followed by WP-O3 at \(1.80\,\mathrm{meV\,atom^{-1}}\), WP-RegTP at \(1.86\,\mathrm{meV\,atom^{-1}}\), and WP-Equi at \(1.90\,\mathrm{meV\,atom^{-1}}\). These values are also below the full-cell e3nn baselines at \(1.92\) and \(2.04\,\mathrm{meV\,atom^{-1}}\) for \(\ell_{\max}=1\) and \(2\), respectively. The exception is WP-InvEq, which reaches \(4.75\pm5.09\,\mathrm{meV\,atom^{-1}}\) and is therefore considerably less stable across seeds. Overall, the energy results show that the additional equivariant complexity is not required for this scalar target: the simplest invariant WP-Inv model gives the lowest mean error.

The same energy-dominated objective does not, however, provide useful conservative forces. The full-system energy residual has a much larger numerical scale than the component-wise force loss, while force supervision is restricted to the non-frozen adsorbate atoms. With equal loss weights, optimization therefore primarily reduces the energy error and the resulting energy gradients tend toward small forces. This motivates treating force prediction as a separate problem. The toy experiments already showed that direct vector-head training generally gives substantially more accurate forces than differentiating the energy, so we first evaluate the same strategy on the real system.

For direct force prediction, WP-O3 gives the lowest error of all ten models, with a relative force error of \(19.22\%\), a component MAE of \(31.3\,\mathrm{meV\,\mathring{A}^{-1}}\), and \(R^2=0.96\). The full-cell e3nn baselines reach \(25.31\%\) and \(25.55\%\) relative error for \(\ell_{\max}=1\) and \(2\), respectively. Among the other WallpaperNet models, WP-InvEq, WP-VecInv, WP-RegTP, and WP-Reg lie in a narrow range of \(25.67\)--\(27.15\%\), whereas WP-Equi reaches \(27.95\%\). WP-Inv performs substantially worse at \(41.35\%\), similar to the full-cell invariant GNN at \(47.22\%\). Thus, the model ordering for forces differs clearly from that for energies: WP-Inv gives the lowest energy error but one of the largest direct-force errors, whereas WP-O3 is competitive for energy and gives the lowest force error.

We next test whether accurate forces can also be obtained conservatively. To prevent the energy term from dominating the optimization, we rescale it using the empirical ratio between the force-MSE and energy-L1 scales and the mean adsorbate-to-full-system atom-count ratio, resulting in an energy weight of approximately \(1.22\times10^{-6}\) relative to a unit force weight. The latter atom-count factor is an additional weighting choice rather than a required normalization, since both losses are already mean-reduced.

The reweighted conservative results show a strong dependence on architecture. WP-O3 reaches \(19.78\%\) relative force error, \(32.9\,\mathrm{meV\,\mathring{A}^{-1}}\) component MAE, and \(R^2=0.96\), compared with \(19.22\%\), \(31.3\,\mathrm{meV\,\mathring{A}^{-1}}\), and \(R^2=0.96\) for its direct head. Thus, for WP-O3, deriving the forces from the energy introduces only a small loss in accuracy once the objective is appropriately balanced. The full-cell e3nn models also perform well conservatively, reaching relative errors of \(25.89\%\) for \(\ell_{\max}=1\) and \(21.43\%\) for \(\ell_{\max}=2\).

The same recovery is not observed for all architectures. WP-Reg and WP-RegTP reach conservative relative errors of \(30.70\%\) and \(30.49\%\), respectively, while WP-VecInv increases to \(62.49\pm15.95\%\). WP-Inv, WP-InvEq, and WP-Equi remain around \(79\)--\(81\%\), and the full-cell invariant GNN reaches \(67.51\%\). In particular, an equivariant force output alone is insufficient: WP-InvEq improves substantially over WP-Inv for direct force regression (\(25.67\%\) versus \(41.35\%\)) but gives almost the same poor conservative performance (\(78.68\%\) versus \(80.79\%\)). The conservative results therefore distinguish architectures that can produce an equivariant vector output from those that can encode sufficient directional information in the scalar energy surface itself. Among the tested models, WP-O3 does this most effectively.

Finally, WallpaperNet obtains these results on a substantially smaller message-passing graph. The WallpaperNet representation contains on average 25 nodes and 512 edges per system, whereas the full-cell representation contains 386 nodes and 22,929 edges, corresponding to approximately \(15\times\) more nodes and \(45\times\) more edges. The \(45\times\) edge reduction should not be interpreted as a \(45\times\) wall-clock speedup, since actual cost also depends on representation size, tensor-product operations, hardware, batch size, and training objective. Peak GPU memory and the accuracy--memory trade-off are reported in Appendix~\ref{sec:memory_footprint}. We additionally benchmark all ten architectures under both conservative and direct-force training at approximately \(90\%\) GPU-memory utilization. These maximum-batch throughput results are reported in Appendix~\ref{sec:real-system-max-batch}.

\paragraph{Additional observations on the averaged results.}
Beyond the discussion above, the averaged results in
Table~\ref{tab:real_system_test_full} show the following. WP-O3 reaches the
highest zero-force score in both training modes, \(96.09\%\) with conservative
and \(96.30\%\) with direct-head training. In direct-force training, the
full-cell invariant GNN is the weakest full-cell model, and WP-Inv reaches a
component MAE of \(75.0\,\mathrm{meV\,\mathring{A}^{-1}}\). The explicit
equivariant output maps of WP-InvEq reduce this to
\(43.5\,\mathrm{meV\,\mathring{A}^{-1}}\), making WP-InvEq the second-best
WallpaperNet model for direct-force component MAE despite being the least
stable model for energy prediction. WP-VecInv, WP-Reg, and WP-RegTP reach a
similar direct-force regime, with component MAEs between
\(44.6\) and \(45.7\,\mathrm{meV\,\mathring{A}^{-1}}\), so increasing the
complexity of the wallpaper-equivariant architecture does not by itself
guarantee improved accuracy. Finally, the contrast between the energy and the
conservative-force rankings indicates that accurate prediction of scalar
energy values and accurate prediction of the local derivatives of the energy
surface are related but not equivalent learning problems: an invariant
representation can be sufficient for the former, whereas the latter benefits
much more strongly from explicit directional equivariant features.

\subsection{Per-seed results and force-weighted conservative objective}
\label{sec:real-system-per-seed}

Tables~\ref{tab:real-system-test-seed-1},
\ref{tab:real-system-test-seed-24}, and
\ref{tab:real-system-test-seed-42} report the test-set results for the three
random seeds separately. Energy metrics are taken from the combined
experiments, whereas force metrics are taken from the direct force-head
experiments. The energy and force-head best epochs are reported as E/F,
respectively. For the real-system conservative force experiment, we found that the unweighted joint objective, \(\lambda_E=\lambda_F=1\), was strongly imbalanced because the numerical scale of the full-system energy residual was much larger than that of the component-wise molecular force MSE. In practice, this caused optimization to be dominated by the energy term and the predicted energy surface to become nearly flat with respect to the adsorbate coordinates, resulting in forces that collapsed toward zero.

We therefore introduced a force-weighted conservative objective. The energy coefficient was determined exclusively from the cached training split as
\begin{equation}
\lambda_E =
\frac{
\left\langle F_{\mathrm{mol}}^2\right\rangle
}{
\left\langle
\left|E_{\mathrm{total}}
-E_{\mathrm{atom-ref}}
-E_{\mathrm{graph-ref}}\right|
\right\rangle
}
\frac{\left\langle N_{\mathrm{mol}}\right\rangle}
{\left\langle N_{\mathrm{full}}\right\rangle},
\qquad
\lambda_F=1.
\end{equation}
The first ratio approximately matches the numerical scales of the energy-L1 and force-MSE terms. The additional atom-count factor accounts for the fact that the energy target corresponds to the complete system, whereas force supervision is restricted to the adsorbate atoms. Tables~\ref{tab:real-system-force-derived-seed-1},
\ref{tab:real-system-force-derived-seed-24}, and
\ref{tab:real-system-force-derived-seed-67} report the corresponding
force-weighted conservative experiments.
Across the three training splits, the resulting coefficients were
\(1.21915\times10^{-6}\), \(1.22801\times10^{-6}\), and
\(1.22606\times10^{-6}\), with a mean of \(1.22441\times10^{-6}\).
The force coefficient was fixed at \(\lambda_F=1\).

The force-weighted conservative experiments show that the poor force performance of the original \(1{:}1\) combined objective was primarily a loss-balancing problem rather than a fundamental limitation of energy-derived force prediction. In contrast to the fixed-size toy systems, the molecule--graphene dataset contains full cells ranging from \(158\) to \(699\) atoms, with a mean of \(386.825\) atoms and a standard deviation of \(51.593\). The corresponding total system energies range from \(-40{,}589.4\) to \(-9{,}197.05\,\mathrm{eV}\). After rescaling the energy contribution, several architectures recover meaningful gradients while retaining forces that are conservative by construction.

WP-O3 benefits most clearly from the reweighted objective. Across the three
force-weighted runs, it reaches relative force errors of \(20.57\%\),
\(19.48\%\), and \(19.28\%\), component MAEs of \(33.7\), \(33.2\), and
\(31.9\,\mathrm{meV}\,\mathring{\mathrm A}^{-1}\), and \(R^2=0.96\) in
all three cases. These values are remarkably close to the direct vector-head
results, for which WP-O3 gives relative force errors of
\(18.46\%\)--\(19.97\%\) across the three reported seeds and component MAEs
of approximately \(31\,\mathrm{meV}\,\mathring{\mathrm A}^{-1}\). Thus,
for WP-O3, most of the accuracy advantage previously associated with direct
force regression can be recovered while enforcing the physically stronger
constraint that forces derive from a scalar energy. Among the force-weighted conservative models, WP-O3 provides the clearest simultaneous satisfaction of both energy and force requirements. Its energy MAE remains below the commonly used \(43\,\mathrm{meV\,atom^{-1}}\) chemical-accuracy threshold in both reported runs. WP-Reg and WP-RegTP also remain below the energy threshold in both runs, but their force errors are substantially larger, at approximately \(30\%\). The full-cell e3nn \(\ell_{\max}=2\) model approaches WP-O3 in force accuracy, but crosses the chemical-accuracy threshold in one of the reported runs. Thus, WP-O3 gives the most consistent balance between chemically accurate energies and accurate conservative forces.

WP-VecInv is particularly unstable under the force-weighted
conservative objective. The results therefore indicate that accurate direct
equivariant force prediction does not automatically imply that the same
architecture can learn an energy surface with equally accurate derivatives.

Among the full-cell baselines, the \(O(3)\)-equivariant models also respond
much more successfully to force-weighted conservative training than the
invariant GNN. In particular, the full-cell e3nn \(\ell_{\max}=2\) model
reaches relative force errors of \(21.21\%\), \(20.81\%\), and \(22.27\%\),
approaching WP-O3 but remaining consistently less accurate in the reported
runs. The \(\ell_{\max}=2\) model also performs noticeably better than the
\(\ell_{\max}=1\) model for conservative forces, suggesting that the
higher-order geometric features become more useful when accurate force
gradients must be encoded through a scalar energy surface. This behavior
differs from the simpler averaged toy benchmark, where increasing
\(\ell_{\max}\) did not provide a systematic advantage. On average, for force prediction, models whose internal message-passing representation remains predominantly invariant and scalar tend to perform worse than architectures that propagate explicit equivariant directional features. This trend is visible both for the WallpaperNet models and for the full-cell baselines. 

The improved conservative forces come at a clear cost in energy accuracy. In
the energy-oriented combined experiments, the strongest models achieve errors
of approximately \(1.7\)--\(2.0\,\mathrm{meV\,atom^{-1}}\), whereas the
force-weighted conservative WP-O3 runs give
\(21.78\)--\(33.24\,\mathrm{meV\,atom^{-1}}\). The reweighted objective should
therefore dose not benefit both, energy and forces, but shifts the optimization
toward the local derivatives of the energy surface. The direct force head
remains the simplest route to high force accuracy, whereas the force-weighted
conservative formulation demonstrates that comparable force accuracy can be
approached while retaining exact energy-force consistency.

Taken together, these experiments reveal a three-way trade-off between energy
accuracy, force accuracy, and conservativity. The unweighted combined
objective favors accurate total energies but fails to resolve the molecular
forces. Direct vector regression gives accurate forces without constraining
them to derive from an energy. The force-weighted conservative objective
sacrifices some energy accuracy to recover accurate gradients. WP-O3 performs
particularly well in this third regime, indicating that its coupled
\(O(3)\)--wallpaper representation is effective not only for predicting
equivariant vectors directly, but also for representing the directional
structure required in the derivatives of a scalar potential-energy surface.

\begin{table*}[t]
\centering
\caption{Test-set performance for random seed 1. Lower is better for the error
metrics, and higher is better for the zero-force score and $R^2$.}
\label{tab:real-system-test-seed-1}
\resizebox{\textwidth}{!}{%
\begin{tabular}{lccccccc}
\toprule
Model &
\shortstack{Energy MAE $\downarrow$\\(meV system$^{-1}$)} &
\shortstack{Energy MAE $\downarrow$\\(meV atom$^{-1}$)} &
\shortstack{Rel. force $L_2$ $\downarrow$\\(\%)} &
\shortstack{Zero-force score $\uparrow$\\(\%)} &
\shortstack{Force comp. MAE $\downarrow$\\(meV\,$\mathring{\mathrm A}^{-1}$)} &
\shortstack{Force comp.\\$R^2$ $\uparrow$} &
\shortstack{Best epoch\\E / F} \\
\midrule
\multicolumn{8}{l}{\textit{Wallpaper-symmetry models}} \\
\addlinespace[2pt]
WP-Inv
& $672.20$ & $1.75$ & $34.58$ & $88.04$ & $61.8$ & $0.88$ & $352$ / $33$ \\
WP-InvEq
& $4077.70$ & $10.62$ & $26.38$ & $93.04$ & $44.6$ & $0.93$ & $46$ / $166$ \\
WP-Equi
& $761.76$ & $2.02$ & $28.66$ & $91.78$ & $49.8$ & $0.92$ & $546$ / $273$ \\
WP-VecInv
& $817.05$ & $2.17$ & $26.23$ & $93.12$ & $44.1$ & $0.93$ & $429$ / $174$ \\
WP-Reg
& $740.18$ & $1.97$ & $25.93$ & $93.28$ & $43.7$ & $0.93$ & $544$ / $246$ \\
WP-RegTP
& $\mathbf{647.53}$ & $\mathbf{1.73}$ & $26.23$ & $93.12$ & $44.4$ & $0.93$ & $525$ / $172$ \\
WP-O3
& $709.95$ & $1.88$ & $\mathbf{19.23}$ & $\mathbf{96.30}$ & $\mathbf{31.4}$ & $\mathbf{0.96}$ & $462$ / $200$ \\
\midrule
\multicolumn{8}{l}{\textit{Full-cell baselines}} \\
\addlinespace[2pt]
Full cell invariant GNN
& $957.12$ & $2.56$ & $52.46$ & $72.48$ & $93.2$ & $0.72$ & $590$ / $47$ \\
Full cell e3nn $\ell_{\max}=1$
& $727.08$ & $1.92$ & $29.25$ & $91.44$ & $52.1$ & $0.91$ & $591$ / $87$ \\
Full cell e3nn $\ell_{\max}=2$
& $771.98$ & $2.02$ & $29.89$ & $91.07$ & $52.7$ & $0.91$ & $584$ / $57$ \\
\bottomrule
\end{tabular}%
}
\end{table*}

\begin{table*}[t]
\centering
\caption{Test-set performance for random seed 24. Lower is better for the error
metrics, and higher is better for the zero-force score and $R^2$.}
\label{tab:real-system-test-seed-24}
\resizebox{\textwidth}{!}{%
\begin{tabular}{lccccccc}
\toprule
Model &
\shortstack{Energy MAE $\downarrow$\\(meV system$^{-1}$)} &
\shortstack{Energy MAE $\downarrow$\\(meV atom$^{-1}$)} &
\shortstack{Rel. force $L_2$ $\downarrow$\\(\%)} &
\shortstack{Zero-force score $\uparrow$\\(\%)} &
\shortstack{Force comp. MAE $\downarrow$\\(meV\,$\mathring{\mathrm A}^{-1}$)} &
\shortstack{Force comp.\\$R^2$ $\uparrow$} &
\shortstack{Best epoch\\E / F} \\
\midrule
\multicolumn{8}{l}{\textit{Wallpaper-symmetry models}} \\
\addlinespace[2pt]
WP-Inv
& $664.42$ & $1.72$ & $49.86$ & $75.14$ & $93.3$ & $0.75$ & $504$ / $9$ \\
WP-InvEq
& $676.98$ & $1.80$ & $25.17$ & $93.66$ & $43.6$ & $0.94$ & $587$ / $188$ \\
WP-Equi
& $753.69$ & $1.98$ & $27.09$ & $92.66$ & $47.9$ & $0.93$ & $587$ / $289$ \\
WP-VecInv
& $864.21$ & $2.28$ & $25.94$ & $93.27$ & $44.6$ & $0.93$ & $352$ / $266$ \\
WP-Reg
& $848.52$ & $2.22$ & $27.59$ & $92.39$ & $47.0$ & $0.92$ & $436$ / $508$ \\
WP-RegTP
& $\mathbf{653.28}$ & $\mathbf{1.71}$ & $25.05$ & $93.72$ & $43.6$ & $0.94$ & $595$ / $157$ \\
WP-O3
& $665.37$ & $1.72$ & $\mathbf{18.46}$ & $\mathbf{96.59}$ & $\mathbf{31.1}$ & $\mathbf{0.97}$ & $595$ / $189$ \\
\midrule
\multicolumn{8}{l}{\textit{Full-cell baselines}} \\
\addlinespace[2pt]
Full cell invariant GNN
& $871.53$ & $2.36$ & $49.20$ & $75.79$ & $87.7$ & $0.76$ & $598$ / $56$ \\
Full cell e3nn $\ell_{\max}=1$
& $703.26$ & $1.85$ & $22.55$ & $94.91$ & $39.0$ & $0.95$ & $591$ / $228$ \\
Full cell e3nn $\ell_{\max}=2$
& $869.10$ & $2.27$ & $18.93$ & $96.41$ & $32.4$ & $0.96$ & $594$ / $208$ \\
\bottomrule
\end{tabular}%
}
\end{table*}

\begin{table*}[t]
\centering
\caption{Test-set performance for random seed 42. Lower is better for the error
metrics, and higher is better for the zero-force score and $R^2$.}
\label{tab:real-system-test-seed-42}
\resizebox{\textwidth}{!}{%
\begin{tabular}{lccccccc}
\toprule
Model &
\shortstack{Energy MAE $\downarrow$\\(meV system$^{-1}$)} &
\shortstack{Energy MAE $\downarrow$\\(meV atom$^{-1}$)} &
\shortstack{Rel. force $L_2$ $\downarrow$\\(\%)} &
\shortstack{Zero-force score $\uparrow$\\(\%)} &
\shortstack{Force comp. MAE $\downarrow$\\(meV\,$\mathring{\mathrm A}^{-1}$)} &
\shortstack{Force comp.\\$R^2$ $\uparrow$} &
\shortstack{Best epoch\\E / F} \\
\midrule
\multicolumn{8}{l}{\textit{Wallpaper-symmetry models}} \\
\addlinespace[2pt]
WP-Inv
& $\mathbf{638.28}$ & $\mathbf{1.70}$ & $39.61$ & $84.31$ & $70.0$ & $0.84$ & $529$ / $25$ \\
WP-InvEq
& $687.08$ & $1.83$ & $25.46$ & $93.52$ & $42.5$ & $0.94$ & $556$ / $213$ \\
WP-Equi
& $639.31$ & $1.71$ & $28.11$ & $92.10$ & $48.9$ & $0.92$ & $584$ / $287$ \\
WP-VecInv
& $852.27$ & $2.27$ & $27.68$ & $92.34$ & $45.6$ & $0.92$ & $400$ / $180$ \\
WP-Reg
& $774.71$ & $2.08$ & $27.93$ & $92.20$ & $46.3$ & $0.92$ & $422$ / $375$ \\
WP-RegTP
& $810.42$ & $2.14$ & $27.34$ & $92.53$ & $45.9$ & $0.93$ & $516$ / $173$ \\
WP-O3
& $670.47$ & $1.81$ & $\mathbf{19.97}$ & $\mathbf{96.01}$ & $\mathbf{31.3}$ & $\mathbf{0.96}$ & $508$ / $205$ \\
\midrule
\multicolumn{8}{l}{\textit{Full-cell baselines}} \\
\addlinespace[2pt]
Full cell invariant GNN
& $1021.78$ & $2.82$ & $40.00$ & $84.00$ & $69.9$ & $0.84$ & $597$ / $86$ \\
Full cell e3nn $\ell_{\max}=1$
& $738.95$ & $1.99$ & $24.12$ & $94.18$ & $40.5$ & $0.94$ & $582$ / $232$ \\
Full cell e3nn $\ell_{\max}=2$
& $678.18$ & $1.83$ & $27.84$ & $92.25$ & $50.1$ & $0.92$ & $597$ / $80$ \\
\bottomrule
\end{tabular}%
}
\end{table*}

\newcommand{\ForceDerivedSeedTable}[3]{%
\begin{table*}[t]
\centering
\caption{Test-set performance for random seed #1 using the force-weighted
energy/force objective. Lower is better for the error metrics, and higher is better
for the zero-force score and $R^2$.}
\label{#2}
\resizebox{\textwidth}{!}{%
\begin{tabular}{lccccccc}
\toprule
Model &
\shortstack{Energy MAE $\downarrow$\\(meV system$^{-1}$)} &
\shortstack{Energy MAE $\downarrow$\\(meV atom$^{-1}$)} &
\shortstack{Rel. force $L_2$ $\downarrow$\\(\%)} &
\shortstack{Zero-force score $\uparrow$\\(\%)} &
\shortstack{Force comp. MAE $\downarrow$\\(meV\,$\mathring{\mathrm A}^{-1}$)} &
\shortstack{Force comp.\\$R^2$ $\uparrow$} &
\shortstack{Best\\epoch} \\
\midrule
#3
\bottomrule
\end{tabular}%
}
\end{table*}%
}

\ForceDerivedSeedTable{1}{tab:real-system-force-derived-seed-1}{%
\multicolumn{8}{l}{\textit{Wallpaper-symmetry models}} \\
\addlinespace[2pt]
WP-Inv & $321{,}562.00$ & $889.75$ & $80.04$ & $35.94$ & $139.6$ & $0.36$ & $380$ \\
WP-InvEq & $290{,}840.00$ & $785.90$ & $79.53$ & $36.75$ & $139.1$ & $0.37$ & $379$ \\
WP-Equi & $290{,}464.00$ & $773.53$ & $79.51$ & $36.79$ & $139.1$ & $0.37$ & $398$ \\
WP-VecInv & $3{,}694{,}400.00$ & $9573.70$ & $52.84$ & $72.08$ & $92.7$ & $0.72$ & $82$ \\
WP-Reg & $\mathbf{5250.41}$ & $\mathbf{13.79}$ & $32.89$ & $89.18$ & $57.1$ & $0.89$ & $399$ \\
WP-RegTP & $8162.55$ & $21.39$ & $29.96$ & $91.02$ & $51.0$ & $0.91$ & $394$ \\
WP-O3 & $11{,}930.60$ & $30.96$ & $\mathbf{20.57}$ & $\mathbf{95.77}$ & $\mathbf{33.7}$ & $\mathbf{0.96}$ & $347$ \\
\midrule
\multicolumn{8}{l}{\textit{Full-cell baselines}} \\
\addlinespace[2pt]
Full cell invariant GNN & $49{,}695.00$ & $128.90$ & $68.67$ & $52.85$ & $122.8$ & $0.53$ & $172$ \\
Full cell e3nn $\ell_{\max}=1$ & $25{,}607.20$ & $65.83$ & $25.60$ & $93.45$ & $43.7$ & $0.93$ & $382$ \\
Full cell e3nn $\ell_{\max}=2$ & $28{,}302.40$ & $73.69$ & $21.21$ & $95.50$ & $35.3$ & $0.96$ & $373$ \\
}

\ForceDerivedSeedTable{24}{tab:real-system-force-derived-seed-24}{%
\multicolumn{8}{l}{\textit{Wallpaper-symmetry models}} \\
\addlinespace[2pt]
WP-Inv & $318{,}885.00$ & $855.62$ & $79.98$ & $36.03$ & $139.5$ & $0.36$ & $282$ \\
WP-InvEq & $295{,}801.00$ & $779.89$ & $77.37$ & $40.14$ & $135.0$ & $0.40$ & $354$ \\
WP-Equi & $308{,}116.00$ & $788.64$ & $81.09$ & $34.25$ & $141.8$ & $0.34$ & $164$ \\
WP-VecInv & $3{,}677{,}060.00$ & $9370.42$ & $53.73$ & $71.13$ & $96.9$ & $0.71$ & $142$ \\
WP-Reg & $12{,}556.60$ & $33.10$ & $29.65$ & $91.21$ & $53.0$ & $0.91$ & $89$ \\
WP-RegTP & $\mathbf{5998.09}$ & $\mathbf{15.28}$ & $29.51$ & $91.29$ & $51.7$ & $0.91$ & $395$ \\
WP-O3 & $12{,}959.50$ & $33.24$ & $\mathbf{19.48}$ & $\mathbf{96.21}$ & $\mathbf{33.2}$ & $\mathbf{0.96}$ & $258$ \\
\midrule
\multicolumn{8}{l}{\textit{Full-cell baselines}} \\
\addlinespace[2pt]
Full cell invariant GNN & $60{,}007.70$ & $148.33$ & $66.72$ & $55.49$ & $119.1$ & $0.55$ & $126$ \\
Full cell e3nn $\ell_{\max}=1$ & $15{,}967.20$ & $40.84$ & $26.89$ & $92.77$ & $47.2$ & $0.93$ & $296$ \\
Full cell e3nn $\ell_{\max}=2$ & $17{,}415.70$ & $46.21$ & $20.81$ & $95.67$ & $35.6$ & $0.96$ & $390$ \\
}

\ForceDerivedSeedTable{67}{tab:real-system-force-derived-seed-67}{%
\multicolumn{8}{l}{\textit{Wallpaper-symmetry models}} \\
\addlinespace[2pt]
WP-Inv & $319{,}690.00$ & $847.65$ & $82.34$ & $32.20$ & $140.7$ & $0.32$ & $381$ \\
WP-InvEq & $357{,}664.00$ & $892.79$ & $79.14$ & $37.36$ & $137.3$ & $0.37$ & $301$ \\
WP-Equi & $297{,}028.00$ & $773.34$ & $81.17$ & $34.12$ & $141.2$ & $0.34$ & $381$ \\
WP-VecInv & $3{,}749{,}750.00$ & $9635.12$ & $80.90$ & $34.56$ & $138.7$ & $0.35$ & $37$ \\
WP-Reg & $13{,}407.80$ & $35.74$ & $29.56$ & $91.26$ & $51.9$ & $0.91$ & $108$ \\
WP-RegTP & $\mathbf{6391.74}$ & $\mathbf{16.49}$ & $32.01$ & $89.75$ & $55.3$ & $0.90$ & $392$ \\
WP-O3 & $8415.36$ & $21.78$ & $\mathbf{19.28}$ & $\mathbf{96.28}$ & $\mathbf{31.9}$ & $\mathbf{0.96}$ & $375$ \\
\midrule
\multicolumn{8}{l}{\textit{Full-cell baselines}} \\
\addlinespace[2pt]
Full cell invariant GNN & $12{,}966.90$ & $33.49$ & $67.14$ & $54.93$ & $118.0$ & $0.55$ & $196$ \\
Full cell e3nn $\ell_{\max}=1$ & $24{,}553.90$ & $63.70$ & $25.18$ & $93.66$ & $43.4$ & $0.94$ & $384$ \\
Full cell e3nn $\ell_{\max}=2$ & $14{,}300.30$ & $36.89$ & $22.27$ & $95.04$ & $37.6$ & $0.95$ & $367$ \\
}

\subsection{Memory footprint of molecule--graphene systems}
\label{sec:memory_footprint}

\begin{figure*}[t]
    \centering
    \includegraphics[width=\textwidth]{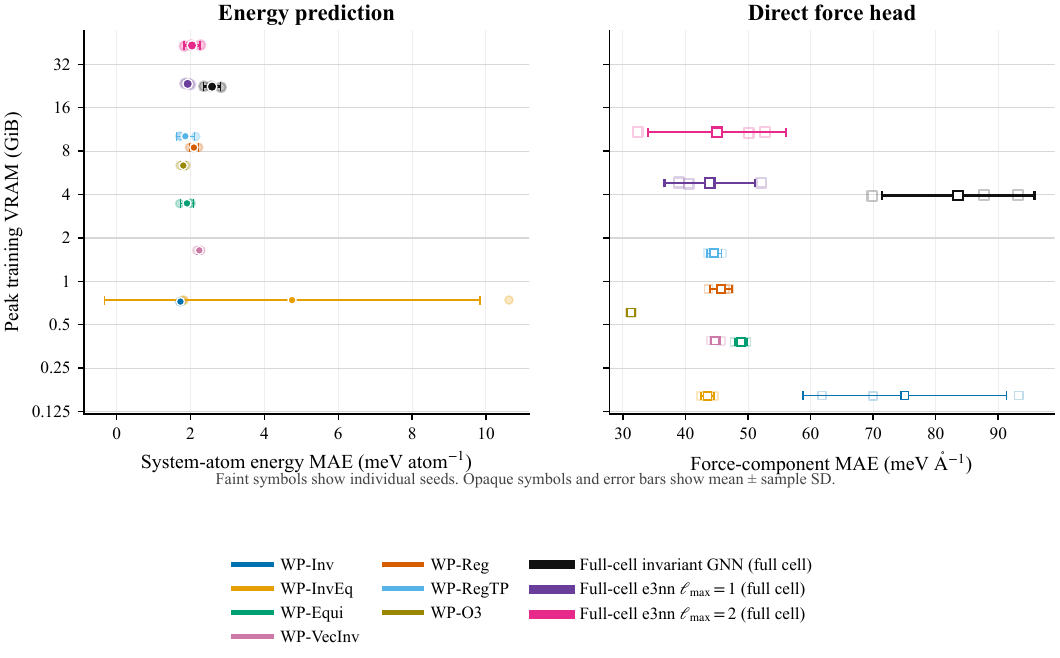}
    \caption{Peak training VRAM as a function of test-set accuracy for energy
    prediction (left) and direct force-head prediction (right). Energy accuracy
    is measured by the system-atom MAE, and force accuracy by the Cartesian
    force-component MAE. Faint markers show the three individual runs, opaque
    markers and error bars show the mean and sample standard deviation. Colors
    identify model architectures, and heavier full-cell markers and legend
    strokes distinguish the full-cell baselines. The VRAM axis is logarithmic.}
    \label{fig:real-system-vram-accuracy}
\end{figure*}

The roughly \(45\times\) smaller edge count of the WallpaperNet graphs
(Section~\ref{sec:real-system}) primarily indicates the reduction in system
size and memory footprint. We report the peak training VRAM consumption across three independently
initialized runs. Figure~\ref{fig:real-system-vram-accuracy} compares this
memory footprint with test-set accuracy for energy prediction and direct
force-head prediction. Peak VRAM is measured during training, while accuracy
is evaluated on the held-out test set. Table~\ref{tab:real-system-peak-vram}
reports the peak-memory values used to construct the figure.

\begin{table*}[t]
\centering
\caption{Peak training VRAM at batch size 16 per GPU for the three random
seeds. Values are in GiB per GPU. The final column in each training block gives
the mean and sample standard deviation across seeds.}
\label{tab:real-system-peak-vram}
\setlength{\tabcolsep}{4pt}
\scriptsize
\begin{tabular}{lcccccccc}
\toprule
& \multicolumn{4}{c}{Combined training} &
  \multicolumn{4}{c}{Direct force-head training} \\
\cmidrule(lr){2-5}\cmidrule(lr){6-9}
Model & Seed 1 & Seed 24 & Seed 42 & Mean $\pm$ SD &
Seed 1 & Seed 24 & Seed 42 & Mean $\pm$ SD \\
\midrule
WP-Inv     & 0.725 & 0.726 & 0.721 & $0.724 \pm 0.002$ & 0.161 & 0.162 & 0.161 & $0.1612 \pm 0.0005$ \\
WP-InvEq   & 0.741 & 0.743 & 0.737 & $0.741 \pm 0.003$ & 0.160 & 0.160 & 0.160 & $0.1601 \pm 0.0004$ \\
WP-Equi    & 3.479 & 3.483 & 3.461 & $3.474 \pm 0.012$ & 0.379 & 0.380 & 0.378 & $0.379 \pm 0.001$ \\
WP-VecInv  & 1.643 & 1.646 & 1.634 & $1.641 \pm 0.006$ & 0.388 & 0.389 & 0.386 & $0.388 \pm 0.001$ \\
WP-Reg     & 8.487 & 8.498 & 8.442 & $8.476 \pm 0.030$ & 0.887 & 0.888 & 0.882 & $0.886 \pm 0.003$ \\
WP-RegTP   & 10.161 & 10.175 & 10.107 & $10.148 \pm 0.036$ & 1.568 & 1.570 & 1.560 & $1.566 \pm 0.005$ \\
WP-O3      & 6.364 & 6.373 & 6.329 & $6.355 \pm 0.023$ & 0.608 & 0.608 & 0.604 & $0.607 \pm 0.002$ \\
\midrule
Full cell invariant GNN & 22.531 & 22.543 & 22.184 & $22.419 \pm 0.204$ & 3.961 & 3.963 & 3.901 & $3.942 \pm 0.036$ \\
Full cell e3nn $\ell_{\max}=1$ & 23.547 & 23.560 & 23.184 & $23.430 \pm 0.214$ & 4.838 & 4.850 & 4.764 & $4.817 \pm 0.047$ \\
Full cell e3nn $\ell_{\max}=2$ & 43.487 & 43.510 & 42.816 & $43.271 \pm 0.394$ & 10.847 & 10.852 & 10.680 & $10.793 \pm 0.098$ \\
\bottomrule
\end{tabular}
\end{table*}

For energy training, the molecule-only WallpaperNet models require
\(0.72\)--\(10.15\,\mathrm{GiB}\), whereas the full-cell invariant, e3nn
\(\ell_{\max}=1\), and e3nn \(\ell_{\max}=2\) baselines require \(22.42\),
\(23.43\), and \(43.27\,\mathrm{GiB}\), respectively. For direct force-head
training, the molecule-only models require \(0.16\)--\(1.57\,\mathrm{GiB}\),
compared with \(3.94\), \(4.82\), and \(10.79\,\mathrm{GiB}\) for the
corresponding full-cell baselines.

WP-O3 provides the best force accuracy, reaching a mean force-component MAE of
\(31.3\,\mathrm{meV\,\mathring{A}^{-1}}\) with a mean peak memory consumption
of \(0.61\,\mathrm{GiB}\). For energy prediction, WP-Inv achieves the lowest
mean system-atom MAE of \(1.73\,\mathrm{meV\,atom^{-1}}\) while requiring only
\(0.72\,\mathrm{GiB}\). These results show that the molecule-only models can
match or improve upon the accuracy of the full-cell baselines while using
substantially less training memory.

\subsection{One-batch maximum utilization speed tests}

\label{sec:real-system-max-batch}

We benchmarked all ten architectures in both combined and direct force-head
training modes on a single NVIDIA GH200 GPU, giving 20 runs in total. For each
architecture and mode, the maximum batch size was estimated from the median
optimizer-training peak measured at a batch size of 16 per GPU. We targeted
90\% of the available \(93\,\mathrm{GiB}\), or \(83.7\,\mathrm{GiB}\), and used
\begin{equation}
    B_{\max}
    = \left\lfloor
        \frac{16 \times 83.7\,\mathrm{GiB}}
             {M_{16}}
      \right\rfloor,
\end{equation}
where \(M_{16}\) is the measured median peak memory at batch size 16. Each
timed fit comprised 30 epochs and included validation, checkpointing, and
offline Weights \& Biases logging. Inference was performed from the best
checkpoint after fitting, and the complete time covers the entire Python run.

\begin{table*}[t]
\centering
\caption{One-GPU maximum-batch timing results for all ten architectures under
combined and direct force-head training. Inference times are measured
from the best checkpoint.}
\label{tab:real-system-max-batch-timing}
\setlength{\tabcolsep}{2.5pt}
\scriptsize
\resizebox{\textwidth}{!}{%
\begin{tabular}{lccccccccc}
\toprule
Model & Batch & \shortstack{Fit\\(min)} &
\shortstack{Infer train\\(s)} & \shortstack{Infer val\\(s)} &
\shortstack{Infer test\\(s)} & \shortstack{Infer total\\(s)} &
\shortstack{Complete\\(min)} & \shortstack{Test\\structures s\(^{-1}\)} &
\shortstack{Test peak\\(GiB GPU\(^{-1}\))} \\
\midrule
\multicolumn{10}{l}{\textit{Combined training}} \\
WP-Inv & 3280 & 1.4 & 2.0 & 0.8 & 0.9 & 3.7 & 1.9 & 10128.1 & 13.82 \\
WP-InvEq & 3196 & 1.4 & 2.0 & 0.7 & 0.9 & 3.5 & 1.7 & 10177.2 & 14.17 \\
WP-Equi & 887 & 8.2 & 6.9 & 1.3 & 1.4 & 9.7 & 8.6 & 1922.9 & 41.26 \\
WP-VecInv & 1547 & 3.9 & 3.7 & 1.0 & 1.1 & 5.7 & 4.3 & 4346.8 & 33.05 \\
WP-Reg & 338 & 63.3 & 48.8 & 6.4 & 6.6 & 61.8 & 64.6 & 257.3 & 40.69 \\
WP-RegTP & 275 & 75.9 & 56.5 & 7.4 & 7.5 & 71.4 & 77.3 & 220.8 & 38.97 \\
WP-O3 & 422 & 49.0 & 37.5 & 5.1 & 5.2 & 47.8 & 50.1 & 335.1 & 38.08 \\
Full cell invariant GNN & 112 & 45.0 & 34.2 & 5.3 & 5.3 & 44.8 & 46.0 & 411.1 & 41.85 \\
Full cell e3nn \(\ell_{\max}=1\) & 117 & 71.6 & 56.3 & 7.8 & 7.9 & 72.0 & 73.2 & 244.6 & 45.47 \\
Full cell e3nn \(\ell_{\max}=2\) & 59 & 189.9 & 131.3 & 19.7 & 17.3 & 168.4 & 193.0 & 96.6 & 44.32 \\
\midrule
\multicolumn{10}{l}{\textit{Direct force-head training}} \\
WP-Inv & 5326 & 0.9 & 2.0 & 0.7 & 0.9 & 3.5 & 1.2 & 22397.3 & 2.67 \\
WP-InvEq & 5256 & 0.9 & 2.0 & 0.6 & 0.7 & 3.3 & 1.2 & 21652.0 & 2.57 \\
WP-Equi & 1241 & 6.0 & 5.1 & 1.1 & 1.2 & 7.3 & 6.3 & 3232.7 & 5.11 \\
WP-VecInv & 3060 & 1.4 & 2.2 & 0.7 & 0.9 & 3.8 & 1.6 & 9223.1 & 7.41 \\
WP-Reg & 568 & 25.3 & 13.2 & 2.0 & 2.1 & 17.4 & 25.8 & 992.7 & 6.80 \\
WP-RegTP & 469 & 30.9 & 15.8 & 2.3 & 2.4 & 20.4 & 31.4 & 815.6 & 10.23 \\
WP-O3 & 771 & 18.6 & 10.7 & 1.7 & 1.8 & 14.2 & 19.0 & 1322.2 & 6.16 \\
Full cell invariant GNN & 252 & 20.2 & 23.1 & 3.3 & 3.6 & 30.0 & 21.0 & 841.0 & 15.30 \\
Full cell e3nn \(\ell_{\max}=1\) & 311 & 23.5 & 27.1 & 4.0 & 4.0 & 35.1 & 24.4 & 640.9 & 22.68 \\
Full cell e3nn \(\ell_{\max}=2\) & 155 & 57.5 & 49.3 & 6.8 & 7.0 & 63.1 & 58.8 & 284.7 & 26.77 \\
\bottomrule
\end{tabular}%
}
\end{table*}

For combined training, the molecule-only models complete in
\(1.7\)--\(77.3\,\mathrm{min}\), with test inference requiring
\(0.9\)--\(7.5\,\mathrm{s}\). The corresponding full-cell ranges are
\(46.0\)--\(193.0\,\mathrm{min}\) and \(5.3\)--\(17.3\,\mathrm{s}\),
respectively. Under direct force-head training, the molecule-only models
complete in \(1.2\)--\(31.4\,\mathrm{min}\) and require
\(0.7\)--\(2.4\,\mathrm{s}\) for test inference, compared with
\(21.0\)--\(58.8\,\mathrm{min}\) and \(3.6\)--\(7.0\,\mathrm{s}\) for the
full-cell baselines.

\begin{figure*}[t]
    \centering
    \includegraphics[width=\textwidth]{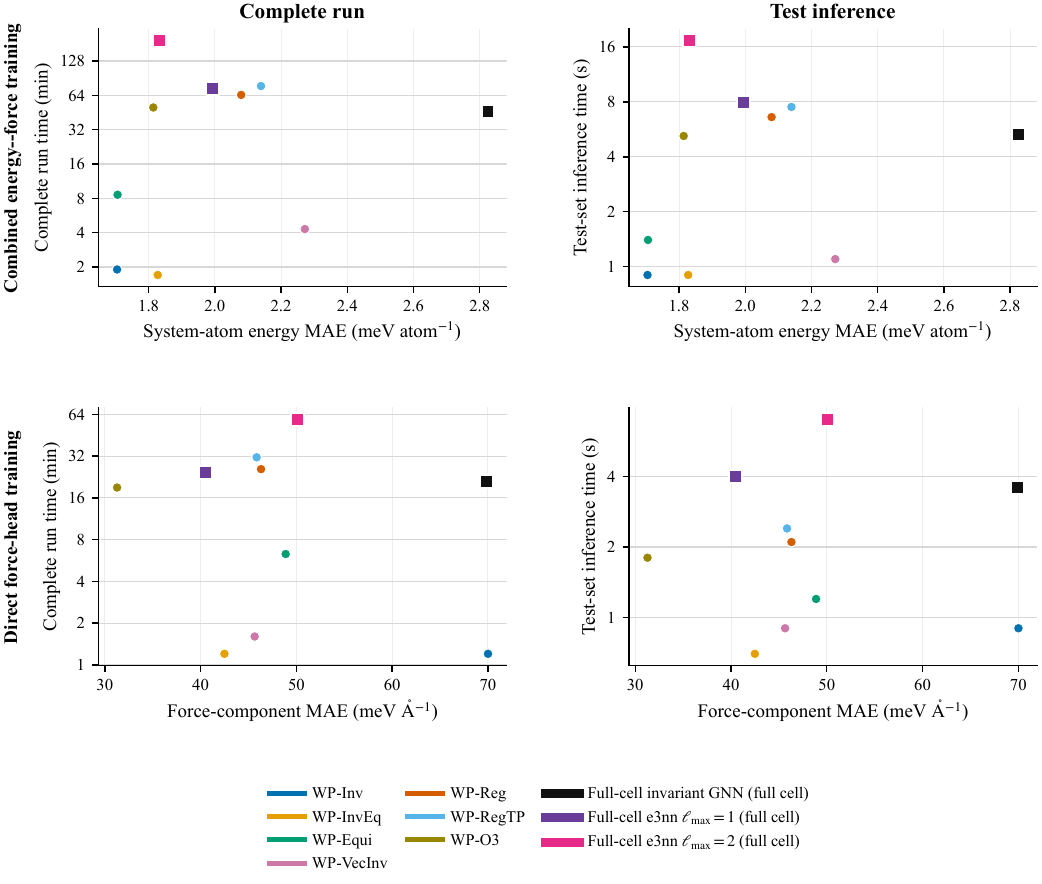}
    \caption{Accuracy--runtime trade-offs for the one-GPU maximum-batch
    experiments. The upper row uses the system-atom energy MAE from combined
    training, and the lower row uses the force-component MAE from direct
    force-head training. The left column reports complete Python-run time and
    the right column reports test-set inference time. Square markers denote
    full-cell baselines and circles denote molecule-only WallpaperNet models.}
    \label{fig:real-system-timing-accuracy}
\end{figure*}

Figure~\ref{fig:real-system-timing-accuracy} relates the timing measurements to
the corresponding seed-42 test errors. The complete runtime includes training,
validation, checkpointing, and inference, whereas the test-inference panels
isolate prediction from the best checkpoint. Logarithmic time axes accommodate
the large runtime range across architectures.

The timing results show that reducing the explicit support graph can lead to substantial computational savings, but that the magnitude of this advantage depends strongly on the operations used within the symmetry-reduced model. The simplest invariant WallpaperNet architectures occupy an especially low-cost regime. Under combined training, WP-Inv and WP-InvEq complete the full 30-epoch run in approximately \(2\,\mathrm{min}\) and process more than \(10^4\) test structures per second, whereas the full-cell baselines require \(46\)--\(193\,\mathrm{min}\) and reach approximately \(10^2\)--\(4\times10^2\) structures per second. This large difference reflects both the substantially smaller molecule-only graph and the comparatively inexpensive scalar message-passing operations.

The reduction in graph size does not, however, make every WallpaperNet architecture faster than every full-cell baseline. WP-Reg and WP-RegTP contain multiple regular-representation pathways and explicit tensor-product operations and are correspondingly much more expensive than the invariant WallpaperNet models. Under combined training, their complete runtimes of \(64.6\) and \(77.3\,\mathrm{min}\), respectively, exceed that of the full-cell invariant GNN (\(46.0\,\mathrm{min}\)). This demonstrates that computational cost is controlled by both graph size and the operations performed on each edge: symmetry reduction decreases the number of interactions that must be processed, but sufficiently expensive equivariant operations can partially or completely offset this advantage.

WP-O3 provides a particularly favorable compromise between these two effects. Despite combining an \(O(3)\)-equivariant tensor pathway with wallpaper-group-equivariant processing, its combined-training run completes in \(50.1\,\mathrm{min}\), compared with \(73.2\,\mathrm{min}\) for the full-cell e3nn \(\ell_{\max}=1\) model and \(193.0\,\mathrm{min}\) for \(\ell_{\max}=2\). This corresponds to approximately \(1.5\times\) and \(3.9\times\) shorter complete runtimes, respectively. The same trend is visible during inference: WP-O3 evaluates approximately \(335\) test structures per second, compared with \(245\) and \(97\) structures per second for the two full-cell e3nn models. The full-cell invariant GNN remains slightly faster than WP-O3 in this mode, illustrating that the benefit should be interpreted relative to models of comparable geometric expressivity rather than as an unconditional speed advantage over every full-cell architecture.

The computational advantage becomes stronger for direct force-head training. Direct force prediction avoids differentiating forces through the coordinate gradient of the scalar energy, substantially reducing both memory requirements and computational cost. Consistent with this, every architecture supports a larger batch and a shorter runtime in the direct-head experiment than in combined training. For WP-O3, the estimated maximum batch size increases from \(422\) to \(771\), while the complete runtime decreases from \(50.1\) to \(19.0\,\mathrm{min}\). The corresponding reductions for the full-cell e3nn models are also large, reflecting the additional cost of gradient-derived force supervision during combined training.

Most importantly, the direct-force timing and accuracy results show that the computational reduction does not come at the expense of predictive accuracy. WP-O3 simultaneously gives the lowest direct-force error in the real-system experiment and is faster than all three full-cell baselines. Its complete runtime is \(19.0\,\mathrm{min}\), compared with \(21.0\), \(24.4\), and \(58.8\,\mathrm{min}\) for the full-cell invariant, e3nn \(\ell_{\max}=1\), and e3nn \(\ell_{\max}=2\) models, respectively. At inference, WP-O3 reaches approximately \(1322\) test structures per second, corresponding to approximately \(1.6\times\), \(2.1\times\), and \(4.6\times\) the throughput of these three baselines. At the same time, its force-component MAE is lower than that of every full-cell model. WP-O3 therefore occupies a favorable region of the accuracy--runtime trade-off rather than achieving its improved force accuracy through increased computational cost.

The comparison also illustrates why parameter count alone is an incomplete measure of computational complexity. All architectures were constructed with approximately matched trainable parameter counts, yet their runtimes, feasible batch sizes, and inference throughputs differ by more than an order of magnitude. Regular-representation tensor products, \(O(3)\) tensor products, the number of explicit graph edges, and differentiation through conservative force predictions all contribute substantially to the actual computational cost. The timing experiment therefore complements the parameter-matched accuracy comparison by showing the practical computational consequences of the different symmetry representations.

Overall, these results support the central motivation of the symmetry-reduced construction: when the substrate symmetry is known, its effect can be represented without explicitly propagating messages through every substrate atom. The largest gains are obtained by the simpler invariant architectures, while more expressive equivariant models trade some of this efficiency for geometric capacity. WP-O3 provides the strongest balance in the present real-system experiment, retaining much of the computational advantage of the molecule-only graph while exceeding the accuracy of the substantially larger full-cell \(O(3)\)-equivariant models. It's important to note that the proposed WP-Nets are not the best performing designs, they are selected as illustrations of application of different parts of the proposed WallpaperNet framework. Additional, keeping the same amount of learnable parameters for all models is a choice that was made to simplify the comparison, but it is not a requirement of the proposed framework, therefore additional gains can be made.

\end{document}